\documentclass{article}

\usepackage[preprint]{neurips_2026}

\usepackage[utf8]{inputenc} % allow utf-8 input
\usepackage[T1]{fontenc}    % use 8-bit T1 fonts
\usepackage{hyperref}       % hyperlinks
\usepackage{url}            % simple URL typesetting
\usepackage{booktabs}       % professional-quality tables
\usepackage{amsfonts}       % blackboard math symbols
\usepackage{nicefrac}       % compact symbols for 1/2, etc.
\usepackage{microtype}      % microtypography
\usepackage{xcolor}         % colors
\usepackage{graphicx}
\usepackage{booktabs}
\usepackage{placeins}
\usepackage{threeparttable}
\usepackage{caption}
\usepackage{subcaption}% The "axessiblity" package can be found at: https://ctan.org/pkg/axessibility?lang=en
\usepackage[accsupp]{axessibility}  % Improves PDF readability for those with disabilities.

\usepackage{orcidlink}
\usepackage{array}
\usepackage{graphicx}
\usepackage{amsmath,amssymb} % define this before the line numbering.
\usepackage{color}

\usepackage{hyperref}
\usepackage{url}
\usepackage{tabularx}
\usepackage{multirow}
\usepackage{booktabs}     % For \toprule, \midrule, \bottomrule
\usepackage{multirow}     % For \multirow in headers
\usepackage{amsmath}      % For general math symbols
\usepackage{xcolor}       % If you plan to use \textcolor or row highlighting
\usepackage{makecell}     % For better control  
\usepackage[table]{xcolor}
\usepackage{colortbl}
\usepackage{wrapfig}
\usepackage{capt-of}

\usepackage{float}
 \usepackage{pifont}
\usepackage{xspace}
\newcommand{\ie}{\textit{i.e.}}
\newcommand{\eg}{\textit{e.g.}}
\newcommand{\etc}{\textit{etc.}\@\xspace}
\usepackage[most]{tcolorbox}

\newtcolorbox{AIbox}[2][]{
  enhanced,
  colback=white,
  colframe=black!50,
  boxrule=0.8pt,
  arc=2mm,
  title={#2},
  #1
}

\newcommand{\ours}{\textsc{FORGE}}%
\newcommand{\taskone}{\textsc{WorkVeri}}%
\newcommand{\tasktwo}{\textsc{SurfInsp}}%
\newcommand{\taskthree}{\textsc{AssyVeri}}%
\newcommand{\messcen}{\textsc{MES scenario}}%
\newcommand{\pesscen}{\textsc{PES scenario}}%
\newcommand{\swnscen}{\textsc{SWN scenario}}%
\newcommand{\cncscen}{\textsc{CNC scenario}}%
\newcommand{\chsscen}{\textsc{CHS scenario}}%
\newcommand{\pcsscen}{\textsc{PCs scenario}}%
\newcommand{\nutsscen}{\textsc{Nuts scenario}}%

\newcommand{\rot}[1]{\rotatebox{60}{\scriptsize\textbf{#1}}}

\newcommand{\homepage}{\raisebox{-1.5pt}{\includegraphics[height=1em]{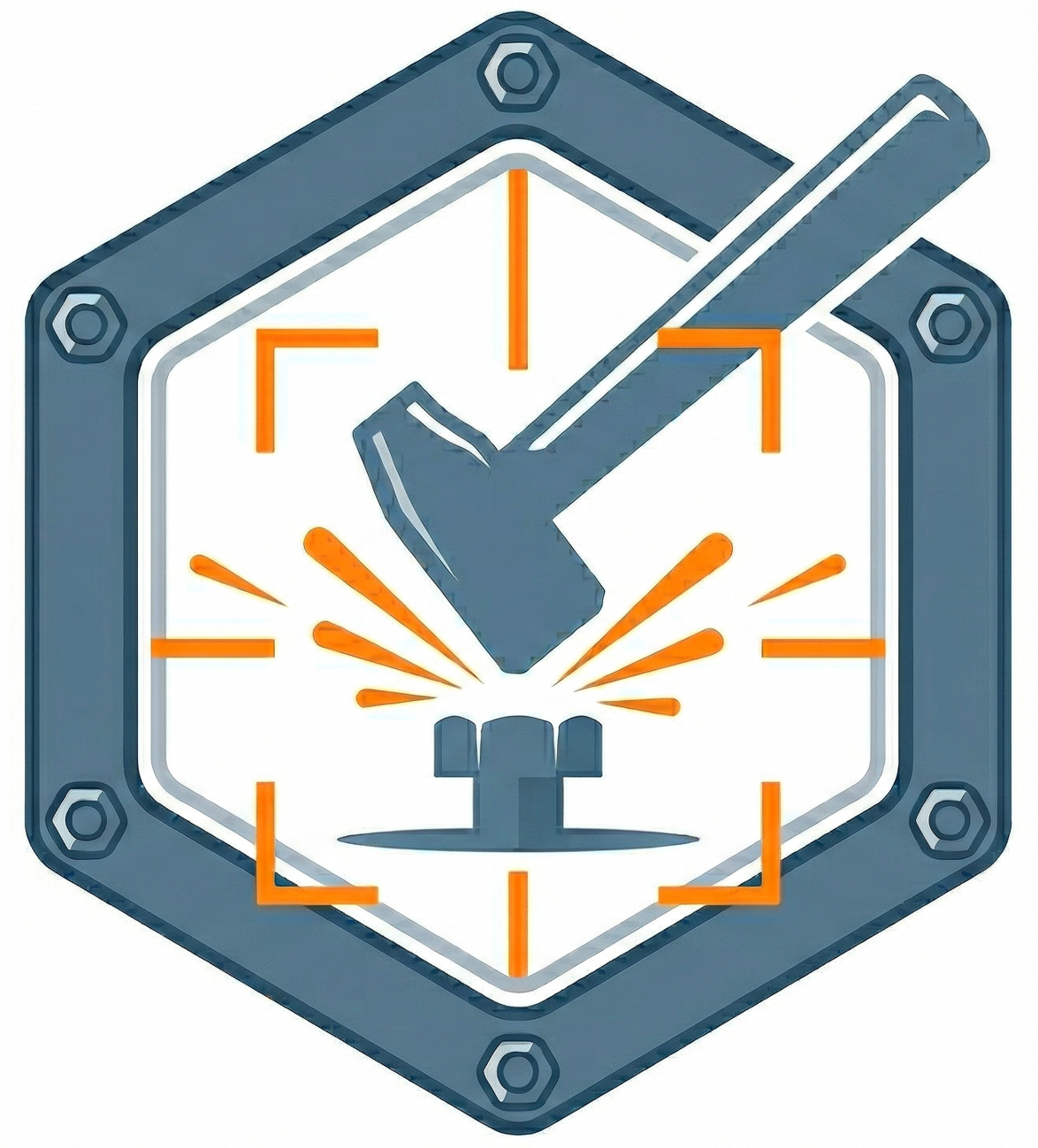}}}
\newcommand{\github}{\raisebox{-1.5pt}{\includegraphics[height=1em]{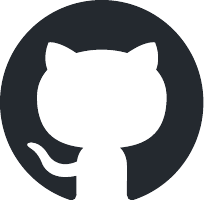}}}
\newcommand{\huggingface}{\raisebox{-1.5pt}{\includegraphics[height=1em]{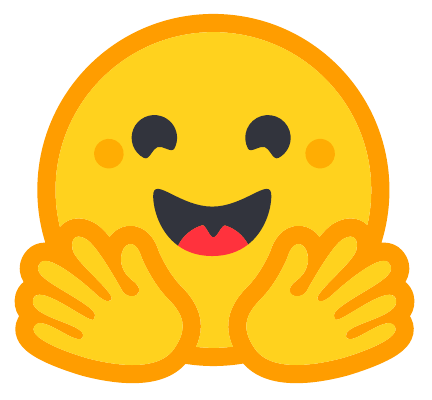}}}
\definecolor{titleblue}{RGB}{0,92,185} % 

\newcommand{\hl}[1]{\textcolor{titleblue}{#1}} 

\usepackage[table]{xcolor}
\usepackage{etoc}
\etocdepthtag.toc{mtchapter}
\etocsettagdepth{mtchapter}{subsection}
\etocsettagdepth{mtappendix}{none}
\definecolor{citecolor}{HTML}{0071bc}
\hypersetup{colorlinks=true, linkcolor=red, citecolor=titleblue}
\title{\raisebox{-.75ex}{\includegraphics[width=0.78cm]{figs/forge_icon.png}}\,\hl{\ours}: \hl{F}ine-grained multim\hl{o}dal evaluation for manufactu\hl{r}in\hl{g} sc\hl{e}narios}

\author{%
  Xiangru Jian\textsuperscript{1, $\ast$},
  Hao Xu\textsuperscript{2,3, $\ast$, $\dagger$},
  Wei Pang\textsuperscript{4, $\ast$},
  Xinjian Zhao\textsuperscript{4},
  Chengyu Tao\textsuperscript{5},
  Qixin Zhang\textsuperscript{6} \\
  \textbf{Xikun Zhang}\textsuperscript{7, $\dagger$},
  \textbf{Chao Zhang}\textsuperscript{1},
  \textbf{Guanzhi Deng}\textsuperscript{8},
  \textbf{Alex Xue}\textsuperscript{1},
  \textbf{Juan Du}\textsuperscript{9}, \\
  \textbf{Tianshu Yu}\textsuperscript{4},
  \textbf{Garth Tarr}\textsuperscript{2},
  \textbf{Linqi Song}\textsuperscript{8, 10}
  \textbf{Qiuzhuang Sun}\textsuperscript{3},
  \textbf{Dacheng Tao}\textsuperscript{6} \\
  \normalfont
  \textsuperscript{1}University of Waterloo, Canada,
  \textsuperscript{2}University of Sydney, Australia,\\
  \textsuperscript{3}Singapore Management University, Singapore 
  \textsuperscript{4}The Chinese University of Hong Kong, Shenzhen, China, \\
  \textsuperscript{5}Hunan University, China, 
  \textsuperscript{6}Nanyang Technological University, Singapore \\
  \textsuperscript{7}Royal Melbourne Institute of Technology, Australia \quad
  \textsuperscript{8}City University of Hong Kong, China \\
  \textsuperscript{9}The Hong Kong University of Science and Technology (Guangzhou), China \\
  \textsuperscript{10}City University of Hong Kong Shenzhen Research Institute, China \vspace{0.3em} \\
  \textsuperscript{$\ast$}Equal contribution \quad \textsuperscript{$\dagger$}Corresponding author
  \\ [1.5mm]
\\
{\homepage\  Project Page: \url{https://ai4manufacturing.github.io/forge-web/}} \\
{\github\ Code: \texttt{\url{https://github.com/AI4Manufacturing/FORGE}}} \\
  {\huggingface\ Dataset: \texttt{\url{https://huggingface.co/datasets/AI4Manufacturing/forge}}}
}

\begin{document}

\maketitle

\begin{abstract}
  
The manufacturing sector is increasingly adopting Multimodal Large Language Models (MLLMs) to transition from simple perception to autonomous execution, yet current evaluations fail to reflect the rigorous demands of real-world manufacturing environments. Progress is hindered by data scarcity and a lack of fine-grained domain semantics in existing datasets. To bridge this gap, we introduce \ours. We first construct a high-quality multimodal dataset that combines real-world 2D images and 3D point clouds, annotated with fine-grained domain semantics (e.g., exact model numbers). We then evaluate 18 state-of-the-art MLLMs across three manufacturing tasks, namely workpiece verification, structural surface inspection, and assembly verification, revealing significant performance gaps. Counter to conventional understanding, the bottleneck analysis shows that visual grounding is not the primary limiting factor. Instead, insufficient domain-specific knowledge is the key bottleneck, setting a clear direction for future research. Beyond evaluation, we show that our structured annotations can serve as an actionable training resource: supervised fine-tuning of a compact 3B-parameter model on our data yields up to 90.8\% relative improvement in accuracy on held-out manufacturing scenarios, providing preliminary evidence for a practical pathway toward domain-adapted manufacturing MLLMs.
 % \keywords{Multimodal Large Language Model, Quality Inspection, and Smart Manufacturing}
 
\end{abstract}

\section{Introduction}
\label{sec:intro}
The manufacturing sector, a critical pillar of the global economy, generates massive amounts of heterogeneous data from production lines and relies on complex decision-making systems~\citep{gautam2025iiot}. The complexity and volume of such data strongly require advanced technologies to address the challenges of integrating and interpreting fragmented multimodal data. Crucially, as modern manufacturing paradigms increasingly rely on data-driven decision-making and sophisticated human-machine collaboration, there is an urgent need for intelligent systems capable of higher-level cognitive tasks~\citep{gautam2025iiot,fan2024enhancing,wang2024llm,yuan2025chat}.

Vision models have been widely applied, primarily functioning as perception modules that focus on information extraction, such as object localization and anomaly detection~\citep{defard2021padim,roth2022towards,Tao2023Anomaly,wen2024foundationpose}. These models typically operate within a modular, pipelined architecture, generating task-specific outputs (\eg, defect type and location) that are subsequently passed to higher-level Manufacturing Execution Systems (MES)~\citep{saenz2009manufacturing,yuan2025chat} for decision-making. Yet, these vision models are fundamentally limited by their inability to reason and execute autonomous control~\citep{zhang2024logicode,Fan2025MaViLa,lin2025vlm}.

\begin{figure*}[t]  
    \begin{subfigure}{\textwidth}
        \centering
        \includegraphics[width=1\linewidth]{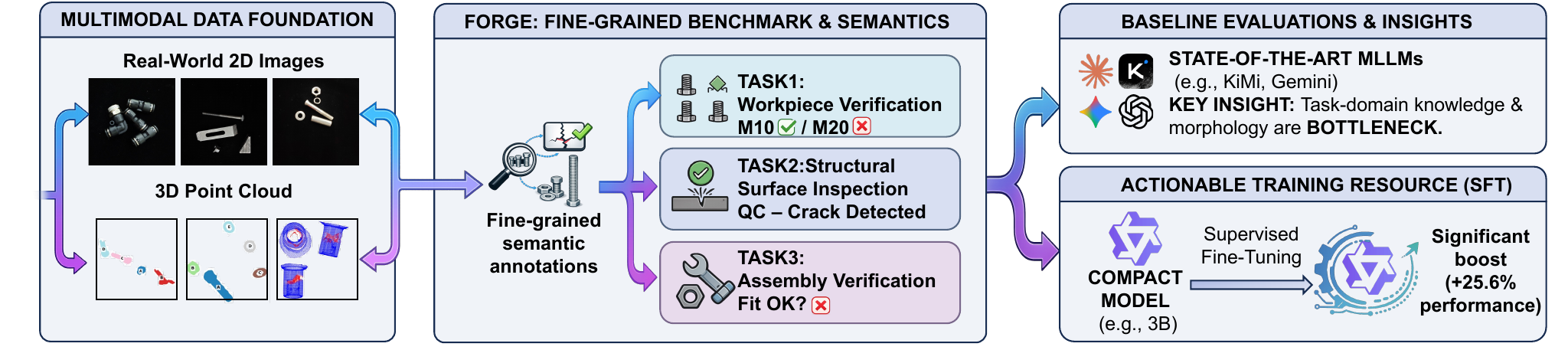} 
        \caption{%The main pipeline of \ours.
        }
        \label{fig:pipeline}
    \end{subfigure}
    
    \begin{subfigure}{0.40\textwidth}  
        \centering
        \includegraphics[width=\linewidth]{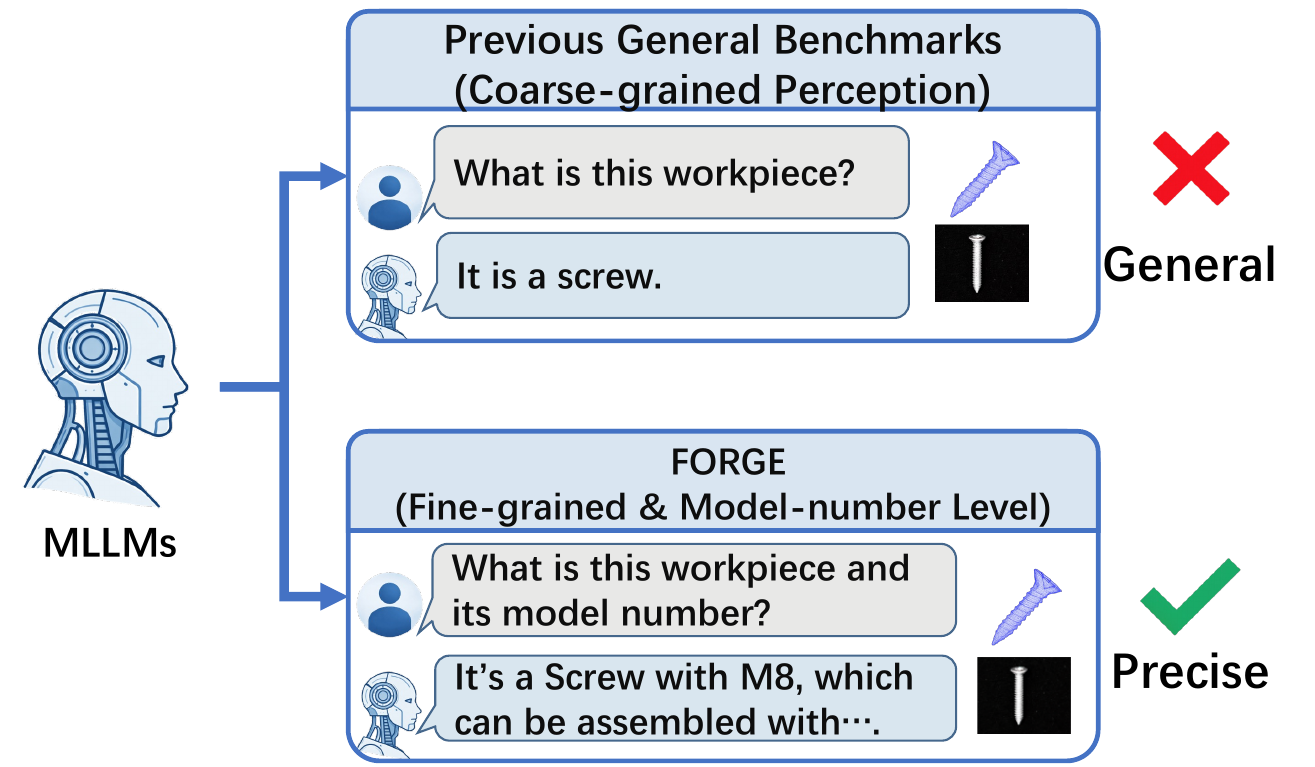}
        \caption{%Example of fine-grained domain semantics for \ours \ 
        }
        \label{fig:intro}
    \end{subfigure}
    %\hfill 
    \begin{subfigure}{0.58\textwidth}
        \centering
        \includegraphics[width=\linewidth]{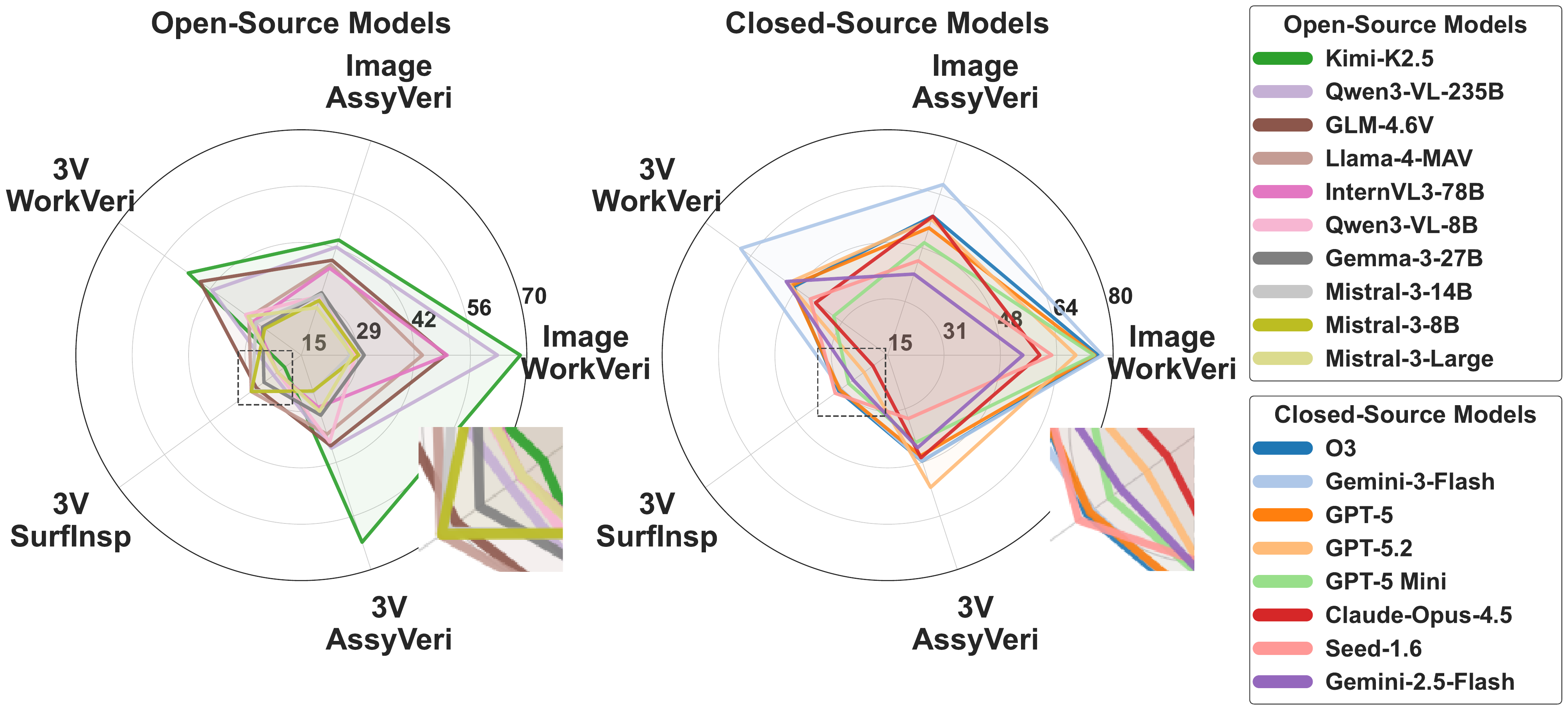} 
        \caption{%Radar map of all models across all given tasks.
        }  
        \label{fig:Radar}
    \end{subfigure}

    %\vspace{-0.3cm} 
    % \begin{subfigure}{\textwidth}
    %     \centering

    %     \includegraphics[width=1\linewidth]{mainfigs/framework_demo.png} 
    %     \caption{The main pipeline of \ours. under preparation}
    %     \label{fig:pipeline}
    % \end{subfigure}

    \caption{ \textbf{Benchmark Overview.} (a) The main pipeline of \ours. (b) Compared to previous benchmarks, \ours \  primarily evaluates the understanding, reasoning, and decision-making capabilities regarding fine-grained domain semantics in manufacturing scenarios.
    (c) The performance of all models across all given tasks. From left to right: open-source models and closed-source models.
} %\vspace{-10pt}
    \label{fig:main_overview}
\end{figure*}

More recently, Large Language Models (LLMs) and Multimodal Large Language Models (MLLMs) have demonstrated remarkable generalization and cross-task transfer capabilities across diverse domains, including multimodal media understanding~\citep{jian2025lazyvlm,rodriguez2025bigdocsopendatasettraining,jian-wang-2023-invgc}, topology analysis~\citep{zhao2026vision,xu2025graphomni,zhao2025underappreciated}, scientific research~\citep{pang2025paper2poster}, GUI navigation~\citep{nayak2025ui-vision,jian2026cuasuite,feizi2025groundingcomputeruseagents} and many other fields~\citep{li2025survey}. Despite this proven versatility, their application in manufacturing remains nascent. The integration of MLLMs presents a critical, yet underexplored, research avenue for manufacturing scenarios~\citep{lee2023unified,picard2025concept,zhang2025llm}. MLLMs can overcome the architectural bottleneck of traditional vision models by integrating heterogeneous data streams with advanced reasoning capabilities. This unique synthesis enables MLLMs to bridge the gap between low-level perception and high-level planning, catalyzing a transition in manufacturing intelligence from extraction to autonomous decision-making~\citep{du2025llm}. Motivated by this transformative potential, we pose the following core inquiry: \textbf{Can MLLMs understand, explain, and execute decisions for tasks inherently characteristic of the manufacturing domain?}

Recent studies have begun to address this question, as summarized in Table \ref{tab:differentiate}. Prior work has examined the performance of MLLMs on specific manufacturing tasks, such as visual anomaly detection~\citep{jiang2024mmad}, engineering documentation comprehension~\citep{doris2025designqa}, and broader manufacturing cognition~\citep{yi2025mme}.
However, current evaluations rarely assess MLLMs understanding of fine-grained domain semantics, and existing benchmarks do not reflect the rigorous demands of the real-world manufacturing domain~\citep{gao2024hierarchical,boysen2009sequencing,zhao2025industrial}. Progress in this domain is currently hindered by several fundamental challenges: \textbf{(i) Data Scarcity Gap.} 
Current manufacturing datasets are constrained by limited scale and diversity, so that many studies rely on simulated or CAD-based data~\citep{tao2024f2pad,khan2025leveraging,Tao2023Anomaly,xu2023ano}. \textbf{(ii) Lack of Fine-Grained Domain Semantics.}
 Many current manufacturing datasets merely treat manufacturing workpieces as generic visual subjects. They fail to integrate explicit, fine-grained domain semantics (\eg, model numbers of workpiece) that are essential to the rigorous demands of real-world manufacturing. \textbf{(iii) Absence of Comprehensive Evaluation Frameworks.} There is a lack of \textit{systematic and representative benchmarks} to assess the reasoning, understanding, and decision-making capabilities of MLLMs in manufacturing scenarios.

\definecolor{HeaderBg}{RGB}{25, 50, 90}        % Deep navy
\definecolor{HeaderFg}{RGB}{255, 255, 255}      % White text
\definecolor{SubHeaderBg}{RGB}{210, 225, 245}   % Light steel blue
\definecolor{RowAlt}{RGB}{242, 246, 252}        % Very light blue tint
\definecolor{OursBg}{RGB}{255, 243, 205}        % Warm amber highlight
\definecolor{OursBorder}{RGB}{200, 150, 30}     % Amber border accent
\definecolor{CheckGreen}{RGB}{34, 139, 34}      % Forest green
\definecolor{CrossRed}{RGB}{180, 30, 30}        % Deep red
\definecolor{MidRuleColor}{RGB}{150, 180, 220}  % Soft blue rule

%% Redefine cmark/xmark with color (place in preamble)
\newcommand{\cmarkc}{\textcolor{CheckGreen}{\ding{51}}}
\newcommand{\xmarkc}{\textcolor{CrossRed}{\ding{55}}}

\begin{table*}[t]
\centering
\caption{\textbf{Comparison of existing manufacturing MLLMs benchmarks with \ours.}}
%\vspace{-10pt}
\setlength{\tabcolsep}{5pt}
\resizebox{\textwidth}{!}{%
\begin{threeparttable}
\begin{tabular}{lcccccccc}
\toprule
%% ── Dark navy header row ──────────────────────────────────────────────────
\rowcolor{HeaderBg}
\textcolor{HeaderFg}{\textbf{Benchmarks}} &
\multicolumn{2}{c}{\textcolor{HeaderFg}{\textbf{Data Modality}}} &
\multicolumn{1}{c}{\textcolor{HeaderFg}{\textbf{Data Source}}} &
\multicolumn{3}{c}{\textcolor{HeaderFg}{\textbf{Granularity \tnote{$\dagger$}}}} &
\multicolumn{2}{c}{\textcolor{HeaderFg}{\textbf{Statistics}}} \\
%% ── Sub-header row ───────────────────────────────────────────────────────
\rowcolor{SubHeaderBg}
 & \textbf{Image} & \textbf{Point cloud} & \textbf{Real / Synthetic}
 & \textbf{Scenario} & \textbf{Workpiece} & \textbf{Model number}
 & \textbf{Availability} & \textbf{Samples} \\
\midrule
%% ── Data rows with alternating shading ───────────────────────────────────
\rowcolor{white}
MMAD~\citep{jiang2024mmad}
  & \cmarkc & \xmarkc & Real-world & \cmarkc & \cmarkc & \xmarkc & \cmarkc & 39{,}672 \\
\rowcolor{RowAlt}
MME-Industry~\citep{yi2025mme}
  & \cmarkc & \xmarkc & Real-world & \xmarkc & \xmarkc & \xmarkc & \cmarkc & 1{,}050 \\
\rowcolor{white}
DesignQA~\citep{doris2025designqa}
  & \xmarkc & \xmarkc & Synthetic  & \cmarkc & \xmarkc & \xmarkc & \xmarkc & 1{,}451 \\
\rowcolor{RowAlt}
FailureSensorIQ~\citep{constantinides2025failuresensoriq}
  & \xmarkc & \xmarkc & Real-world & \cmarkc & \xmarkc & \xmarkc & \cmarkc & 8{,}296 \\
\rowcolor{white}
EngDesign~\citep{guo2025toward}
  & \xmarkc & \xmarkc & Synthetic  & \cmarkc & \xmarkc & \xmarkc & \cmarkc & 1{,}717 \\
%% ── Separator before our method ──────────────────────────────────────────
\arrayrulecolor{MidRuleColor}\midrule\arrayrulecolor{black}
%% ── Highlighted "Ours" row ───────────────────────────────────────────────
\rowcolor{OursBg}
\textbf{\ours}
  & \cmarkc & \cmarkc & \textbf{Real-world}
  & \textbf{\cmarkc} & \textbf{\cmarkc} & \textbf{\cmarkc}
  & \cmarkc & \textbf{12{,}972} \\
\bottomrule
\end{tabular}
\begin{tablenotes}[flushleft]
\footnotesize
\item[$\dagger$] For Granularity, \textit{Workpiece-level} focuses on manufacturing workpieces;
         \textit{Scenario-level} focuses on specific manufacturing scenarios;
         \textit{Model-number-level} requires not only identifying the workpiece type
         but also the model number.
\end{tablenotes}
\end{threeparttable}
}%\vspace{-10pt}
\label{tab:differentiate}
\end{table*}

To address these challenges, we introduce \textbf{\ours}, a comprehensive benchmark tailored for the manufacturing domain. First, we collect, construct, and annotate a large-scale multimodal manufacturing dataset comprising image and point cloud data of representative workpieces across diverse \textbf{model numbers} (\eg, nuts ranging from M10 to M20), thereby capturing the \textbf{fine-grained domain semantics} of the real-world manufacturing domain. Furthermore, we adopt three evaluation tasks 
%, \textbf{Workpiece Verification(\taskone)}, \textbf{Structural Surface Inspection(\tasktwo)}, and \textbf{Assembly Verification(\taskthree)}, 
aligned with key manufacturing applications, including \textbf{material sorting}, \textbf{quality inspection}, and \textbf{assembly recognition}, providing a systematic and comprehensive framework for assessing MLLMs performance in manufacturing scenarios. Beyond benchmarking, we further propose a dedicated dataset for domain-specific fine-tuning. MLLMs fine-tuned on this dataset achieved substantial performance gains on unseen manufacturing tasks, validating the dataset's ability to bridge the domain knowledge gap and enhance model generalization in manufacturing settings.
 
In summary, our major contributions include:
%\vspace{-5pt}
\begin{itemize}  
\item \textbf{High-Quality Multimodal Manufacturing Dataset.} We present the first large-scale fine-grained manufacturing dataset that integrates aligned 2D images and 3D point clouds. The dataset provides rich multimodal annotations designed to support systematic evaluation and development of MLLMs for manufacturing perception and reasoning.

 \item \textbf{Real-World Manufacturing Cognitive Tasks.} Based on the collected fine-grained manufacturing dataset, we design three core manufacturing tasks, \textbf{Workpiece Verification (\taskone)}, \textbf{Structural Surface Inspection (\tasktwo)}, and \textbf{Assembly Verification (\taskthree)}, which demand MLLMs to perform fine-grained visual discrimination (in Figure \ref{fig:intro}) and complex logical reasoning (\eg, verifying assembly compatibility).
 
 \item \textbf{Extensive Benchmarking and Critical Insights.} We conduct a rigorous evaluation of state-of-the-art MLLMs under different evaluation settings. Our extensive experiments reveal significant performance gaps in microscopic surface analysis and manufacturing task reasoning, and identify morphology understanding and domain knowledge as the major bottlenecks when deploying MLLMs in manufacturing scenarios.

 \item \textbf{Actionable Training Resource.} Beyond evaluation, we demonstrate that our structured annotations can serve as training data for domain-specific fine-tuning. Supervised fine-tuning (SFT)~\citep{ouyang2022training} of a compact open-weight model on our training split yields substantial accuracy improvements on held-out manufacturing scenarios unseen during training, providing preliminary evidence that the dataset can help close the domain knowledge gap identified by our analysis.
\end{itemize}

% %\vspace{-15pt}

\section{Related Work} \label{sec:rel}

% \subsection{Benchmark in Manufacturing Scenario}\label{sec:rel_bench_manu}
Evaluating MLLMs in manufacturing scenarios is a critical yet nascent area.
MMAD~\citep{jiang2024mmad} establishes a standardized framework for visual anomaly detection and evaluates fine-grained perceptual ability on datasets such as MVTec-AD~\citep{bergmann2019mvtec}. However, manufacturing demands more than visual pattern recognition, as it requires domain-specific knowledge and reasoning. Recent works expand into specific cognitive domains: \textbf{MME-Industry}~\citep{yi2025mme} addresses manufacturing cognition and safety regulations; \textbf{DesignQA}~\citep{doris2025designqa} focuses on technical blueprints and standards; \textbf{EngDesign}~\citep{guo2025toward} targets design synthesis and constraint trade-offs; and \textbf{FailureSensorIQ}~\citep{constantinides2025failuresensoriq} evaluates reliability engineering and failure diagnosis.

Despite this progress, existing methods rarely validate adherence to highly structured and standardized manufacturing settings, which impose more stringent precision demands~\citep{gao2024hierarchical,boysen2009sequencing,zhao2025industrial,maji2013fine,yadav2018modelling,jiao2007product,gray1993synthesis}. Paradigms such as mixed-model assembly lines~\citep{boysen2009sequencing}, flexible manufacturing systems~\citep{yadav2018modelling}, product family design~\citep{jiao2007product}, and tool management~\citep{gray1993synthesis} all require not only classifying the general workpiece category but also distinguishing specific variants, demanding precise matching of fine-grained attributes such as model numbers. Current frameworks, constrained by the scarcity of multimodal data and limited fine-grained semantic annotations, fail to meet these requirements. We propose \textbf{\ours}, built on a \textbf{multimodal} dataset with \textbf{fine-grained annotations} that include various \textbf{workpiece attributes (\eg, model numbers)}. Three evaluation tasks are designed to assess MLLMs against these strict standards, providing an authentic reflection of their ability to handle real-world manufacturing complexity.

\section{\ours}

\begin{figure}[htbp]
  \centering
  \includegraphics[width=0.95\linewidth]{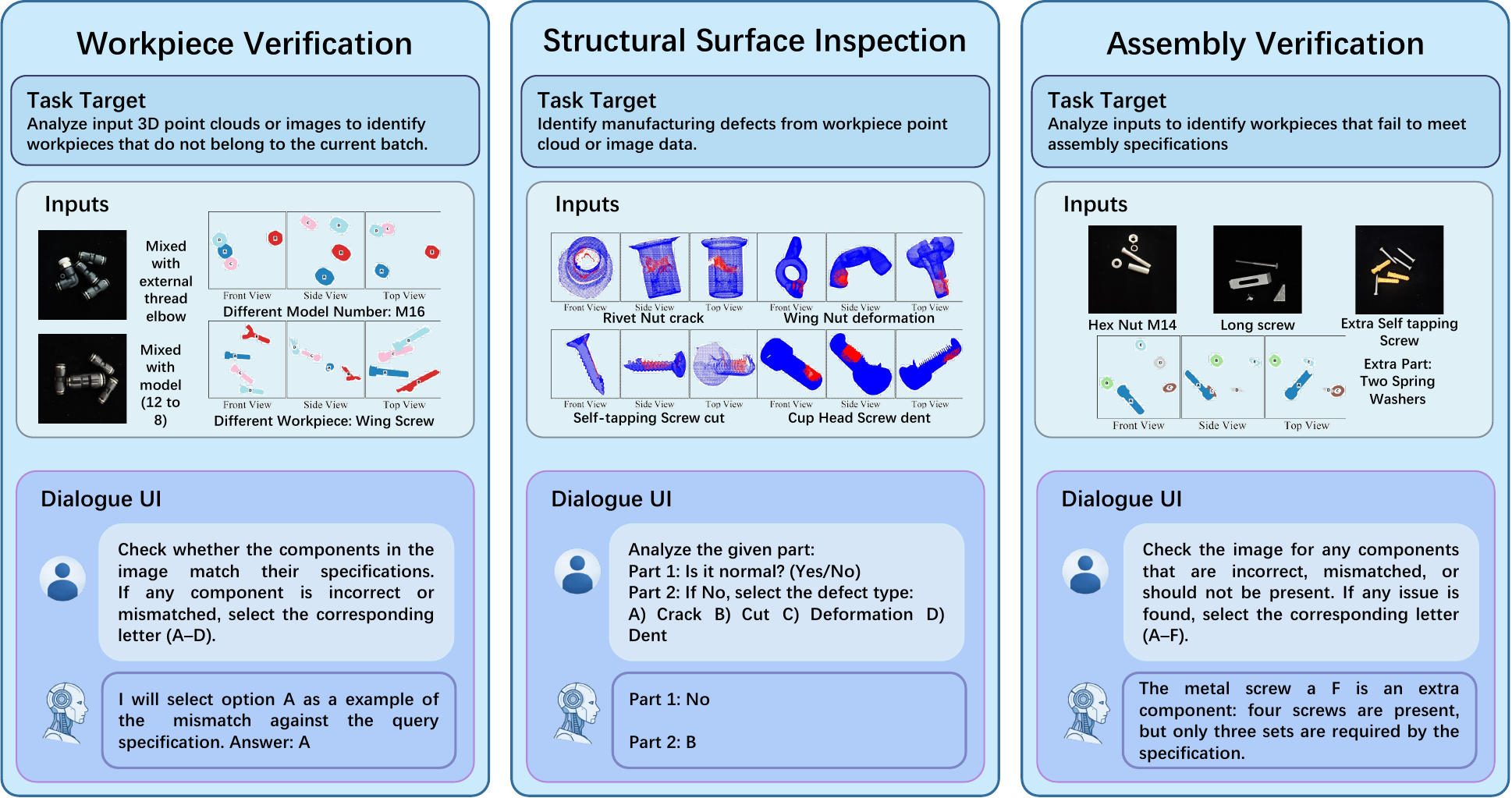}
  \caption{Task descriptions, input data, and dialogue examples of \ours.}%\vspace{-10pt}
  \label{fig:qaexample_input}
\end{figure}
\subsection{Dataset Curation}\label{subsec:datacuration}

%   To facilitate the evaluation of MLLMs in manufacturing settings, we constructed a comprehensive dataset comprising raw data from authentic manufacturing components.
\noindent\textbf{Dataset Collection.} We constructed a comprehensive dataset comprising raw data from authentic manufacturing components. The dataset comprises two subsets. \textbf{(i) 3D Point Cloud Subset}: Contains high-fidelity geometric data covering 14 workpiece categories across 90 distinct models. This subset supports tasks such as \taskone, \tasktwo, and \taskthree. \textbf{(ii) Image Subset}: Consists of approximately 3,000 images capturing four distinct manufacturing scenarios (\eg, expansion screw assemblies), including both normal and abnormal samples. All data were captured with fine-grained geometric and visual details.

\noindent\textbf{Dataset Processing.} The raw data, comprising both 2D images and 3D point clouds, required preprocessing before being utilized in \ours. For 2D images, ground-truth labels were established through a two-step process: automated contour and coordinate extraction, followed by manual refinement. For 3D point clouds, strategies varied by task. For \taskone \ and \taskthree, we synthesized batch samples by stitching 4-5 individual point clouds with random orientations, automatically generating labels during assembly. For \tasktwo, we simulated four typical manufacturing defects (Crack, Deformation, Dent, and Cut) using morphology-based algorithms and non-rigid deformation to ensure realism. 

The primary objective of \ours \ is to evaluate general MLLMs, as their advanced reasoning and versatility make them highly promising for autonomous manufacturing. A central challenge in this evaluation is that these models typically lack native 3D encoders. As validated by our bottleneck analysis (\textbf{C} in Section~\ref{subsec:bottleneckana}), directly feeding raw 3D data via text-based coordinate serialization is highly ineffective. To bridge this modality gap, we adopt a multi-view projection strategy: all 3D point cloud samples are rendered as three-view (3V) images (front, side, and top orthogonal projections). This approach preserves the essential geometric structure across complementary viewpoints while remaining fully compatible with standard visual inputs. While specialized 3D-language models (\eg, those utilizing PointNet++~\citep{qi2017pointnet++} or 3D Transformers~\citep{guo2021pct}) can process point clouds natively, focusing on them diverges from our core goal of benchmarking the generalizable cognitive capabilities of foundational MLLMs. The final dataset comprises approximately 12,000 samples across all tasks. Detailed specifications of the dataset statistics, specific object categories, and the data collection process are provided in the Appendix \ref{app: taskdescrip}. %Appendix \ref{app: data collect}.  

%\xjtd{we missing the two things here: 1. how we collect task 1 and 3 image, and how we render point cloud into three view}

%\noindent\textbf{Dataset Processing}. 

\begin{figure}[htbp]
  \centering
  \captionsetup{font={scriptsize}} 
  \captionsetup[subfigure]{font={scriptsize}, skip=2pt} 
  %\vspace{-12pt}  
  \begin{minipage}{0.68\textwidth} 
    \centering
    
    \begin{subfigure}[b]{\linewidth}
      \centering

      \includegraphics[width=\linewidth]{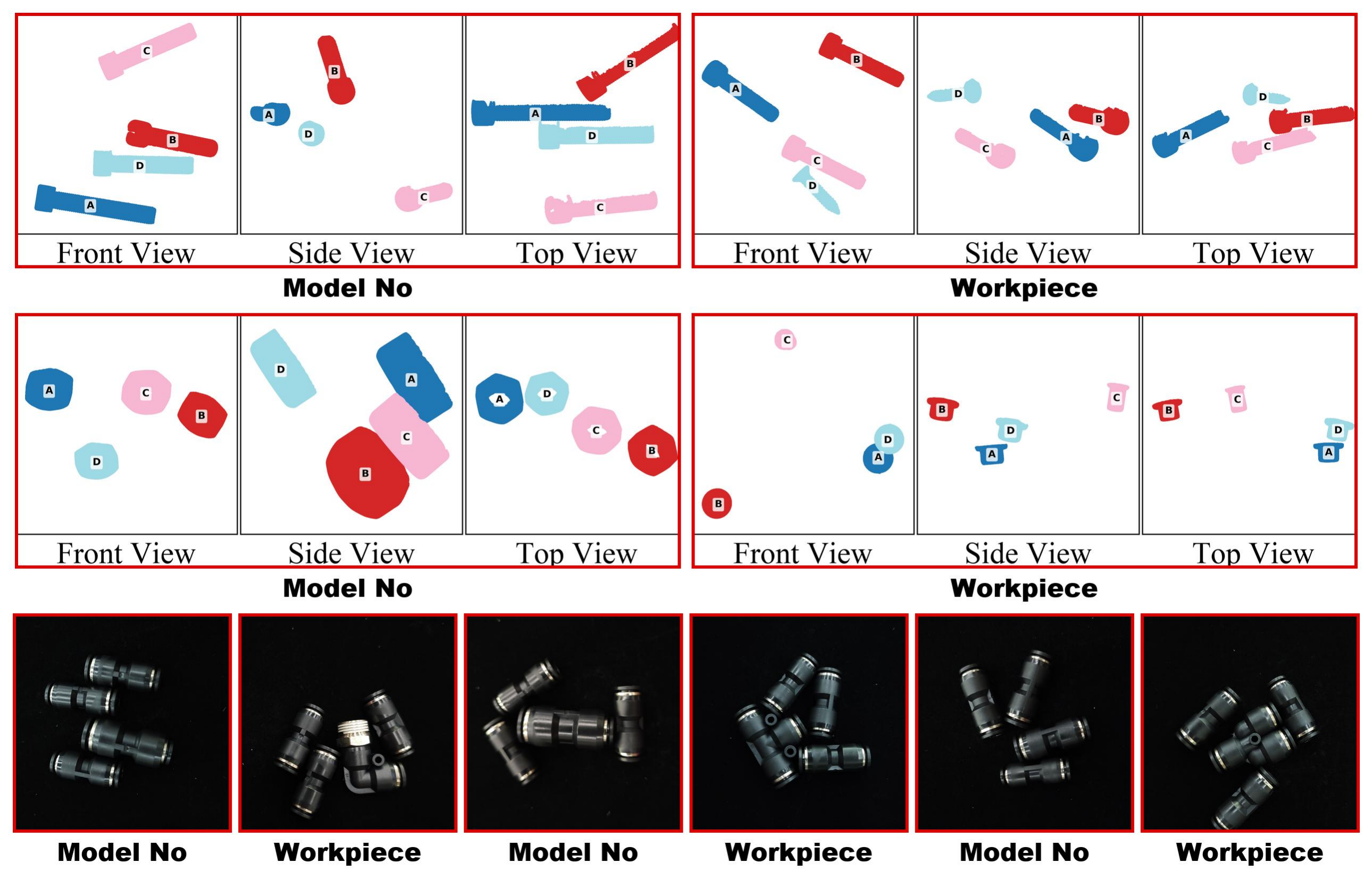}
      \caption{Data examples of \taskone} 
      \label{fig:task-a} 
    \end{subfigure}

    \begin{subfigure}[b]{\linewidth}
      \centering
      \includegraphics[width=\linewidth]{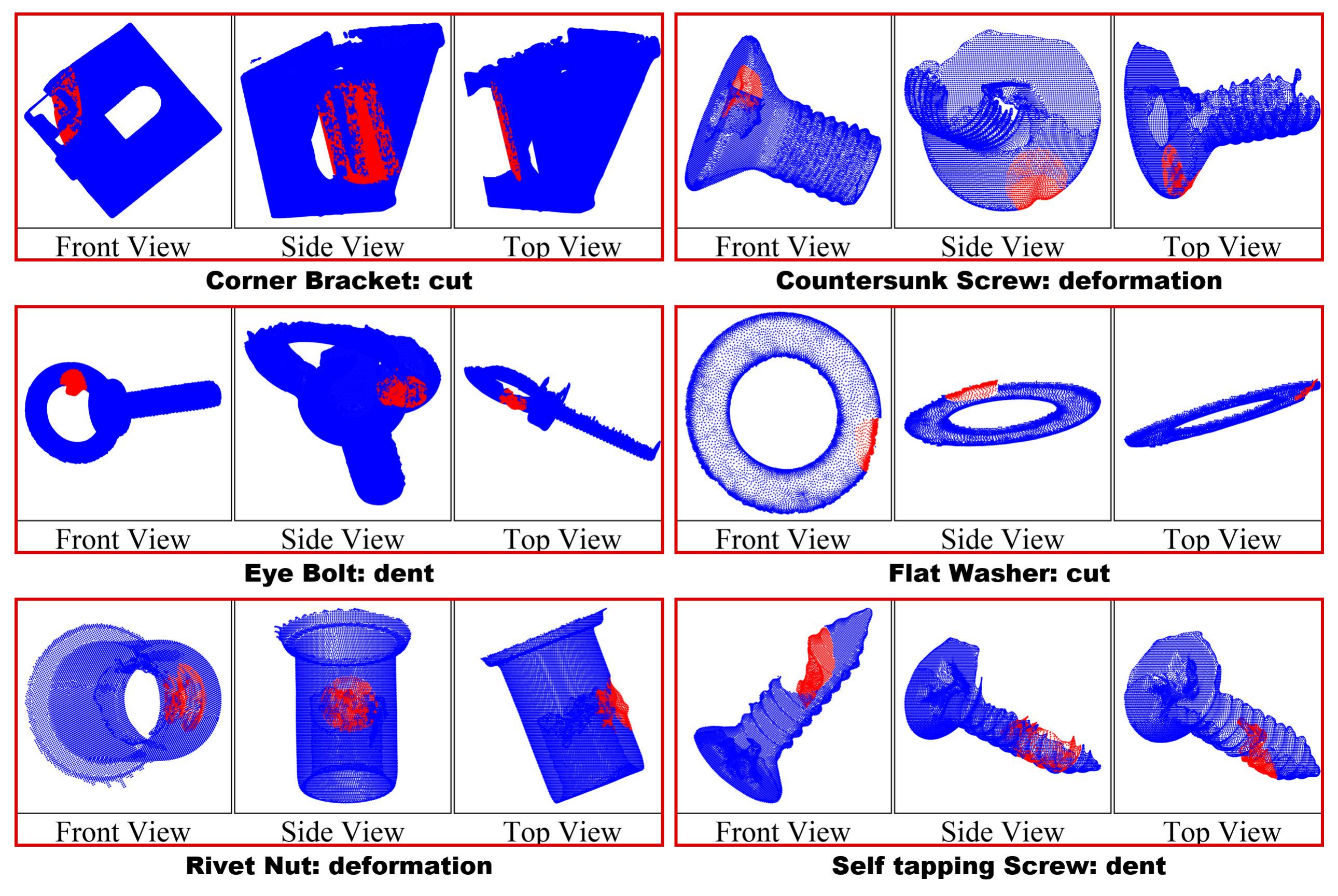}
      \caption{Data examples of \tasktwo}
      \label{fig:task-b}
    \end{subfigure}

    \begin{subfigure}[b]{\linewidth}
      \centering
      \includegraphics[width=\linewidth]{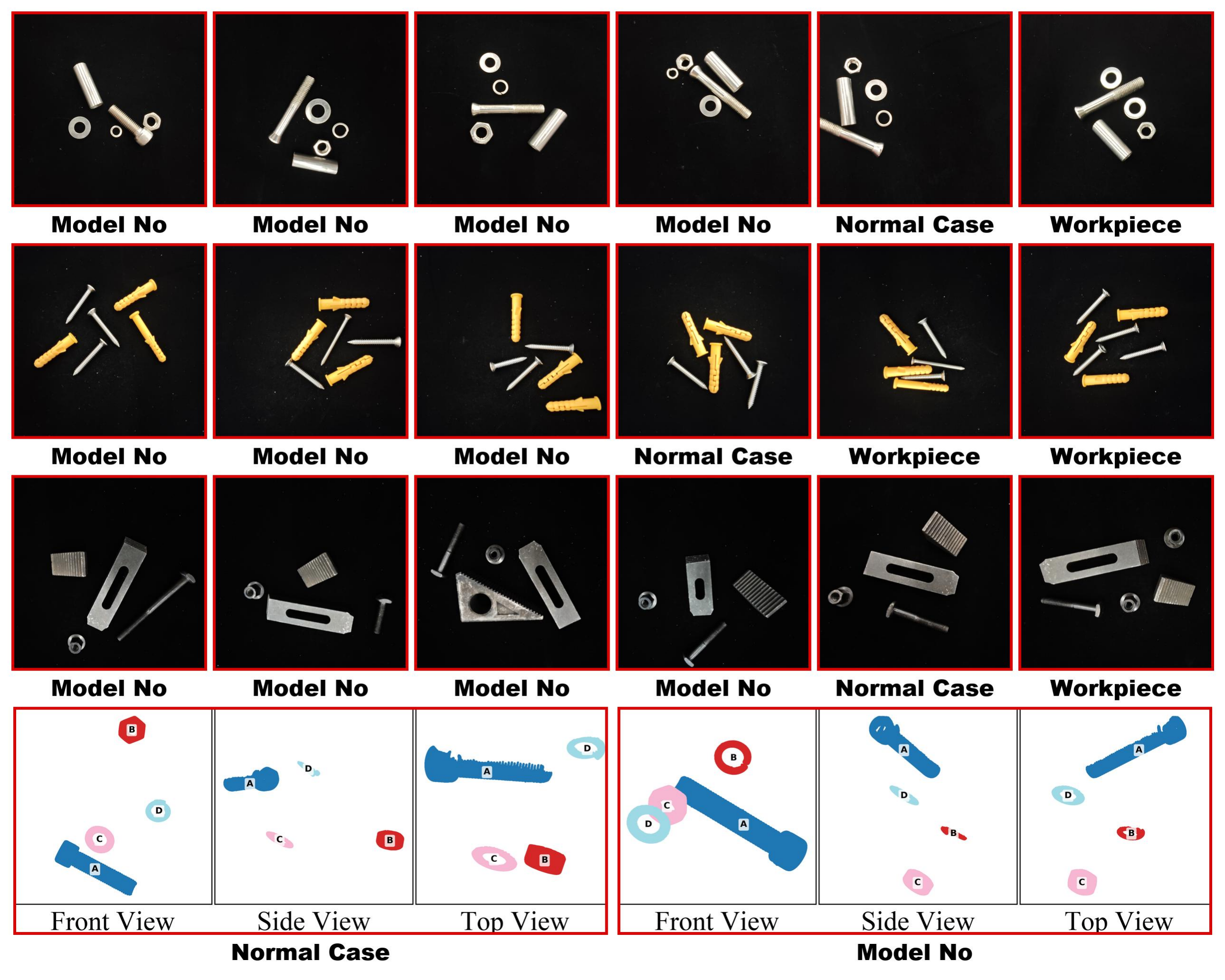}
      \caption{Data examples of \taskthree} 
      \label{fig:task-c}
    \end{subfigure}

    %\vspace{-8pt} 
    
    \caption{Qualitative results and analysis. 
    (a) \taskone: The top row shows examples of \chsscen, the second rows are from \nutsscen, and the bottom row is from \pcsscen. 
    (b) \tasktwo: Randomly selected defect examples across different workpieces.
    (c) \taskthree: The first three rows correspond to \messcen, \pesscen, and \cncscen, respectively, while the last row shows examples of \swnscen. 
    }
    %\vspace{-18pt}  
    
    \label{fig:combined_results}    
  \end{minipage}
\end{figure}

%\vspace{-10pt}
\subsection{Task Description}\label{subsc: taskdescri} 
%\xjtd{we try to say we "adopt" not "design", trying to show the sense that we use the real world setting, not random design by ourselves}
From raw material intake to final workpiece delivery, \textbf{material sorting}, \textbf{quality inspection}, and \textbf{assembly recognition} serve as the critical pillars of automation, driving nearly every stage of the manufacturing automation~\citep{zhao2025industrial,wang2018deep}. Recognizing their ubiquity and importance, \ours\  adopts three corresponding tasks, \textbf{Workpiece Verification(\taskone)}, \textbf{Structural Surface Inspection(\tasktwo)}, and \textbf{Assembly Verification(\taskthree)} to comprehensively evaluate MLLMs capabilities. Furthermore, to align with the stringent requirements of real-world manufacturing scenarios, we categorize potential error scenarios into two primary classes: \textit{Different workpiece} and \textit{Different Model Number} for \taskone \ and \taskthree. The former denotes coarse-grained failures, such as workpiece mismatches or missing components, while the latter targets fine-grained errors arising from subtle model variations. Finally, we designed \textbf{3} scenarios for \taskone, \textbf{4} scenarios for \taskthree, and \textbf{14} workpieces for \tasktwo. An example of the input for the three tasks, along with the corresponding outputs, is shown in Figure~\ref{fig:qaexample_input}. Additional data examples of the three tasks are presented in Figures~\ref{fig:task-a}, \ref{fig:task-b}, and \ref{fig:task-c}.

% \textbf{Task1: \taskone.}\taskone \ is designed to evaluate MLLMs ability in material sorting. Given explicit specifications, MLLMs must analyze input 3D point clouds or images to identify workpieces that do not satisfy the requirements. This task includes three scenarios covering two from the image subset, \nutsscen{} and image Pneumatic Connectors(\pcsscen), and one for point cloud, for Cup Head Screws(\chsscen). \textbf{Task2: \tasktwo.} The core objective of \tasktwo \ is to identify manufacturing defects from workpiece point cloud or image data. The task involves two steps: (1) defect detection and (2) defect type classification (\eg, Crack, Dent). Our evaluation covers 14 distinct manufacturing components with 3D point cloud data. \textbf{Task3: \taskthree.} \taskthree \ aims to assess MLLMs capability in understanding assembly relationships and compatibility constraints. Unlike basic sorting, this task requires reasoning over a given complex set of assembly rules. MLLMs must analyze inputs to identify workpieces that fail to meet assembly specifications. This task covers four scenarios involving three from the image subset, \ie, metal expansion screws(\messcen), plastic expansion screws(\pesscen), CNC fixtures(\cncscen), and one for point cloud data for the compatibility relationships among metal screws, washers, and nuts(\swnscen). %(see Appendix \ref{app: task3} for full dataset descriptions).

\noindent\textbf{Task 1: \taskone.}\label{subsc: task1} 
 \taskone\ is designed to evaluate how effectively MLLMs perform material sorting. Given explicit specifications, MLLMs must analyze 3D point clouds or images to identify workpieces that do not belong to the current batch. This task includes \textbf{three} scenarios: one from the image subset, pneumatic connectors (\pcsscen), and two point cloud datasets, cup head screws (\chsscen) and nuts (\nutsscen{}). 
 
\noindent\textbf{Task 2: \tasktwo.} \label{subsc: task2} 
The core objective of \tasktwo \ is to identify manufacturing defects from workpiece point cloud or image. The task involves two steps: (1) defect detection and (2) defect type classification (\eg, Crack, Dent). Our evaluation covers \textbf{14} distinct manufacturing components with 3D point cloud. 

\noindent\textbf{Task 3: \taskthree.}\label{subsc: task3} \taskthree \ aims to assess the capability of MLLMs in understanding assembly relationships and compatibility constraints. Unlike sorting, this task requires reasoning over a given complex set of assembly rules. MLLMs must analyze inputs to identify workpieces that fail to meet assembly specifications. This task covers \textbf{four} scenarios involving three from the image subset: metal expansion screws (\messcen), plastic expansion screws (\pesscen), CNC fixtures (\cncscen), and one for point cloud data for the compatibility among metal screws, washers, and nuts (\swnscen).

\noindent This task design establishes a challenging benchmark that goes beyond basic perceptual evaluation, rigorously assessing the fine-grained logical reasoning and judgment capabilities of MLLMs in real-world manufacturing scenarios. Detailed task description and examples are provided in the Appendix \ref{app: taskdescrip}.%Appendix \ref{app: taskdescrip} and Table \ref{table:raw_image_scenarios}.

\section{Experiments}\label{sc: Expe}
We evaluate 18 MLLMs on three manufacturing tasks, report main results with error case analysis, perform a bottleneck analysis to disentangle visual-perception from domain-knowledge limitations, and show the training potential of our benchmark by fine-tuning small models for cross-scenario generalization.
% \xj{one or two overall sentences, xhgg check section 4.1 and 4.2}

% %\vspace{-5pt}
\subsection{Evaluated Models}\label{subsc: baselines}

To provide a comprehensive assessment of the current MLLM landscape, we evaluate 18 representative models across open- and closed-source families (Table~\ref{tab:model_list}).

% %\vspace{-8pt}

%We include \textbf{8} open-source models from 5 organizations: \textbf{Gemma-3-27B}, \textbf{InternVL3-78B}, \textbf{Llama-4-Maverick} (MoE), three Mistral AI models at varying scales: \textbf{Ministral-3-8B}, \textbf{Mistral-3-14B}, and \textbf{Mistral-3-Large}, and two Alibaba Qwen models, \textbf{Qwen3-VL-235B-A22B-Thinking} (235B MoE with 22B active parameters and extended reasoning) and \textbf{Qwen3-VL-8B-Thinking} (8B compact variant with reasoning). Meanwhile, we evaluate \textbf{10} frontier and mid-scale proprietary models: four OpenAI models, \ie, \textbf{GPT-5}, \textbf{GPT-5.2}, \textbf{GPT-5-Mini}, and the reasoning-specialized \textbf{O3}; two generations of Google's efficient models. \textbf{Gemini-2.5-Flash} and \textbf{Gemini-3-Flash}; \textbf{Claude-4.5-Opus} (Anthropic), \textbf{GLM-4.6V} (Zhipu AI), \textbf{Kimi-K2.5} (Moonshot AI), and \textbf{Seed-1.6} (ByteDance). This selection ensures broad coverage across model families, parameter scales, and design philosophies, from compact 8B open-weight models to frontier closed-source ones.

%\vspace{-10pt}

\subsection{Evaluation Settings and Metrics}\label{subsc: metrics}

\noindent\textbf{Evaluation Protocol.}
All tasks in \ours\ are formulated as multiple-choice questions (MCQs). For \taskone\ and \taskthree, each assembly contains 4--6 components. In image-based evaluation, each MCQ option corresponds to a part identified by its normalized center coordinate (\eg, ``A.~Part at [0.70, 0.44]''), enabling the model to ground each choice to a specific spatial location in the image. In three-view evaluation, components are annotated with letter labels (A--F) using the Set-of-Mark~\citep{yang2023set} visual prompting strategy. In both cases, the model must select the letter corresponding to the anomalous component. For \tasktwo, the model must classify the surface condition of a given workpiece into one of five categories (\textit{crack}, \textit{cut}, \textit{deformation}, \textit{dent}, or \textit{good}).
We evaluate under three progressively informative settings: \textbf{i. Zero-Shot.} The model receives only the test image (or three-view rendering) and a task-specific query. No additional examples or references are provided. This setting measures the model's inherent ability to perform the task using only its pretrained knowledge and visual understanding. \textbf{ii. Reference-Conditioned (Ref-Cond).} In addition to the test case, the model is provided with reference images of correct, normal assemblies (or defect-free surfaces for \tasktwo). These references establish a visual baseline for what ``correct'' looks like, enabling the model to detect deviations by comparison. This setting evaluates whether explicit visual references improve anomaly detection. \textbf{iii. In-Context Demonstration (ICD).} Building upon Ref-Cond, the model additionally receives complete solved examples presented as multi-turn dialogue pairs, each containing a query, input data, and the correct answer. This setting tests whether full reasoning demonstrations can bridge the gap in domain-specific task understanding. We use 2 examples by default in this study, unless stated otherwise. To enable efficient and thorough evaluation of \tasktwo, we default to the 6 most representative and commonly used workpieces (60\% of \tasktwo\ in count). Additionally, error scenarios in \taskone\ and \taskthree\ are categorized into two difficulty levels: \textit{Different workpiece} (coarse-grained discrepancies, \eg, entirely wrong workpiece types or missing components) and \textit{Different Model Number} (fine-grained inconsistencies involving subtle variations within the same production line, such as different screw lengths or thread pitches).

% \begin{itemize}
%     \item \textbf{Zero-Shot.} The model receives only the test image (or three-view rendering) and a task-specific query. No additional examples or references are provided. This setting measures the model's inherent ability to perform the task using only its pretrained knowledge and visual understanding.
%     \item \textbf{Reference-Conditioned (Ref-Cond).} In addition to the test case, the model is provided with reference images of correct, normal assemblies (or defect-free surfaces for \tasktwo). These references establish a visual baseline for what "correct" looks like, enabling the model to detect deviations by comparison. This setting evaluates whether explicit visual references improve anomaly detection.
%     \item \textbf{In-Context Demonstration (ICD).} Building upon Ref-Cond, the model additionally receives complete solved examples presented as multi-turn user--assistant dialogue pairs, each containing a query, input data, and the correct answer. This setting tests whether full reasoning demonstrations can bridge the gap in domain-specific task understanding.
% \end{itemize}

\begin{wraptable}{r}{0.55\linewidth}
%\vspace{-60pt}
\centering
\caption{List of evaluated open-source and closed-source MLLMs.}
\label{tab:model_list}
\resizebox{\linewidth}{!}{%
\begin{tabular}{ll @{\hspace{6mm}} ll}
\toprule
\multicolumn{2}{c}{\textbf{Open-Source / Weights Models}} & \multicolumn{2}{c}{\textbf{Closed-Source Models}} \\
\cmidrule(lr){1-2} \cmidrule(lr){3-4}
\textbf{Provider} & \textbf{Model} & \textbf{Provider} & \textbf{Model} \\
\midrule
Google     & Gemma-3-27B          & OpenAI    & GPT-5 / 5.2 \\
OpenGVLab  & InternVL3-78B        & OpenAI    & GPT-5-Mini \\
Meta       & Llama-4-Maverick     & OpenAI        & O3\\
Mistral    & Mi(ni)stral-3-8B/14B/Large  & Google    & Gemini-2.5/3-Flash \\
Alibaba    & Qwen3-VL-8B/235B        & Anthropic & Claude-4.5-Opus \\
Zhipu AI   & GLM-4.6V             &  ByteDance & Seed-1.6 \\ 
Moonshot   & Kimi-K2.5            &           & \\
\bottomrule
\end{tabular}%
}
%\vspace{-30pt}
\end{wraptable}
\noindent\textbf{Metric.}
We adopt \textbf{exact-match accuracy} as the evaluation metric. For each test case, the model's predicted MCQ letter is extracted from its free-form response and compared with the ground-truth label. Accuracy is computed as the percentage of cases where the prediction exactly matches the correct answer. The random-chance baseline is the weighted average of a random guess across tasks and serves as a critical benchmark.

%\vspace{-8pt}

\subsection{Research Results and Analysis}\label{sec:results and analysis}
%\vspace{-12pt}
%\subsubsection{Main results }
Table~\ref{tab:mainresults} summarizes the main results across all tasks, modalities, and evaluation settings. Detailed extended results are provided in the Appendix \ref{app: extenresults}. From these results, we distill four key findings on current MLLM capabilities and limitations in manufacturing.

\begin{table*}[htbp]
\centering
\definecolor{deepblue}{RGB}{198,219,239}
\definecolor{deeporange}{RGB}{253,208,162}
\definecolor{lightblue}{RGB}{230,218,240}
\setlength{\tabcolsep}{3.2pt}
%\vspace{-10pt}
\caption{Main benchmark results (accuracy \%) across three manufacturing tasks. \colorbox{deeporange}{\textbf{Bold orange}} / \colorbox{deepblue}{\underline{underlined blue}} / \colorbox{lightblue}{light purple} indicate best / second / third within each model category (open- or closed-source) per row. \textbf{Rand.}\ = weighted random-guess baseline.}
%\vspace{-10pt}
\resizebox{\textwidth}{!}{%
\begin{tabular}{llc|cccccccccc|cccccccc|c}
\toprule
\multirow{2}{*}{\textbf{Task}} & \multirow{2}{*}{\textbf{Mod.}} & \multirow{2}{*}{\textbf{Setting}} & \multicolumn{10}{c|}{\textbf{Open-source Models}} & \multicolumn{8}{c|}{\textbf{Closed-source Models}} & \multirow{2}{*}{Random} \\
\cmidrule(lr){4-13}
\cmidrule(lr){14-21}
 &  &  & \rot{Gemma-3-27B} & \rot{InternVL3-78B} & \rot{Llama-4-MAV} & \rot{Mistral-3-14B} & \rot{Mistral-3-8B} & \rot{Mistral-3-Large} & \rot{Qwen3-VL-235B} & \rot{Qwen3-VL-8B} & \rot{Kimi-K2.5} & \rot{GLM-4.6V} & \rot{Claude-4.5-Opus} & \rot{Gemini-2.5-Flash} & \rot{Gemini-3-Flash} & \rot{GPT-5.2} & \rot{GPT-5} & \rot{GPT-5 Mini} & \rot{O3} & \rot{Seed-1.6} \\
\midrule
\multirow{6}{*}{\taskone} & \multirow{3}{*}{3V} & Zero Shot & 27.62 & 32.59 & 36.36 & 32.46 & 30.97 & 34.78 & \cellcolor{deeporange}\textbf{52.36} & 42.01 & \cellcolor{deepblue}\underline{50.00} & \cellcolor{lightblue}{47.42} & 39.19 & 56.05 & \cellcolor{deeporange}\textbf{69.56} & \cellcolor{deepblue}\underline{61.29} & \cellcolor{lightblue}{61.05} & 37.98 & 58.22 & 54.44 & 25.0 \\
 &  & Ref-Cond & 23.59 & 24.90 & 20.36 & 25.00 & 18.26 & 20.93 & \cellcolor{lightblue}{35.67} & 27.20 & \cellcolor{deepblue}\underline{36.76} & \cellcolor{deeporange}\textbf{44.29} & 27.88 & \cellcolor{deepblue}\underline{43.55} & \cellcolor{deeporange}\textbf{65.25} & 28.34 & 30.02 & \cellcolor{lightblue}{33.94} & 29.55 & 26.87 & 25.0 \\
 &  & ICD & 28.83 & 30.16 & 35.69 & 28.54 & 28.86 & 37.53 & \cellcolor{lightblue}{37.86} & 26.32 & \cellcolor{deeporange}\textbf{60.56} & \cellcolor{deepblue}\underline{44.90} & 55.04 & 53.85 & \cellcolor{deeporange}\textbf{67.34} & \cellcolor{deepblue}\underline{61.69} & 56.68 & 30.71 & \cellcolor{lightblue}{57.91} & 46.45 & 25.0 \\
 & \multirow{3}{*}{Image} & Zero Shot & 25.94 & 32.59 & 38.80 & 33.56 & 30.61 & 25.78 & \cellcolor{deepblue}\underline{64.08} & 35.41 & \cellcolor{deeporange}\textbf{66.75} & \cellcolor{lightblue}{50.44} & 59.42 & 55.78 & 72.22 & 72.50 & \cellcolor{deepblue}\underline{74.72} & \cellcolor{lightblue}{73.78} & \cellcolor{deeporange}\textbf{76.18} & 67.04 & 25.0 \\
 &  & Ref-Cond & 30.51 & \cellcolor{lightblue}{53.88} & 40.65 & 20.22 & 24.28 & 25.50 & \cellcolor{deepblue}\underline{58.01} & 24.04 & \cellcolor{deeporange}\textbf{69.67} & 50.33 & 56.22 & 51.66 & \cellcolor{deeporange}\textbf{76.27} & 55.48 & 64.22 & \cellcolor{deepblue}\underline{72.73} & \cellcolor{lightblue}{68.32} & 49.66 & 25.0 \\
 &  & ICD & 34.59 & \cellcolor{lightblue}{65.26} & 54.19 & 27.20 & 32.07 & 31.18 & \cellcolor{deepblue}\underline{66.44} & 25.28 & \cellcolor{deeporange}\textbf{68.86} & 50.00 & 60.98 & 54.55 & \cellcolor{lightblue}{82.26} & 79.87 & \cellcolor{deeporange}\textbf{85.23} & 77.01 & \cellcolor{deepblue}\underline{82.49} & 70.29 & 25.0 \\
\midrule
\multirow{6}{*}{\taskthree} & \multirow{3}{*}{3V} & Zero Shot & 27.18 & 34.30 & 38.19 & 28.16 & 24.60 & 31.64 & \cellcolor{deepblue}\underline{42.33} & 39.16 & \cellcolor{deeporange}\textbf{78.26} & \cellcolor{lightblue}{39.60} & 42.07 & 44.98 & 47.25 & \cellcolor{deeporange}\textbf{54.22} & \cellcolor{deepblue}\underline{53.90} & 49.51 & \cellcolor{lightblue}{52.38} & 41.69 & 32.8 \\
 &  & Ref-Cond & 32.04 & 22.01 & 30.10 & 19.09 & 18.45 & 29.54 & 31.01 & \cellcolor{lightblue}{33.77} & \cellcolor{deeporange}\textbf{57.14} & \cellcolor{deepblue}\underline{37.29} & \cellcolor{deepblue}\underline{50.81} & 40.13 & \cellcolor{lightblue}{46.28} & \cellcolor{deeporange}\textbf{56.96} & 30.74 & 29.55 & 36.46 & 26.71 & 32.8 \\
 &  & ICD & 32.04 & 28.80 & 37.54 & 26.06 & 29.55 & 27.03 & \cellcolor{deepblue}\underline{43.26} & \cellcolor{lightblue}{38.96} & \cellcolor{deeporange}\textbf{53.54} & 37.86 & 45.45 & 44.01 & 47.57 & \cellcolor{deeporange}\textbf{54.07} & \cellcolor{lightblue}{51.95} & 45.31 & \cellcolor{deepblue}\underline{52.46} & 34.30 & 32.8 \\
 & \multirow{3}{*}{Image} & Zero Shot & 28.47 & 33.14 & \cellcolor{deepblue}\underline{36.42} & 30.15 & 29.64 & 29.12 & \cellcolor{deeporange}\textbf{36.97} & 31.89 & 33.83 & \cellcolor{lightblue}{35.46} & \cellcolor{deepblue}\underline{52.10} & 39.56 & \cellcolor{deeporange}\textbf{58.11} & 48.18 & \cellcolor{lightblue}{50.06} & 43.69 & 48.93 & 39.86 & 29.5 \\
 &  & Ref-Cond & 31.70 & 36.40 & 32.12 & 32.12 & 26.84 & 23.74 & \cellcolor{deepblue}\underline{40.82} & 28.50 & \cellcolor{deeporange}\textbf{52.08} & \cellcolor{lightblue}{37.28} & \cellcolor{lightblue}{56.36} & 32.32 & \cellcolor{deeporange}\textbf{70.61} & 56.28 & 49.77 & 51.35 & \cellcolor{deepblue}\underline{60.16} & 42.49 & 29.5 \\
 &  & ICD & 32.94 & 42.46 & \cellcolor{lightblue}{46.13} & 29.46 & 30.65 & 28.67 & \cellcolor{deeporange}\textbf{50.43} & 30.32 & \cellcolor{deepblue}\underline{47.72} & 45.33 & \cellcolor{lightblue}{62.92} & 46.89 & \cellcolor{deeporange}\textbf{71.50} & \cellcolor{deepblue}\underline{63.99} & 60.94 & 52.29 & 62.34 & 48.52 & 29.5 \\
\midrule
\multirow{3}{*}{\tasktwo} & \multirow{3}{*}{3V} & Zero Shot & 21.75 & 19.19 & \cellcolor{deepblue}\underline{27.02} & \cellcolor{deeporange}\textbf{28.30} & \cellcolor{lightblue}{24.26} & 19.83 & 19.16 & 19.40 & 13.19 & 23.45 & 8.72 & 17.23 & 18.51 & 16.63 & \cellcolor{deepblue}\underline{22.01} & 17.02 & \cellcolor{lightblue}{21.11} & \cellcolor{deeporange}\textbf{22.60} & 20.0 \\
 &  & Ref-Cond & 23.88 & 21.63 & \cellcolor{lightblue}{24.09} & \cellcolor{deeporange}\textbf{27.72} & \cellcolor{deepblue}\underline{27.08} & 19.79 & 18.74 & 21.28 & 16.81 & 23.83 & 7.66 & 26.38 & 29.57 & 21.91 & \cellcolor{lightblue}{35.74} & 33.40 & \cellcolor{deeporange}\textbf{36.25} & \cellcolor{deepblue}\underline{36.17} & 20.0 \\
 &  & ICD & 33.19 & 25.74 & \cellcolor{deeporange}\textbf{39.15} & 33.19 & \cellcolor{deepblue}\underline{38.94} & 26.65 & 32.15 & 25.75 & 30.06 & \cellcolor{lightblue}{38.38} & \cellcolor{deepblue}\underline{44.35} & 38.09 & \cellcolor{deeporange}\textbf{47.12} & 31.70 & 38.30 & 36.25 & 40.00 & \cellcolor{lightblue}{42.31} & 20.0 \\
\bottomrule
\end{tabular}}
\label{tab:mainresults} %\vspace{-10pt}
\end{table*}

\noindent \textbf{A.Current MLLMs demonstrate better understanding of semantics than in morphological analysis.} 
In \taskone \ and \taskthree, based on Table \ref{tab:mainresults} and Figure \ref{fig:Radar}, the open-source model Kimi-K2.5 and the closed-source model Gemini-3-Flash both achieved leading performance. However, this relative competence did not extend to \tasktwo. Although the objectives of \tasktwo \ were relatively simple, it yielded the lowest performance among the three tasks. This performance disparity indicates that MLLMs exhibit fundamentally different capabilities for macroscopic part discrimination (recognition) and microscopic surface morphology analysis (perception). The poor performance on \tasktwo \ further indicates significant room for improvement in current MLLMs' understanding of physical details and microscopic features of manufacturing workpieces.

\noindent \textbf{B.Limited comprehension of domain knowledge is the bottleneck for current MLLMs.} By comparing evaluation results across Zero-Shot, Ref-Cond, and ICD-based settings based on image modality, we observed that in \taskone \ and \taskthree, simple Ref-Cond strategies did not consistently yield performance gains and, in several instances, even led to degradation. Conversely, ICD methods incorporating complete reasoning demonstrations achieved universal improvements over Zero-Shot baselines. This suggests that for complex manufacturing tasks, MLLMs do not lack simple sample references, but rather a deep understanding of task logic and reasoning paths, a gap that ICD bridges but Ref-Cond cannot. Notably, Section \ref{grounding_not} shows that visual grounding is not the bottleneck: MLLMs can identify these workpieces. Error case 1 in the Appendix \ref{app: errexemplars} also corroborates that MLLMs reveal a disconnect between perception and comprehension. %Supplementary error example corroborates this: MLLMs frequently fail tasks despite accurate object localization, indicating a disconnect between perception and comprehension.

%in Task 2, where the task logic is relatively simple, performance showed a gradual upward trend with the transition from Zero-Shot, Ref-Cond to ICD. This contrast further corroborates the conclusion that when tasks involve complex reasoning, MLLMs insufficient capability to understand the "task itself" constitutes the primary bottleneck.

%\vspace{-12pt}

\noindent \textbf{C. Given limited perceptual understanding of 3D spatial contexts, the introduction of additional examples hinders comprehension of MLLMs.}
Based on Table \ref{tab:mainresults}, for three-view modality, MLLMs achieved optimal performance under the Zero-Shot setting, whereas performance declined following the introduction of Ref-Cond and ICD. Given that the fundamental distinction between the three-view and image modalities lies in their multi-angular spatial representations, this counterintuitive phenomenon suggests that the contextual examples provided in Ref-Cond and ICD do not provide effective guidance. Instead, they induced spatial confusion within MLLMs, thereby impeding comprehension of manufacturing domain knowledge. 
Consequently, compared with model-number-level recognition, which relies less on spatial visual features, workpiece-level recognition, which is highly dependent on visual perception, is more severely affected by Ref-Cond and ICD. 
Furthermore, the experimental results regarding point cloud inputs in the Bottleneck Analysis \ref{pointfail} provide additional evidence of the current MLLMs limitations in explicit spatial perception capabilities.

\begin{figure}[htbp]
  \centering
  %\vspace{-10pt}
  \includegraphics[width=0.75\linewidth]{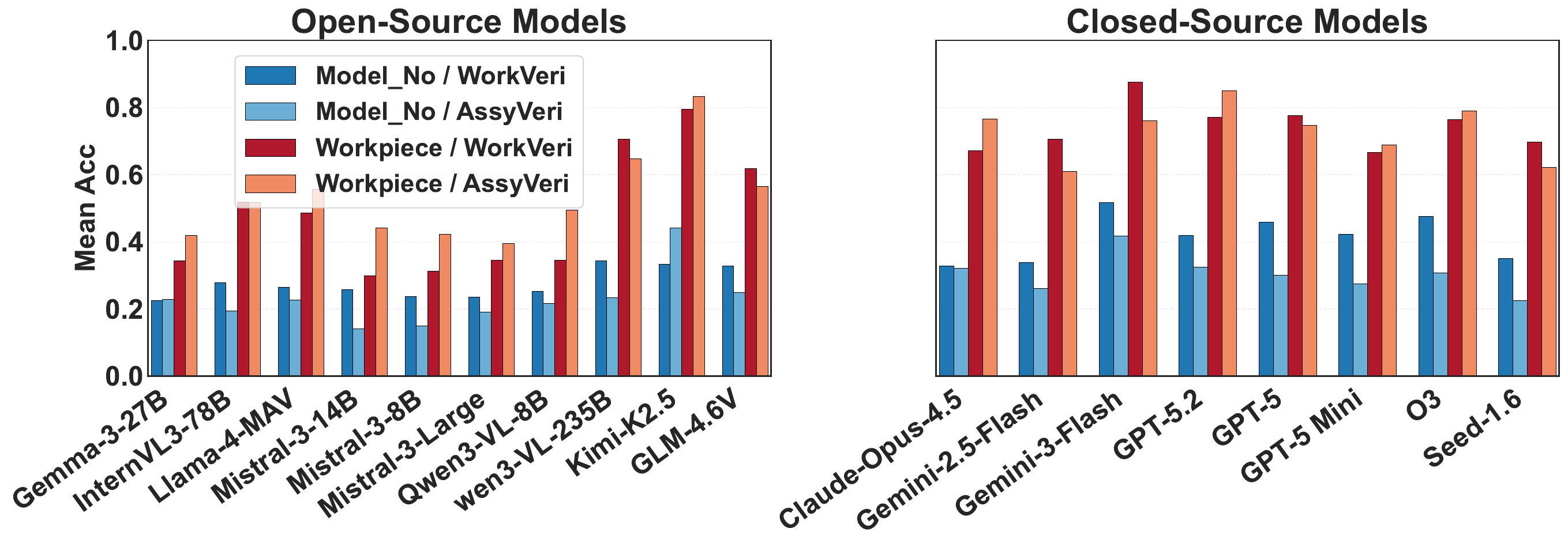} 
 %\vspace{-5pt}
 \caption{Performance of Open-Source vs. Closed-Source models.}%\vspace{-15pt}
  \label{fig:aveperf}
\end{figure}
% \subsubsection{finegrained}

\noindent \textbf{D. Model-number-level tasks are more challenging for MLLMs compared to workpiece-level tasks.}
A comprehensive analysis of \taskone\ and \taskthree\ based on Figure \ref{fig:aveperf} reveals a distinct performance disparity: MLLMs consistently outperform on workpiece-level tasks compared to model-number-level tasks. Whether for open-source or closed-source models, performance on model-number-level tasks (blue) is worse than on workpiece-level tasks (red).
%When examining the top-performing models, accuracy on workpiece-level identification reaches 92.01\% for open-source models (Qwen3-VL-235B) and 97.79\% for closed-source models (Kimi-K2.5). In sharp contrast, the best performance on model-number-level tasks peaks at only 33.45\% and 67.50\%, respectively.
This substantial gap indicates that while MLLMs have established a certain level of understanding of general manufactured workpieces, there remains significant room for improvement in capturing fine-grained domain specificity. Nevertheless, the capability to conduct fine-grained analysis is crucial in manufacturing scenarios, as manufacturing systems and tasks are discussed in Section~\ref{sec:rel}.

\subsection{Qualitative error cases analysis}\label{ab:erroranalysis}
% %\vspace{-2pt}
Here, we provide a qualitative analysis of error cases. Even with limited classification precision, MLLMs demonstrate distinct reasoning capabilities in manufacturing scenarios. For instance, the models exhibit latent reasoning potential in identifying workpiece materials and assessing workpiece service status. More detailed error cases are included in the Appendix  \ref{app: errexemplars}.

%Specific responses for the selected error cases are illustrated in Figure \ref{fig:error_exa}.
% \xjtd{qualitative analysis/ error analysis}
% \subsection{Results of Bottleneck Analysis}\label{ab:results }

% % \iffalse
% \begin{figure}[htbp]
%   \centering
%   %\vspace{-12pt}
%   \includegraphics[width=0.95\linewidth]{mainfigs/error_analysis.pdf}
%   \caption{Partial responses of two error examples.}
%   %\vspace{-12pt}
%   \label{fig:error_exa}
% \end{figure}
% % \fi

\begin{wrapfigure}{r}{0.65\linewidth}
%\vspace{-12pt}
\centering
\includegraphics[width=0.95\linewidth]{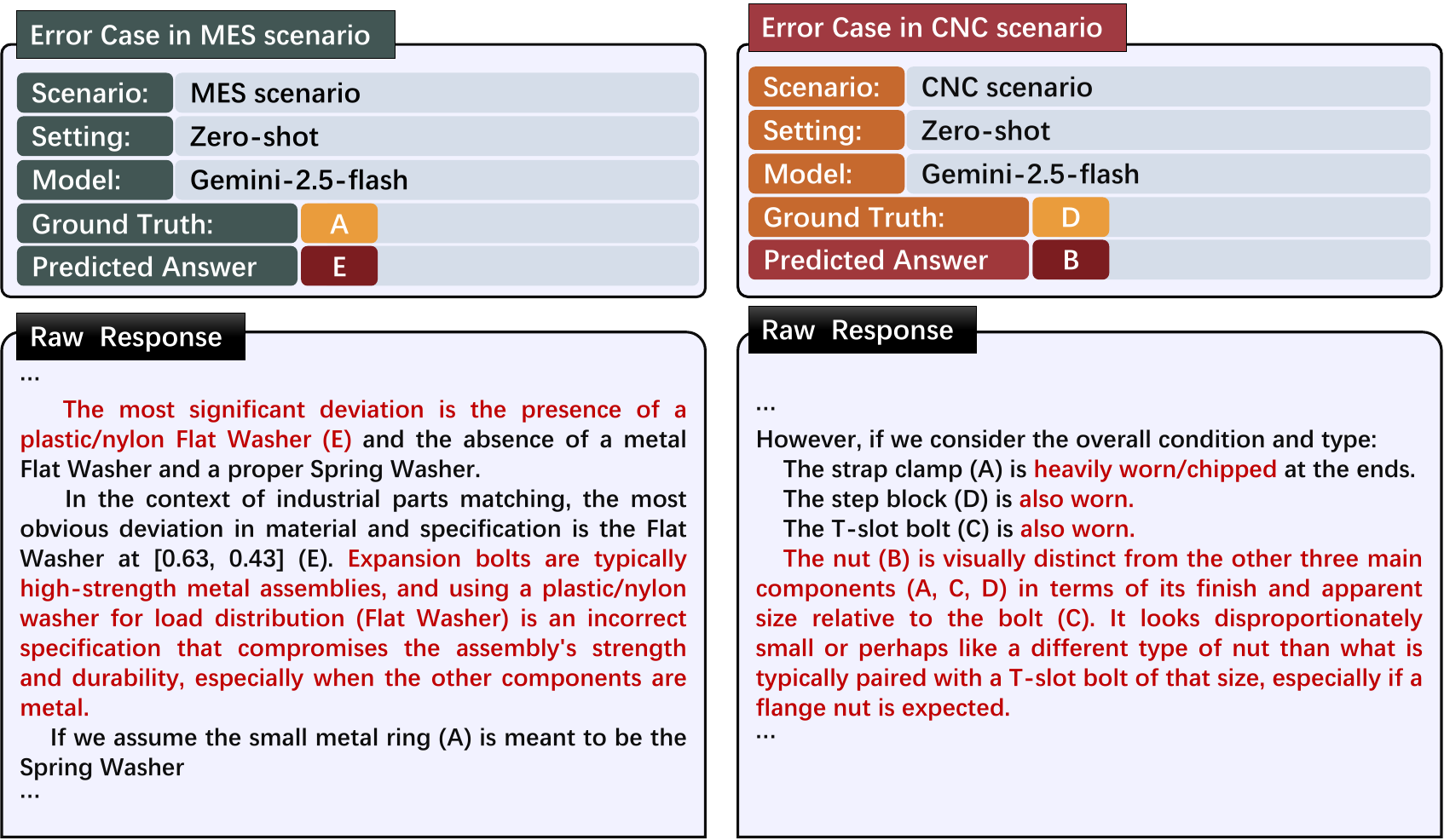}
\caption{Two error cases of \taskthree.}
\label{fig:error_exa}
 
\end{wrapfigure}

\noindent \textbf{A. Misjudging and over-relying on material properties.}
As shown in Figure \ref{fig:error_exa}, for the error case in \messcen, MLLMs incorrectly assess and unnecessarily factor in material properties. The model attempts to infer material composition from visual textures (\eg, "a plastic/nylon Flat Washer (E)") but misidentifies the material. Furthermore, it overcomplicates the reasoning process by relying on these erroneous material properties to make a judgment, even when the specific problem does not require material considerations. 
Despite the incorrect response, this indicates MLLMs are developing the potential to autonomously recognize workpiece materials and integrate inferred physical properties into manufacturing reasoning.

% Despite leading to an incorrect response, this result indicates that MLLMs are developing the potential to autonomously recognize workpiece materials from visual cues. The fact that the model proactively attempts to integrate these inferred physical properties into the specific manufacturing context demonstrates a promising trajectory toward deeper physical understanding.

\noindent \textbf{B. Failure on model number recognition but showing emerging capabilities in the service condition.}
In the error case in \cncscen of Figure \ref{fig:error_exa}, while MLLMs successfully identify the workpiece type, the model incorrectly concludes that Nut (B) is too small for the \cncscen, when in reality, Step Block (D) is too large. This typical error response directly echoes point B in Section \ref{sec:results and analysis}. 
Despite this misunderstanding of the model number recognition, the model's intermediate reasoning process reveals capabilities for evaluating service conditions. During its analysis, the model notes "the strap clamp (A) is heavily worn/chipped... block (D) is also worn." The ability to casually extract nuanced physical features, such as "heavy wear" or "chipping", indicates a potential for graded degradation assessment of workpieces. By recognizing these wear levels and integrating them into diagnostic reasoning, MLLMs may provide perceptual feedback that supports \textbf{Predictive Maintenance (PdM)}~\citep{sun2023robust,sun2017optimal}.

\subsection{Bottleneck Analysis}\label{subsec:bottleneckana}
% %\vspace{-3pt}
Our main benchmark requires multi-stage reasoning: a model must first \emph{ground} individual components, optionally \emph{compare} across reference and test images, and finally apply \emph{domain-specific logic} to reach a verdict. When a model fails, it is unclear which stage is the bottleneck. To disentangle these factors, we design three complementary Bottleneck Analysis and evaluate five representative models: Gemini-3-Flash, GPT-5.2, Qwen3-VL-235B, Seed-1.6, and Mistral-3-8B.

\noindent\textbf{A.~Visual grounding is not the bottleneck.}\label{grounding_not}
To disentangle perceptual failures from reasoning failures, we isolate each model's visual grounding ability using dedicated probing tasks. In our benchmark, each assembly image is annotated with letter labels (A, B, C, \etc) on individual parts, \ie Set-of-Mark~\citep{yang2023set} trick. We test whether models can correctly map between these labels and spatial coordinates in two settings: (i)~\emph{Single-image grounding}: given one annotated image, the model must either locate a part from its letter (\textbf{L$\to$C}) or identify a letter from given coordinates (\textbf{C$\to$L}); (ii)~\emph{Cross-image correspondence}: given two annotated images of assemblies from the same scenario, a part is identified in the first image and the model must find the visually corresponding part in the second image, using either letters (\textbf{L$\to$L}) or coordinates (\textbf{C$\to$C}). If models perform well on these grounding tasks but poorly on the full benchmark, the bottleneck lies in domain reasoning rather than perception. Presented in Table~\ref{tab:grounding_bottleneck}, Gemini-3-Flash achieves 98.9\% average on single-image grounding, and four of five models exceed 97.6\% on L$\to$C, the direction most relevant to our benchmark. These near-ceiling results confirm that failures on the full \taskone\ and \taskthree\ evaluations cannot be attributed to poor visual localization. Cross-image comparison is harder (84.3\% for Gemini, 80.5\% for GPT-5.2) but remains well above chance for the top four models on letter-based matching (more than 79.3\%), indicating it is a \emph{contributing} but not \emph{dominant} factor in the performance gap between Zero-Shot and Ref-Cond/ICD settings, as it validates MLLMs' ability to compare between images.

\begin{wraptable}{r}{0.55\linewidth}
%\vspace{-30pt}\centering
\definecolor{deepblue}{RGB}{198,219,239}
\definecolor{deeporange}{RGB}{253,208,162}
\definecolor{lightblue}{RGB}{230,218,240}
\setlength{\tabcolsep}{3.5pt}

\caption{\scriptsize \textbf{Visual grounding Bottleneck Analysis results (accuracy \%).} \emph{Single-image}: coordinate-to-letter (C$\to$L) and letter-to-coordinate (L$\to$C) on 500 cases. \emph{Cross-image}: letter-to-letter (L$\to$L) and coordinate-to-coordinate (C$\to$C) on 513 pairs.} 
\resizebox{0.95\linewidth}{!}{% 
\begin{tabular}{ll|>{\centering\arraybackslash}p{0.9cm}>{\centering\arraybackslash}p{0.9cm}>{\centering\arraybackslash}p{0.9cm}|>{\centering\arraybackslash}p{0.9cm}>{\centering\arraybackslash}p{0.9cm}>{\centering\arraybackslash}p{0.9cm}}
\toprule
\multirow{2}{*}{\textbf{Model}} & \multirow{2}{*}{\textbf{Type}} & \multicolumn{3}{c|}{\textbf{Single-Image}} & \multicolumn{3}{c}{\textbf{Cross-Image}} \\
\cmidrule(lr){3-5} \cmidrule(lr){6-8}
 & & \textbf{C$\to$L} & \textbf{L$\to$C} & \textbf{Avg.} & \textbf{L$\to$L} & \textbf{C$\to$C} & \textbf{Avg.} \\
\midrule
Gemini-3-Flash   & Closed & \cellcolor{deeporange}\textbf{98.2} & \cellcolor{deeporange}\textbf{99.6} & \cellcolor{deeporange}\textbf{98.9} & \cellcolor{deeporange}\textbf{88.7} & \cellcolor{deeporange}\textbf{79.9} & \cellcolor{deeporange}\textbf{84.3} \\
GPT-5.2          & Closed & \cellcolor{lightblue}{74.6} & 97.6 & \cellcolor{lightblue}{86.1} & \cellcolor{deepblue}\underline{85.6} & \cellcolor{deepblue}\underline{75.4} & \cellcolor{deepblue}\underline{80.5} \\
Qwen3-VL-235B    & Open   & \cellcolor{deepblue}\underline{85.4} & \cellcolor{lightblue}{98.8} & \cellcolor{deepblue}\underline{92.1} & \cellcolor{lightblue}{80.3} & \cellcolor{lightblue}{72.2} & \cellcolor{lightblue}{76.3} \\
Seed 1.6         & Closed & 42.0 & \cellcolor{deepblue}\underline{99.2} & 70.6 & 79.3 & 71.2 & 75.2 \\
Mistral-3-8B     & Open   & 66.0 & 70.6 & 68.3 & 62.0 & 33.9 & 48.0 \\
\bottomrule
\end{tabular}}
% \caption{\textbf{Visual grounding Bottleneck Analysis results (accuracy \%).} \emph{Single-image}: coordinate-to-letter (C$\to$L) and letter-to-coordinate (L$\to$C) on 500 cases. \emph{Cross-image}: letter-to-letter (L$\to$L) and coordinate-to-coordinate (C$\to$C) on 513 pairs.} 

% \colorbox{deeporange}{\textbf{Bold orange}}/\colorbox{deepblue}{\underline{Underlined blue}}/\colorbox{lightblue}{Light purple} = best/2nd/3rd
\label{tab:grounding_bottleneck}
%\vspace{-15pt}
\end{wraptable}

% \noindent\textbf{Coordinate-based tasks consistently lag letter-based tasks.}\xj{this might not be useful later, just put here for now}
% Across all models and both Bottleneck Analysis types, tasks involving raw coordinates are harder than their letter-based counterparts. In cross-image comparison, letter-to-letter outperforms coordinate-to-coordinate by 8--28 percentage points. This confirms that letter labels serve as effective visual anchors, while raw coordinate interpretation imposes an additional spatial reasoning burden that varies dramatically across models (\eg, Seed~1.6 achieves 99.2\% on L$\to$C but only 42.0\% on C$\to$L).

% ── Missing Part Bottleneck Analysis Table ──
%\begin{table}[t]

%  Models are given an explicit assembly description and must identify which component is absent from an MCQ. 
%\end{table}

\noindent\textbf{B. Fine-grained part identification remains a domain-knowledge bottleneck.}\label{missing_part_bottleneck}
To further disentangle \emph{domain-specific reasoning} from explicit \emph{ground} capabilities, we conduct a bottleneck analysis focusing on the missing part scenario. In this setup, each model is provided with an explicit assembly specification (\ie, a comprehensive list of required parts, their counts, and functional descriptions) and must identify the absent component via an MCQ. This directly probes domain knowledge: detecting that ``something is missing'' requires only counting, but pinpointing \emph{which} part is absent demands understanding the visual and functional distinctions between components. As shown in Table~\ref{tab:missing_part_bottleneck} (The column headers indicate the specific missing workpiece type in that scenario), the top four models achieve 74.9--90.7\% overall accuracy on images, which is well above the 23.3\% random baseline and thus demonstrates that MLLMs can reason about assembly completeness when given structured descriptions. Performance is near-perfect on most part types where components are visually distinctive (\ie, screws, nuts, anchors, wedges). However, a systematic failure is observed in flat washer detection (23.3--60.0\% on images, 8.3--74.5\% on three-view), where all five models struggle to some extent. Error analysis reveals the models can reliably detect that \emph{a washer} is absent but cannot determine \emph{which} washer, despite the two having distinct physical forms (further discussion and examples in the Appendix \ref{app: bottleneck}). %Error analysis reveals that when the Flat Washer is missing, Gemini~3~Flash, Qwen3-VL-235B, and Seed~1.6 predict Spring Washer in more than 90\% of their errors, \ie, the models reliably detect that \emph{a washer} is absent but cannot determine \emph{which} washer, despite the two having distinct physical forms (flat disc vs.\ helical split ring). 
Since the grounding ability analysis (Table~\ref{tab:grounding_bottleneck}) confirms that these models can accurately localize individual components, these confusion points indicate insufficient fine-grained manufacturing knowledge of the functional and morphological differences between part variants rather than a perceptual failure. A secondary finding concerns normal-case recognition: Seed~1.6 achieves only 43.3\% on normal cases overall (as low as 10\% on \messcen) while maintaining 89.5\% on most missing-part subcases, revealing a bias toward predicting a missing component rather than confirming completeness, again a reasoning rather than perception limitation.

\begin{table*}[htbp]
\centering
\definecolor{deepblue}{RGB}{198,219,239}
\definecolor{deeporange}{RGB}{253,208,162}
\definecolor{lightblue}{RGB}{230,218,240}
%\vspace{-15pt}

\begin{minipage}[t]{0.6\linewidth}
\centering
\setlength{\tabcolsep}{2.5pt}
\caption{\scriptsize \textbf{Zero-shot missing-part detection.} Superscripts denote scenario: $\,^1$\messcen\ (6 options), $\,^2$\pesscen\ (3 options), $\,^3$\cncscen\ (5 options) . Three-view covers \swnscen\ only (5 options). FW=Flat Washer, SW=Spring Washer, Sc=Screw, An=Anchor, Nu=Nut, We=Wedge, Norm=No missing part.}
\label{tab:missing_part_bottleneck}
\resizebox{0.95\linewidth}{!}{% 
\begin{tabular}{l | c c | c c | c c c | c | c || c c | c | c}
\toprule
\multirow{2}{*}{\textbf{Model}} & \multicolumn{9}{c||}{\textbf{Image (240 cases)}} & \multicolumn{4}{c}{\textbf{Three-View (137)}} \\
\cmidrule(lr){2-10} \cmidrule(lr){11-14}
 & \textbf{FW\textsuperscript{1}} & \textbf{SW\textsuperscript{1}} & \textbf{Sc\textsuperscript{2}} & \textbf{An\textsuperscript{2}} & \textbf{Nu\textsuperscript{3}} & \textbf{Sc\textsuperscript{3}} & \textbf{We\textsuperscript{3}} & \textbf{Norm} & \textbf{All} & \textbf{FW} & \textbf{SW} & \textbf{Norm} & \textbf{All} \\
\midrule
Gemini-3-Flash   & \cellcolor{deepblue}\underline{36.7} & \cellcolor{deeporange}\textbf{100} & \cellcolor{deeporange}\textbf{100} & \cellcolor{deepblue}\underline{96.7} & \cellcolor{deeporange}\textbf{100} & \cellcolor{deeporange}\textbf{100} & \cellcolor{deepblue}\underline{95.0} & \cellcolor{deeporange}\textbf{98.3} & \cellcolor{deeporange}\textbf{90.7} & 8.3 & \cellcolor{lightblue}{84.6} & \cellcolor{deeporange}\textbf{100} & 63.5 \\
GPT-5.2          & \cellcolor{deeporange}\textbf{60.0} & \cellcolor{deepblue}\underline{83.3} & \cellcolor{deepblue}\underline{90.0} &\cellcolor{deepblue}\underline{96.7} & \cellcolor{deeporange}\textbf{100} & \cellcolor{deeporange}\textbf{100} & \cellcolor{lightblue}85.0 & \cellcolor{lightblue}{80.0} & \cellcolor{lightblue}{85.0} & \cellcolor{deeporange}\textbf{74.5} & \cellcolor{deepblue}\underline{87.2} & \cellcolor{deeporange}\textbf{100} & \cellcolor{deeporange}\textbf{87.5} \\
Qwen3-VL-235B    & 23.3 & \cellcolor{deeporange}\textbf{100} & \cellcolor{deeporange}\textbf{100} & \cellcolor{deeporange}\textbf{100} & \cellcolor{deepblue}\underline{95.0} & \cellcolor{deeporange}\textbf{100} & \cellcolor{deeporange}\textbf{100} & \cellcolor{deepblue}\underline{85.0} & \cellcolor{deepblue}\underline{86.2} & \cellcolor{lightblue}{29.2} & \cellcolor{lightblue}{84.6} & \cellcolor{lightblue}86.0 & \cellcolor{lightblue}{65.7} \\
Seed 1.6         & \cellcolor{lightblue}26.7 & \cellcolor{deeporange}\textbf{100} & \cellcolor{deeporange}\textbf{100} & \cellcolor{deeporange}\textbf{100} & \cellcolor{lightblue}90.0 & \cellcolor{deepblue}\underline{89.5} & \cellcolor{deeporange}\textbf{100} & 43.3 & 74.9 & \cellcolor{deepblue}\underline{41.7} & \cellcolor{deeporange}\textbf{89.7} & \cellcolor{lightblue}{88.0} & \cellcolor{deepblue}\underline{72.3} \\
Mistral-3-8B     & \cellcolor{deepblue}\underline{36.7} & \cellcolor{lightblue}40.0 & \cellcolor{lightblue}44.8 & \cellcolor{lightblue}23.3 & 5.0 & \cellcolor{lightblue}0.0 & 80.0 & 8.3 & 27.2 & 10.4 & 56.4 & 4.0 & 21.2 \\
\bottomrule
\end{tabular}}
\end{minipage}
\hfill
\begin{minipage}[t]{0.38\linewidth}
\centering
\setlength{\tabcolsep}{3.5pt}
\caption{\scriptsize \textbf{Raw point cloud text input Bottleneck Analysis (accuracy \%).} 3D coordinates are serialized as integer-scaled text tables and fed directly to MLLMs, bypassing visual rendering. ZS\,=\,Zero-Shot, RC\,=\,Ref-Cond.}
\label{tab:raw_point}
\resizebox{0.95\linewidth}{!}{% 
\begin{tabular}{l l | c c c | c c c | c c c}
\toprule
\multirow{2}{*}{\textbf{Model}} & \multirow{2}{*}{\textbf{Type}} & \multicolumn{3}{c|}{\textbf{\taskthree}} & \multicolumn{3}{c|}{\textbf{\tasktwo}} & \multicolumn{3}{c}{\textbf{\taskone}} \\
\cmidrule(lr){3-5} \cmidrule(lr){6-8} \cmidrule(lr){9-11}
 &  & \textbf{ZS} & \textbf{RC} & \textbf{ICD} & \textbf{ZS} & \textbf{RC} & \textbf{ICD} & \textbf{ZS} & \textbf{RC} & \textbf{ICD} \\
\midrule
Gemini-3-Flash & Closed & 25.2 & 32.7 & 35.0 & 22.6 & 19.8 & 20.2 & \textbf{53.6} & \textbf{48.1} & \textbf{37.1} \\
Qwen3-235B     & Open   & 25.2 & \textbf{34.2} & 32.7 & 11.1 & 10.9 & 17.7 & 35.0 & 39.1 & 36.7 \\
\midrule
\multicolumn{2}{l|}{\emph{Random baseline}} & \multicolumn{3}{c|}{25.0} & \multicolumn{3}{c|}{20.0} & \multicolumn{3}{c}{25.0} \\
\bottomrule
\end{tabular}}
\end{minipage}

%\vspace{-10pt}
\end{table*}

% \begin{wraptable}{r}{0.60\linewidth}
% %\vspace{-12pt}
% %\begin{table}[ht]
% \centering
% \setlength{\tabcolsep}{3.5pt}
% \resizebox{0.95\linewidth}{!}{% 
% \begin{tabular}{l l | c c c | c c c | c c c}
% \toprule
% \multirow{2}{*}{\textbf{Model}} & \multirow{2}{*}{\textbf{Type}} & \multicolumn{3}{c|}{\textbf{\taskthree}} & \multicolumn{3}{c|}{\textbf{\tasktwo}} & \multicolumn{3}{c}{\textbf{\taskone}} \\
% \cmidrule(lr){3-5} \cmidrule(lr){6-8} \cmidrule(lr){9-11}
%  &  & \textbf{ZS} & \textbf{RC} & \textbf{ICD} & \textbf{ZS} & \textbf{RC} & \textbf{ICD} & \textbf{ZS} & \textbf{RC} & \textbf{ICD} \\
% \midrule
% Gemini-3-Flash & Closed & 25.2 & 32.7 & 35.0 & 22.6 & 19.8 & 20.2 & \textbf{53.6} & \textbf{48.1} & \textbf{37.1} \\
% Qwen3-235B     & Open   & 25.2 & \textbf{34.2} & 32.7 & 11.1 & 10.9 & 17.7 & 35.0 & 39.1 & 36.7 \\
% \midrule
% \multicolumn{2}{l|}{\emph{Random baseline}} & \multicolumn{3}{c|}{20.0} & \multicolumn{3}{c|}{20.0} & \multicolumn{3}{c}{20.0} \\
% \bottomrule
% \end{tabular}}
% \caption{\textbf{Raw point cloud text input Bottleneck Analysis (accuracy \%).} 3D coordinates are serialized as integer-scaled text tables and fed directly to MLLMs, bypassing visual rendering. ZS\,=\,Zero-Shot, RC\,=\,Ref-Cond.} % ICD\,=\,In-Context Demonstration.
% \label{tab:all-rows}
% %\end{table}
% %\vspace{-15pt}
% \end{wraptable}

\noindent\textbf{C. Visual projection is necessary: the text channel cannot replace it for generic MLLMs.} \label{pointfail}
Since our benchmark targets general-purpose MLLMs that lack native 3D encoders (Section~\ref{subsec:datacuration}), a natural question is whether the \emph{text} modality can serve as an alternative 3D interface by feeding raw coordinates directly as token sequences. We test this by serializing point clouds as integer-scaled text tables and querying two representative models: Gemini-3-Flash (multimodal) and Qwen3-235B (text-only). As shown in Table~\ref{tab:raw_point}, both models perform near the 20\% random baseline on \tasktwo\ (surface defect classification), with per-class analysis revealing that models default to a dominant prediction rather than genuinely discriminating defect types. Only \taskone\ shows a moderate signal above chance (Gemini-3-Flash: 53.6\% zero-shot), suggesting that coarse-grained shape comparison can partially exploit coordinate distributions. These results confirm that, among the input channels available to general-purpose MLLMs, visual rendering via multi-view projection is a relatively more effective interface for 3D manufacturing data. 

% We note that specialized 3D-language architectures with dedicated point cloud encoders represent a complementary research direction; however, they fall outside the scope of this benchmark, which evaluates the broad class of general-purpose MLLMs.

% \xjtd{we will need to point the related section in appendix eventually for all the Bottleneck Analysis here}

% Detailed Bottleneck Analysis, including per-task breakdowns and representative error cases, is provided in Supplementary Materials. %Appendix~\ref{subsec:bottleneckana_appendix}.

% %\vspace{-10pt}

\subsection{From Benchmark to Training Resource}\label{subsec:training}

The preceding analyses identify insufficient manufacturing domain knowledge as the primary bottleneck (Section~\ref{grounding_not}). We investigate whether \ours\ annotations can also serve as an \emph{actionable training resource} to close this gap. We fine-tune Qwen2.5-VL-3B-Instruct using task-specific SFT and adopt a \emph{scenario-based} train/eval split: for \taskone\ three-view, training on the \chsscen\ and evaluating on the held-out \nutsscen, and for \taskthree\ image, training on the \messcen\ and \pesscen\ and evaluating on the held-out \cncscen. This out-of-distribution protocol ensures that observed improvements reflect genuine acquisition of transferable manufacturing reasoning rather than memorization of specific assembly layouts. Full configuration details are provided in the Appendix \ref{app:training_details}.
\iffalse
\begin{wraptable}{r}{0.50\linewidth}
%\vspace{-12pt}
\centering
\definecolor{deeporange}{RGB}{253,208,162}
\setlength{\tabcolsep}{3.5pt}
\resizebox{0.95\linewidth}{!}{%
\begin{tabular}{l c c}
\toprule
 & \textbf{\taskone\ 3V} & \textbf{\taskthree\ Img} \\
 & \textbf{\nutsscen\ (195)} & \textbf{\cncscen\ (254)} \\
\midrule
Qwen2.5-VL-3B & 28.2 & 24.0 \\
\cellcolor{deeporange}\textbf{+\,SFT} & \cellcolor{deeporange}\textbf{53.8} & \cellcolor{deeporange}\textbf{30.5} \\
\midrule
$\Delta$ & \textbf{+25.6} & \textbf{+6.5} \\
\bottomrule
\end{tabular}}
\caption{Domain-specific SFT on held-out scenarios (zero-shot acc.\,\%).}
\label{tab:training_results}
%\vspace{-20pt}
\end{wraptable}
\fi

\begin{wrapfigure}{r}{0.45\linewidth}
%\vspace{-20pt}
\centering
\includegraphics[width=0.95\linewidth]{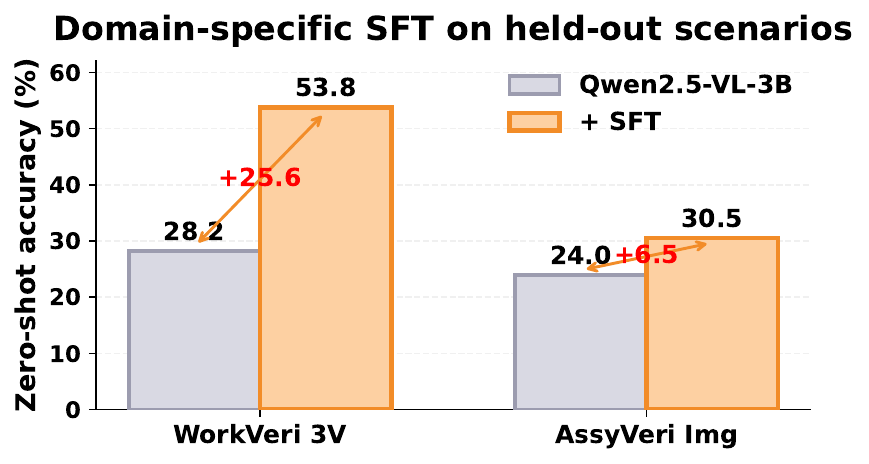}
%\vspace{-12pt}
\caption{Domain-specific SFT on held-out scenarios (zero-shot acc.\,\%).}
\label{fig:training_results}
%\vspace{-20pt}
\end{wrapfigure}

As shown in Figure~\ref{fig:training_results}, SFT yields a \textbf{90.8\%} improvement on \taskone\ three-view, bringing the 3B model on par with Qwen3-VL-235B (54.4\%), a model 78$\times$ larger. On \taskthree\ image, SFT achieves a 27.1\% relative gain, surpassing all reference models except Gemini-3-Flash and GPT-5.2. Because these gains are measured on product categories absent from training, they confirm that \ours\ annotations encode \emph{transferable} manufacturing knowledge. This establishes our dataset as not merely a static benchmark but an \textbf{actionable resource}: even modest amounts of domain-specific training data enable compact models to approach top performance on out-of-distribution manufacturing tasks.

% \subsection{Results of Training }\label{subsc: results of Training }
%\vspace{-5pt}

\section{Conclusion}

We present \ours, a fine-grained multimodal benchmark of real-world 2D images and 3D point clouds covering three manufacturing tasks%: Workpiece Verification, Structural Surface Inspection, and Assembly Verification
. Evaluating 18 state-of-the-art MLLMs, we find that current models handle macroscopic part recognition but fall short on fine-grained-level reasoning and microscopic surface analysis. Insights of evaluation and further bottleneck analysis reveals that visual grounding is not the primary limiting factor; rather, insufficient manufacturing domain knowledge and morphology understanding are the key gaps. Beyond evaluation, we show that \ours\ serves as an actionable training resource: domain-specific fine-tuning on our annotations enables a compact 3B-parameter model to approach frontier-scale performance on held-out scenarios. These findings establish \ours\ as both a rigorous evaluation baseline and a practical starting point for closing the domain knowledge gap in manufacturing intelligence.
 
\bibliographystyle{plain}
\bibliography{main}

%%%%%%%%%%%%%%%%%%%%%%%%%%%%%%%%%%%%%%%%%%%%%%%%%%%%%%%%%%%%
\newpage
\appendix
\begin{center}
	\LARGE \bf {Appendix}
\end{center}

\etocdepthtag.toc{mtappendix}
\etocsettagdepth{mtchapter}{none}
\etocsettagdepth{mtappendix}{subsubsection}
\tableofcontents
\clearpage
%\section{Technical appendices and supplementary material} Technical appendices with additional results, figures, graphs, and proofs may be submitted with the paper submission before the full submission deadline (see above). You can upload a ZIP file for videos or code, but do not upload a separate PDF file for the appendix. There is no page limit for the technical appendices. Note: Think of the appendix as ``optional reading'' for reviewers. The paper must be able to stand alone without the appendix; for example, adding critical experiments that support the main claims to an appendix is inappropriate. 

%%%%%%%%%%%%%%%%%%%%%%%%%%%%%%%%%%%%%%%%%%%%%%%%%%%%%%%%%%%%

\section{Extended Related Works, Main Results and Task Description}\label{app:appendix}
% You may include additional sections here.

\subsection{Related Works}\label{app: relatedworks}

\subsubsection{ CV models in Manufacturing Scenario.}\label{app:cv_models}

Computer Vision (CV) has widely served as the "eyes" of modern manufacturing automation, evolving from traditional image processing to deep learning-based approaches. Existing literature in this domain can be broadly categorized into three primary tasks: surface defect detection, geometric measurement, and robotic guidance.

Surface anomaly detection represents the most extensively studied subfield. Early works used supervised Convolutional Neural Networks (CNNs) to classify defects using large-scale annotated datasets \cite{he2016deep, ren2015faster}. To mitigate the reliance on defective samples, recent trends have shifted towards unsupervised paradigms, such as reconstruction-based methods (e.g., AutoEncoders, GANs) \cite{bergmann2019mvtec} and feature-embedding approaches \cite{roth2022towards}, which identify anomalies by measuring deviations from a normal distribution. Recently, to leverage geometric information for more robust detection, researchers have increasingly focused on 3D-aware approaches. \textit{G2SF} \cite{tao2025g2sf} is a multimodal industrial anomaly detection framework that utilizes geometry-guided score fusion to effectively combine 2D appearance and 3D depth information. Furthermore, to address the challenges of unstructured 3D data, \textit{PointSGRADE} \cite{tao2025pointsgrade} employs graph representations and sparse learning to detect anomalies directly from 3D point clouds.

CV models also play a critical role in geometric metrology and robotic manipulation. Key applications include 6D pose estimation for component assembly \cite{xiang2017posecnn} and precise dimensional measurement for quality control. These models provide spatial coordinates and geometric parameters, enabling automated systems to execute pre-defined mechanical tasks with high precision.

However, despite their perceptual precision, traditional CV models are fundamentally limited by their inability to reason and execute autonomous control. They typically operate as closed-set systems, mapping pixel inputs to fixed output classes without understanding the underlying semantics or physical causality. Consequently, they struggle to interpret complex, high-level instructions or adapt to unseen manufacturing scenarios that require logical deduction rather than mere pattern matching. This limitation underscores the urgent need to introduce MLLMs equipped with cognitive reasoning capabilities.

\subsubsection{MLLMs in Manufacturing Scenario.}\label{app:mllms_manu}

Recent advances in manufacturing anomaly detection leverage MLLMs to shift from simple perception to complex reasoning. LogiCode\cite{zhang2024logicode} prompts LLMs to generate executable Python code for verifying logical constraints, though it relies on manual annotations. LogicQA\cite{kwon2025logicqa} takes a more scalable, annotation-free approach, using VLMs to generate question-based checklists from standard samples for zero- or few-shot detection. EMIT\cite{guan2025emit} proposes Difficulty-Aware GRPO to align MLLMs with challenging anomaly detection tasks via response resampling and advantage reweighting. Despite their innovations, these works operate within existing datasets and predefined tasks, failing to explore novel formulations that reflect real-world factory complexity.

% \subsection{Detailed Ablation Results}\label{subsec:ablation_appendix}
% Full ablation analysis is in sections/ablation.tex.
% To include the full version here, uncomment the line below and resolve any duplicate labels
% (the main text Section~\ref{subsec:ablation} contains a condensed version with the grounding table).
% \input{sections/ablation}

\subsection{Extended Main Results}\label{app: extenresults}

\begin{table*}[htbp]
\centering
\definecolor{deepblue}{RGB}{198,219,239}
\definecolor{deeporange}{RGB}{253,208,162}
\definecolor{lightblue}{RGB}{230,218,240}
\setlength{\tabcolsep}{3.5pt}
\resizebox{\textwidth}{!}{
\begin{tabular}{lllc|cccccccccc|cccccccc}
\toprule
\multirow{2}{*}{\textbf{Task}} & \multirow{2}{*}{\textbf{Case}} & \multirow{2}{*}{\textbf{Mod.}} & \multirow{2}{*}{\textbf{Setting}} & \multicolumn{10}{c|}{\textbf{Open-source Models}} & \multicolumn{8}{c}{\textbf{Closed-source Models}} \\
\cmidrule(lr){5-14}
\cmidrule(lr){15-22}
 &  &  &  & \rot{Gemma-3-27B} & \rot{InternVL3-78B} & \rot{Llama-4-MAV} & \rot{Mistral-3-14B} & \rot{Mistral-3-8B} & \rot{Mistral-3-Large} & \rot{Qwen3-VL-235B} & \rot{Qwen3-VL-8B} & \rot{Kimi-K2.5} & \rot{GLM-4.6V} & \rot{Claude-4.5-Opus} & \rot{Gemini-2.5-Flash} & \rot{Gemini-3-Flash} & \rot{GPT-5.2} & \rot{GPT-5} & \rot{GPT-5 Mini} & \rot{O3} & \rot{Seed-1.6} \\
\midrule
\multirow{12}{*}{\taskone} & \multirow{6}{*}{Model No} & \multirow{3}{*}{Image} & Zero-Shot & 23.18 & 27.04 & 24.55 & 25.78 & 20.09 & 24.46 & \cellcolor{deepblue}\underline{36.91} & 21.12 & \cellcolor{lightblue}{30.41} & \cellcolor{deeporange}\textbf{41.84} & 30.04 & 33.48 & 46.78 & 48.03 & \cellcolor{lightblue}{51.93} & \cellcolor{deepblue}\underline{53.88} & \cellcolor{deeporange}\textbf{55.70} & 38.96 \\
 &  &  & Ref-Cond & 25.97 & \cellcolor{lightblue}{30.47} & 21.33 & 18.61 & 23.81 & 19.40 & \cellcolor{deepblue}\underline{33.64} & 25.54 & \cellcolor{deeporange}\textbf{37.04} & 27.90 & 34.33 & 30.04 & \cellcolor{deeporange}\textbf{55.79} & 32.86 & 40.52 & \cellcolor{deepblue}\underline{49.36} & \cellcolor{lightblue}{45.66} & 27.07 \\
 &  &  & ICD & 18.03 & \cellcolor{deeporange}\textbf{41.56} & 31.28 & 24.24 & 26.84 & 24.03 & \cellcolor{lightblue}{35.34} & 22.41 & \cellcolor{deepblue}\underline{38.76} & 29.00 & 30.90 & 31.33 & \cellcolor{lightblue}{65.67} & 60.87 & \cellcolor{deeporange}\textbf{71.43} & 61.90 & \cellcolor{deepblue}\underline{66.51} & 44.21 \\
 &  & \multirow{3}{*}{3V} & Zero-Shot & 22.87 & 20.18 & 29.15 & 27.35 & 26.24 & 27.91 & \cellcolor{deeporange}\textbf{37.79} & 33.33 & \cellcolor{lightblue}{35.00} & \cellcolor{deepblue}\underline{37.50} & 30.49 & 39.01 & \cellcolor{deeporange}\textbf{51.12} & \cellcolor{lightblue}{43.50} & 40.09 & 31.39 & \cellcolor{deepblue}\underline{44.09} & 40.81 \\
 &  &  & Ref-Cond & 20.18 & 21.17 & 19.28 & \cellcolor{deeporange}\textbf{30.94} & 19.73 & 20.36 & \cellcolor{deepblue}\underline{30.05} & 25.69 & 22.64 & \cellcolor{lightblue}{28.96} & \cellcolor{lightblue}{29.60} & \cellcolor{deepblue}\underline{31.39} & \cellcolor{deeporange}\textbf{49.10} & 22.07 & 25.91 & 27.35 & 26.82 & 22.87 \\
 &  &  & ICD & 25.11 & 27.03 & \cellcolor{deepblue}\underline{33.63} & 28.38 & 25.45 & 25.56 & \cellcolor{lightblue}{32.49} & 24.22 & \cellcolor{deeporange}\textbf{36.30} & 32.43 & 41.26 & 38.12 & \cellcolor{deeporange}\textbf{49.33} & 43.95 & \cellcolor{lightblue}{45.50} & 30.04 & \cellcolor{deepblue}\underline{46.51} & 36.65 \\
 & \multirow{6}{*}{Workpiece} & \multirow{3}{*}{Image} & Zero-Shot & 28.90 & 38.53 & 54.07 & 41.90 & 41.98 & 27.19 & \cellcolor{deepblue}\underline{93.12} & 50.69 & \cellcolor{deeporange}\textbf{100.00} & \cellcolor{lightblue}{57.03} & 90.83 & 79.72 & \cellcolor{deeporange}\textbf{99.54} & \cellcolor{lightblue}{99.05} & \cellcolor{deepblue}\underline{99.08} & 94.95 & 97.70 & 96.79 \\
 &  &  & Ref-Cond & 35.32 & \cellcolor{lightblue}{78.90} & 61.54 & 21.96 & 24.77 & 32.09 & \cellcolor{deepblue}\underline{85.13} & 22.43 & \cellcolor{deeporange}\textbf{95.59} & 74.31 & 79.72 & 74.77 & \cellcolor{deeporange}\textbf{98.17} & 78.74 & 89.45 & \cellcolor{deepblue}\underline{97.71} & \cellcolor{lightblue}{92.65} & 73.39 \\
 &  &  & ICD & 52.29 & \cellcolor{lightblue}{90.37} & 75.36 & 30.15 & 37.61 & 38.89 & \cellcolor{deepblue}\underline{99.54} & 28.37 & \cellcolor{deeporange}\textbf{100.00} & 72.35 & 93.12 & 79.36 & \cellcolor{deeporange}\textbf{100.00} & \cellcolor{deepblue}\underline{100.00} & \cellcolor{lightblue}{100.00} & 93.09 & 100.00 & 98.17 \\
 &  & \multirow{3}{*}{3V} & Zero-Shot & 31.50 & 42.49 & 42.28 & 36.63 & 34.80 & 40.30 & \cellcolor{deeporange}\textbf{64.07} & 49.07 & \cellcolor{deepblue}\underline{60.71} & \cellcolor{lightblue}{55.39} & 46.32 & 69.96 & \cellcolor{deeporange}\textbf{84.62} & \cellcolor{lightblue}{75.82} & \cellcolor{deepblue}\underline{78.23} & 43.38 & 69.60 & 65.57 \\
 &  &  & Ref-Cond & 26.37 & 27.94 & 21.25 & 20.15 & 17.04 & 21.40 & \cellcolor{lightblue}{40.16} & 28.41 & \cellcolor{deepblue}\underline{45.78} & \cellcolor{deeporange}\textbf{56.88} & 26.47 & \cellcolor{deepblue}\underline{53.48} & \cellcolor{deeporange}\textbf{78.39} & 33.46 & 33.33 & \cellcolor{lightblue}{39.34} & 31.82 & 30.15 \\
 &  &  & ICD & 31.87 & 32.72 & 37.36 & 28.68 & 31.62 & \cellcolor{lightblue}{47.41} & 42.06 & 28.04 & \cellcolor{deeporange}\textbf{75.11} & \cellcolor{deepblue}\underline{55.22} & 66.30 & 66.79 & \cellcolor{deeporange}\textbf{82.05} & \cellcolor{deepblue}\underline{76.19} & 65.81 & 31.25 & \cellcolor{lightblue}{66.91} & 54.41 \\
\midrule
\multirow{12}{*}{\taskthree} & \multirow{6}{*}{Model No} & \multirow{3}{*}{Image} & Zero-Shot & 19.27 & 18.94 & 20.82 & 17.86 & 16.76 & \cellcolor{deepblue}\underline{21.39} & 15.29 & 17.52 & \cellcolor{deeporange}\textbf{27.12} & \cellcolor{lightblue}{21.34} & \cellcolor{deepblue}\underline{33.70} & 25.87 & \cellcolor{deeporange}\textbf{40.44} & 27.94 & \cellcolor{lightblue}{29.87} & 21.17 & 26.52 & 19.05 \\
 &  &  & Ref-Cond & 23.50 & 24.72 & 21.71 & \cellcolor{deepblue}\underline{25.14} & 16.39 & 16.27 & 24.71 & 18.83 & \cellcolor{deeporange}\textbf{43.56} & \cellcolor{lightblue}{24.73} & \cellcolor{lightblue}{41.71} & 21.02 & \cellcolor{deeporange}\textbf{61.86} & 38.76 & 36.13 & 37.11 & \cellcolor{deepblue}\underline{46.69} & 27.74 \\
 &  &  & ICD & 23.72 & 22.26 & 30.05 & 19.17 & 20.33 & 19.12 & \cellcolor{lightblue}{31.55} & 22.53 & \cellcolor{deeporange}\textbf{39.28} & \cellcolor{deepblue}\underline{35.15} & \cellcolor{lightblue}{47.63} & 32.23 & \cellcolor{deeporange}\textbf{64.30} & \cellcolor{deepblue}\underline{48.90} & 44.89 & 39.85 & 46.44 & 27.88 \\
 &  & \multirow{3}{*}{3V} & Zero-Shot & 16.78 & \cellcolor{deepblue}\underline{26.17} & 19.46 & 15.44 & 14.77 & 19.63 & \cellcolor{lightblue}{23.57} & 19.46 & \cellcolor{deeporange}\textbf{100.00} & 22.07 & 13.42 & 20.13 & \cellcolor{deeporange}\textbf{32.89} & 22.97 & \cellcolor{deepblue}\underline{25.68} & 21.48 & 20.00 & \cellcolor{lightblue}{23.13} \\
 &  &  & Ref-Cond & \cellcolor{deepblue}\underline{26.85} & 6.04 & 21.48 & 0.00 & 0.67 & 19.83 & 21.05 & \cellcolor{lightblue}{25.52} & \cellcolor{deeporange}\textbf{42.86} & 25.17 & \cellcolor{lightblue}{26.17} & 26.17 & \cellcolor{deeporange}\textbf{30.87} & \cellcolor{deepblue}\underline{28.86} & 18.12 & 21.62 & 24.26 & 14.29 \\
 &  &  & ICD & \cellcolor{deeporange}\textbf{26.85} & 18.79 & 22.82 & 7.43 & 20.81 & 18.25 & \cellcolor{lightblue}{24.06} & \cellcolor{deepblue}\underline{26.35} & 12.77 & 21.48 & \cellcolor{deepblue}\underline{30.87} & \cellcolor{deeporange}\textbf{31.54} & \cellcolor{lightblue}{30.87} & 27.89 & 26.35 & 23.49 & 20.69 & 22.82 \\
 & \multirow{6}{*}{Workpiece} & \multirow{3}{*}{Image} & Zero-Shot & 44.92 & 58.44 & \cellcolor{lightblue}{65.28} & 52.70 & 52.60 & 42.86 & \cellcolor{deepblue}\underline{76.08} & 57.70 & \cellcolor{deeporange}\textbf{91.84} & 58.37 & 85.02 & 63.96 & \cellcolor{deepblue}\underline{89.61} & 84.67 & \cellcolor{lightblue}{86.04} & 83.77 & \cellcolor{deeporange}\textbf{89.90} & 76.87 \\
 &  &  & Ref-Cond & 46.41 & 57.00 & 49.80 & 44.59 & 45.45 & 37.01 & \cellcolor{deepblue}\underline{69.79} & 46.03 & \cellcolor{deeporange}\textbf{100.00} & \cellcolor{lightblue}{59.61} & 82.47 & 52.44 & \cellcolor{deepblue}\underline{86.27} & \cellcolor{deeporange}\textbf{89.93} & 74.03 & 76.97 & \cellcolor{lightblue}{84.04} & 69.08 \\
 &  &  & ICD & 49.35 & \cellcolor{lightblue}{78.50} & 73.98 & 47.52 & 49.01 & 45.60 & \cellcolor{deepblue}\underline{84.48} & 44.26 & \cellcolor{deeporange}\textbf{91.89} & 63.52 & \cellcolor{lightblue}{90.23} & 72.96 & 84.36 & \cellcolor{deepblue}\underline{91.09} & 90.07 & 74.51 & \cellcolor{deeporange}\textbf{92.12} & 84.92 \\
 &  & \multirow{3}{*}{3V} & Zero-Shot & 36.88 & 41.88 & 55.62 & 40.00 & 33.75 & 40.27 & \cellcolor{deepblue}\underline{58.75} & \cellcolor{lightblue}{57.50} & \cellcolor{deeporange}\textbf{77.78} & 55.70 & 68.75 & 68.12 & 60.62 & \cellcolor{deeporange}\textbf{83.12} & \cellcolor{deepblue}\underline{80.00} & 75.62 & \cellcolor{lightblue}{79.87} & 58.75 \\
 &  &  & Ref-Cond & 36.88 & 36.88 & 38.12 & 36.88 & 35.00 & 36.88 & 39.61 & \cellcolor{lightblue}{41.40} & \cellcolor{deeporange}\textbf{60.71} & \cellcolor{deepblue}\underline{48.72} & \cellcolor{deepblue}\underline{73.75} & 53.12 & \cellcolor{lightblue}{60.62} & \cellcolor{deeporange}\textbf{83.12} & 42.50 & 36.88 & 47.37 & 38.12 \\
 &  &  & ICD & 36.88 & 38.12 & 51.25 & 43.40 & 37.74 & 34.59 & \cellcolor{deepblue}\underline{60.40} & 50.62 & \cellcolor{deeporange}\textbf{77.50} & \cellcolor{lightblue}{53.12} & 59.12 & 55.62 & 63.12 & \cellcolor{deepblue}\underline{78.12} & \cellcolor{lightblue}{75.62} & 65.62 & \cellcolor{deeporange}\textbf{81.25} & 45.00 \\
\bottomrule
\end{tabular}}
\caption{Performance of different cases for  \taskone \ and \taskthree, different workpieces, and different model numbers.\colorbox{deeporange}{\textbf{Bold orange}} / \colorbox{deepblue}{\underline{underlined blue}} / \colorbox{lightblue}{light purple} indicate best / second / third within each model category (open- or closed-source) per row.}
\label{tab:case-modality-method}
\end{table*}

In this section, we present more detailed results compared to the main text. Specifically, Table \ref{tab:case-modality-method} details the performance of \taskone \ and \taskthree \ across different workpieces and model numbers, while Table \ref{tab:task2-native-cases} breaks down the performance across various anomaly cases. Overall, these findings are consistent with the results discussed in the main text: \textbf{Current MLLMs face substantial hurdles, where overall performance remains far from acceptable.}
Results across the three tasks reveal that the insufficient capability to internalize and reason about complex manufacturing standards makes this domain an arduous challenge for future MLLM development. Notably, most open-source models perform near the random baseline. While closed-source models such as Gemini-3-Flash and GPT-5 demonstrate superior capabilities, achieving state-of-the-art results in \taskone \ and \taskthree, even the most advanced models struggle to surpass 50\% accuracy in the highly demanding \tasktwo. This disparity underscores that, although closed-source models exhibit stronger general-purpose capabilities, performance in highly specialized manufacturing scenarios remains a significant challenge.

However, analyzing the results for \taskone \ and \taskthree \ across different workpieces and model numbers in Table \ref{tab:case-modality-method}, we observe a distinct pattern across the three settings. Specifically, for the three-view (3V) modality, both model-number-level and workpiece-level tasks expose a significant bottleneck in 3D spatial understanding. Interestingly, under the 3V setting, both Ref-Cond and ICD exhibit a performance degradation compared to the zero-shot baseline. However, this decrease is significantly more pronounced in workpiece-level tasks than in model-number-level tasks. This observation further corroborates Conclusion C in Section 4.3. The root cause lies in the fact that the similarity among workpieces is substantially lower than that among different model numbers. Consequently, the visual features of workpieces exhibit much higher variance under the inherent spatial and angular shifts of 3V images. This extreme variability severely hinders MLLMs from aligning spatial semantics when integrating reference examples, ultimately leading to a steeper performance drop for Ref-Cond and ICD in workpiece-level tasks.

More granular results for \tasktwo are presented in Table \ref{tab:task2-native-cases}. Although the task comprehension requirement for \tasktwo \ is relatively low, and performance generally improves as the number of references increases across all three settings, the anomalies of DEFORMATION and DENT do not conform to this trend. This suggests that even for seemingly simple tasks, varying domain knowledge requirements cause MLLMs to interpret the task differently, leading to distinct performance trajectories across different anomaly types.

\begin{table*}[htbp]
\centering
\definecolor{deepblue}{RGB}{198,219,239}
\definecolor{deeporange}{RGB}{253,208,162}
\definecolor{lightblue}{RGB}{230,218,240}
\setlength{\tabcolsep}{3.2pt}

\resizebox{\textwidth}{!}{%
\begin{tabular}{ll|cccccccccc|cccccccc}
\toprule
\multirow{2}{*}{\textbf{Case}} & \multirow{2}{*}{\textbf{Method}} & \multicolumn{10}{c|}{\textbf{Open-Source}} & \multicolumn{8}{c}{\textbf{Closed-Source}} \\
\cmidrule(lr){3-12}
\cmidrule(lr){13-20}
 &  & \rot{Gemma-3-27B} & \rot{InternVL3-78B} & \rot{Llama-4-MAV} & \rot{Mistral-3-14B} & \rot{Mistral-3-8B} & \rot{Mistral-3-Large} & \rot{Qwen3-VL-235B} & \rot{Qwen3-VL-8B} & \rot{Kimi-K2.5} & \rot{GLM-4.6V} & \rot{Claude-4.5-Opus} & \rot{Gemini-2.5-Flash} & \rot{Gemini-3-Flash} & \rot{GPT-5.2} & \rot{GPT-5} & \rot{GPT-5 Mini} & \rot{O3} & \rot{Seed-1.6} \\
\midrule
\multirow{3}{*}{CRACK} & Zero-Shot & 31.58 & 6.32 & 24.21 & \cellcolor{deeporange}\textbf{56.84} & \cellcolor{lightblue}{44.21} & 4.21 & 4.35 & 22.11 & 21.05 & \cellcolor{deepblue}\underline{53.68} & \cellcolor{deepblue}\underline{16.84} & 1.05 & \cellcolor{lightblue}{7.37} & 5.26 & 0.00 & 5.26 & 5.26 & \cellcolor{deeporange}\textbf{42.55} \\
 & Ref-Cond & 15.79 & 4.21 & 13.68 & \cellcolor{deeporange}\textbf{60.64} & \cellcolor{deepblue}\underline{60.00} & 9.47 & 8.14 & 27.37 & 14.74 & \cellcolor{lightblue}{44.21} & \cellcolor{lightblue}{11.58} & 1.05 & \cellcolor{deepblue}\underline{34.74} & 5.26 & 3.16 & 2.11 & 7.37 & \cellcolor{deeporange}\textbf{46.32} \\
 & ICD & 37.89 & 9.47 & 44.21 & \cellcolor{deepblue}\underline{59.78} & \cellcolor{deeporange}\textbf{69.47} & 10.64 & 21.79 & 41.49 & 40.00 & \cellcolor{lightblue}{51.58} & \cellcolor{deepblue}\underline{57.45} & 47.37 & \cellcolor{deeporange}\textbf{62.11} & 20.00 & 10.53 & 14.74 & 25.26 & \cellcolor{lightblue}{56.38} \\
\midrule
\multirow{3}{*}{CUT} & Zero-Shot & 3.37 & 1.14 & 5.62 & \cellcolor{deepblue}\underline{15.73} & 4.49 & 0.00 & \cellcolor{deeporange}\textbf{25.58} & \cellcolor{lightblue}{12.50} & 12.36 & 2.25 & 5.62 & 4.49 & 7.87 & \cellcolor{deeporange}\textbf{30.68} & \cellcolor{lightblue}{22.47} & \cellcolor{deepblue}\underline{28.09} & 14.61 & 4.49 \\
 & Ref-Cond & 1.12 & 4.49 & 5.62 & \cellcolor{deepblue}\underline{16.85} & 2.25 & 1.12 & \cellcolor{deeporange}\textbf{19.05} & \cellcolor{lightblue}{11.24} & 11.24 & 2.25 & 6.74 & 4.49 & 17.98 & \cellcolor{deepblue}\underline{32.58} & \cellcolor{lightblue}{29.21} & \cellcolor{deeporange}\textbf{47.19} & 20.45 & 7.87 \\
 & ICD & 16.85 & 4.49 & 12.36 & \cellcolor{deepblue}\underline{20.22} & 10.11 & \cellcolor{lightblue}{17.98} & 9.09 & 7.95 & \cellcolor{deeporange}\textbf{29.21} & 5.68 & 21.35 & 13.48 & 34.83 & \cellcolor{lightblue}{39.33} & \cellcolor{deepblue}\underline{41.57} & \cellcolor{deeporange}\textbf{64.04} & 34.83 & 24.72 \\
\midrule
\multirow{3}{*}{DEFORMATION} & Zero-Shot & 81.82 & \cellcolor{deepblue}\underline{93.59} & 8.97 & \cellcolor{lightblue}{83.33} & 83.33 & \cellcolor{deeporange}\textbf{96.10} & 72.37 & 64.10 & 29.49 & 70.51 & 19.23 & \cellcolor{deeporange}\textbf{91.03} & \cellcolor{deepblue}\underline{88.46} & 53.85 & 57.69 & 60.26 & 67.95 & \cellcolor{lightblue}{74.36} \\
 & Ref-Cond & 56.41 & \cellcolor{deeporange}\textbf{94.81} & 62.34 & 69.23 & 73.08 & \cellcolor{deepblue}\underline{78.21} & 72.22 & 57.69 & 53.85 & \cellcolor{lightblue}{78.21} & 20.51 & \cellcolor{deeporange}\textbf{97.44} & \cellcolor{lightblue}{71.79} & 62.82 & 50.00 & 55.13 & 55.13 & \cellcolor{deepblue}\underline{84.62} \\
 & ICD & 33.33 & \cellcolor{deeporange}\textbf{87.18} & 30.77 & 28.21 & 32.05 & 48.72 & 49.28 & \cellcolor{deepblue}\underline{57.14} & 37.66 & \cellcolor{lightblue}{52.56} & 12.82 & \cellcolor{deeporange}\textbf{62.82} & 32.05 & \cellcolor{lightblue}{51.28} & 32.05 & 12.82 & 25.64 & \cellcolor{deepblue}\underline{54.55} \\
\midrule
\multirow{3}{*}{DENT} & Zero-Shot & 7.06 & \cellcolor{deepblue}\underline{11.76} & 1.18 & 0.00 & 0.00 & \cellcolor{deeporange}\textbf{17.65} & 4.94 & \cellcolor{lightblue}{10.59} & 1.18 & 1.18 & 3.53 & \cellcolor{lightblue}{5.88} & 4.71 & 4.71 & \cellcolor{deeporange}\textbf{20.48} & 3.53 & \cellcolor{deepblue}\underline{14.12} & 1.18 \\
 & Ref-Cond & \cellcolor{deepblue}\underline{22.62} & 7.14 & 7.06 & 4.71 & 5.88 & \cellcolor{deeporange}\textbf{25.88} & 4.00 & \cellcolor{lightblue}{16.47} & 14.12 & 1.18 & 3.53 & 4.71 & \cellcolor{lightblue}{12.94} & 8.24 & \cellcolor{deeporange}\textbf{18.82} & 5.88 & \cellcolor{deepblue}\underline{18.82} & 1.18 \\
 & ICD & 11.76 & 4.71 & \cellcolor{deepblue}\underline{28.24} & 5.88 & 7.06 & \cellcolor{deeporange}\textbf{38.82} & 17.39 & \cellcolor{lightblue}{26.19} & 12.94 & 9.41 & 22.35 & \cellcolor{deepblue}\underline{29.41} & \cellcolor{deeporange}\textbf{36.47} & 10.59 & 25.88 & 2.35 & \cellcolor{lightblue}{29.41} & 11.76 \\
\midrule
\multirow{3}{*}{GOOD} & Zero-Shot & 0.00 & 0.00 & \cellcolor{deeporange}\textbf{73.98} & 0.00 & \cellcolor{lightblue}{2.44} & 0.00 & 1.68 & 0.00 & \cellcolor{deepblue}\underline{5.69} & 0.82 & 1.63 & 0.00 & 0.00 & 0.00 & \cellcolor{deeporange}\textbf{17.07} & 0.00 & \cellcolor{deepblue}\underline{13.11} & \cellcolor{lightblue}{2.44} \\
 & Ref-Cond & \cellcolor{deepblue}\underline{26.83} & \cellcolor{lightblue}{11.48} & \cellcolor{deeporange}\textbf{33.33} & 0.00 & 4.92 & 0.00 & 1.82 & 4.07 & 0.81 & 4.88 & 0.00 & 31.71 & 18.70 & 10.57 & \cellcolor{deepblue}\underline{68.29} & \cellcolor{lightblue}{52.85} & \cellcolor{deeporange}\textbf{69.92} & 42.28 \\
 & ICD & 56.10 & 29.27 & \cellcolor{deeporange}\textbf{67.48} & 44.72 & \cellcolor{deepblue}\underline{62.60} & 22.76 & 55.88 & 6.50 & 30.08 & \cellcolor{lightblue}{62.60} & \cellcolor{deeporange}\textbf{86.18} & 39.02 & 61.48 & 37.40 & 69.92 & \cellcolor{lightblue}{71.31} & \cellcolor{deepblue}\underline{71.54} & 57.72 \\
\bottomrule
\end{tabular}}
\caption{Performance of 4 anomaly cases for \tasktwo, different workpieces, and different model numbers.\colorbox{deeporange}{\textbf{Bold orange}} / \colorbox{deepblue}{\underline{underlined blue}} / \colorbox{lightblue}{light purple} indicate best / second / third within each model category (open- or closed-source) per row.}
\label{tab:task2-native-cases}

\end{table*}

\subsection{Task description}\label{app: taskdescrip}

In \ours, we primarily designed three tasks related to manufacturing scenarios to evaluate the key capabilities of MLLMs in typical manufacturing contexts such as \textbf{material sorting}, \textbf{quality inspection}, and \textbf{assembly recognition}. This Section systematically introduces the detailed design framework of these three tasks, including task design, construction of testing and evaluation data, evaluation settings, and the corresponding data collection process.

\subsubsection{Raw Data Collection.}\label{app: data collect}

In this work, to facilitate the evaluation of MLLMs and provide data for our tasks, we collected raw data from authentic manufacturing workpieces. The data acquisition setup was established using a platform equipped with a precision rotary table and a custom-designed fixture. The data collection process was divided into two distinct phases: point cloud acquisition and image data collection.

\textbf{Point Cloud Data Acquisition:} A handheld 3D scanner was utilized for data collection with a scanning precision configured to 0.02 mm. Before scanning, positioning markers were affixed to the platform to facilitate the establishment of a local coordinate system via marker recognition. To mitigate the surface reflectivity inherent in metallic workpieces, a 3D scanning developer spray was applied as a pretreatment. Once the workpiece was securely positioned, the scanning procedure was initiated. Subsequently, proprietary software was employed to perform post-processing operations, including registration, denoising, mesh reconstruction, and texture mapping, yielding high-fidelity 3D model reconstructions.

\textbf{Image Data Acquisition:}
Specific experimental scenarios were pre-designed for image data collection. Metal workpieces of corresponding models were randomly selected from the sample library and placed in random orientations, while maintaining consistency in shooting angles and distances. Images were captured using a 50-megapixel sensor with a Leica Vario-Summilux optical lens system to ensure high resolution.

Then, the raw dataset is categorized into \textbf{Point Cloud Subset} and \textbf{Image Subset}, capturing both geometric and visual aspects of manufacturing components.

\textbf{Point Cloud Subset:} This subset covers 14 component categories and 90 model numbers. Each sample is represented as a 3D point cloud, providing high-fidelity geometric structure information. Through data preprocessing and manual annotation, this subset enables a variety of applications, including \taskone, \tasktwo, and \ \taskthree. The detailed collected 3D point cloud data are summarized in Table \ref{table:raw_point}.

\textbf{Image Subset:} This subset comprises approximately 3000 images collected from four manufacturing scenarios, including expansion screw assemblies and positioners. Each scenario contains normal samples and abnormal samples. This subset primarily supports \taskone \ and \taskthree \ under real-world manufacturing conditions. The detailed collected image data are summarized in Table \ref{table:raw_image_scenarios}.

While these two subsets provide the comprehensive raw material for \taskone, \tasktwo, and \taskthree, raw data requires specific curation before it can be used for standardized evaluation. Therefore, the detailed data preparation pipelines and specific definitions for each task are presented in the following sections.

\textbf{ Data Processing:}  It is important to note that the raw data—comprising both 2D images and 3D point clouds—could not be directly utilized for \ours. Therefore, data preprocessing pipelines were designed for each modality. For 2D image data, we first employed a Python-based algorithm to extract the precise contour and coordinate information of each artifact. Subsequently, through manual calibration, we mapped these spatial coordinates to the corresponding manufacturing model information to establish ground-truth labels. For 3D point cloud data, distinct strategies were applied based on the task requirements:
 
For \taskone and \taskthree: Since the collected data consisted of single normal artifacts, we utilized CloudCompare to synthesize batch samples. Specifically, we stitched 4-5 individual point clouds together, applying random orientations and relative positions within a constrained range. Corresponding labels were automatically generated based on the artifact models and workpiece information during this assembly process. For \tasktwo: We focused on synthetic defect generation. Four typical manufacturing defects were simulated: Crack, Deformation, Dent, and Cut. For each type, we designed a tailored algorithm based on its distinctive morphological characteristics to produce a large set of initial shapes. Non-rigid deformation was then applied to enhance realism and variability. The proportion of defect points per sample was constrained between 5\% and 15\%. To address data scarcity, we augmented the dataset by applying 20 random rotations (uniformly sampled from $[0^\circ, 360^\circ)$) to each sample. In total, we constructed a dataset containing approximately 30,000 samples (including training data) for all three tasks. Furthermore, our preliminary tests revealed that feeding raw point cloud files directly into the LLM resulted in sub-optimal performance. Consequently, we adopted a multi-view projection strategy: all point clouds were rendered into 3V images (orthogonal projections) to serve as the actual input for MLLMs.

\textbf{Eval setting:} For all tasks, the evaluation settings are divided into three categories: \textbf{Zero-Shot}, \textbf{ Reference-Conditioned(Ref-Cond)}, and \textbf{ In-Context Demonstration(ICD)}.In the \textbf{Zero-Shot} setting, only the test image and the corresponding textual query are provided as input during evaluation. In the \textbf{Ref-Cond} setting, three correct normal cases are additionally included as reference examples for the MLLMs model. In the \textbf{In-Context Demonstration} setting, one more example that is similar to the test case, consisting of an image, its corresponding query, and the correct answer, is added on top of the Ref-Cond examples. The final test query is then provided based on these contextual demonstrations.  To further systematize the evaluation setting, we categorize potential error scenarios into two primary classes: \textit{Different workpiece} and \textit{Different Model Number}. The former refers to coarse-grained discrepancies, including workpiece mismatches, component absence, or other workpiece-level anomalies. The latter addresses fine-grained inconsistencies, specifically focusing on errors arising from distinct model variants despite the workpiece category being correct.

\begin{table}[t]
\centering
\small
\caption{Simple data summary of raw point cloud data. The dataset contains 14 workpieces.}
\label{table:raw_point}
\setlength{\tabcolsep}{8pt}
\renewcommand{\arraystretch}{1.2}
\begin{tabularx}{\linewidth}{>{\centering\arraybackslash}X>{\raggedleft\arraybackslash}m{3cm}>{\raggedleft\arraybackslash}m{3cm}}
\toprule
\textbf{Workpieces} & \textbf{Model Number} & \textbf{No. of Samples} \\
\midrule
Corner Bracket & 5 & 25 \\
Countersunk Screw & 2 & 20 \\
Cup Head Screw & 22 & 110 \\
Eye Bolt & 5 & 45 \\
Flat Washer & 9 & 45 \\
Hex Nut & 4 & 33 \\
Rivet Nut & 3 & 30 \\
Self-tapping Screw & 6 & 60 \\
Spring Washer & 9 & 50 \\
T Bolt Half thread Screw & 4 & 40 \\
T Nut & 4 & 20 \\
T Screw & 10 & 50 \\
Wing Nut & 3 & 27 \\
Wing Screw & 4 & 30 \\
\midrule
\textbf{Total} & \textbf{90} & \textbf{585} \\
\bottomrule
\end{tabularx}
\end{table}

\begin{table}[t]
\centering
\small
\renewcommand{\arraystretch}{1.15}
\caption{Simple data summary of raw image data (scenario-level).}
\label{table:raw_image_scenarios}
\begin{tabularx}{\linewidth}{@{}>{\raggedright\arraybackslash}p{2.0cm}>{\raggedright\arraybackslash}X>{\centering\arraybackslash}p{2.2cm}@{}}
\toprule
\textbf{Workpiece} & \textbf{Wrong types} & \textbf{No. of samples} \\
\midrule
\multirow[t]{9}{*}{\messcen} &  No Spring Washers & 56 \\
 & Hex Nut M14 & 52 \\
 & Flat Washer M14 & 51 \\
 & No Flat Washers & 50 \\
 & Two Flat Washers & 50 \\
 & Two Spring Washers & 49 \\
 & Spring Washer M8 & 48 \\
 & Cup Head Screw M12 40 & 46 \\
\addlinespace[2pt]
\multirow[t]{9}{*}{\pesscen} &  One missing plastic expansion anchor & 52 \\
 & One missing plastic expansion Self-tapping Screw & 51 \\
 & One extra plastic expansion Self-tapping Screw & 49 \\
 & Screws belonging to model 860 expansion anchors were found mixed in & 49 \\
 & Screws shorter 30 than the specified length were found mixed in & 49 \\
 & One extra plastic expansion anchor & 47 \\
 & Model 860 expansion screws were mistakenly included & 46 \\
 & Plastic anchors belonging to the expansion screw model 860 were mixed  & 44 \\
\addlinespace[2pt]
\multirow[t]{9}{*}{\cncscen} &  Missing one nut & 57 \\
 & Extra one nut & 56 \\
 & Missing triangular part & 56 \\
 & retainer block & 56 \\
 & Missing screw & 53 \\
 & Long screw & 51 \\
 & Short screw & 50 \\
 & Triangular part & 48 \\
\addlinespace[2pt]
\multirow[t]{9}{*}{\pcsscen} & 2-way pneumatic tube connectors (8 to 6), mixed with model (10 to 6) & 67 \\
 & Three 2-way pneumatic tube connectors (8mm to 6mm), mixed with T-type 3-way & 63 \\
 & 2-way pneumatic tube connectors (8 to 6), mixed with model (6 to 4) & 57 \\
 & Three 2-way pneumatic tube connectors (8mm to 6mm), mixed with Y-type 3-way & 57 \\
 & 2-way pneumatic tube connectors (8 to 6), mixed with model (12 to 8) & 56 \\
 & 2-way pneumatic tube connectors (8 to 6), mixed with model (8 to 4) & 56 \\
 & Three 2-way pneumatic tube connectors (8mm to 6mm), mixed with throttle valve & 55 \\
 & Three 2-way pneumatic tube connectors (8 to 6), mixed with external-thread elbow & 50 \\
\addlinespace[2pt]
\midrule
\textbf{Total} &  & \textbf{3115} \\
\bottomrule
\end{tabularx}
\end{table}

\begin{table*}[htbp]
\centering
\scriptsize

\setlength{\tabcolsep}{4pt}
\begin{tabular}{>{\centering\arraybackslash}p{1.5cm} >{\raggedright\arraybackslash}p{1.5cm} >{\raggedright\arraybackslash}p{1.5cm} >{\raggedright\arraybackslash}p{4.0cm} >{\centering\arraybackslash}p{1.0cm}}
\toprule
\textbf{Task} & \textbf{Scenario} & \textbf{Case Type} & \textbf{Error Type} & \textbf{Samples} \\
\midrule
\multirow{21}{*}{\taskone} & \multirow{8}{*}{\chsscen} & \multirow{4}{*}{Model No} & M16 longer length 100 & 42 \\
 &  &  & M18 longer length 100 & 35 \\
 &  &  & M12 shorter length 50 & 34 \\
 &  &  & M10 shorter length 50 & 32 \\
 &  & \multirow{4}{*}{Workpiece} & Self tapping Screw & 37 \\
 &  &  & T Bolt Half thread Screw & 44 \\
 &  &  & T Screw 45 M8 & 36 \\
 &  &  & Wing Screw & 41 \\
 & \multirow{5}{*}{\nutsscen} & \multirow{2}{*}{Model No} & Mixed with M16 & 75 \\
 &  &  & Mixed with M20 & 75 \\
 &  & \multirow{3}{*}{Workpiece} & Rivet Nut & 78 \\
 &  &  & T Nut & 81 \\
 &  &  & Wing Nut & 81 \\
 & \multirow{8}{*}{\pcsscen} & \multirow{4}{*}{Model No} &  Mixed with (10 to 6) & 67 \\
 &  &  & Mixed with (12 to 8) & 59 \\
 &  &  & Mixed with (6 to 4) & 59 \\
 &  &  & Mixed with (8 to 4) & 56 \\
 &  & \multirow{4}{*}{Workpiece} & Mixed with external-thread elbow & 50 \\
 &  &  & Mixed with T-type 3-way & 63 \\
 &  &  & Mixed with throttle valve & 55 \\
 &  &  & Mixed with Y-type 3-way & 58 \\
\midrule
\multirow{14}{*}{\tasktwo} & \multirow{14}{*}{ } & \multirow{1}{*}{Corner Bracket} &   & 48 \\
 &  & \multirow{1}{*}{Countersunk Screw} &   & 40 \\
 &  & \multirow{1}{*}{Cup Head Screw} &   & 234 \\
 &  & \multirow{1}{*}{Eye Bolt} &   & 86 \\
 &  & \multirow{1}{*}{Flat Washer} &   & 62 \\
 &  & \multirow{1}{*}{Hex Nut} &   & 58 \\
 &  & \multirow{1}{*}{Rivet Nut} &   & 58 \\
 &  & \multirow{1}{*}{Self tapping Screw} &   & 116 \\
 &  & \multirow{1}{*}{Spring Washer} &   & 88 \\
 &  & \multirow{1}{*}{T Bolt Half thread Screw} &   & 76 \\
 &  & \multirow{1}{*}{T Nut} &   & 62 \\
 &  & \multirow{1}{*}{T Screw} &   & 160 \\
 &  & \multirow{1}{*}{Wing Nut} &   & 54 \\
 &  & \multirow{1}{*}{Wing Screw} &   & 58 \\
\midrule
\multirow{25}{*}{\taskthree} & \multirow{6}{*}{\messcen} & \multirow{4}{*}{Model No} & Cup Head Screw M12 40 & 51 \\
 &  &  & Flat Washer M14 & 52 \\
 &  &  & Hex Nut M14 & 52 \\
 &  &  & Spring Washer M8 & 51 \\
 &  & \multirow{2}{*}{Workpiece} & Extra Flat Washers & 55 \\
 &  &  & Extra Spring Washers & 52 \\
 & \multirow{8}{*}{\swnscen} & \multirow{4}{*}{Model No} & Cup Head Screw M18 100 & 52 \\
 &  &  & Flat Washer M14 & 15 \\
 &  &  & Hex Nut M16 & 37 \\
 &  &  & Spring Washer M20 & 46 \\
 &  & \multirow{4}{*}{Workpiece} & Extra Cup Head Screw & 32 \\
 &  &  & Extra Flat Washers & 44 \\
 &  &  & Extra Hex Nut & 38 \\
 &  &  & Extra Spring Washers & 46 \\
 & \multirow{6}{*}{\pesscen} & \multirow{3}{*}{Model No} & Plastic anchors belonging to expansion screw model 860 were mixed & 51 \\
 &  &  & Screws belonging to model 860 expansion anchors were found mixed in & 56 \\
 &  &  & Shorter screws  were found mixed in & 50 \\
 &  & \multirow{3}{*}{Workpiece} & Model 860 expansion screws were mistakenly included & 51 \\
 &  &  & One extra plastic expansion anchor & 52 \\
 &  &  & One extra plastic expansion Self-tapping Screw & 54 \\
 & \multirow{5}{*}{\cncscen} & \multirow{4}{*}{Model No} & Long screw & 51 \\
 &  &  & retainer block & 56 \\
 &  &  & Short screw & 53 \\
 &  &  & Triangular part & 48 \\
 &  & \multirow{1}{*}{Workpiece} & Extra one nut & 56 \\
\midrule
 &  &  &  & 3559 \\
\bottomrule
\end{tabular}
\caption{Simple data summary of three tasks (scenario-level).}
\label{table:raw_3vimage_scenarios}
\end{table*}

\subsubsection{ Task Description of \taskone.}\label{app: task1}

In manufacturing scenarios, material sorting refers to the process of identifying, selecting, classifying, assembling, matching, and delivering materials from inventory to designated locations in accordance with specific requirements. Among these steps, material identification and verification are the most fundamental and essential. To evaluate the capability of MLLMs in material identification and verification within manufacturing environments, we design a task termed \taskone. Given explicit workpiece specifications or model requirements, the task requires MLLMs to analyze 3D point clouds or image data of manufacturing workpieces and identify those that do not satisfy the specified requirements. \textbf{Task design and details:} Based on this task, we construct three representative manufacturing application scenarios. Two of them are based on point cloud data, with the research objects being common manufacturing components, namely Nuts and Cup Head Screws. The third scenario is based on image data and focuses on commonly used Pneumatic Connectors (PCs). The illustration of \pcsscen \ is presented in Figure \ref{fig:  Manufacturing_task1_scenario_3_saved}, and the illustration of 3D point cloud scenario (\chsscen \ and \nutsscen) is presented in Figure \ref{fig: Manufacturing_task1_scenario_1_pointcloud_saved} and Figure \ref{fig: Manufacturing_task1_scenario_2_pointcloud_saved}.

\subsubsection{ Task Description of \tasktwo.}\label{app: task2}

Quality inspection has long been a fundamental research problem in manufacturing. Its core objective is to systematically inspect and measure workpieces, components, or production processes to ensure compliance with design specifications, process standards, and customer requirements. In this paper, we propose a \tasktwo \ task to evaluate the capability of MLLMs in identifying manufacturing defects from workpiece point cloud data in manufacturing scenarios. \textbf{Task design and details:} Specifically, the task involves: (1) determining whether a workpiece contains defects; and (2) further identifying the type of defect. The experimental evaluation covers 14 categories of manufacturing components, and the considered defect types are Crack, Cut, Deformation, and Dent. The point cloud data of different mechanical components are illustrated in Figures~\ref{fig: Corner_Bracket_saved}--\ref{fig: Wing_Screw_saved}.

%The illustration of point cloud data is in Figure \ref{fig: Corner_Bracket_saved},Figure \ref{fig: Cup_Head_Screw_saved},Figure \ref{fig: Eye_Bolt_saved},Figure \ref{fig: Flat_Washer_saved},Figure \ref{fig: Countersunk_Screw_saved},Figure \ref{fig: Hex_Nut_saved},Figure \ref{fig: Rivet_Nut_saved},Figure \ref{fig: Self_tapping_Screw_saved},Figure \ref{fig: Spring_Washer_saved},Figure \ref{fig: T_Bolt_Half_thread_Screw_saved},Figure \ref{fig: T_Nut_saved},Figure \ref{fig: T_Screw_saved},Figure \ref{fig: Wing_Nut_saved},and Figure \ref{fig: Wing_Screw_saved}.

\subsubsection{ Task Description of \taskthree.}\label{app: task3}

Assembly recognition is the automatic identification and understanding of assembly relationships, structural hierarchies, and compatibility constraints among workpieces or materials. Compared with material sorting, this task imposes greater requirements on MLLMs, as it requires them to reason about more complex assembly rules and compatibility relationships. To this end, we design an evaluation task termed \taskthree, which aims to systematically assess MLLMs capability in understanding assembly relationships.
Given specific compatibility rules, workpiece specifications, or model requirements, MLLMs analyze point clouds or images of manufacturing workpieces and identify those that do not meet assembly requirements. \textbf{Task design and details:} This task consists of four representative manufacturing scenarios. Among them, three scenarios are based on image data and involve common manufacturing applications, including metal expansion screws(\messcen), plastic expansion screws (\pesscen), and CNC fixtures (\cncscen). The remaining scenario is based on point cloud data and focuses on the compatibility relationships among metal screws, washers, and nuts. The illustration of image and point cloud data is in Figure \ref{fig: Manufacturing_scenario_1_saved}--\ref{fig: manufacturing_scenario_1_pointcloud_saved}.
%The illustration of image data is in Figure \ref{fig: Manufacturing_scenario_1_saved},Figure \ref{fig: Manufacturing_scenario_2_saved}, and Figure \ref{fig: Manufacturing_scenario_4_saved} and the illustration of point cloud data is in Figure \ref{fig: manufacturing_scenario_1_pointcloud_saved}

In summary, by designing three core tasks,\taskone, \tasktwo, and \taskthree, combined with high-quality multimodal data (point clouds and images), we have constructed a comprehensive and challenging benchmark corresponding to typical manufacturing contexts such as \textbf{material sorting}, \textbf{quality inspection}, and \textbf{assembly recognition}. This design not only addresses the fundamental perceptual requirements of manufacturing scenarios but also aims to rigorously evaluate models' logical judgment and reasoning capabilities under complex manufacturing rules. Building upon this framework, the following sections will systematically evaluate and discuss the reasoning mechanisms and performance of MLLMs in addressing real-world manufacturing challenges. Full visualization results are presented in Section \ref{app:image_examples}, and complete dialogue examples are provided in Section \ref{app:dataxemplars}.

\section{Training Details and Extended Results}\label{app:training_details}

\paragraph{Training Configurations.}
We train \texttt{Qwen2.5-VL-3B-Instruct} using \emph{full-parameter} supervised fine-tuning with the following configuration: bfloat16 precision, 3 epochs, per-device batch size 1, gradient accumulation steps 8, effective batch size 32 (1 $\times$ 8 $\times$ 4 GPUs), learning rate $1\times10^{-5}$, warmup ratio 0.05, weight decay 0.01, SDPA attention, frozen ViT, gradient checkpointing enabled, and response standardization enabled. Each task-specific model is trained independently. For \taskone\ 3V, training uses 8,337 examples from the \chsscen\ (Cup Head Screws). For \taskthree\ image, training uses 627 examples from the \messcen\ and \pesscen. Training data is converted to the standard single-turn chat format with the grounding-annotated image as input and the MCQ answer letter as the target response.

\paragraph{Scenario-Based Splitting.}
Unlike random train/test splits, our scenario-based protocol ensures the evaluation assemblies belong to an entirely different product category than those seen during training. For \taskone\ 3V, the \chsscen\ (Cup Head Screws) is used for training and the \nutsscen\ for evaluation. For \taskthree\ image, the \messcen\ and \pesscen\ (metal and plastic expansion screw assemblies) are used for training, and the \cncscen\ (CNC fixture assemblies) for evaluation. This tests whether the model learns \emph{transferable} manufacturing reasoning principles rather than assembly-specific visual patterns.

\paragraph{Full Comparison with Frontier Models.}
Table~\ref{tab:training_full_comparison} compares the finetuned Qwen2.5-VL-3B with five ablation models evaluated on the same held-out scenarios.

\begin{table}[h]
\centering
\definecolor{deepblue}{RGB}{198,219,239}
\definecolor{deeporange}{RGB}{253,208,162}
\definecolor{lightblue}{RGB}{230,218,240}
\setlength{\tabcolsep}{5pt}
\resizebox{\textwidth}{!}{
\begin{tabular}{l l c c}
\toprule
\textbf{Model} & \textbf{Type} & \textbf{\taskone\ 3V (\nutsscen)} & \textbf{\taskthree\ Img (\cncscen)} \\
\midrule
Gemini 3 Flash  & Closed & \cellcolor{deeporange}\textbf{63.6} & \cellcolor{deeporange}\textbf{53.5} \\
GPT-5.2         & Closed & \cellcolor{deepblue}\underline{62.6} & \cellcolor{deepblue}\underline{37.0} \\
Seed 1.6        & Closed & \cellcolor{lightblue}{60.0} & 23.6 \\
Qwen3-VL-235B   & Open   & 54.4 & 19.7 \\
Mistral 3 8B    & Open   & 35.4 & \cellcolor{lightblue}{26.0} \\
\midrule
Qwen2.5-VL-3B (base) & Open & 28.2 & 24.0 \\
\textbf{Qwen2.5-VL-3B (SFT)} & Open & \textbf{53.8} & \textbf{30.5} \\
\bottomrule
\end{tabular}}
\caption{\textbf{Finetuned 3B model vs.\ frontier models on held-out scenarios (zero-shot accuracy \%).} The finetuned Qwen2.5-VL-3B nearly matches Qwen3-VL-235B on \taskone\ (53.8\% vs.\ 54.4\%) and surpasses Seed~1.6, Qwen3-VL-235B, and Mistral~3~8B on \taskthree. \colorbox{deeporange}{\textbf{Bold}}/\colorbox{deepblue}{\underline{Underlined}}/\colorbox{lightblue}{Shaded} = best/2nd/3rd among the five reference models.}
\label{tab:training_full_comparison}
\end{table}

\paragraph{Task-Level Analysis.}
The asymmetry between the two tasks is informative. \taskone\ 3V requires distinguishing a single mismatched part among geometrically similar components, a skill that transfers well across scenarios because the training data teaches the model \emph{what constitutes a specification mismatch} in general, regardless of the specific assembly type. In contrast, \taskthree\ image involves diverse error modes (different model numbers and extra parts) across assembly categories with varying visual complexity; a scenario-held-out evaluation is therefore inherently harder because the model encounters part types and spatial layouts it has never seen. Nevertheless, even 627 training examples produce a measurable improvement on a completely novel product category, demonstrating that \ours\ annotations encode transferable manufacturing knowledge.

\paragraph{Per-Category Breakdown.}
On \taskone\ 3V (\nutsscen), the finetuned model shows a notable shift in behavior: accuracy on Normal cases improves dramatically from 24.3\% to 74.8\%, while accuracy on Different Model Number cases decreases from 33.8\% to 23.8\%. This indicates the model learns a more conservative decision boundary, correctly recognizing conforming assemblies far more often, at the cost of some sensitivity to subtle specification mismatches. Since Normal cases constitute 59\% of the evaluation set (115/195), the net effect is a large overall improvement. Future work could explore loss weighting or balanced sampling to mitigate this calibration trade-off.

On \taskthree\ image (\cncscen), improvements are more uniform: Different Model Number accuracy increases from 20.5\% to 27.0\% and Extra Part from 37.0\% to 44.4\%, suggesting the training signal transfers consistently across error types to a novel product category.

\section{Extended Error Cases and Bottleneck Analysis}\label{app: bandexemplars}

This section supplements the qualitative error analysis (Section 4.4) and the bottleneck analysis (Section 4.5) of the main text with complete transcripts and detailed experimental configurations. Section~\ref{app: errexemplars} presents the full prompt--response pairs for the two representative error cases discussed in the main paper, allowing readers to trace each model's reasoning chain from input to incorrect prediction. Section~\ref{app: bottleneck} then provides the experimental settings and configurations for all three bottleneck analyses (visual grounding, missing-part detection, and raw point cloud text input), followed by additional discussion of the findings.

\subsection{Error Cases for Main Experiment}\label{app: errexemplars}

This section presents the complete prompt and response transcripts for the two error cases analyzed qualitatively in Section 4.4 of the main text. Both examples are drawn from zero-shot \taskthree\ evaluation on Gemini-2.5-Flash and illustrate recurring failure modes discussed in the main paper.

\begin{figure}[htbp]
\centering

\begin{subfigure}{0.40\textwidth}
    \centering
    \includegraphics[width=\linewidth]{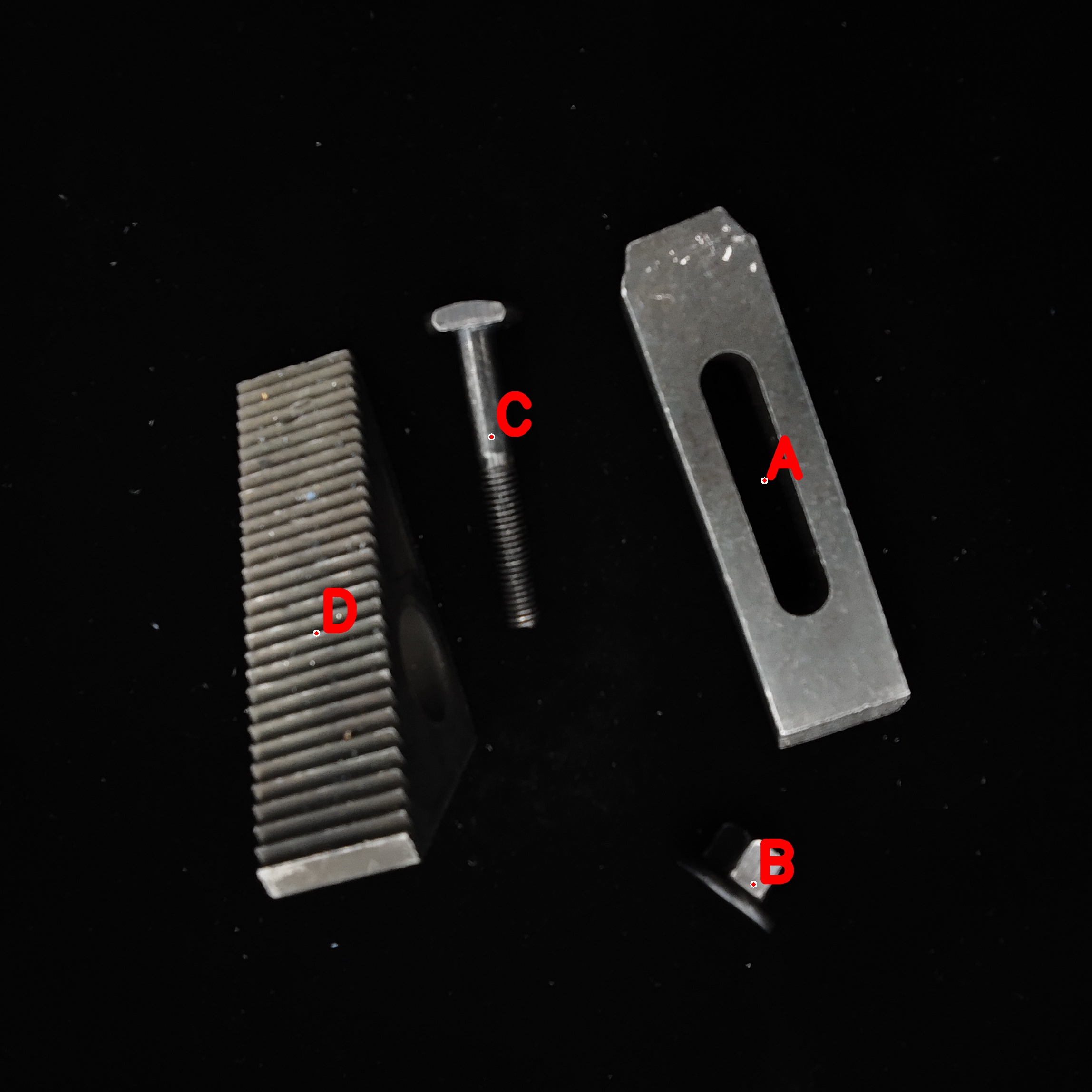}
    \caption{Error Case 1}
\end{subfigure}
%\hfill
\begin{subfigure}{0.40\textwidth}
    \centering
    \includegraphics[width=\linewidth]{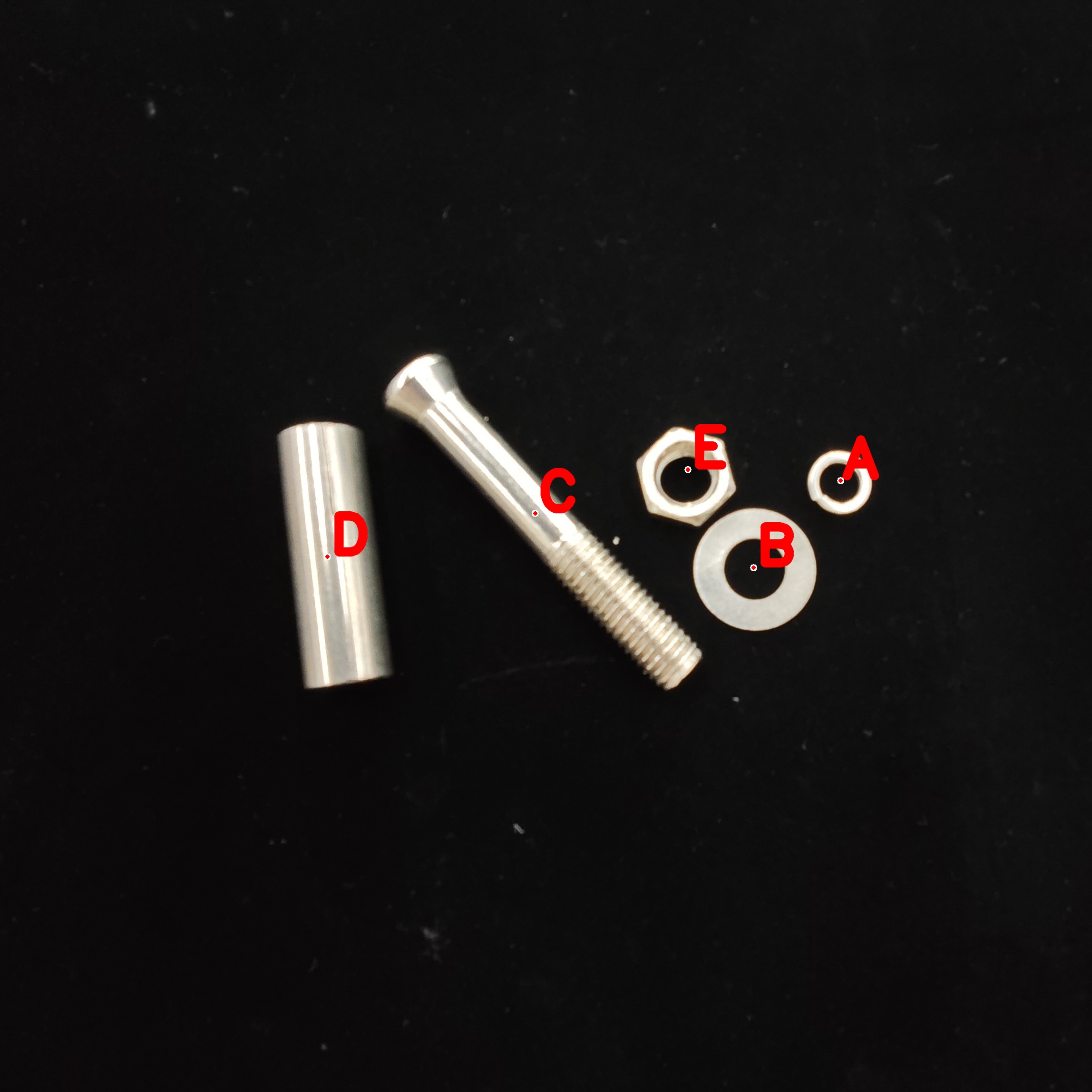}
    \caption{Error Case 2}
\end{subfigure}

\caption{Two error cases with red annotations added for clearer interpretation.}
\label{fig:errorimages}
\end{figure}

\textbf{Error Case~1} (\cncscen) corresponds to error pattern~B in the main text: the model correctly identifies all four workpiece types in the CNC fixture assembly but misjudges which component has the wrong specification. Specifically, it concludes that the Hex Nut~(B) is undersized relative to the T-slot bolt, when the actual anomaly is an oversized Step Block~(D). The full response reveals extended deliberation about size compatibility and material condition, yet the model ultimately anchors on the wrong component. The image input is in Figure~\ref{fig:errorimages}(a).

\textbf{Error Case~2} (\messcen) corresponds to error pattern~A: the model hallucinates material properties from visual textures, misidentifying a metal Flat Washer(E) as "plastic/nylon", and then uses this erroneous material inference to select it as the anomalous part (predicting~E instead of the correct answer~A). Despite the incorrect conclusion, the response demonstrates MLLM's inclination to reason about material composition, a capability that could become valuable if grounded in accurate perception. The image input is in Figure~\ref{fig:errorimages}(b).

Together, these transcripts provide the full reasoning chains behind the error patterns summarized in the main text, enabling readers to inspect the exact prompts, intermediate reasoning steps, and final predictions.

\begin{AIbox}{\taskthree \ \cncscen  \ Error Cases 1} \label{example:image CNC Zero-Shot}
  \small
  \begin{itemize}
     \item \textbf{System Prompt:}
          
          General task:
          
          - You will be given a single image of an assembly containing multiple industrial parts that match each other well.
          
          - Your job is to identify which part in the provided test image uses the wrong model number (incorrect size or shape) and thus deviates from the rest.
          
          - Instead of returning a coordinate directly, you will be presented with a list of multiple-choice options (A, B, C, ...), each describing one candidate part location.
          
          - Choose the letter corresponding to the mismatched part.
          
          Input format:
          
          - Query: a single line stating the task.
          
          - Image: the image to evaluate.
          
          - Choices: a list of options, each with a normalized coordinate for one part in the image (normalized to [0,1]).
          
          Output format:
          
          - Return exactly one letter in the format " Answer: X " at the end of your response.
          
          - You could include your reasoning before the answer.
  \end{itemize}
\end{AIbox}

\begin{AIbox}{\taskthree \ \cncscen  \ Error Cases 1} 
  \small
  \begin{itemize}
  \item \textbf{Setting:}  Zero-Shot
    \item \textbf{User Prompt:}
    
    In CNC machining, the primary role of fixtures is to ensure that the workpiece is accurately positioned and firmly secured during various cutting, milling, or drilling operations.

    A typical CNC fixture is composed of several fundamental elements that work together to deliver reliable clamping performance. These include: (1) the flange nut, (2) the T-bolt half-thread screw, (3) the pressure plate, and (4) the triangular wedge.
    
    Based on the provided reference images of correctly assembled Computer Numerical Control (CNC) fixture, check the test image for any extra, unnecessary, or incorrectly specified parts. Determine whether the test image contains any components that should not be present or have incorrect specifications. Identify any such problematic parts.
    
    Below are candidate parts with their normalized coordinates in [0,1].
    
    Pick the letter corresponding to the mismatched part.
    
    A. Part at [0.70, 0.44]
    
    B. Part at [0.69, 0.81]
    
    C. Part at [0.45, 0.40]
    
    D. Part at [0.29, 0.58]
    
    Answer:
  \end{itemize}

\end{AIbox}

\begin{AIbox}{\taskthree \ \cncscen  \ Error Cases 1} 
  \small
  \begin{itemize}
      \item \textbf{Model:} Gemini-2.5-Flash
  \item   
   Gt answers:  D
   
   \item    Pred answer:B
    \item \textbf{Raw Response 1:} 
    
    The image displays components typically found in a clamping set for a CNC machine (T-slot clamp set). These usually include a T-slot bolt, a step block (or serrated block), a strap clamp (or clamping strap), and a flange nut.
    
    The parts present are:
    
    1.  A serrated/step block (left side, [0.29, 0.58]).
    
    2.  A T-slot bolt (center, [0.45, 0.40]).
    
    3.  A strap clamp (right side, [0.70, 0.44]).
    
    4.  A nut (bottom right, [0.69, 0.81]).
    
    The standard nut used in a T-slot clamping set is a **flange nut** (or sometimes a coupling nut or standard hex nut, depending on the specific kit). A flange nut has a wide flange at one end that acts as an integrated washer, distributing the pressure over a larger area.
    
    The part at location B ([0.69, 0.81]) is a nut. However, it appears to be a standard **hex nut** or possibly a **T-nut** (though it looks more like a hex nut based on the visible facets and lack of a clear T-shape profile for a T-slot).
    
    Let's re-examine the parts based on typical clamping sets:

    *   A ([0.70, 0.44]): Strap clamp. Correct component.
    
    *   B ([0.69, 0.81]): Nut. This appears to be a standard hex nut, which is often used, but sometimes a flange nut is preferred for better clamping force distribution. However, looking closely at the shape, it seems to be a standard hex nut, which is a common component.
    
    *   C ([0.45, 0.40]): T-slot bolt. Correct component.
    
    *   D ([0.29, 0.58]): Step block/serrated block. Correct component.
    
    In a typical T-slot clamping set, all four components are necessary. If we are looking for a part with an *incorrect specification* (wrong size or shape relative to the others), we must compare them.
    
    The T-slot bolt (C) and the strap clamp (A) appear to be sized appropriately for each other (the slot in A looks like it would fit the bolt C). The step block (D) is also standard.
    
    Let's look closely at the nut (B). It is a small hex nut. If this kit is intended to be a complete, matched set, the nut should match the thread size of the bolt (C). Without knowing the exact thread size, it's hard to confirm a mismatch based on size alone.

  \end{itemize}
\end{AIbox}

\begin{AIbox}{\taskthree \ \cncscen  \ Error Cases 1} 
  \small
  \begin{itemize}
  
    \item \textbf{Raw Response 2:} 
    
    However, if we consider the overall condition and type:
    
    The strap clamp (A) is heavily worn/chipped at the ends.
    
    The step block (D) is also worn.
    
    The T-slot bolt (C) is also worn.
    
    The nut (B) is visually distinct from the other three main components (A, C, D) in terms of its finish and apparent size relative to the bolt (C). It looks disproportionately small or perhaps like a different type of nut than what is typically paired with a T-slot bolt of that size, especially if a flange nut is expected.
    
    Let's assume the task requires identifying the part that is visually the most likely mismatch in terms of model/specification.
    
    The part at B is a small hex nut. Given the size of the T-slot bolt (C), this nut seems undersized for heavy-duty clamping, or it might be a nut for a different, smaller bolt. If the bolt (C) is M12 or M16, the nut (B) looks too small to be the corresponding standard hex nut.
    
    Comparing the parts again:
    
    A, C, and D are clearly industrial clamping components. B is a fastener (nut). If B is intended to mate with C, its size relative to the threaded portion of C seems questionable. The diameter of the nut (B) appears significantly smaller than the head of the bolt (C), and possibly smaller than the threaded shaft diameter, suggesting it might be the wrong size or thread pitch for the bolt C.
    
    Therefore, the nut at B is the most likely candidate for an incorrectly specified part (wrong size/model number) relative to the T-slot bolt (C).
    
    Answer: B
  \end{itemize}
\end{AIbox}

\begin{AIbox}{\taskthree \ \messcen \ Error Cases 2} \label{error:(MEB) Image Zero-Shot}
  \small%      "normal_case": "raw_full_data/Manufacturing_scenario_1/Not_Normal_Case/Different_Model_number/Spring_Washer_M8/image_35.jpg",       "grounding_case": "raw_full_data/Manufacturing_scenario_1/Not_Normal_Case/Different_Model_number/Spring_Washer_M8/image_35_grounding.jpg",
  \begin{itemize}
  
  \item \textbf{Setting:}  Zero-Shot
 
  \item  \textbf{Query}

  Based on the provided reference images of correctly assembled Metal expansion bolt, check the test image for any extra, unnecessary, or incorrectly specified parts. A standard assembly consists of five key components, each contributing to its performance and reliability: (1)Bolt - the main fastening element that provides tensile strength; (2)Nut - used to tighten and secure the bolt in place; (3)Flat Washer - distributes the load evenly across the surface, reducing the risk of damage to the base material; (4)Spring Washer - prevents loosening caused by vibrations or dynamic loads; (5)Sleeve - the expansion element that deforms outward under tightening, ensuring strong anchorage in the drilled hole. Determine whether the test image contains any components that should not be present or have incorrect specifications. Identify any such problematic parts.
  
  Below are candidate parts with their normalized coordinates in [0,1].
  
  Pick the letter corresponding to the mismatched part.
  
  A. Part at [0.77, 0.44]
  
  B. Part at [0.69, 0.52]
  
  C. Part at [0.49, 0.47]
  
  D. Part at [0.30, 0.51]
  
  E. Part at [0.63, 0.43]
  
  Answer:

  \end{itemize}
\end{AIbox}

\begin{AIbox}{\taskthree \ \messcen \ Error Cases 2}  
  \small
  \begin{itemize} 
        \item \textbf{Model:} Gemini-2.5-Flash

  \item   
   Gt answers:  A
   
   \item    Pred answer: E
  \item  \textbf{Raw Response:}
  
    The standard assembly for a Metal expansion bolt consists of five components: Bolt, Nut, Flat Washer, Spring Washer, and Sleeve.
    
    Let's identify the parts present in the image:
    
    1.  Sleeve (leftmost cylindrical part): Part at [0.30, 0.51] (D). This is a required component.
    
    2.  Bolt (long threaded part with a conical head): Part at [0.49, 0.47] (C). This is a required component.
    
    3.  Nut (large hexagonal part): Part at [0.69, 0.52] (B). This is a required component.
    
    4.  Flat Washer (large white/translucent ring): Part at [0.63, 0.43] (E). This is a required component.
    
    5.  Small Ring (small metal ring): Part at [0.77, 0.44] (A). This appears to be a small metal washer, likely a Spring Washer (or a second, smaller flat washer). A Spring Washer is a required component.
    
    The image contains five parts, which matches the expected number of components (Bolt, Nut, Flat Washer, Spring Washer, Sleeve).
    
    However, upon closer inspection of the components:

    *   The part at [0.63, 0.43] (E) is a white/translucent plastic or nylon washer (Flat Washer).
    
    *   The part at [0.77, 0.44] (A) is a small, thin metal ring/washer. This is likely intended to be the Spring Washer, but it is a simple flat washer, not a split/spring washer, and it appears to be significantly smaller than the bolt's diameter, suggesting it might be the wrong size or type.
    
    *   The part at [0.69, 0.52] (B) is the Nut.
    
    *   The part at [0.49, 0.47] (C) is the Bolt.
    
    *   The part at [0.30, 0.51] (D) is the Sleeve.
    
    The most significant deviation is the presence of a plastic/nylon Flat Washer (E) and the absence of a metal Flat Washer and a proper Spring Washer.
    
    In the context of industrial parts matching, the most obvious deviation in material and specification is the Flat Washer at [0.63, 0.43] (E). Expansion bolts are typically high-strength metal assemblies, and using a plastic/nylon washer for load distribution (Flat Washer) is an incorrect specification that compromises the assembly's strength and durability, especially when the other components are metal.
    
    If we assume the small metal ring (A) is meant to be the Spring Washer, then the plastic washer (E) is the incorrect component due to its material.
    
    Therefore, the part at [0.63, 0.43] is incorrectly specified (wrong material).
    
    Answer: E

  \end{itemize}
\end{AIbox}

\iffalse
\noindent
\begin{AIbox}{Graph Tasks}
  \small
  \begin{itemize}
    \item xx
    \item xxx
  \end{itemize}
\end{AIbox}
\fi

\subsection{Details of Bottleneck Analysis}\label{app: bottleneck}

\subsubsection{ Settings and Configurations.}\label{app: abdescrip}

This section details the experimental setup for each of the three bottleneck analyses introduced in the main text. Unless stated otherwise, all analyses use a temperature of 0.0 for deterministic outputs and evaluate five representative models spanning both open- and closed-source families: Gemini-3-Flash (closed), GPT-5.2 (closed), Qwen3-VL-235B (open), Seed-1.6 (closed), and Mistral-3-8B (open). All models are accessed via the OpenRouter API.

\paragraph{A. Visual Grounding (Table~3 in the main text).}
This analysis isolates each model's ability to ground and localize individual components, decoupled from downstream reasoning. Each assembly image is annotated with Set-of-Mark~\cite{yang2023set} letter labels (A, B, C, \etc) on individual parts.

\begin{itemize}
  \item \textbf{Single-image grounding} (500 sampled cases from 4{,}257 total). Two complementary sub-tasks are tested: (i)~\emph{Coordinate-to-Letter} (C$\to$L): given a normalized coordinate $[x,y] \in [0,1]^2$ (origin at the top-left corner), the model must identify the letter of the corresponding part; (ii)~\emph{Letter-to-Coordinate} (L$\to$C): given a letter label, the model must select the correct coordinates from an MCQ. Images are resized so that the longest side is at most 1{,}568 pixels.
  \item \textbf{Cross-image correspondence} (513 sampled pairs from 14{,}992 total). Two images from the same assembly scenario are shown, each with independently assigned letter labels. A part is identified in the first (reference) image, and the model must locate the visually corresponding part in the second (test) image. Two modes are evaluated: (i)~\emph{Letter-to-Letter} (L$\to$L) and (ii)~\emph{Coordinate-to-Coordinate} (C$\to$C).
\end{itemize}
All grounding evaluations use zero-shot prompting with \texttt{max\_tokens}$\,=\,$512 and reasoning disabled. Grounding image examples are shown in Figures \ref{figab: Manufacturing_scenario_1_saved}--\ref{figab: Manufacturing_task1_scenario_3_saved} in Section \ref{app:grounding_exemplars}

\paragraph{B. Missing Part Detection (Table~4 in the main text).}
This analysis probes domain-specific reasoning by providing models with explicit assembly specifications. The model receives a complete description of the assembly (part names, quantities, and functional descriptions) and must identify which component is absent from an MCQ. Each case includes a "No part is missing" option.

\begin{itemize}
  \item \textbf{Image modality}: 240 test cases across three scenarios (\messcen\ with 6 MCQ options, \pesscen\ with 3 options, and \cncscen\ with 5 options).
  \item \textbf{Three-view modality}: 137 test cases from the \swnscen\ (5 MCQ options).
\end{itemize}
All missing-part evaluations use zero-shot prompting with \texttt{max\_tokens}$\,=\,$8{,}192 and extended reasoning enabled.

\paragraph{C. Raw Point Cloud Text Input (Table~5 in the main text).}
This analysis investigates whether the text modality can serve as an alternative 3D interface for general-purpose MLLMs. Point clouds are serialized as integer-scaled text tables and fed directly to the model, bypassing visual rendering. Two representative models are evaluated: Gemini-3-Flash and Qwen3-235B (text-only mode).

The serialization procedure is as follows: each part's 3D coordinates are randomly sampled to $n_{\text{points}}=500$ points, then scaled by $10^{d}$ where $d=2$ (round digits), yielding integer coordinates. The resulting table format is:
\begin{verbatim}
x, y, z
1896, 750, 31970
3732, -3451, 32051
...
\end{verbatim}
For \taskone\ and \taskthree, assemblies are segmented into individual parts via DBSCAN clustering ($\varepsilon=3.0$, $\text{min\_samples}=50$), each assigned a random letter label (A, B, C, \etc). For \tasktwo, a single part's coordinates are provided. All three evaluation modes are tested (Zero-Shot, Ref-Cond, and ICD) with \texttt{max\_tokens}$\,=\,$1{,}024 across \taskone\ (420 cases), \tasktwo\ (460 cases), and \taskthree\ (266 cases).

\subsubsection{Additional Discussion of Bottleneck Analyses.}\label{app:ab_bas}

Detailed examples of bottleneck analyses A, B, and C discussed in the main text are presented in Sections \ref{app:exemplars_BA1}, \ref{app:exemplars_BA2}, and \ref{app:exemplars_BA3} in this Appendix, respectively.

Recall that the motivation for the "Missing Part" scenario in Bottleneck Analysis B is to further disentangle domain-specific reasoning from explicit grounding capabilities. As demonstrated in Example \ref{app:exemplars_BA2}, this analysis provides a more direct evaluation of the model's intrinsic task comprehension. Because the System and User Prompts strictly exclude coordinate information and constrain the output format (e.g., "Answer with ONLY the letter of your choice"), the model is forced to rely entirely on logical deduction. Consequently, the raw response reveals an extensive reasoning process where the model continuously verifies its assumptions and corrects its own logical errors. This internal debate is vividly captured by phrases such as: " Wait, the assembly needs four parts: Bolt (B?), Nut (maybe C?), Flat Washer (A?), Spring Washer (D?).", "But wait, the problem is to check if all four are present", and "Wait, but a Metal Expansion Bolt might not require a nut?". This pronounced hesitation and reliance on iterative self-correction underscore the model's fragility when deprived of explicit spatial anchors. It highlights a critical gap between merely recognizing individual components and robustly comprehending their relational assembly logic. Meanwhile, the image presentation of the missing part can follow the images (Figure \ref{fig: Manufacturing_scenario_1_saved}--\ref{fig: Manufacturing_scenario_4_saved} ) shown in Section  \ref{app:image_examples}.

The analysis of example \ref{app:exemplars_BA3} reveals that in the raw response, LLMs spend a significant amount of time repeatedly analyzing and processing point cloud coordinate data, relying on range-related calculations to determine the relative sizes of the corresponding parts. For instance, "Part C's z span is \textasciitilde6200, which is 3.26 larger than Part A's (1900), 4.77 larger than Part B's (1300), and 2.38 larger than Part D's (2600)." This suggests that when provided with point cloud coordinates as input, LLMs exhibit the potential to assess relative sizes. However, this method of processing coordinate data struggles to analyze microscopic morphology, which explains why MLLMs achieves a certain level of performance in \taskone, whereas performance on the remaining tasks is noticeably inferior. Nevertheless, the responses of MLLMs also reveal certain inherent capabilities for processing point clouds; therefore, analyzing coordinate information and absolute values can provide valuable insights from an alternative perspective.

\newpage
\section{Image examples}\label{app:image_examples}

\begin{figure}[htbp]
\centering

\begin{minipage}{0.23\textwidth}
    \centering
    \parbox[c][3.0cm][c]{\linewidth}{
        \centering\includegraphics[width=\linewidth,height=3.0cm,keepaspectratio]{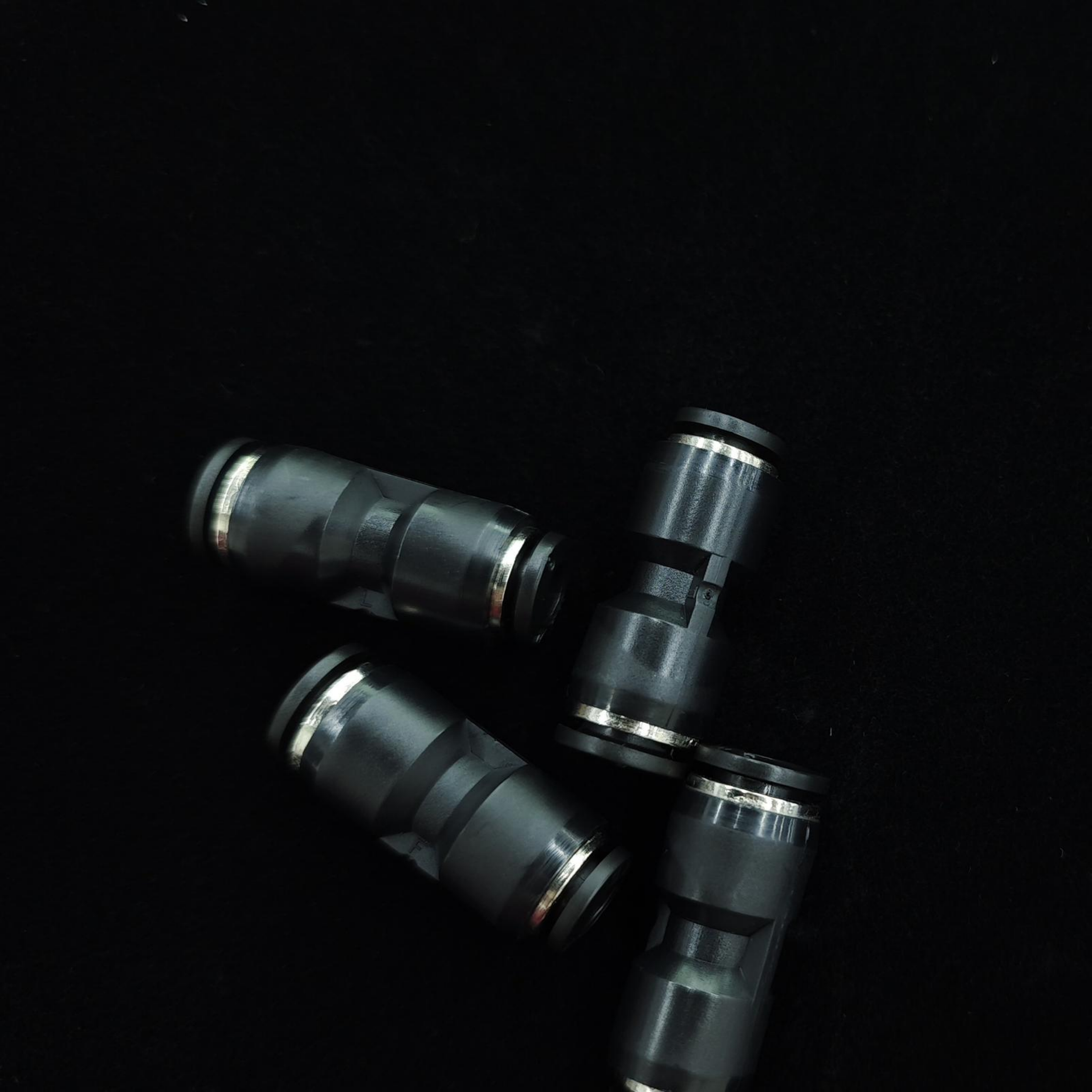}
    }
    \parbox[t][1.5cm][t]{\linewidth}{\small Normal Case}
\end{minipage}\hfill
\begin{minipage}{0.23\textwidth}
    \centering
    \parbox[c][3.0cm][c]{\linewidth}{
        \centering\includegraphics[width=\linewidth,height=3.0cm,keepaspectratio]{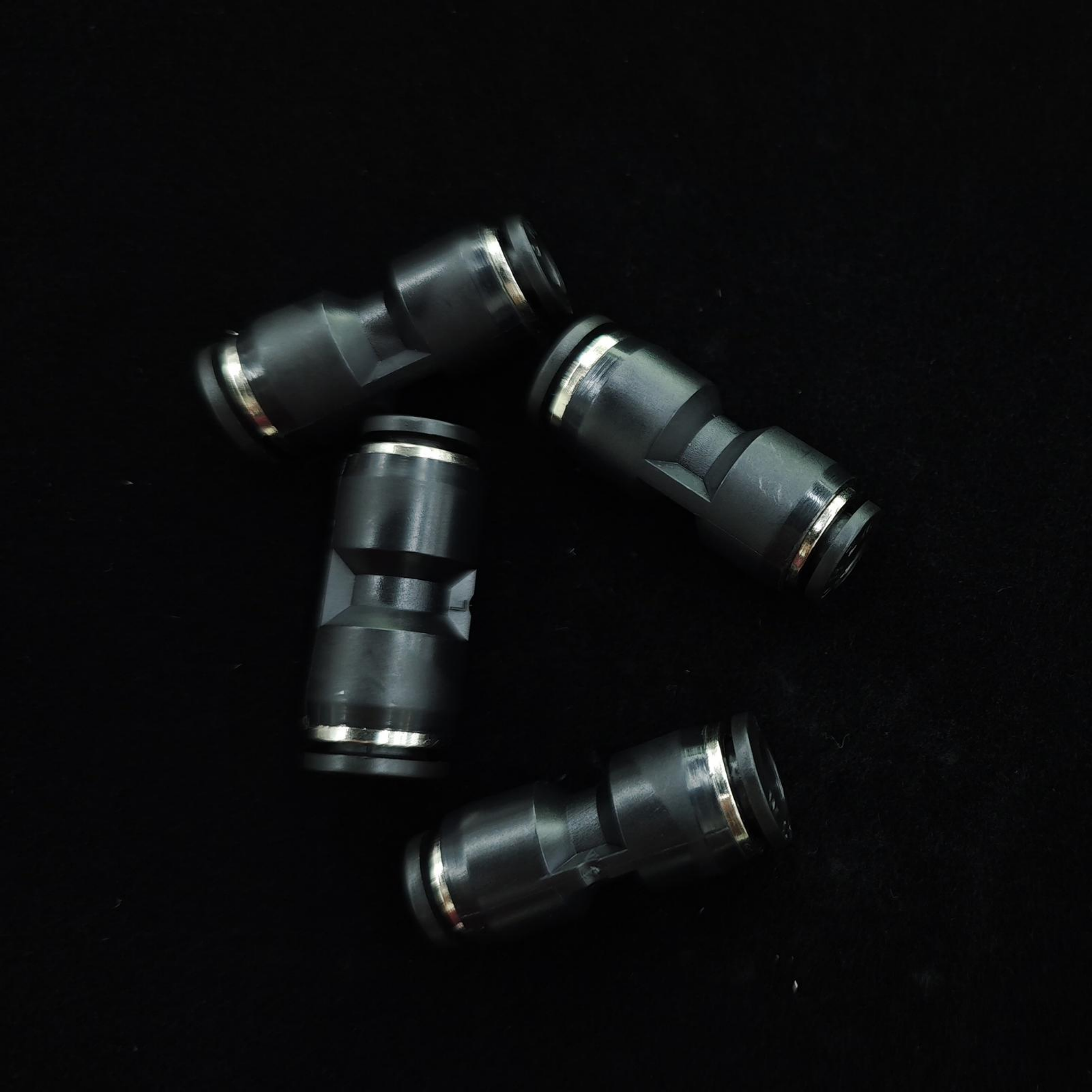}
    }
    \parbox[t][1.5cm][t]{\linewidth}{\small Normal Case}
\end{minipage}\hfill
\begin{minipage}{0.23\textwidth}
    \centering
    \parbox[c][3.0cm][c]{\linewidth}{
        \centering\includegraphics[width=\linewidth,height=3.0cm,keepaspectratio]{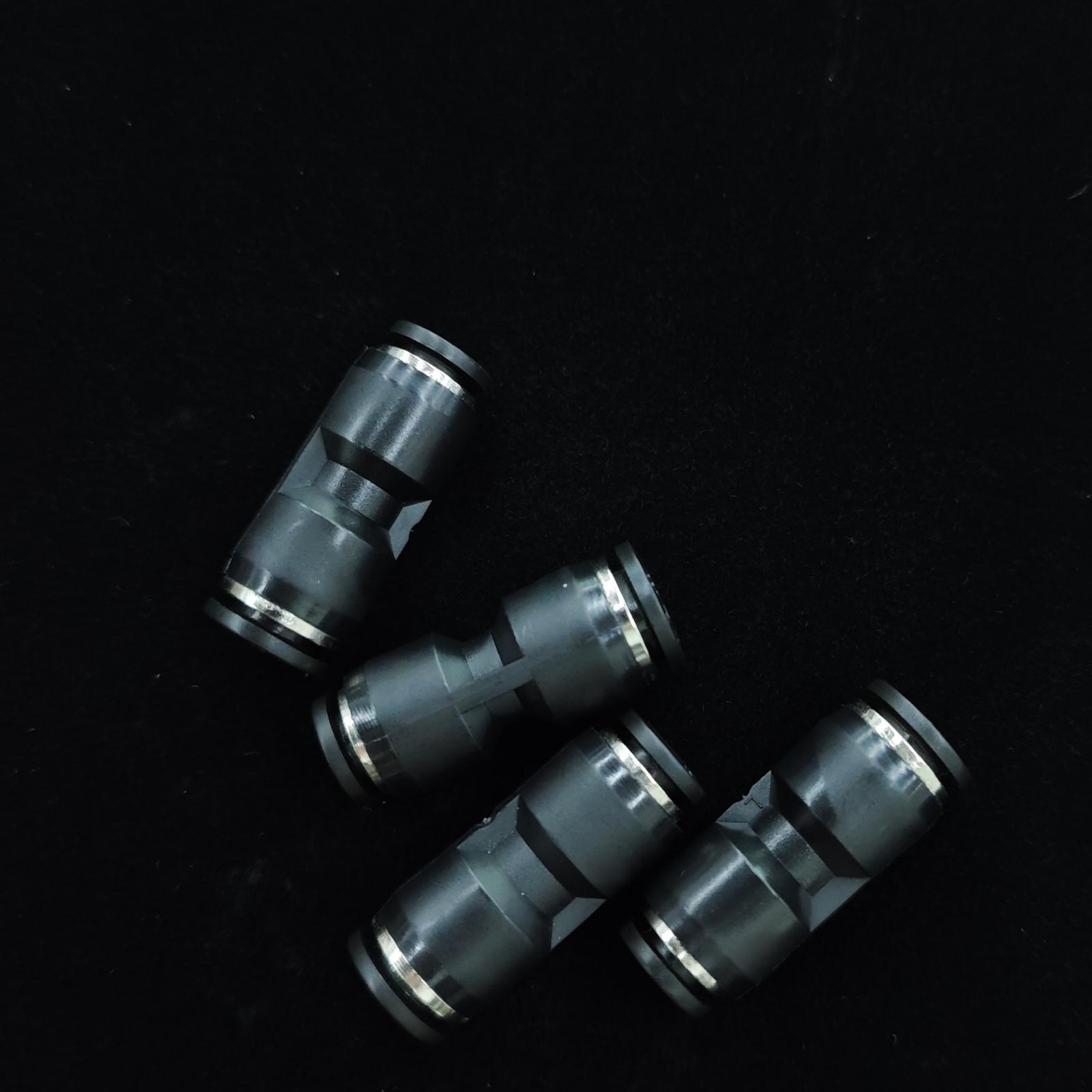}
    }
    \parbox[t][1.5cm][t]{\linewidth}{\small Normal Case}
\end{minipage}\hfill
\begin{minipage}{0.23\textwidth}
    \centering
    \parbox[c][3.0cm][c]{\linewidth}{
        \centering\includegraphics[width=\linewidth,height=3.0cm,keepaspectratio]{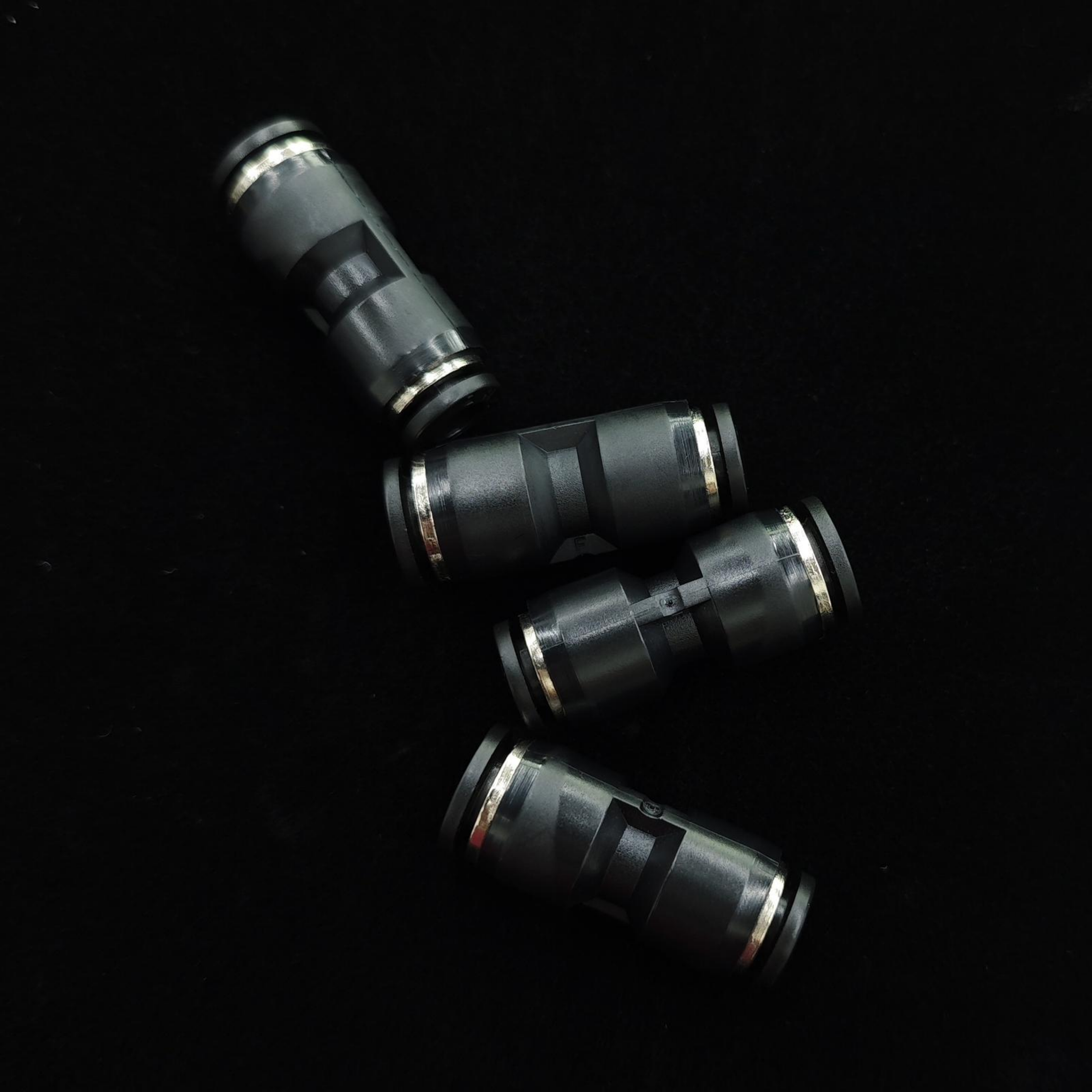}
    }
    \parbox[t][1.5cm][t]{\linewidth}{\small Normal Case}
\end{minipage}

\begin{minipage}{0.23\textwidth}
    \centering
    \parbox[c][3.0cm][c]{\linewidth}{
        \centering\includegraphics[width=\linewidth,height=3.0cm,keepaspectratio]{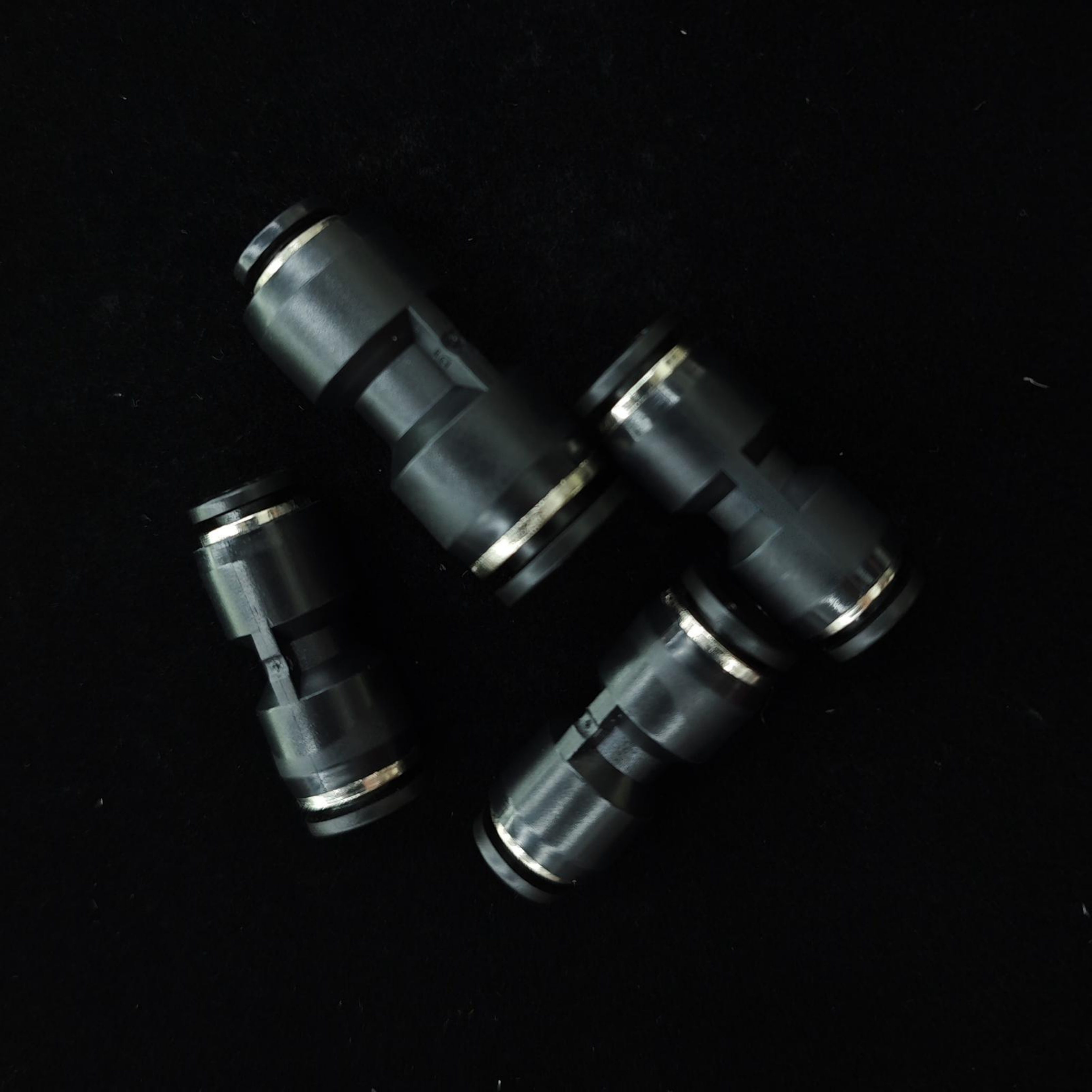}
    }
    \parbox[t][1.5cm][t]{\linewidth}{\small   mixed with model (10 to 6)}
\end{minipage}\hfill
\begin{minipage}{0.23\textwidth}
    \centering
    \parbox[c][3.0cm][c]{\linewidth}{
        \centering\includegraphics[width=\linewidth,height=3.0cm,keepaspectratio]{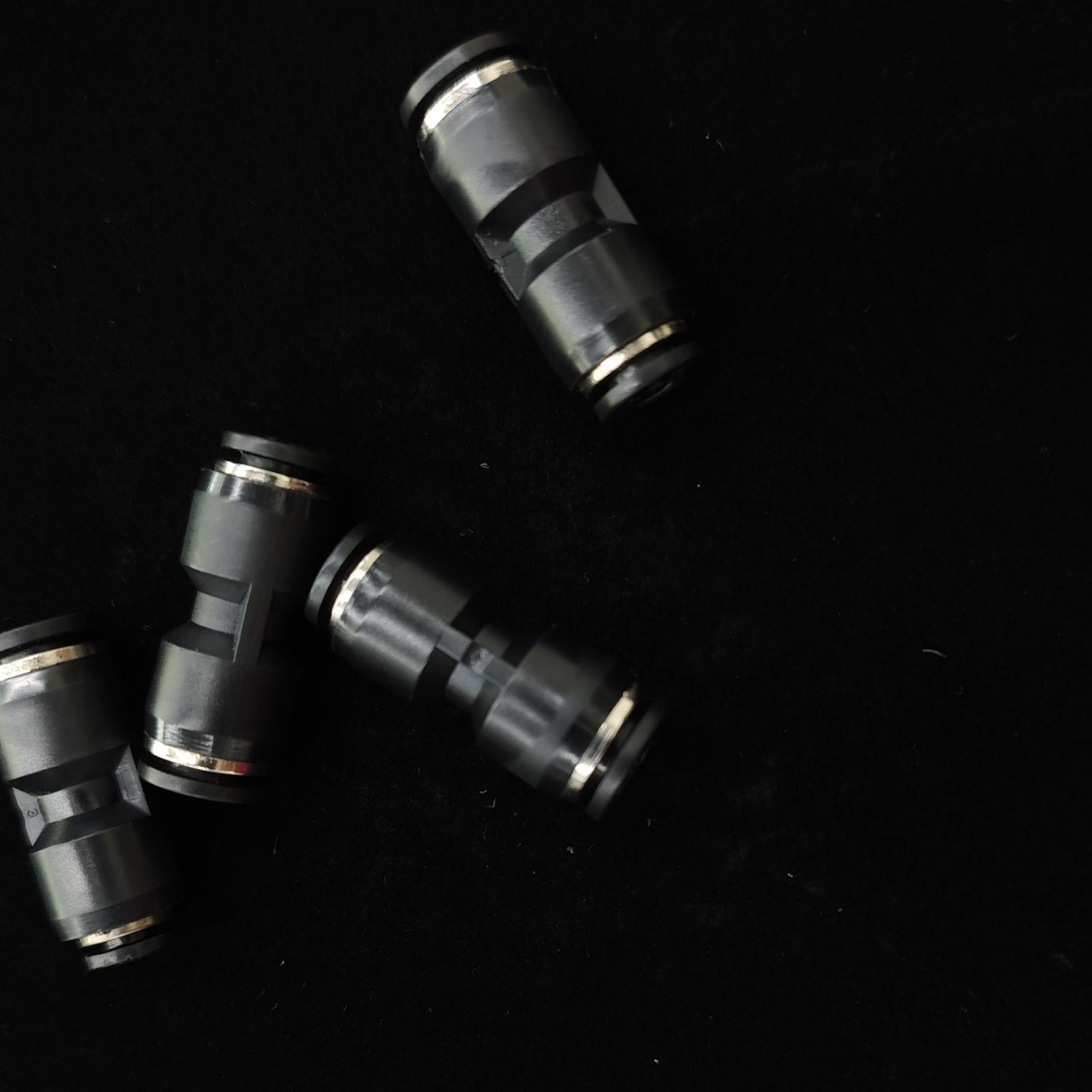}
    }
    \parbox[t][1.5cm][t]{\linewidth}{\small   mixed with model (8 to 4)}
\end{minipage}\hfill
\begin{minipage}{0.23\textwidth}
    \centering
    \parbox[c][3.0cm][c]{\linewidth}{
        \centering\includegraphics[width=\linewidth,height=3.0cm,keepaspectratio]{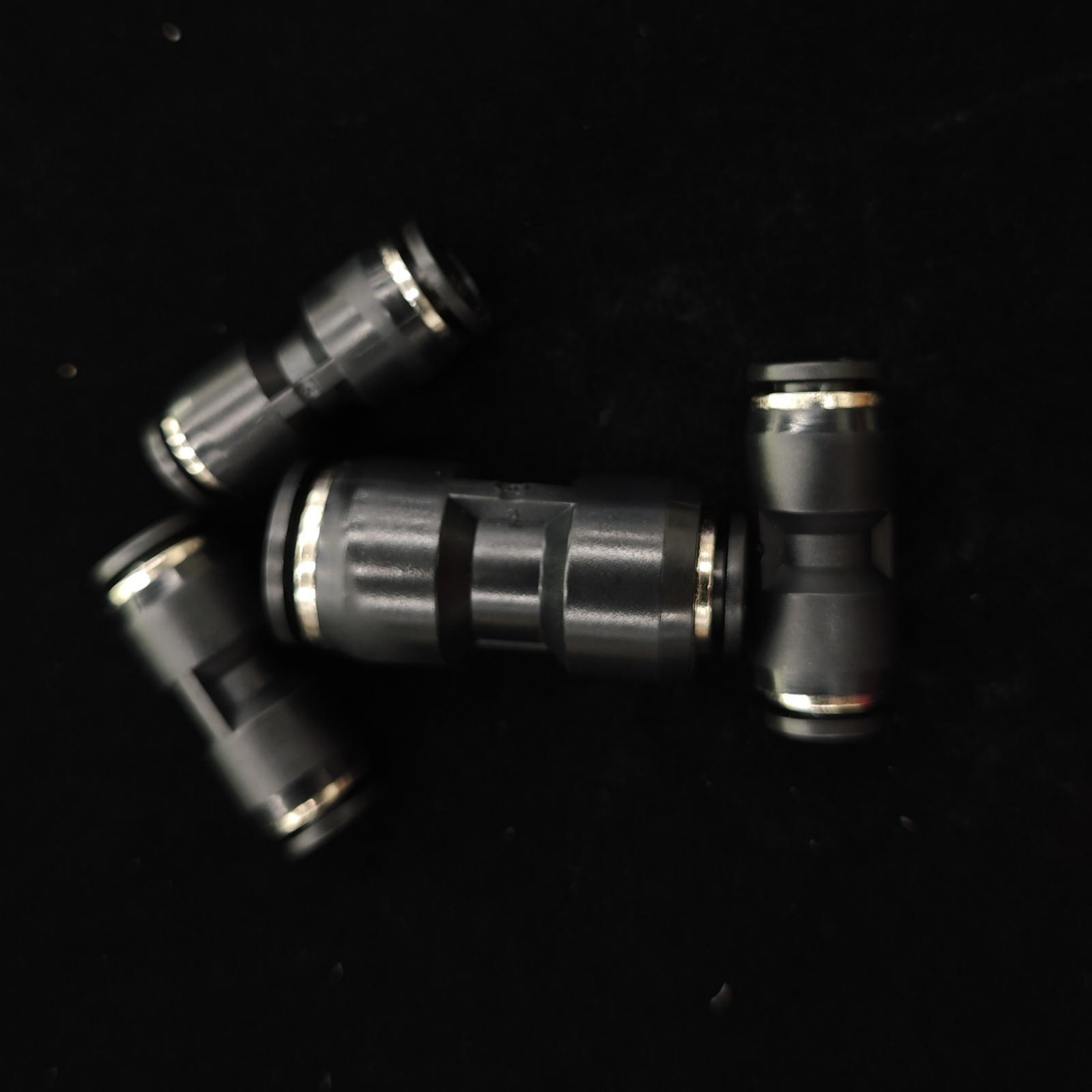}
    }
    \parbox[t][1.5cm][t]{\linewidth}{\small   mixed with model (12 to 8)}
\end{minipage}\hfill
\begin{minipage}{0.23\textwidth}
    \centering
    \parbox[c][3.0cm][c]{\linewidth}{
        \centering\includegraphics[width=\linewidth,height=3.0cm,keepaspectratio]{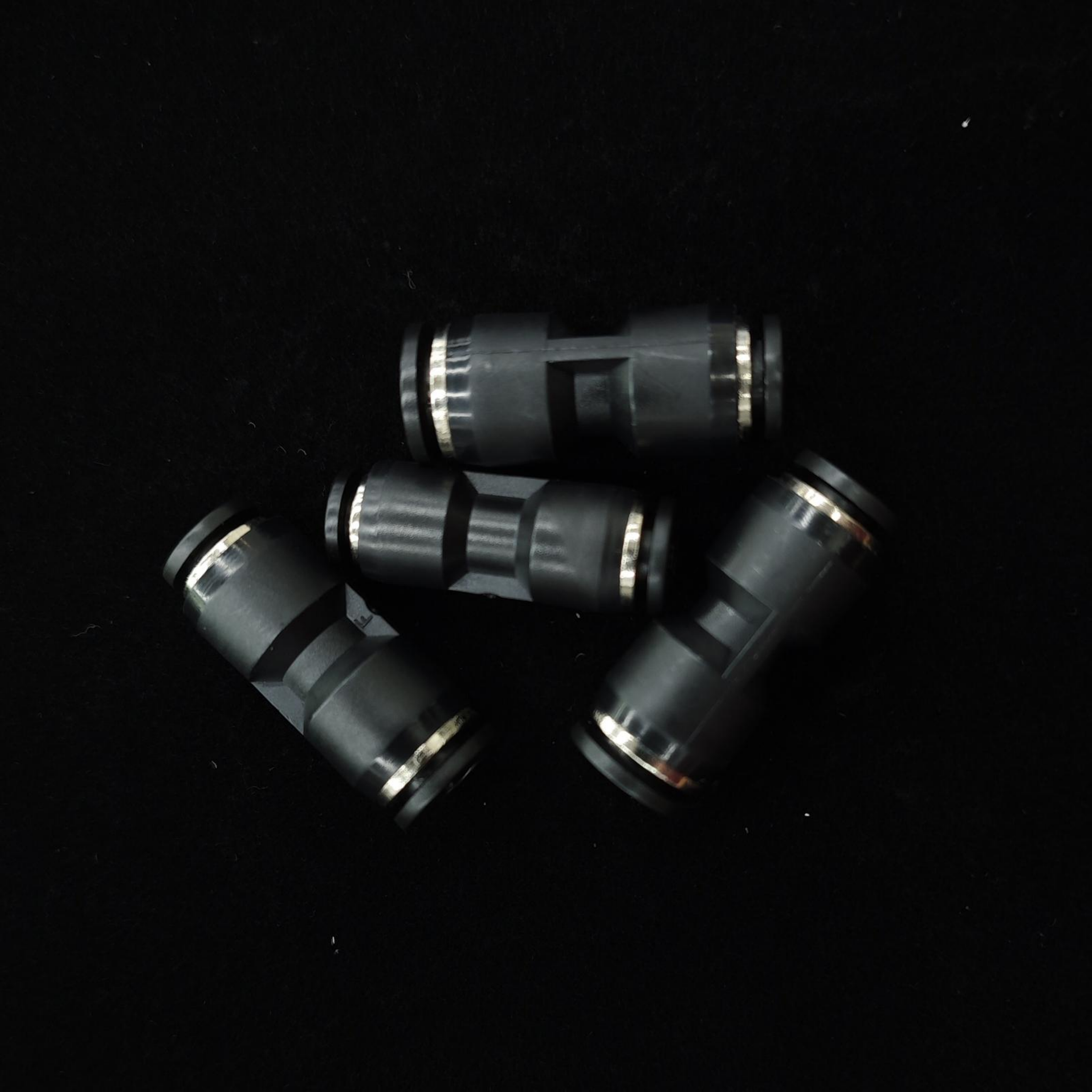}
    }
    \parbox[t][1.5cm][t]{\linewidth}{\small   mixed with model (6 to 4)}
\end{minipage}

\begin{minipage}{0.23\textwidth}
    \centering
    \parbox[c][3.0cm][c]{\linewidth}{
        \centering\includegraphics[width=\linewidth,height=3.0cm,keepaspectratio]{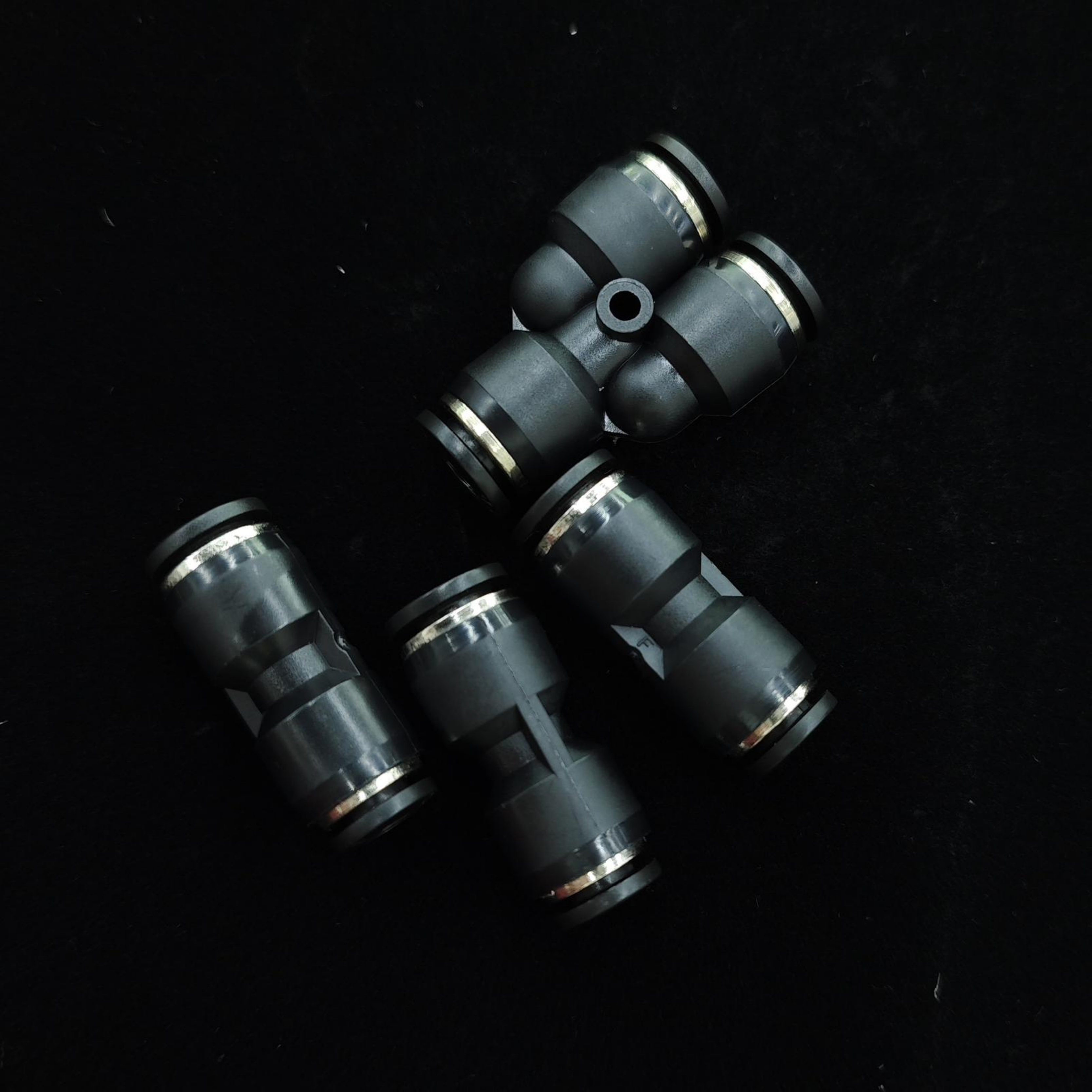}
    }
    \parbox[t][1.5cm][t]{\linewidth}{\small   mixed with Y-type 3-way}
\end{minipage}\hfill
\begin{minipage}{0.23\textwidth}
    \centering
    \parbox[c][3.0cm][c]{\linewidth}{
        \centering\includegraphics[width=\linewidth,height=3.0cm,keepaspectratio]{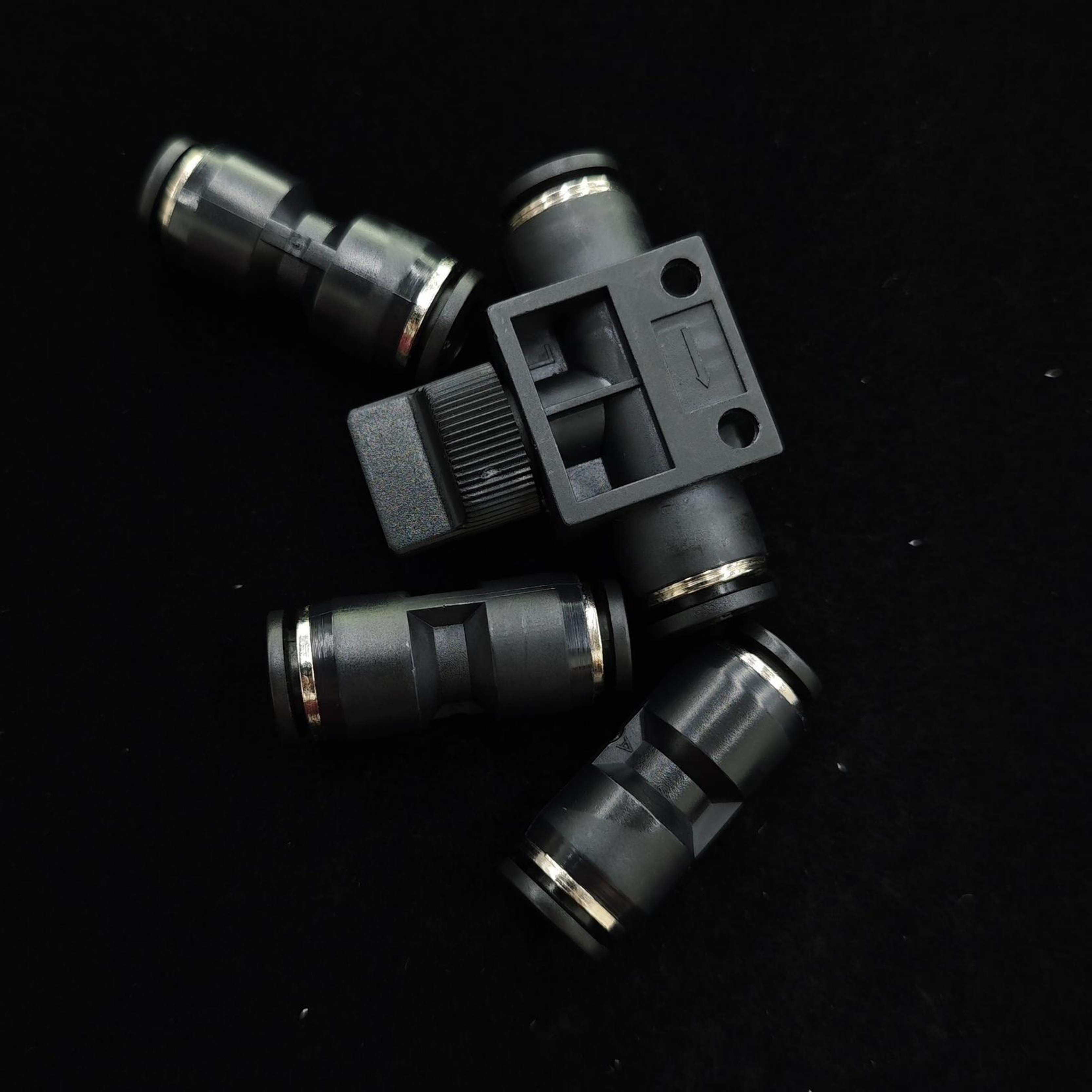}
    }
    \parbox[t][1.5cm][t]{\linewidth}{\small   mixed with throttle valve}
\end{minipage}\hfill
\begin{minipage}{0.23\textwidth}
    \centering
    \parbox[c][3.0cm][c]{\linewidth}{
        \centering\includegraphics[width=\linewidth,height=3.0cm,keepaspectratio]{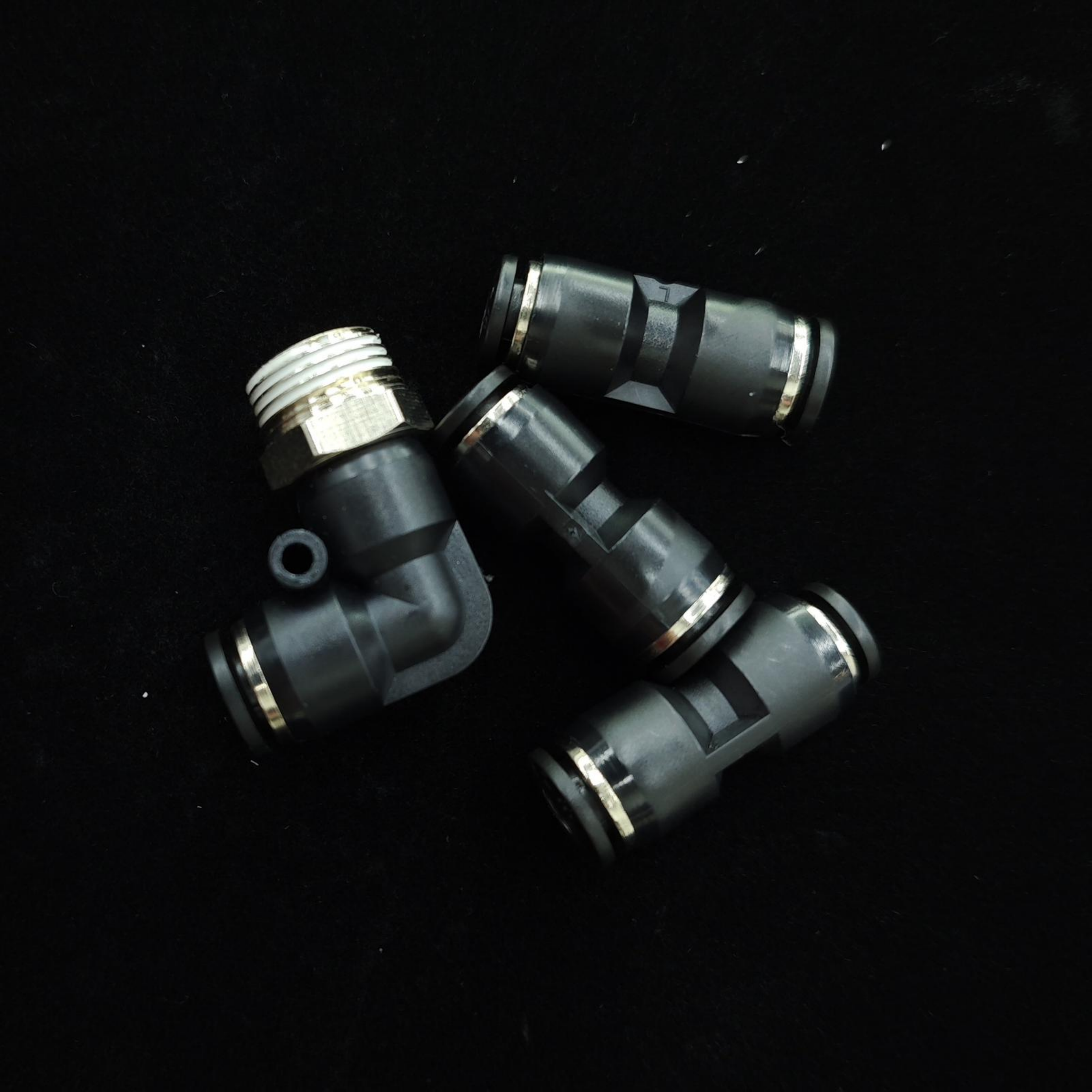}
    }
    \parbox[t][1.5cm][t]{\linewidth}{\small   mixed with external-thread elbow}
\end{minipage}\hfill
\begin{minipage}{0.23\textwidth}
    \centering
    \parbox[c][3.0cm][c]{\linewidth}{
        \centering\includegraphics[width=\linewidth,height=3.0cm,keepaspectratio]{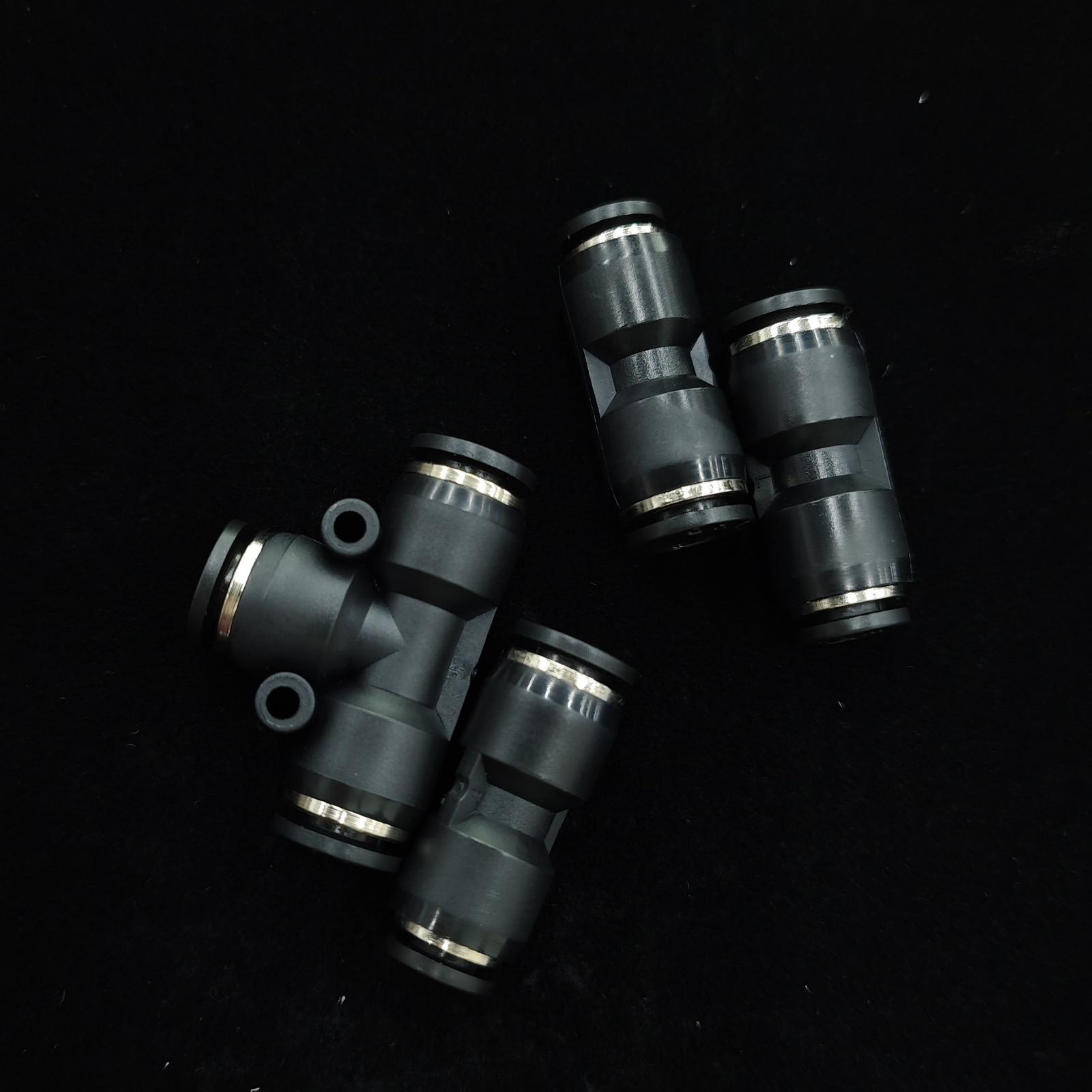}
    }
    \parbox[t][1.5cm][t]{\linewidth}{\small   mixed with T-type 3-way}
\end{minipage}

\caption{Random examples from \pcsscen}
\label{fig: Manufacturing_task1_scenario_3_saved}
\end{figure}

\begin{figure}[htbp]
\centering

\begin{minipage}{0.48\textwidth}
    \centering
    \includegraphics[width=\linewidth]{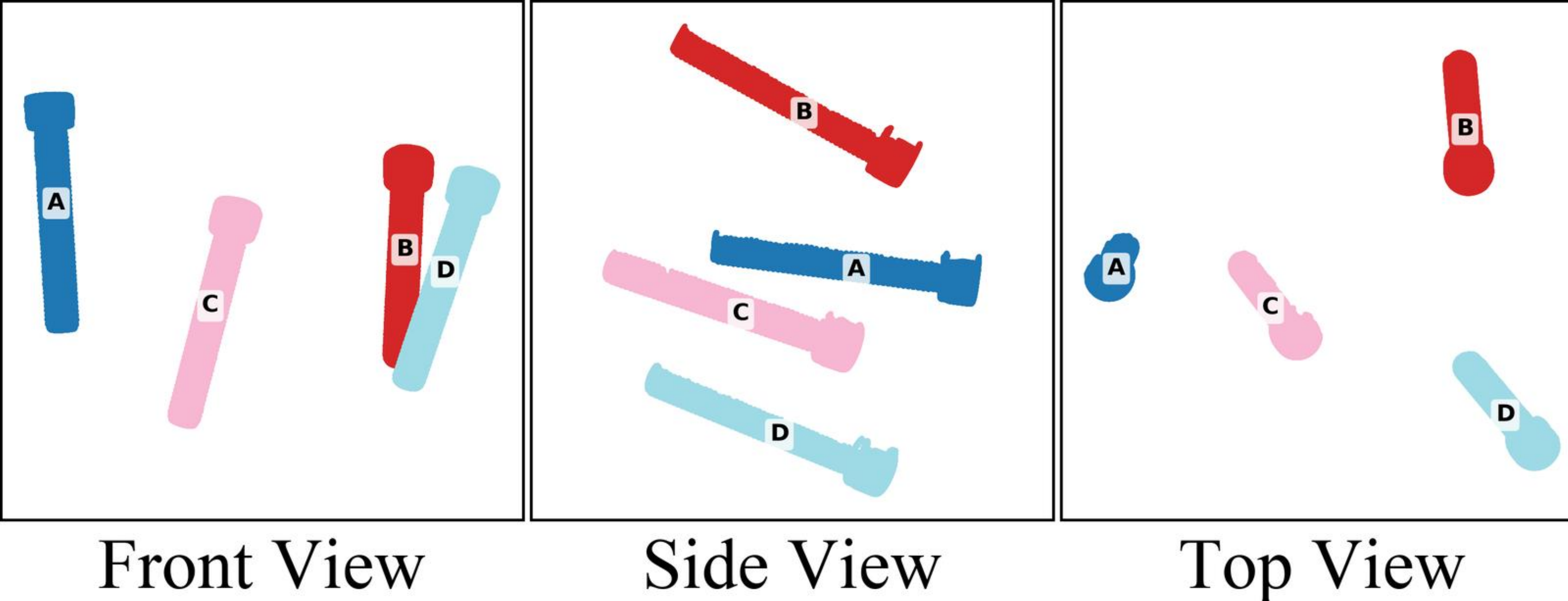}\\
    \small Normal Case 
\end{minipage}\hfill
\begin{minipage}{0.48\textwidth}
    \centering
    \includegraphics[width=\linewidth]{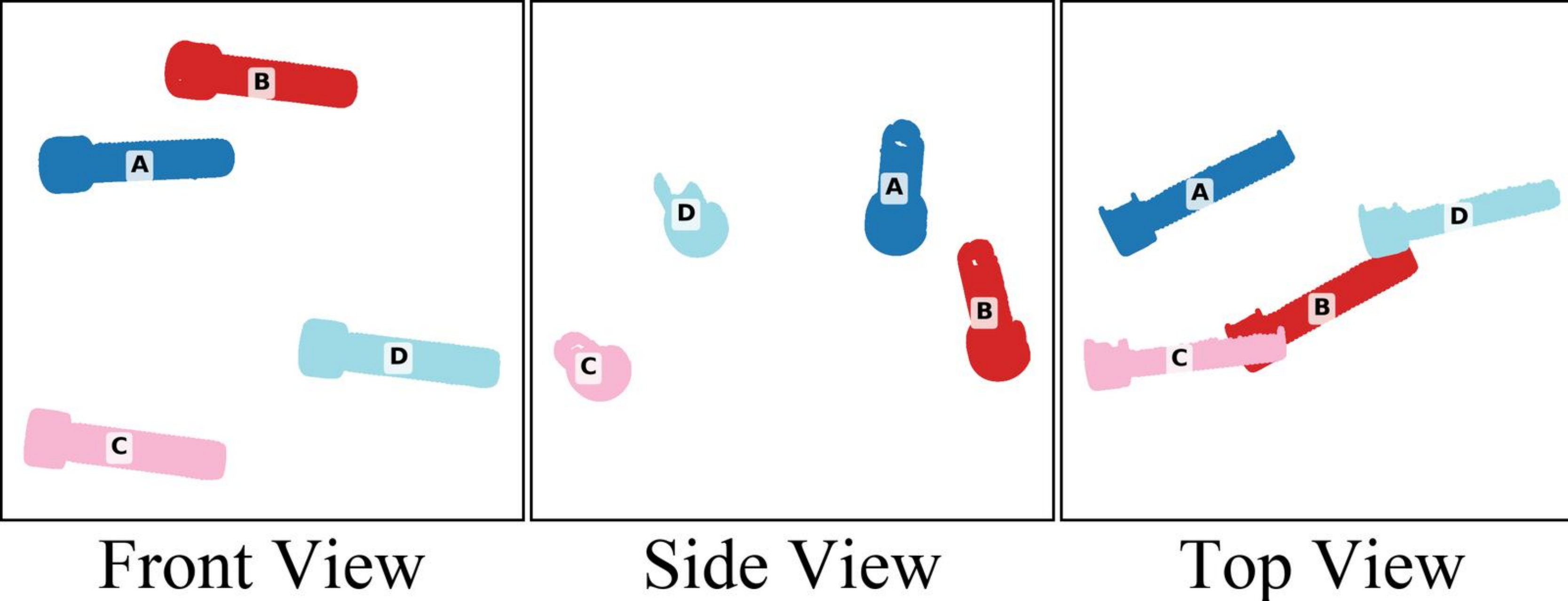}\\
    \small Normal Case 
\end{minipage}

\begin{minipage}{0.48\textwidth}
    \centering
    \includegraphics[width=\linewidth]{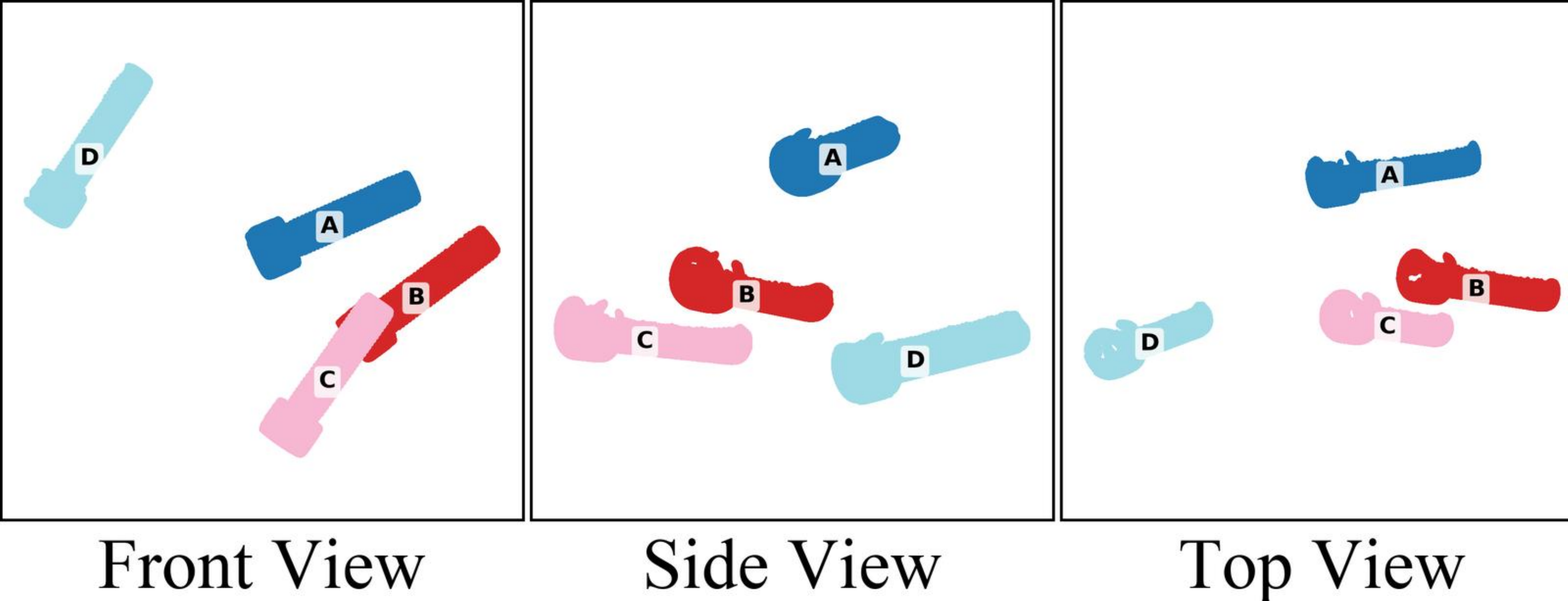}\\
    \small Normal Case 
\end{minipage}\hfill
\begin{minipage}{0.48\textwidth}
    \centering
    \includegraphics[width=\linewidth]{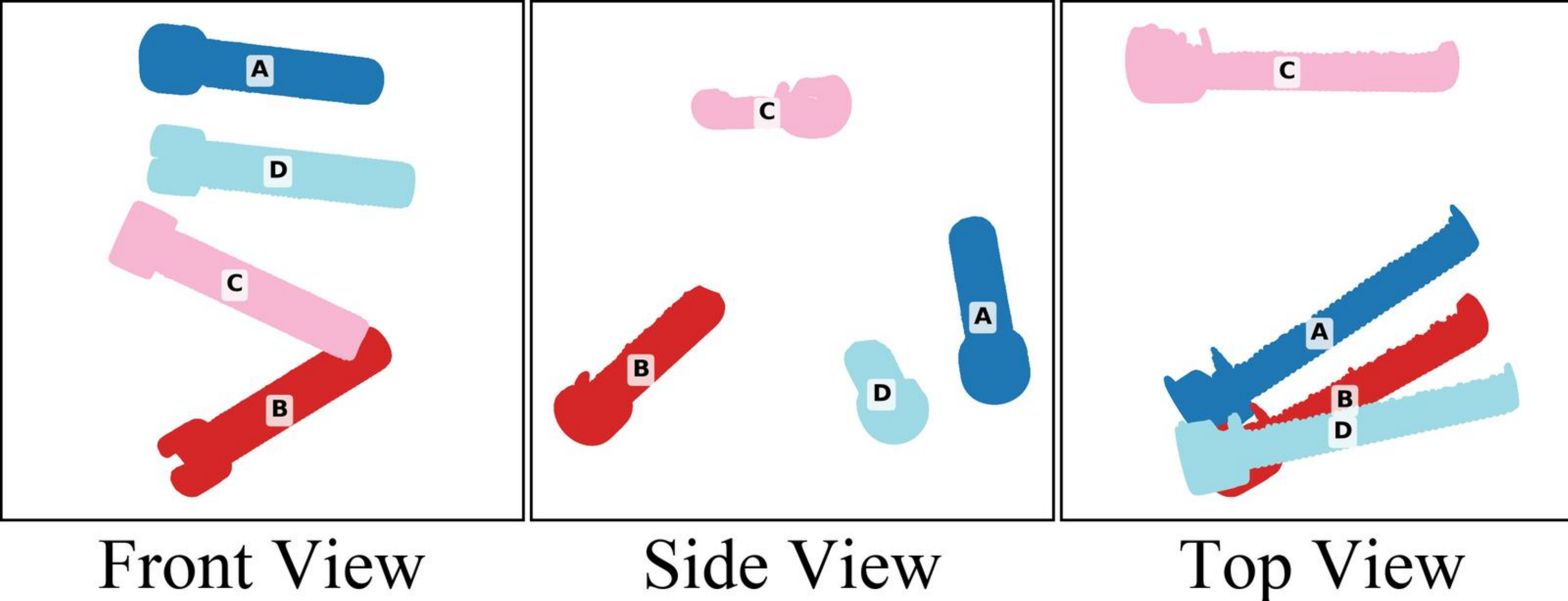}\\
    \small Normal Case 
\end{minipage}

\begin{minipage}{0.48\textwidth}
    \centering
    \includegraphics[width=\linewidth]{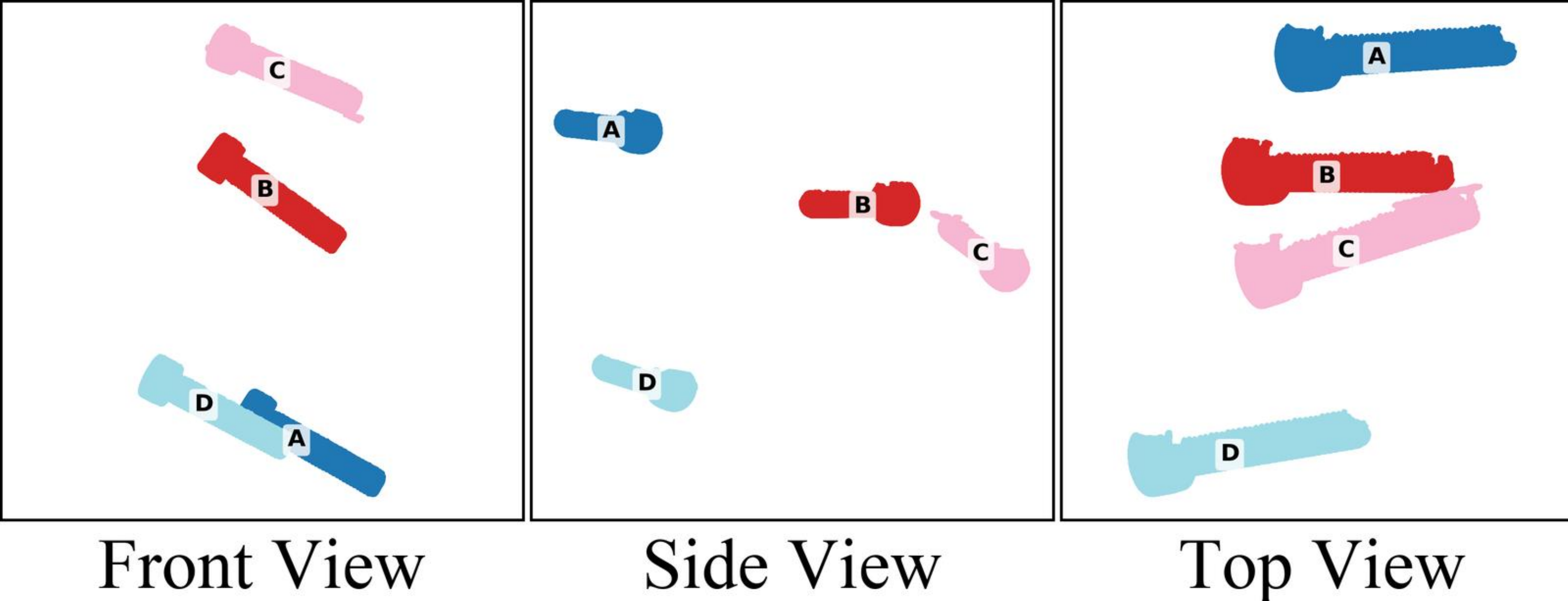}\\
    \small Normal Case 
\end{minipage}\hfill
\begin{minipage}{0.48\textwidth}
    \centering
    \includegraphics[width=\linewidth]{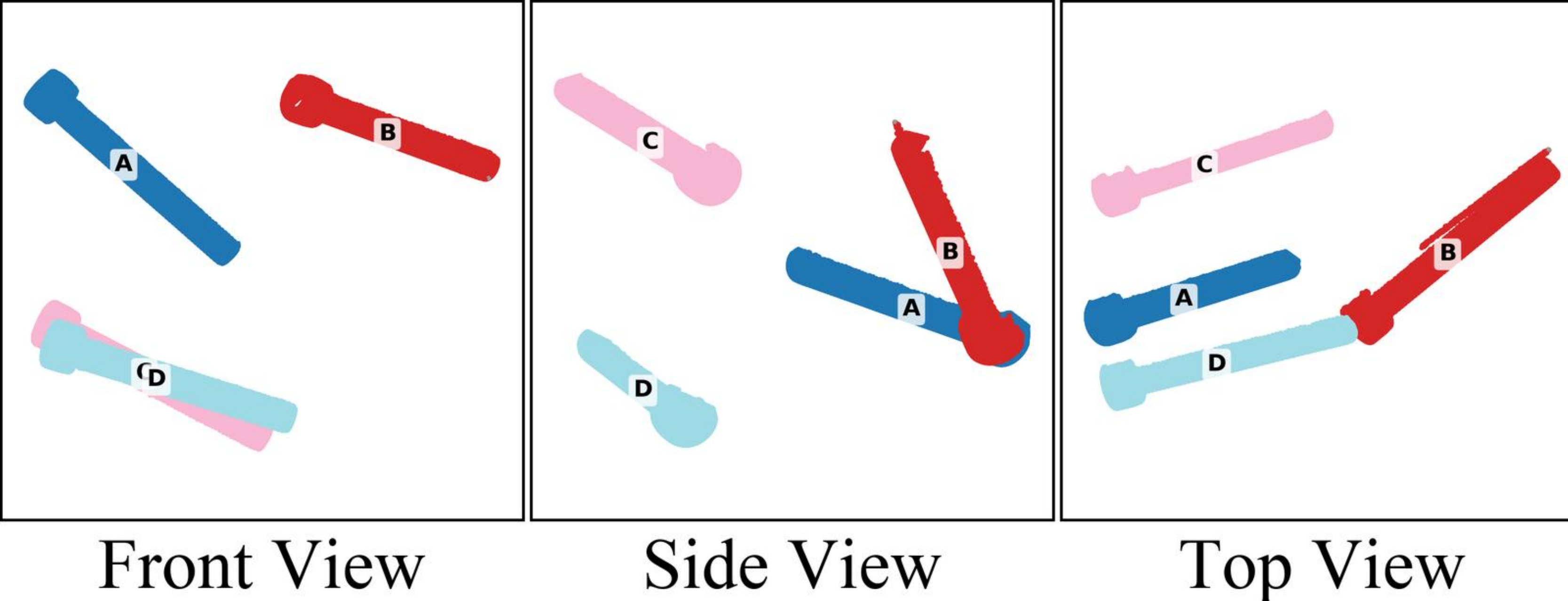}\\
    \small Normal Case 
\end{minipage}

\begin{minipage}{0.48\textwidth}
    \centering
    \includegraphics[width=\linewidth]{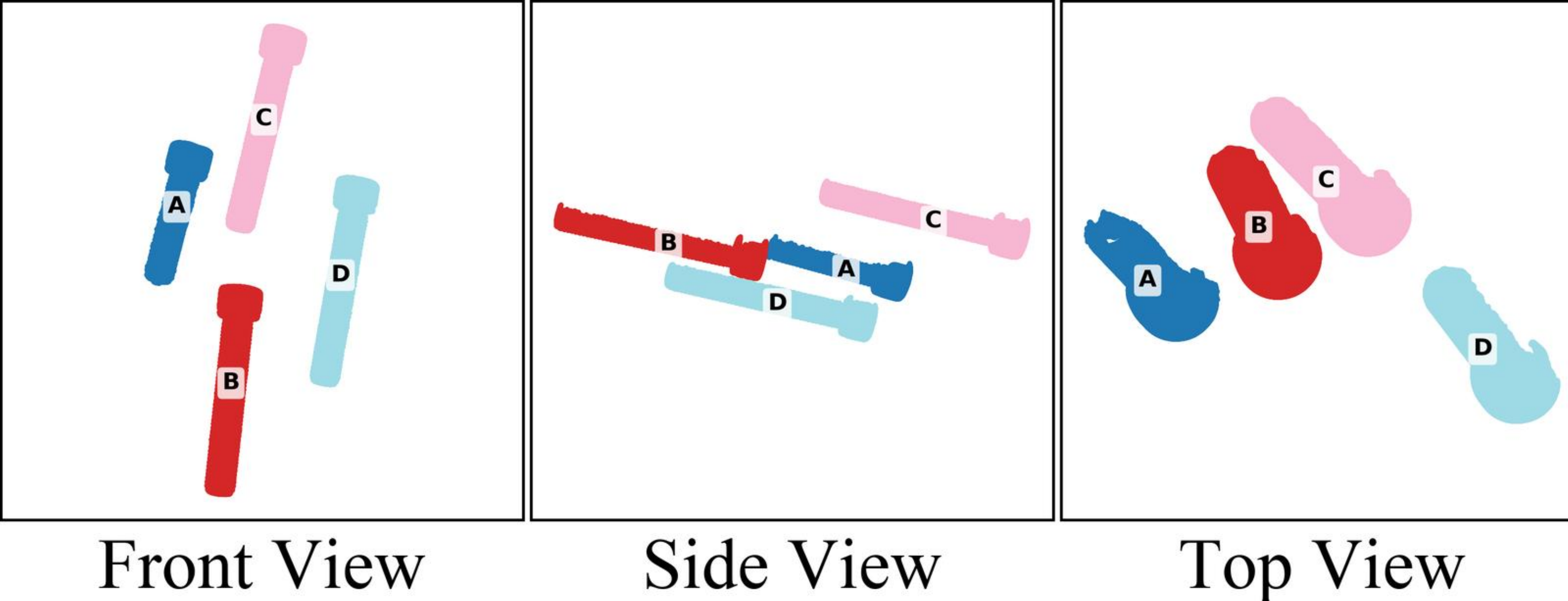}\\
    \small   M12  shorter length 50
\end{minipage}\hfill
\begin{minipage}{0.48\textwidth}
    \centering
    \includegraphics[width=\linewidth]{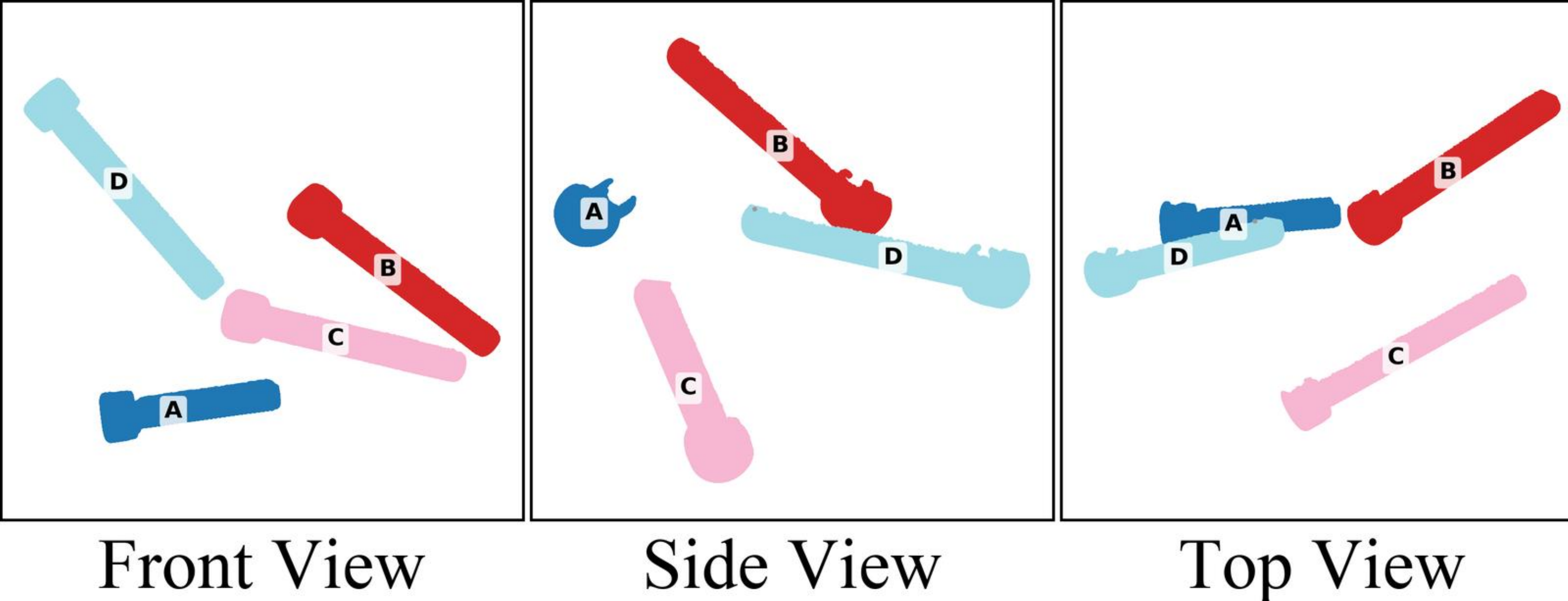}\\
    \small   M10  shorter length 50
\end{minipage}

\begin{minipage}{0.48\textwidth}
    \centering
    \includegraphics[width=\linewidth]{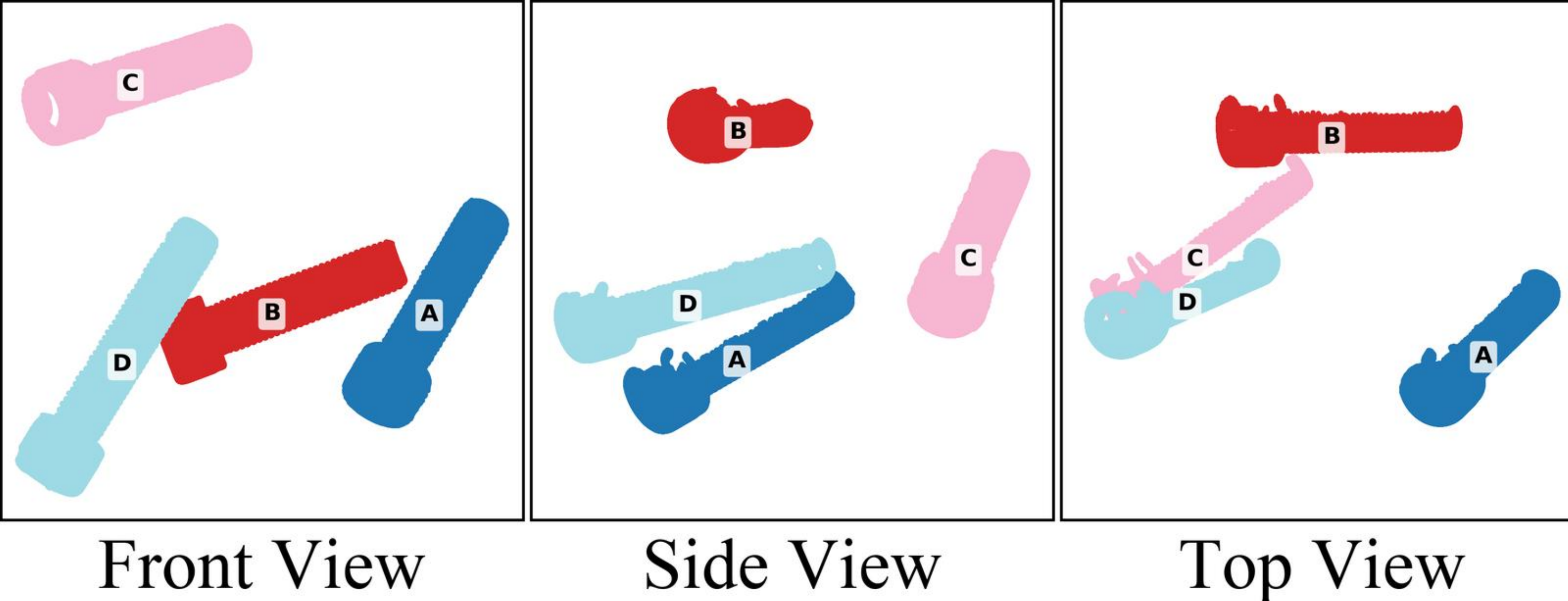}\\
    \small   M20 longer length 100  
\end{minipage}\hfill
\begin{minipage}{0.48\textwidth}
    \centering
    \includegraphics[width=\linewidth]{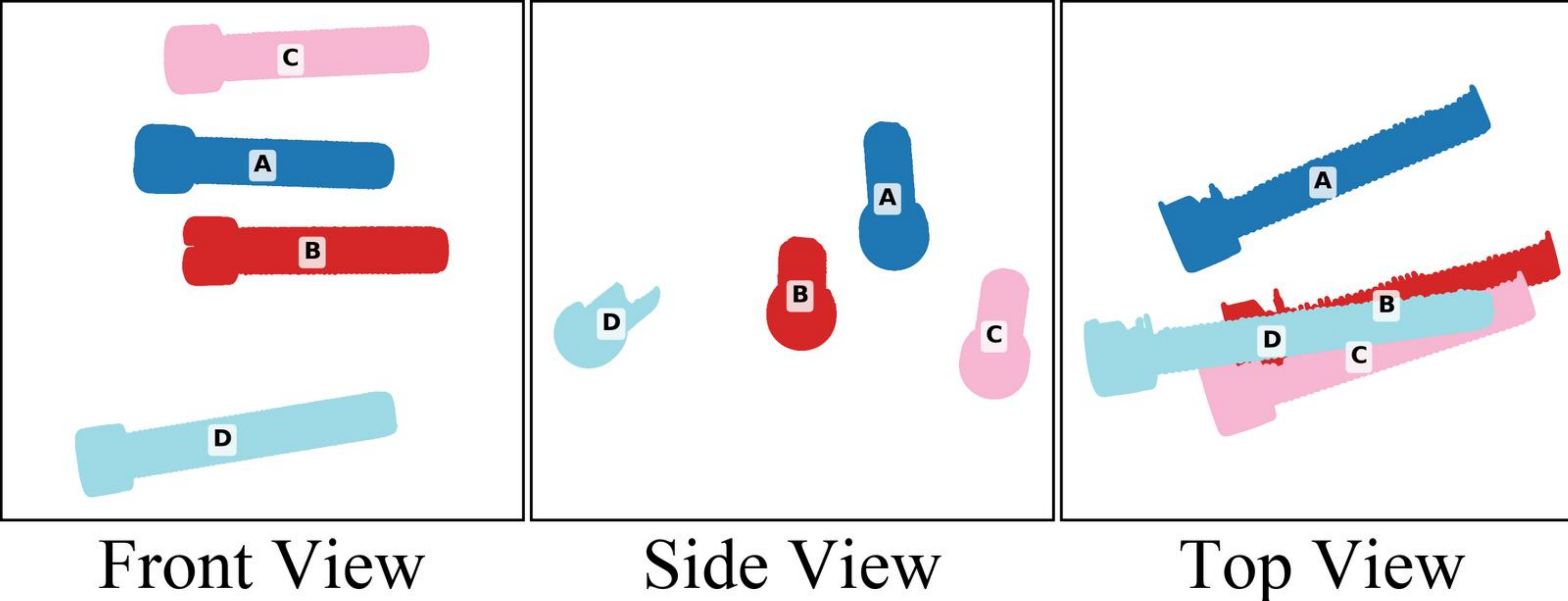}\\
    \small   M16 longer length 100  
\end{minipage}

\begin{minipage}{0.48\textwidth}
    \centering
    \includegraphics[width=\linewidth]{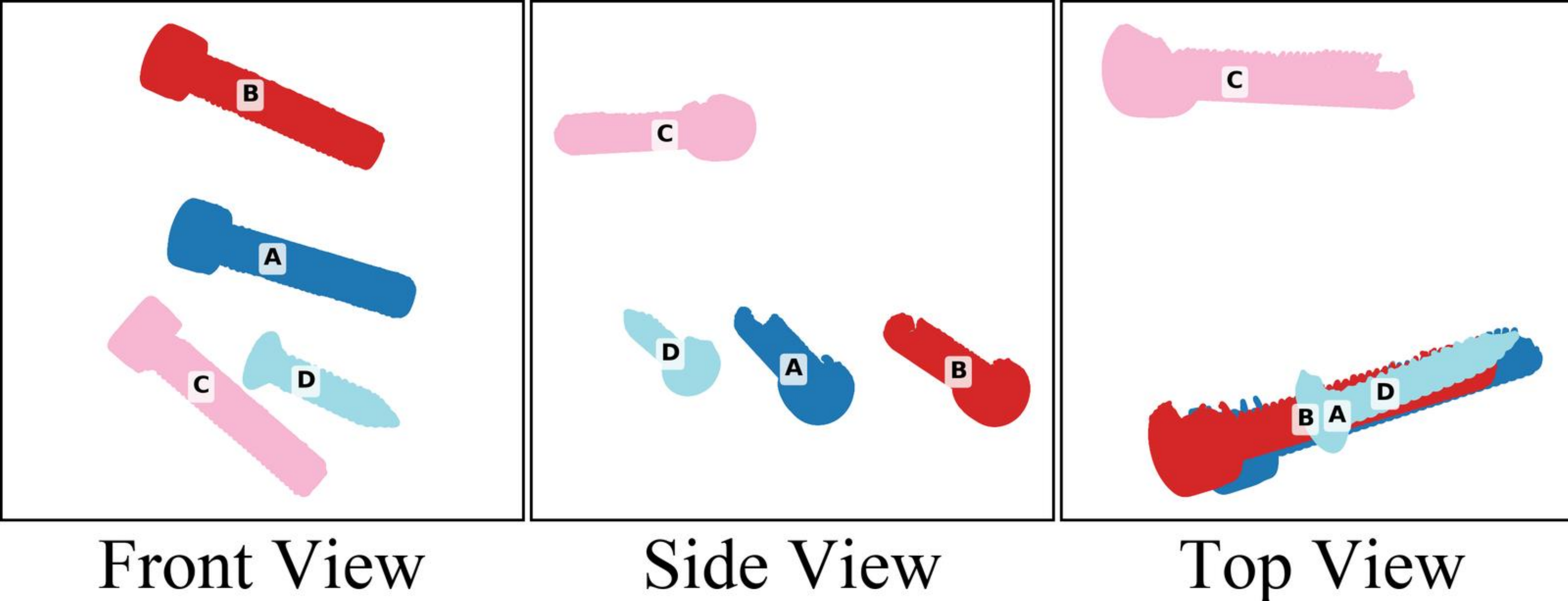}\\
    \small   Self-tapping Screw 
\end{minipage}\hfill
\begin{minipage}{0.48\textwidth}
    \centering
    \includegraphics[width=\linewidth]{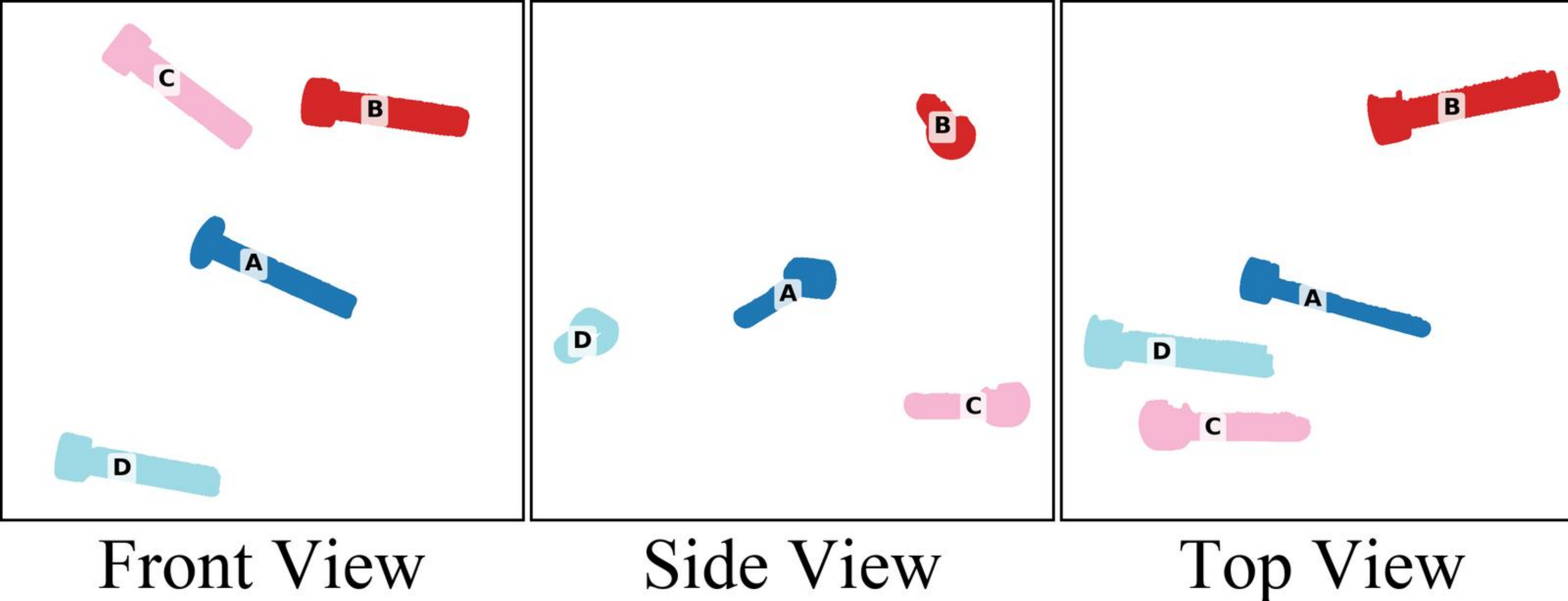}\\
    \small   T Bolt Half thread Screw 
\end{minipage}

\begin{minipage}{0.48\textwidth}
    \centering
    \includegraphics[width=\linewidth]{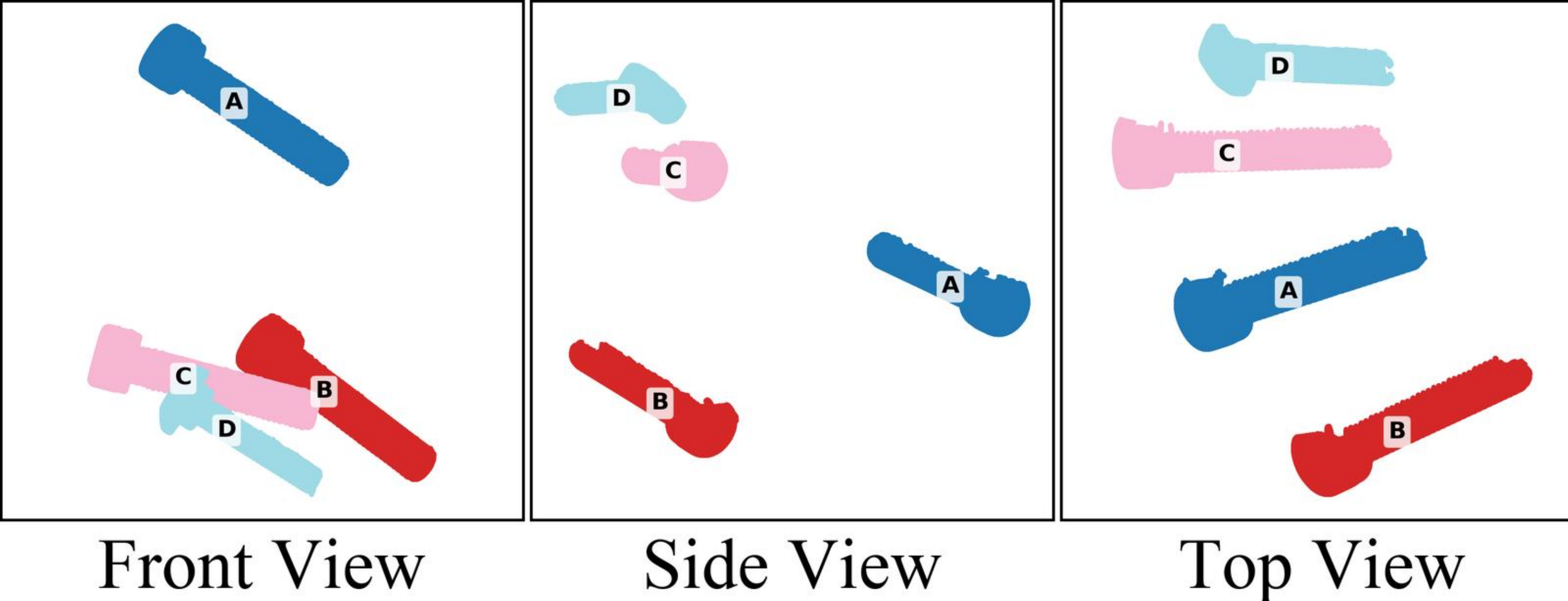}\\
    \small   T Screw 
\end{minipage}\hfill
\begin{minipage}{0.48\textwidth}
    \centering
    \includegraphics[width=\linewidth]{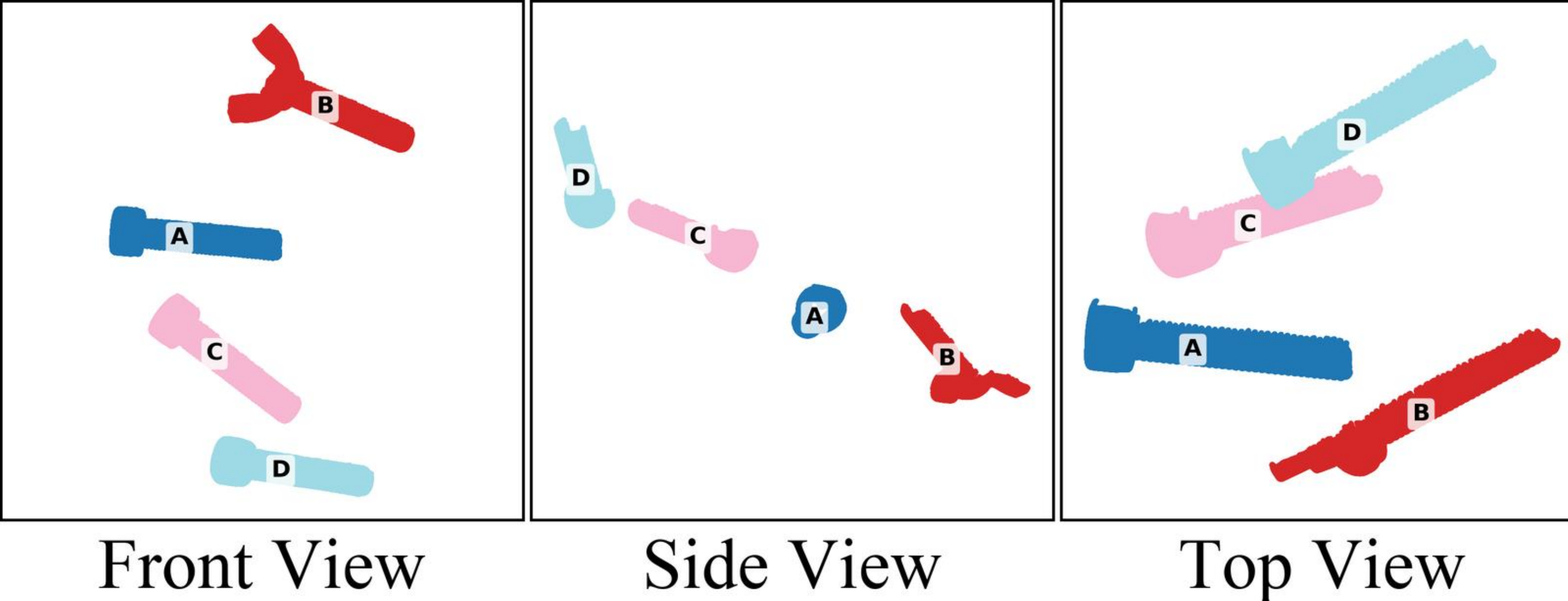}\\
    \small   Wing Screw 
\end{minipage}

\caption{Random examples from \chsscen}
\label{fig: Manufacturing_task1_scenario_1_pointcloud_saved}
\end{figure}

\begin{figure}[htbp]
\centering

\begin{minipage}{0.48\textwidth}
    \centering
    \includegraphics[width=\linewidth]{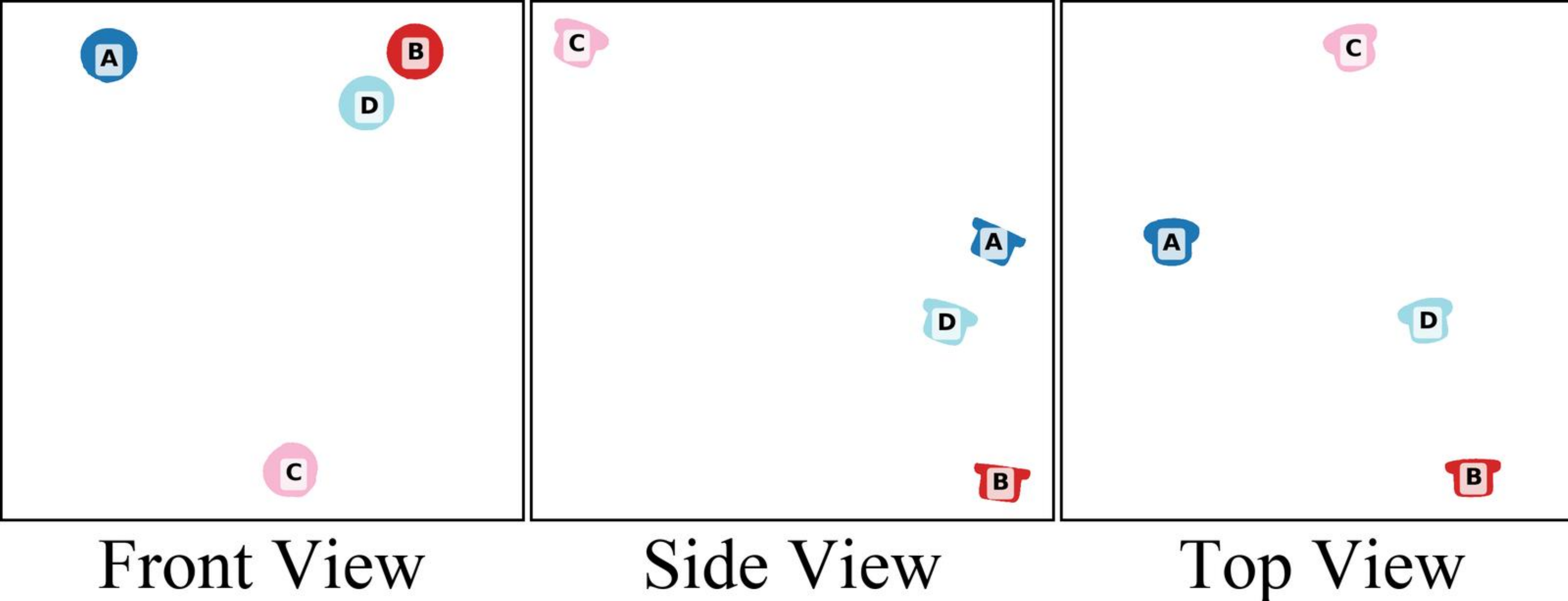}\\
    \small Normal Case: Flange Nut
\end{minipage}\hfill
\begin{minipage}{0.48\textwidth}
    \centering
    \includegraphics[width=\linewidth]{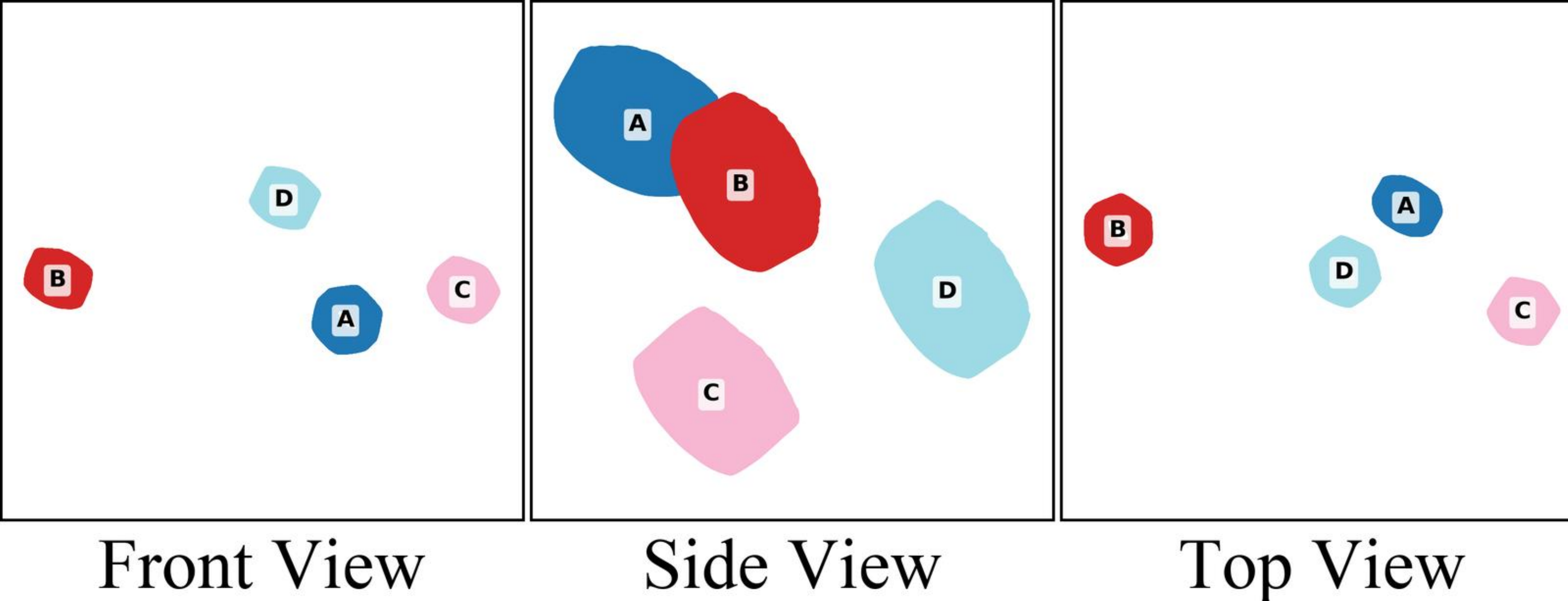}\\
    \small Normal Case: M18
\end{minipage}

\begin{minipage}{0.48\textwidth}
    \centering
    \includegraphics[width=\linewidth]{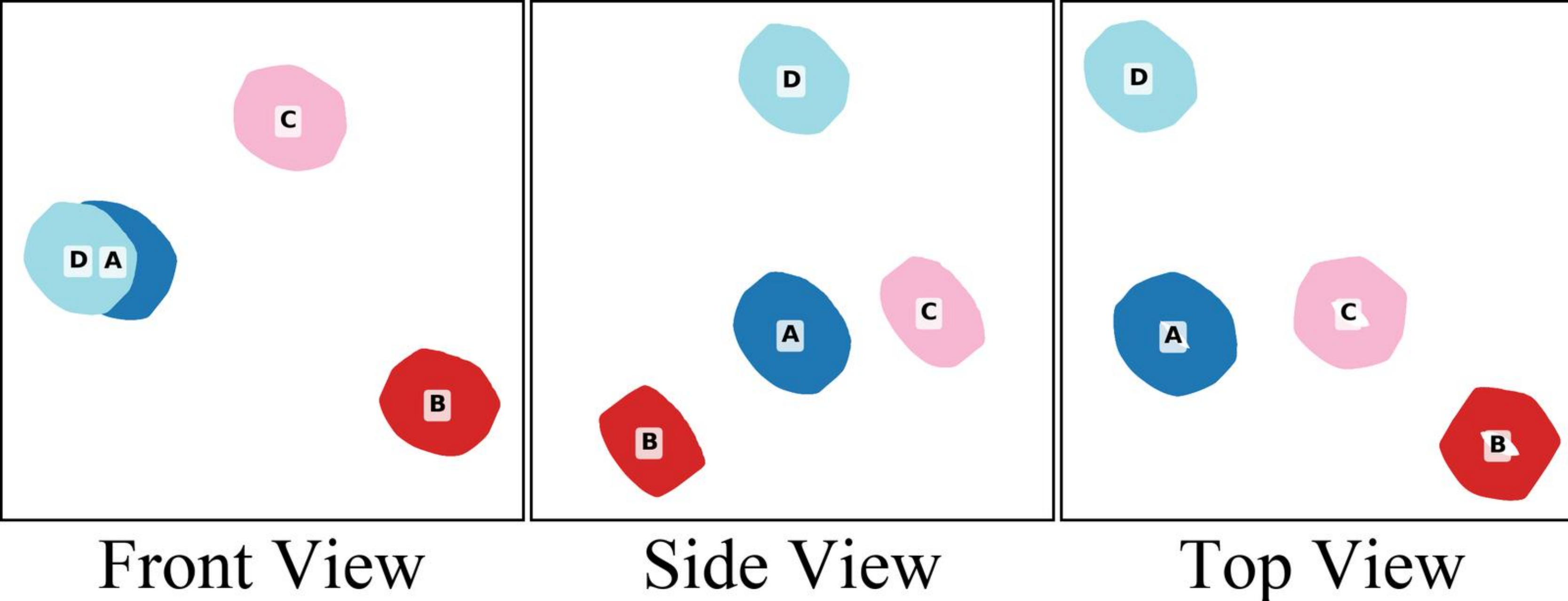}\\
    \small   Mixed with M20
\end{minipage}\hfill
\begin{minipage}{0.48\textwidth}
    \centering
    \includegraphics[width=\linewidth]{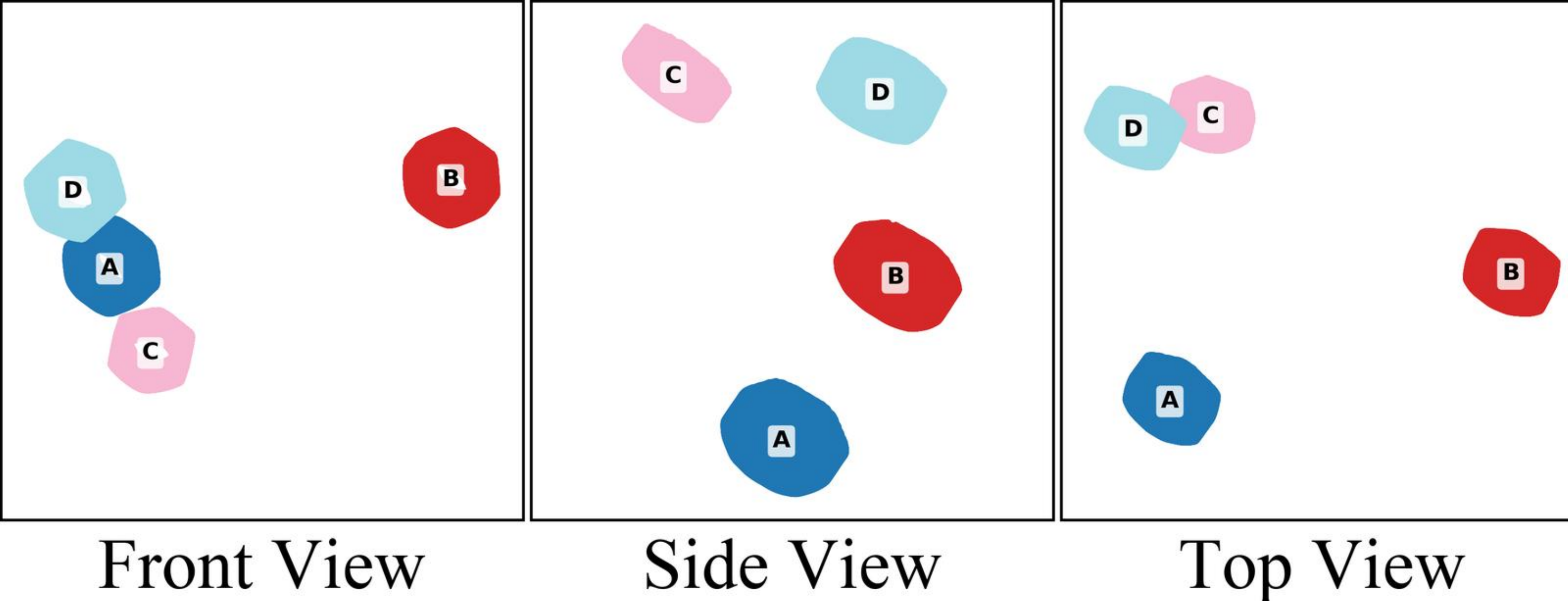}\\
    \small   Mixed with M16
\end{minipage}

\begin{minipage}{0.48\textwidth}
    \centering
    \includegraphics[width=\linewidth]{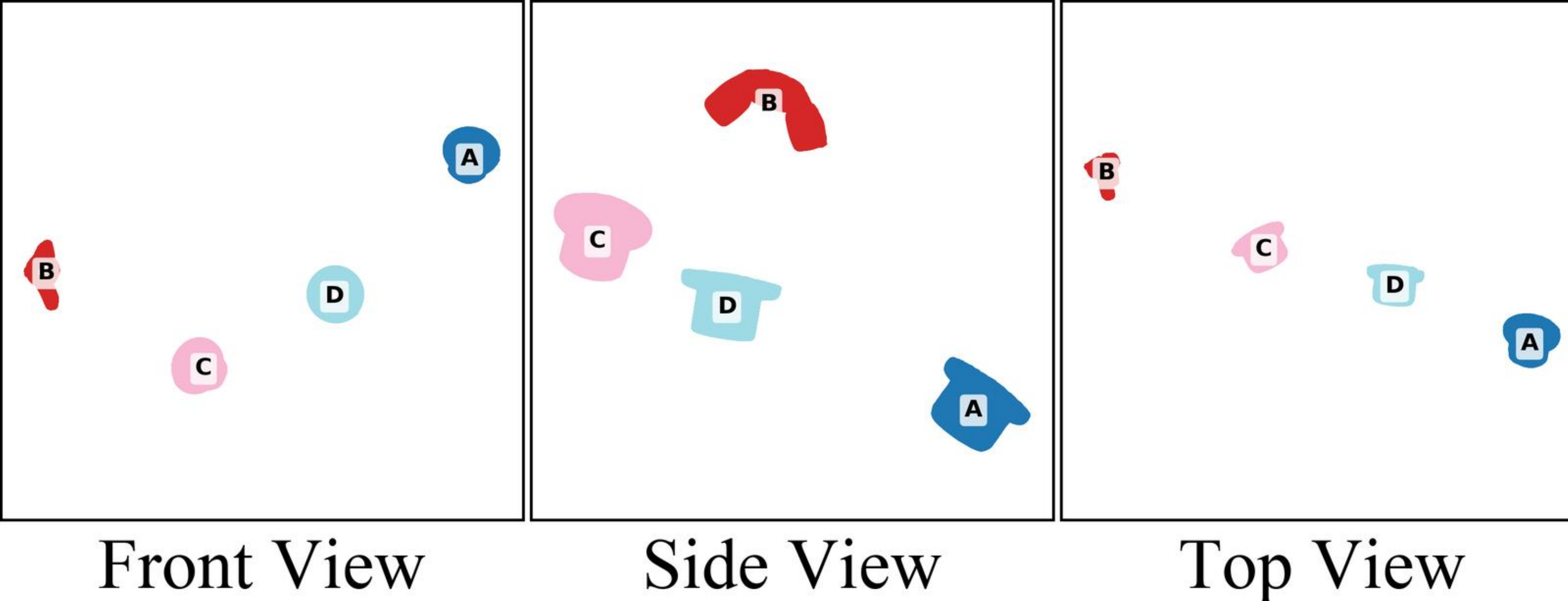}\\
    \small   Wing Nut 
\end{minipage}\hfill
\begin{minipage}{0.48\textwidth}
    \centering
    \includegraphics[width=\linewidth]{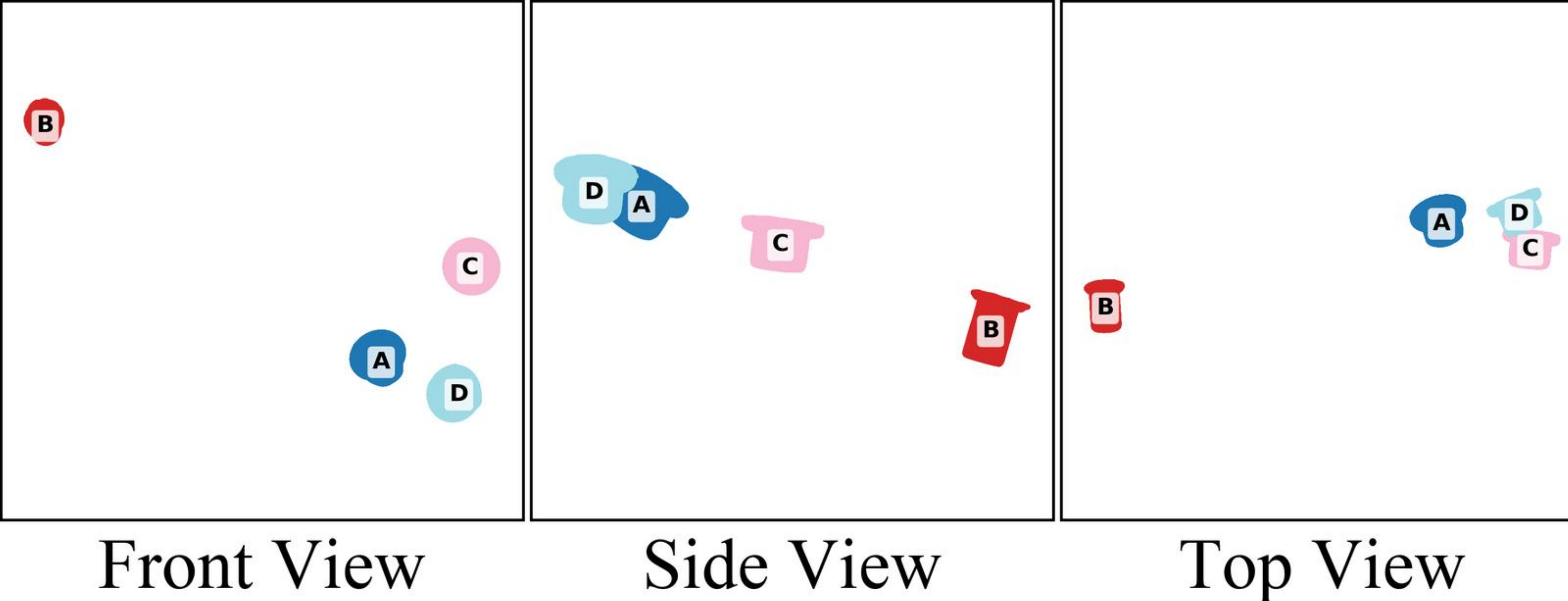}\\
    \small   Rivet Nut 
\end{minipage}

\begin{minipage}{0.48\textwidth}
    \centering
    \includegraphics[width=\linewidth]{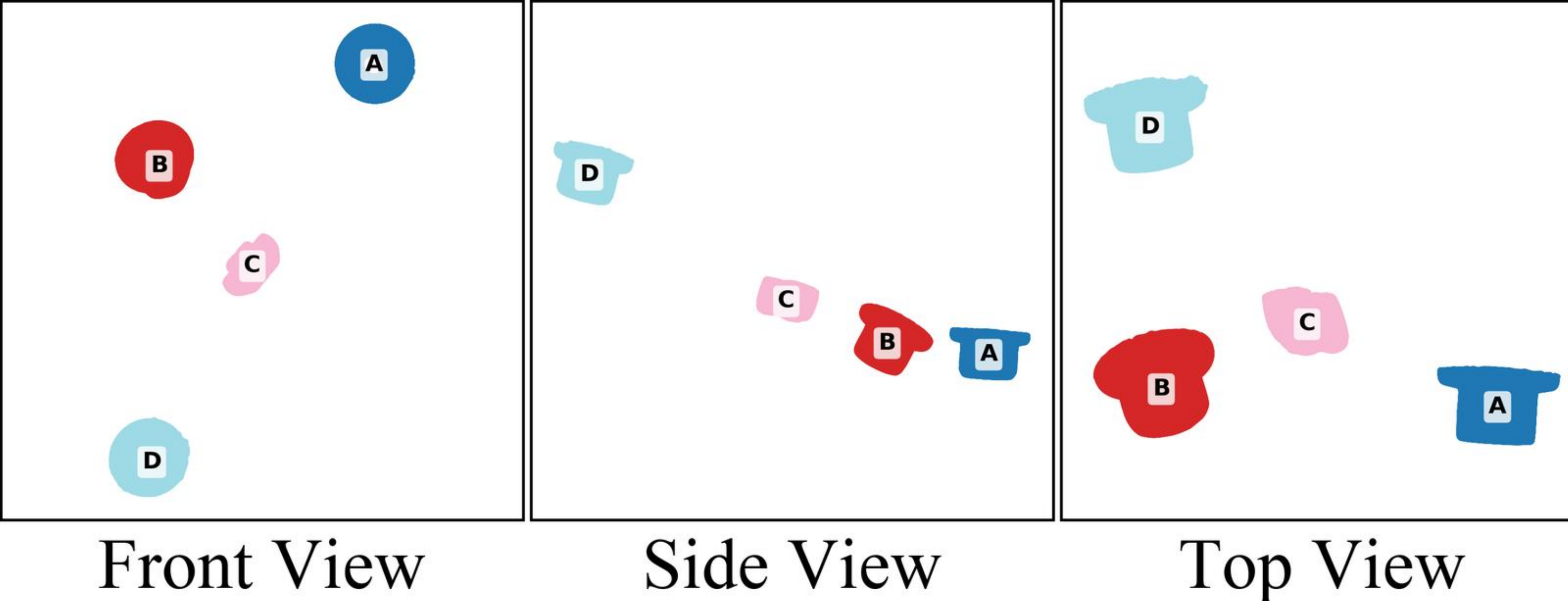}\\
    \small   T Nut 
\end{minipage}

\caption{Random examples from \nutsscen}
\label{fig: Manufacturing_task1_scenario_2_pointcloud_saved}
\end{figure}

\begin{figure}[htbp]
\centering

\begin{minipage}{0.48\textwidth}
    \centering
    \includegraphics[width=\linewidth]{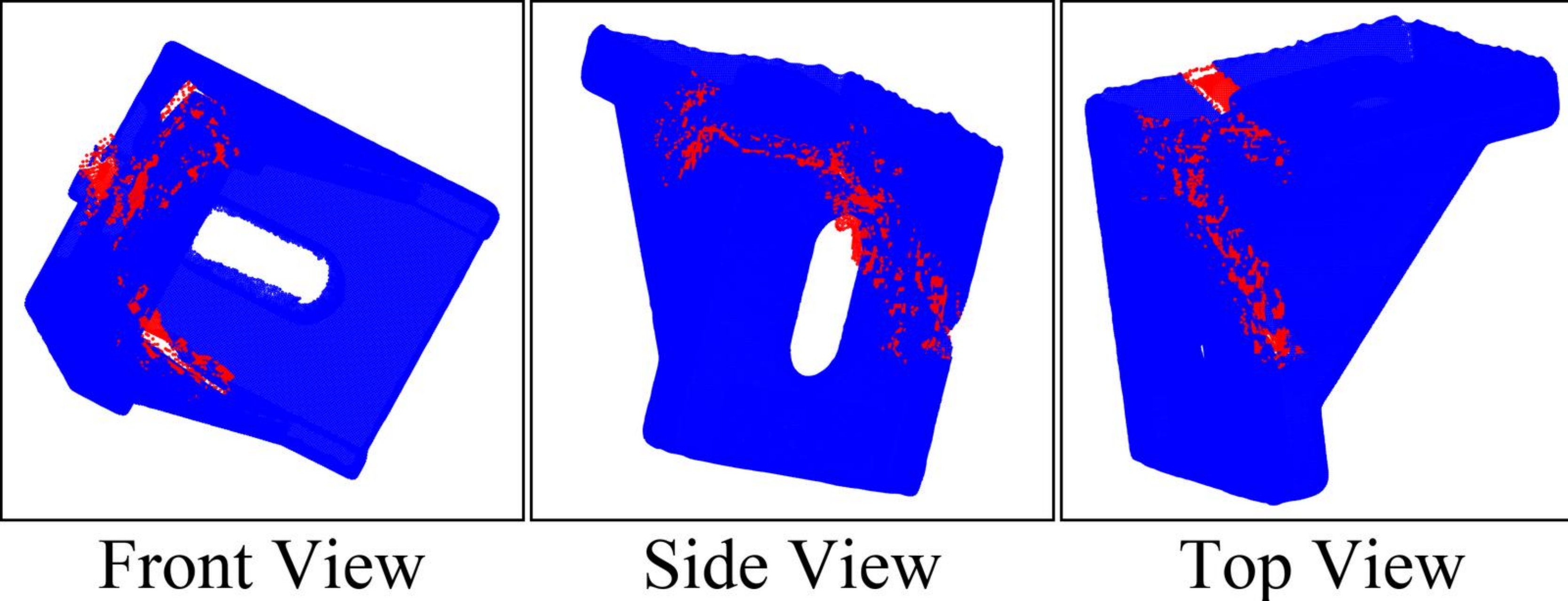}\\
    \small Corner Bracket crack
\end{minipage}\hfill
\begin{minipage}{0.48\textwidth}
    \centering
    \includegraphics[width=\linewidth]{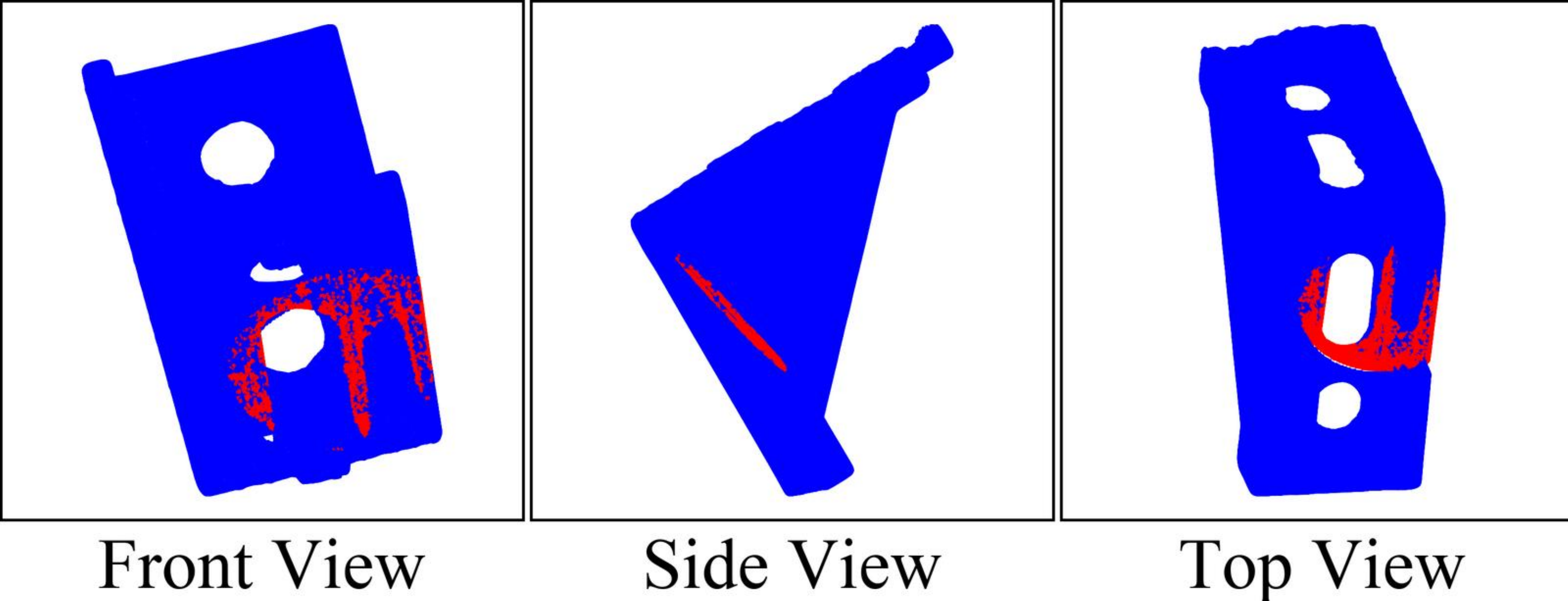}\\
    \small Corner Bracket cut
\end{minipage}

\begin{minipage}{0.48\textwidth}
    \centering
    \includegraphics[width=\linewidth]{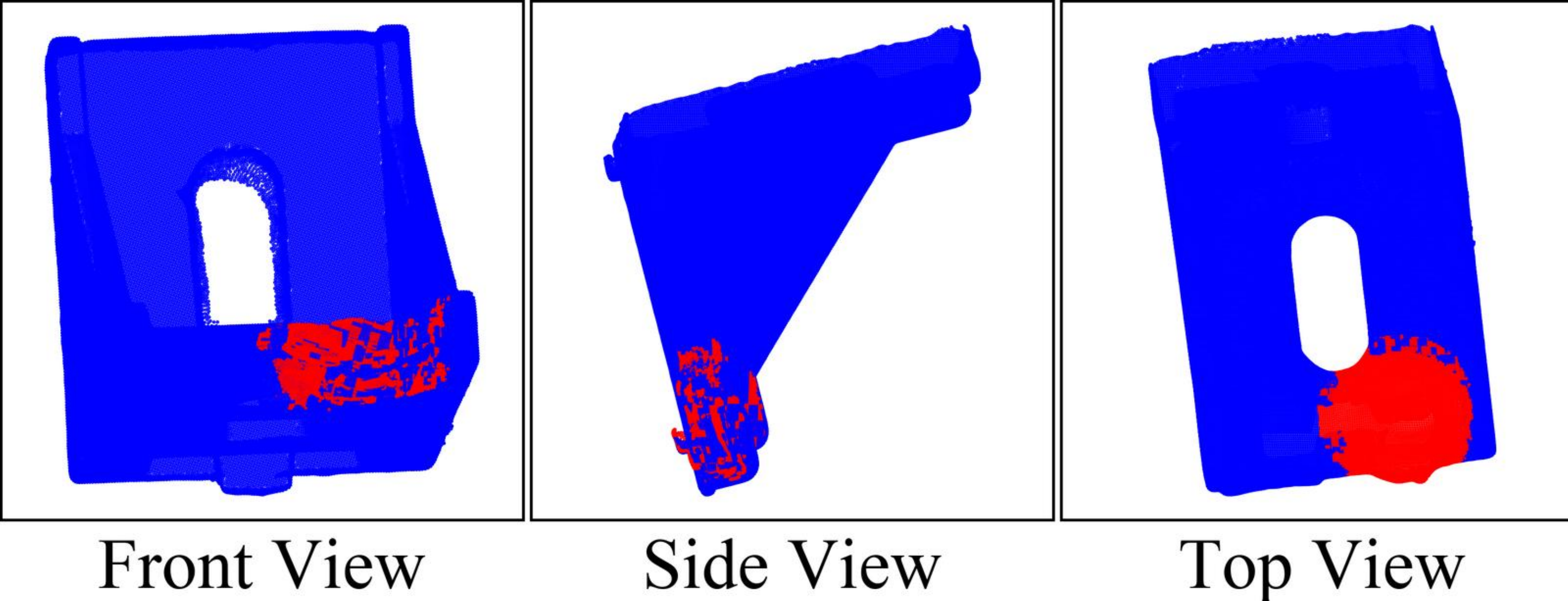}\\
    \small Corner Bracket deformation
\end{minipage}\hfill
\begin{minipage}{0.48\textwidth}
    \centering
    \includegraphics[width=\linewidth]{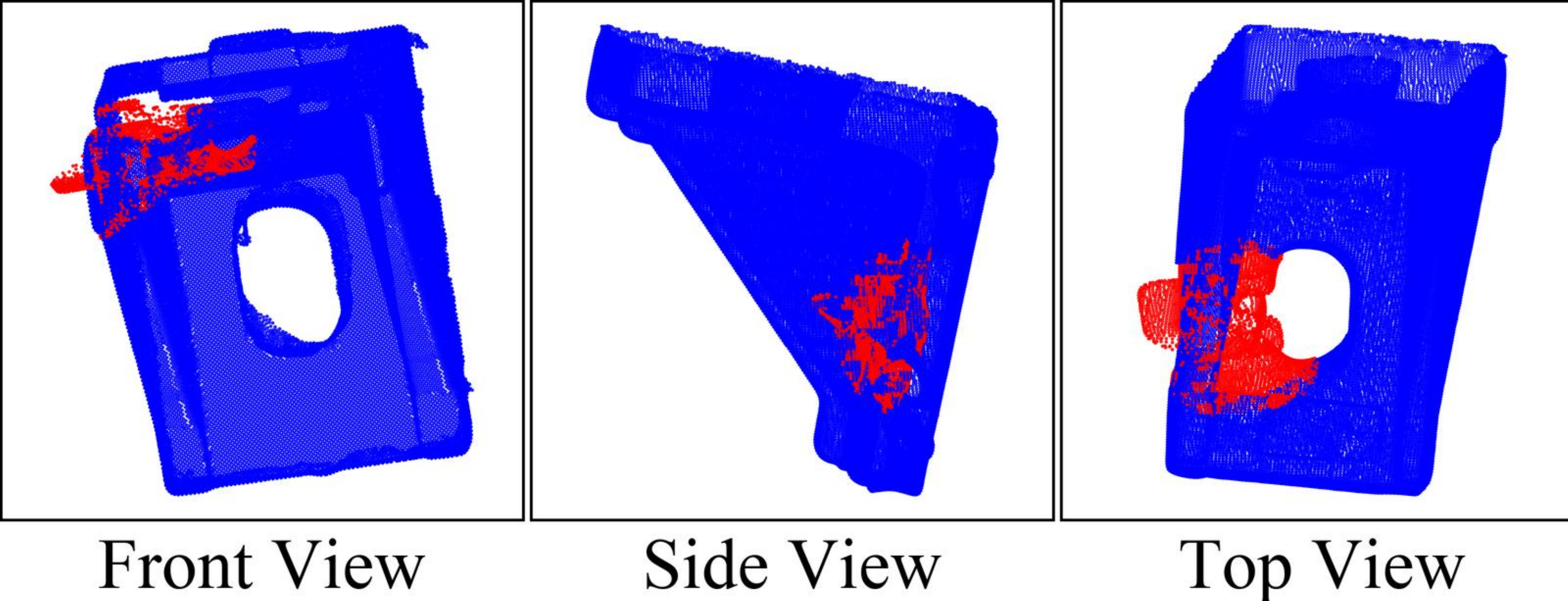}\\
    \small Corner Bracket dent
\end{minipage}

\caption{Random examples from Corner Bracket}
\label{fig: Corner_Bracket_saved}
\end{figure}

\begin{figure}[htbp]
\centering

\begin{minipage}{0.48\textwidth}
    \centering
    \includegraphics[width=\linewidth]{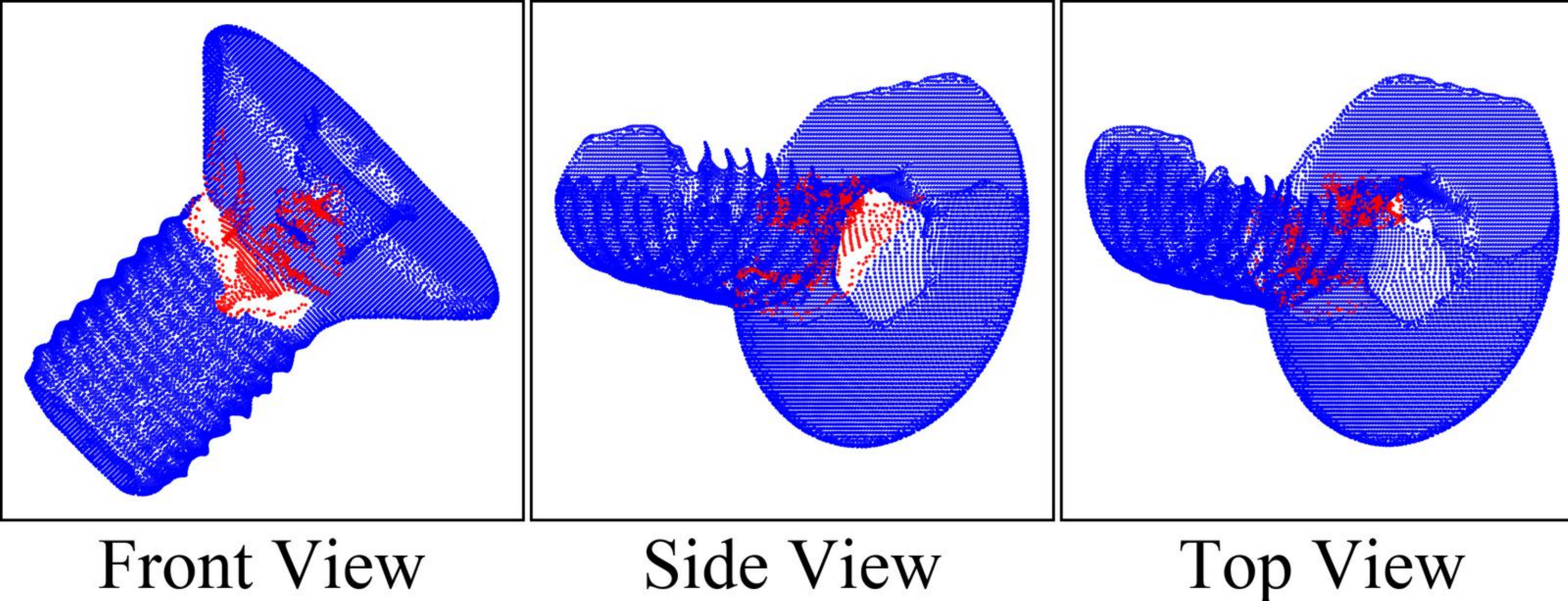}\\
    \small Countersunk Screw crack
\end{minipage}\hfill
\begin{minipage}{0.48\textwidth}
    \centering
    \includegraphics[width=\linewidth]{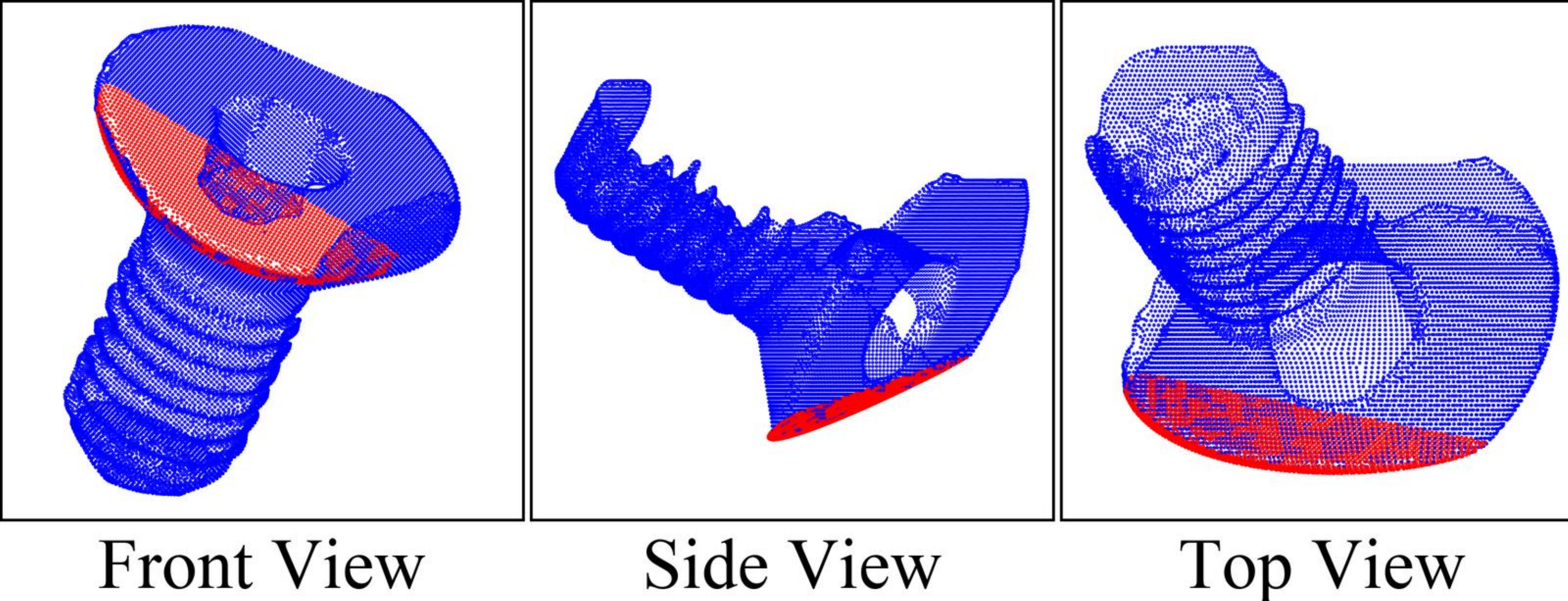}\\
    \small Countersunk Screw cut
\end{minipage}

\begin{minipage}{0.48\textwidth}
    \centering
    \includegraphics[width=\linewidth]{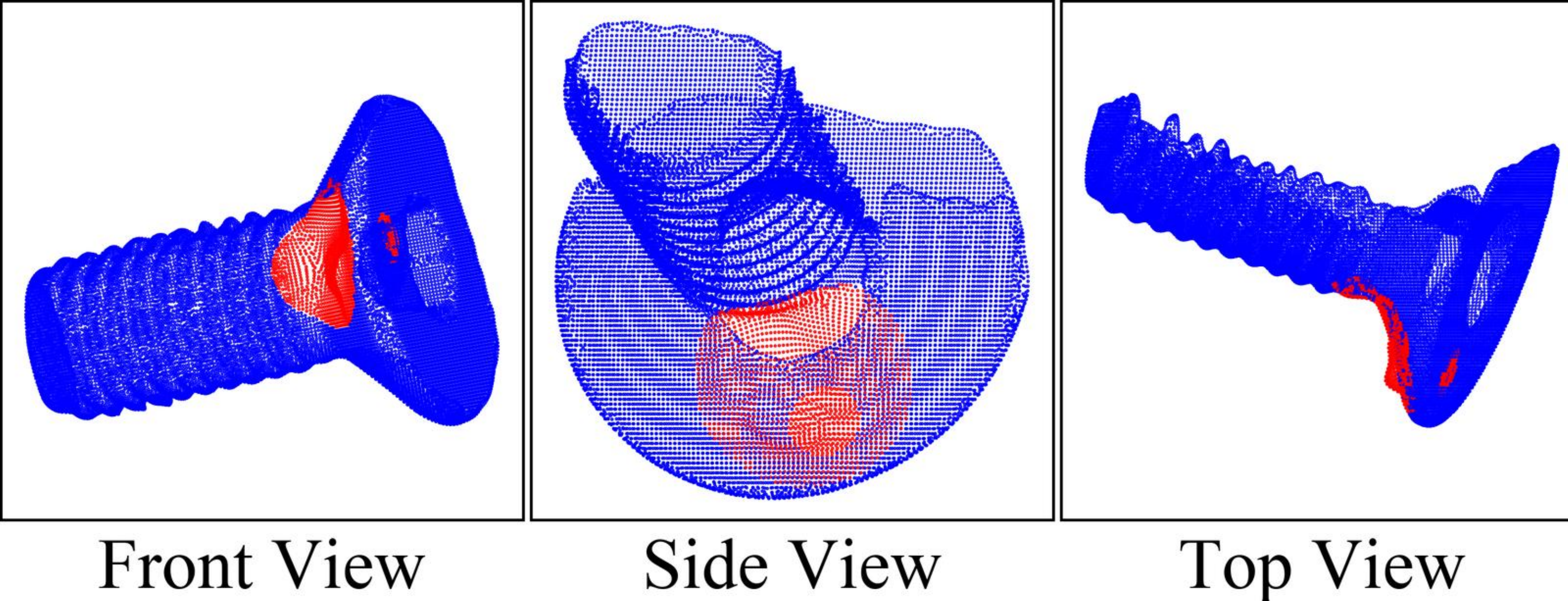}\\
    \small Countersunk Screw deformation
\end{minipage}\hfill
\begin{minipage}{0.48\textwidth}
    \centering
    \includegraphics[width=\linewidth]{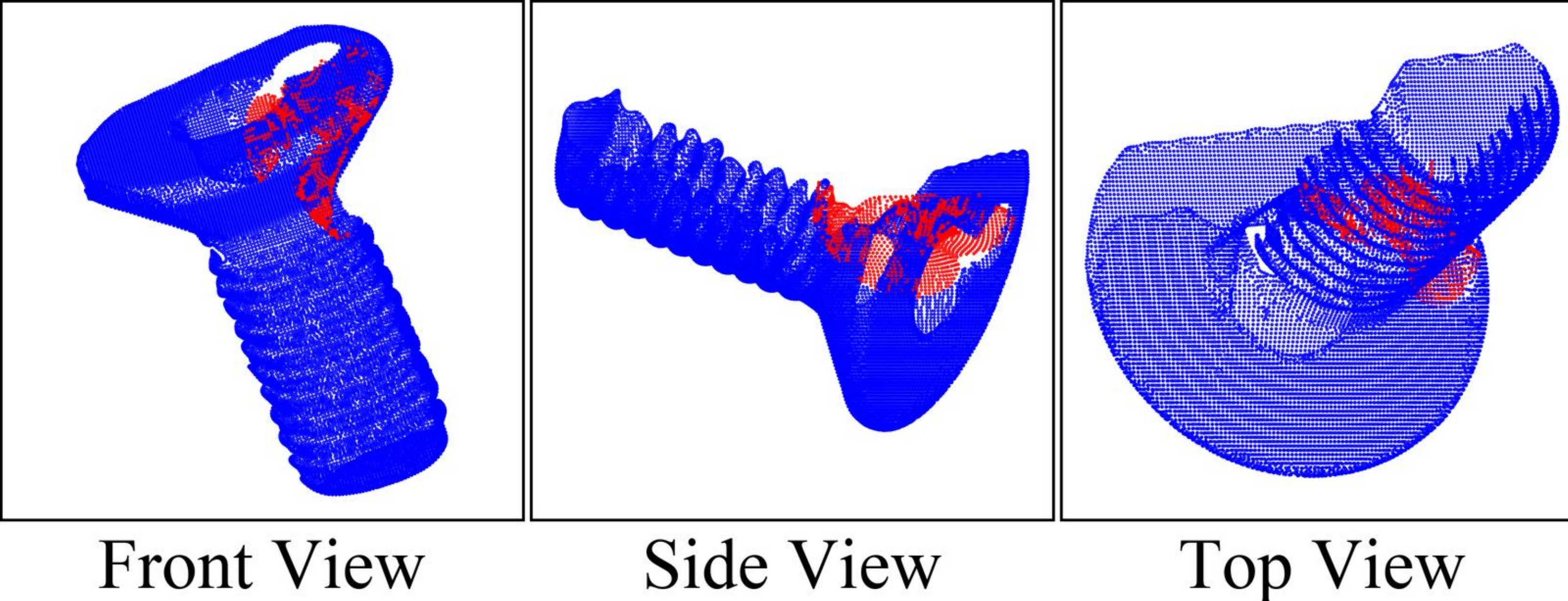}\\
    \small Countersunk Screw dent
\end{minipage}

\caption{Random examples from Countersunk Screw}
\label{fig: Countersunk_Screw_saved}
\end{figure}

\begin{figure}[htbp]
\centering

\begin{minipage}{0.48\textwidth}
    \centering
    \includegraphics[width=\linewidth]{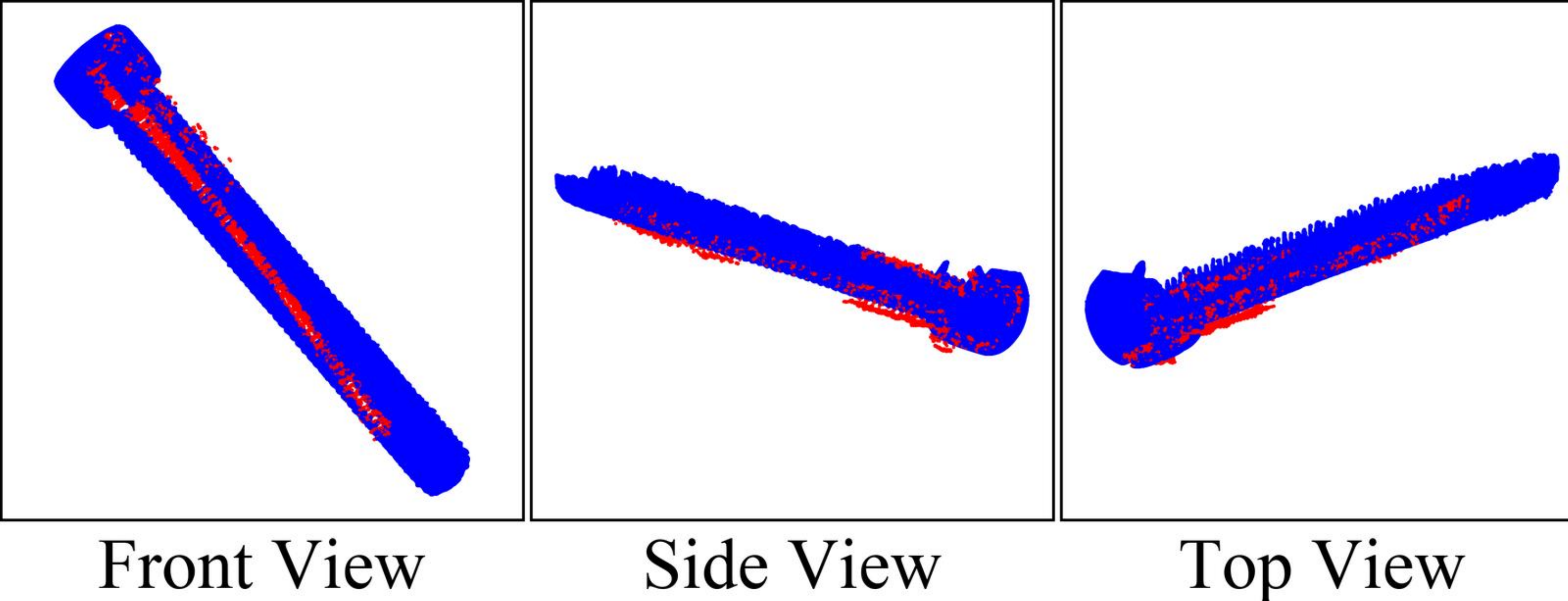}\\
    \small Cup Head Screw crack
\end{minipage}\hfill
\begin{minipage}{0.48\textwidth}
    \centering
    \includegraphics[width=\linewidth]{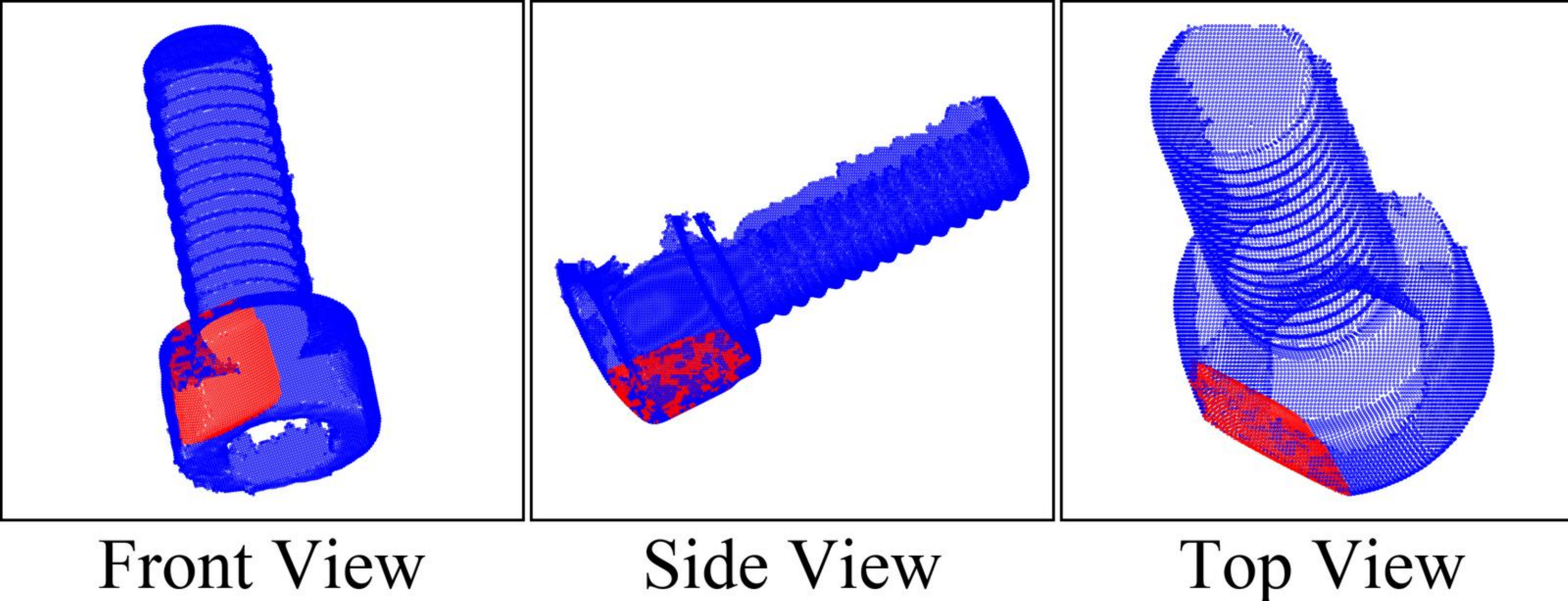}\\
    \small Cup Head Screw cut
\end{minipage}

\begin{minipage}{0.48\textwidth}
    \centering
    \includegraphics[width=\linewidth]{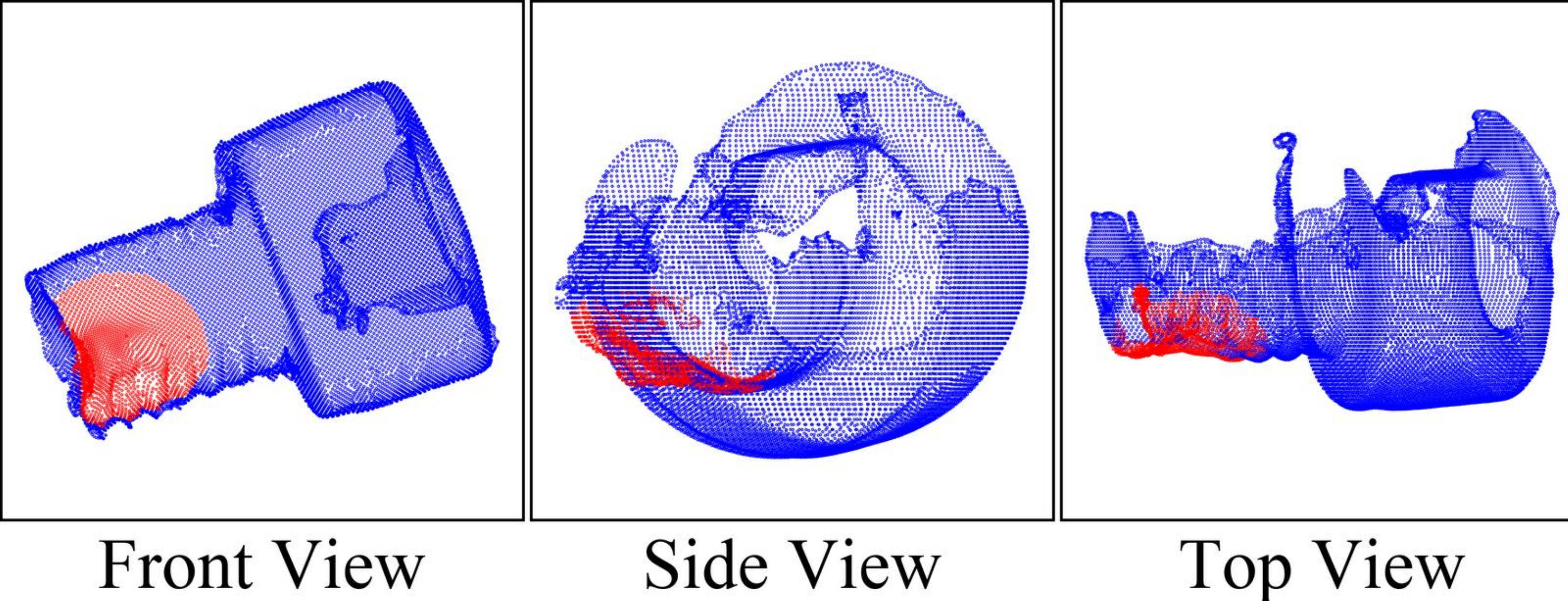}\\
    \small Cup Head Screw deformation
\end{minipage}\hfill
\begin{minipage}{0.48\textwidth}
    \centering
    \includegraphics[width=\linewidth]{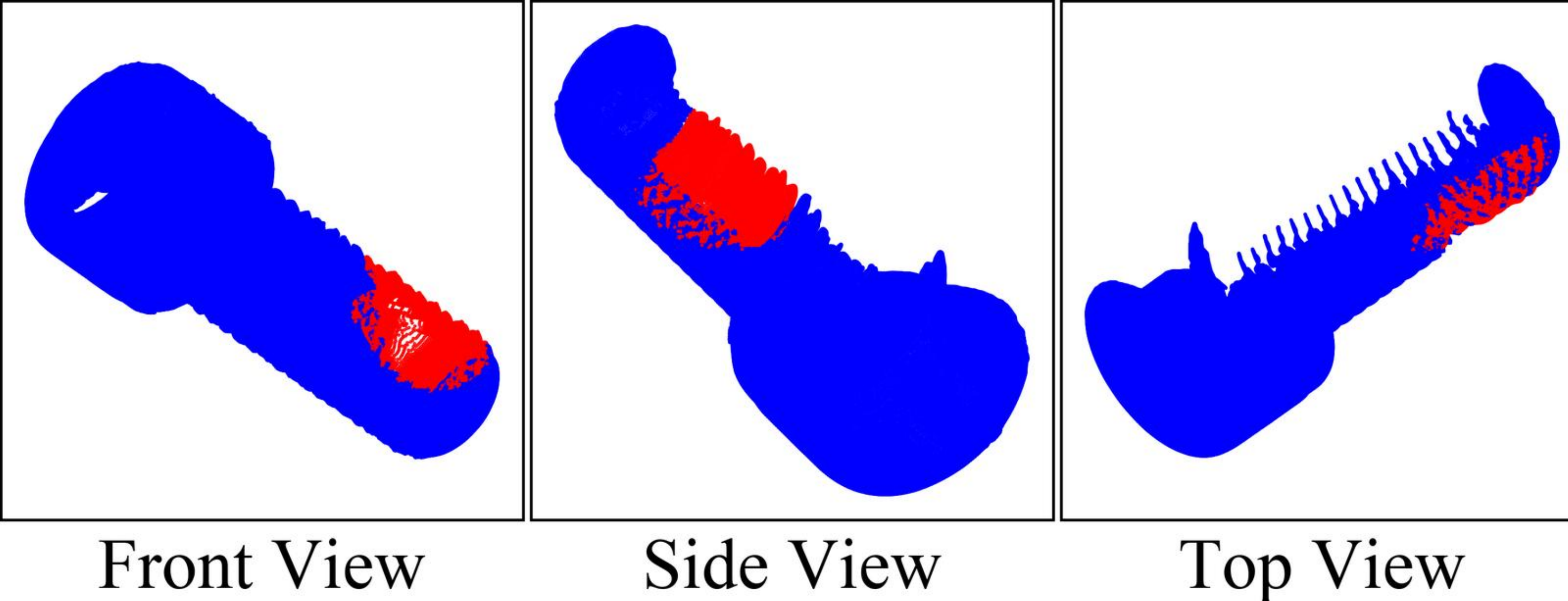}\\
    \small Cup Head Screw dent
\end{minipage}

\caption{Random examples from Cup Head Screw}
\label{fig: Cup_Head_Screw_saved}
\end{figure}

\begin{figure}[htbp]
\centering

\begin{minipage}{0.48\textwidth}
    \centering
    \includegraphics[width=\linewidth]{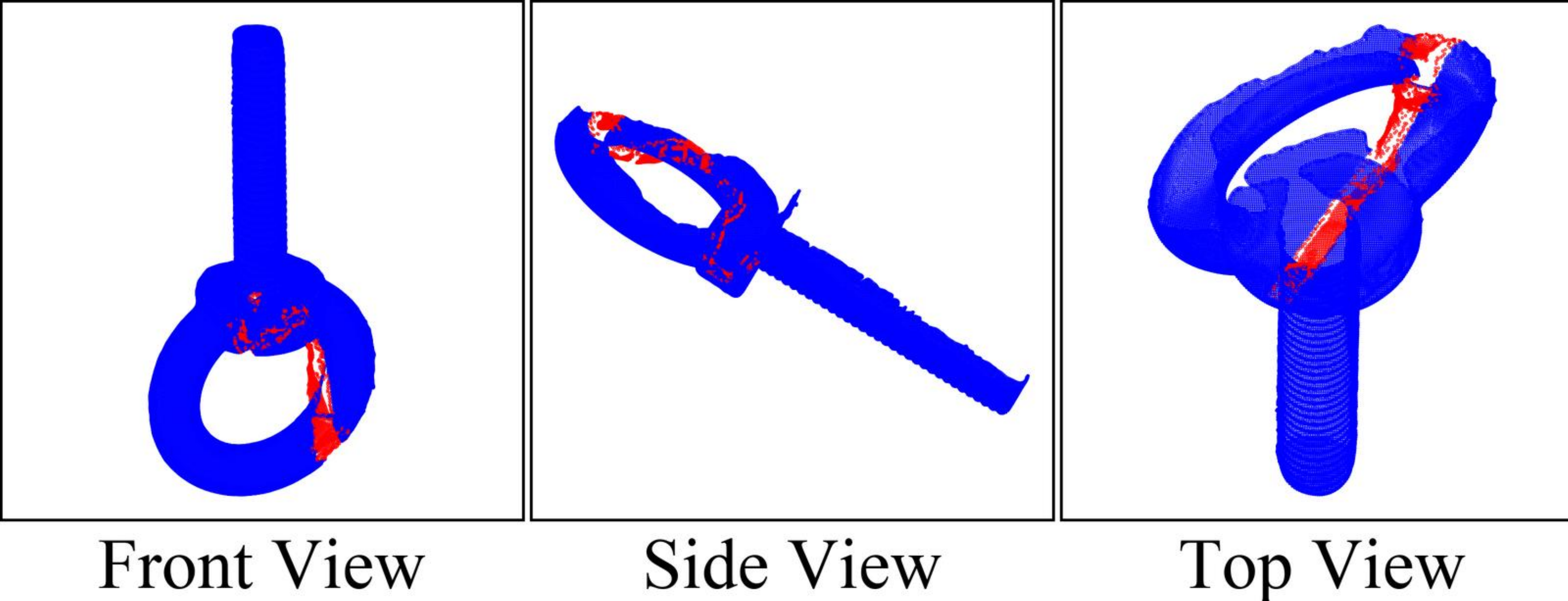}\\
    \small Eye Bolt crack
\end{minipage}\hfill
\begin{minipage}{0.48\textwidth}
    \centering
    \includegraphics[width=\linewidth]{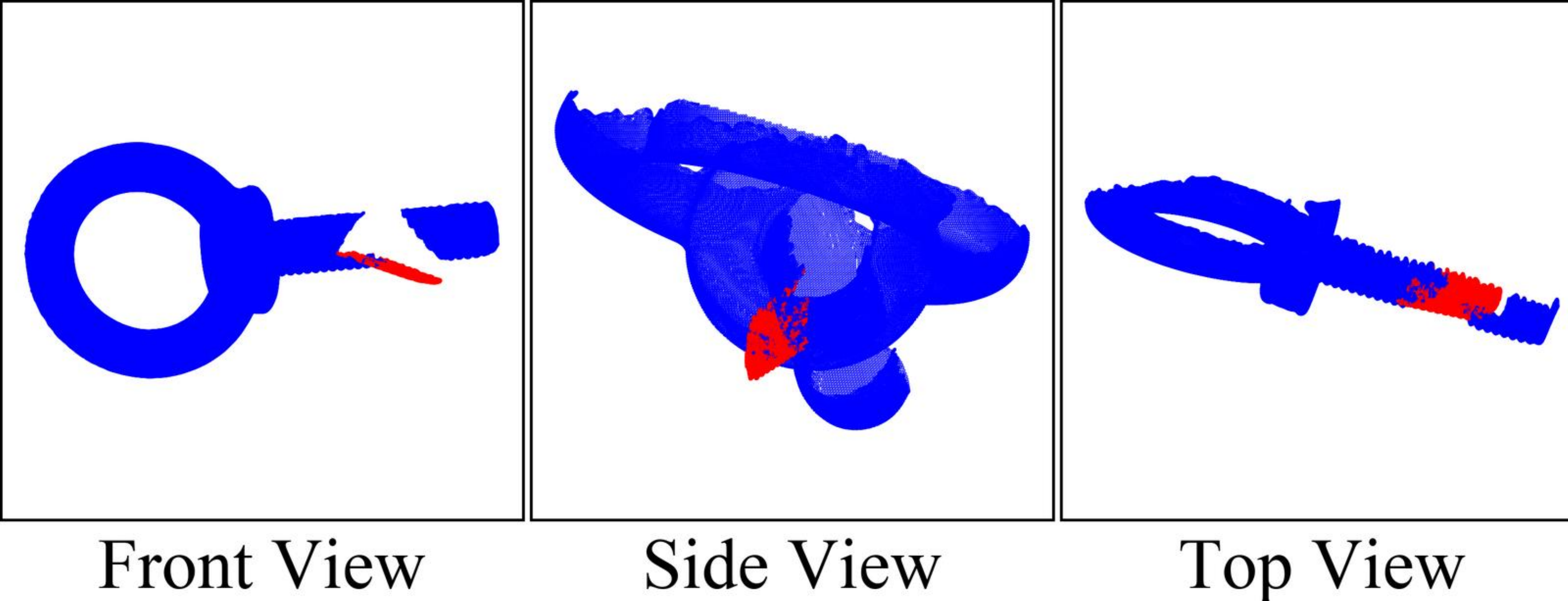}\\
    \small Eye Bolt cut
\end{minipage}

\begin{minipage}{0.48\textwidth}
    \centering
    \includegraphics[width=\linewidth]{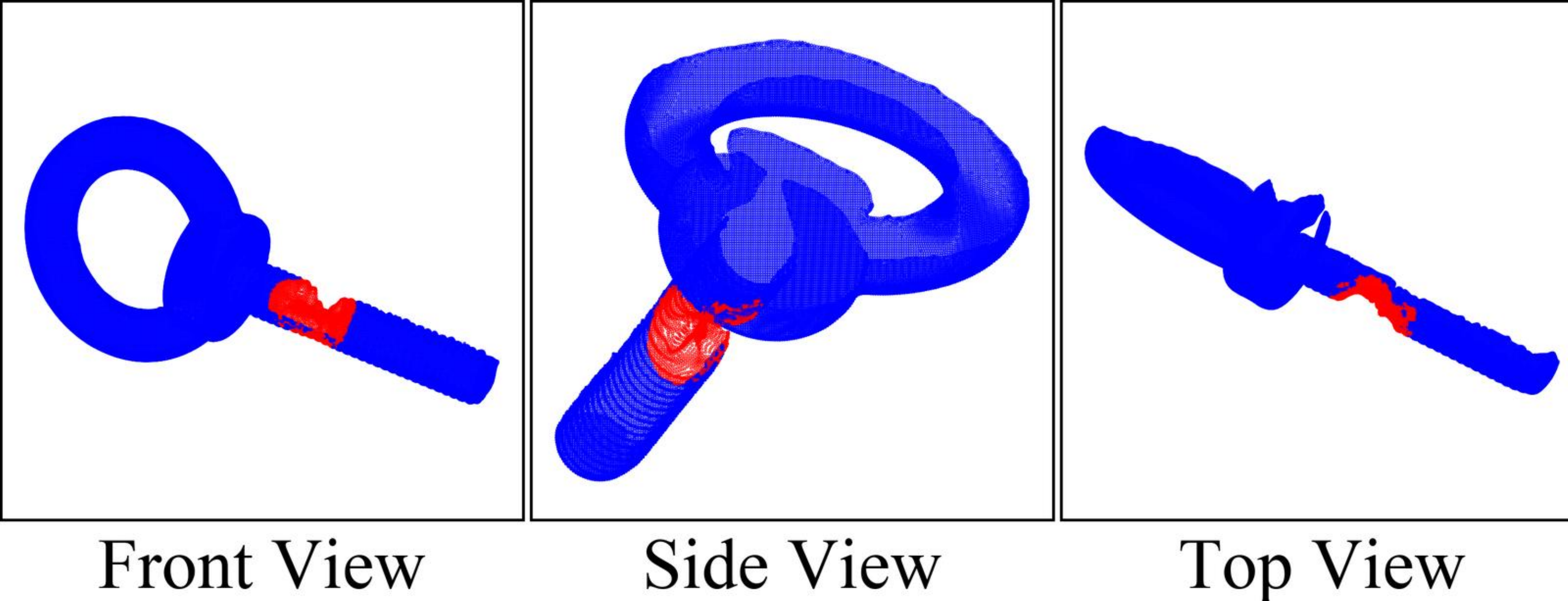}\\
    \small Eye Bolt deformation
\end{minipage}\hfill
\begin{minipage}{0.48\textwidth}
    \centering
    \includegraphics[width=\linewidth]{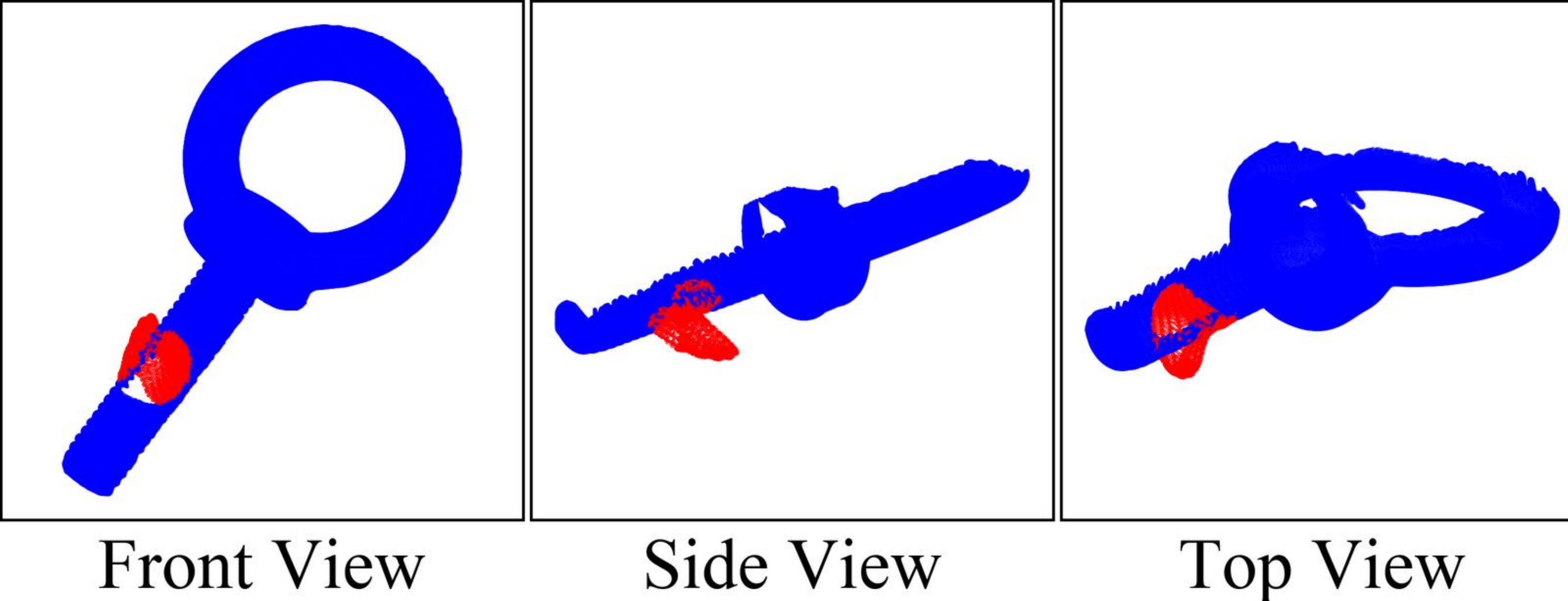}\\
    \small Eye Bolt dent
\end{minipage}

\caption{Random examples from Eye Bolt}
\label{fig: Eye_Bolt_saved}
\end{figure}

\begin{figure}[htbp]
\centering

\begin{minipage}{0.48\textwidth}
    \centering
    \includegraphics[width=\linewidth]{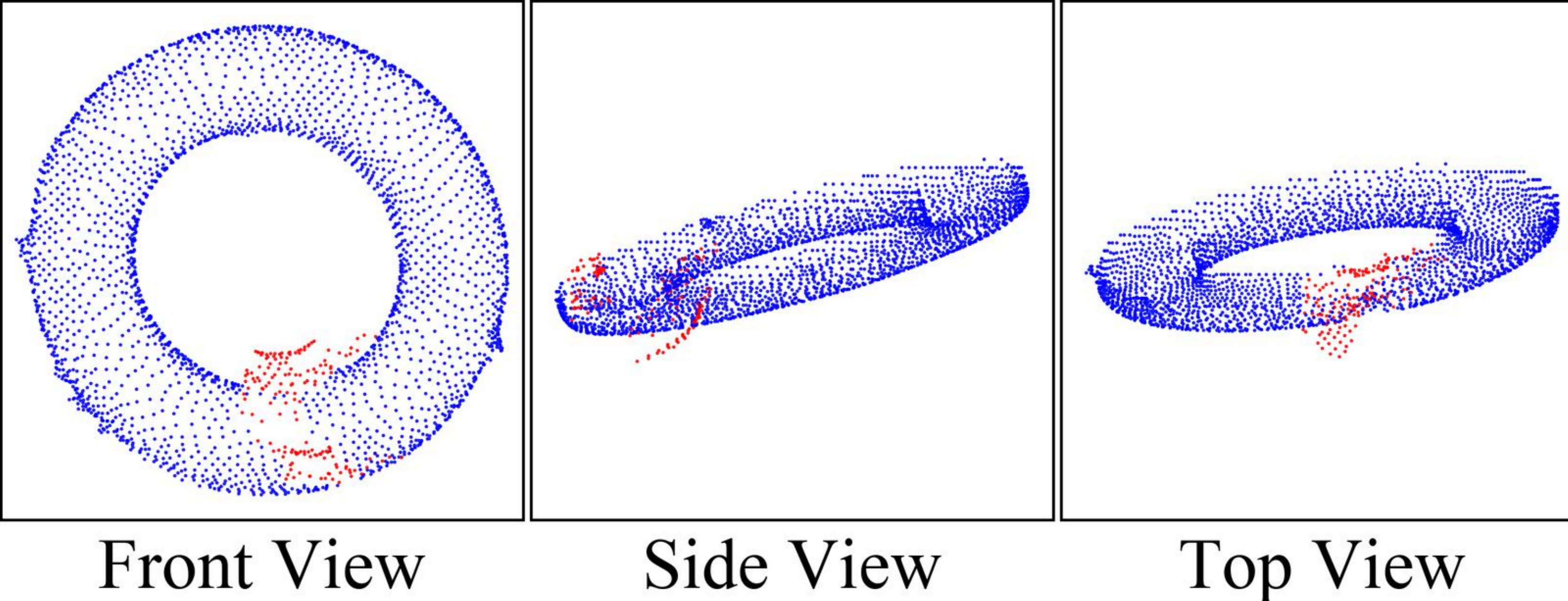}\\
    \small Flat Washer crack
\end{minipage}\hfill
\begin{minipage}{0.48\textwidth}
    \centering
    \includegraphics[width=\linewidth]{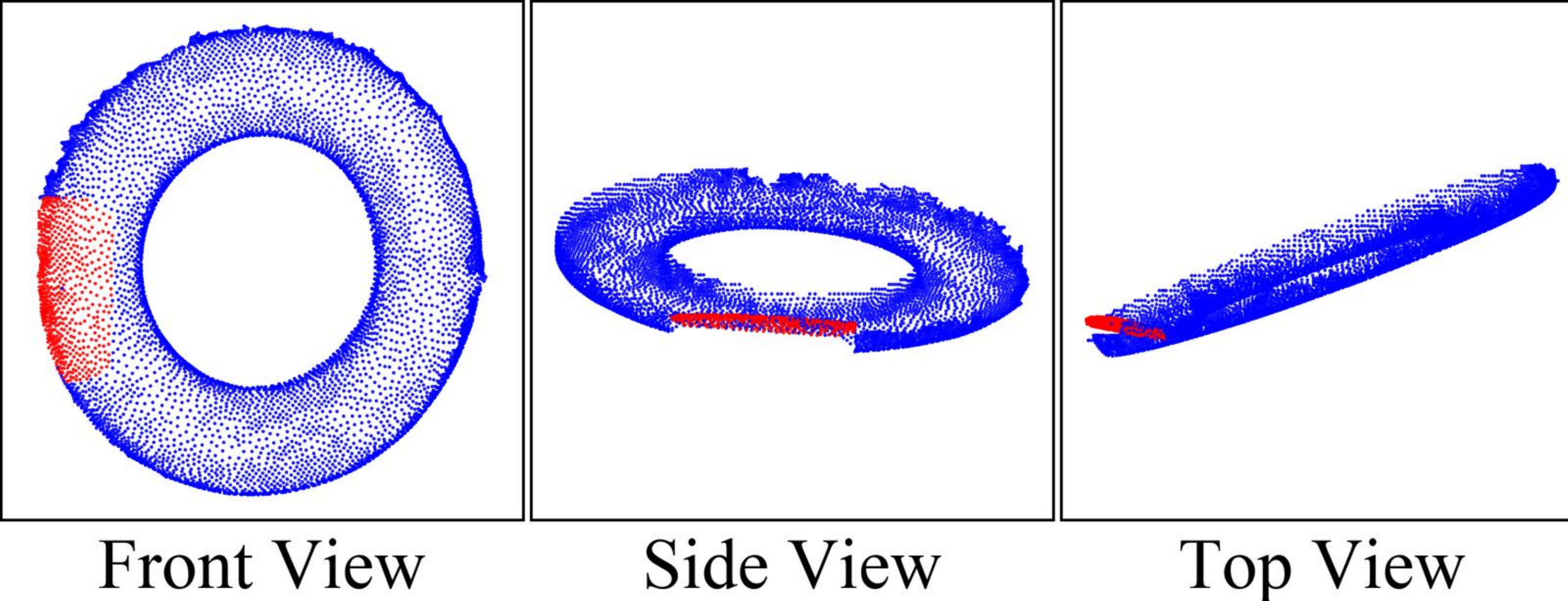}\\
    \small Flat Washer cut
\end{minipage}

\begin{minipage}{0.48\textwidth}
    \centering
    \includegraphics[width=\linewidth]{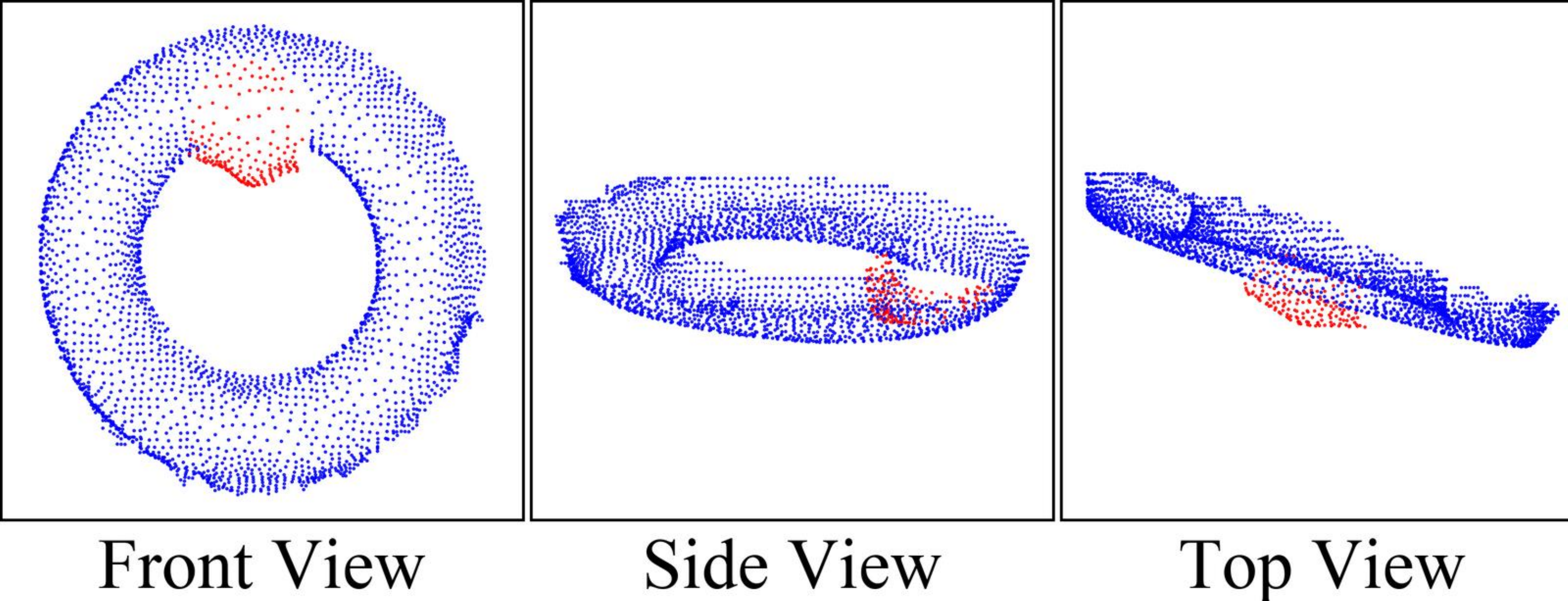}\\
    \small Flat Washer deformation
\end{minipage}\hfill
\begin{minipage}{0.48\textwidth}
    \centering
    \includegraphics[width=\linewidth]{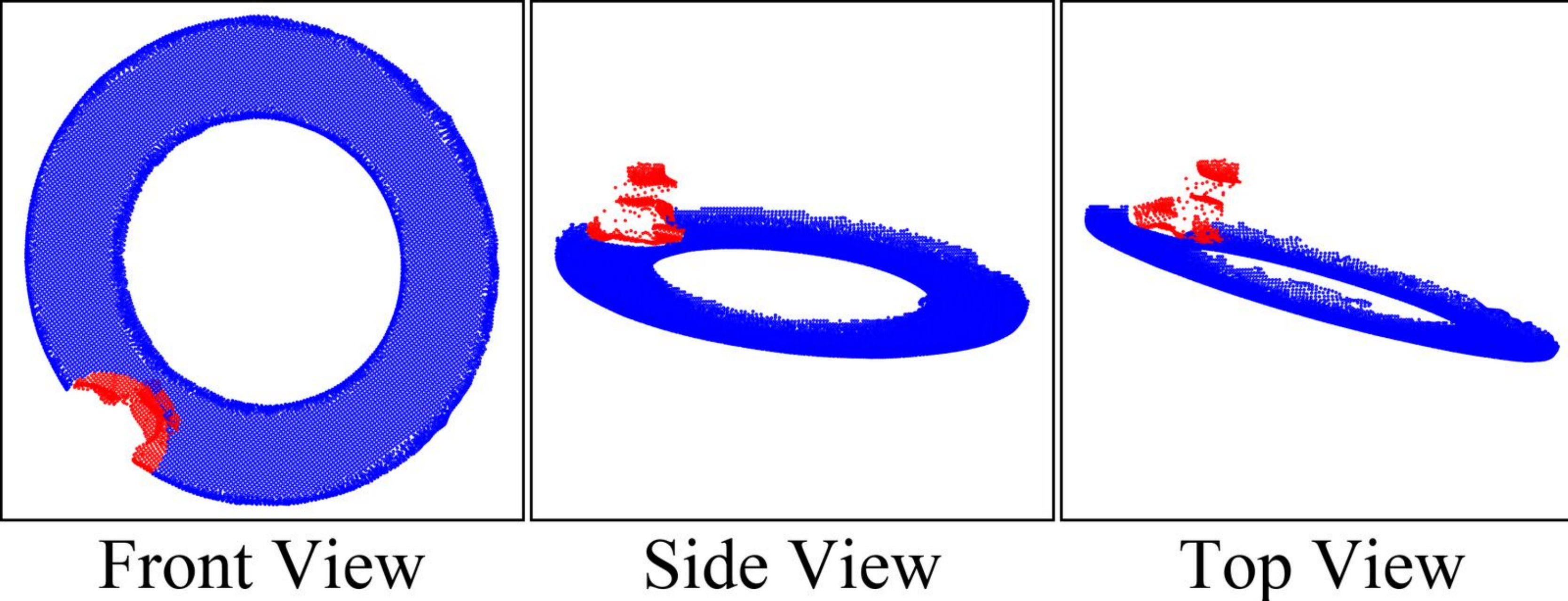}\\
    \small Flat Washer dent
\end{minipage}

\caption{Random examples from Flat Washer}
\label{fig: Flat_Washer_saved}
\end{figure}

\begin{figure}[htbp]
\centering

\begin{minipage}{0.48\textwidth}
    \centering
    \includegraphics[width=\linewidth]{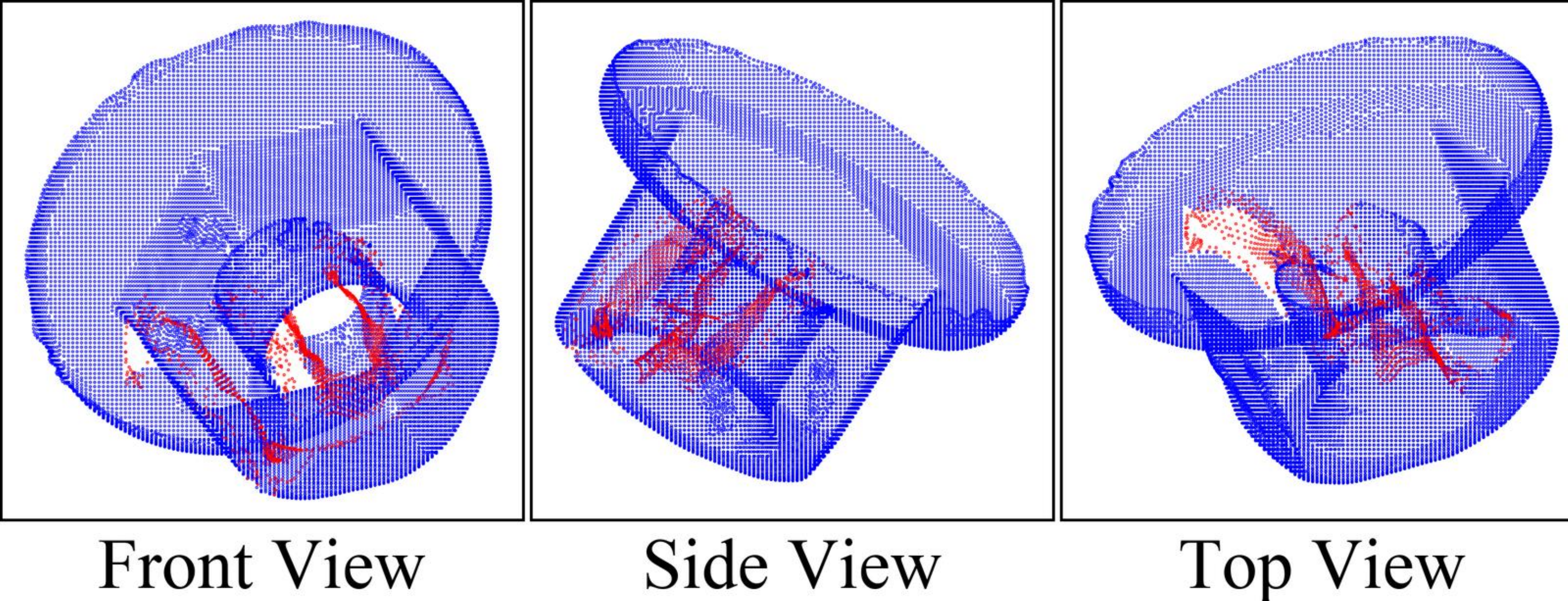}\\
    \small Hex Nut crack
\end{minipage}\hfill
\begin{minipage}{0.48\textwidth}
    \centering
    \includegraphics[width=\linewidth]{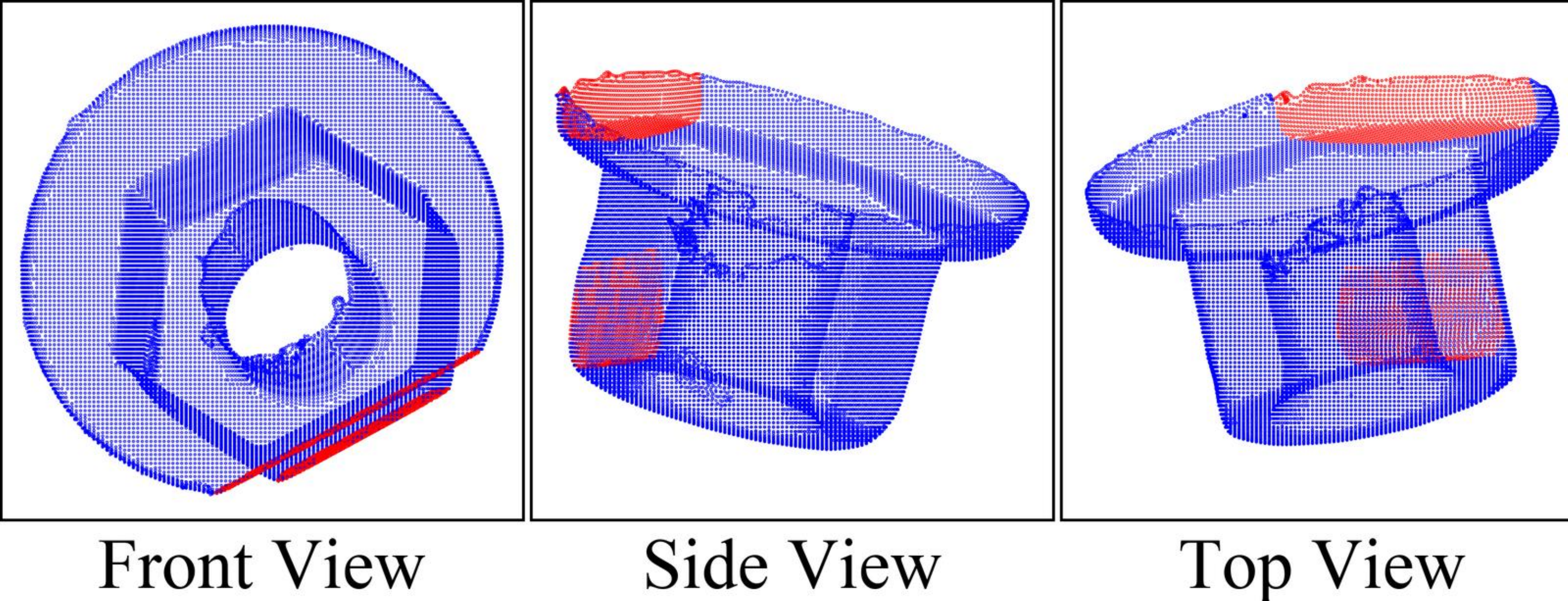}\\
    \small Hex Nut cut
\end{minipage}

\begin{minipage}{0.48\textwidth}
    \centering
    \includegraphics[width=\linewidth]{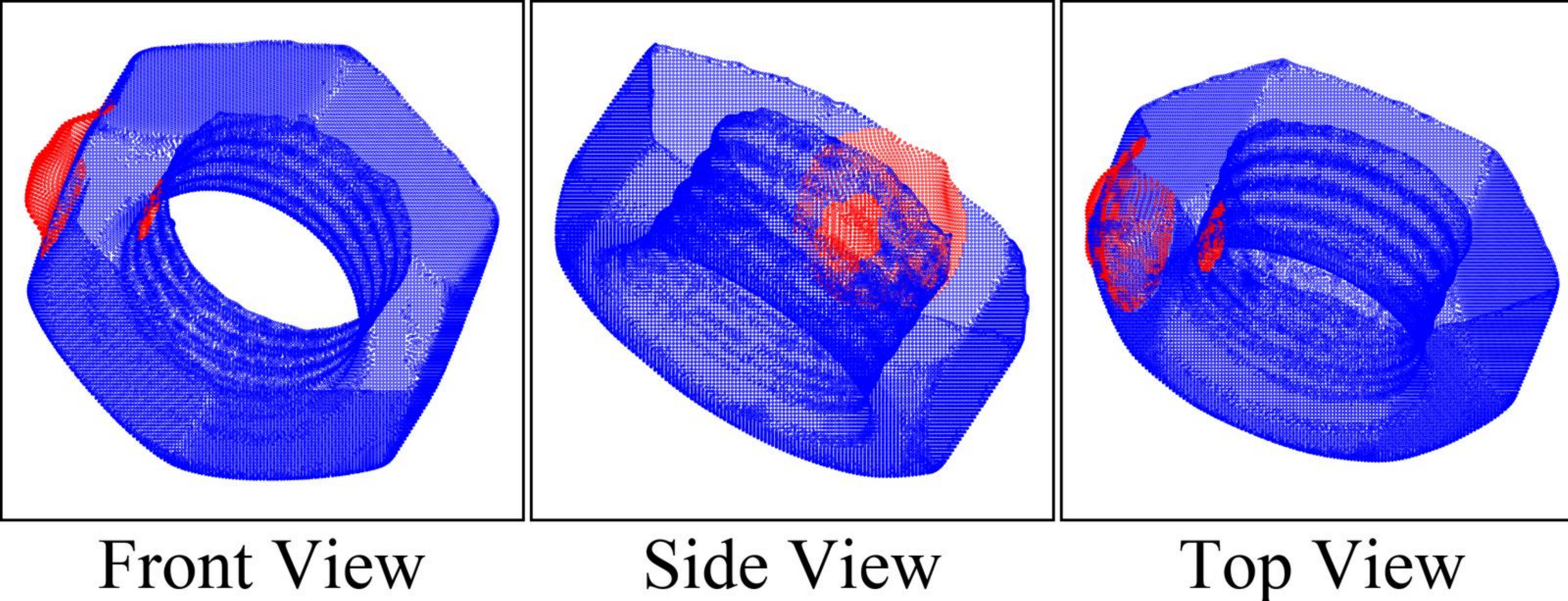}\\
    \small Hex Nut deformation
\end{minipage}\hfill
\begin{minipage}{0.48\textwidth}
    \centering
    \includegraphics[width=\linewidth]{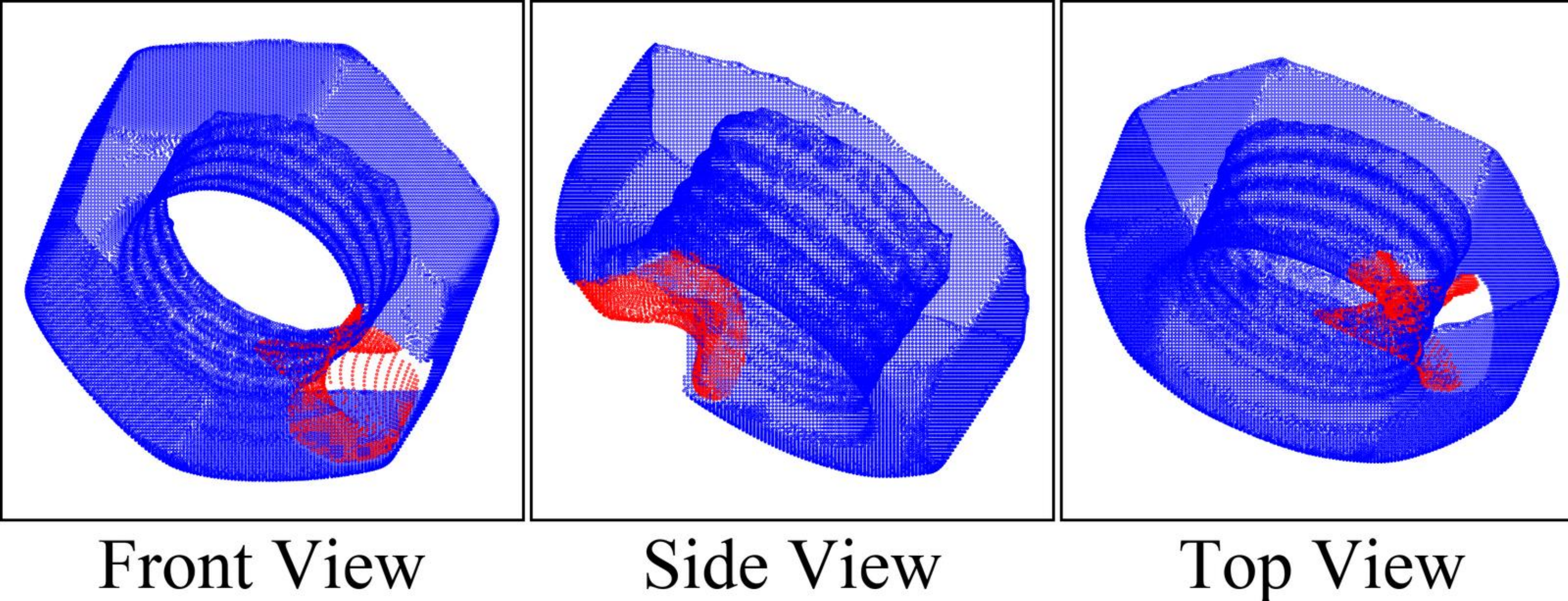}\\
    \small Hex Nut dent
\end{minipage}

\caption{Random examples from Hex Nut}
\label{fig: Hex_Nut_saved}
\end{figure}

\begin{figure}[htbp]
\centering

\begin{minipage}{0.48\textwidth}
    \centering
    \includegraphics[width=\linewidth]{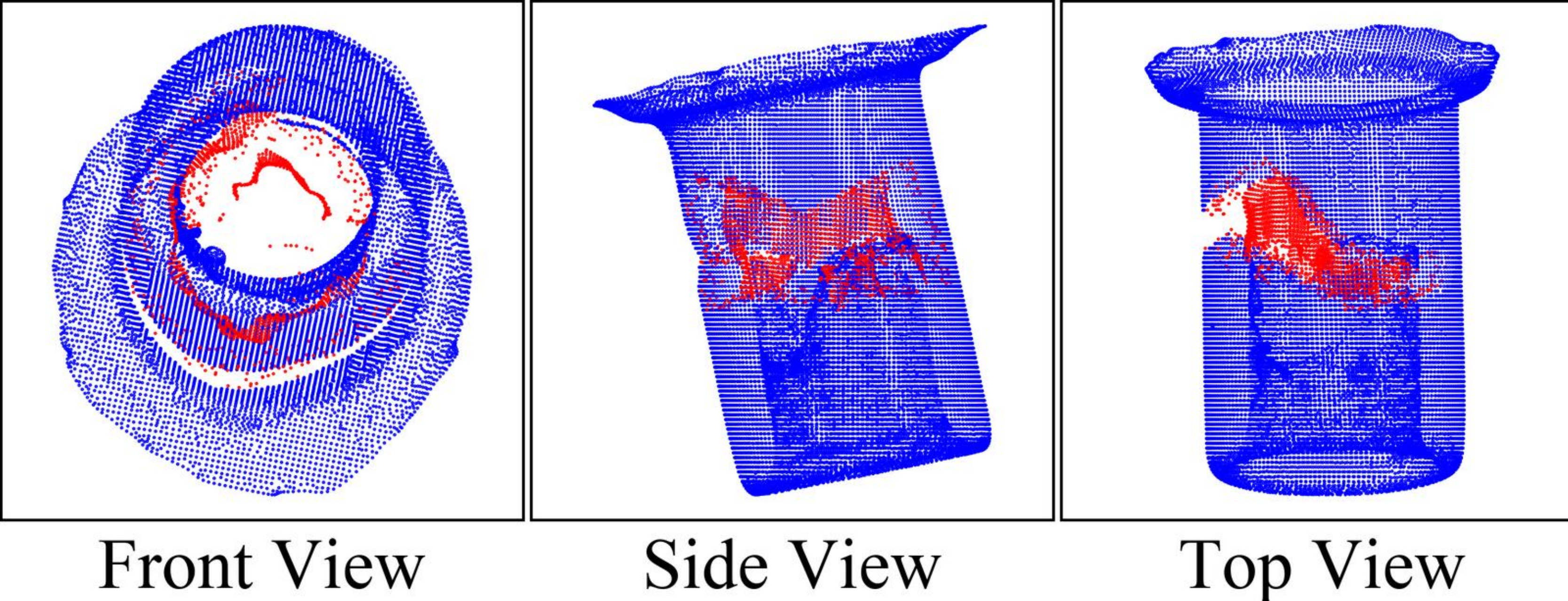}\\
    \small Rivet Nut crack
\end{minipage}\hfill
\begin{minipage}{0.48\textwidth}
    \centering
    \includegraphics[width=\linewidth]{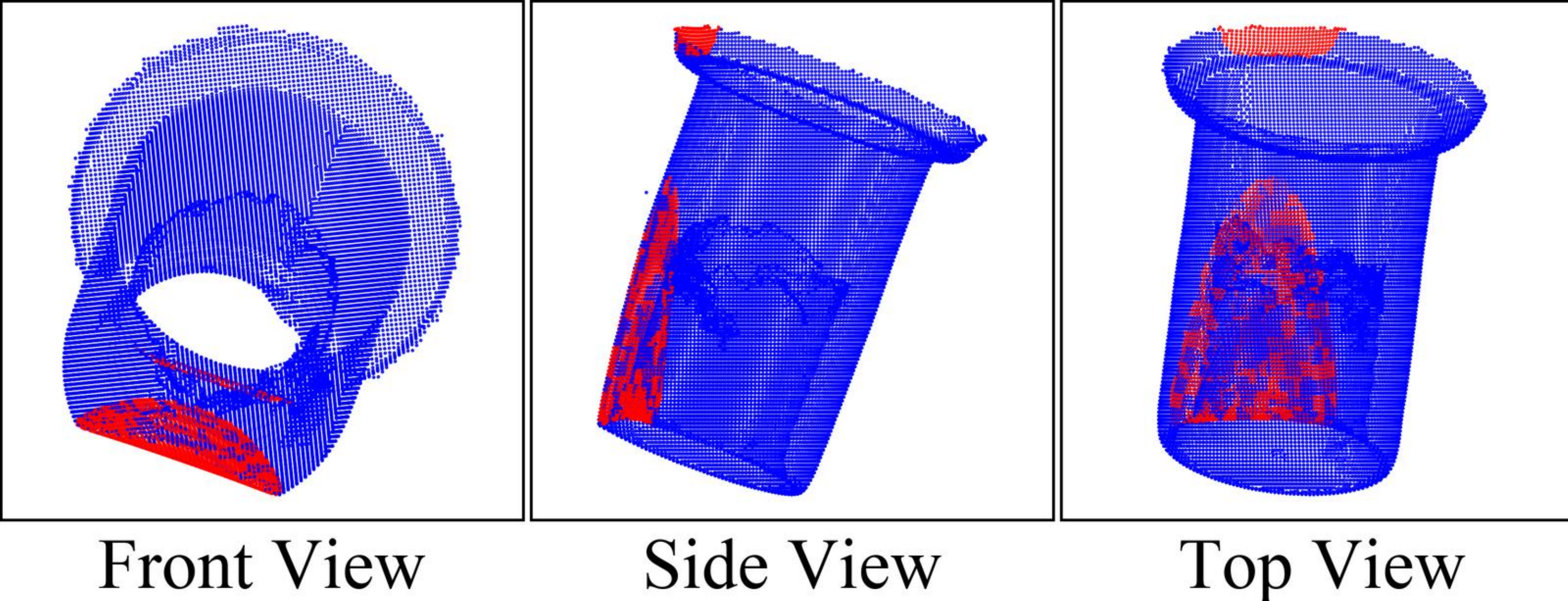}\\
    \small Rivet Nut cut
\end{minipage}

\begin{minipage}{0.48\textwidth}
    \centering
    \includegraphics[width=\linewidth]{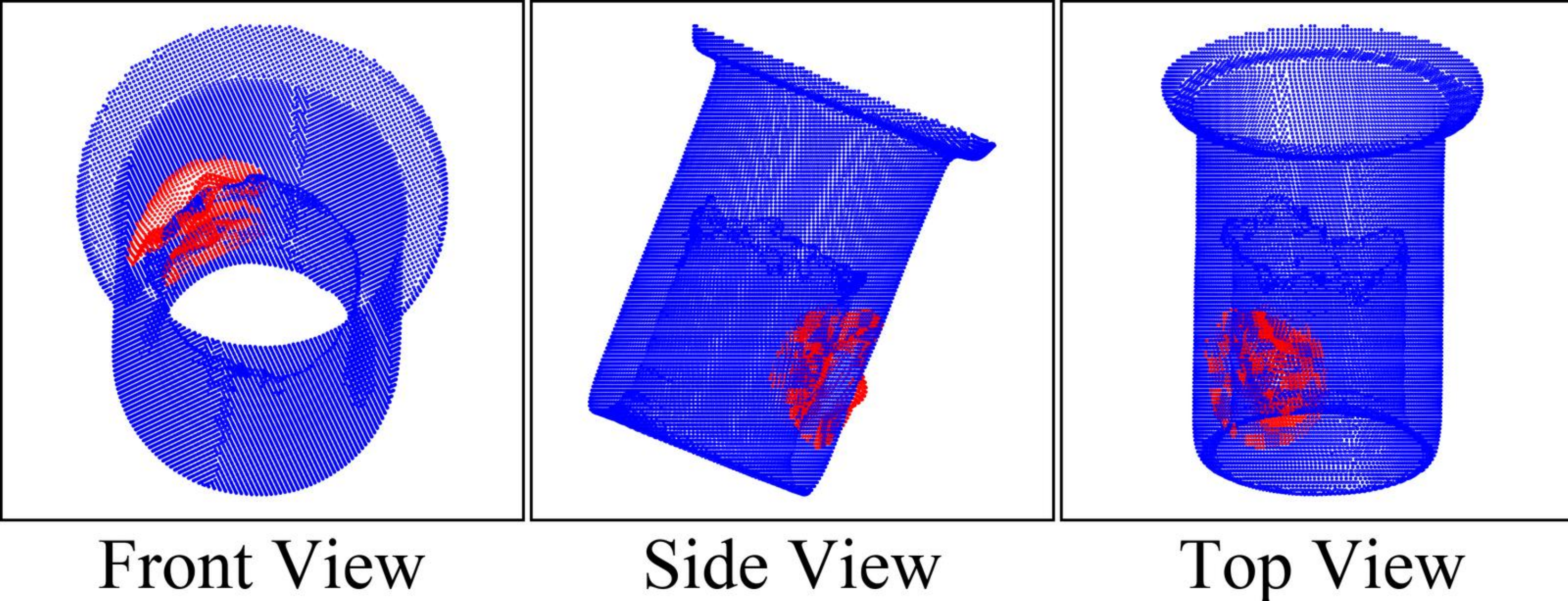}\\
    \small Rivet Nut deformation
\end{minipage}\hfill
\begin{minipage}{0.48\textwidth}
    \centering
    \includegraphics[width=\linewidth]{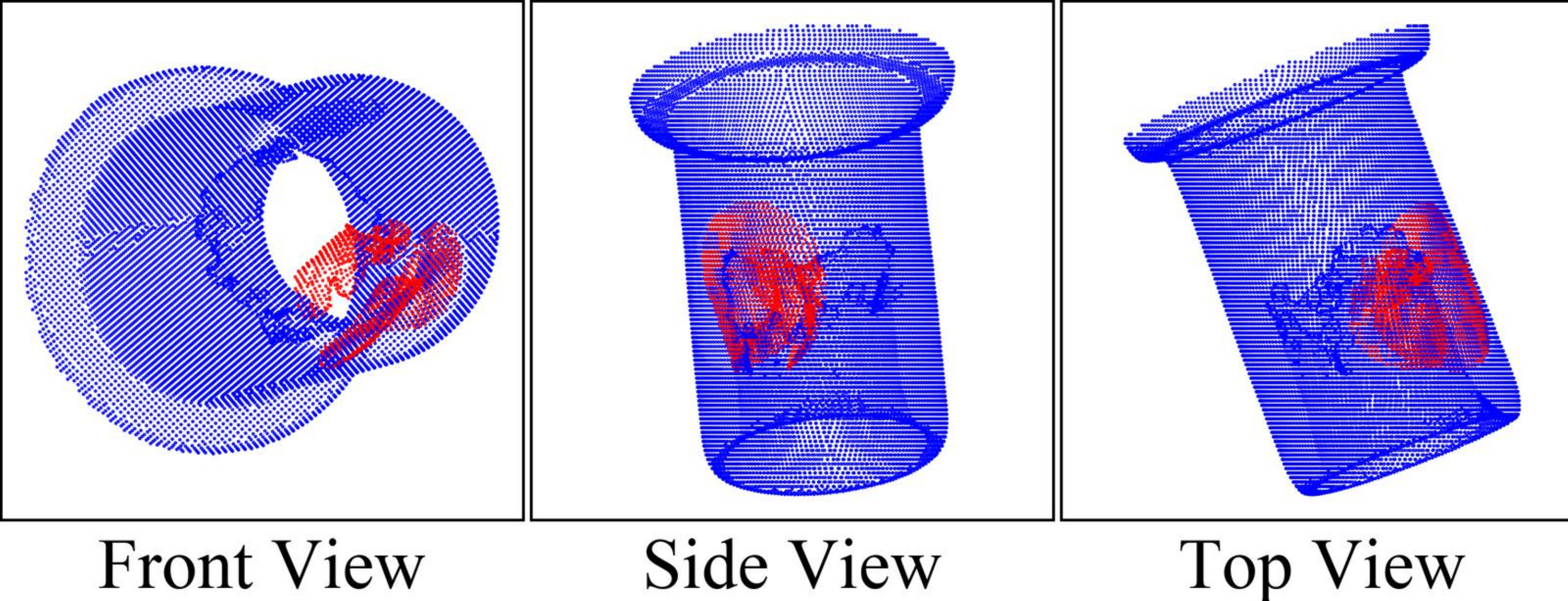}\\
    \small Rivet Nut dent
\end{minipage}

\caption{Random examples from Rivet Nut}
\label{fig: Rivet_Nut_saved}
\end{figure}

\begin{figure}[htbp]
\centering

\begin{minipage}{0.48\textwidth}
    \centering
    \includegraphics[width=\linewidth]{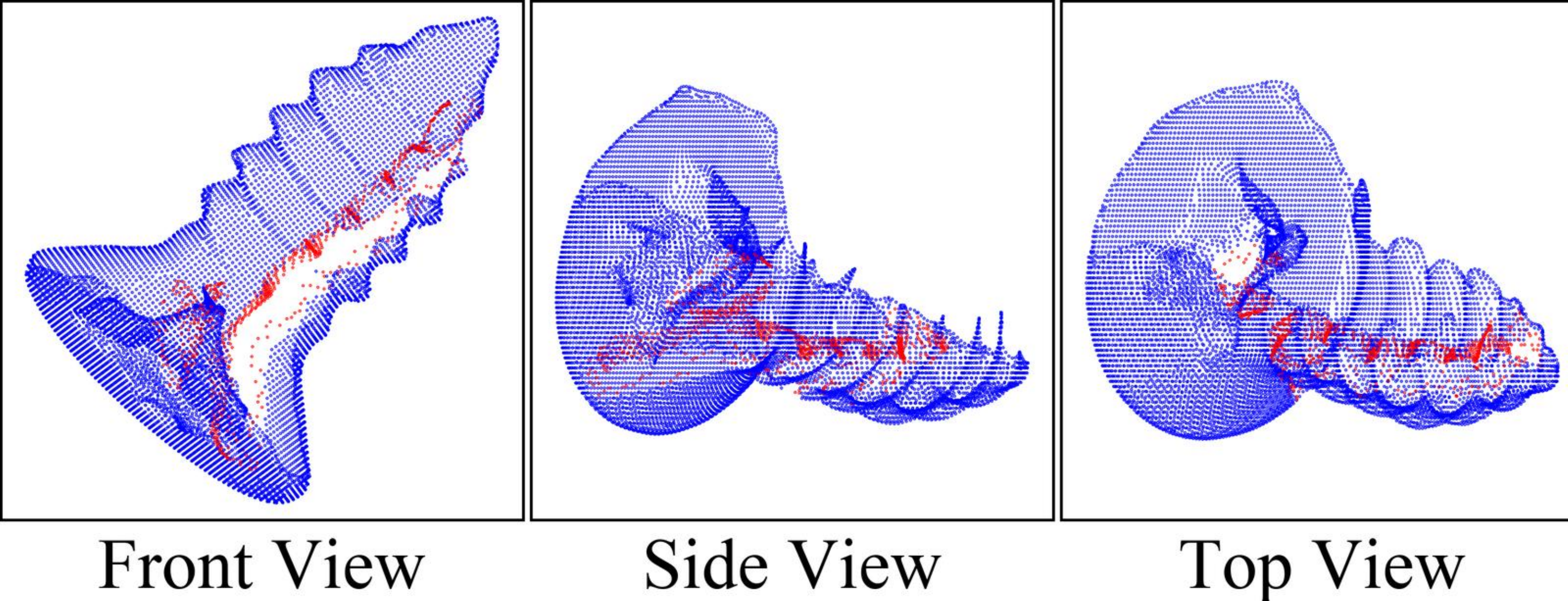}\\
    \small Self-tapping Screw crack
\end{minipage}\hfill
\begin{minipage}{0.48\textwidth}
    \centering
    \includegraphics[width=\linewidth]{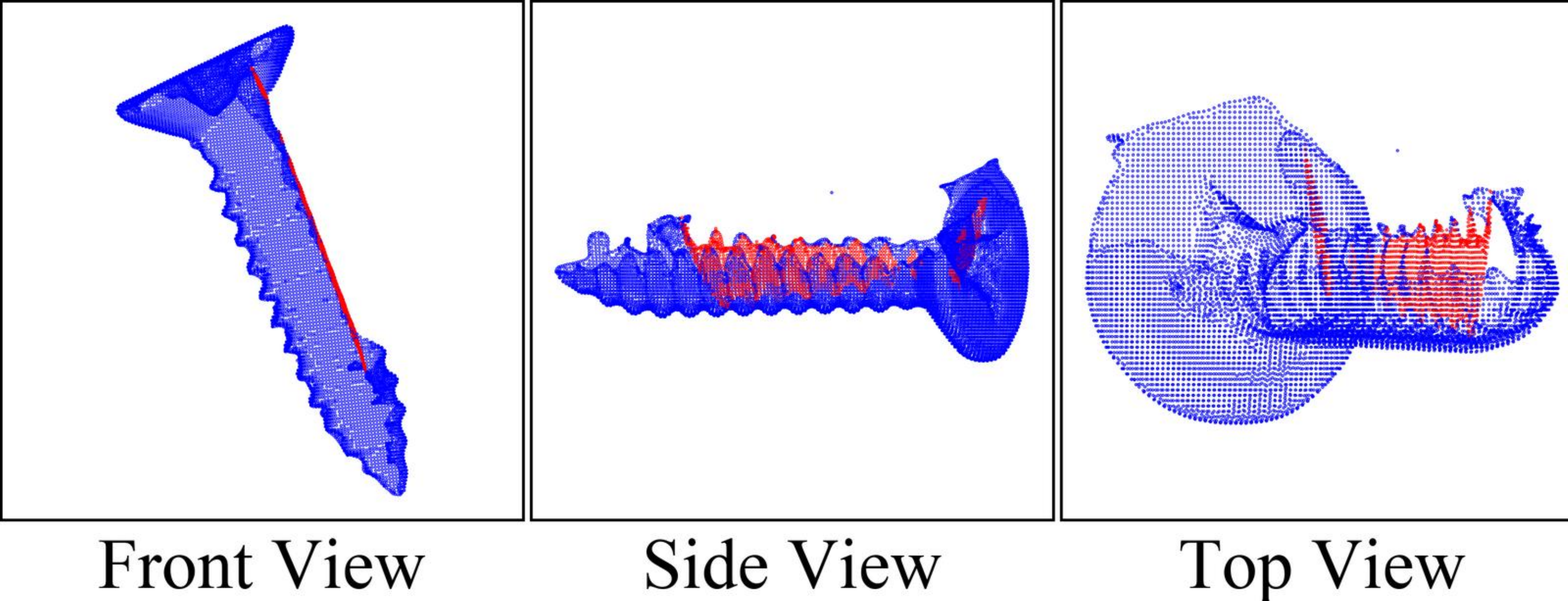}\\
    \small Self-tapping Screw cut
\end{minipage}

\begin{minipage}{0.48\textwidth}
    \centering
    \includegraphics[width=\linewidth]{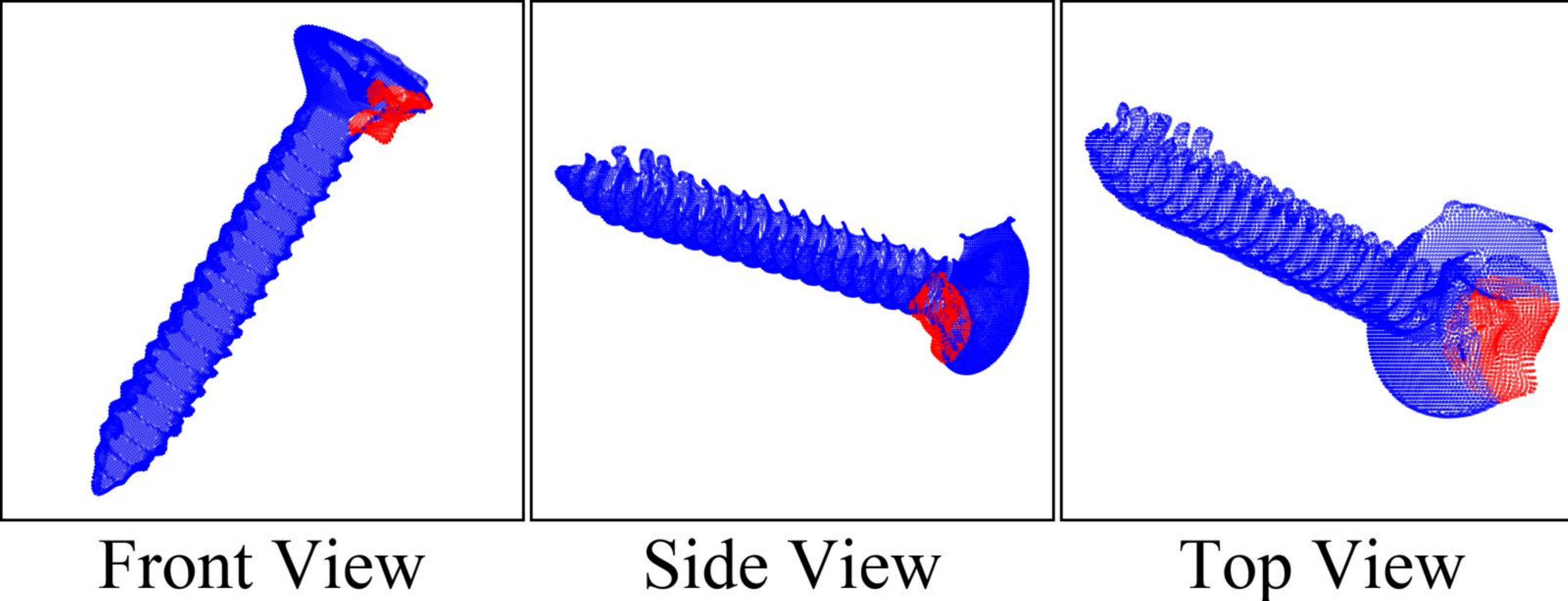}\\
    \small Self-tapping Screw deformation
\end{minipage}\hfill
\begin{minipage}{0.48\textwidth}
    \centering
    \includegraphics[width=\linewidth]{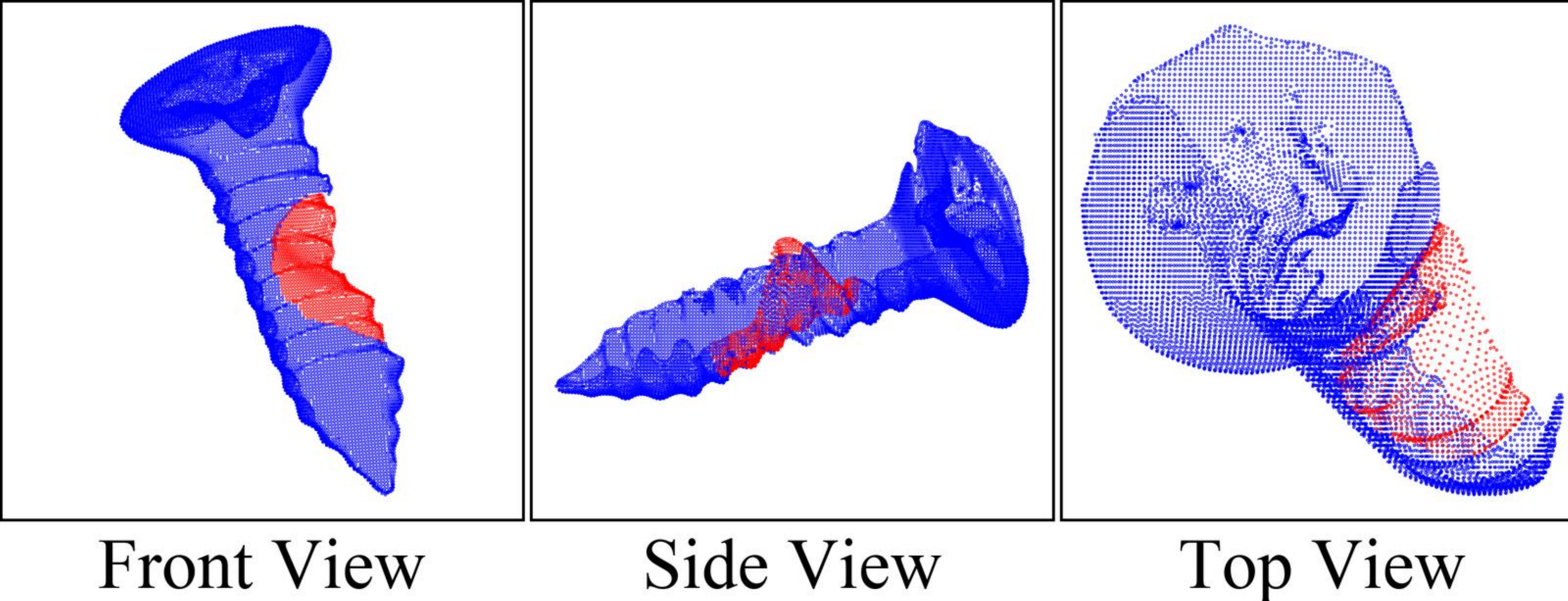}\\
    \small Self-tapping Screw dent
\end{minipage}

\caption{Random examples from Self-tapping Screw}
\label{fig: Self_tapping_Screw_saved}
\end{figure}

\begin{figure}[htbp]
\centering

\begin{minipage}{0.48\textwidth}
    \centering
    \includegraphics[width=\linewidth]{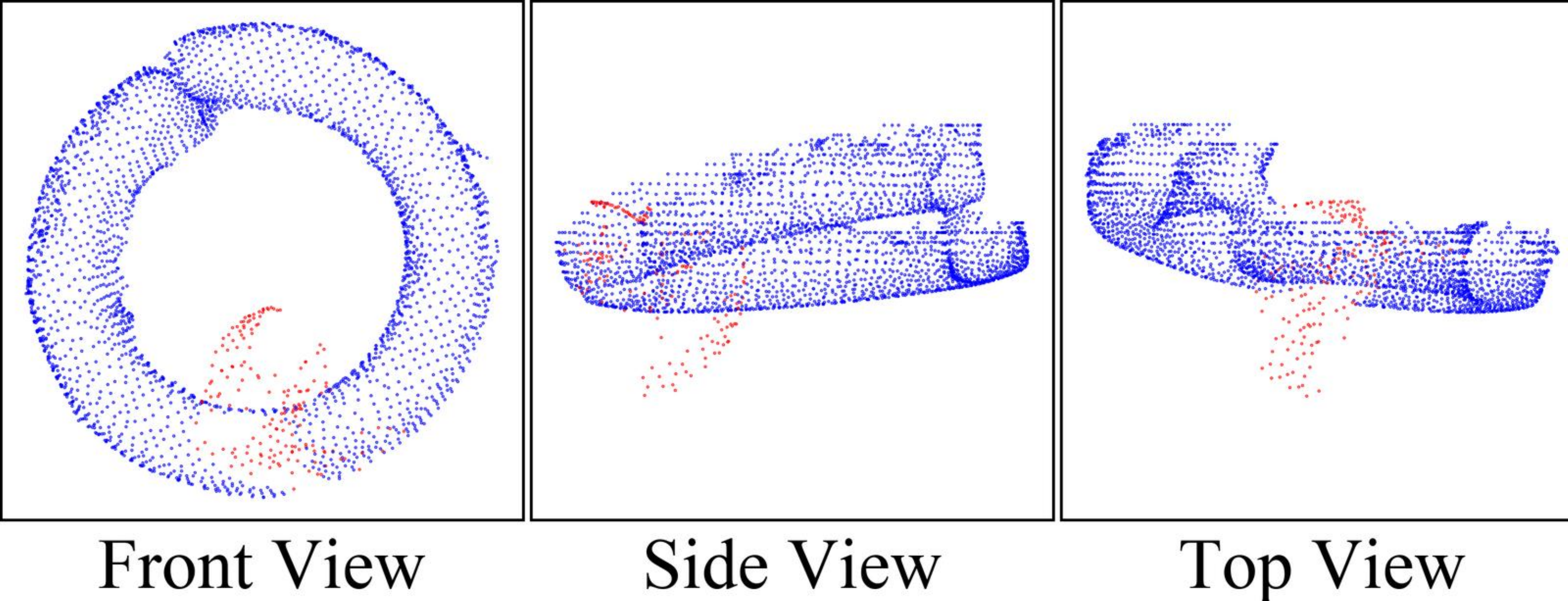}\\
    \small Spring Washer crack
\end{minipage}\hfill
\begin{minipage}{0.48\textwidth}
    \centering
    \includegraphics[width=\linewidth]{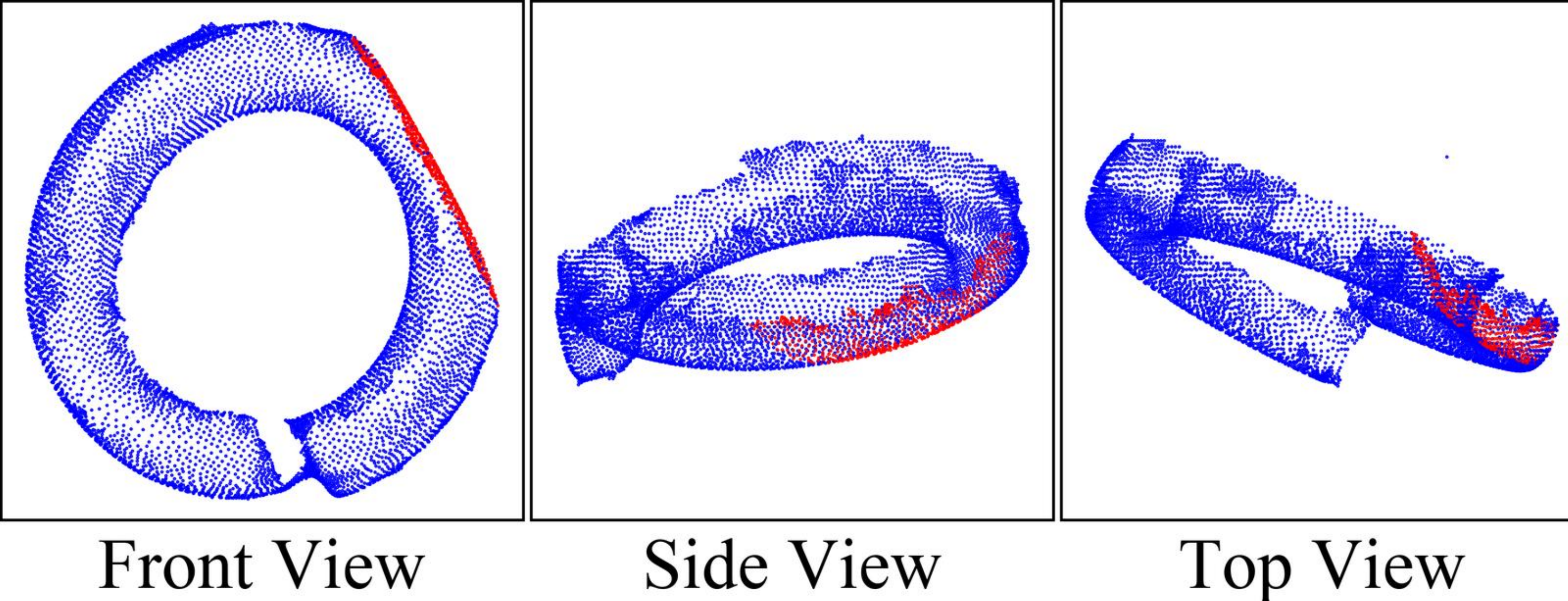}\\
    \small Spring Washer cut
\end{minipage}

\begin{minipage}{0.48\textwidth}
    \centering
    \includegraphics[width=\linewidth]{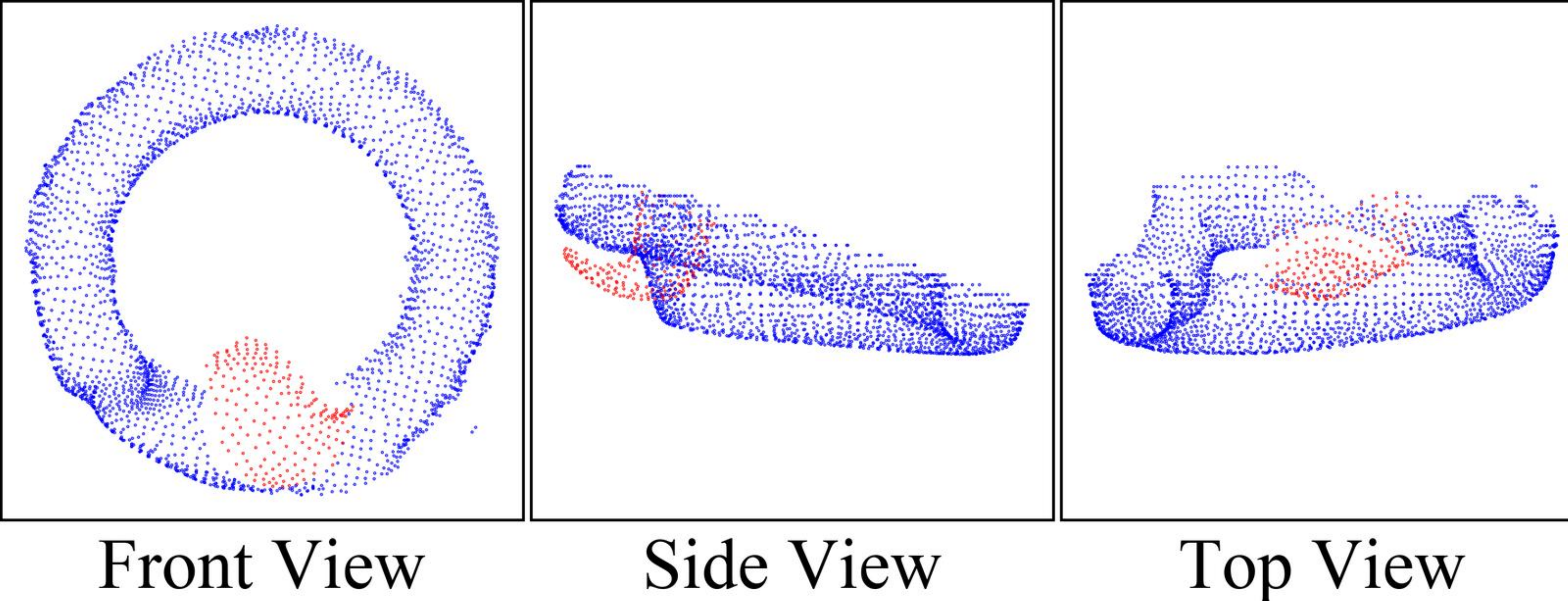}\\
    \small Spring Washer deformation
\end{minipage}\hfill
\begin{minipage}{0.48\textwidth}
    \centering
    \includegraphics[width=\linewidth]{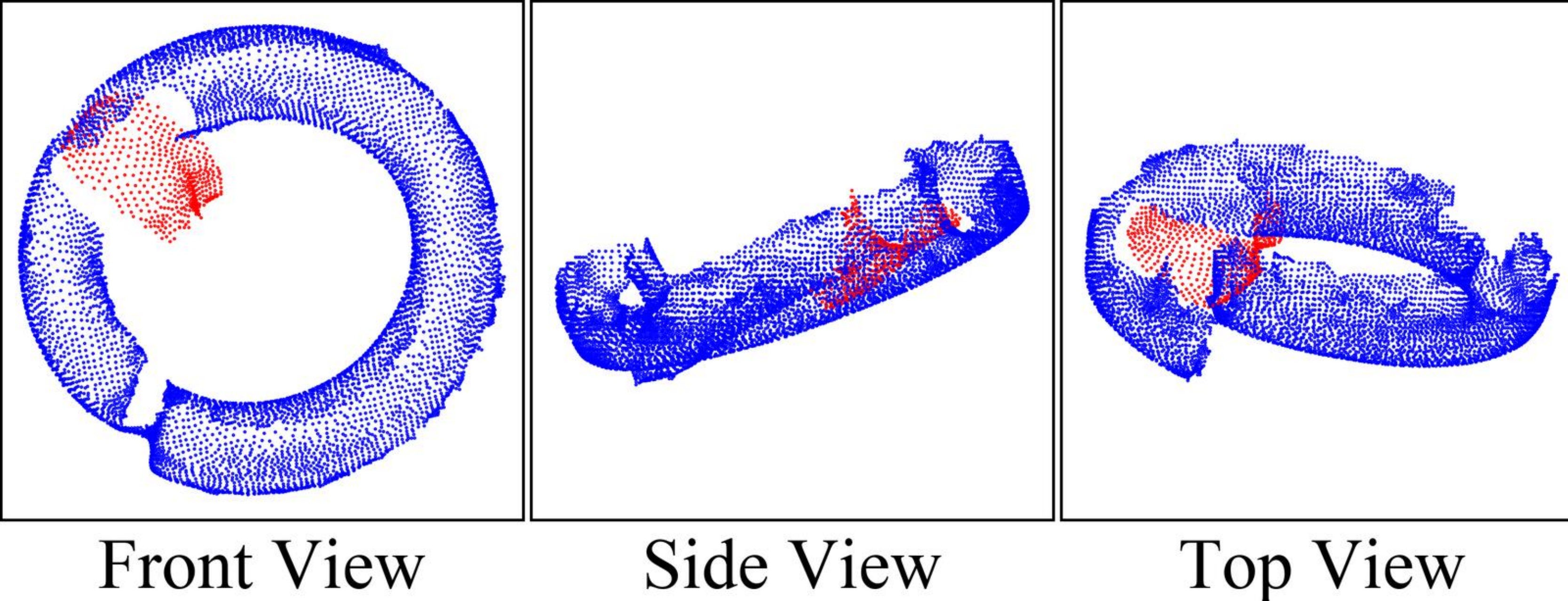}\\
    \small Spring Washer dent
\end{minipage}

\caption{Random examples from Spring Washer}
\label{fig: Spring_Washer_saved}
\end{figure}

\begin{figure}[htbp]
\centering

\begin{minipage}{0.48\textwidth}
    \centering
    \includegraphics[width=\linewidth]{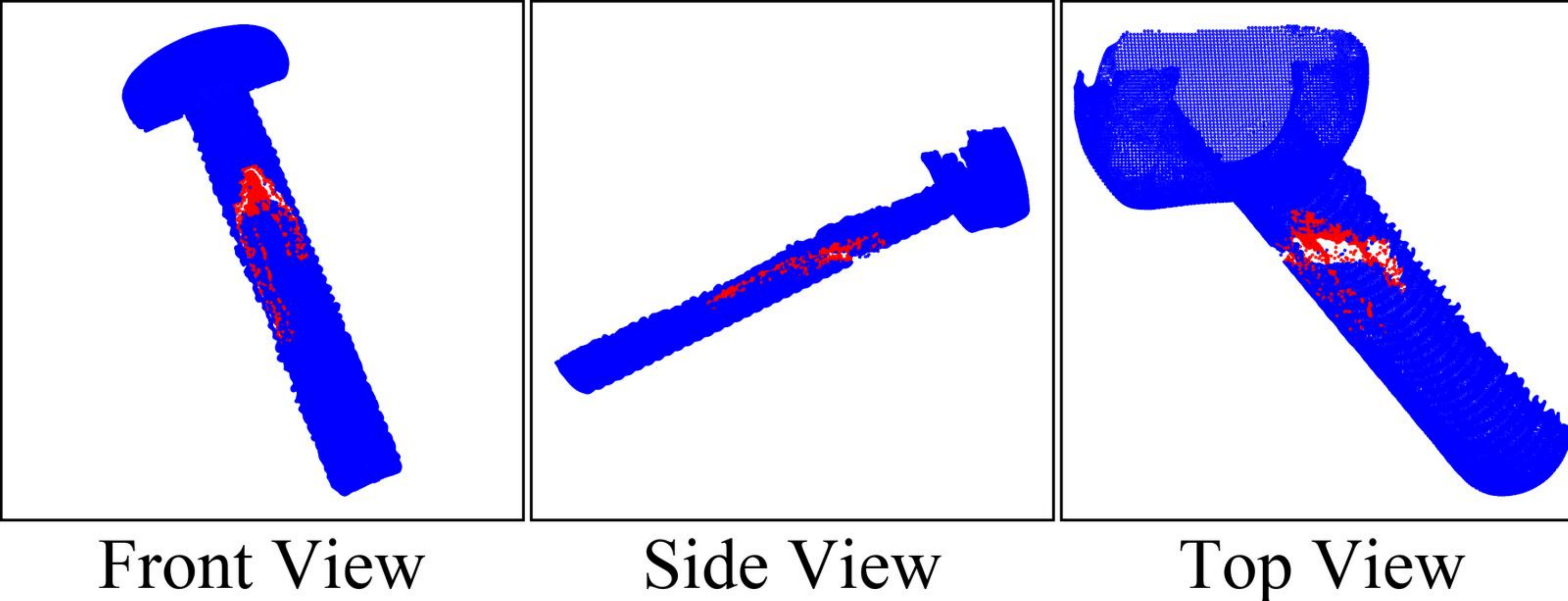}\\
    \small T Bolt Half thread Screw crack
\end{minipage}\hfill
\begin{minipage}{0.48\textwidth}
    \centering
    \includegraphics[width=\linewidth]{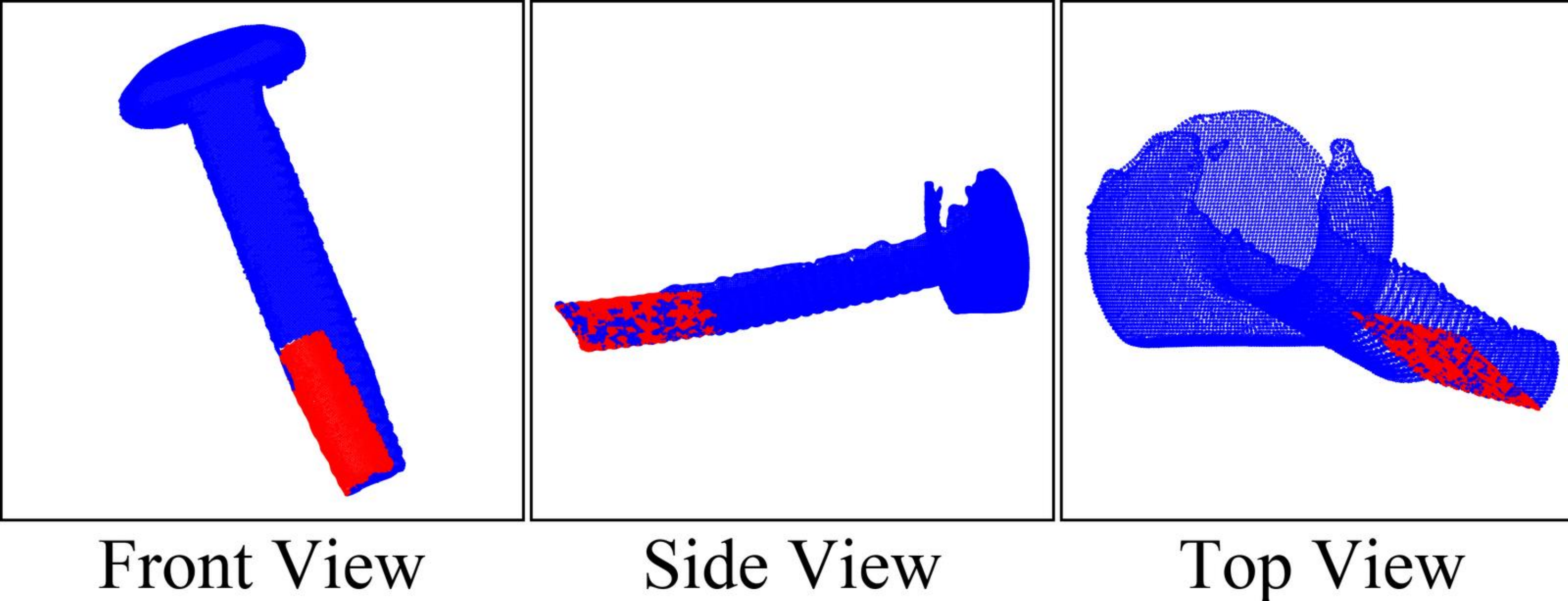}\\
    \small T Bolt Half thread Screw cut
\end{minipage}

\begin{minipage}{0.48\textwidth}
    \centering
    \includegraphics[width=\linewidth]{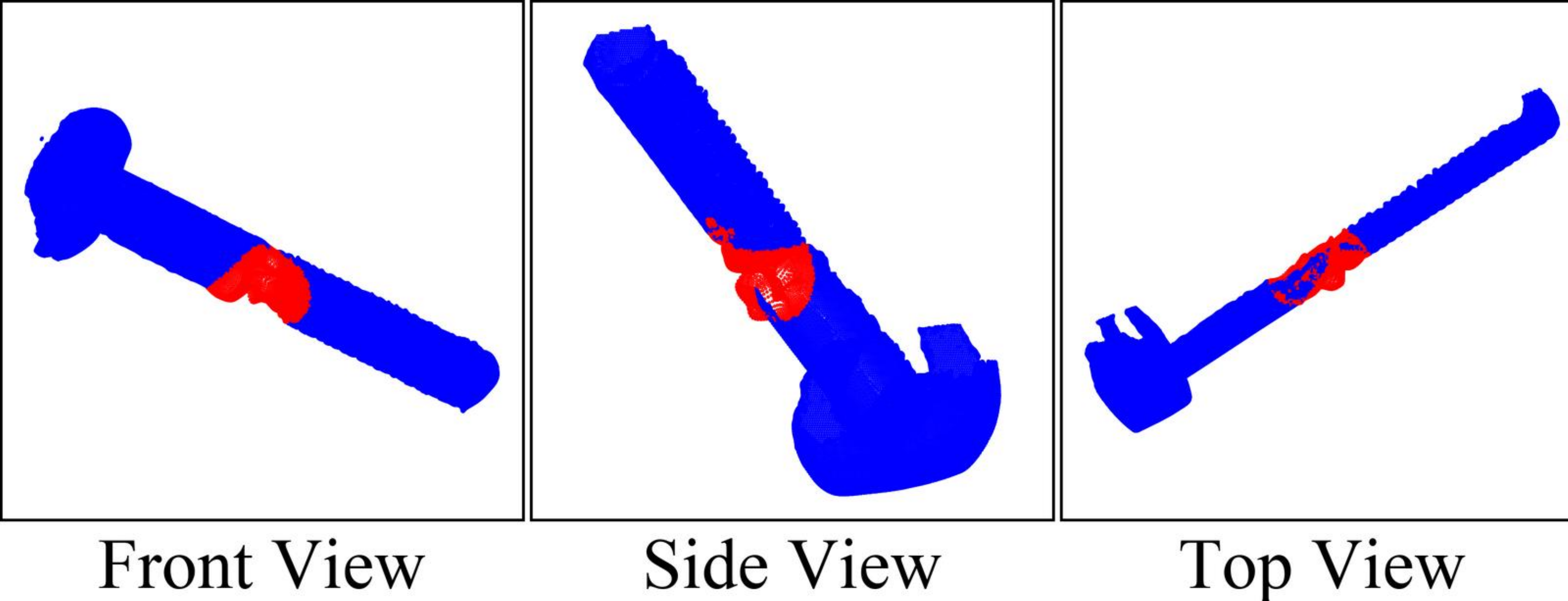}\\
    \small T Bolt Half thread Screw deformation
\end{minipage}\hfill
\begin{minipage}{0.48\textwidth}
    \centering
    \includegraphics[width=\linewidth]{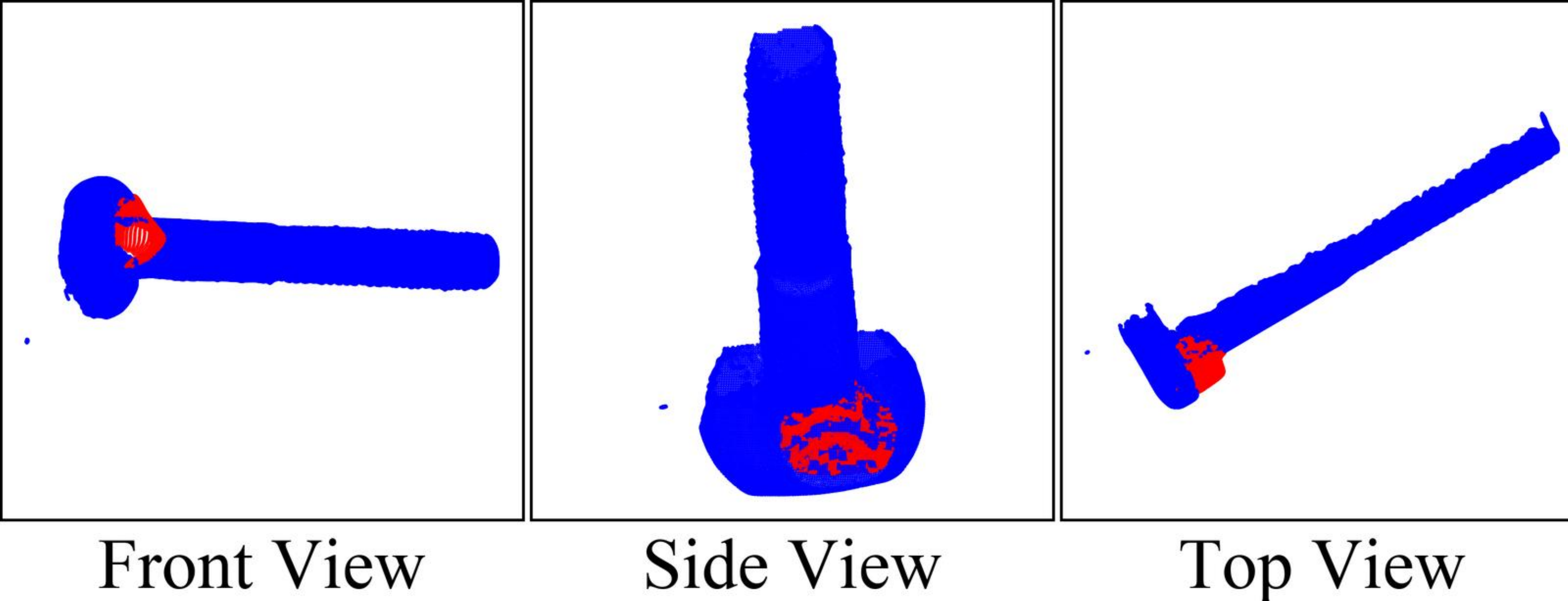}\\
    \small T Bolt Half thread Screw dent
\end{minipage}

\caption{Random examples from T Bolt Half thread Screw}
\label{fig: T_Bolt_Half_thread_Screw_saved}
\end{figure}

\begin{figure}[htbp]
\centering

\begin{minipage}{0.48\textwidth}
    \centering
    \includegraphics[width=\linewidth]{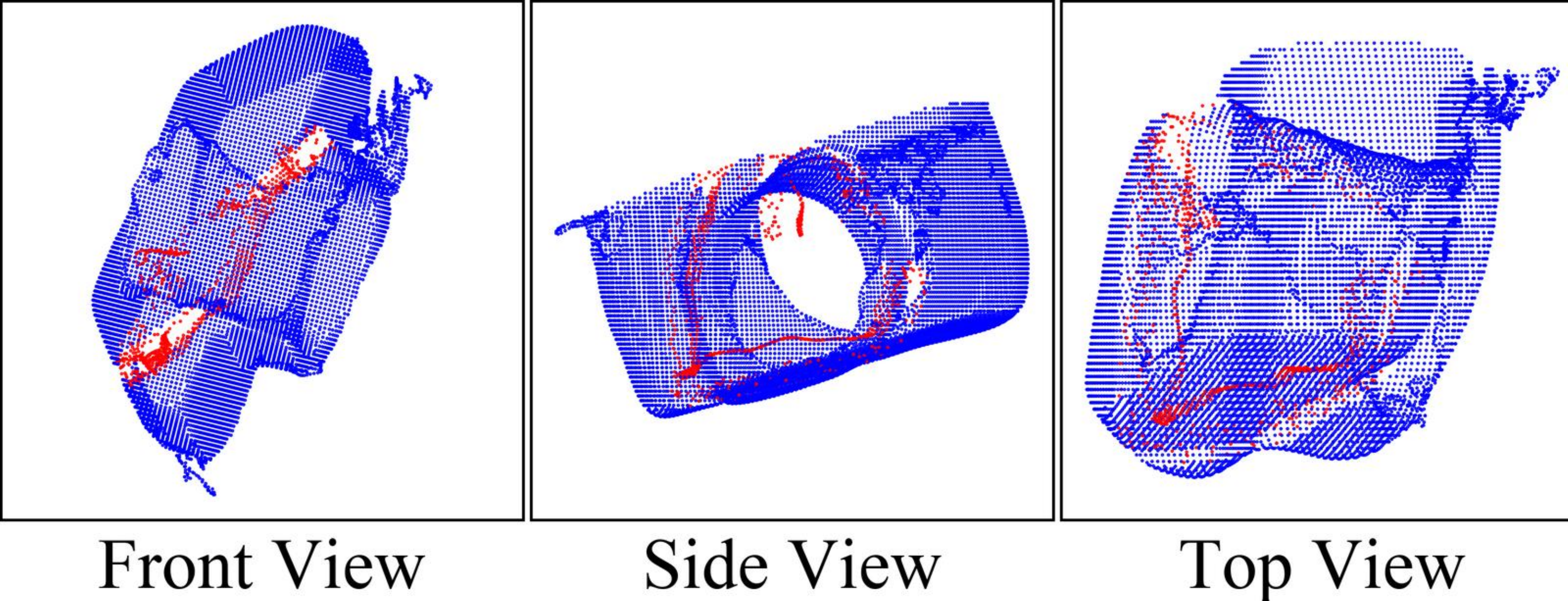}\\
    \small T Nut crack
\end{minipage}\hfill
\begin{minipage}{0.48\textwidth}
    \centering
    \includegraphics[width=\linewidth]{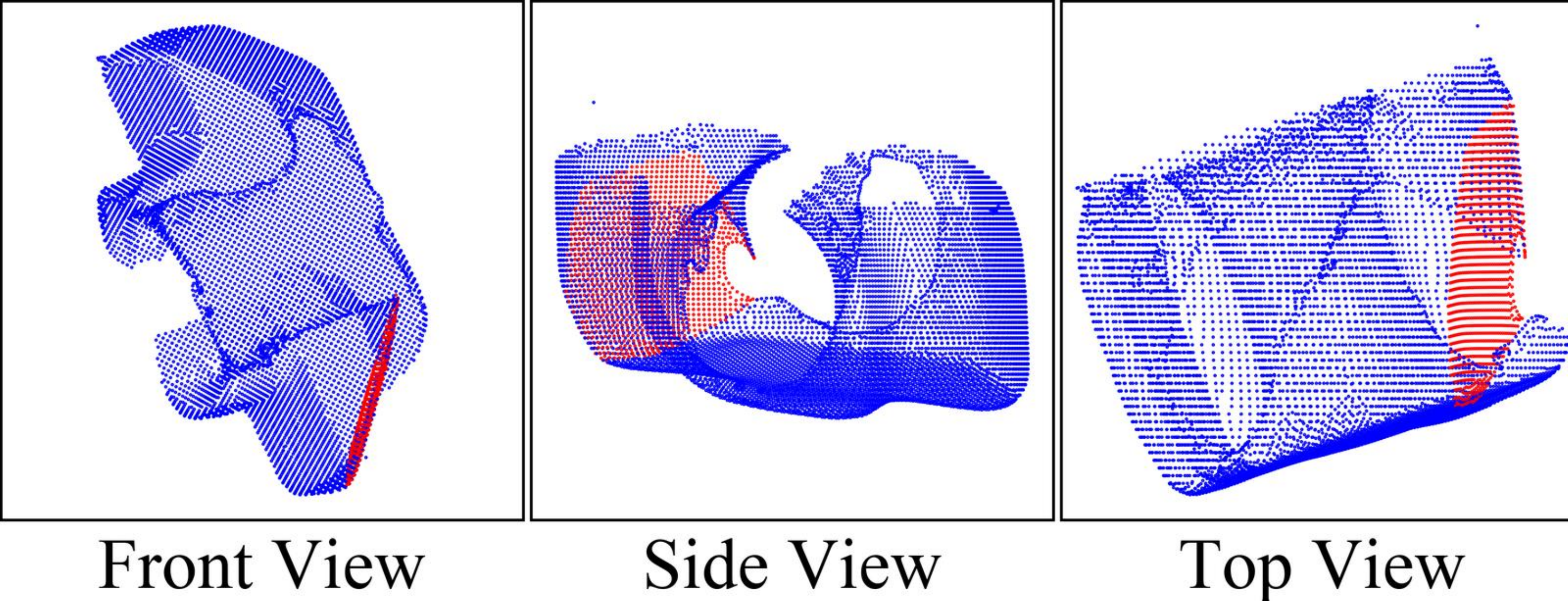}\\
    \small T Nut cut
\end{minipage}

\begin{minipage}{0.48\textwidth}
    \centering
    \includegraphics[width=\linewidth]{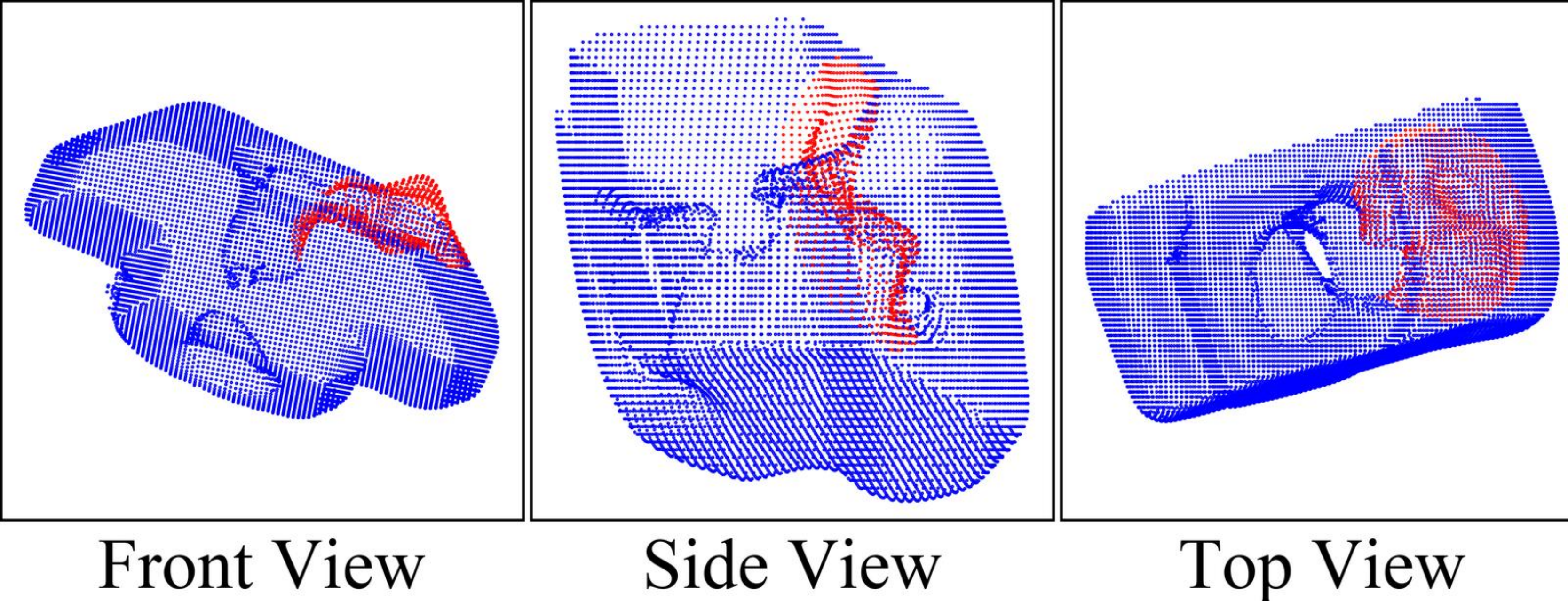}\\
    \small T Nut deformation
\end{minipage}\hfill
\begin{minipage}{0.48\textwidth}
    \centering
    \includegraphics[width=\linewidth]{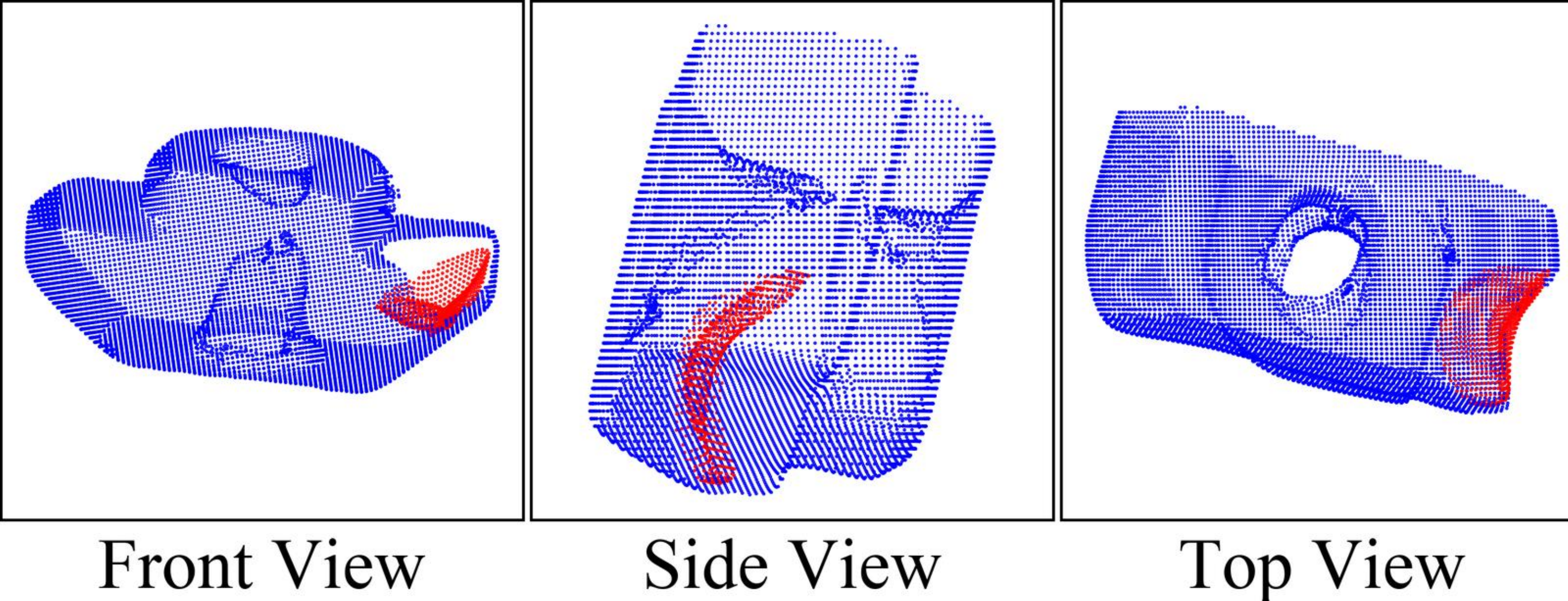}\\
    \small T Nut dent
\end{minipage}

\caption{Random examples from T Nut}
\label{fig: T_Nut_saved}
\end{figure}

\begin{figure}[htbp]
\centering

\begin{minipage}{0.48\textwidth}
    \centering
    \includegraphics[width=\linewidth]{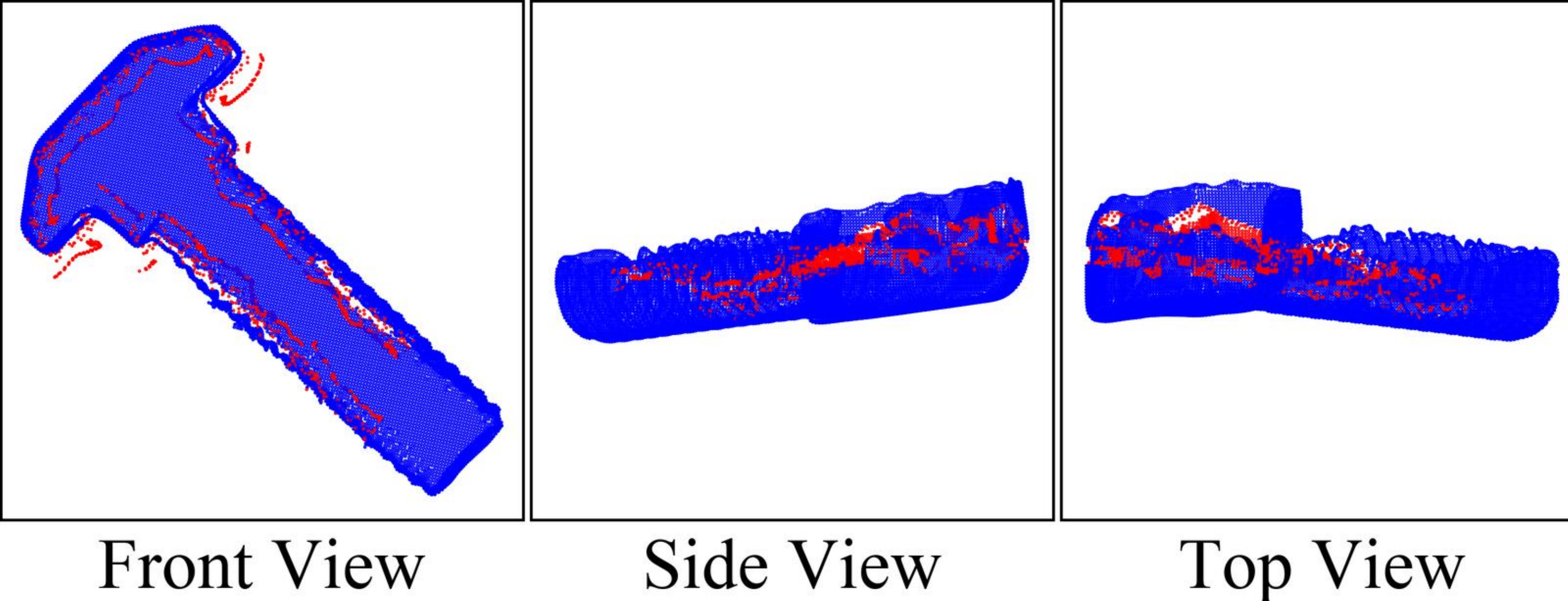}\\
    \small T Screw crack
\end{minipage}\hfill
\begin{minipage}{0.48\textwidth}
    \centering
    \includegraphics[width=\linewidth]{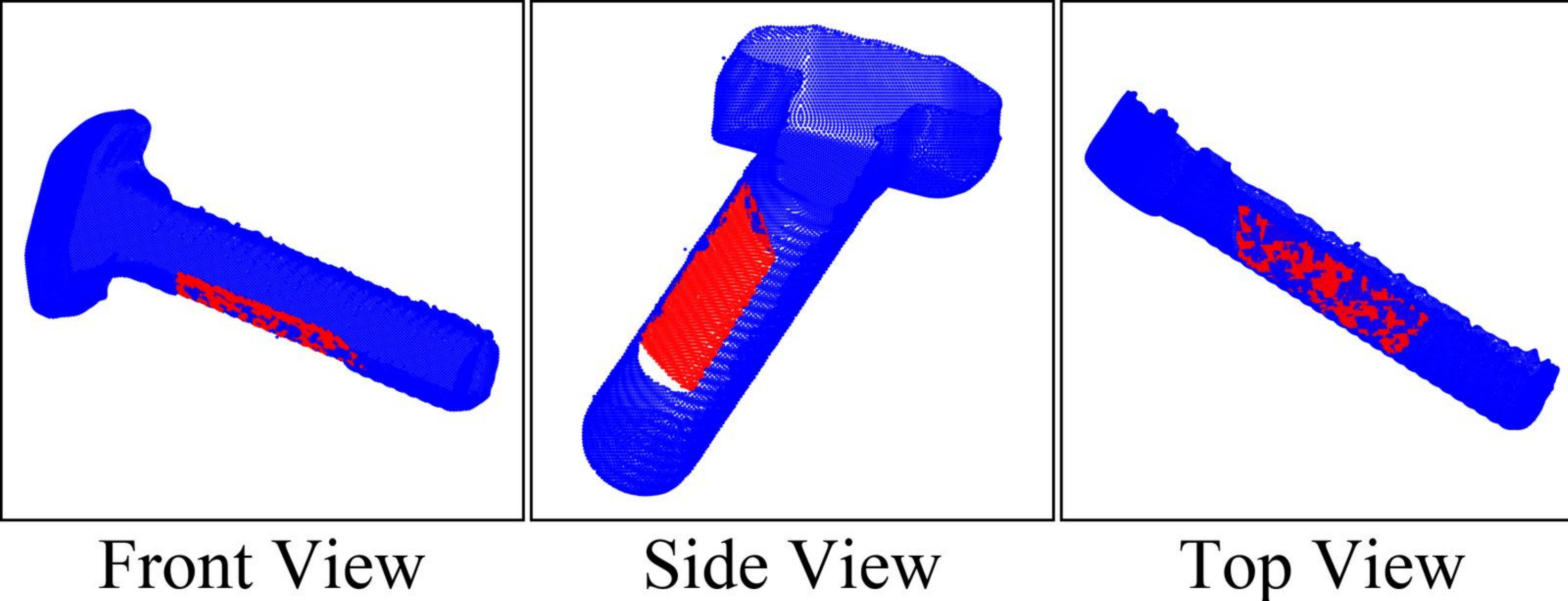}\\
    \small T Screw cut
\end{minipage}

\begin{minipage}{0.48\textwidth}
    \centering
    \includegraphics[width=\linewidth]{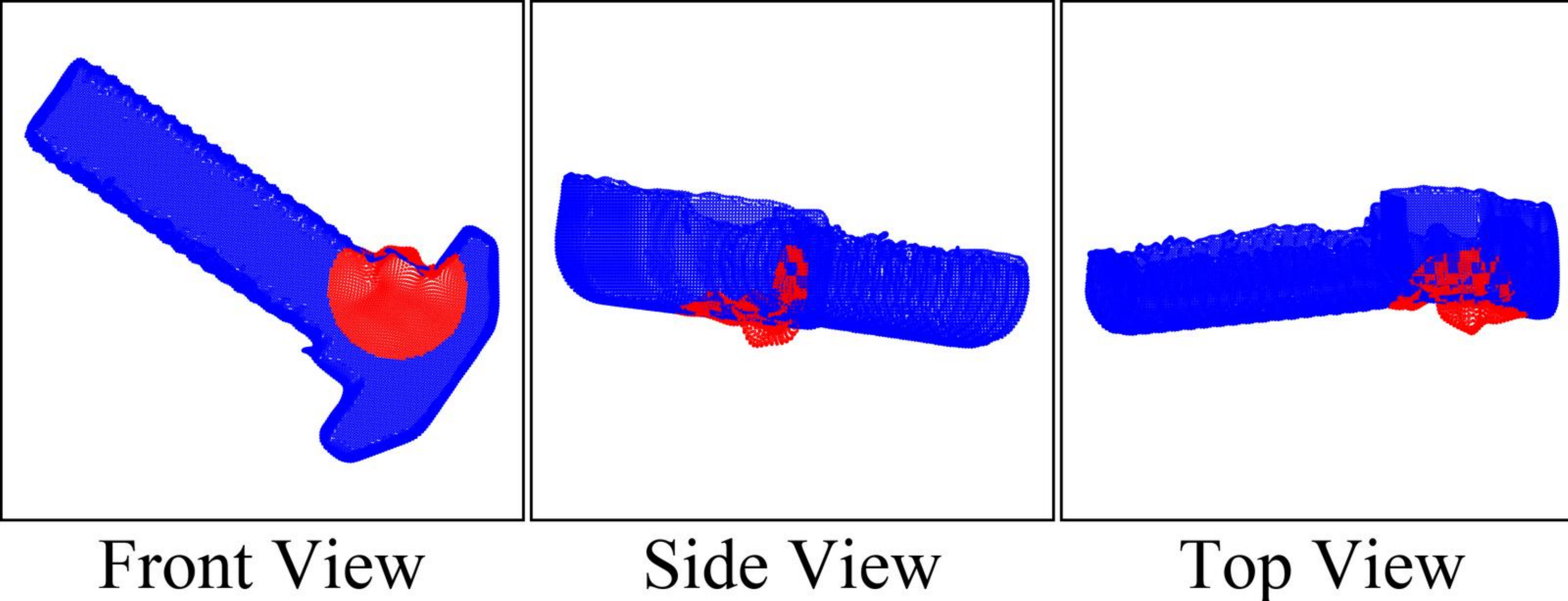}\\
    \small T Screw deformation
\end{minipage}\hfill
\begin{minipage}{0.48\textwidth}
    \centering
    \includegraphics[width=\linewidth]{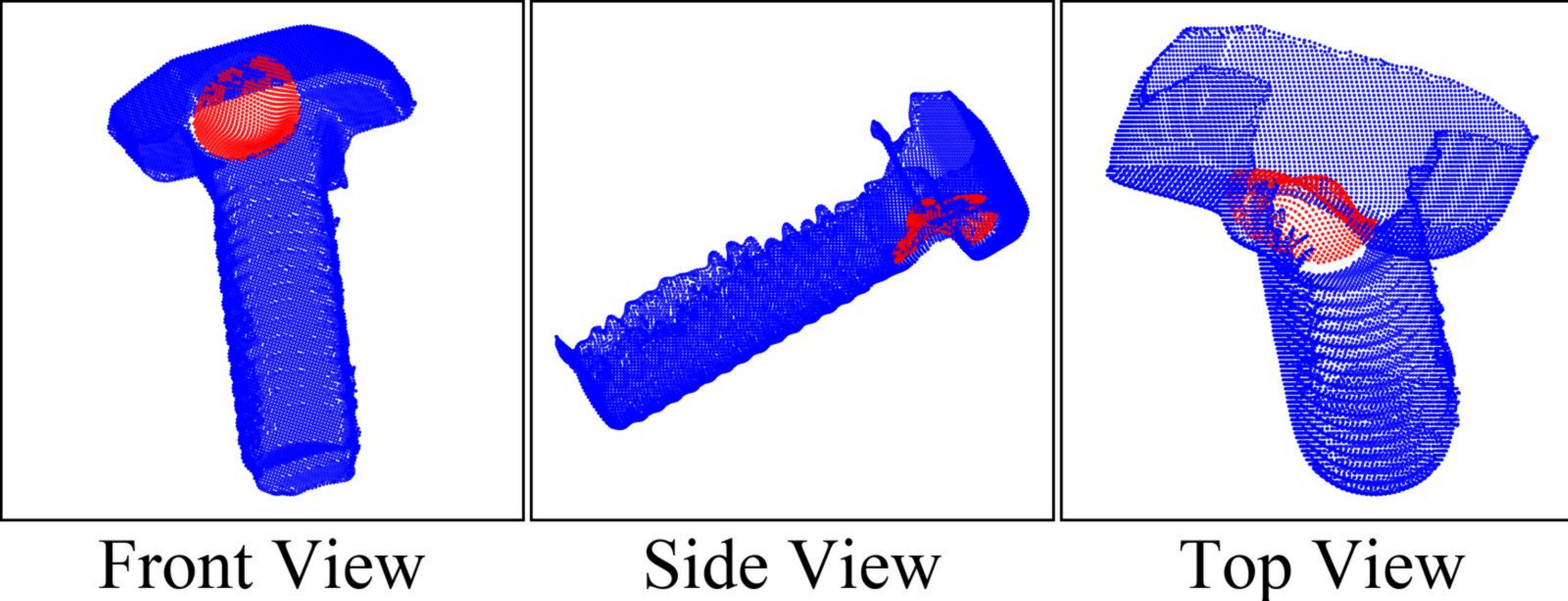}\\
    \small T Screw dent
\end{minipage}

\caption{Random examples from T Screw}
\label{fig: T_Screw_saved}
\end{figure}

\begin{figure}[htbp]
\centering

\begin{minipage}{0.48\textwidth}
    \centering
    \includegraphics[width=\linewidth]{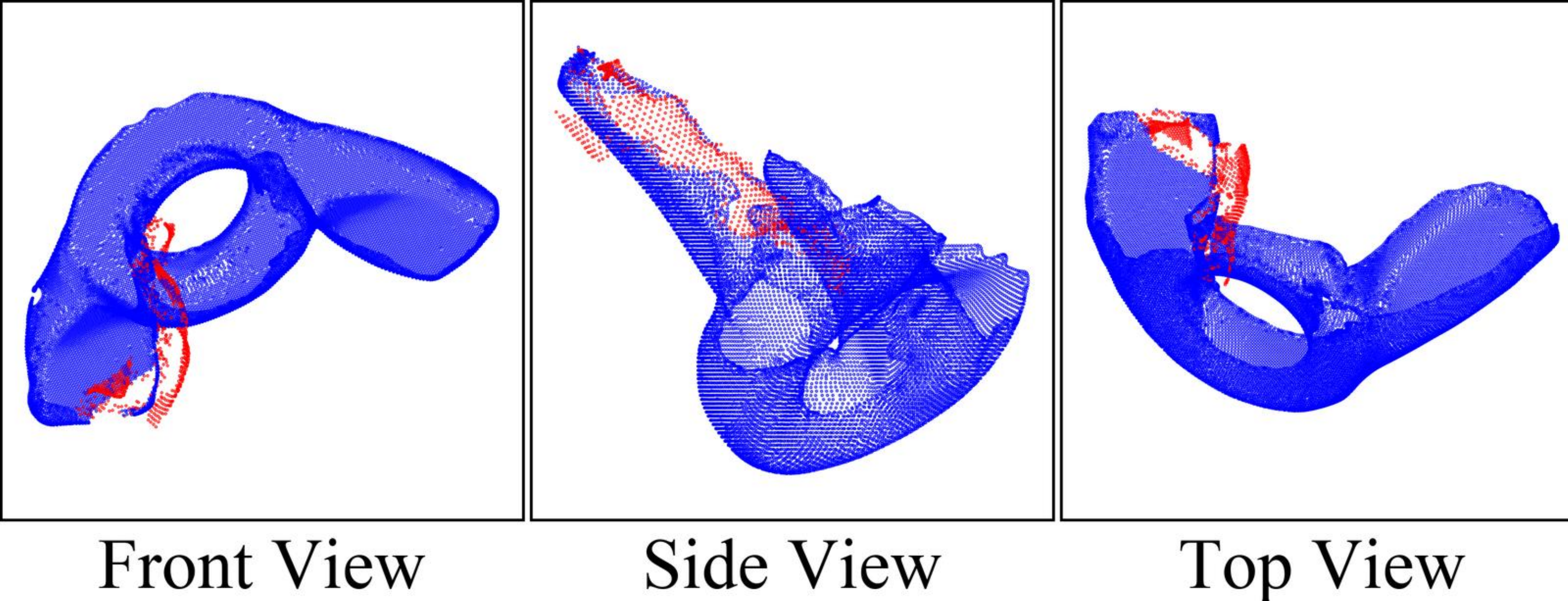}\\
    \small Wing Nut crack
\end{minipage}\hfill
\begin{minipage}{0.48\textwidth}
    \centering
    \includegraphics[width=\linewidth]{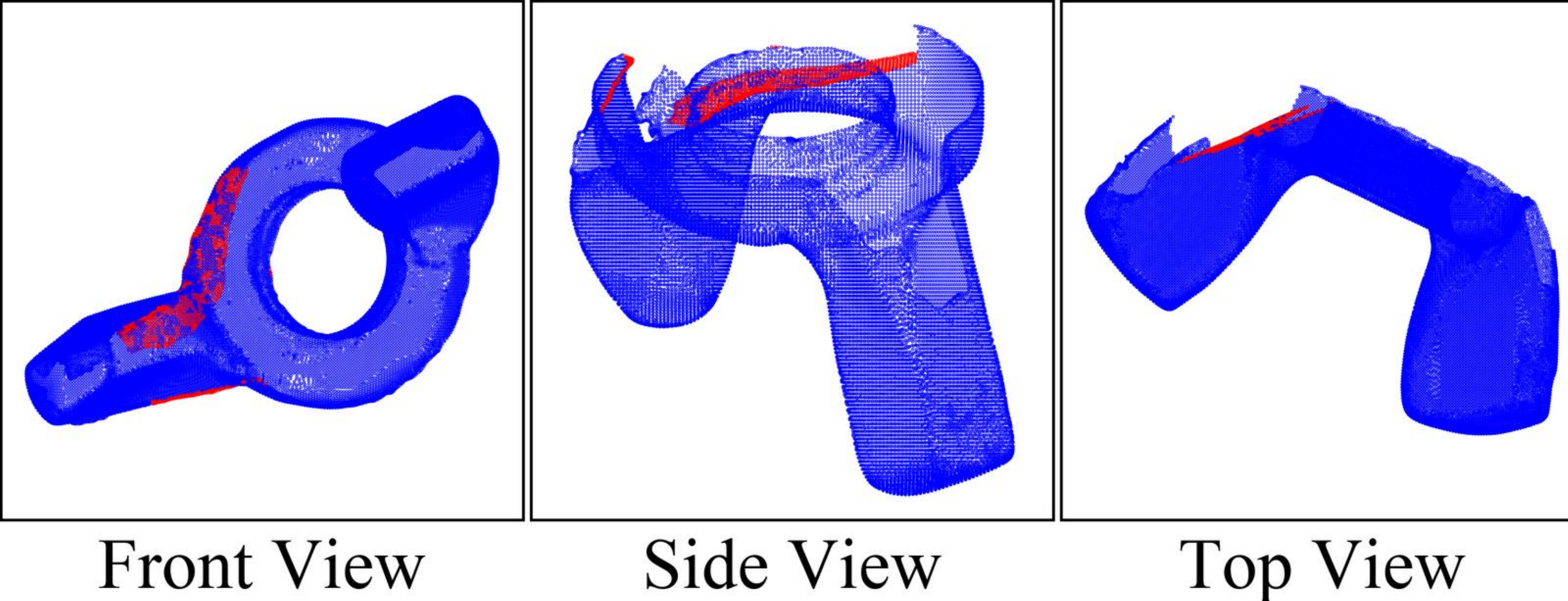}\\
    \small Wing Nut cut
\end{minipage}

\begin{minipage}{0.48\textwidth}
    \centering
    \includegraphics[width=\linewidth]{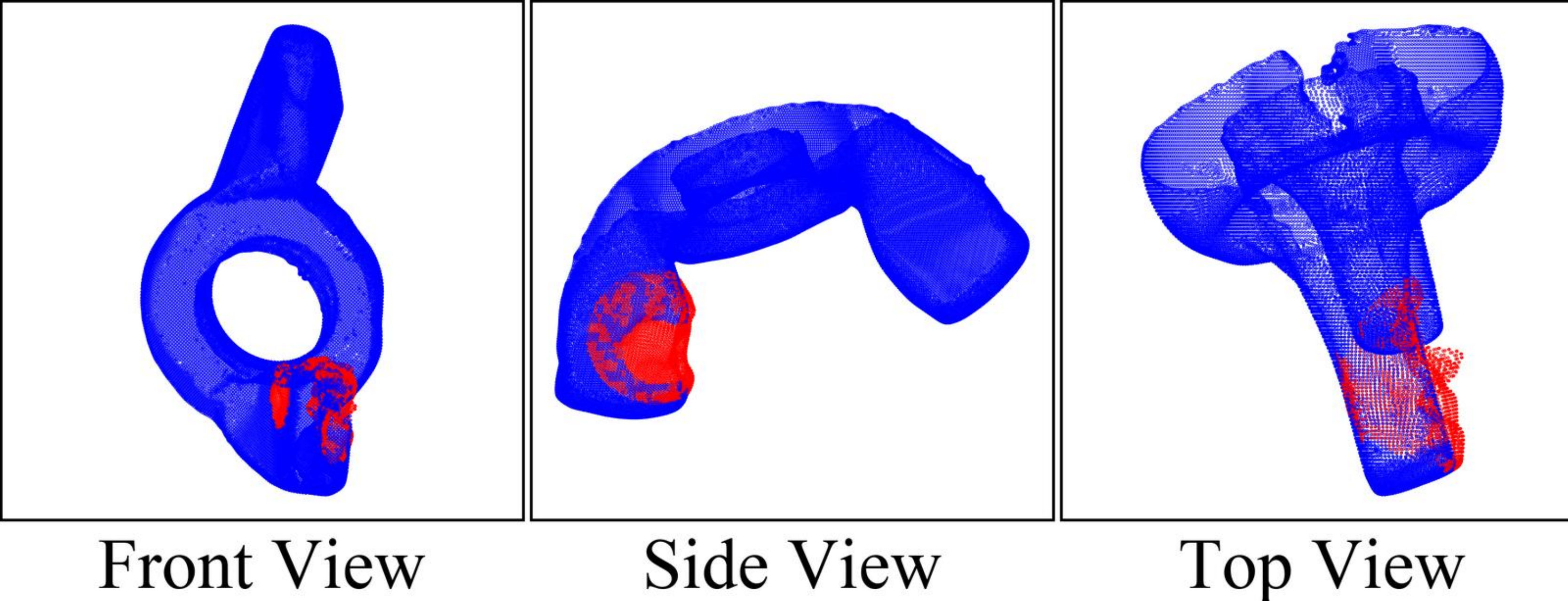}\\
    \small Wing Nut deformation
\end{minipage}\hfill
\begin{minipage}{0.48\textwidth}
    \centering
    \includegraphics[width=\linewidth]{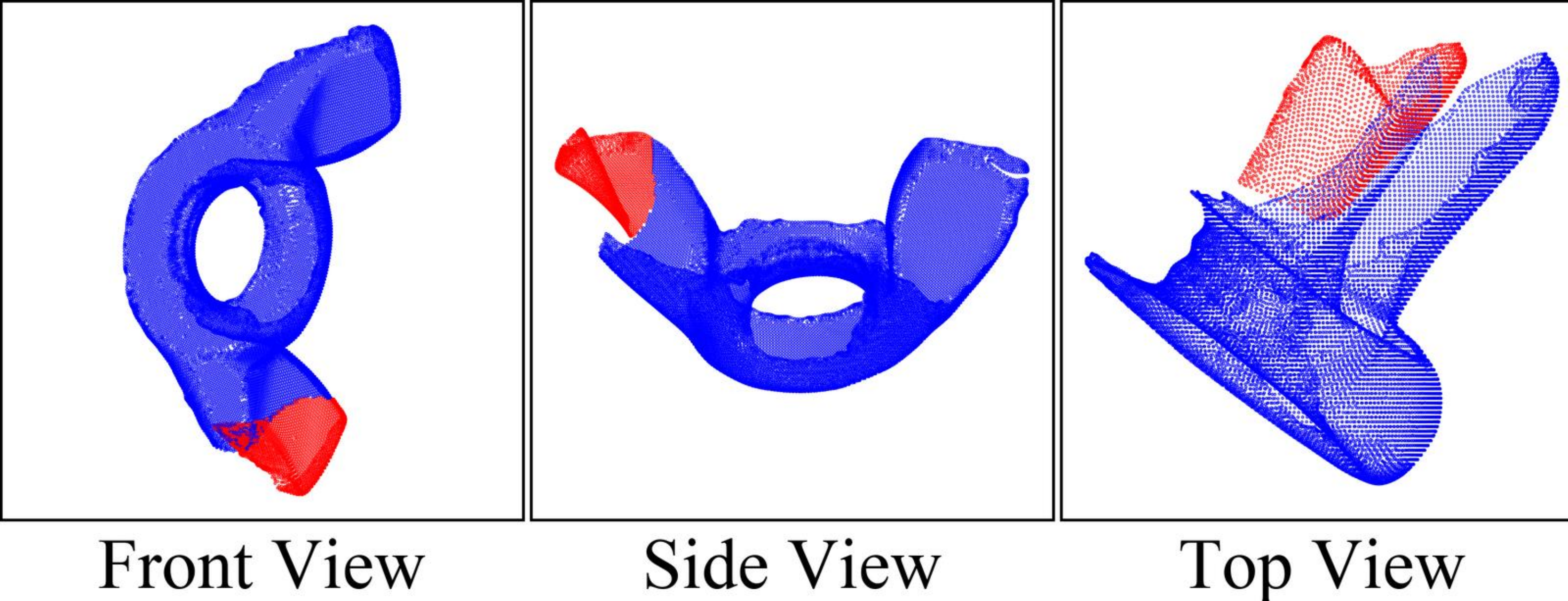}\\
    \small Wing Nut dent
\end{minipage}

\caption{Random examples from Wing Nut}
\label{fig: Wing_Nut_saved}
\end{figure}

\begin{figure}[htbp]
\centering

\begin{minipage}{0.48\textwidth}
    \centering
    \includegraphics[width=\linewidth]{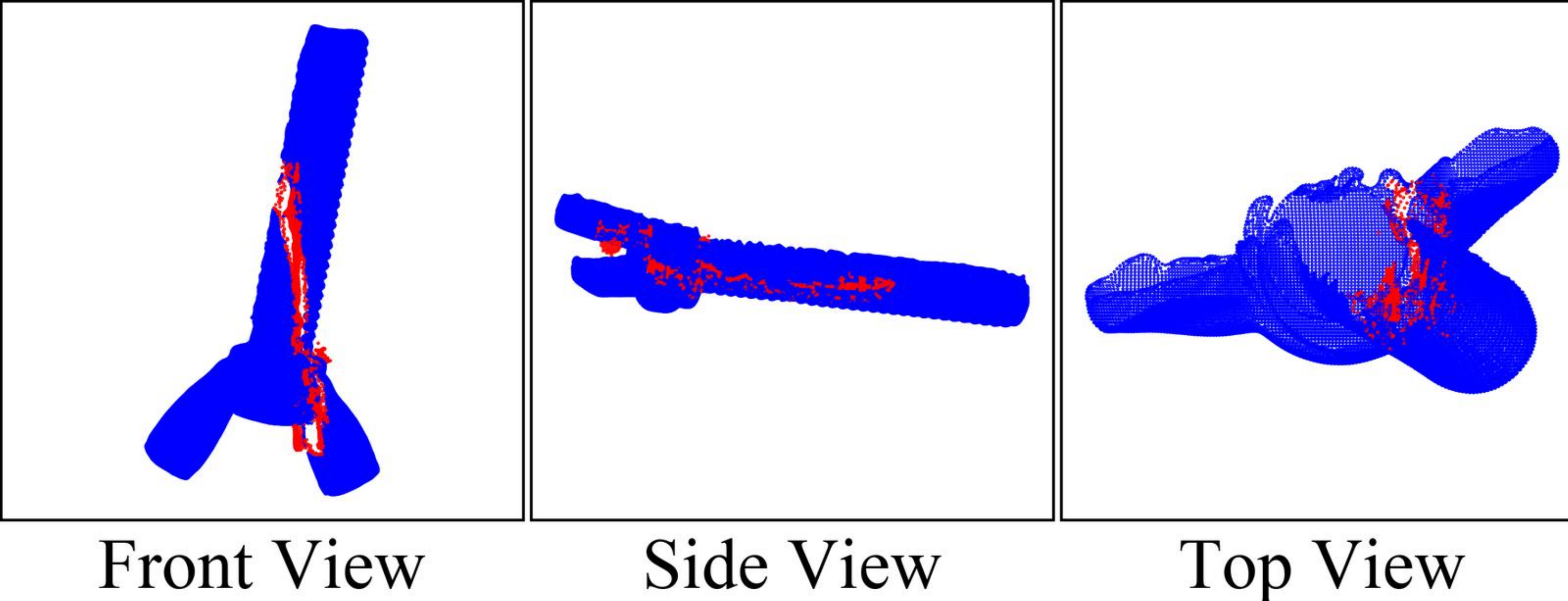}\\
    \small Wing Screw crack
\end{minipage}\hfill
\begin{minipage}{0.48\textwidth}
    \centering
    \includegraphics[width=\linewidth]{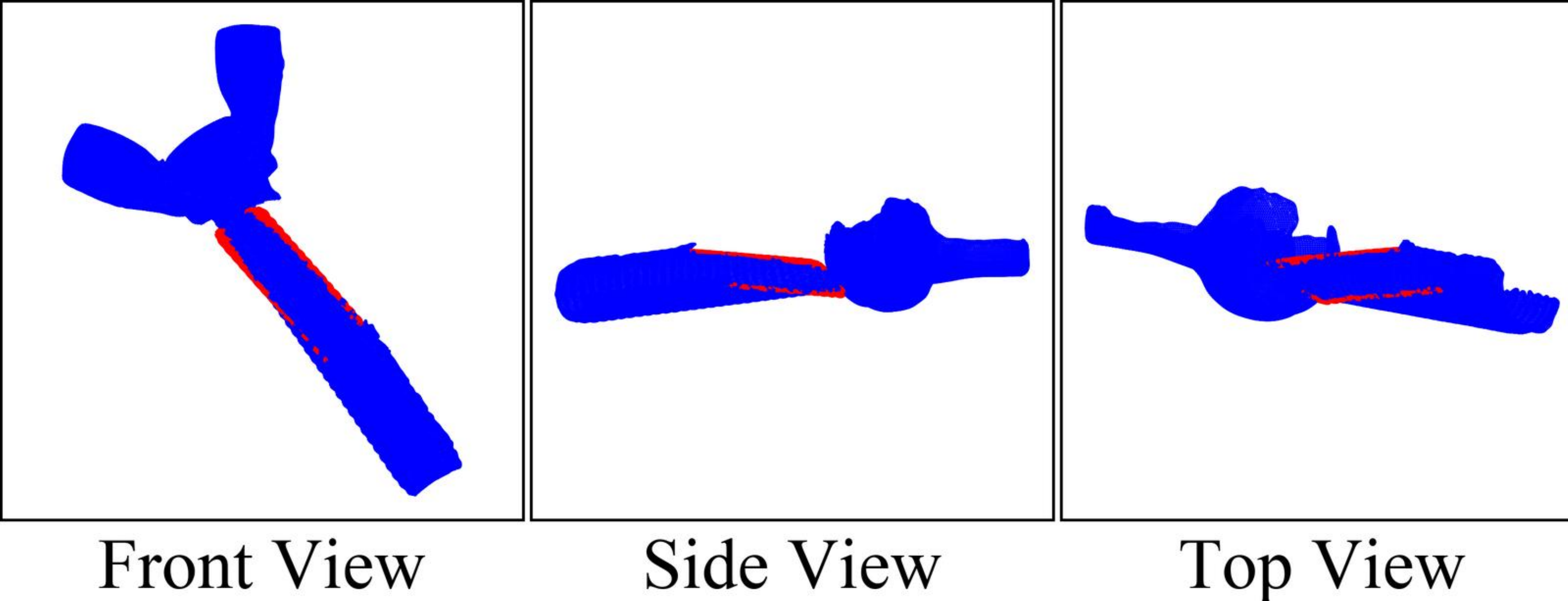}\\
    \small Wing Screw cut
\end{minipage}

\begin{minipage}{0.48\textwidth}
    \centering
    \includegraphics[width=\linewidth]{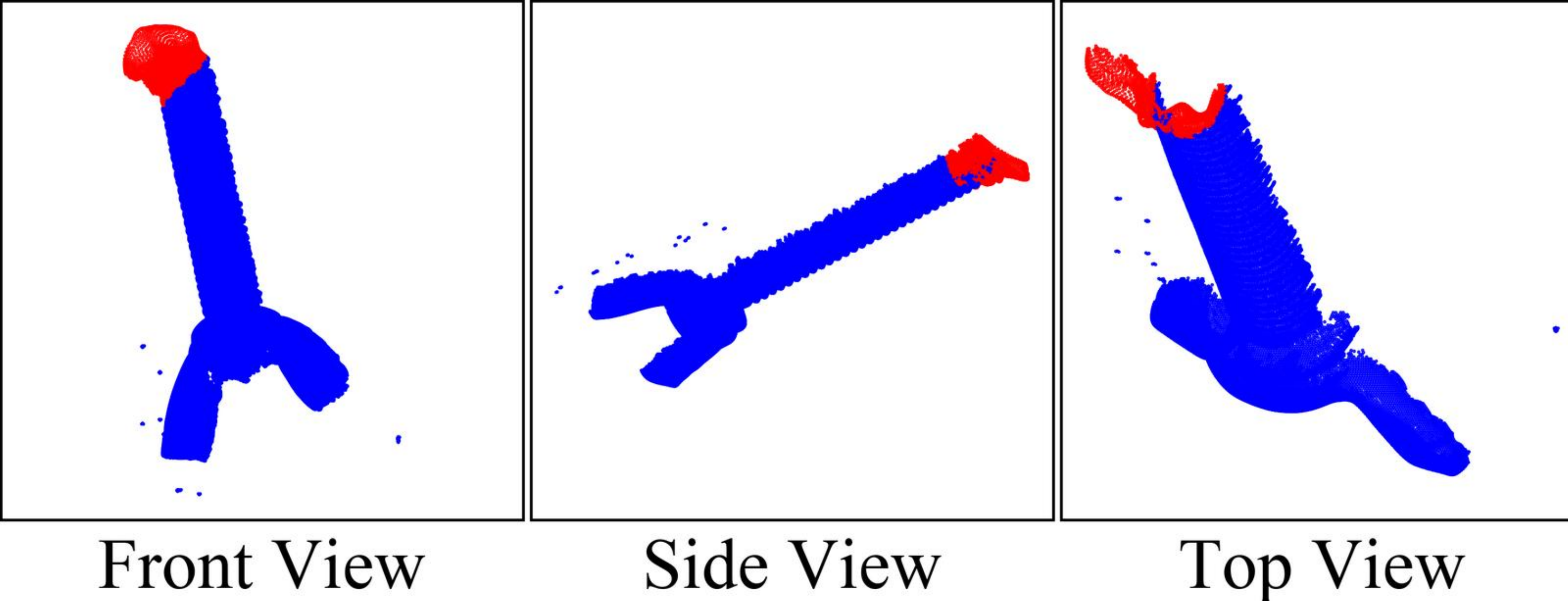}\\
    \small Wing Screw deformation
\end{minipage}\hfill
\begin{minipage}{0.48\textwidth}
    \centering
    \includegraphics[width=\linewidth]{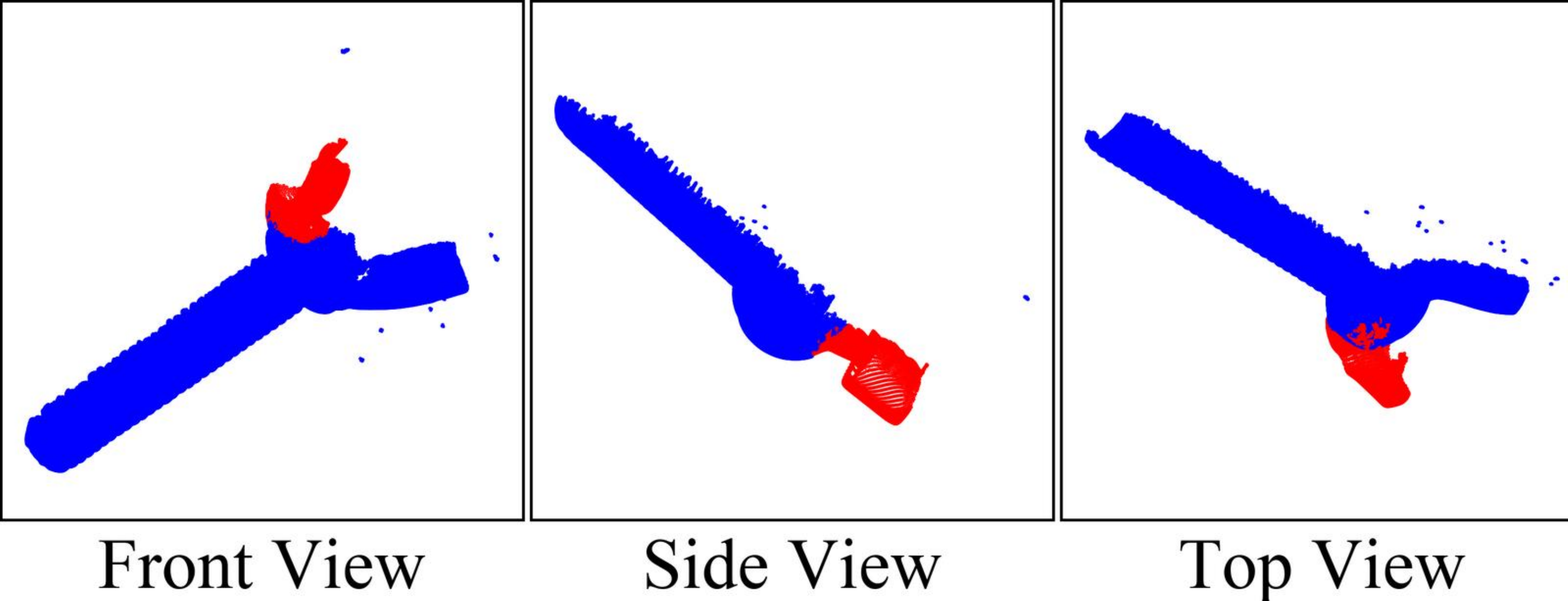}\\
    \small Wing Screw dent
\end{minipage}

\caption{Random examples from Wing Screw}
\label{fig: Wing_Screw_saved}
\end{figure}

\begin{figure}[htbp]
\centering

\begin{minipage}{0.23\textwidth}
    \centering
    \parbox[c][3.0cm][c]{\linewidth}{
        \centering\includegraphics[width=\linewidth,height=3.0cm,keepaspectratio]{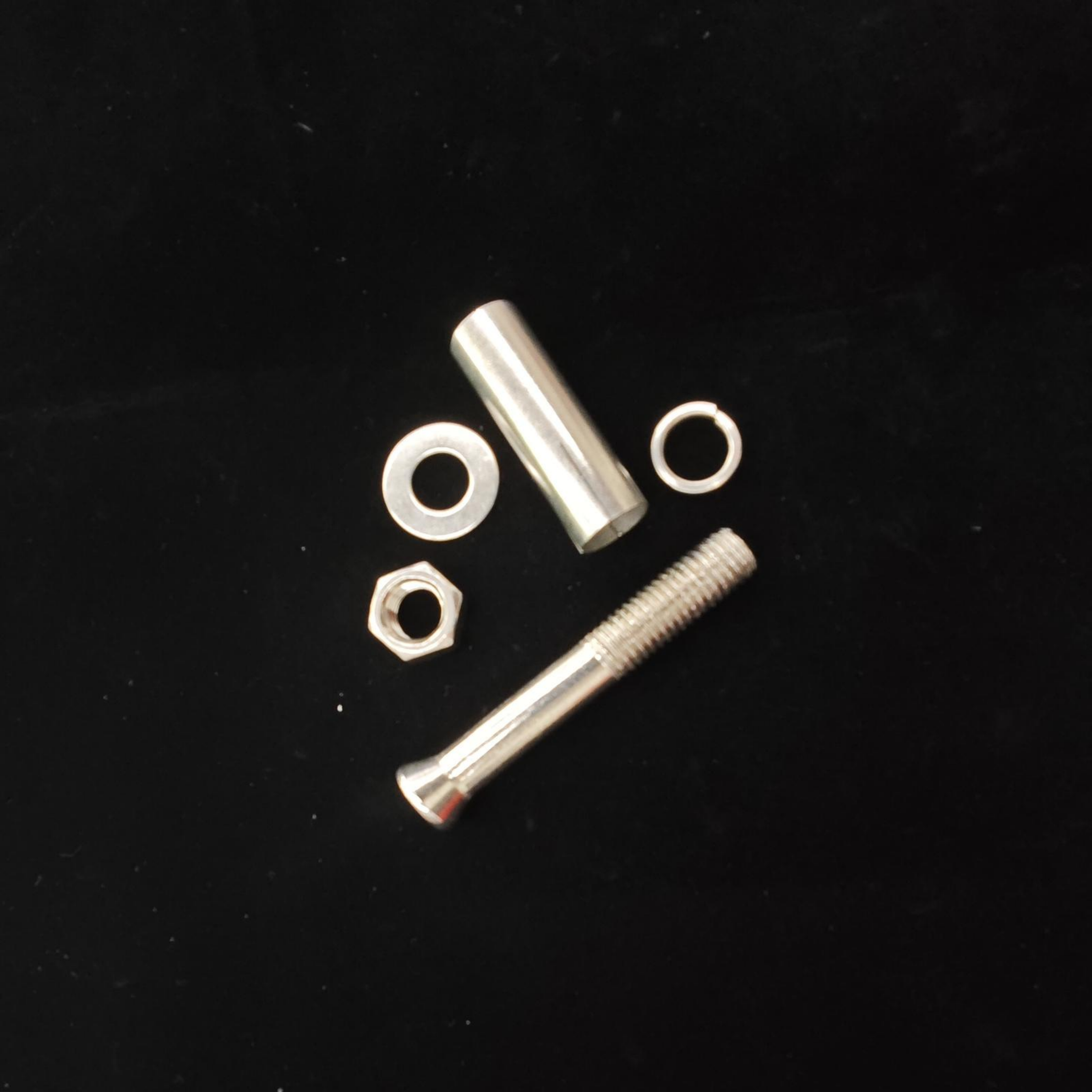}
    }
    \parbox[t][1.5cm][t]{\linewidth}{\small Normal Case}
\end{minipage}\hfill
\begin{minipage}{0.23\textwidth}
    \centering
    \parbox[c][3.0cm][c]{\linewidth}{
        \centering\includegraphics[width=\linewidth,height=3.0cm,keepaspectratio]{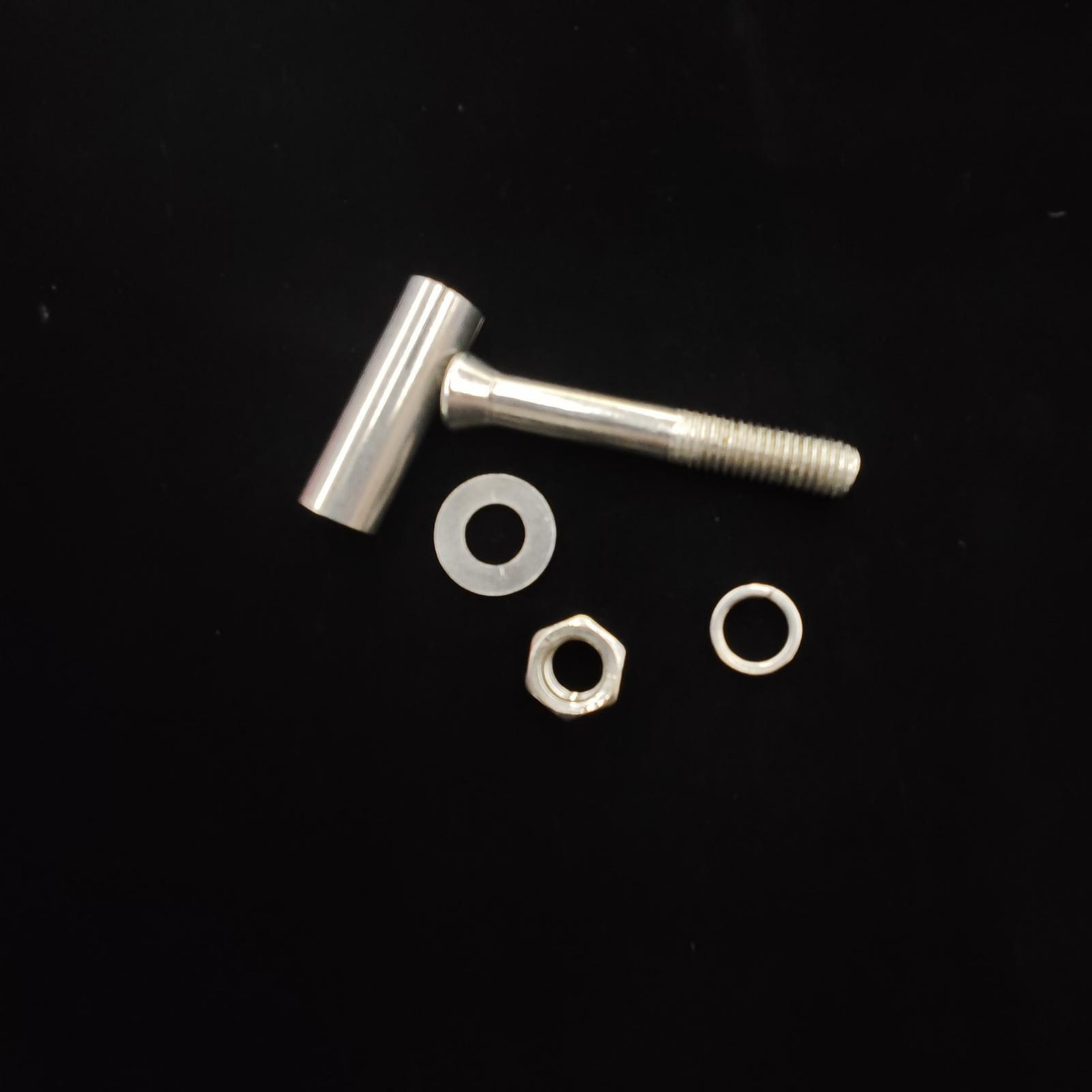}
    }
    \parbox[t][1.5cm][t]{\linewidth}{\small Normal Case}
\end{minipage}\hfill
\begin{minipage}{0.23\textwidth}
    \centering
    \parbox[c][3.0cm][c]{\linewidth}{
        \centering\includegraphics[width=\linewidth,height=3.0cm,keepaspectratio]{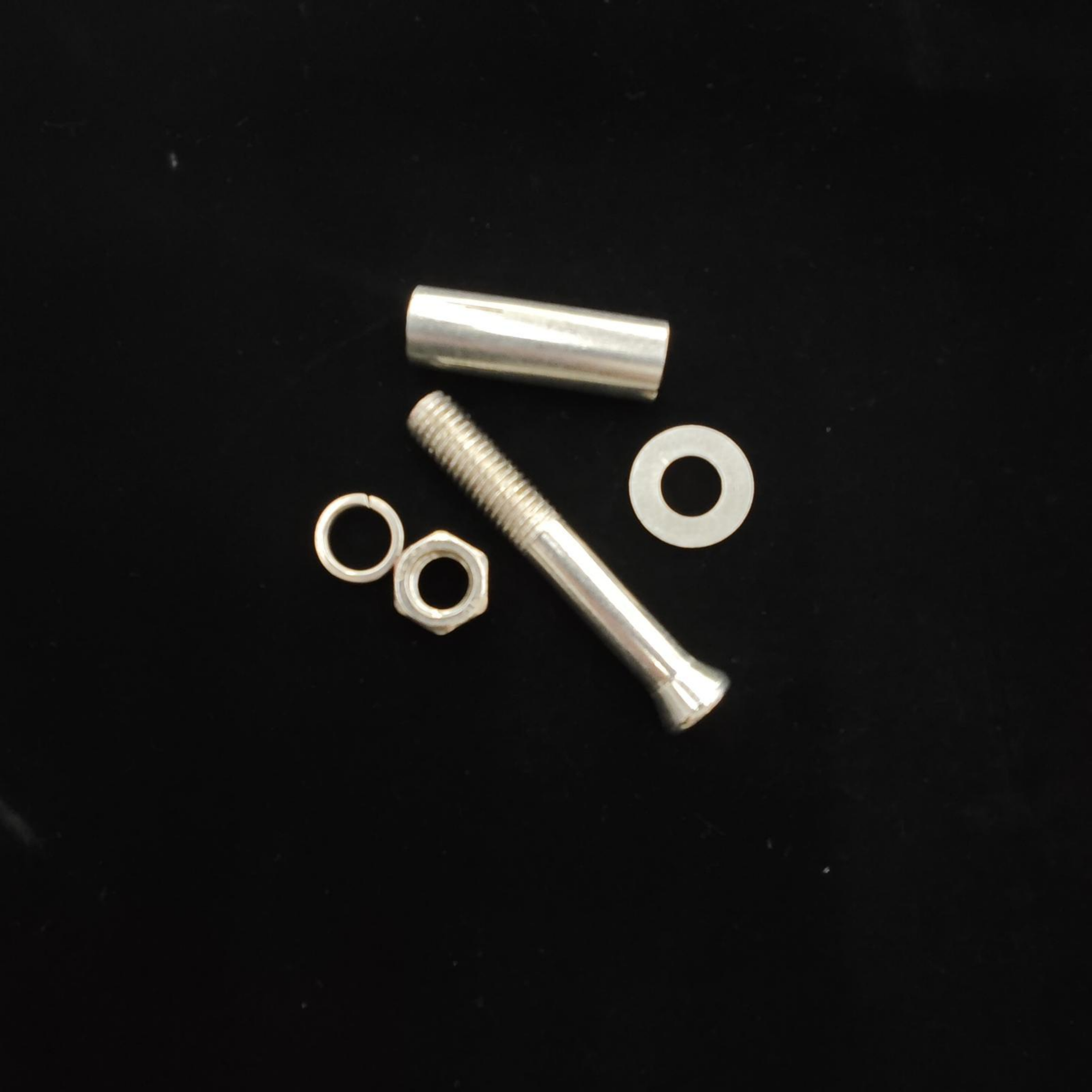}
    }
    \parbox[t][1.5cm][t]{\linewidth}{\small Normal Case}
\end{minipage}\hfill
\begin{minipage}{0.23\textwidth}
    \centering
    \parbox[c][3.0cm][c]{\linewidth}{
        \centering\includegraphics[width=\linewidth,height=3.0cm,keepaspectratio]{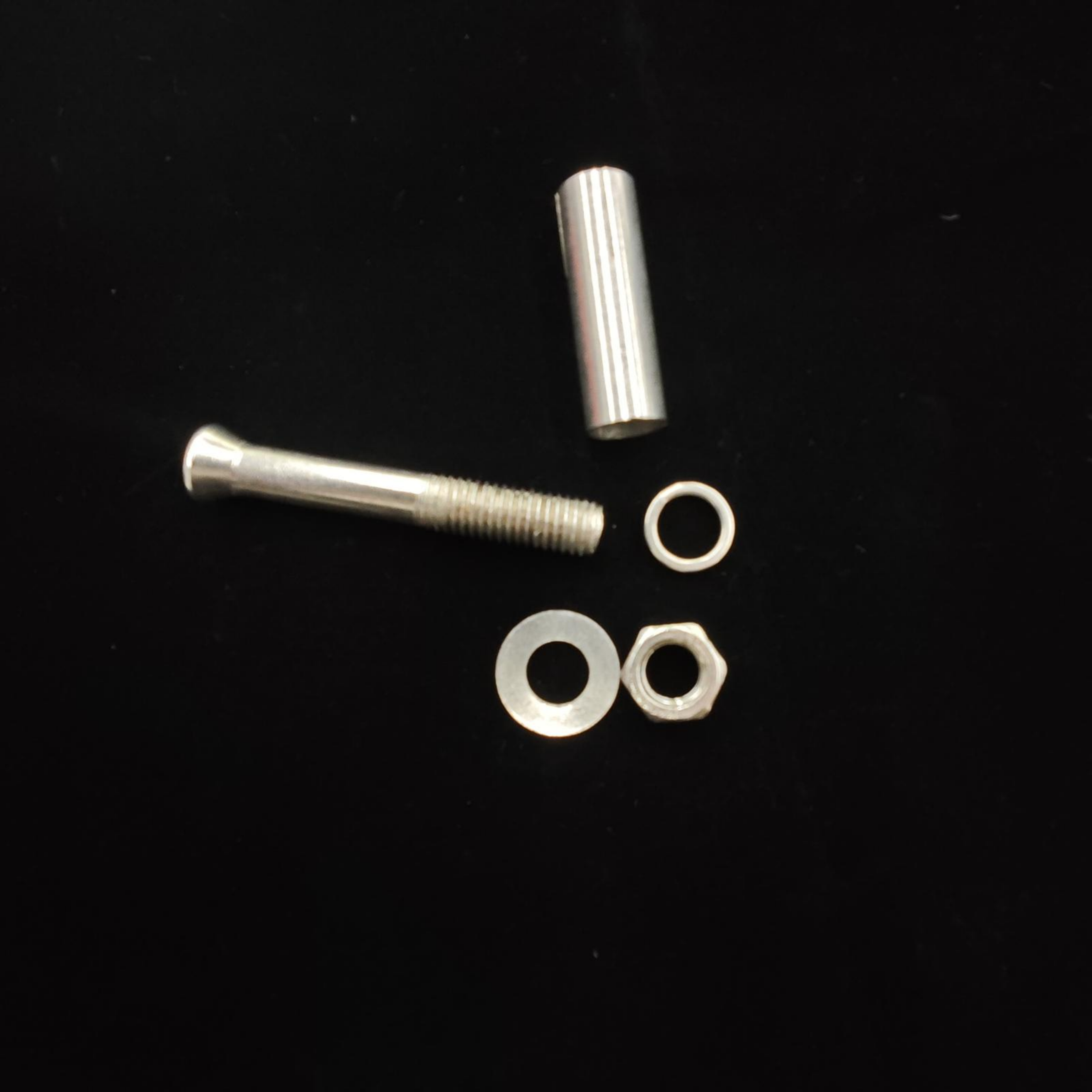}
    }
    \parbox[t][1.5cm][t]{\linewidth}{\small Normal Case}
\end{minipage}

\begin{minipage}{0.23\textwidth}
    \centering
    \parbox[c][3.0cm][c]{\linewidth}{
        \centering\includegraphics[width=\linewidth,height=3.0cm,keepaspectratio]{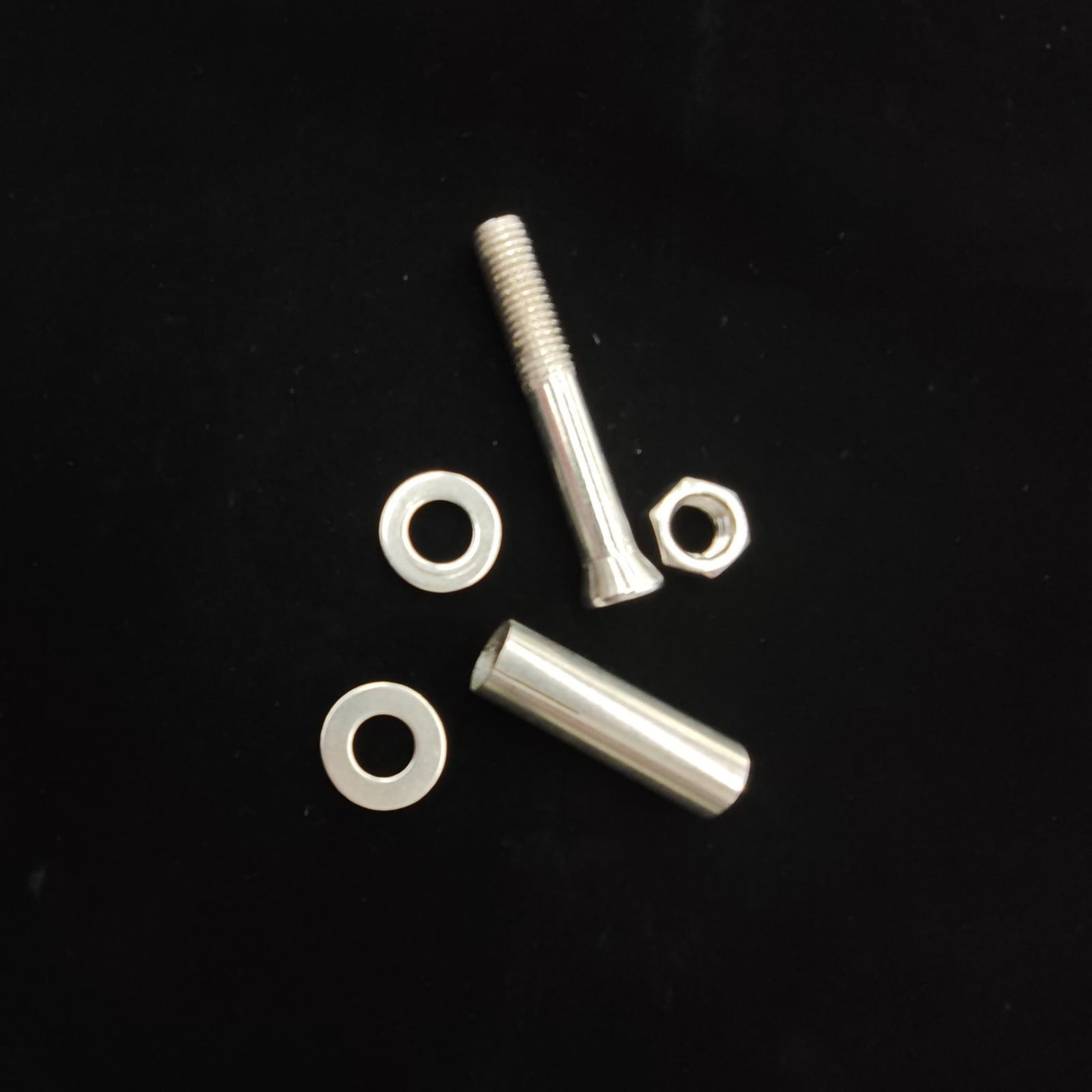}
    }
    \parbox[t][1.5cm][t]{\linewidth}{\small Two Flat Washers}
\end{minipage}\hfill
\begin{minipage}{0.23\textwidth}
    \centering
    \parbox[c][3.0cm][c]{\linewidth}{
        \centering\includegraphics[width=\linewidth,height=3.0cm,keepaspectratio]{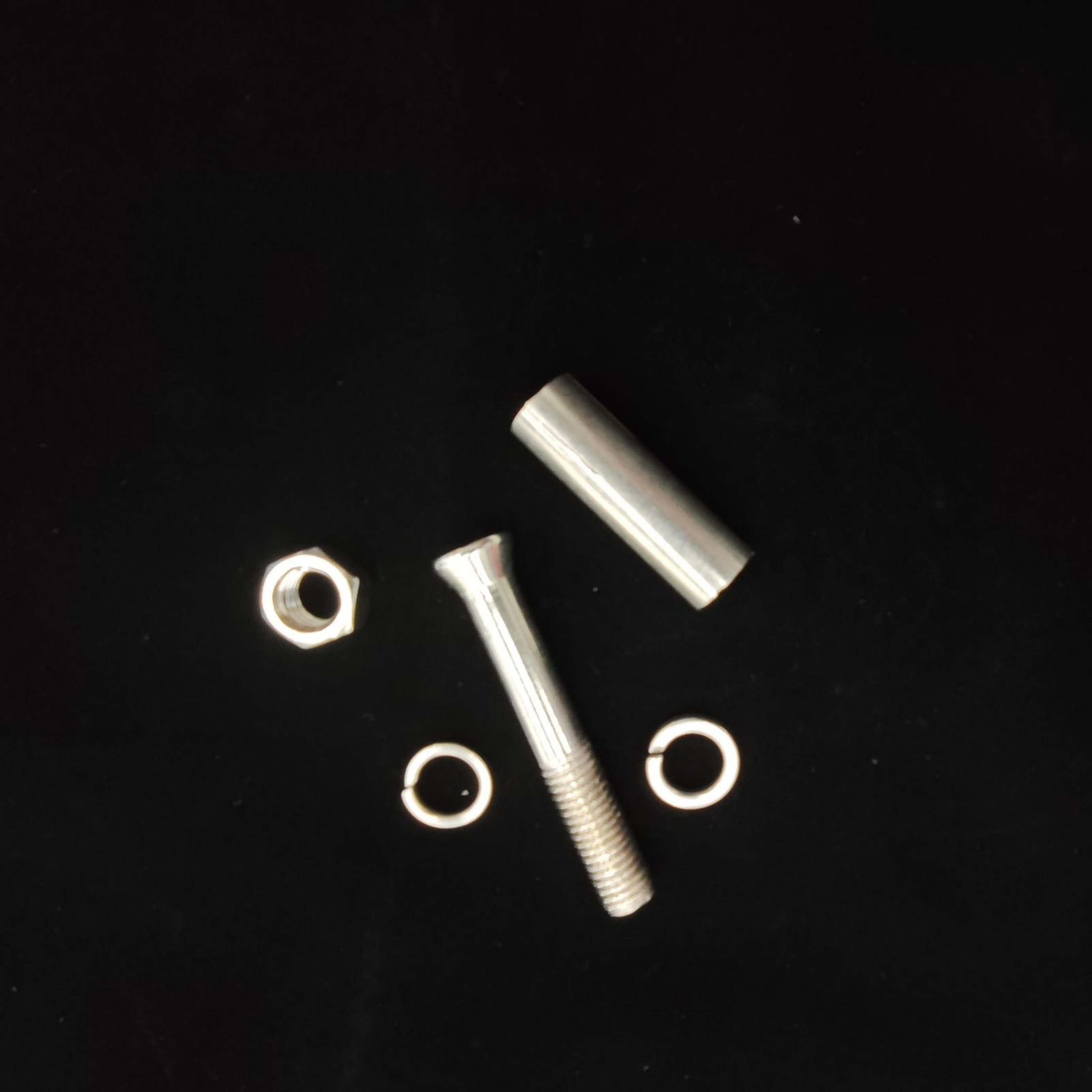}
    }
    \parbox[t][1.5cm][t]{\linewidth}{\small Two Spring Washers}
\end{minipage}\hfill
\begin{minipage}{0.23\textwidth}
    \centering
    \parbox[c][3.0cm][c]{\linewidth}{
        \centering\includegraphics[width=\linewidth,height=3.0cm,keepaspectratio]{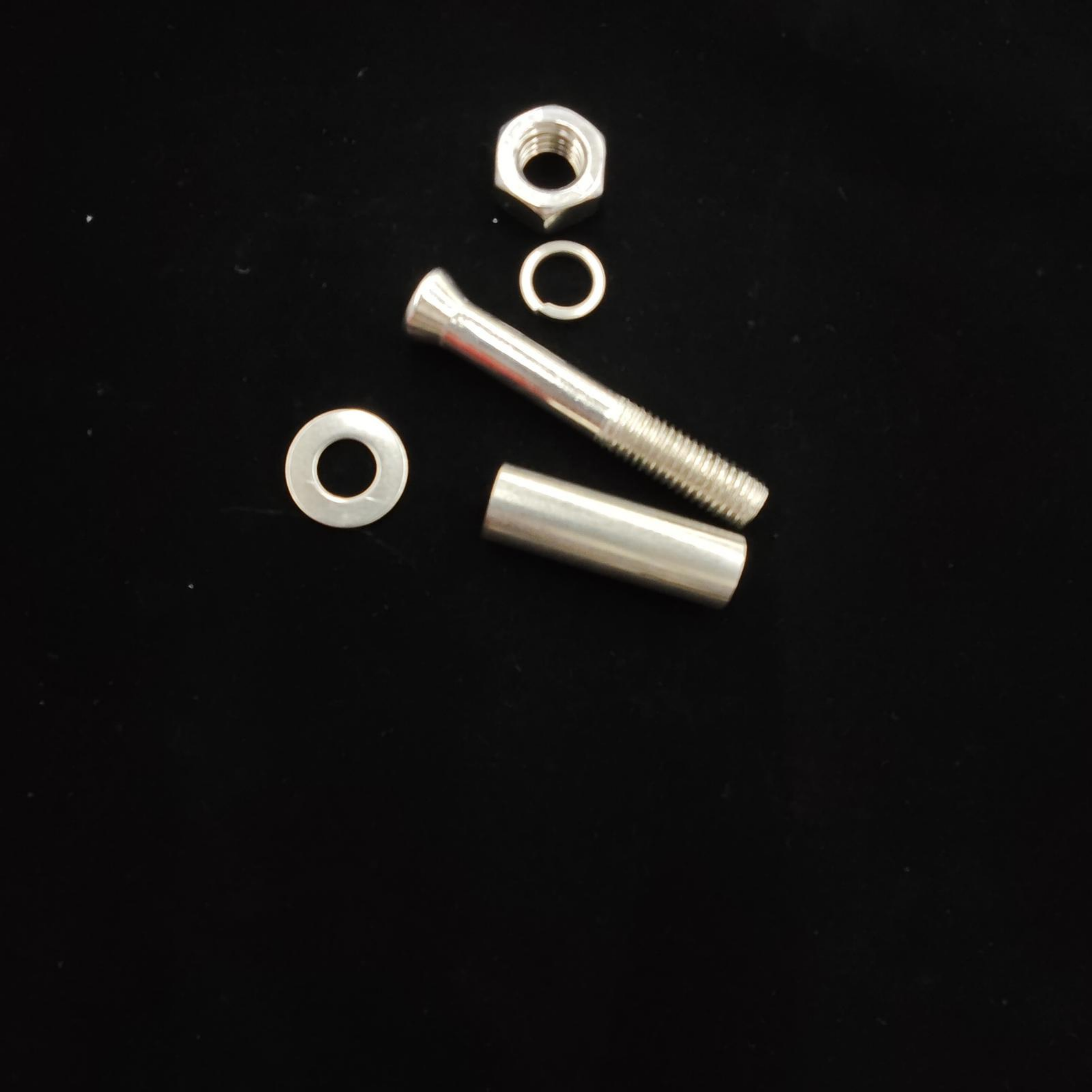}
    }
    \parbox[t][1.5cm][t]{\linewidth}{\small Hex Nut M14}
\end{minipage}\hfill
\begin{minipage}{0.23\textwidth}
    \centering
    \parbox[c][3.0cm][c]{\linewidth}{
        \centering\includegraphics[width=\linewidth,height=3.0cm,keepaspectratio]{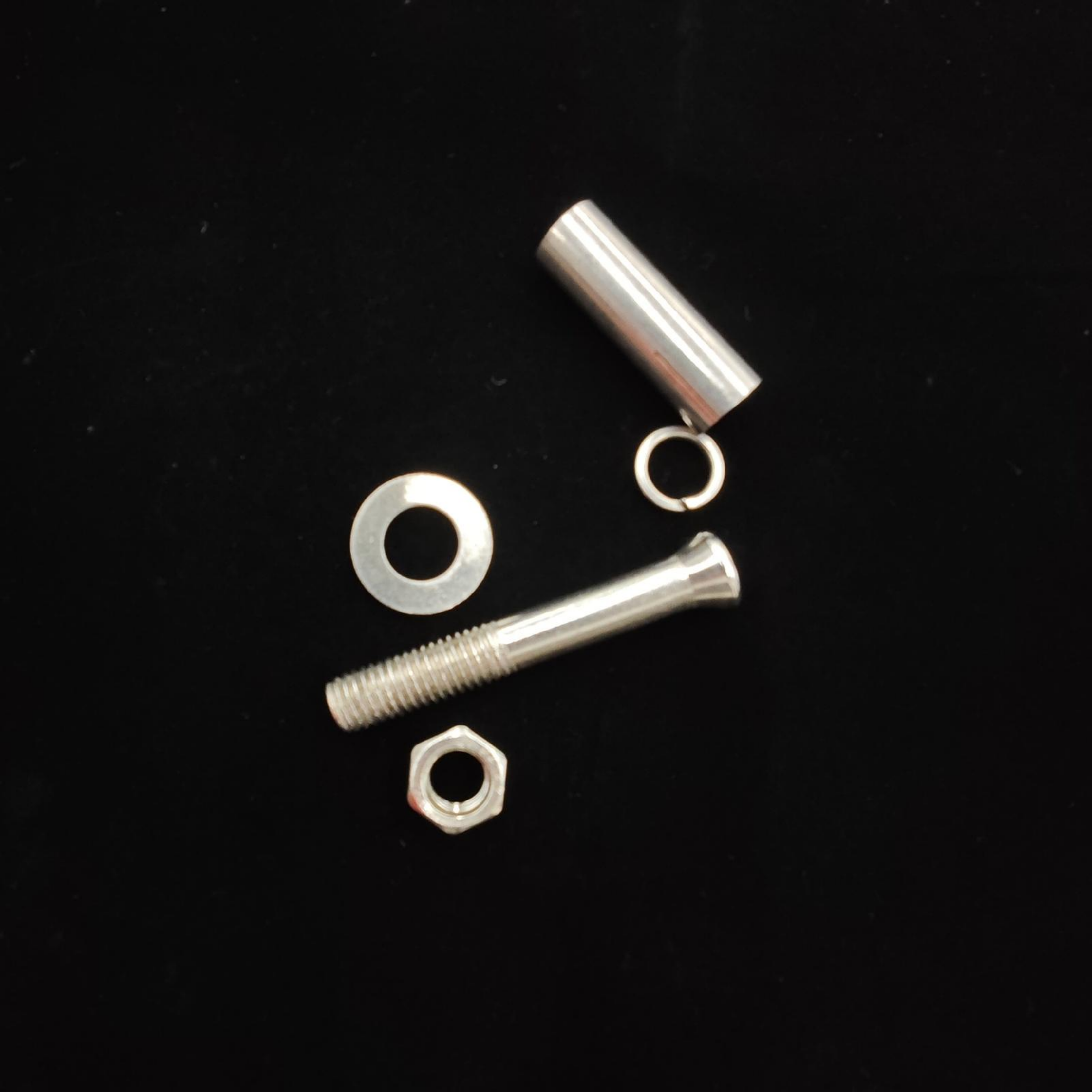}
    }
    \parbox[t][1.5cm][t]{\linewidth}{\small Flat Washer M14}
\end{minipage}

\begin{minipage}{0.23\textwidth}
    \centering
    \parbox[c][3.0cm][c]{\linewidth}{
        \centering\includegraphics[width=\linewidth,height=3.0cm,keepaspectratio]{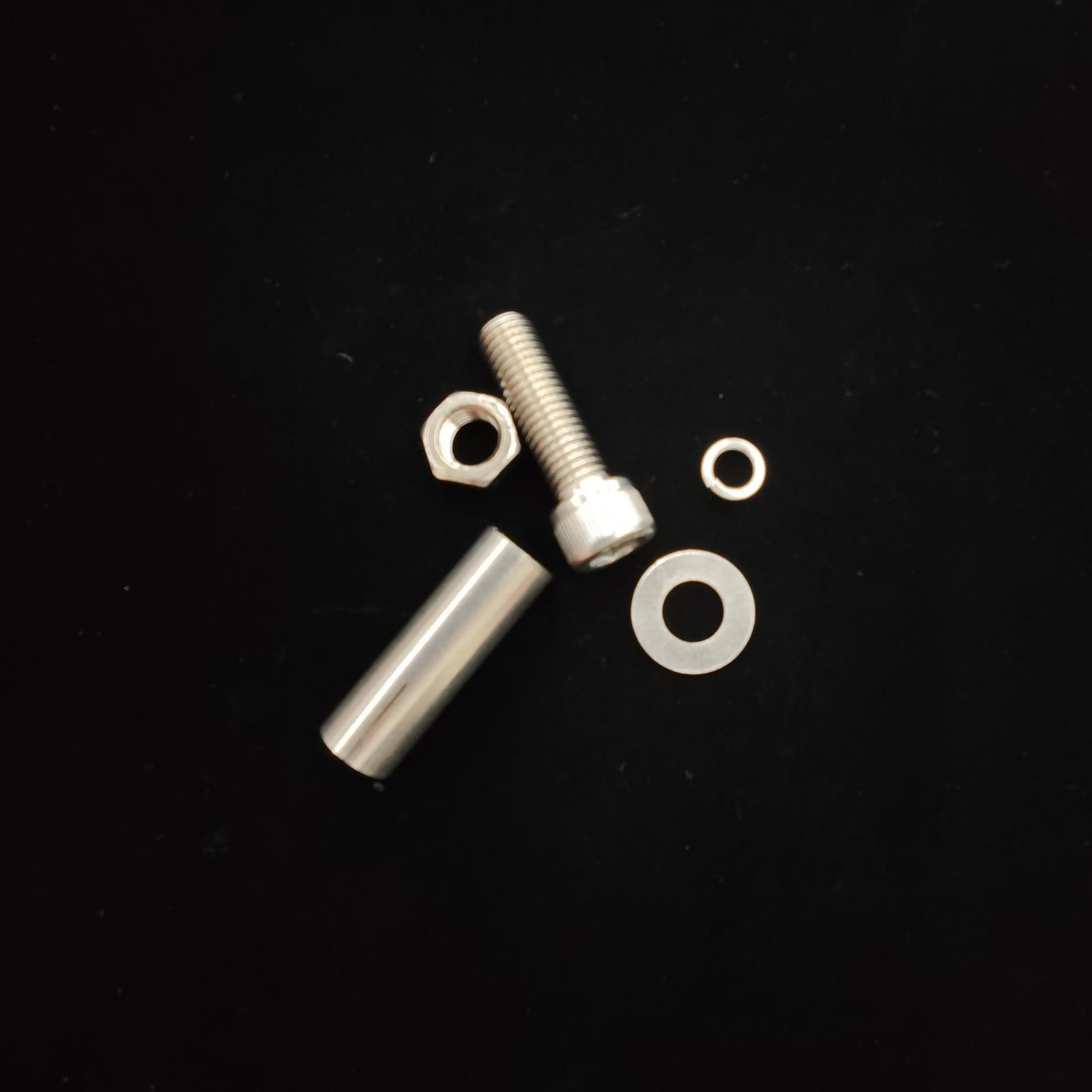}
    }
    \parbox[t][1.5cm][t]{\linewidth}{\small Cup Head Screw M12 40}
\end{minipage}\hfill
\begin{minipage}{0.23\textwidth}
    \centering
    \parbox[c][3.0cm][c]{\linewidth}{
        \centering\includegraphics[width=\linewidth,height=3.0cm,keepaspectratio]{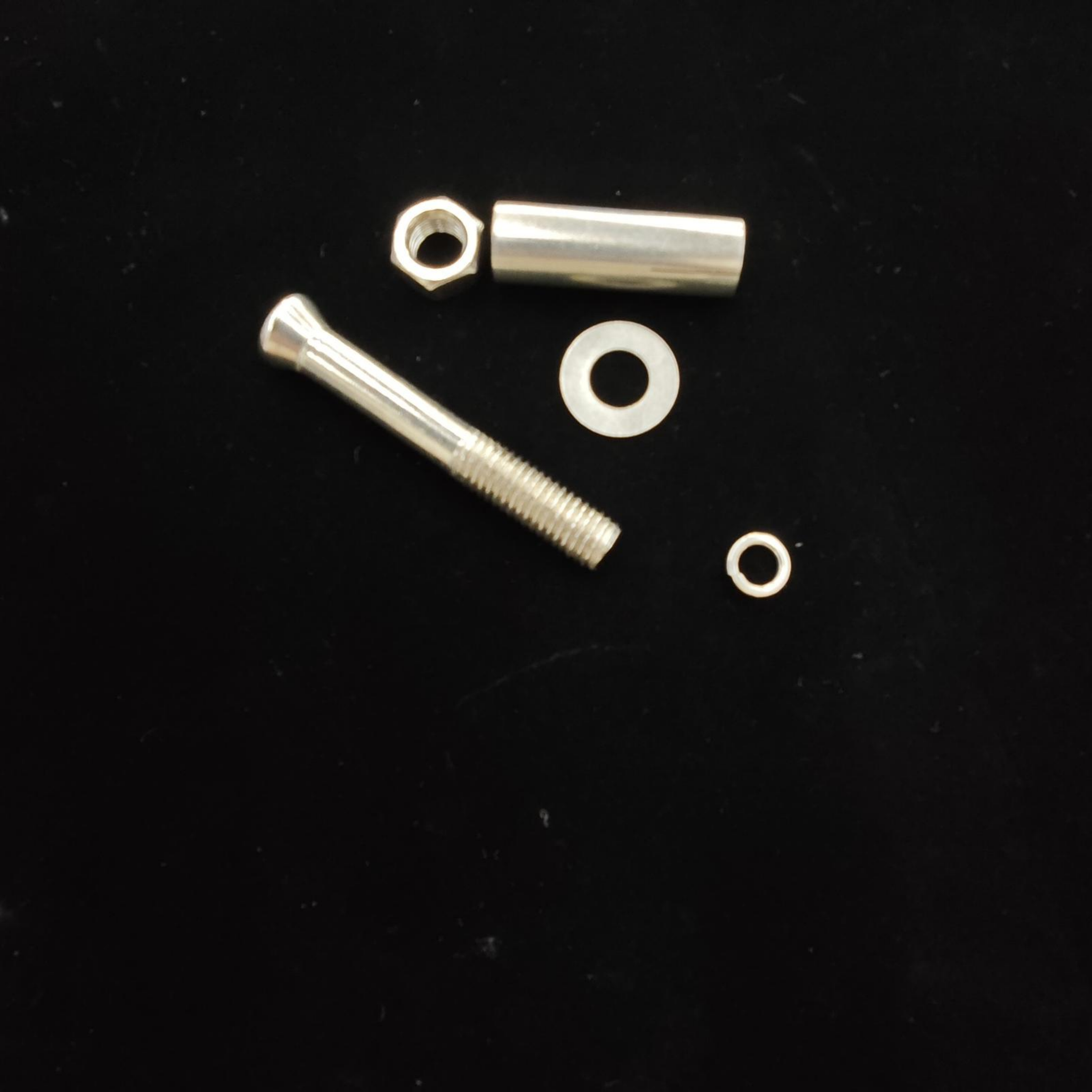}
    }
    \parbox[t][1.5cm][t]{\linewidth}{\small Spring Washer M8}
\end{minipage}\hfill
\begin{minipage}{0.23\textwidth}
    \centering
    \parbox[c][3.0cm][c]{\linewidth}{
        \centering\includegraphics[width=\linewidth,height=3.0cm,keepaspectratio]{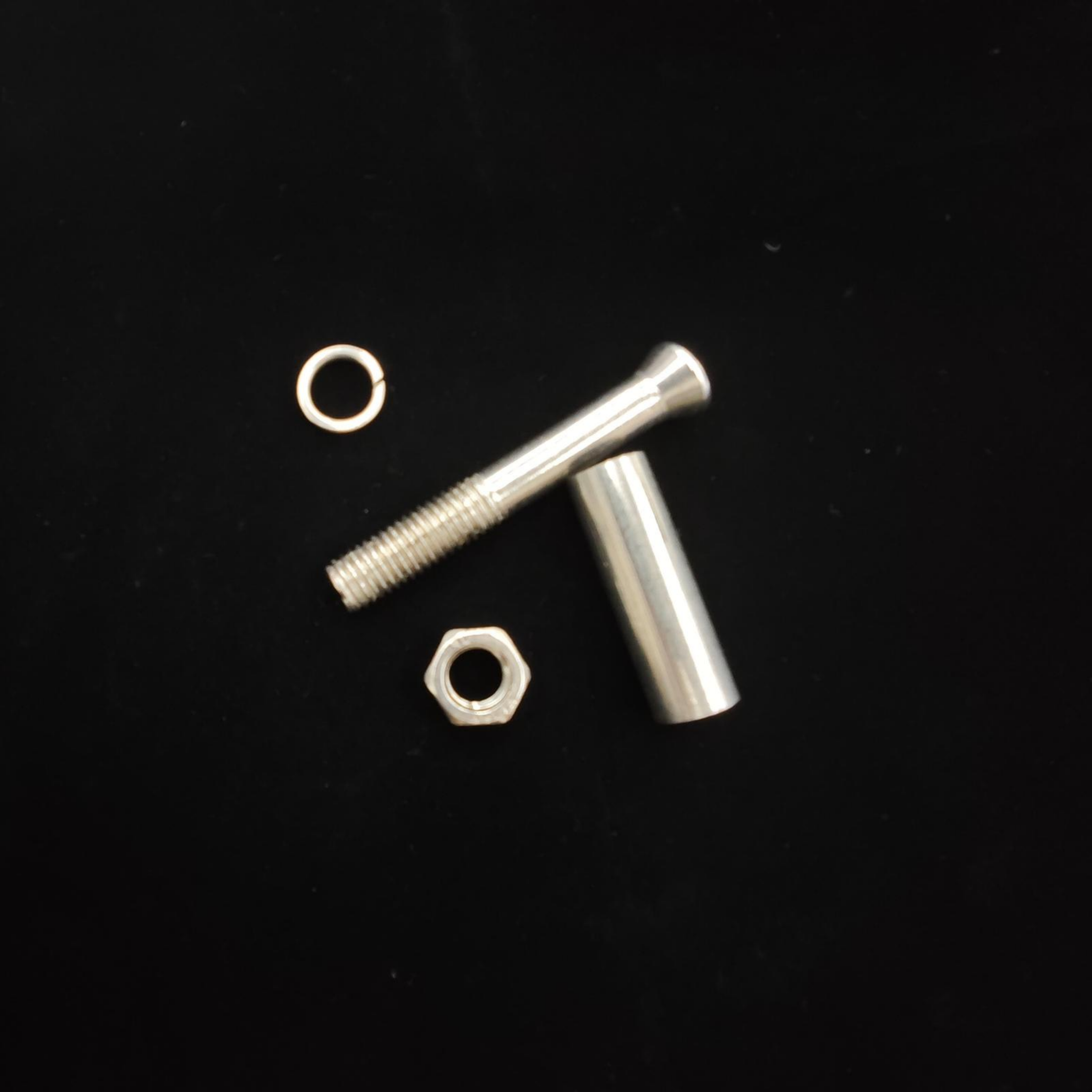}
    }
    \parbox[t][1.5cm][t]{\linewidth}{\small No Flat Washers}
\end{minipage}\hfill
\begin{minipage}{0.23\textwidth}
    \centering
    \parbox[c][3.0cm][c]{\linewidth}{
        \centering\includegraphics[width=\linewidth,height=3.0cm,keepaspectratio]{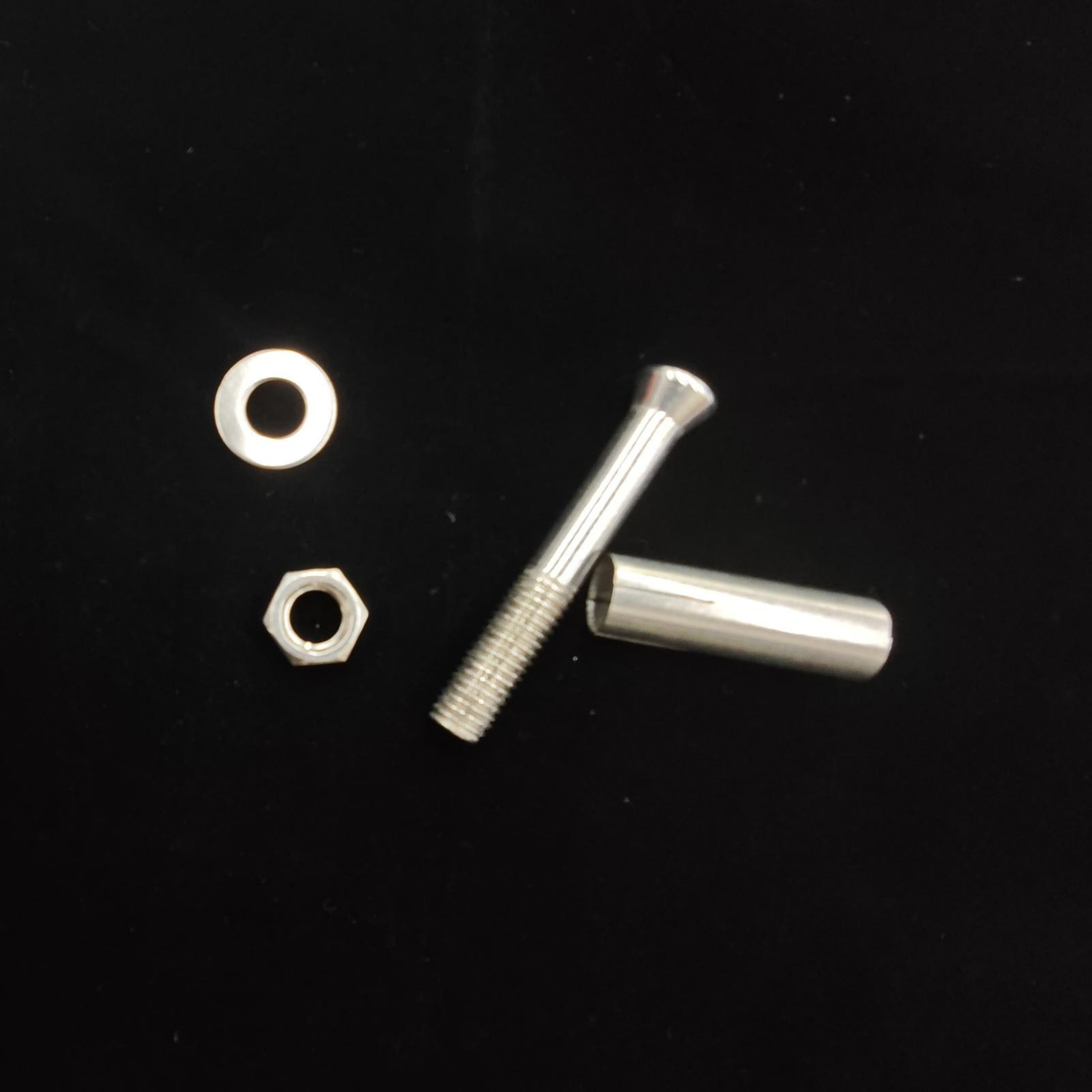}
    }
    \parbox[t][1.5cm][t]{\linewidth}{\small No Spring Washers}
\end{minipage}

\caption{Random examples from \messcen}
\label{fig: Manufacturing_scenario_1_saved}
\end{figure}

\begin{figure}[htbp]
\centering

\begin{minipage}{0.23\textwidth}
    \centering
    \parbox[c][3.0cm][c]{\linewidth}{
        \centering\includegraphics[width=\linewidth,height=3.0cm,keepaspectratio]{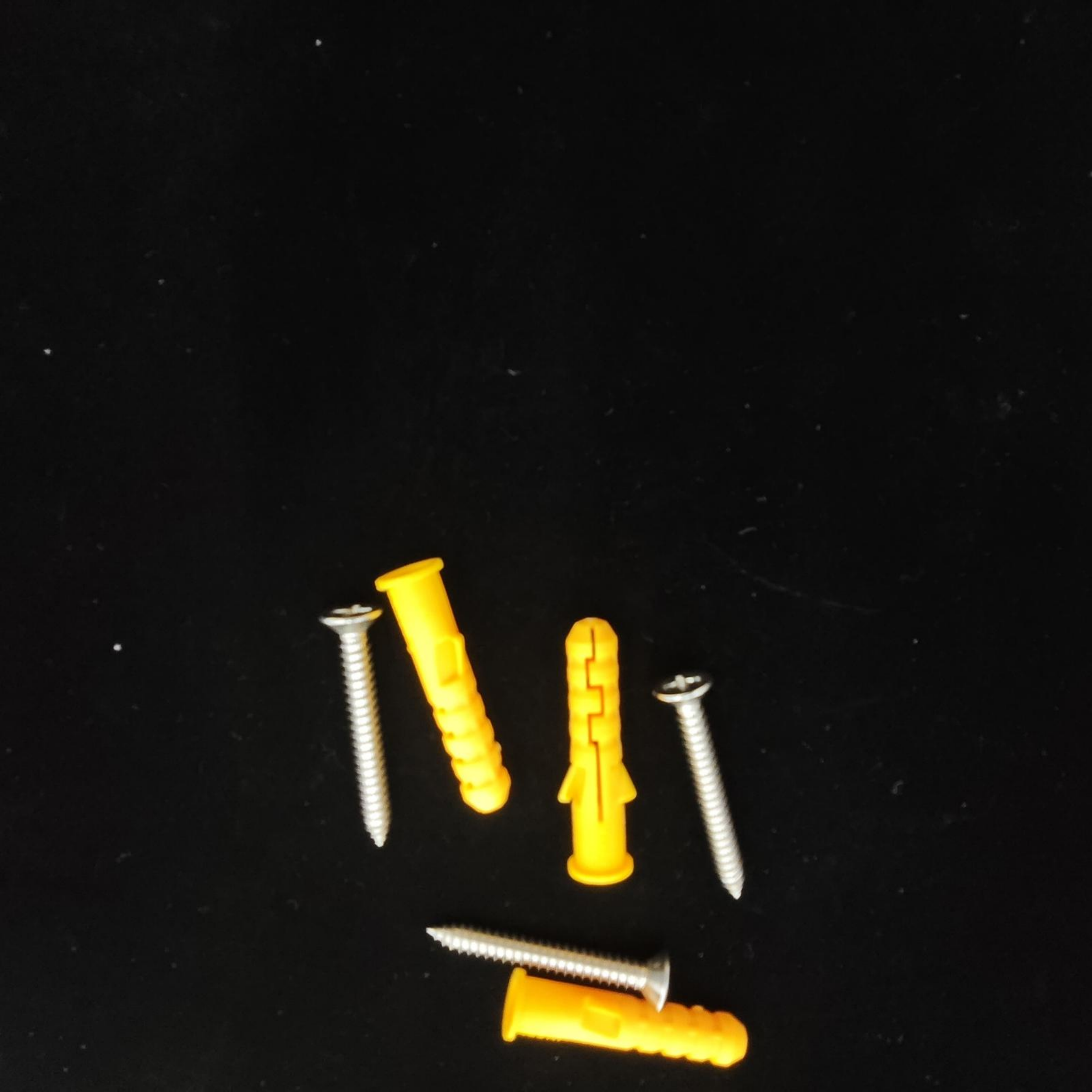}
    }
    \parbox[t][1.5cm][t]{\linewidth}{\small Normal Case}
\end{minipage}\hfill
\begin{minipage}{0.23\textwidth}
    \centering
    \parbox[c][3.0cm][c]{\linewidth}{
        \centering\includegraphics[width=\linewidth,height=3.0cm,keepaspectratio]{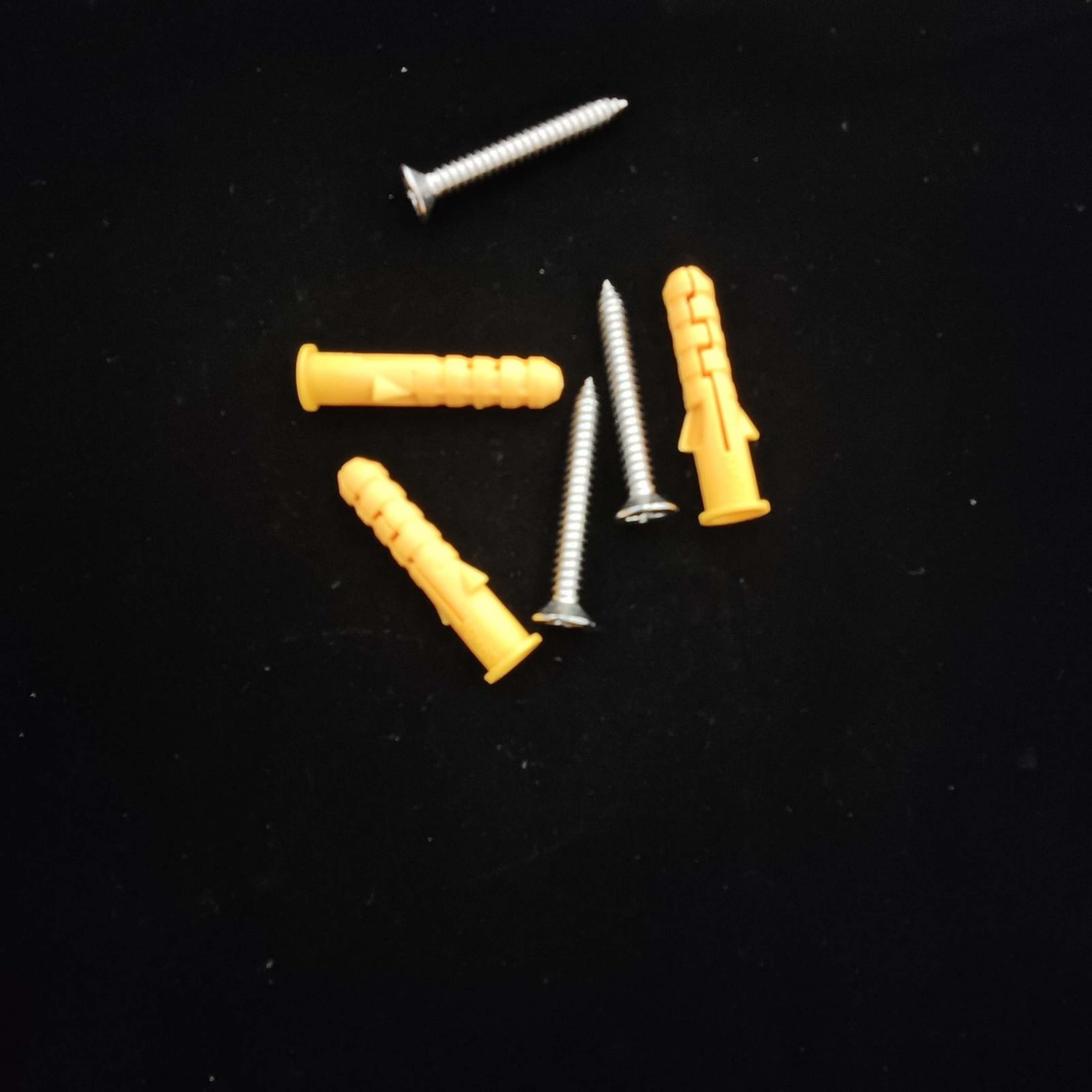}
    }
    \parbox[t][1.5cm][t]{\linewidth}{\small Normal Case}
\end{minipage}\hfill
\begin{minipage}{0.23\textwidth}
    \centering
    \parbox[c][3.0cm][c]{\linewidth}{
        \centering\includegraphics[width=\linewidth,height=3.0cm,keepaspectratio]{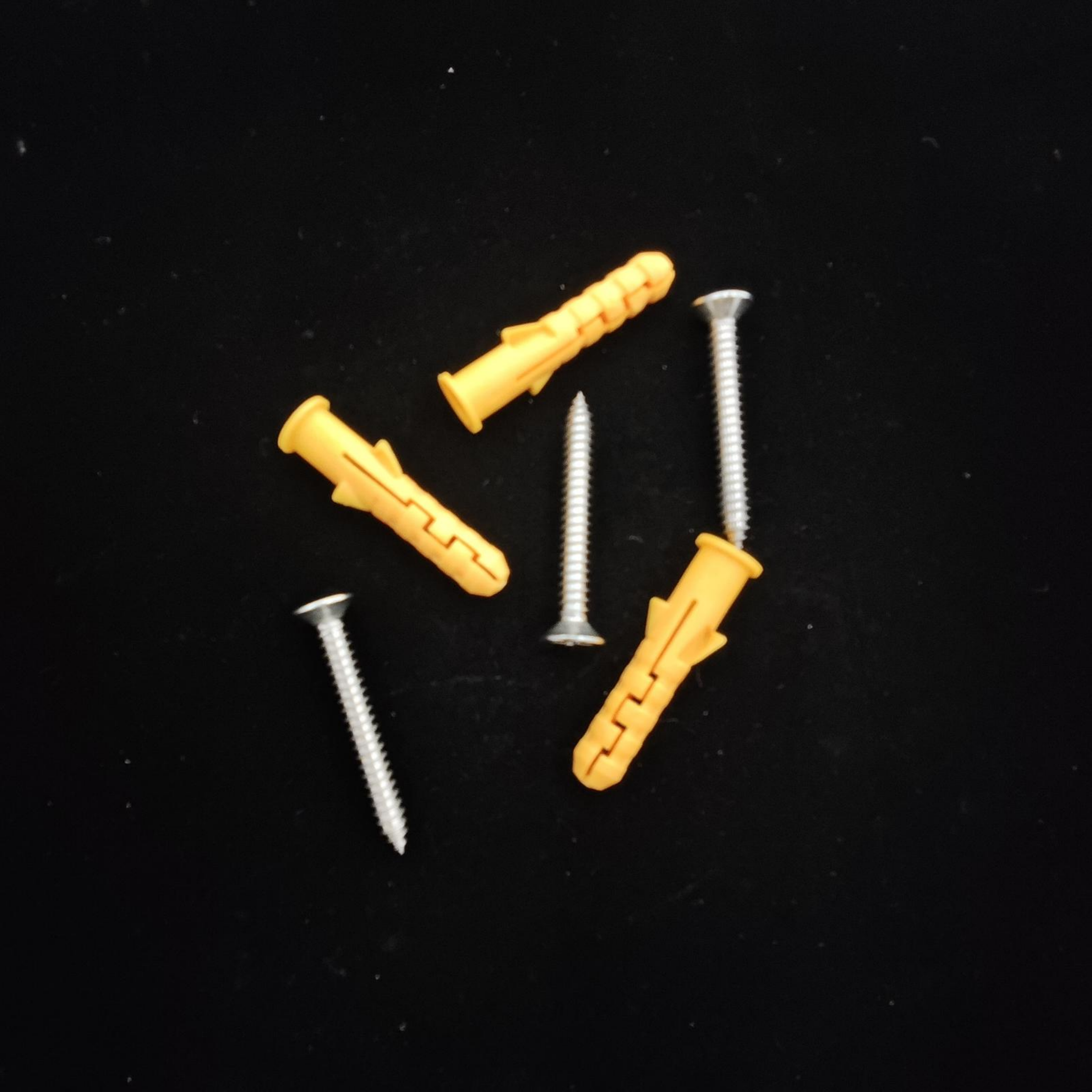}
    }
    \parbox[t][1.5cm][t]{\linewidth}{\small Normal Case}
\end{minipage}\hfill
\begin{minipage}{0.23\textwidth}
    \centering
    \parbox[c][3.0cm][c]{\linewidth}{
        \centering\includegraphics[width=\linewidth,height=3.0cm,keepaspectratio]{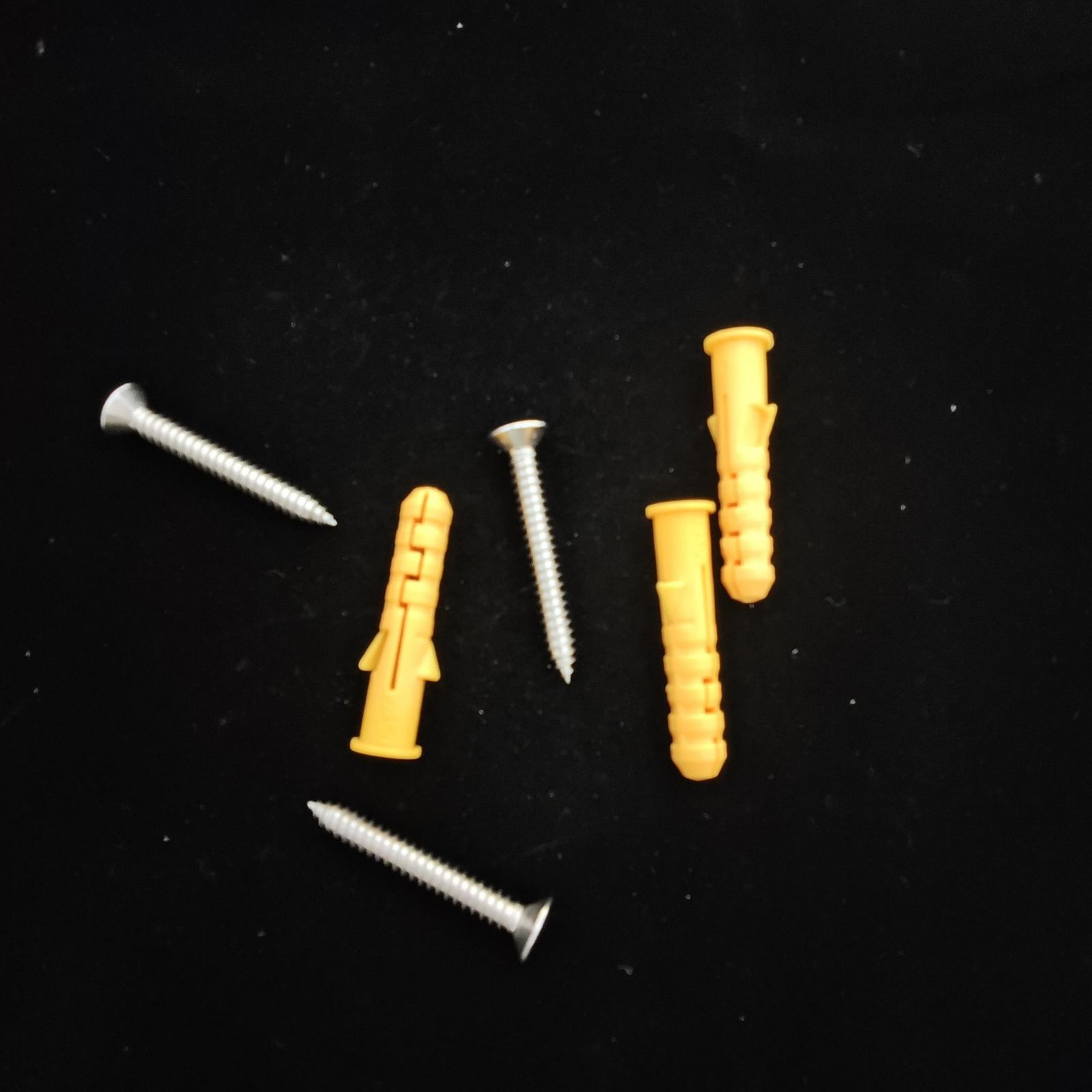}
    }
    \parbox[t][1.5cm][t]{\linewidth}{\small Normal Case}
\end{minipage}

\begin{minipage}{0.23\textwidth}
    \centering
    \parbox[c][3.0cm][c]{\linewidth}{
        \centering\includegraphics[width=\linewidth,height=3.0cm,keepaspectratio]{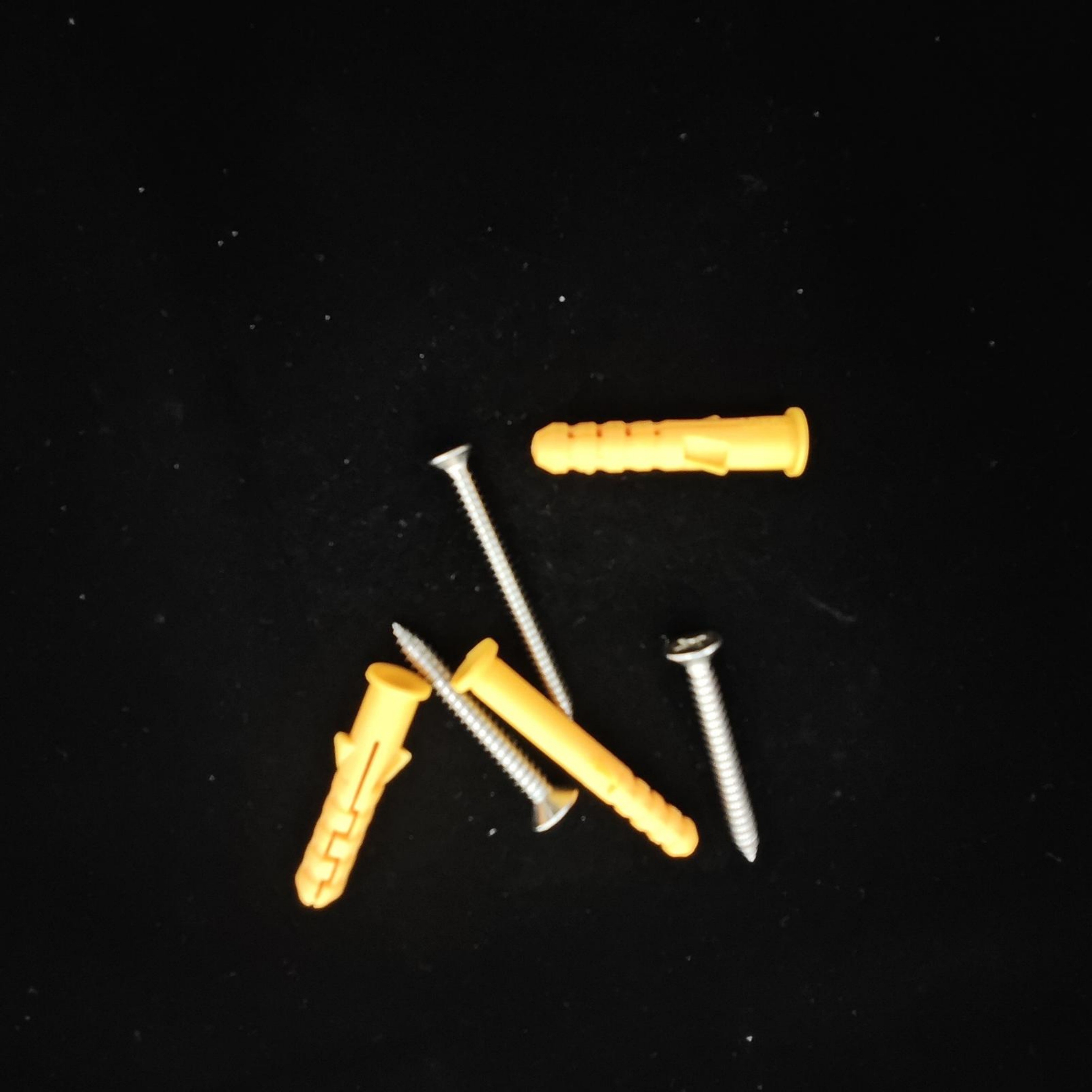}
    }
    \parbox[t][1.5cm][t]{\linewidth}{\small Model 860 expansion screws were mistakenly included}
\end{minipage}\hfill
\begin{minipage}{0.23\textwidth}
    \centering
    \parbox[c][3.0cm][c]{\linewidth}{
        \centering\includegraphics[width=\linewidth,height=3.0cm,keepaspectratio]{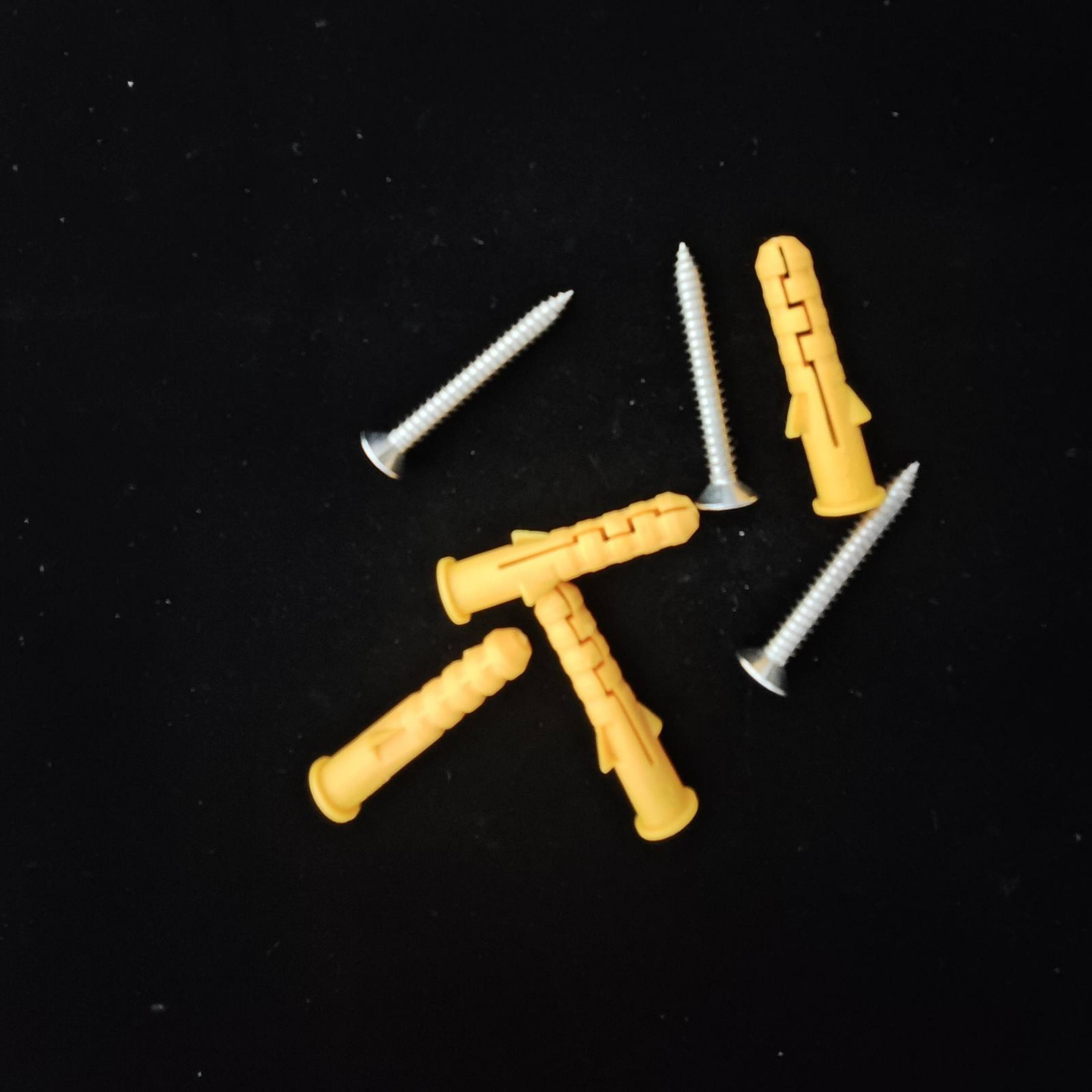}
    }
    \parbox[t][1.5cm][t]{\linewidth}{\small One extra plastic expansion anchor}
\end{minipage}\hfill
\begin{minipage}{0.23\textwidth}
    \centering
    \parbox[c][3.0cm][c]{\linewidth}{
        \centering\includegraphics[width=\linewidth,height=3.0cm,keepaspectratio]{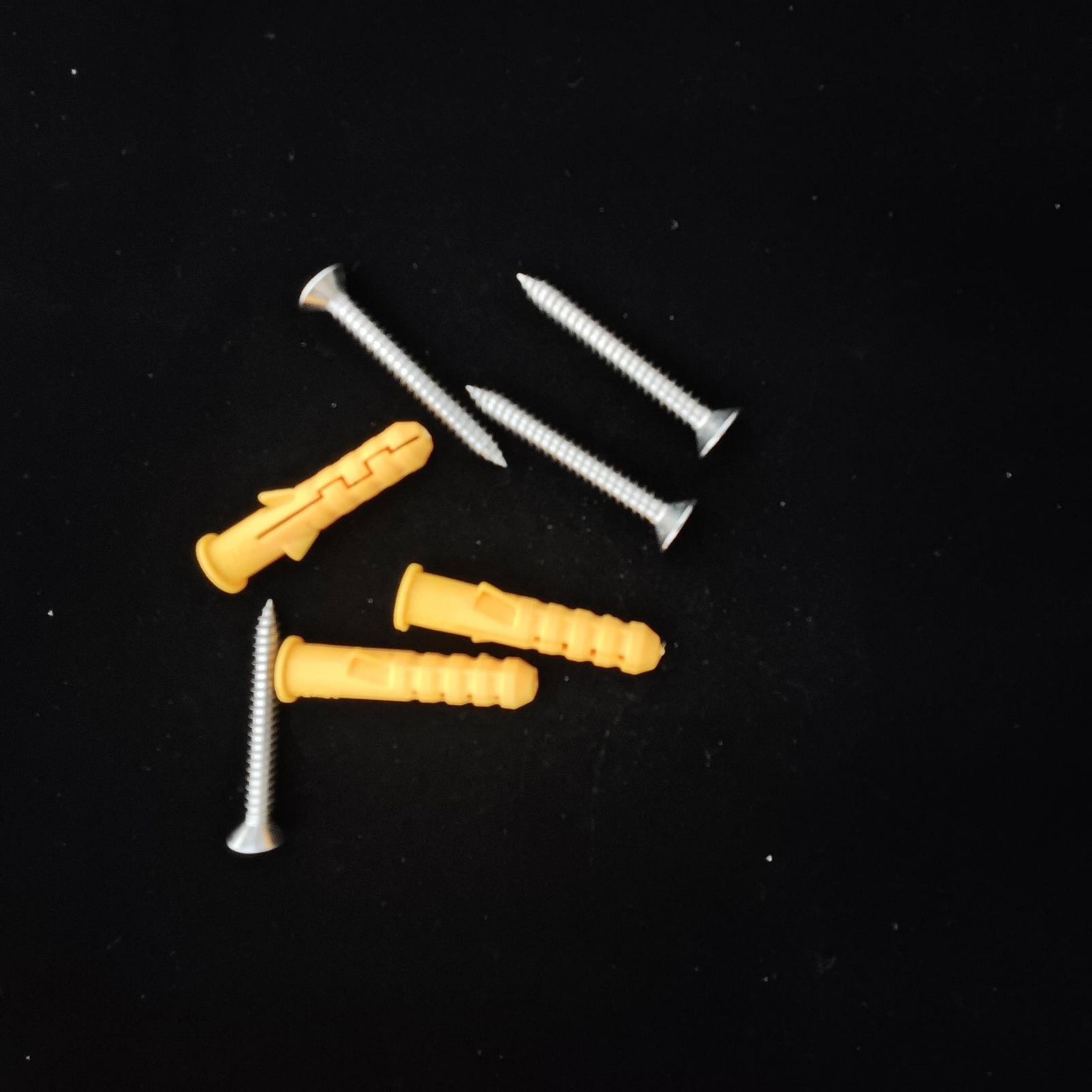}
    }
    \parbox[t][1.5cm][t]{\linewidth}{\small One extra plastic expansion Self-tapping Screw}
\end{minipage}\hfill
\begin{minipage}{0.23\textwidth}
    \centering
    \parbox[c][3.0cm][c]{\linewidth}{
        \centering\includegraphics[width=\linewidth,height=3.0cm,keepaspectratio]{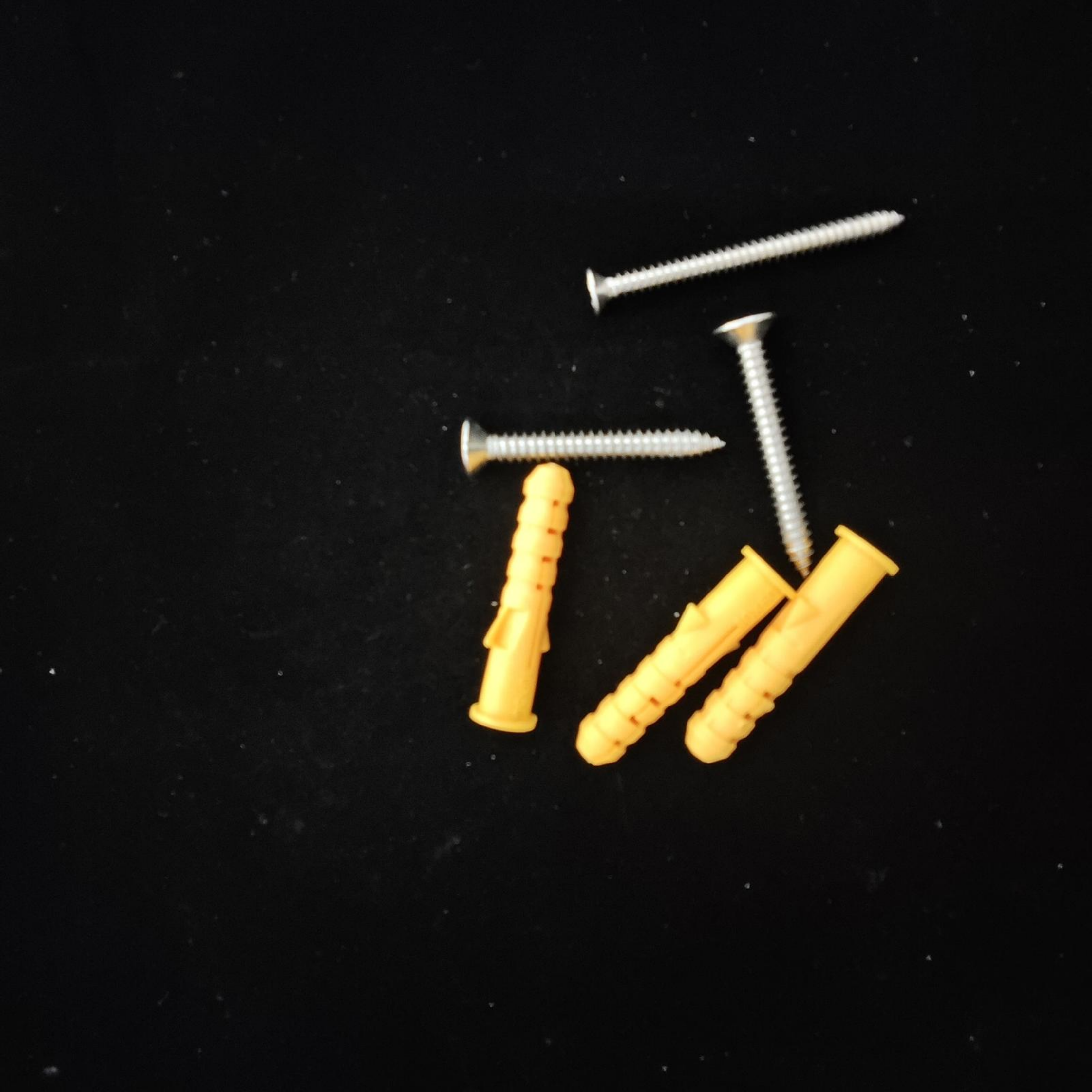}
    }
    \parbox[t][1.5cm][t]{\linewidth}{\small Screws belonging to model 860 expansion anchors were found mixed in}
\end{minipage}

\begin{minipage}{0.23\textwidth}
    \centering
    \parbox[c][3.0cm][c]{\linewidth}{
        \centering\includegraphics[width=\linewidth,height=3.0cm,keepaspectratio]{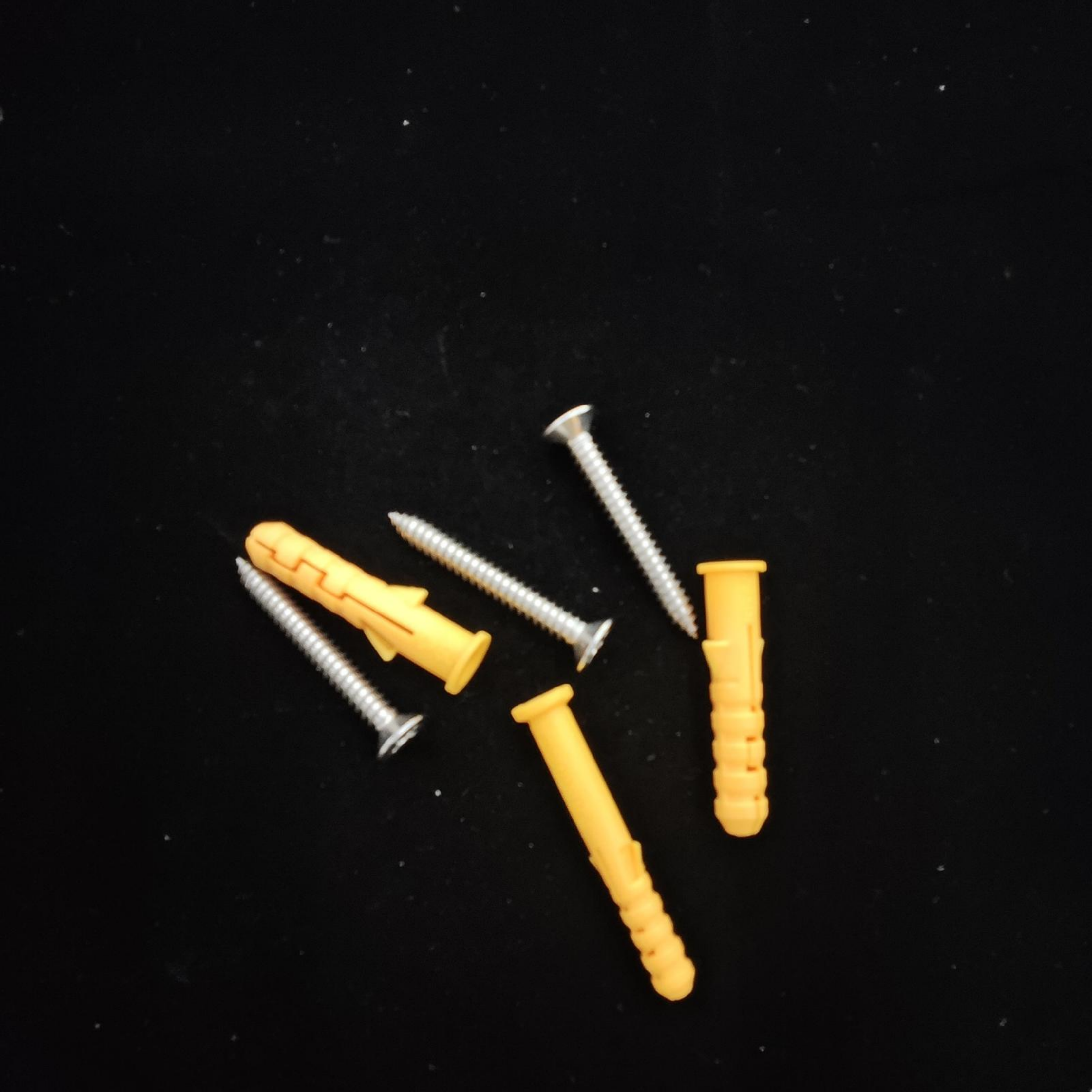}
    }
    \parbox[t][1.5cm][t]{\linewidth}{\small Plastic anchors belonging to expansion screw model 860 were mixed }
\end{minipage}\hfill
\begin{minipage}{0.23\textwidth}
    \centering
    \parbox[c][3.0cm][c]{\linewidth}{
        \centering\includegraphics[width=\linewidth,height=3.0cm,keepaspectratio]{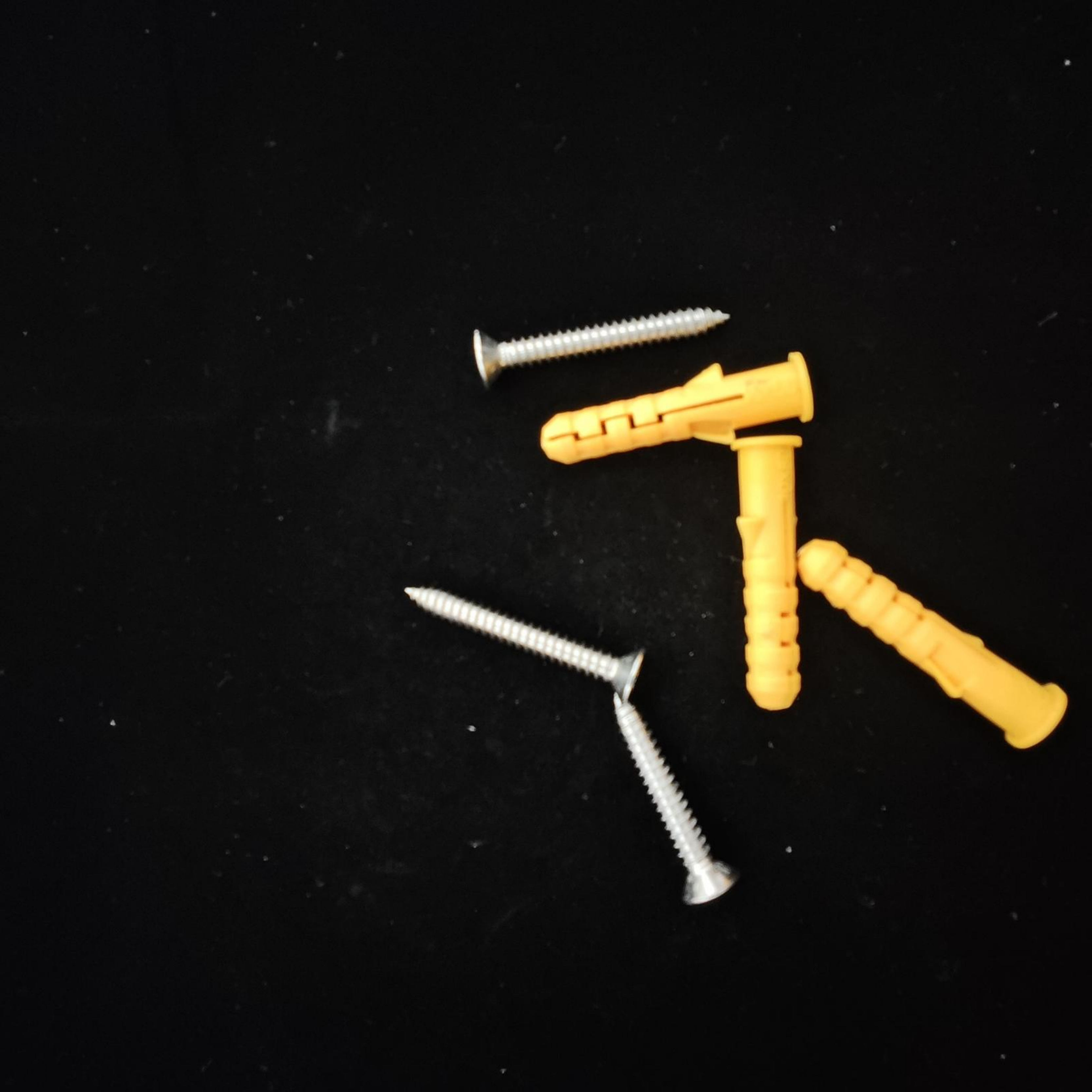}
    }
    \parbox[t][1.5cm][t]{\linewidth}{\small Shorter screws  were found mixed in}
\end{minipage}\hfill
\begin{minipage}{0.23\textwidth}
    \centering
    \parbox[c][3.0cm][c]{\linewidth}{
        \centering\includegraphics[width=\linewidth,height=3.0cm,keepaspectratio]{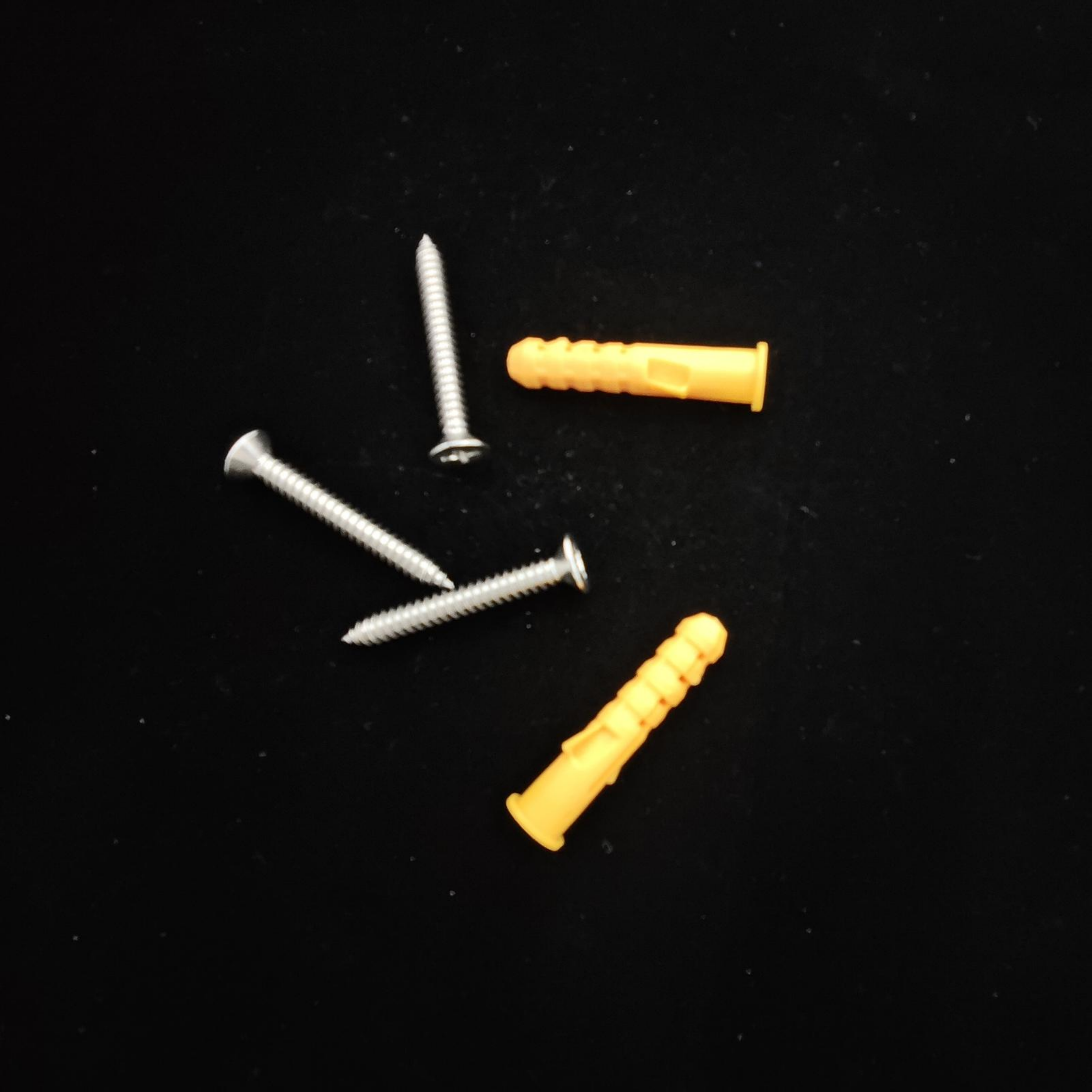}
    }
    \parbox[t][1.5cm][t]{\linewidth}{\small One missing plastic expansion anchor}
\end{minipage}\hfill
\begin{minipage}{0.23\textwidth}
    \centering
    \parbox[c][3.0cm][c]{\linewidth}{
        \centering\includegraphics[width=\linewidth,height=3.0cm,keepaspectratio]{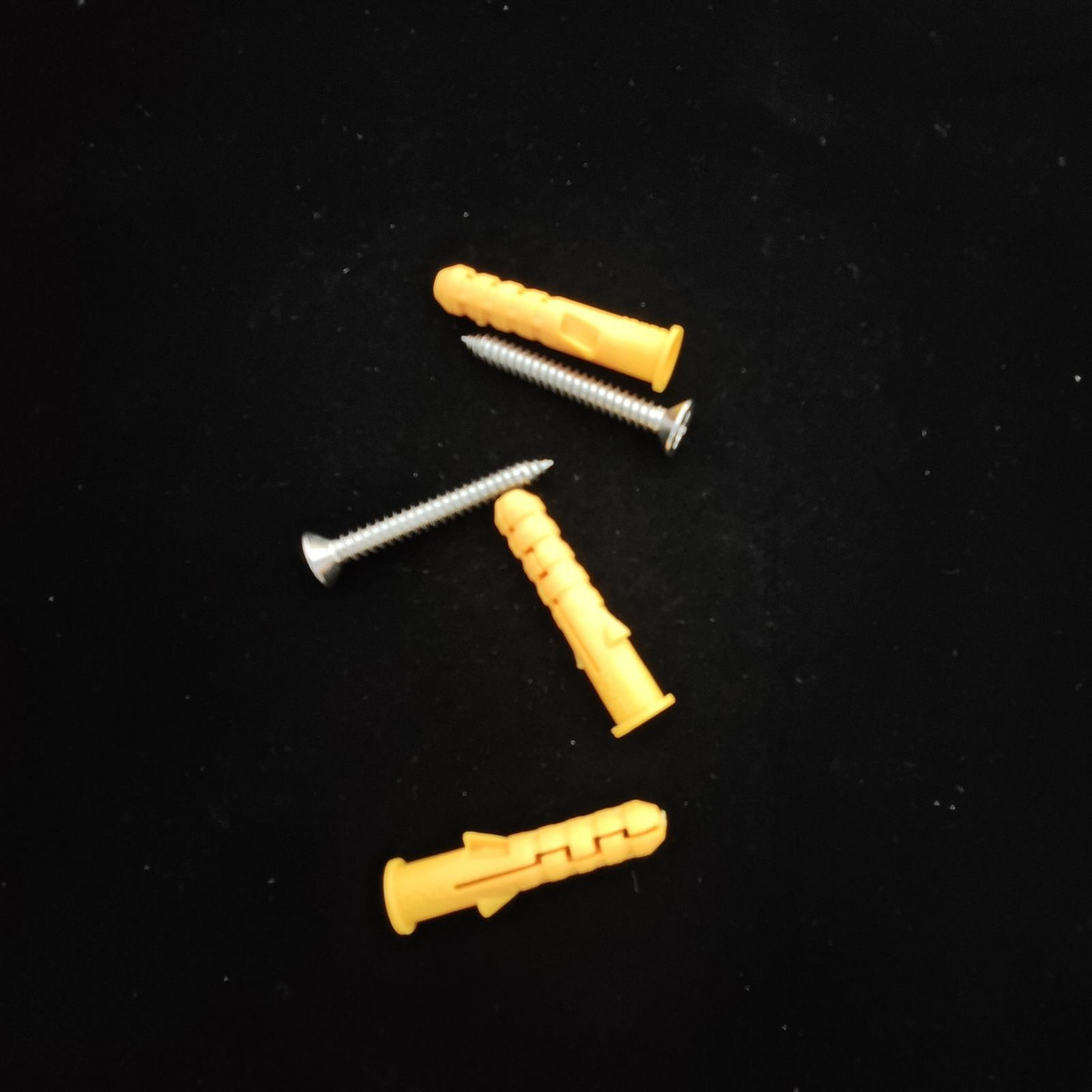}
    }
    \parbox[t][1.5cm][t]{\linewidth}{\small One missing plastic expansion Self-tapping Screw}
\end{minipage}

\caption{Random examples from \pesscen}
\label{fig: Manufacturing_scenario_2_saved}
\end{figure}

\begin{figure}[htbp]
\centering

\begin{minipage}{0.23\textwidth}
    \centering
    \parbox[c][3.0cm][c]{\linewidth}{
        \centering\includegraphics[width=\linewidth,height=3.0cm,keepaspectratio]{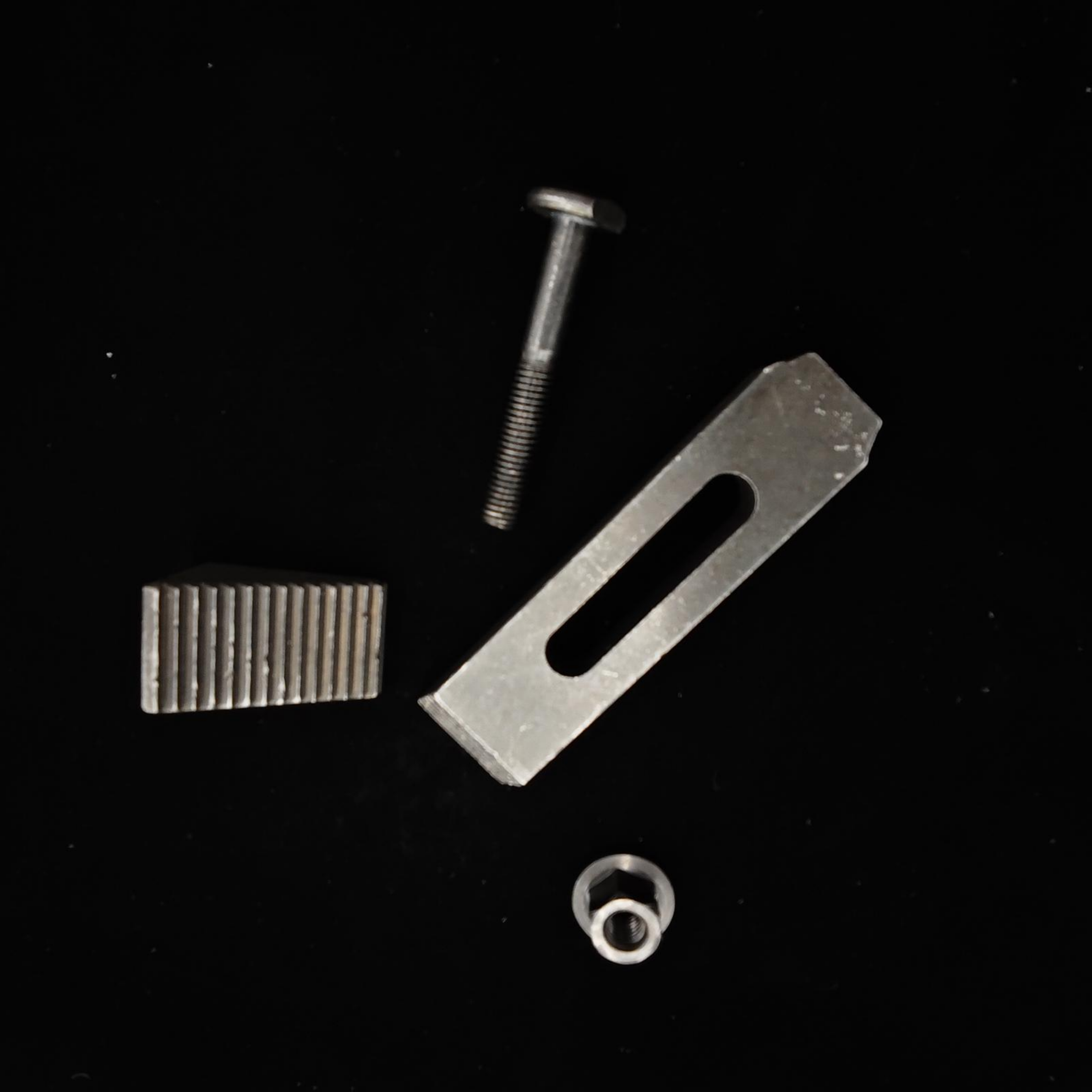}
    }
    \parbox[t][1.5cm][t]{\linewidth}{\small Normal Case }
\end{minipage}\hfill
\begin{minipage}{0.23\textwidth}
    \centering
    \parbox[c][3.0cm][c]{\linewidth}{
        \centering\includegraphics[width=\linewidth,height=3.0cm,keepaspectratio]{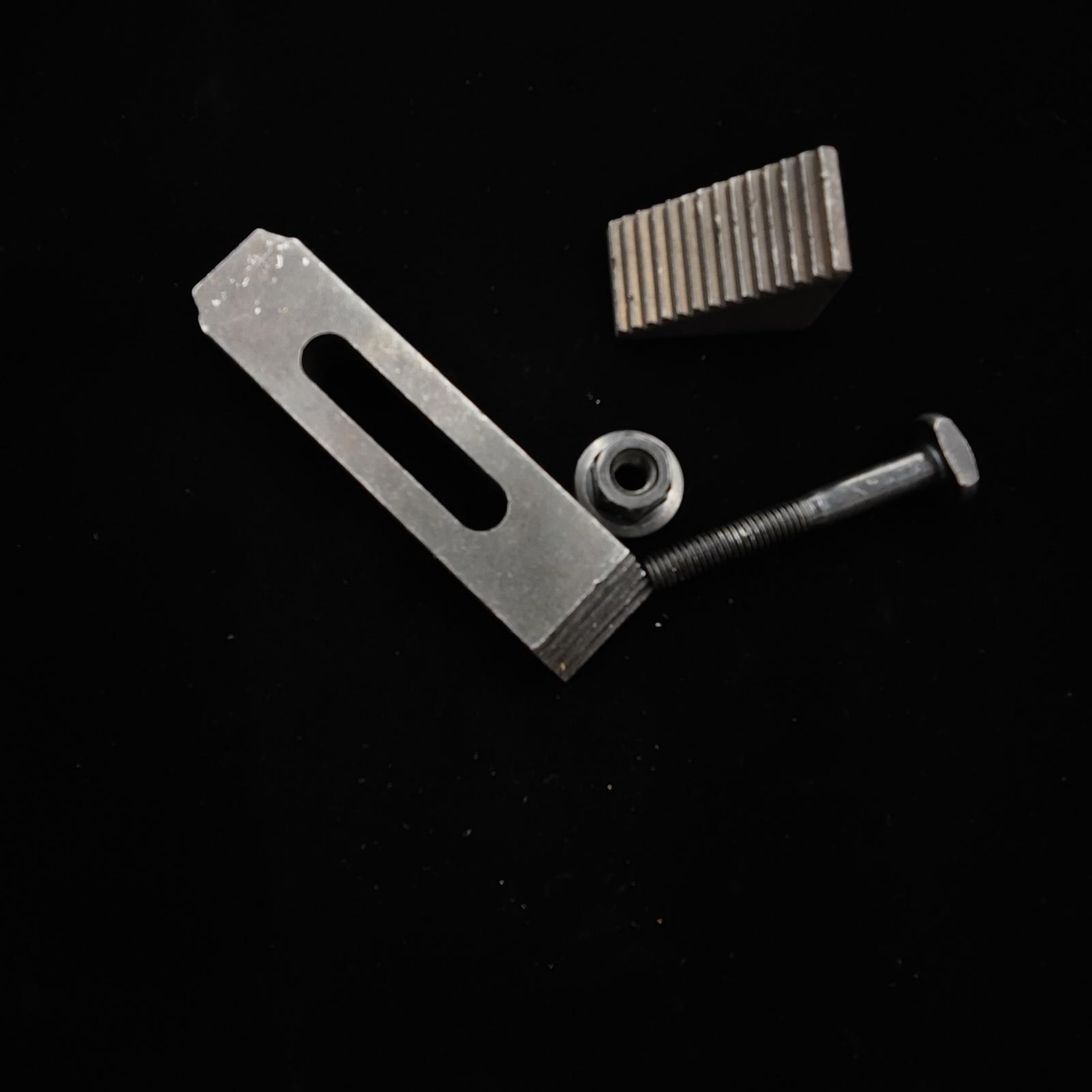}
    }
    \parbox[t][1.5cm][t]{\linewidth}{\small Normal Case }
\end{minipage}\hfill
\begin{minipage}{0.23\textwidth}
    \centering
    \parbox[c][3.0cm][c]{\linewidth}{
        \centering\includegraphics[width=\linewidth,height=3.0cm,keepaspectratio]{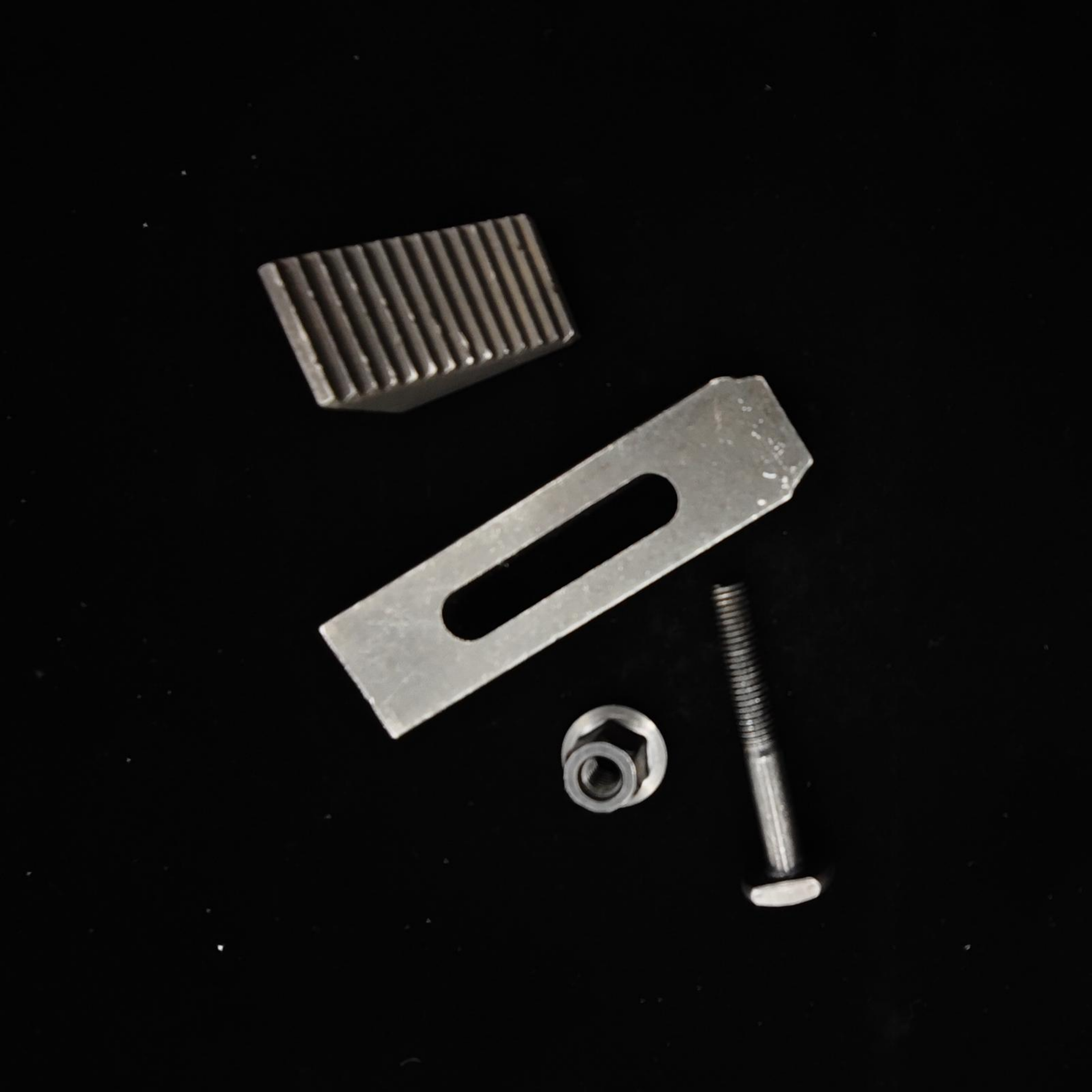}
    }
    \parbox[t][1.5cm][t]{\linewidth}{\small Normal Case }
\end{minipage}\hfill
\begin{minipage}{0.23\textwidth}
    \centering
    \parbox[c][3.0cm][c]{\linewidth}{
        \centering\includegraphics[width=\linewidth,height=3.0cm,keepaspectratio]{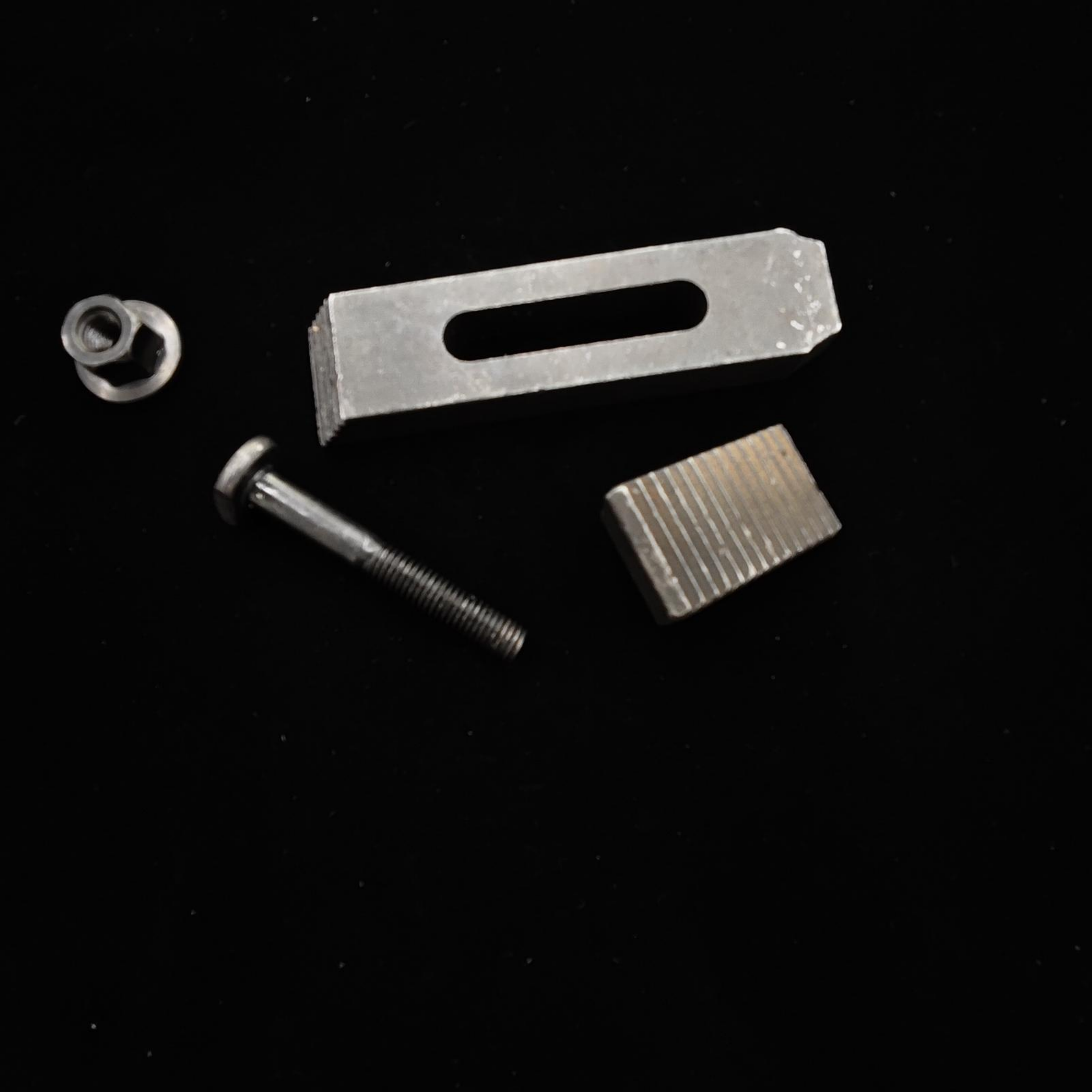}
    }
    \parbox[t][1.5cm][t]{\linewidth}{\small Normal Case }
\end{minipage}

\begin{minipage}{0.23\textwidth}
    \centering
    \parbox[c][3.0cm][c]{\linewidth}{
        \centering\includegraphics[width=\linewidth,height=3.0cm,keepaspectratio]{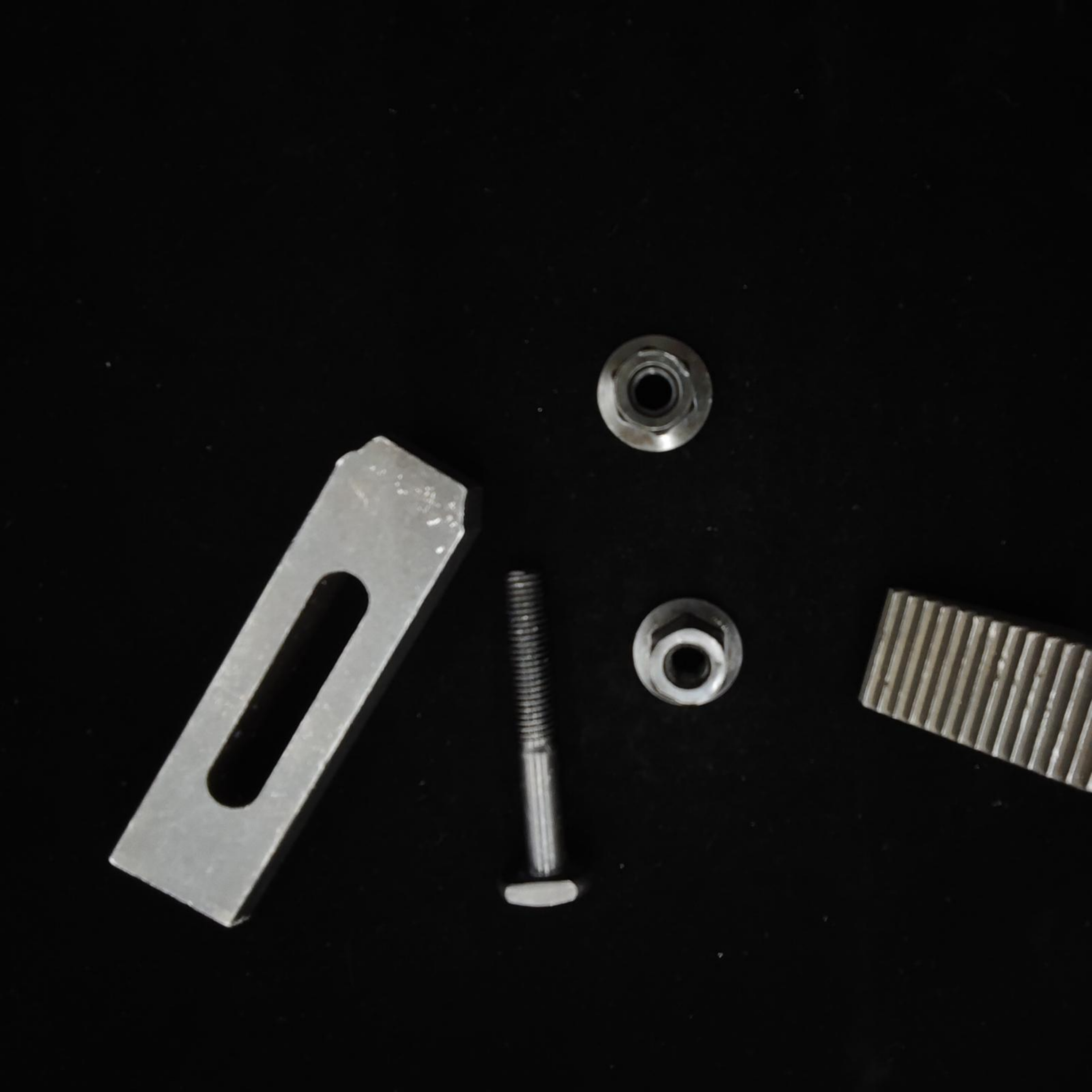}
    }
    \parbox[t][1.5cm][t]{\linewidth}{\small Extra one nut}
\end{minipage}\hfill
\begin{minipage}{0.23\textwidth}
    \centering
    \parbox[c][3.0cm][c]{\linewidth}{
        \centering\includegraphics[width=\linewidth,height=3.0cm,keepaspectratio]{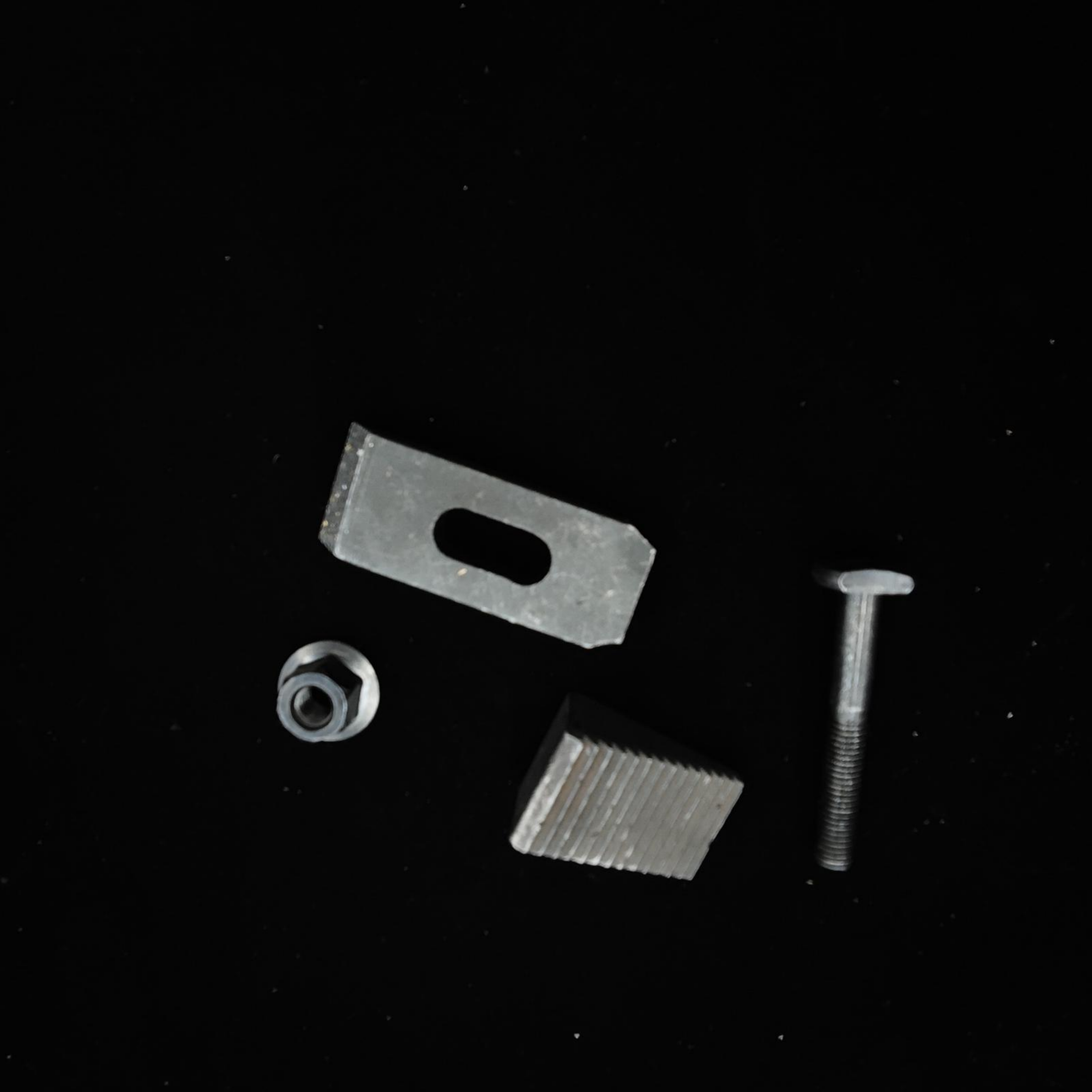}
    }
    \parbox[t][1.5cm][t]{\linewidth}{\small Retainer block}
\end{minipage}\hfill
\begin{minipage}{0.23\textwidth}
    \centering
    \parbox[c][3.0cm][c]{\linewidth}{
        \centering\includegraphics[width=\linewidth,height=3.0cm,keepaspectratio]{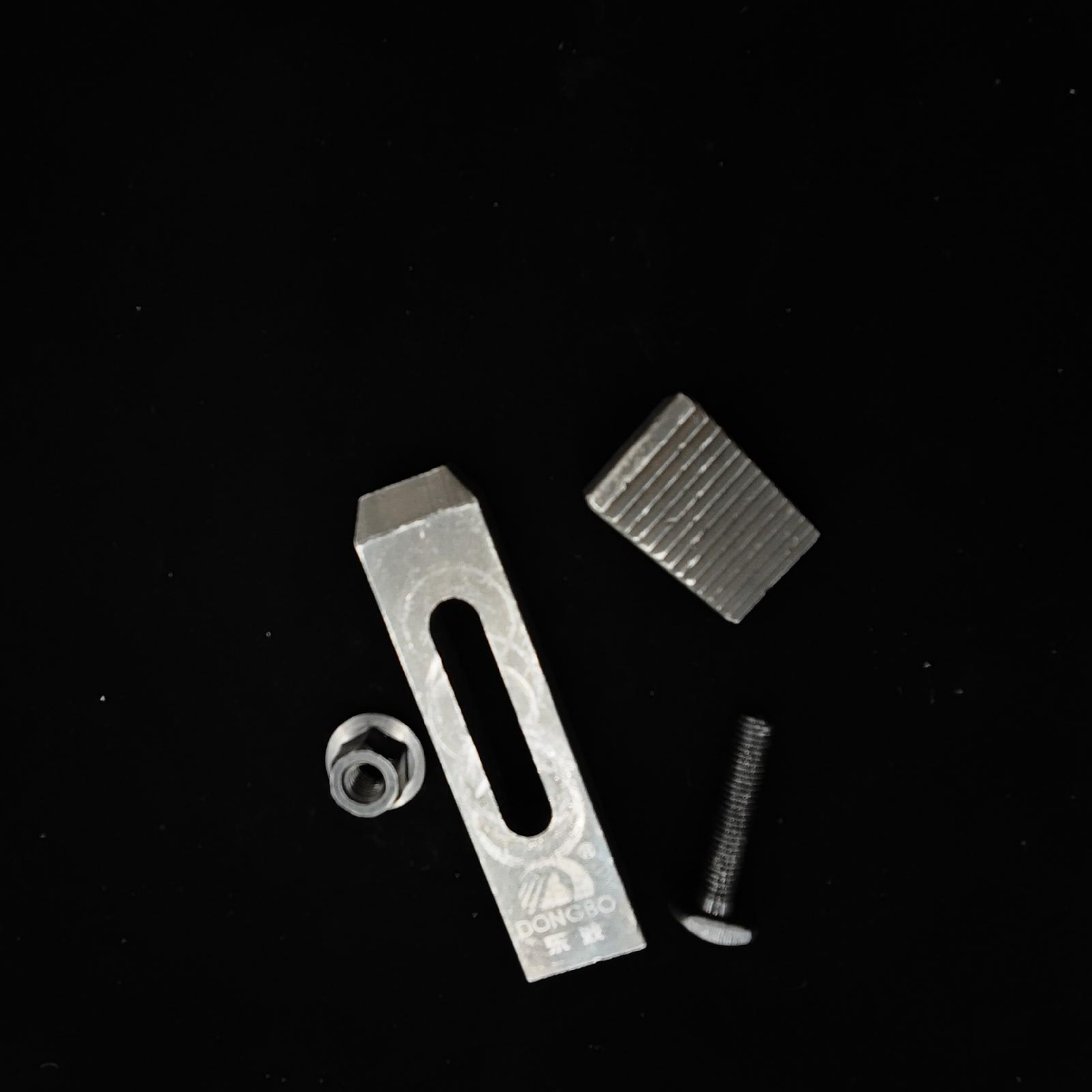}
    }
    \parbox[t][1.5cm][t]{\linewidth}{\small Short screw}
\end{minipage}\hfill
\begin{minipage}{0.23\textwidth}
    \centering
    \parbox[c][3.0cm][c]{\linewidth}{
        \centering\includegraphics[width=\linewidth,height=3.0cm,keepaspectratio]{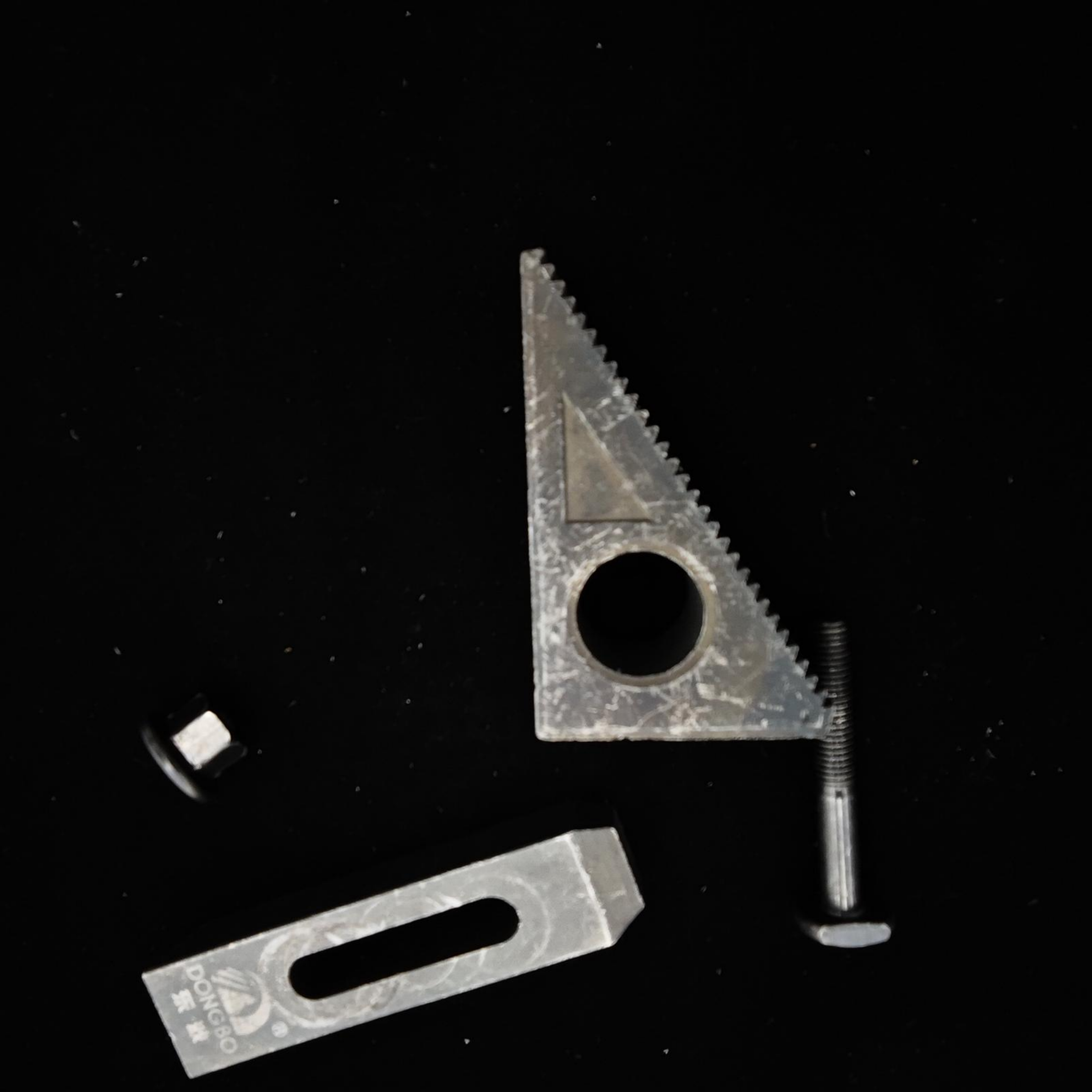}
    }
    \parbox[t][1.5cm][t]{\linewidth}{\small Triangular part}
\end{minipage}

\begin{minipage}{0.23\textwidth}
    \centering
    \parbox[c][3.0cm][c]{\linewidth}{
        \centering\includegraphics[width=\linewidth,height=3.0cm,keepaspectratio]{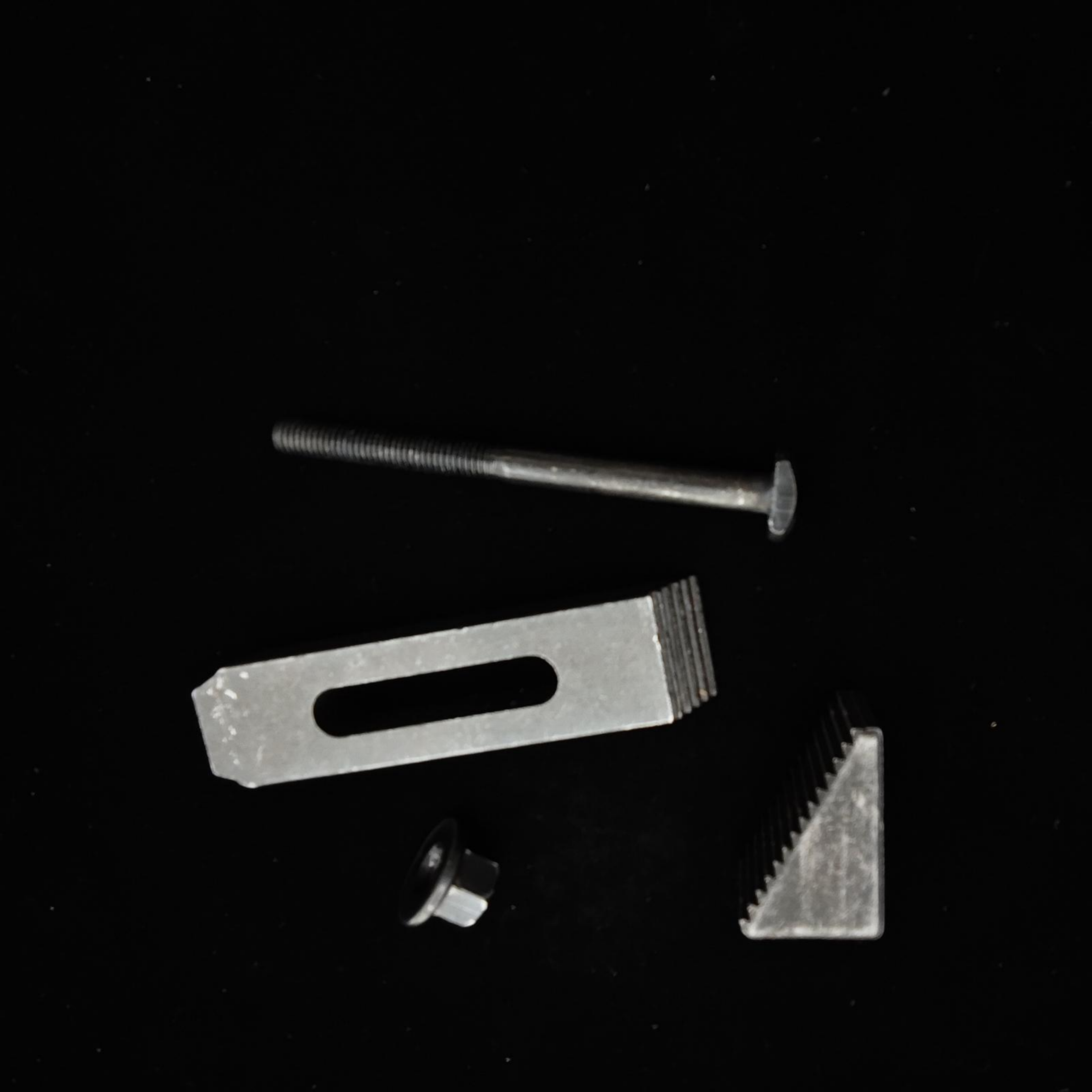}
    }
    \parbox[t][1.5cm][t]{\linewidth}{\small Long screw}
\end{minipage}\hfill
\begin{minipage}{0.23\textwidth}
    \centering
    \parbox[c][3.0cm][c]{\linewidth}{
        \centering\includegraphics[width=\linewidth,height=3.0cm,keepaspectratio]{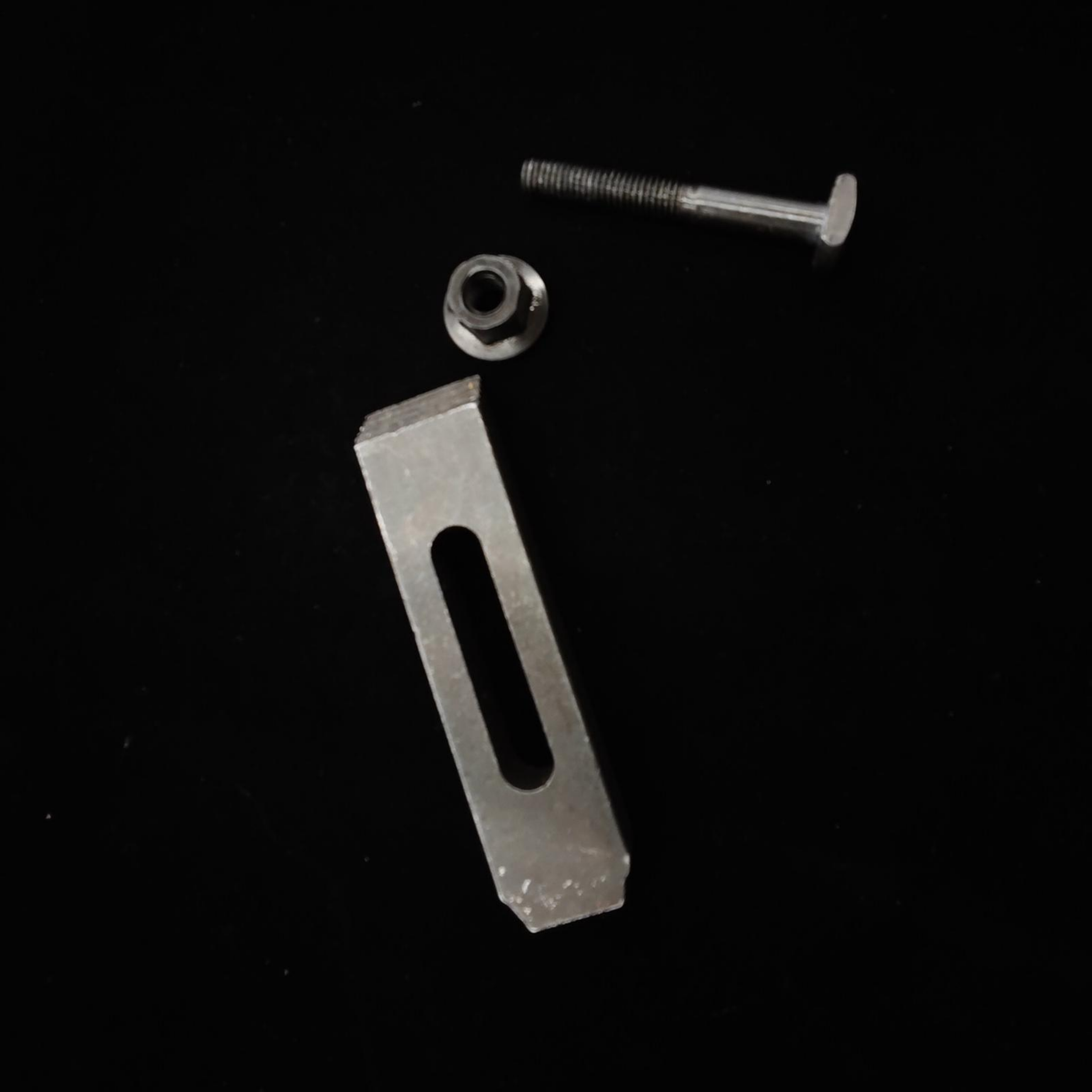}
    }
    \parbox[t][1.5cm][t]{\linewidth}{\small Missing triangular part}
\end{minipage}\hfill
\begin{minipage}{0.23\textwidth}
    \centering
    \parbox[c][3.0cm][c]{\linewidth}{
        \centering\includegraphics[width=\linewidth,height=3.0cm,keepaspectratio]{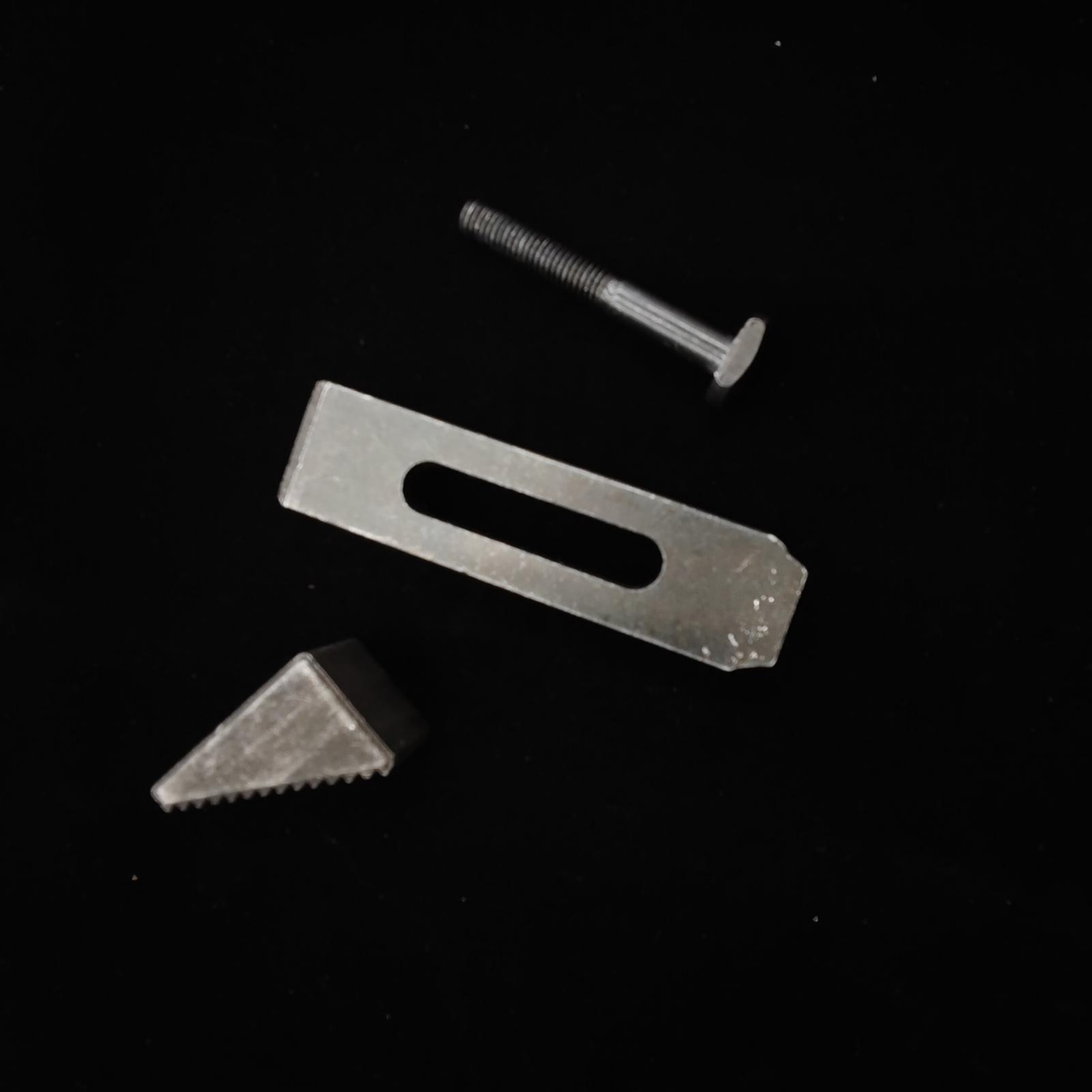}
    }
    \parbox[t][1.5cm][t]{\linewidth}{\small Missing one nut}
\end{minipage}\hfill
\begin{minipage}{0.23\textwidth}
    \centering
    \parbox[c][3.0cm][c]{\linewidth}{
        \centering\includegraphics[width=\linewidth,height=3.0cm,keepaspectratio]{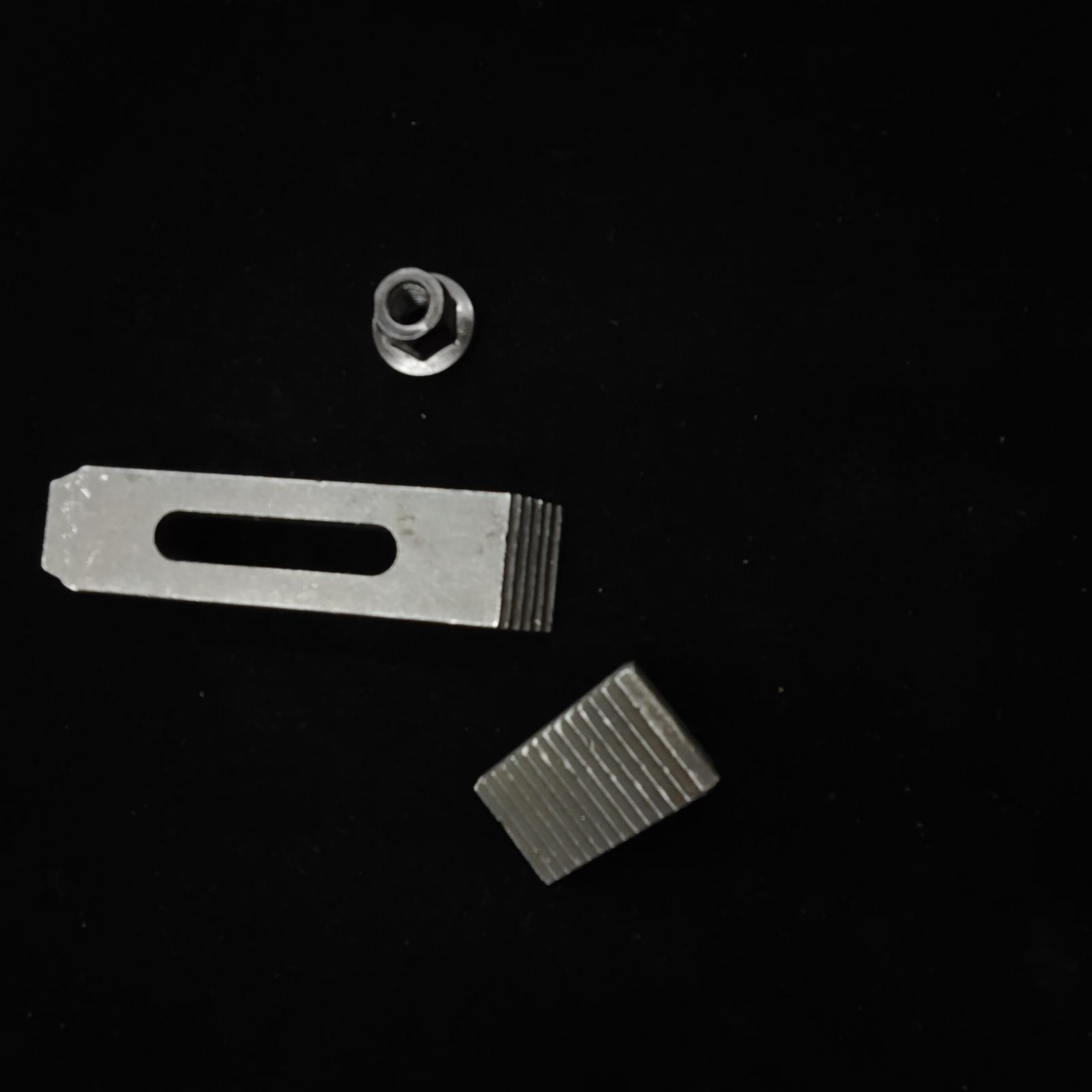}
    }
    \parbox[t][1.5cm][t]{\linewidth}{\small Missing screw}
\end{minipage}

\caption{Random examples from \cncscen}
\label{fig: Manufacturing_scenario_4_saved}
\end{figure}

\begin{figure}[htbp]
\centering

\begin{minipage}{0.48\textwidth}
    \centering
    \includegraphics[width=\linewidth]{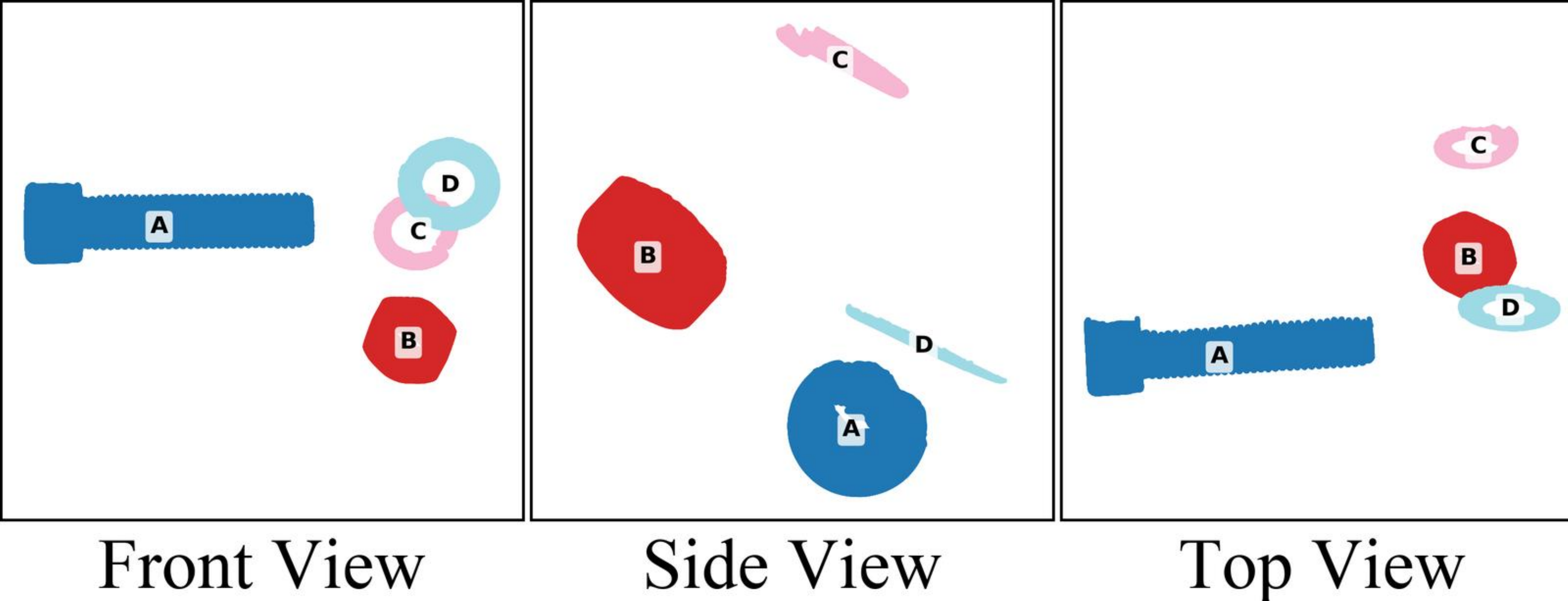}\\
    \small Normal Case
\end{minipage}\hfill
\begin{minipage}{0.48\textwidth}
    \centering
    \includegraphics[width=\linewidth]{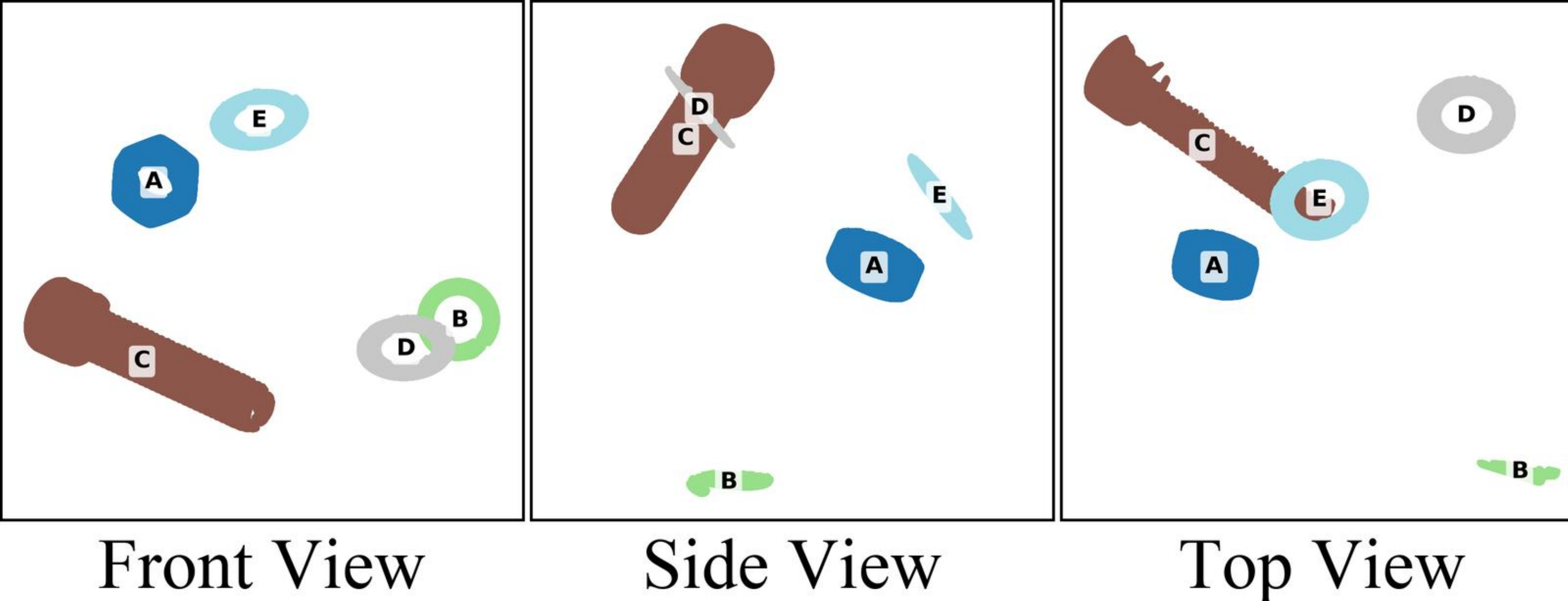}\\
    \small Extra Part: Two Flat Washers
\end{minipage}

\begin{minipage}{0.48\textwidth}
    \centering
    \includegraphics[width=\linewidth]{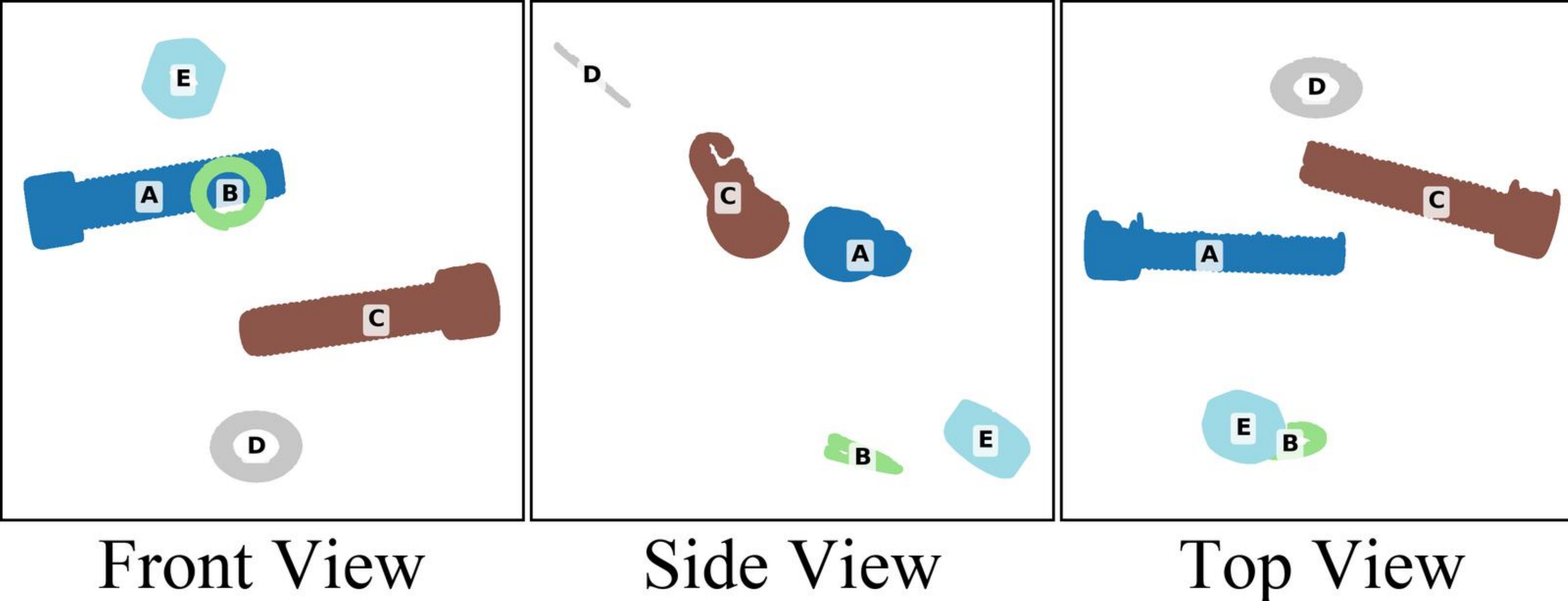}\\
    \small Extra Part: Two Cup Head Screw
\end{minipage}\hfill
\begin{minipage}{0.48\textwidth}
    \centering
    \includegraphics[width=\linewidth]{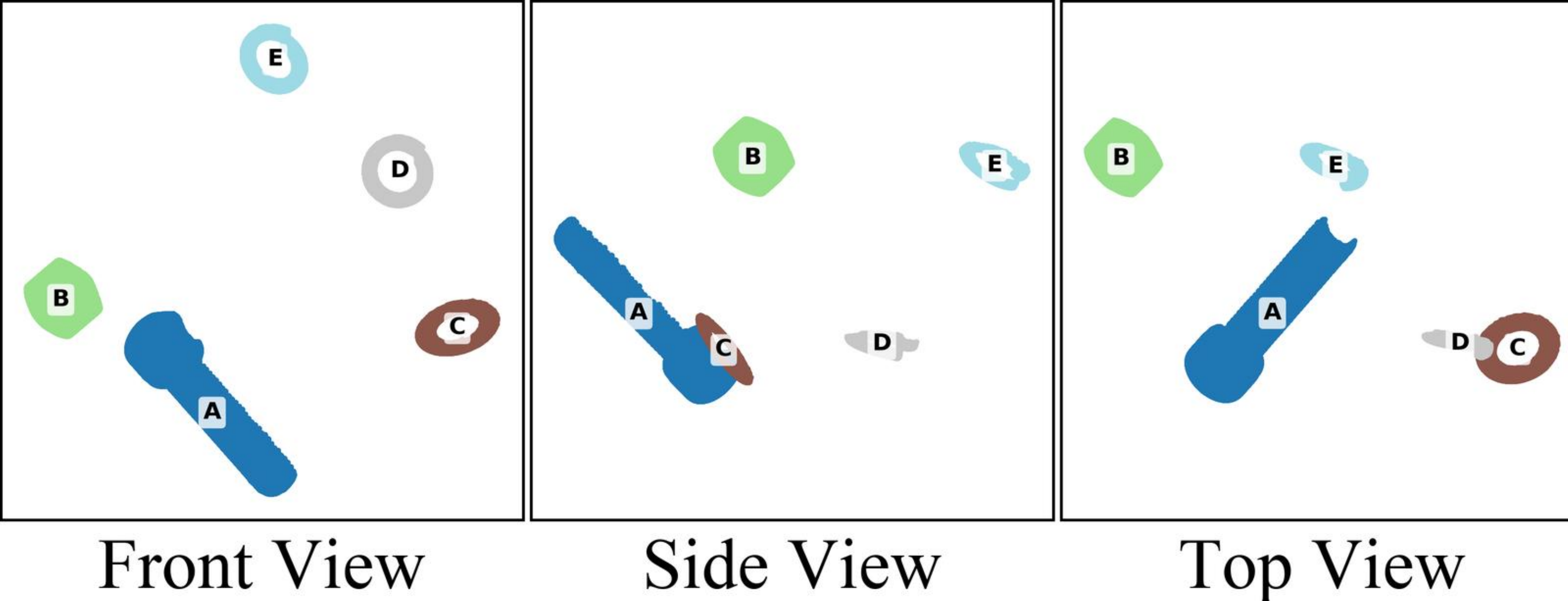}\\
    \small Extra Part: Two Spring Washers
\end{minipage}

\begin{minipage}{0.48\textwidth}
    \centering
    \includegraphics[width=\linewidth]{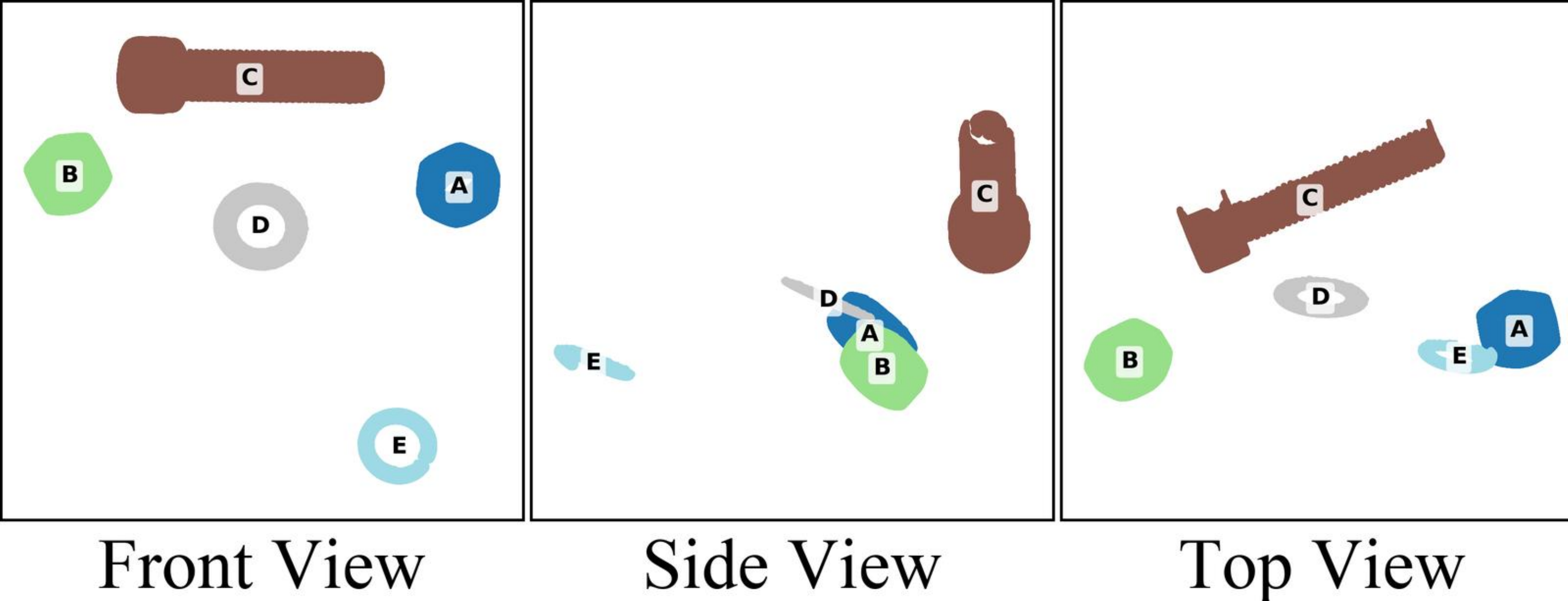}\\
    \small Extra Part: Two Hex Nuts
\end{minipage}\hfill
\begin{minipage}{0.48\textwidth}
    \centering
    \includegraphics[width=\linewidth]{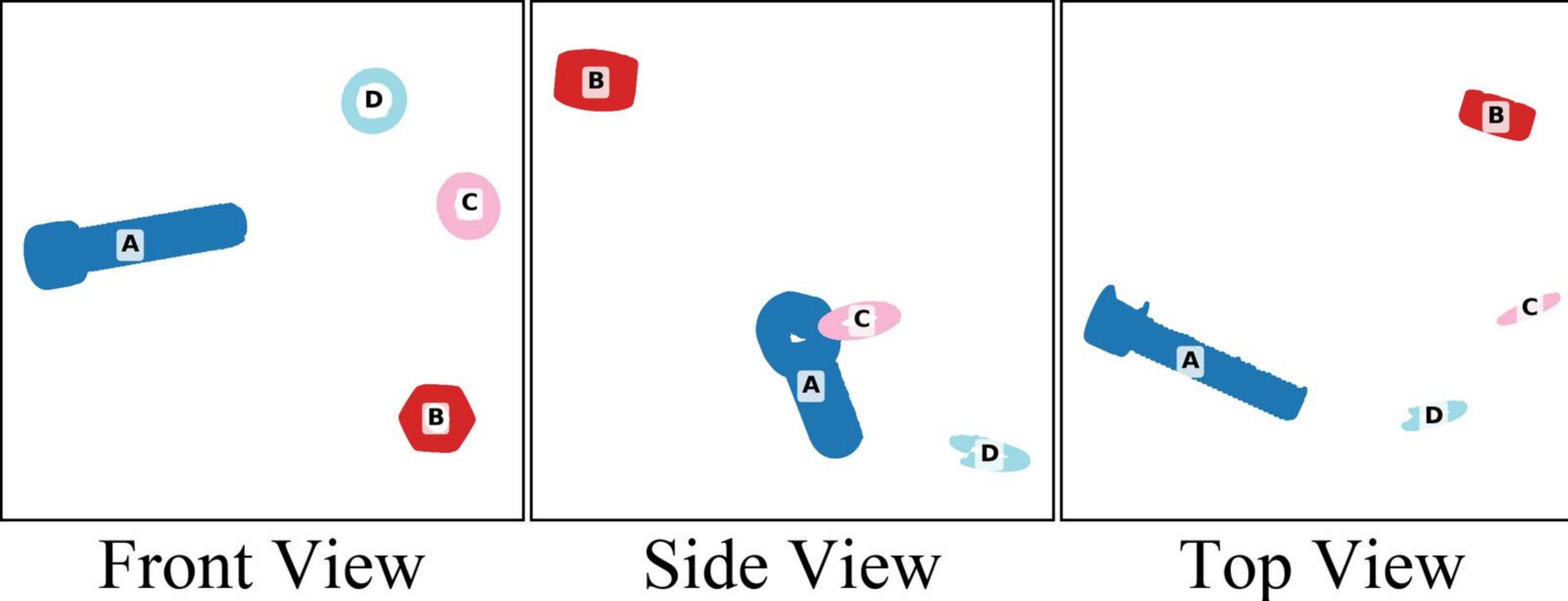}\\
    \small   Flat Washer M14
\end{minipage}

\begin{minipage}{0.48\textwidth}
    \centering
    \includegraphics[width=\linewidth]{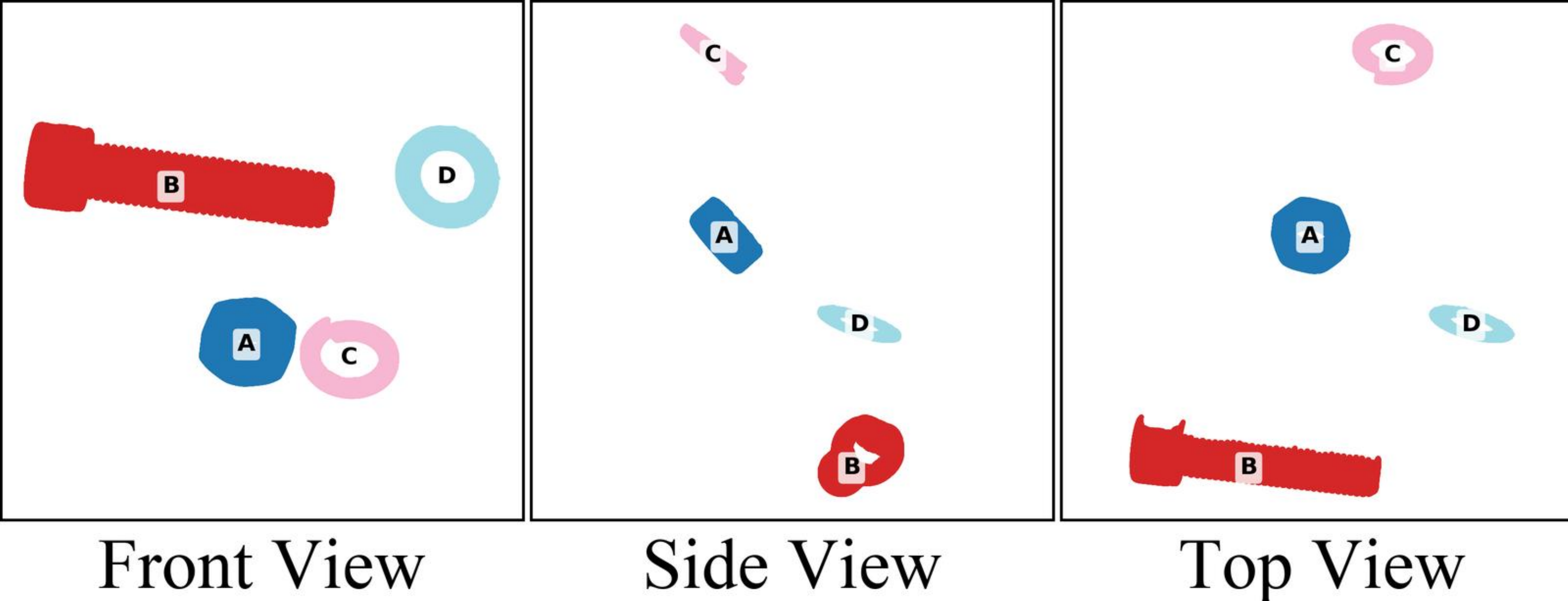}\\
    \small   Spring Washer M20
\end{minipage}\hfill
\begin{minipage}{0.48\textwidth}
    \centering
    \includegraphics[width=\linewidth]{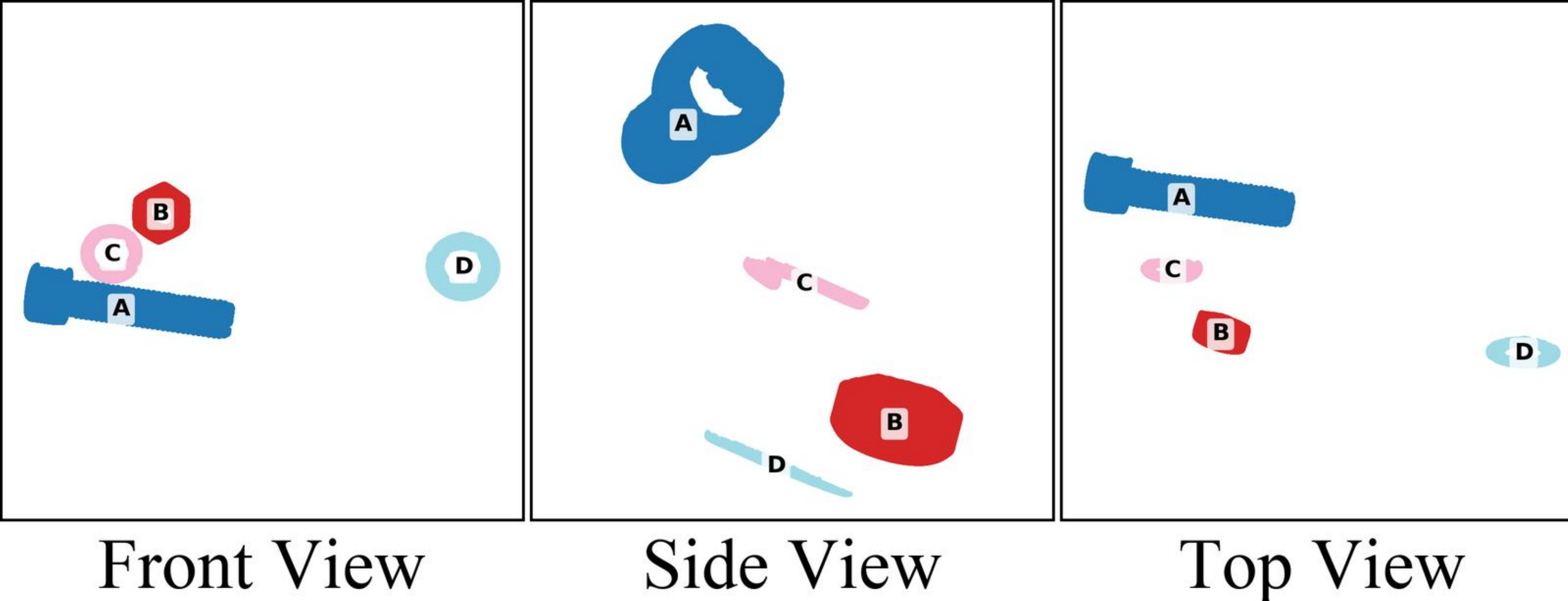}\\
    \small   Hex Nut M16
\end{minipage}

\begin{minipage}{0.48\textwidth}
    \centering
    \includegraphics[width=\linewidth]{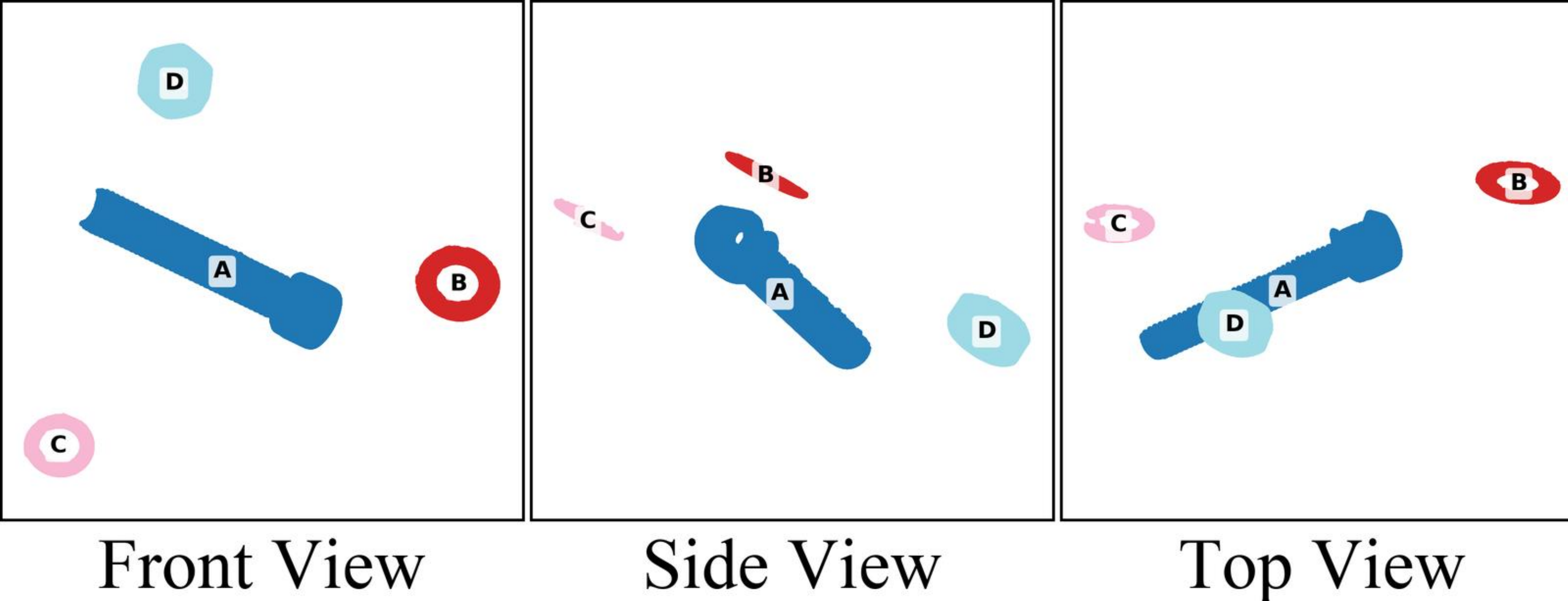}\\
    \small   Cup Head Screw M18 100
\end{minipage}\hfill
\begin{minipage}{0.48\textwidth}
    \centering
    \includegraphics[width=\linewidth]{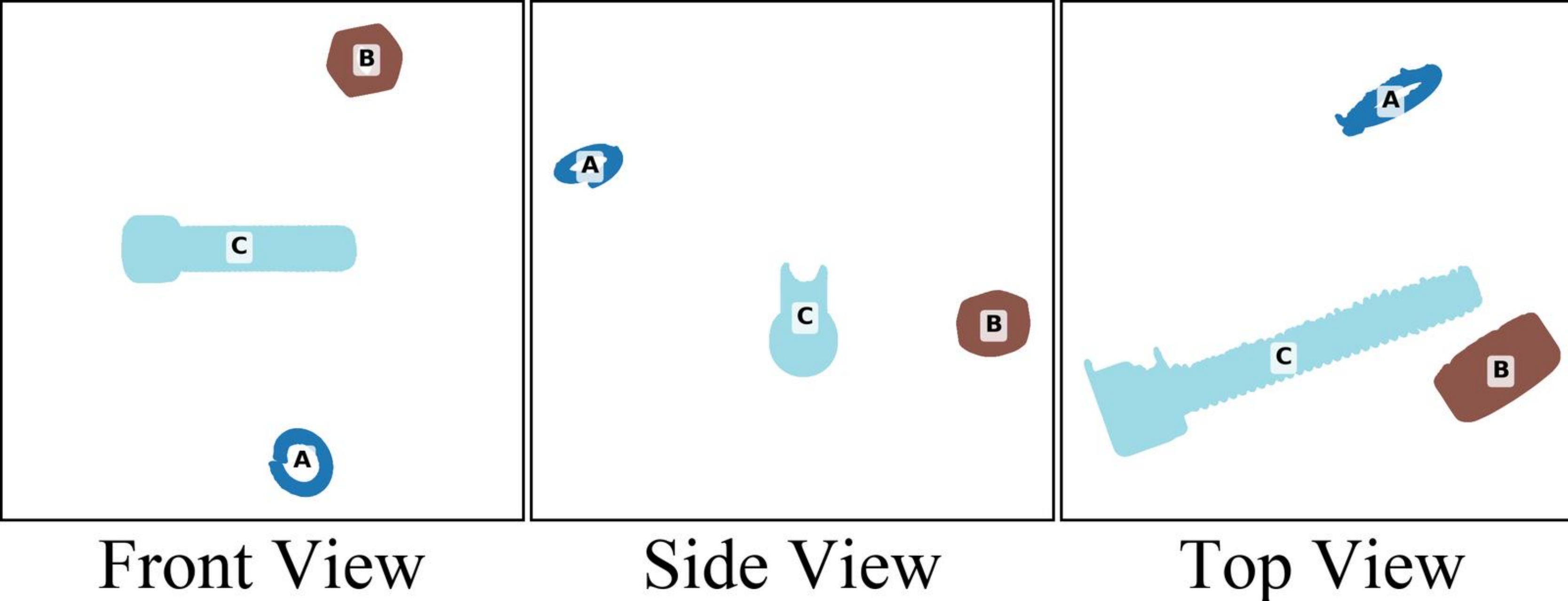}\\
    \small Missing Part: No Flat Washers
\end{minipage}

\begin{minipage}{0.48\textwidth}
    \centering
    \includegraphics[width=\linewidth]{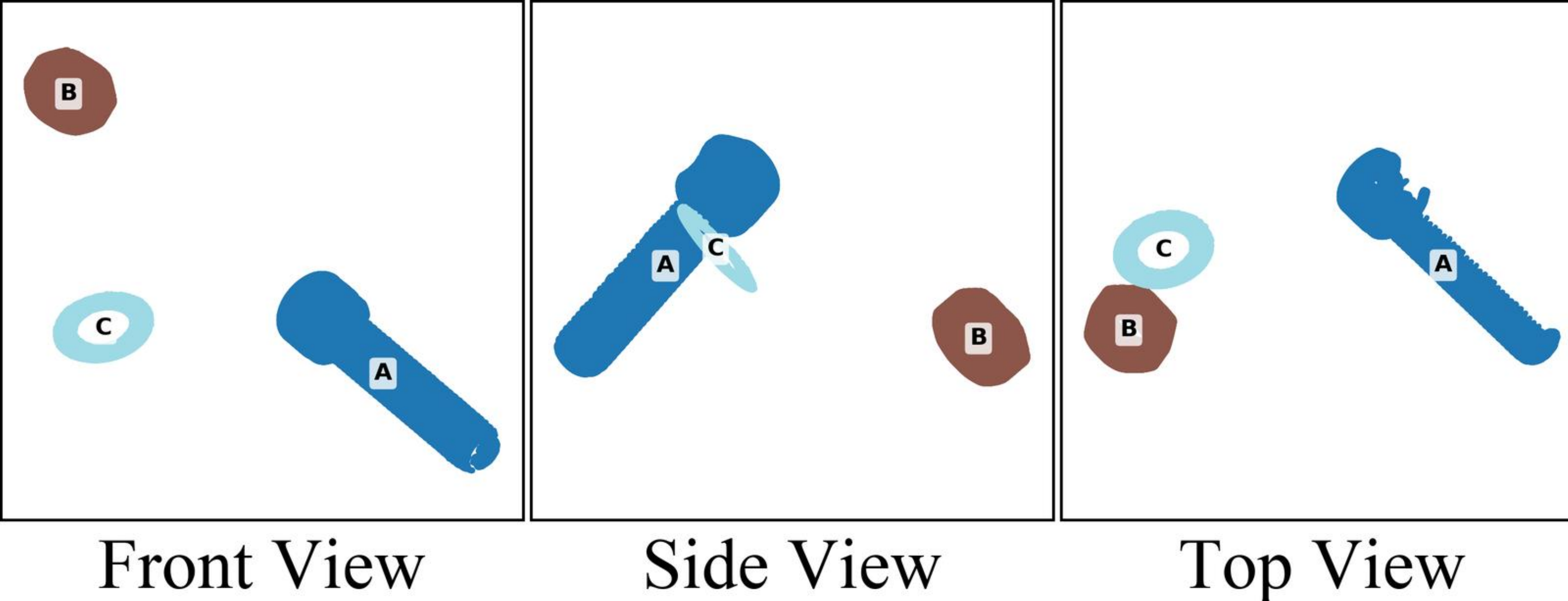}\\
    \small Missing Part: No Spring Washers
\end{minipage}

\caption{Random examples from \swnscen}
\label{fig: manufacturing_scenario_1_pointcloud_saved}
\end{figure}

This section presents randomly sampled images from the benchmark dataset to illustrate the visual characteristics and diversity of each scenario. We show examples for all three tasks: \taskone\ scenarios (Figures~\ref{fig: Manufacturing_task1_scenario_3_saved}--\ref{fig: Manufacturing_task1_scenario_2_pointcloud_saved}), \tasktwo\ workpiece types (Figures~\ref{fig: Corner_Bracket_saved}--\ref{fig: Wing_Screw_saved}), and \taskthree\ scenarios (Figures~\ref{fig: Manufacturing_scenario_1_saved}--\ref{fig: manufacturing_scenario_1_pointcloud_saved}). Each figure includes both normal cases and representative anomalous cases. Section~\ref{app:grounding_exemplars} additionally shows the Set-of-Mark annotated grounding images used in Bottleneck Analysis~A.

\newpage

\FloatBarrier

\subsection{The Grounding Examples of Image Data (Bottleneck Analysis A)}\label{app:grounding_exemplars}

\begin{figure}[htbp]
\centering

\begin{minipage}{0.23\textwidth}
    \centering
    \parbox[c][3.0cm][c]{\linewidth}{
        \centering\includegraphics[width=\linewidth,height=3.0cm,keepaspectratio]{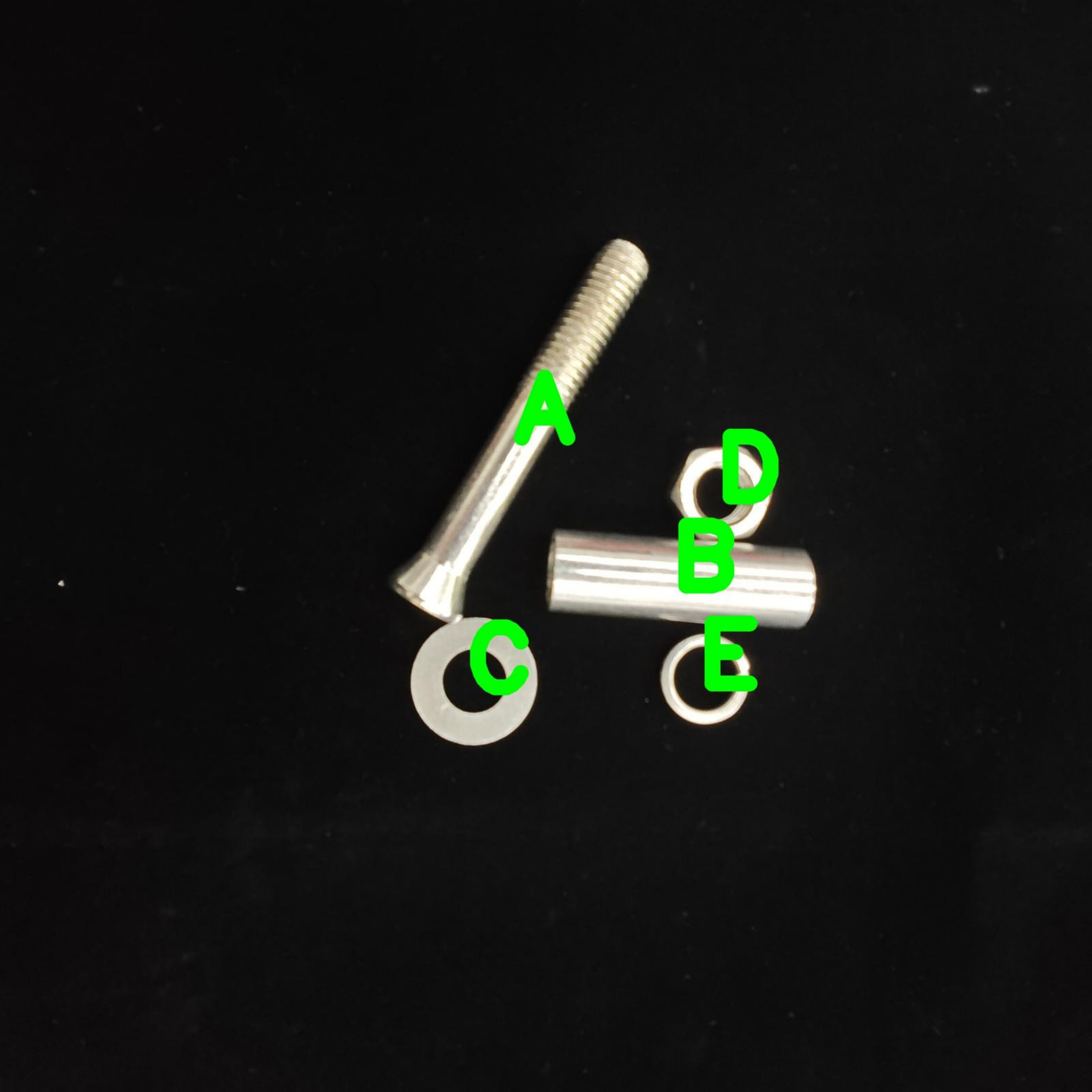}
    }
    \parbox[t][1.5cm][t]{\linewidth}{\small Normal Case}
\end{minipage}\hfill
\begin{minipage}{0.23\textwidth}
    \centering
    \parbox[c][3.0cm][c]{\linewidth}{
        \centering\includegraphics[width=\linewidth,height=3.0cm,keepaspectratio]{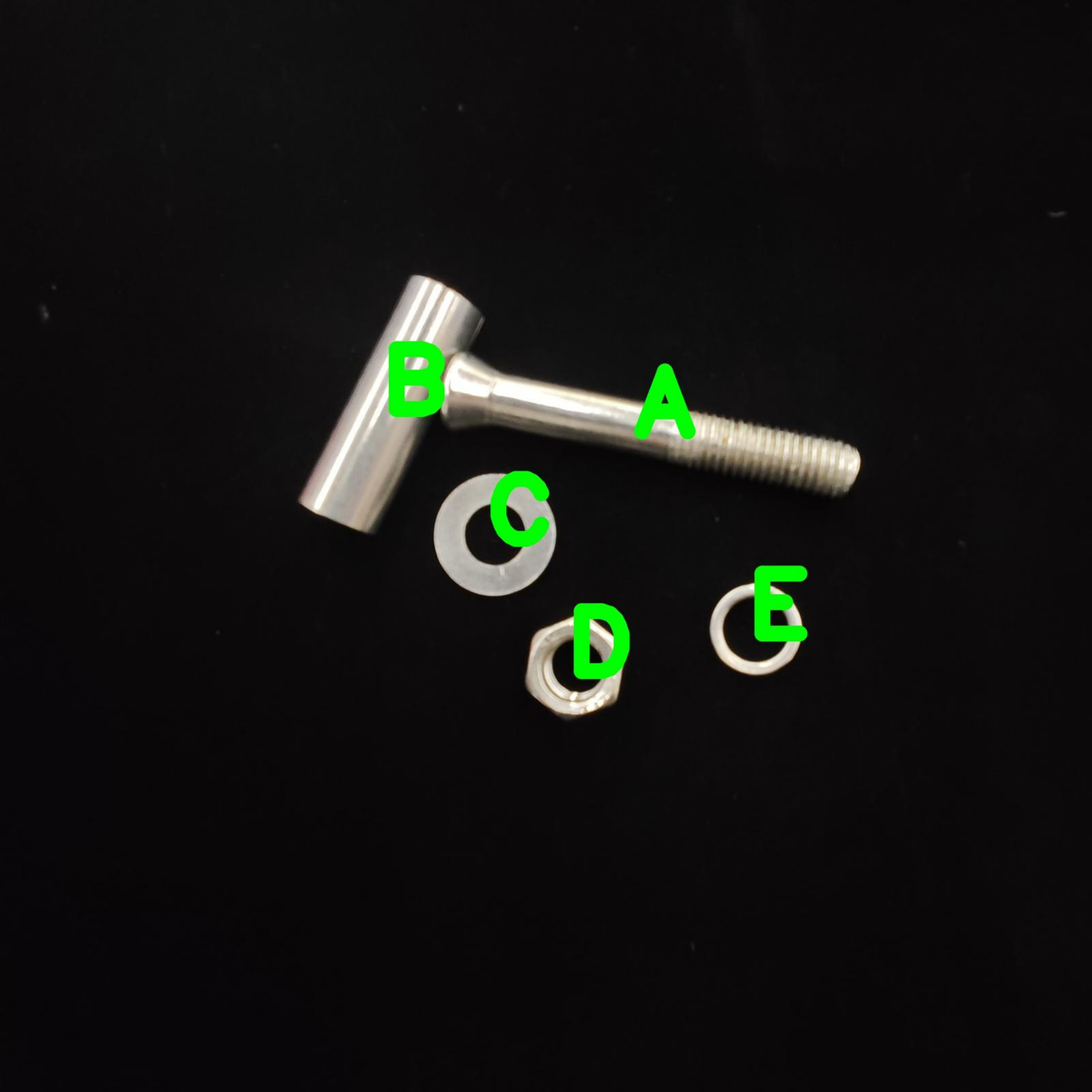}
    }
    \parbox[t][1.5cm][t]{\linewidth}{\small Normal Case}
\end{minipage}\hfill
\begin{minipage}{0.23\textwidth}
    \centering
    \parbox[c][3.0cm][c]{\linewidth}{
        \centering\includegraphics[width=\linewidth,height=3.0cm,keepaspectratio]{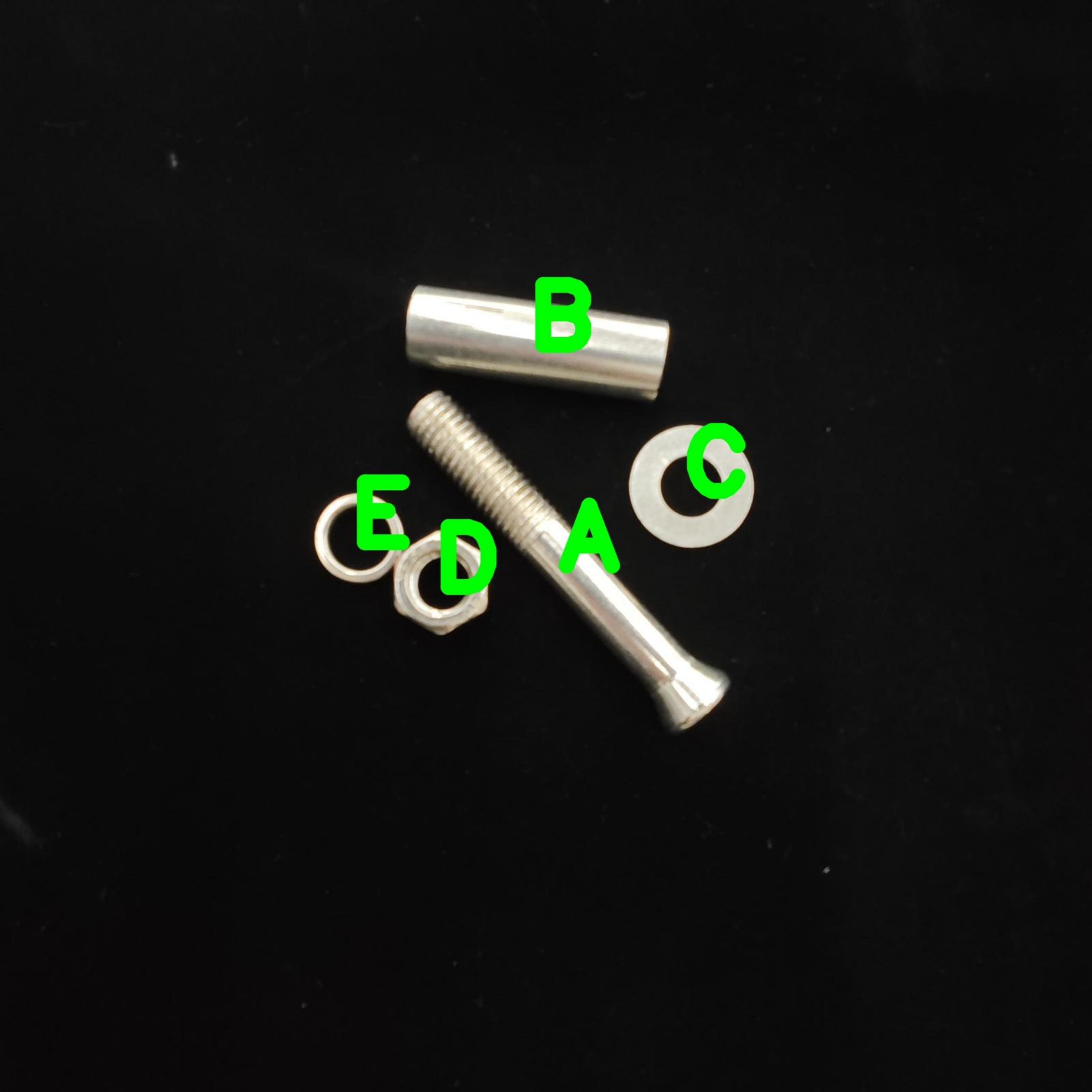}
    }
    \parbox[t][1.5cm][t]{\linewidth}{\small Normal Case}
\end{minipage}\hfill
\begin{minipage}{0.23\textwidth}
    \centering
    \parbox[c][3.0cm][c]{\linewidth}{
        \centering\includegraphics[width=\linewidth,height=3.0cm,keepaspectratio]{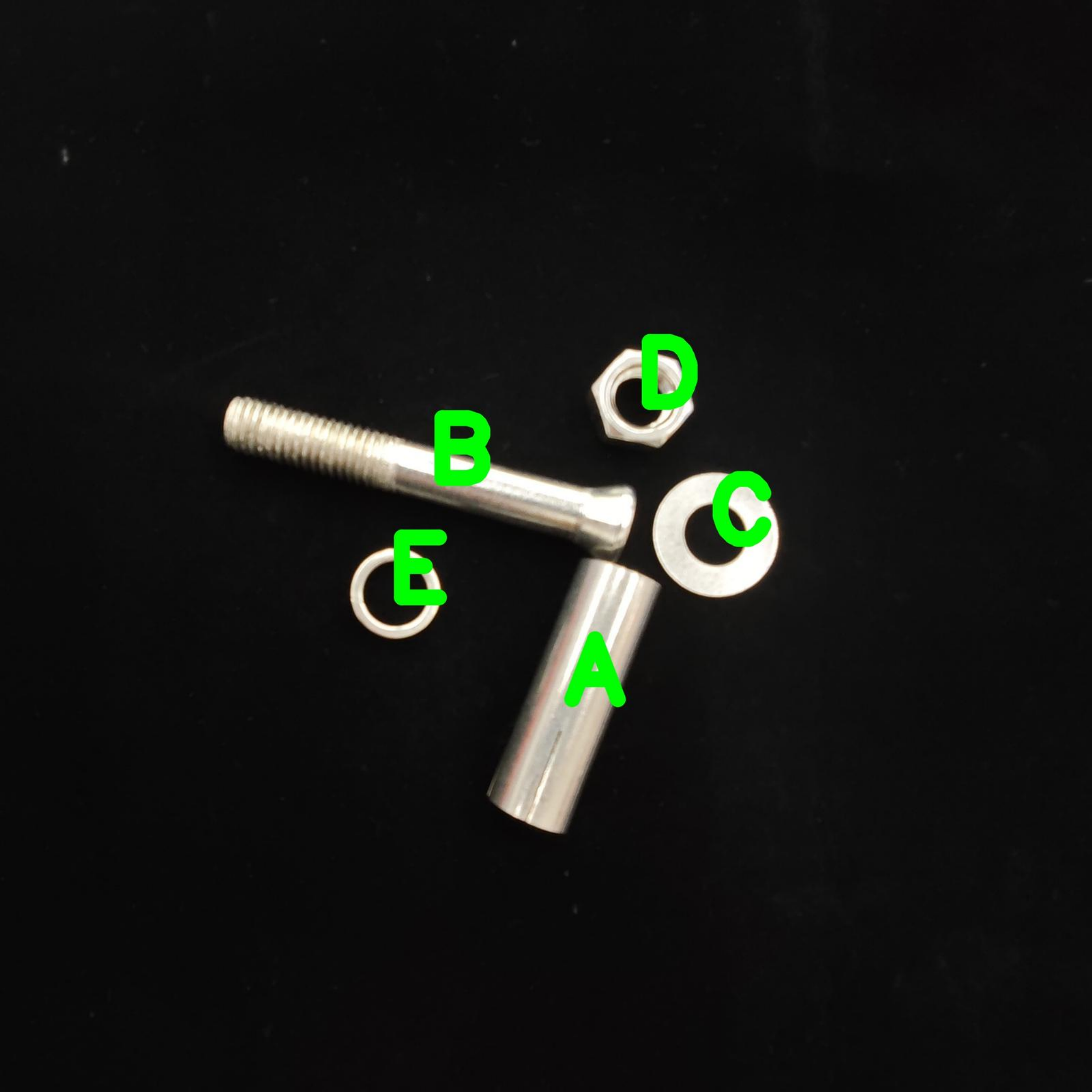}
    }
    \parbox[t][1.5cm][t]{\linewidth}{\small Normal Case}
\end{minipage}

\begin{minipage}{0.23\textwidth}
    \centering
    \parbox[c][3.0cm][c]{\linewidth}{
        \centering\includegraphics[width=\linewidth,height=3.0cm,keepaspectratio]{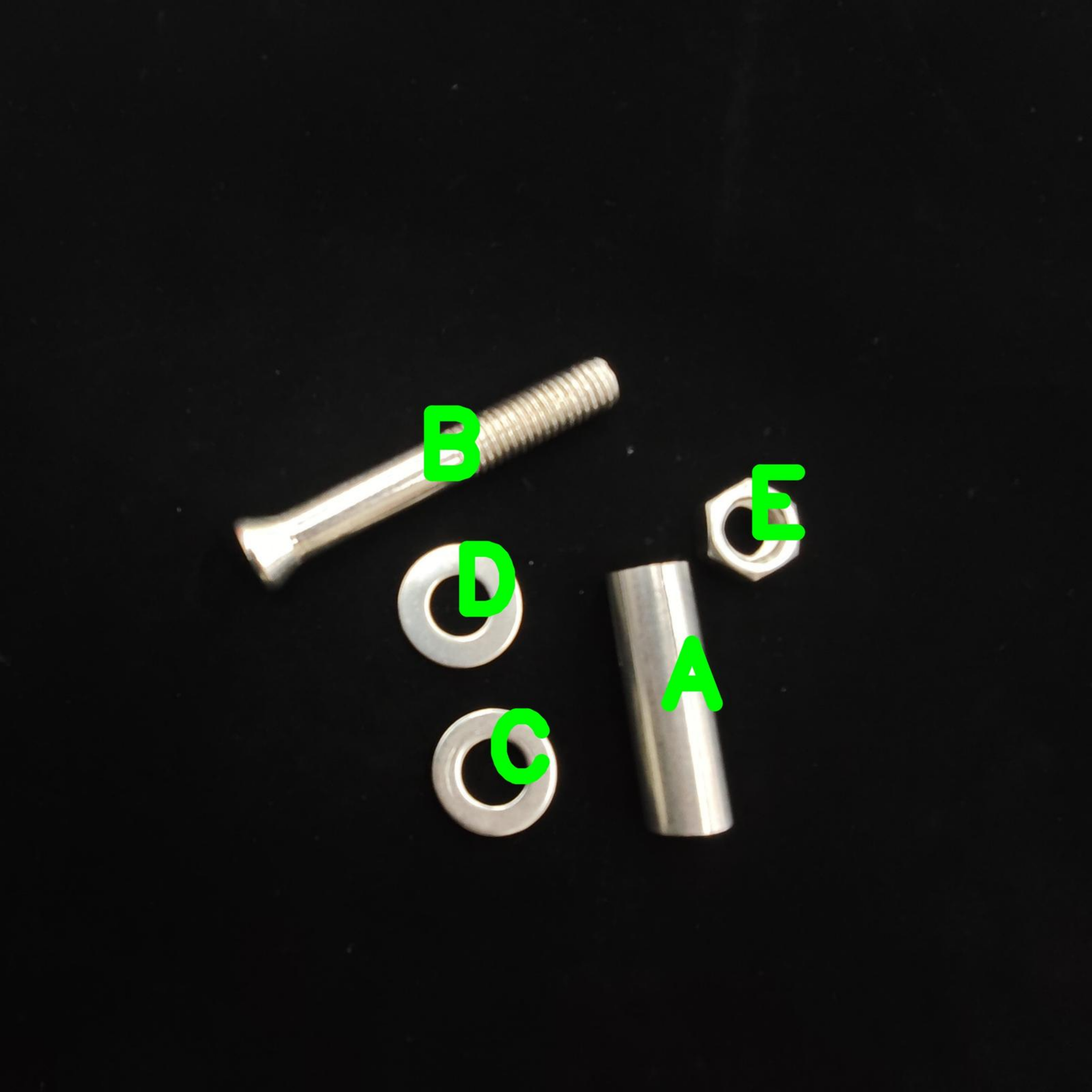}
    }
    \parbox[t][1.5cm][t]{\linewidth}{\small Two Flat Washers}
\end{minipage}\hfill
\begin{minipage}{0.23\textwidth}
    \centering
    \parbox[c][3.0cm][c]{\linewidth}{
        \centering\includegraphics[width=\linewidth,height=3.0cm,keepaspectratio]{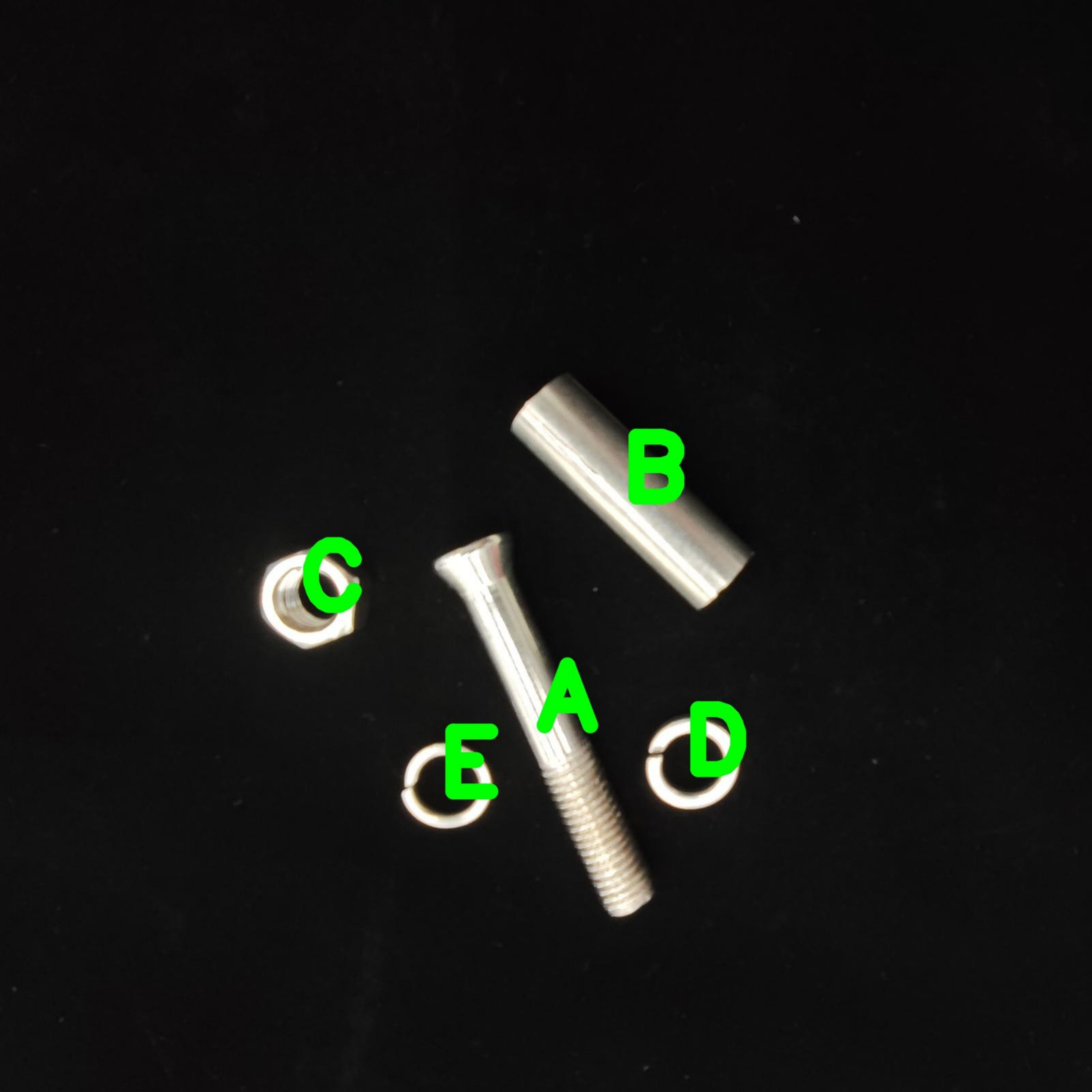}
    }
    \parbox[t][1.5cm][t]{\linewidth}{\small Two Spring Washers}
\end{minipage}\hfill
\begin{minipage}{0.23\textwidth}
    \centering
    \parbox[c][3.0cm][c]{\linewidth}{
        \centering\includegraphics[width=\linewidth,height=3.0cm,keepaspectratio]{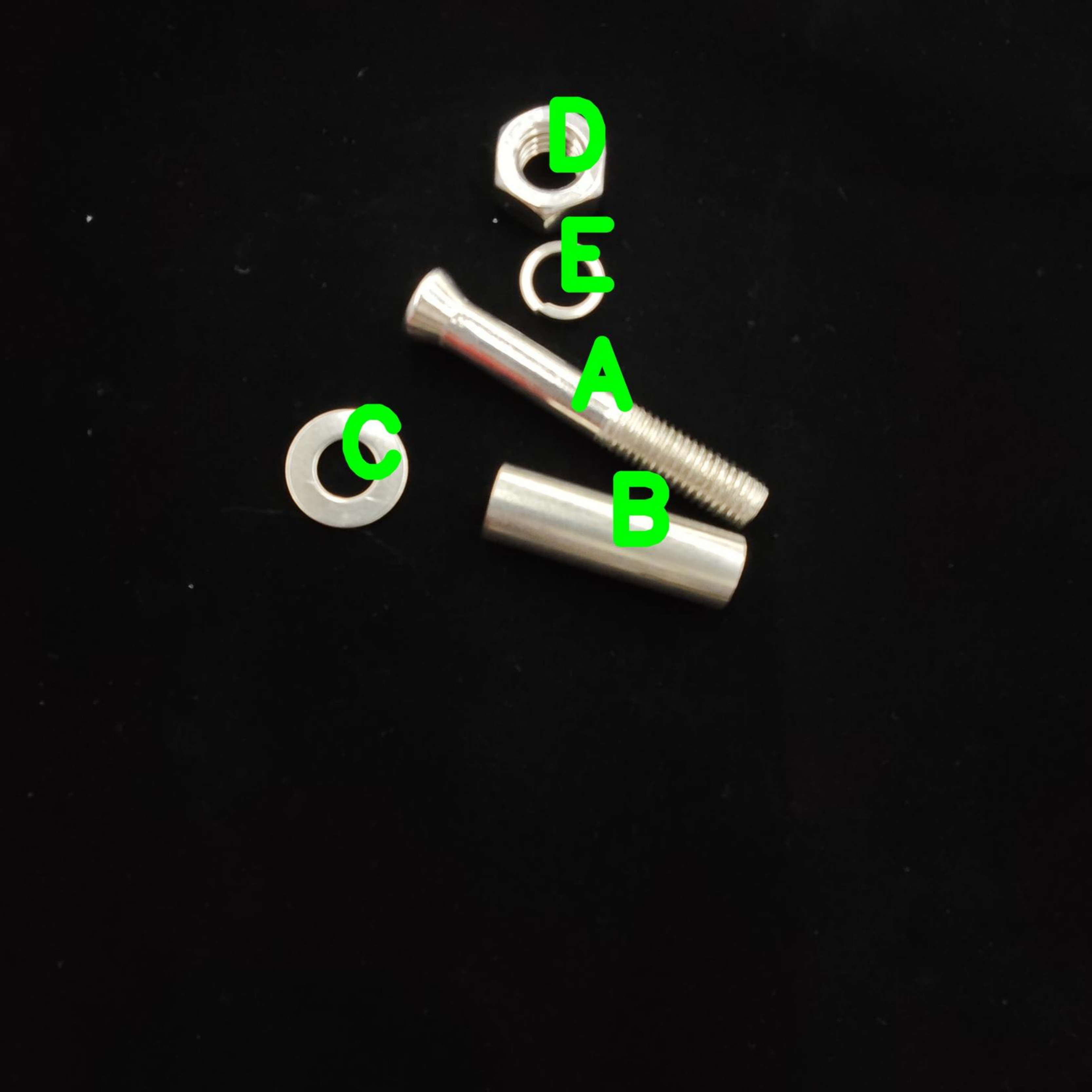}
    }
    \parbox[t][1.5cm][t]{\linewidth}{\small Hex Nut M14}
\end{minipage}\hfill
\begin{minipage}{0.23\textwidth}
    \centering
    \parbox[c][3.0cm][c]{\linewidth}{
        \centering\includegraphics[width=\linewidth,height=3.0cm,keepaspectratio]{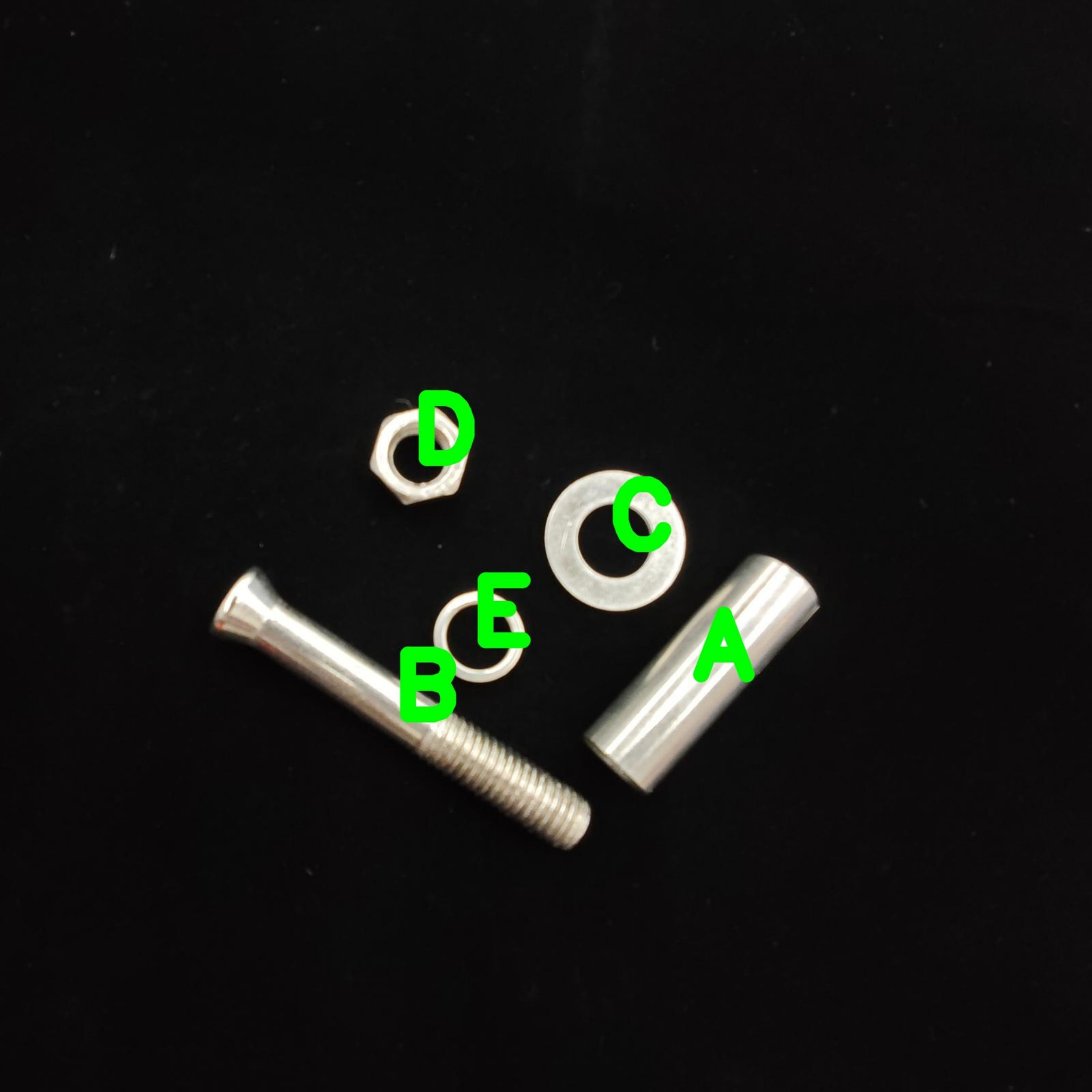}
    }
    \parbox[t][1.5cm][t]{\linewidth}{\small Flat Washer M14}
\end{minipage}

\begin{minipage}{0.23\textwidth}
    \centering
    \parbox[c][3.0cm][c]{\linewidth}{
        \centering\includegraphics[width=\linewidth,height=3.0cm,keepaspectratio]{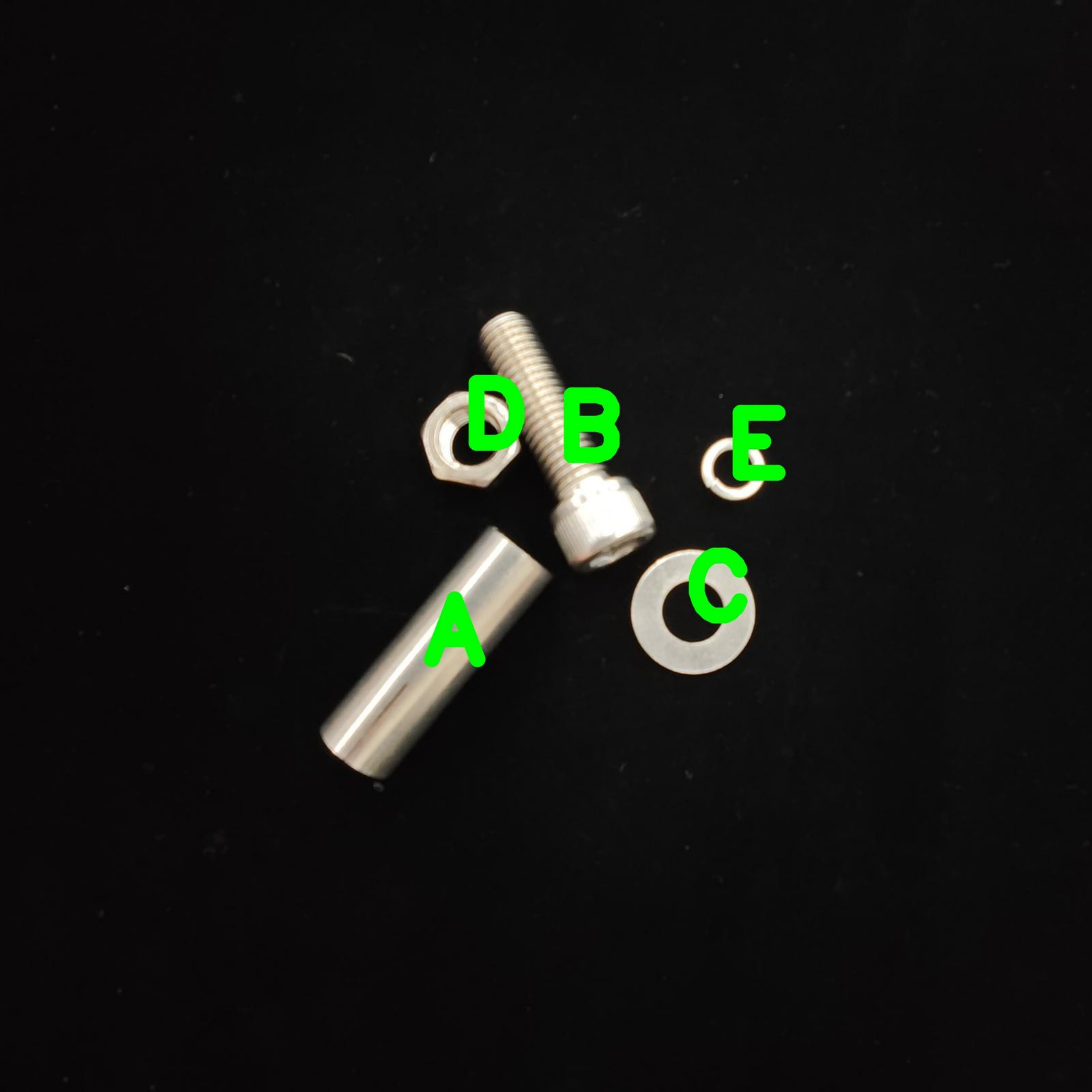}
    }
    \parbox[t][1.5cm][t]{\linewidth}{\small Cup Head Screw M12 40}
\end{minipage}\hfill
\begin{minipage}{0.23\textwidth}
    \centering
    \parbox[c][3.0cm][c]{\linewidth}{
        \centering\includegraphics[width=\linewidth,height=3.0cm,keepaspectratio]{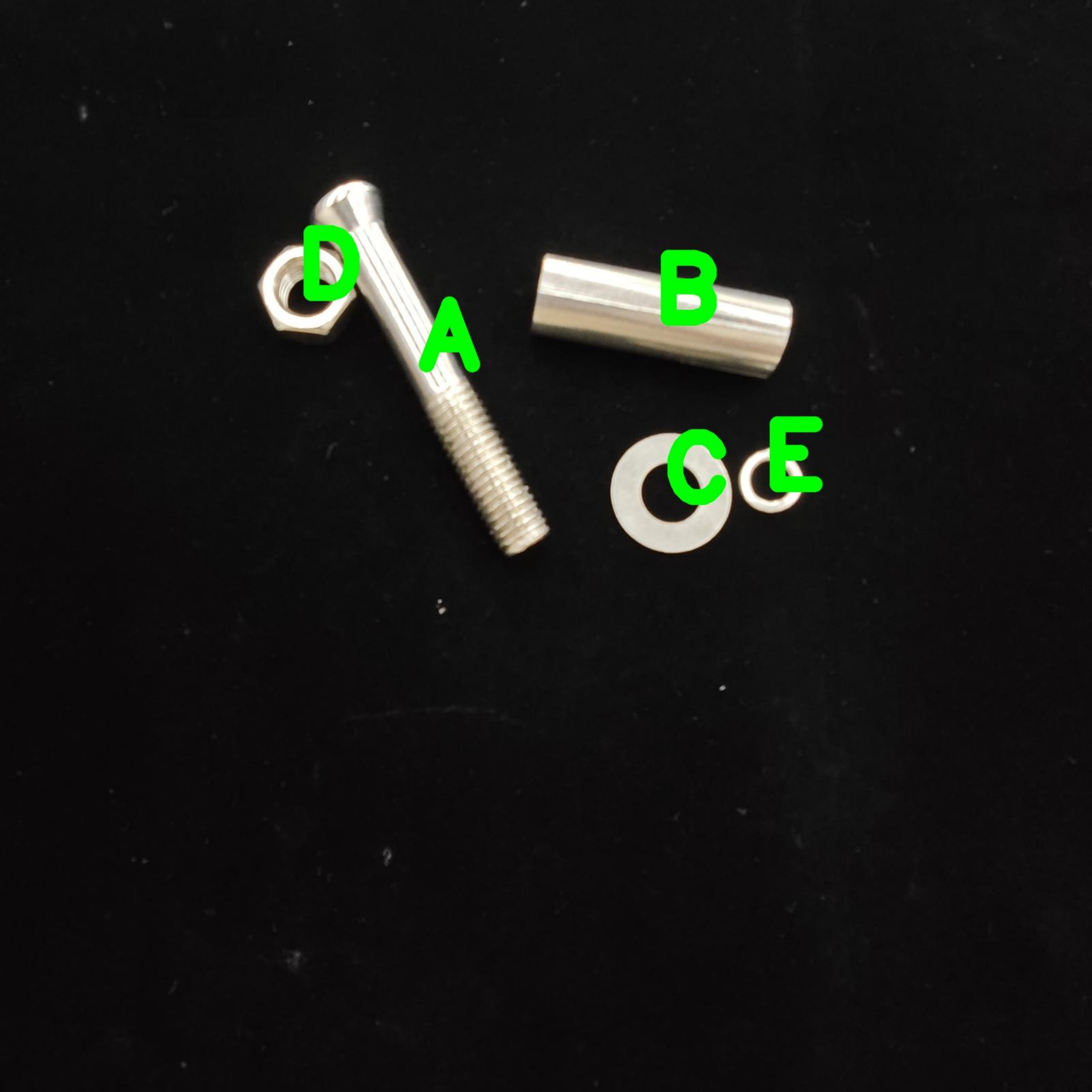}
    }
    \parbox[t][1.5cm][t]{\linewidth}{\small Spring Washer M8}
\end{minipage}\hfill
\begin{minipage}{0.23\textwidth}
    \centering
    \parbox[c][3.0cm][c]{\linewidth}{
        \centering\includegraphics[width=\linewidth,height=3.0cm,keepaspectratio]{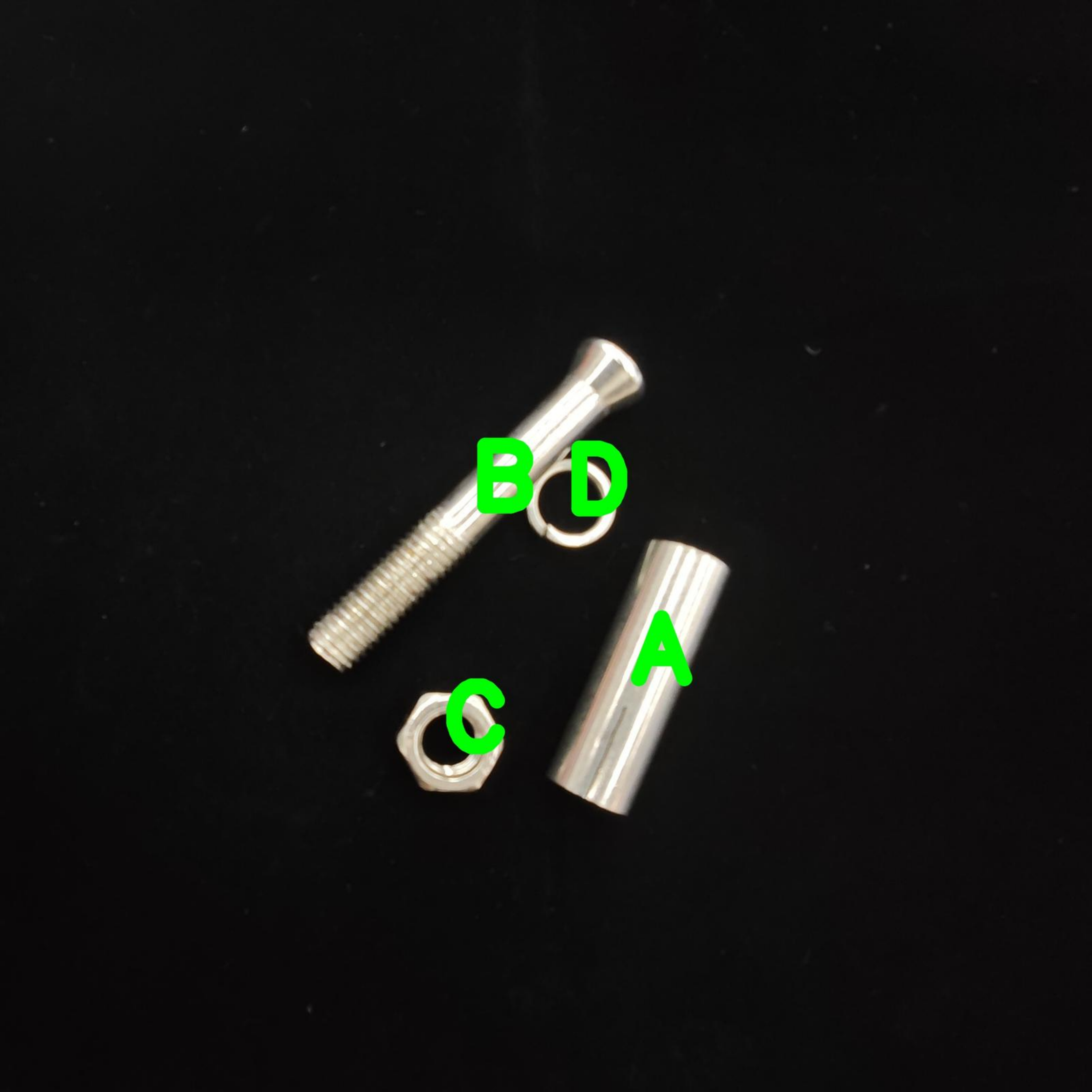}
    }
    \parbox[t][1.5cm][t]{\linewidth}{\small No Flat Washers}
\end{minipage}\hfill
\begin{minipage}{0.23\textwidth}
    \centering
    \parbox[c][3.0cm][c]{\linewidth}{
        \centering\includegraphics[width=\linewidth,height=3.0cm,keepaspectratio]{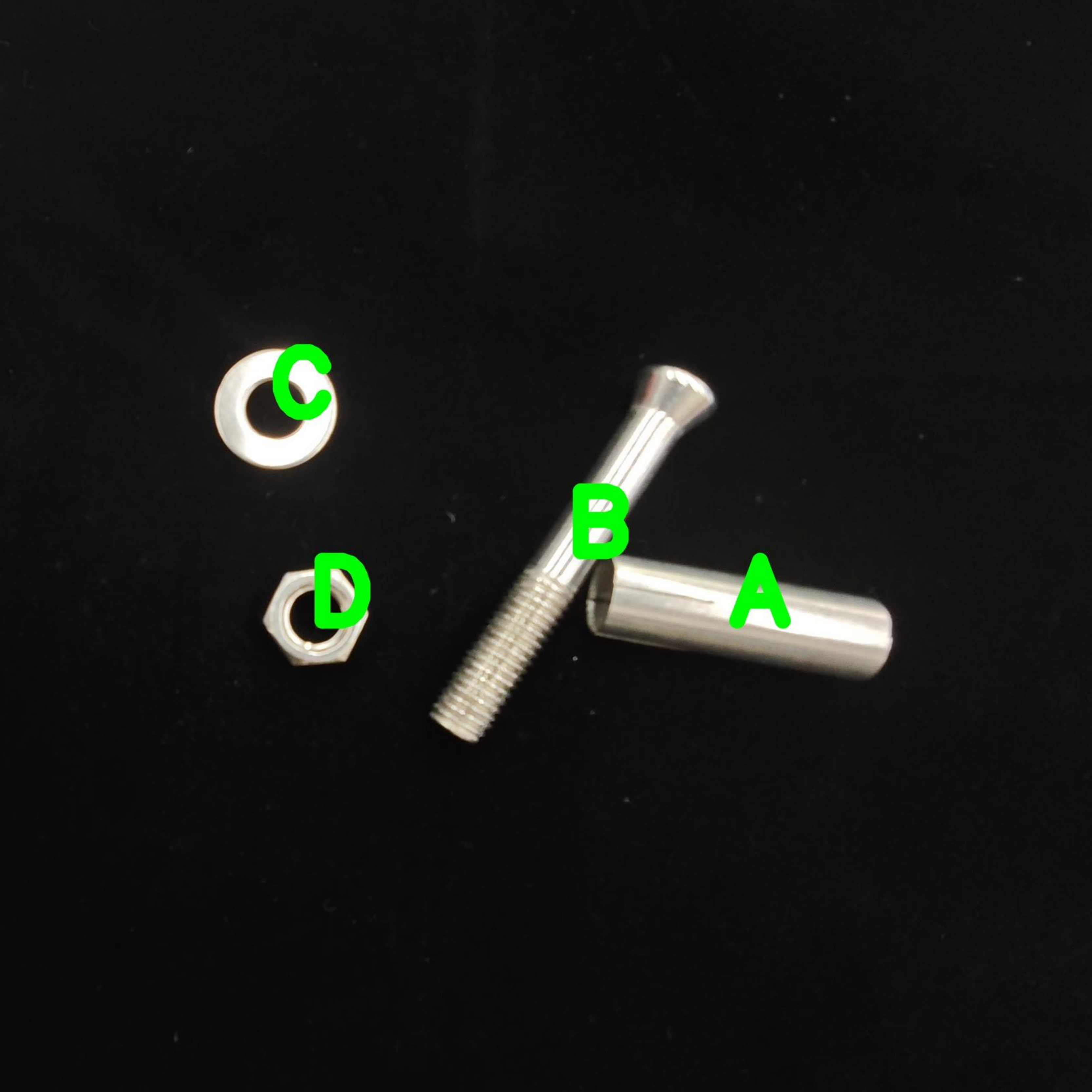}
    }
    \parbox[t][1.5cm][t]{\linewidth}{\small No Spring Washers}
\end{minipage}

\caption{Random grounding examples from \messcen}
\label{figab: Manufacturing_scenario_1_saved}
\end{figure}

\begin{figure}[htbp]
\centering

\begin{minipage}{0.23\textwidth}
    \centering
    \parbox[c][3.0cm][c]{\linewidth}{
        \centering\includegraphics[width=\linewidth,height=3.0cm,keepaspectratio]{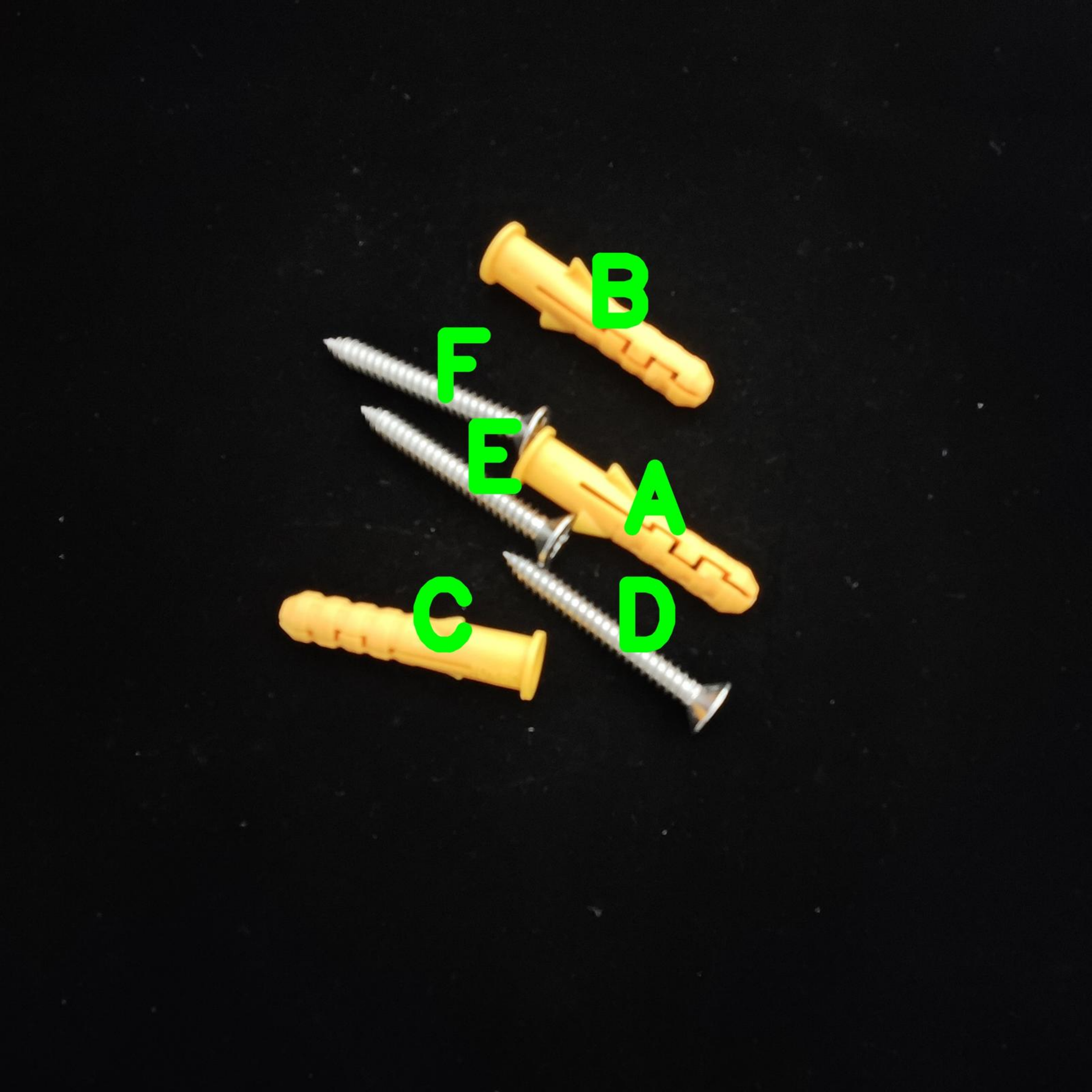}
    }
    \parbox[t][1.5cm][t]{\linewidth}{\small Normal Case}
\end{minipage}\hfill
\begin{minipage}{0.23\textwidth}
    \centering
    \parbox[c][3.0cm][c]{\linewidth}{
        \centering\includegraphics[width=\linewidth,height=3.0cm,keepaspectratio]{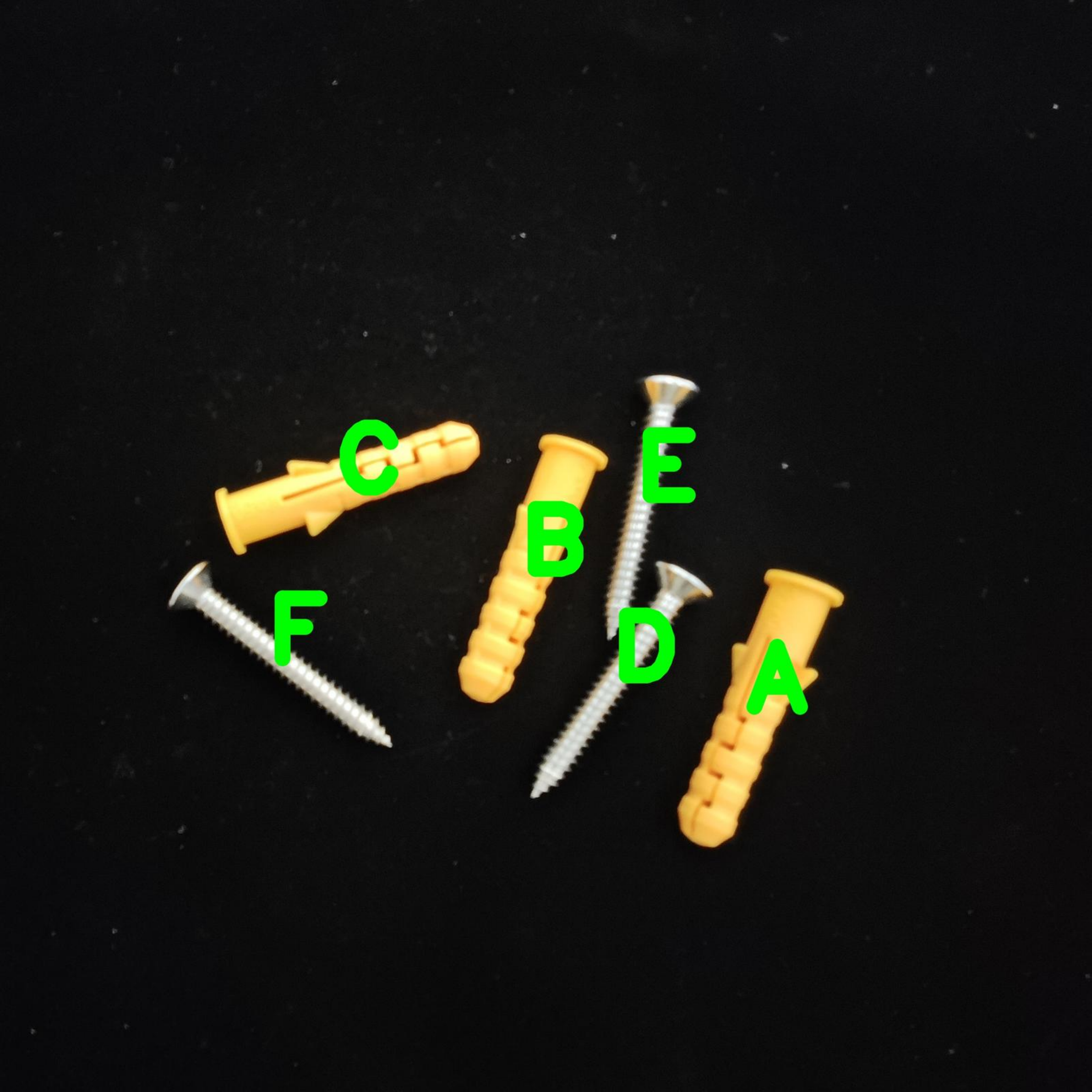}
    }
    \parbox[t][1.5cm][t]{\linewidth}{\small Normal Case}
\end{minipage}\hfill
\begin{minipage}{0.23\textwidth}
    \centering
    \parbox[c][3.0cm][c]{\linewidth}{
        \centering\includegraphics[width=\linewidth,height=3.0cm,keepaspectratio]{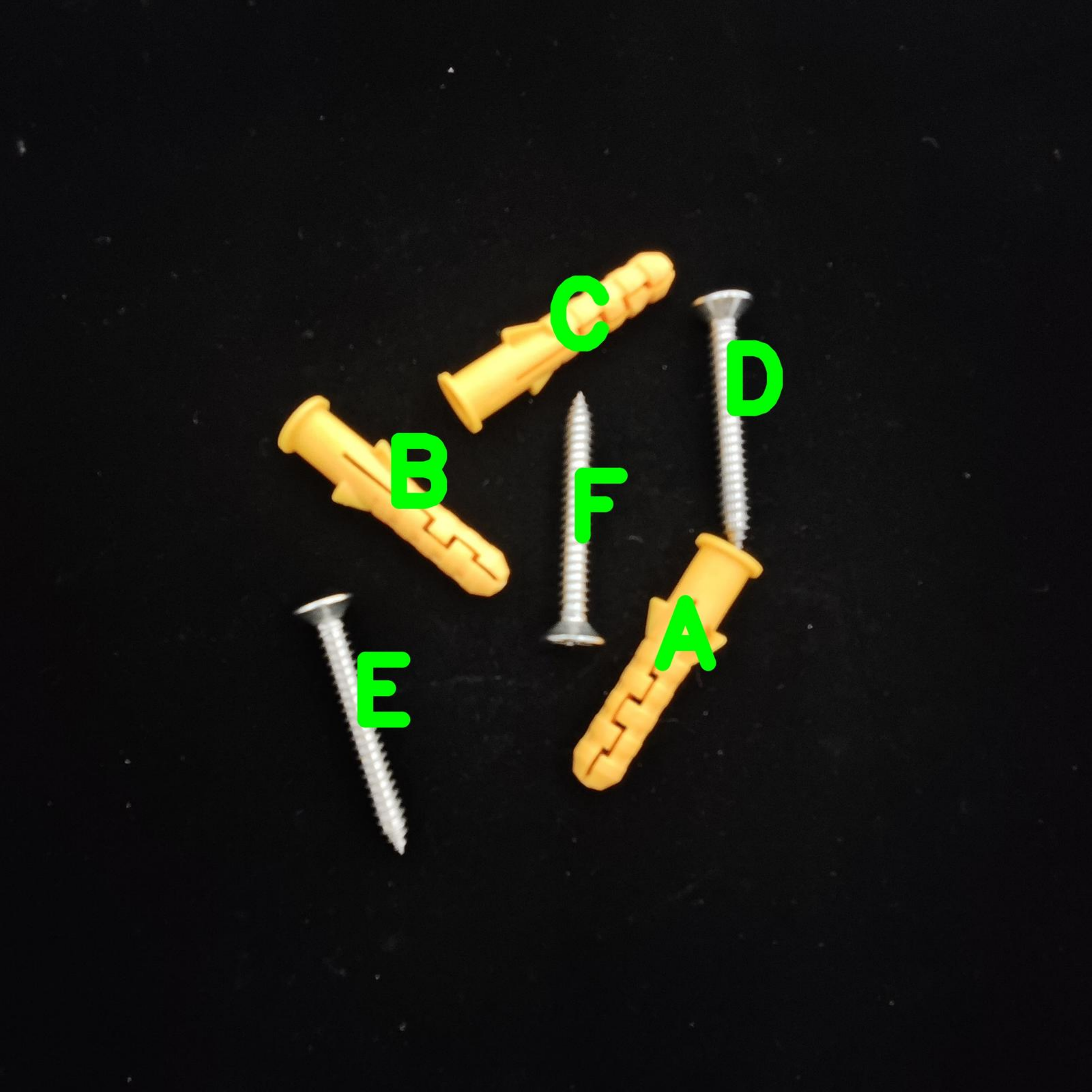}
    }
    \parbox[t][1.5cm][t]{\linewidth}{\small Normal Case}
\end{minipage}\hfill
\begin{minipage}{0.23\textwidth}
    \centering
    \parbox[c][3.0cm][c]{\linewidth}{
        \centering\includegraphics[width=\linewidth,height=3.0cm,keepaspectratio]{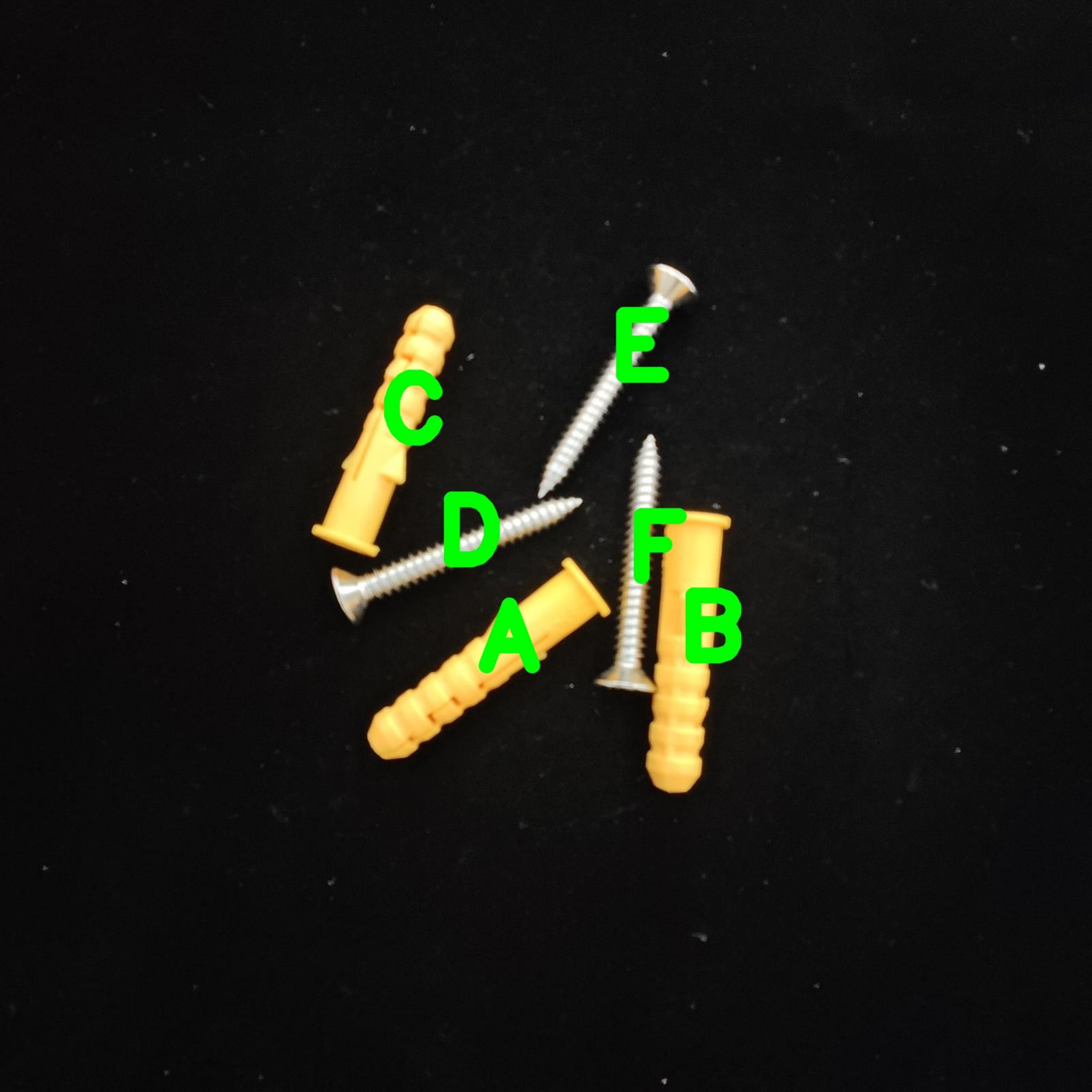}
    }
    \parbox[t][1.5cm][t]{\linewidth}{\small Normal Case}
\end{minipage}

\begin{minipage}{0.23\textwidth}
    \centering
    \parbox[c][3.0cm][c]{\linewidth}{
        \centering\includegraphics[width=\linewidth,height=3.0cm,keepaspectratio]{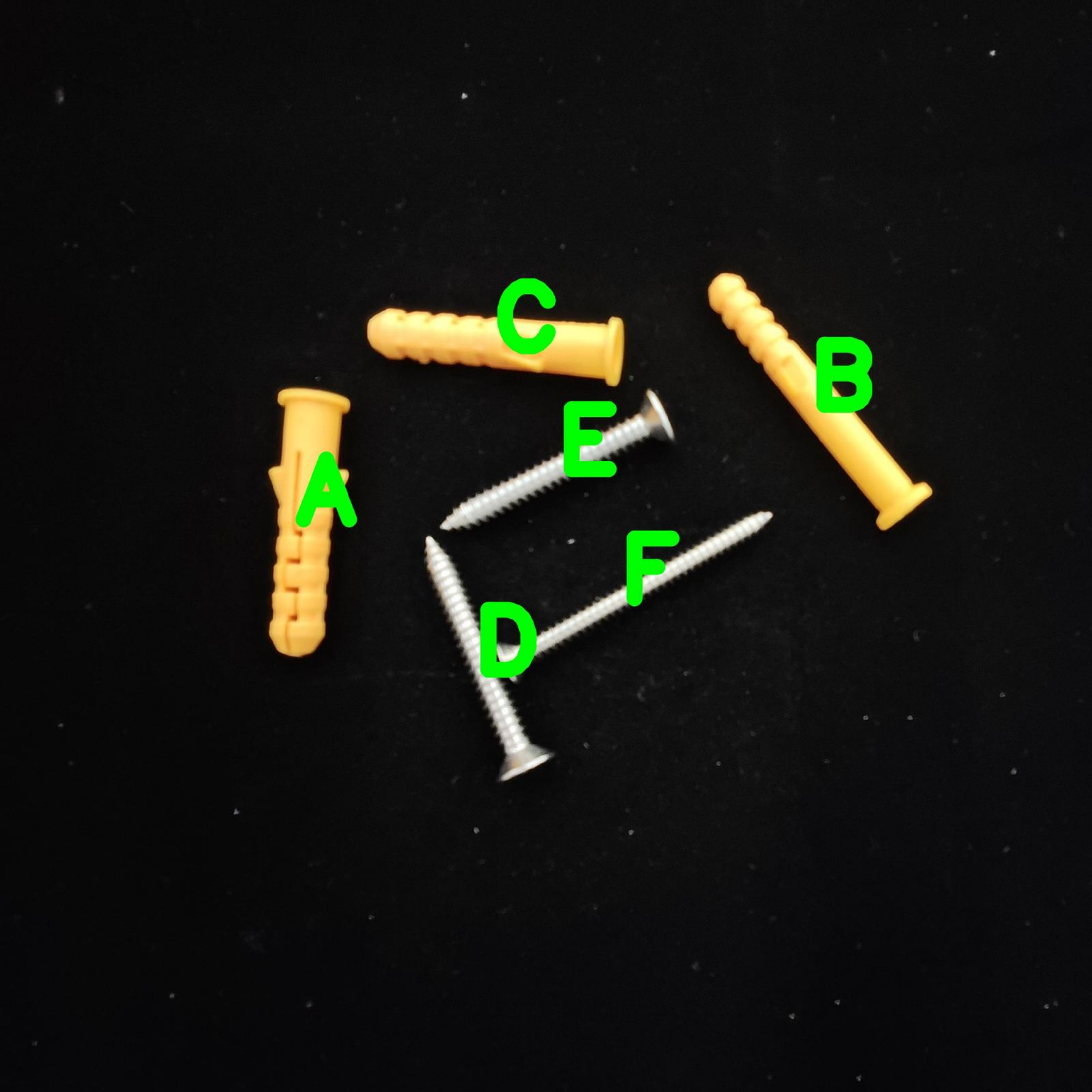}
    }
    \parbox[t][1.5cm][t]{\linewidth}{\small Model 860 expansion screws were mistakenly included}
\end{minipage}\hfill
\begin{minipage}{0.23\textwidth}
    \centering
    \parbox[c][3.0cm][c]{\linewidth}{
        \centering\includegraphics[width=\linewidth,height=3.0cm,keepaspectratio]{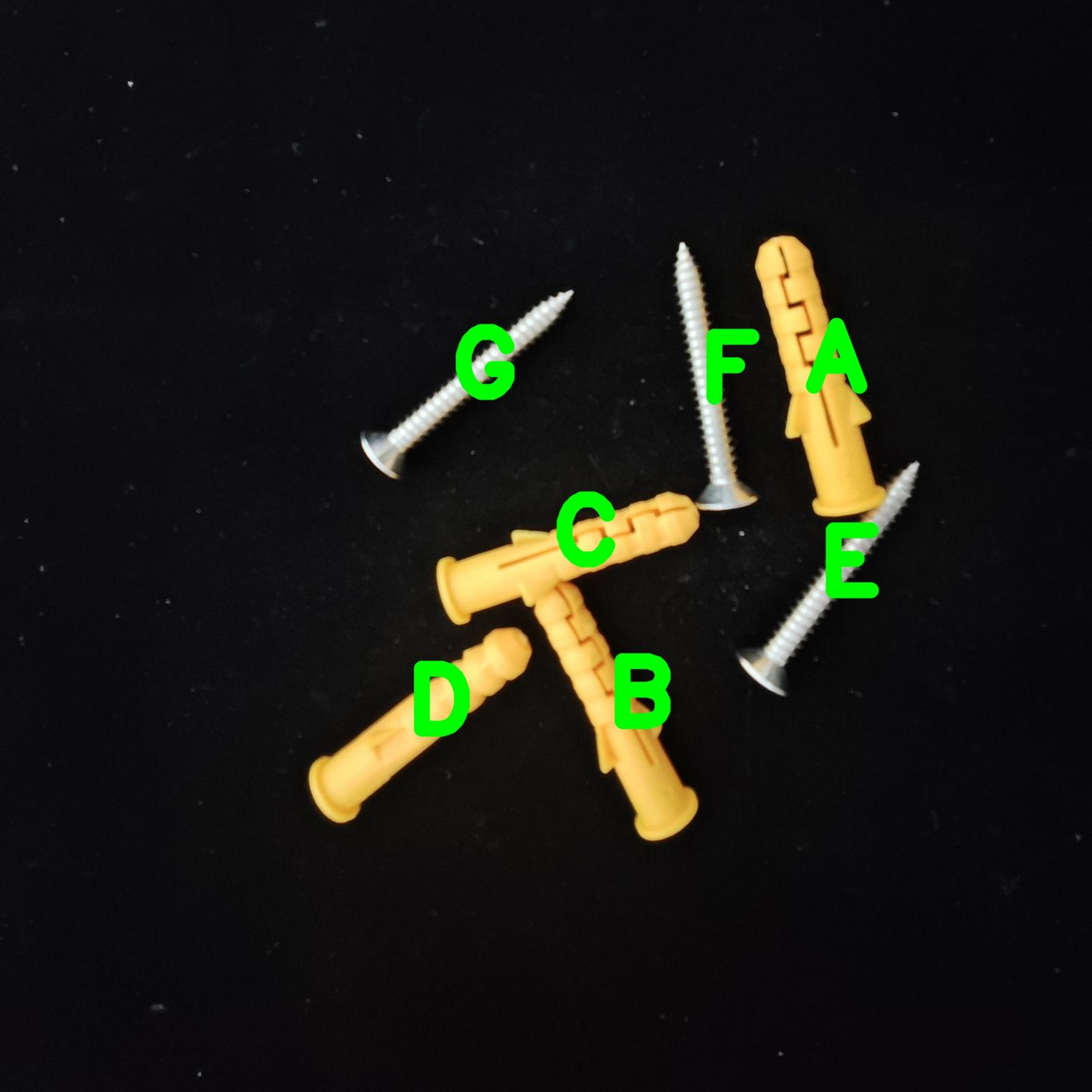}
    }
    \parbox[t][1.5cm][t]{\linewidth}{\small One extra plastic expansion anchor}
\end{minipage}\hfill
\begin{minipage}{0.23\textwidth}
    \centering
    \parbox[c][3.0cm][c]{\linewidth}{
        \centering\includegraphics[width=\linewidth,height=3.0cm,keepaspectratio]{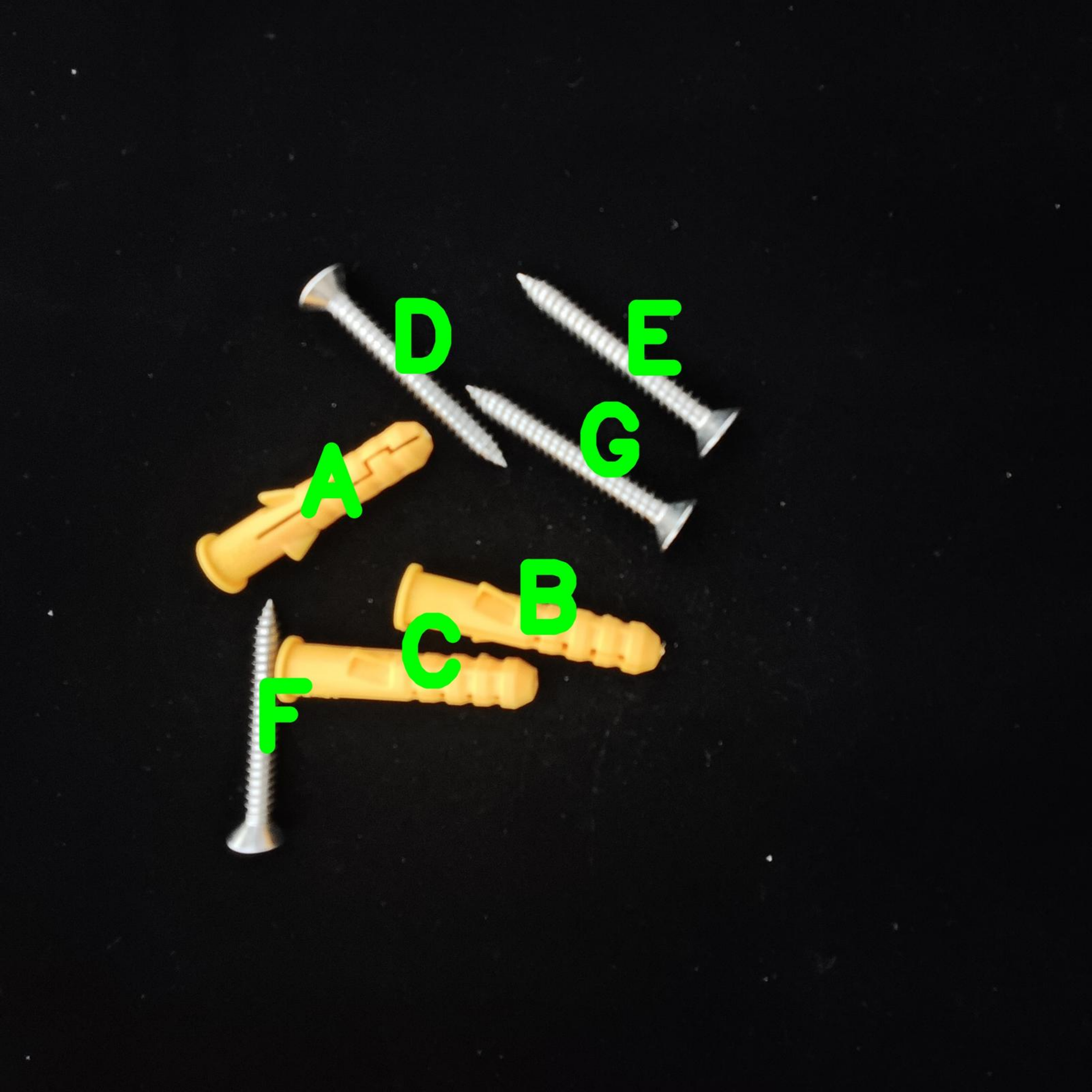}
    }
    \parbox[t][1.5cm][t]{\linewidth}{\small One extra plastic expansion Self-tapping Screw}
\end{minipage}\hfill
\begin{minipage}{0.23\textwidth}
    \centering
    \parbox[c][3.0cm][c]{\linewidth}{
        \centering\includegraphics[width=\linewidth,height=3.0cm,keepaspectratio]{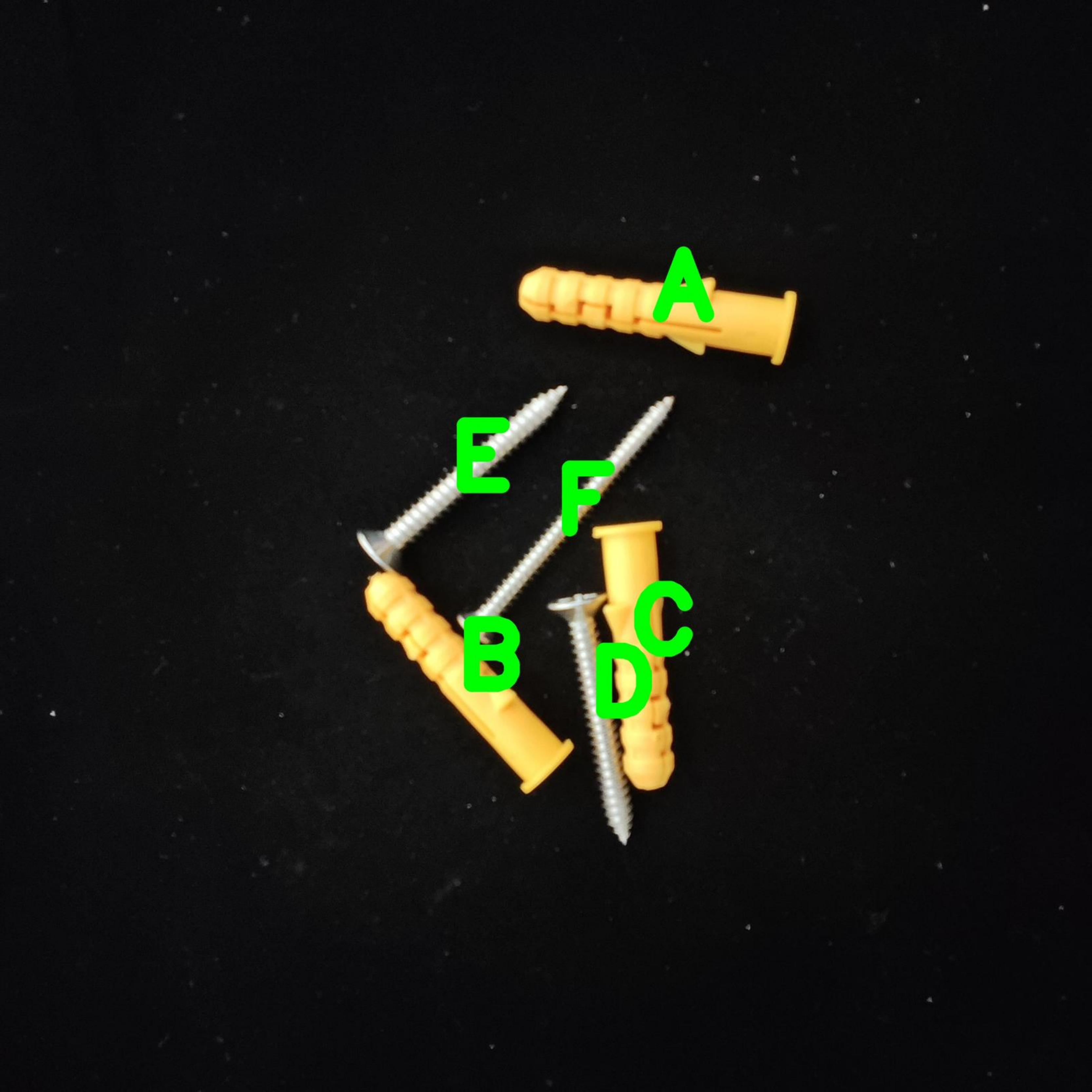}
    }
    \parbox[t][1.5cm][t]{\linewidth}{\small Screws belonging to model 860 expansion anchors were found mixed in}
\end{minipage}

\begin{minipage}{0.23\textwidth}
    \centering
    \parbox[c][3.0cm][c]{\linewidth}{
        \centering\includegraphics[width=\linewidth,height=3.0cm,keepaspectratio]{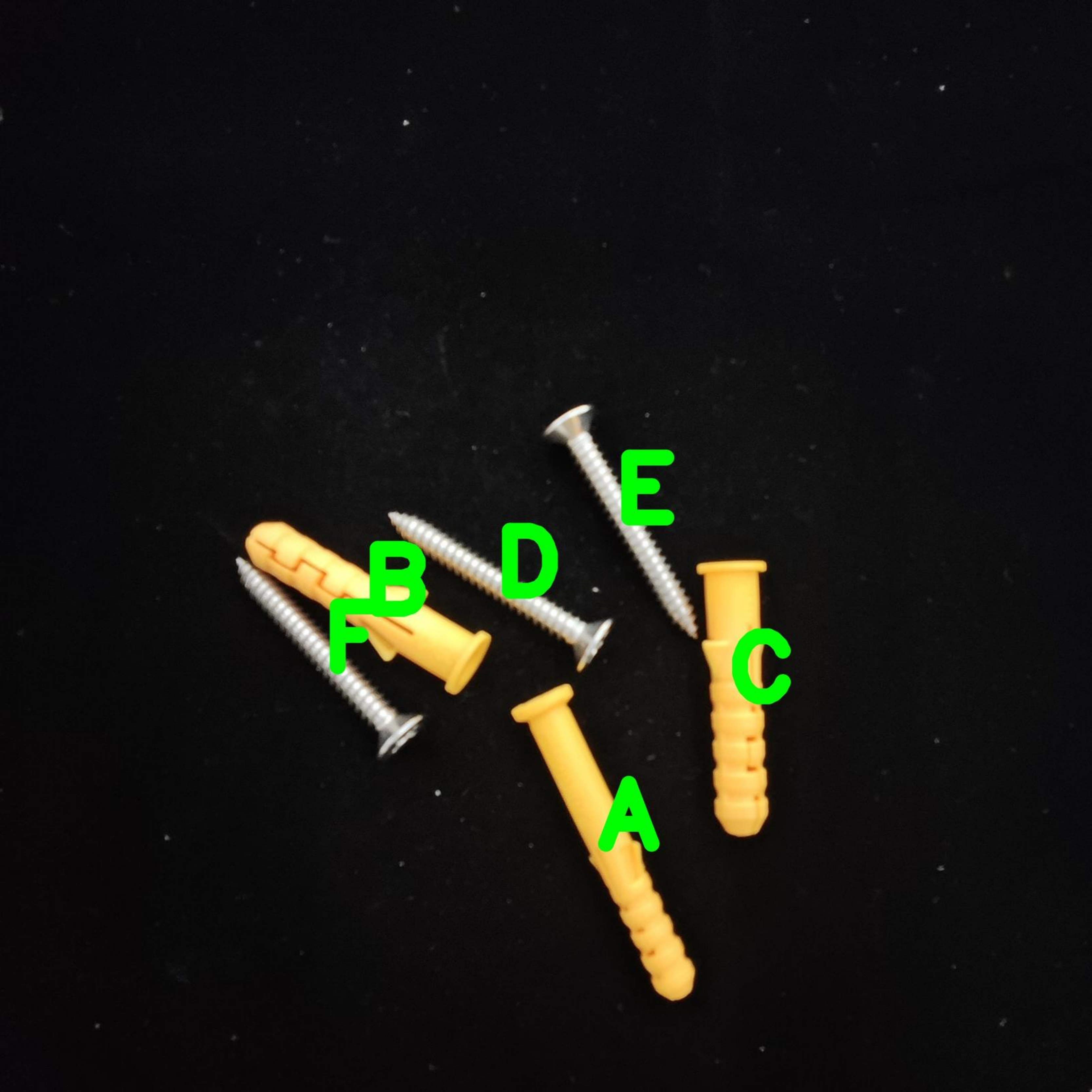}
    }
    \parbox[t][1.5cm][t]{\linewidth}{\small Plastic anchors belonging to expansion screw model 860 were mixed into the ba}
\end{minipage}\hfill
\begin{minipage}{0.23\textwidth}
    \centering
    \parbox[c][3.0cm][c]{\linewidth}{
        \centering\includegraphics[width=\linewidth,height=3.0cm,keepaspectratio]{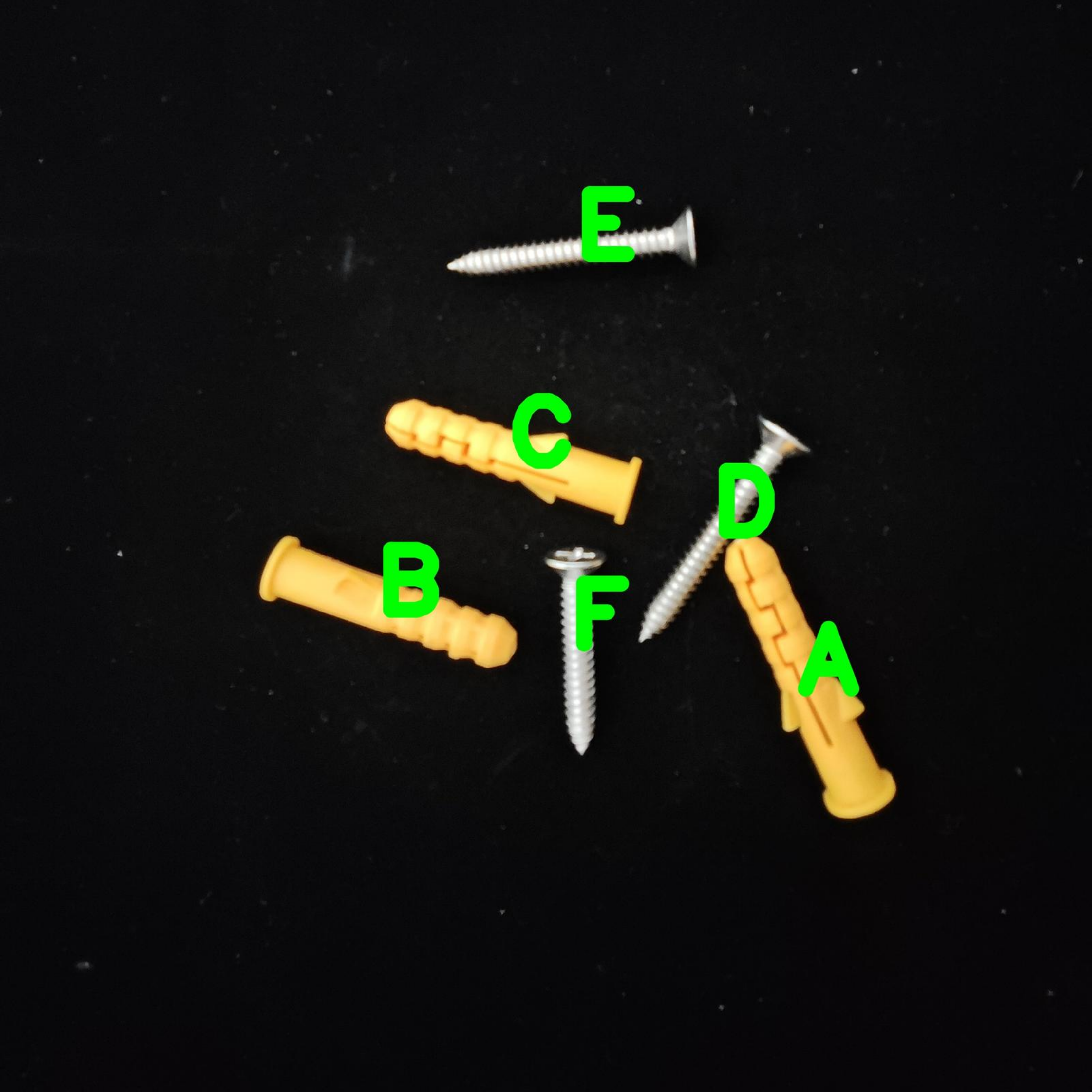}
    }
    \parbox[t][1.5cm][t]{\linewidth}{\small Screws shorter 30 than the specified length were found mixed in}
\end{minipage}\hfill
\begin{minipage}{0.23\textwidth}
    \centering
    \parbox[c][3.0cm][c]{\linewidth}{
        \centering\includegraphics[width=\linewidth,height=3.0cm,keepaspectratio]{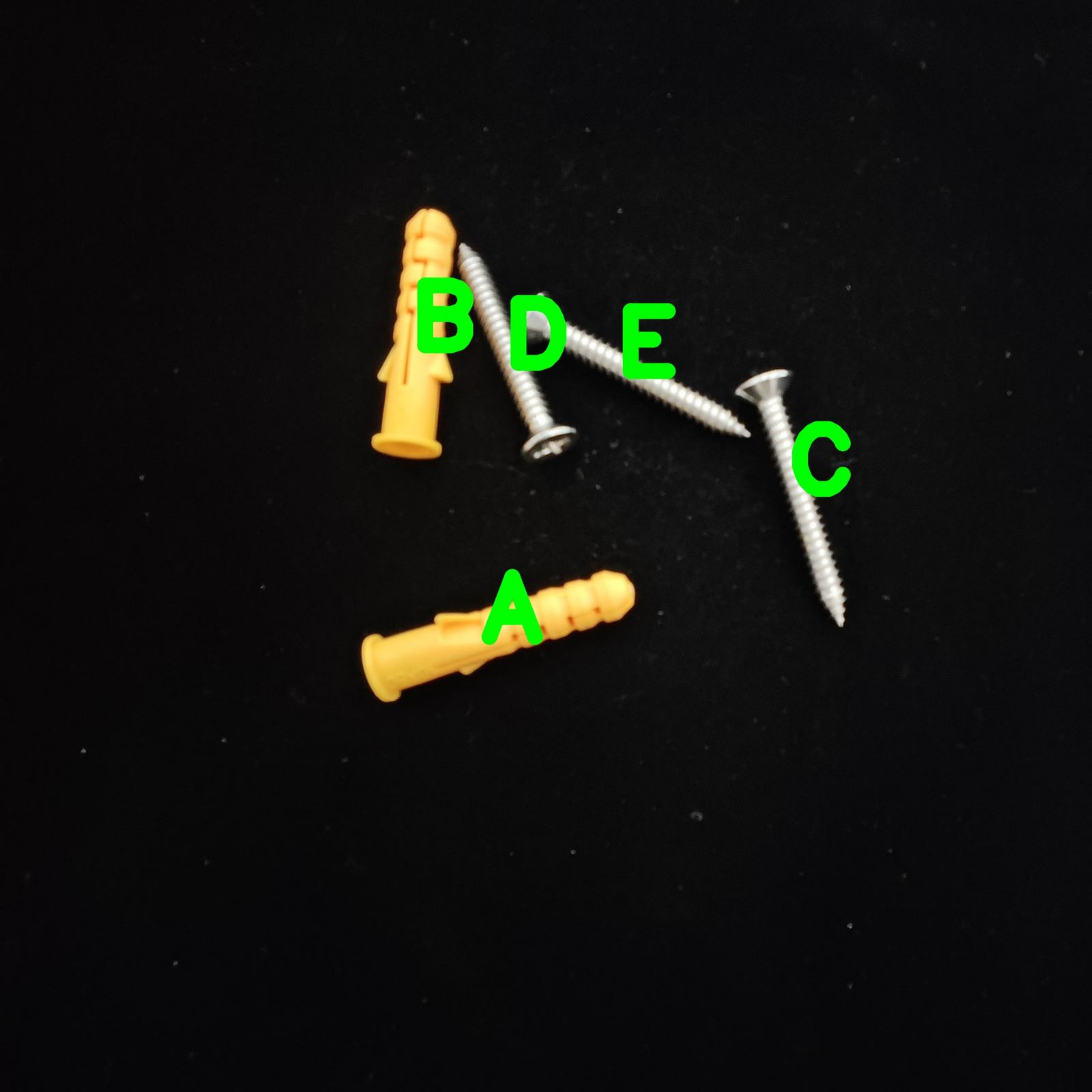}
    }
    \parbox[t][1.5cm][t]{\linewidth}{\small One missing plastic expansion anchor}
\end{minipage}\hfill
\begin{minipage}{0.23\textwidth}
    \centering
    \parbox[c][3.0cm][c]{\linewidth}{
        \centering\includegraphics[width=\linewidth,height=3.0cm,keepaspectratio]{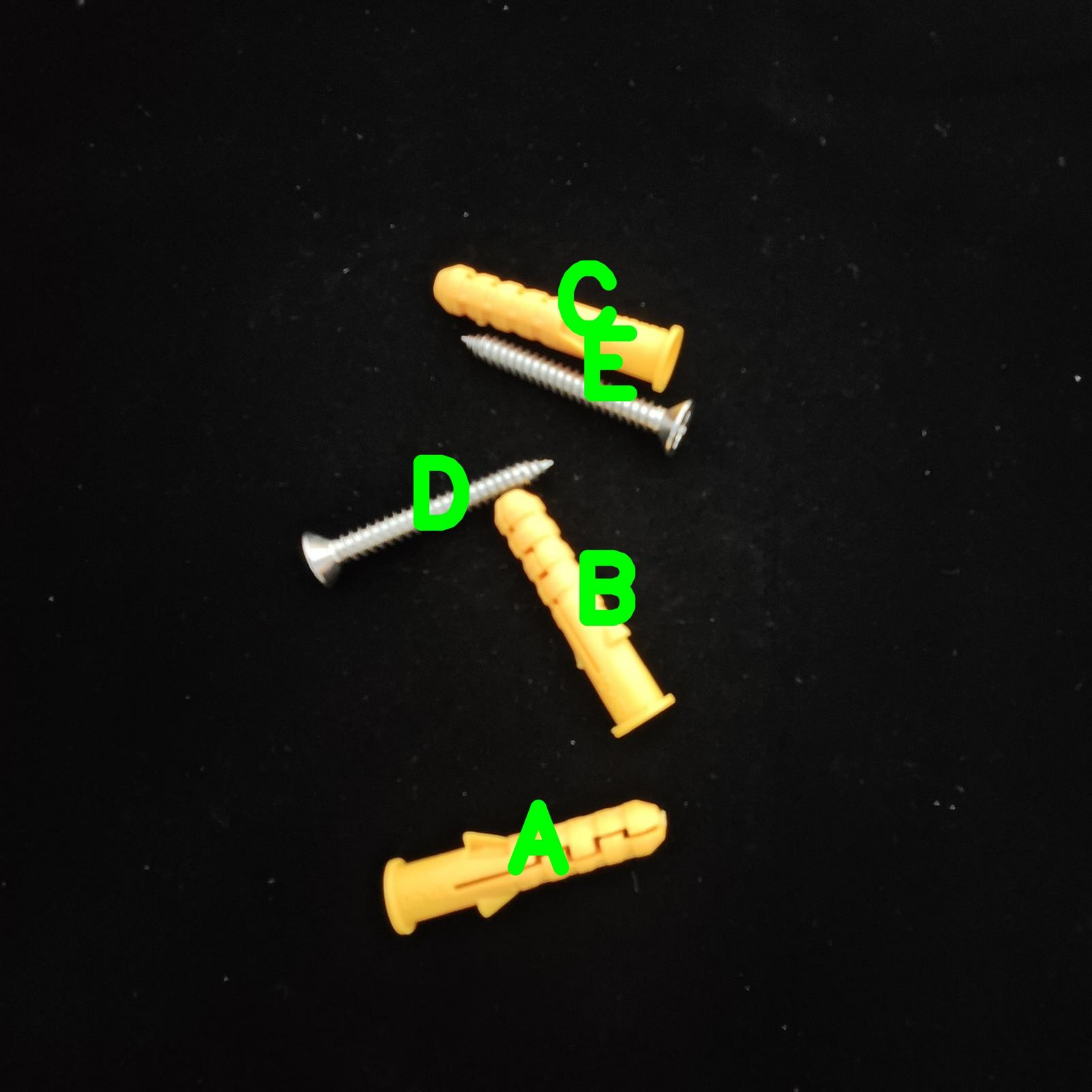}
    }
    \parbox[t][1.5cm][t]{\linewidth}{\small One missing plastic expansion Self-tapping Screw}
\end{minipage}

\caption{Random grounding examples from \pesscen}
\label{figab: Manufacturing_scenario_2_saved}
\end{figure}

\begin{figure}[htbp]
\centering

\begin{minipage}{0.23\textwidth}
    \centering
    \parbox[c][3.0cm][c]{\linewidth}{
        \centering\includegraphics[width=\linewidth,height=3.0cm,keepaspectratio]{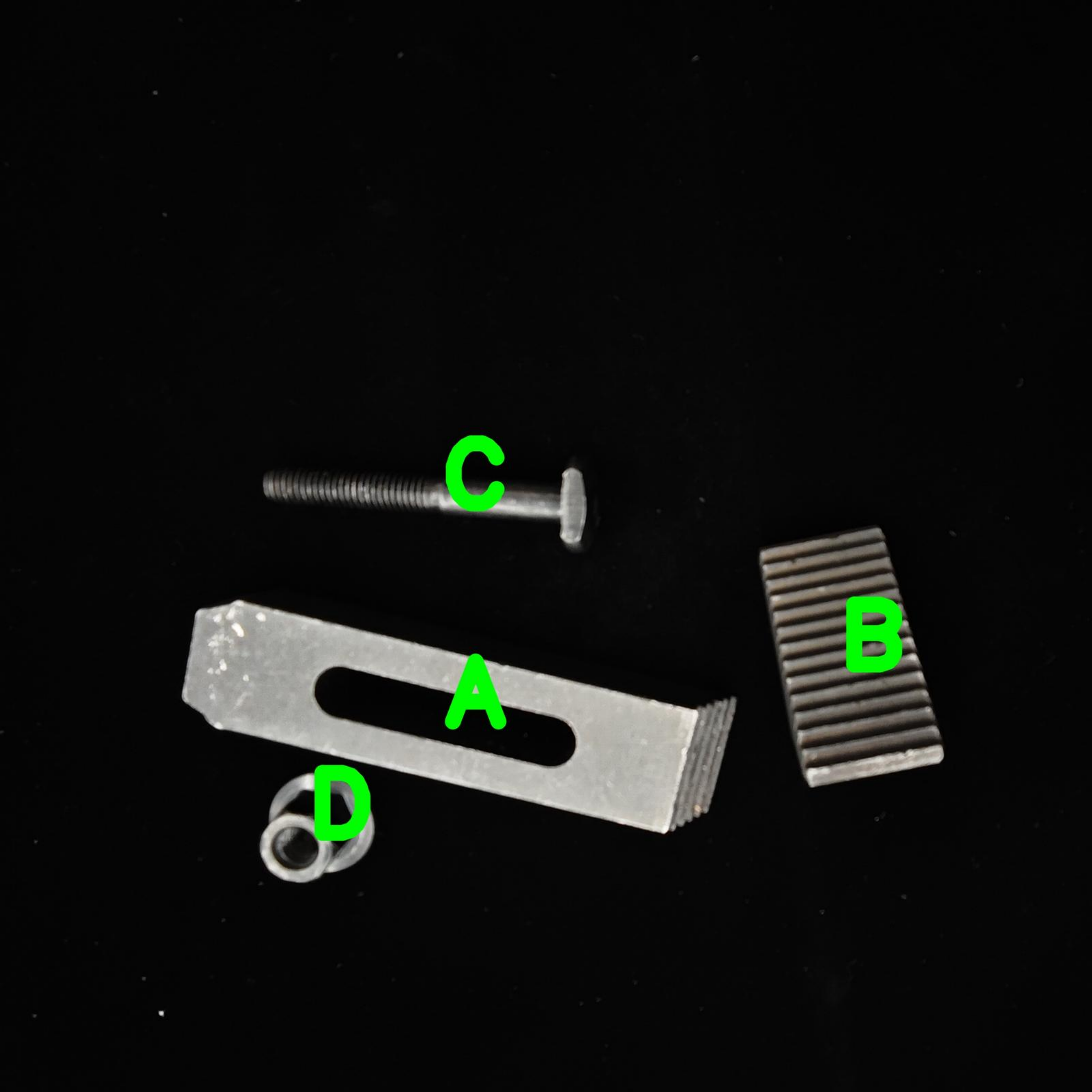}
    }
    \parbox[t][1.5cm][t]{\linewidth}{\small Normal Case }
\end{minipage}\hfill
\begin{minipage}{0.23\textwidth}
    \centering
    \parbox[c][3.0cm][c]{\linewidth}{
        \centering\includegraphics[width=\linewidth,height=3.0cm,keepaspectratio]{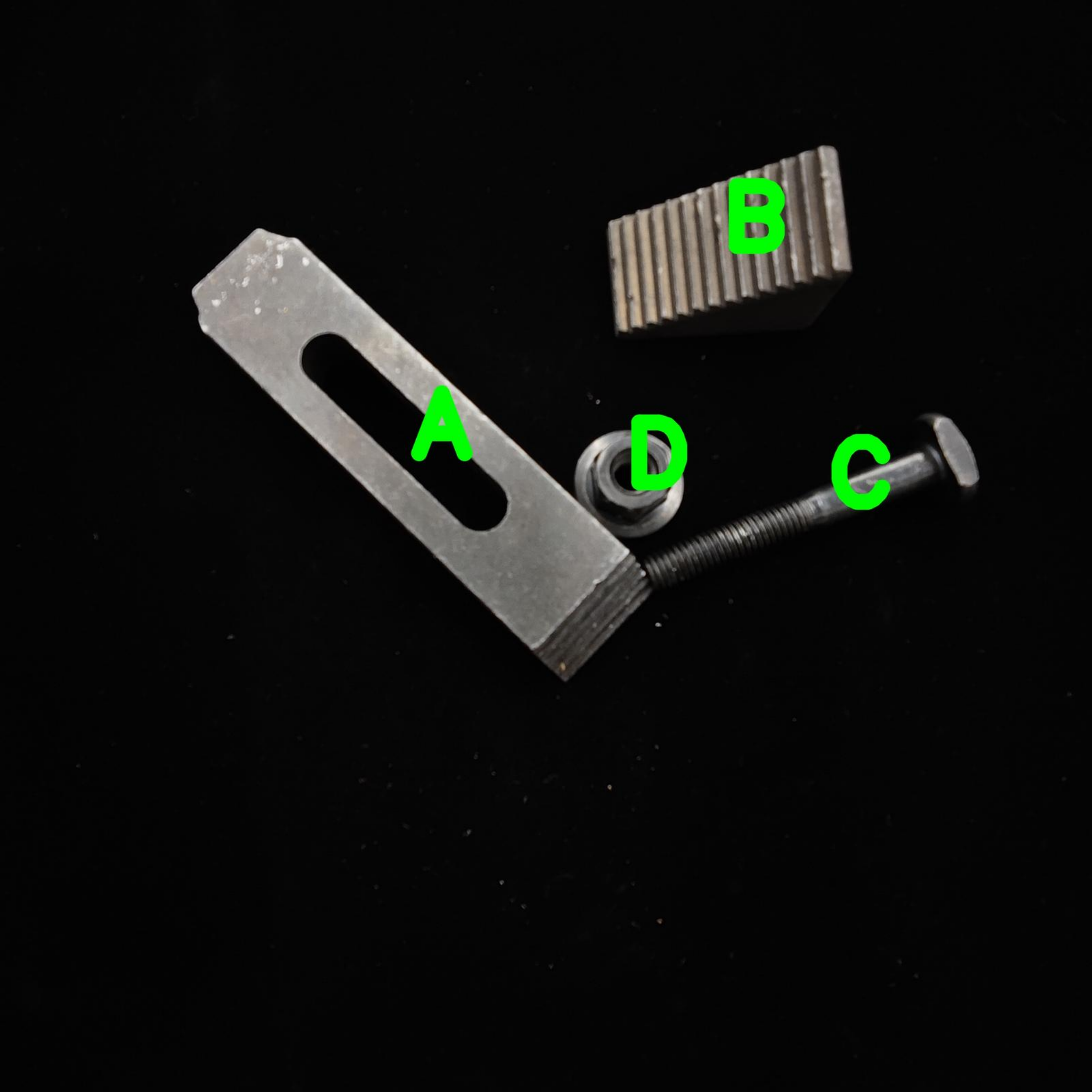}
    }
    \parbox[t][1.5cm][t]{\linewidth}{\small Normal Case }
\end{minipage}\hfill
\begin{minipage}{0.23\textwidth}
    \centering
    \parbox[c][3.0cm][c]{\linewidth}{
        \centering\includegraphics[width=\linewidth,height=3.0cm,keepaspectratio]{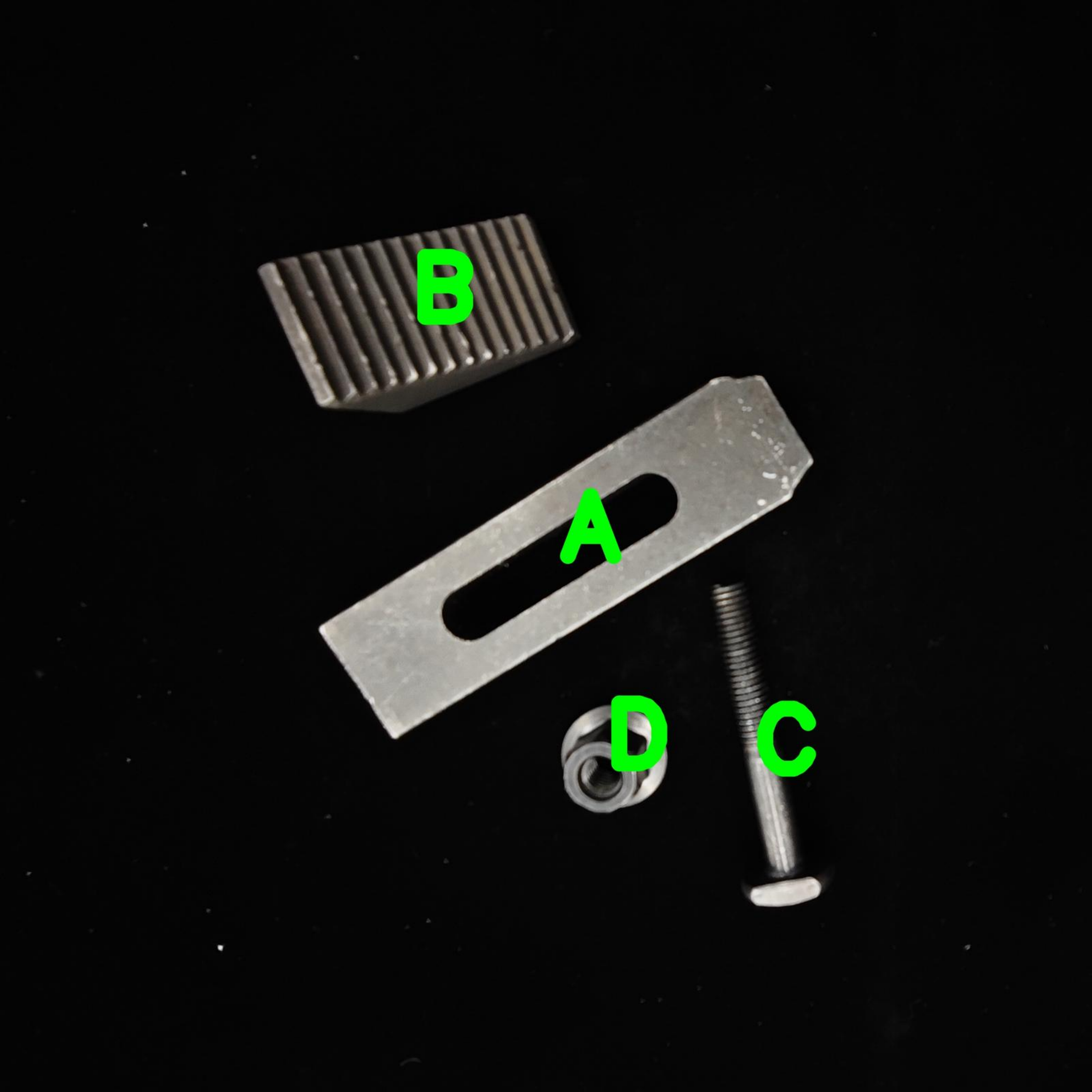}
    }
    \parbox[t][1.5cm][t]{\linewidth}{\small Normal Case }
\end{minipage}\hfill
\begin{minipage}{0.23\textwidth}
    \centering
    \parbox[c][3.0cm][c]{\linewidth}{
        \centering\includegraphics[width=\linewidth,height=3.0cm,keepaspectratio]{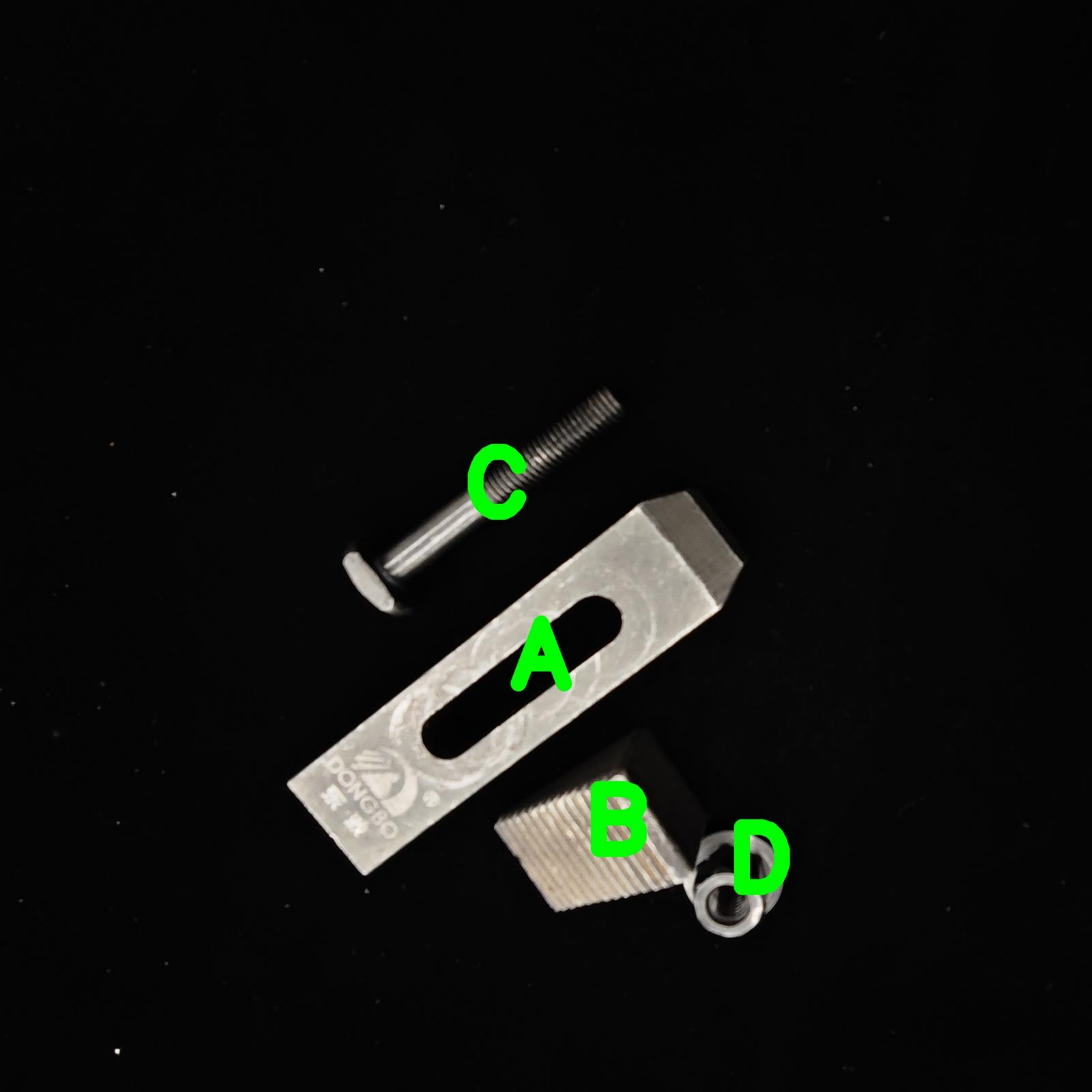}
    }
    \parbox[t][1.5cm][t]{\linewidth}{\small Normal Case }
\end{minipage}

\begin{minipage}{0.23\textwidth}
    \centering
    \parbox[c][3.0cm][c]{\linewidth}{
        \centering\includegraphics[width=\linewidth,height=3.0cm,keepaspectratio]{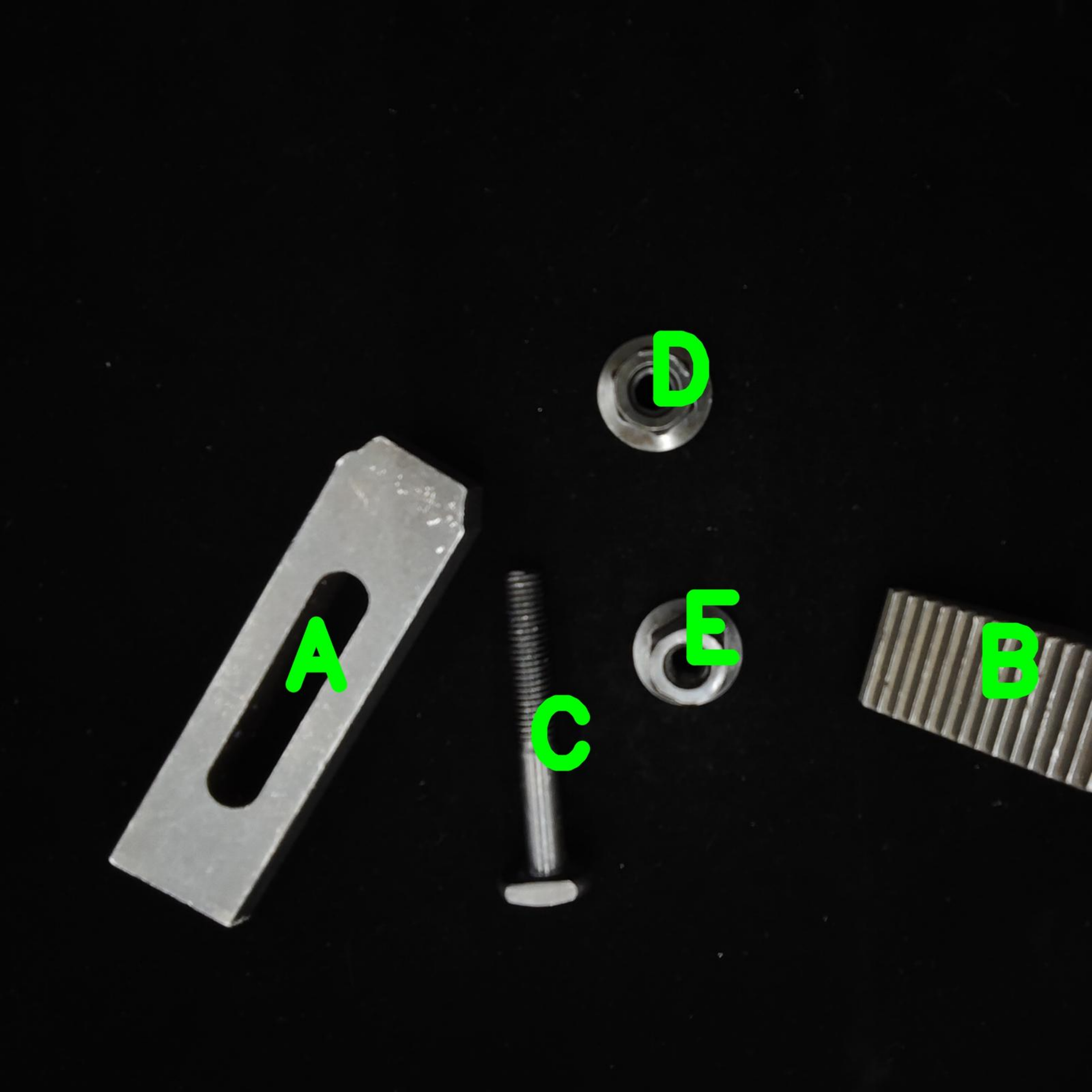}
    }
    \parbox[t][1.5cm][t]{\linewidth}{\small Extra one nut}
\end{minipage}\hfill
\begin{minipage}{0.23\textwidth}
    \centering
    \parbox[c][3.0cm][c]{\linewidth}{
        \centering\includegraphics[width=\linewidth,height=3.0cm,keepaspectratio]{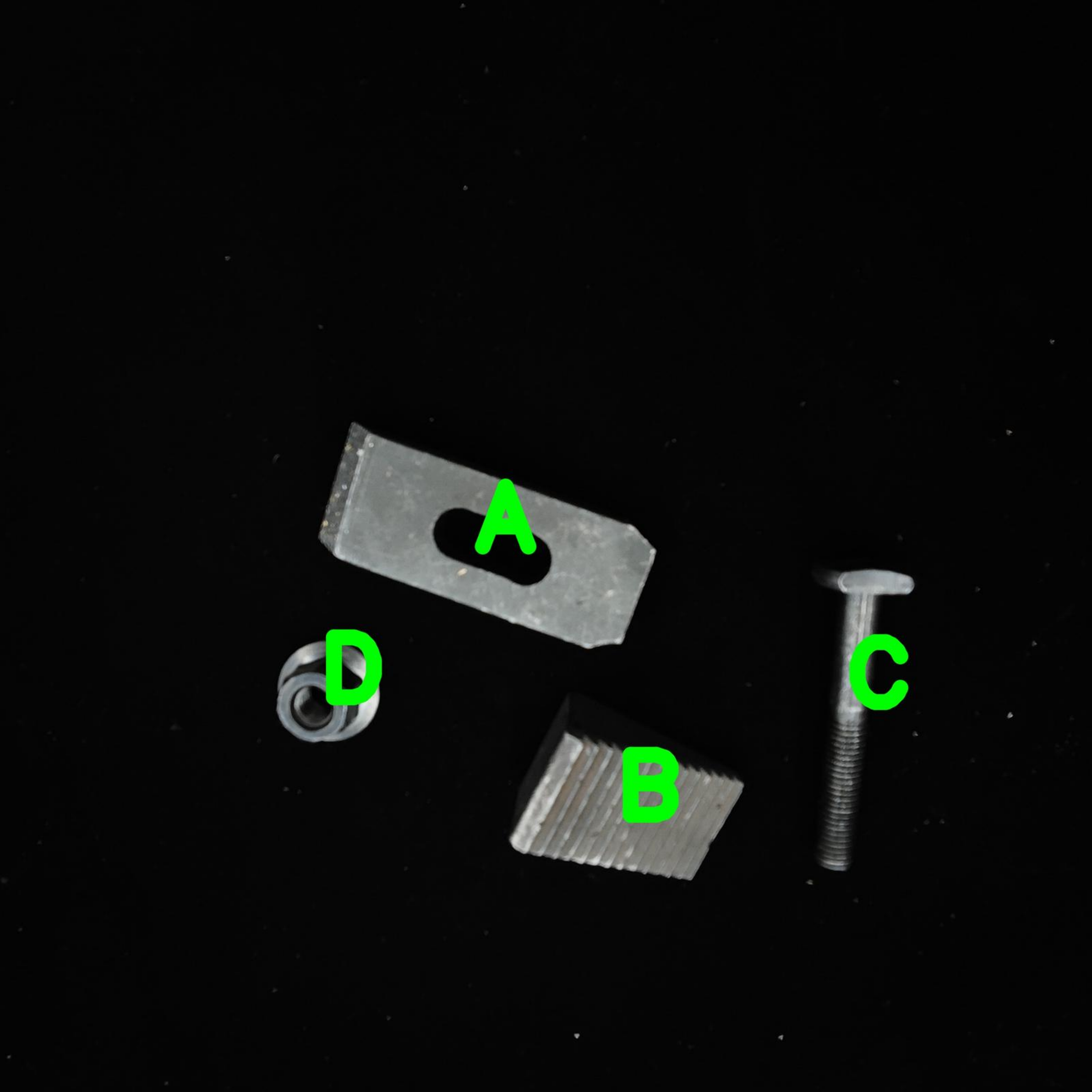}
    }
    \parbox[t][1.5cm][t]{\linewidth}{\small retainer block}
\end{minipage}\hfill
\begin{minipage}{0.23\textwidth}
    \centering
    \parbox[c][3.0cm][c]{\linewidth}{
        \centering\includegraphics[width=\linewidth,height=3.0cm,keepaspectratio]{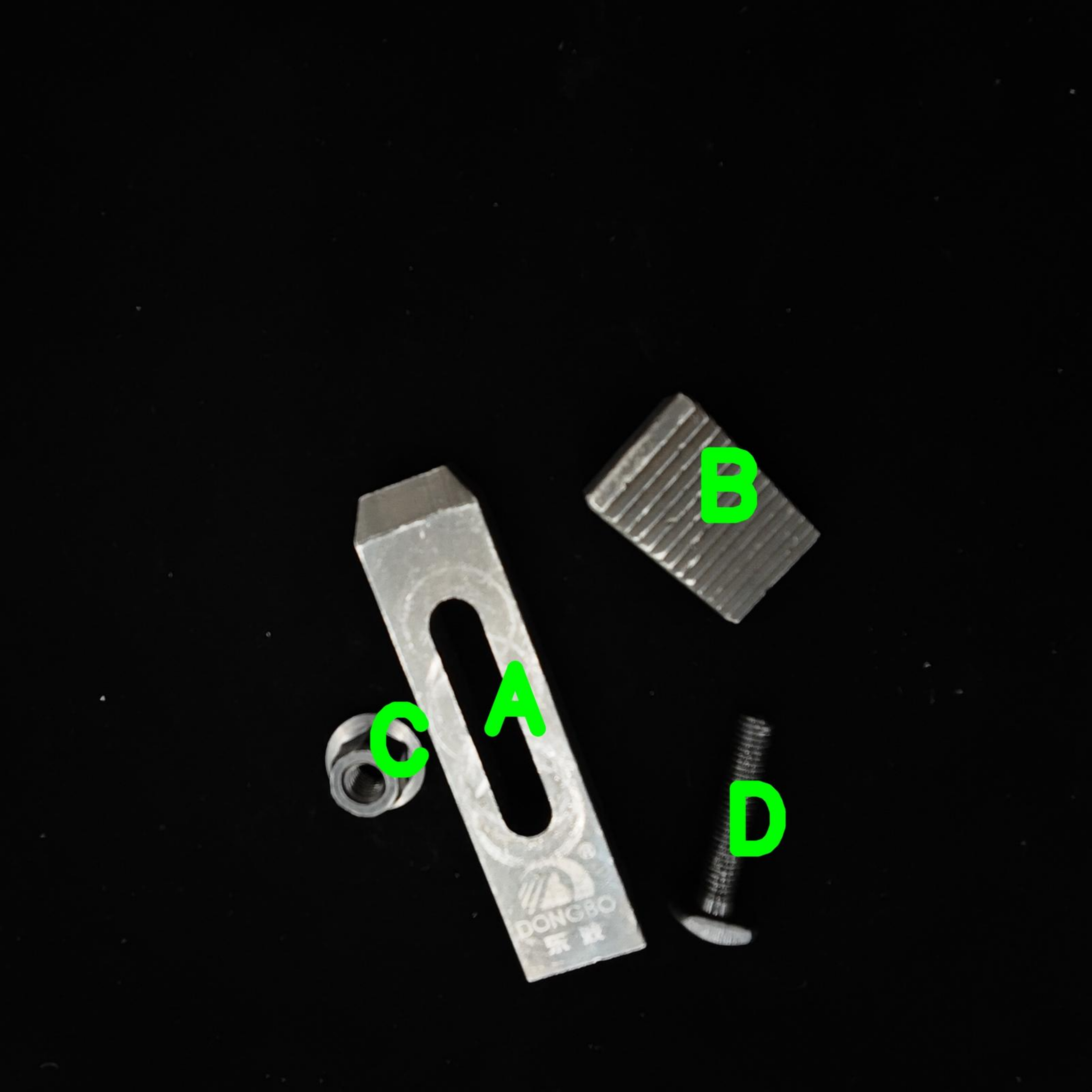}
    }
    \parbox[t][1.5cm][t]{\linewidth}{\small Short screw}
\end{minipage}\hfill
\begin{minipage}{0.23\textwidth}
    \centering
    \parbox[c][3.0cm][c]{\linewidth}{
        \centering\includegraphics[width=\linewidth,height=3.0cm,keepaspectratio]{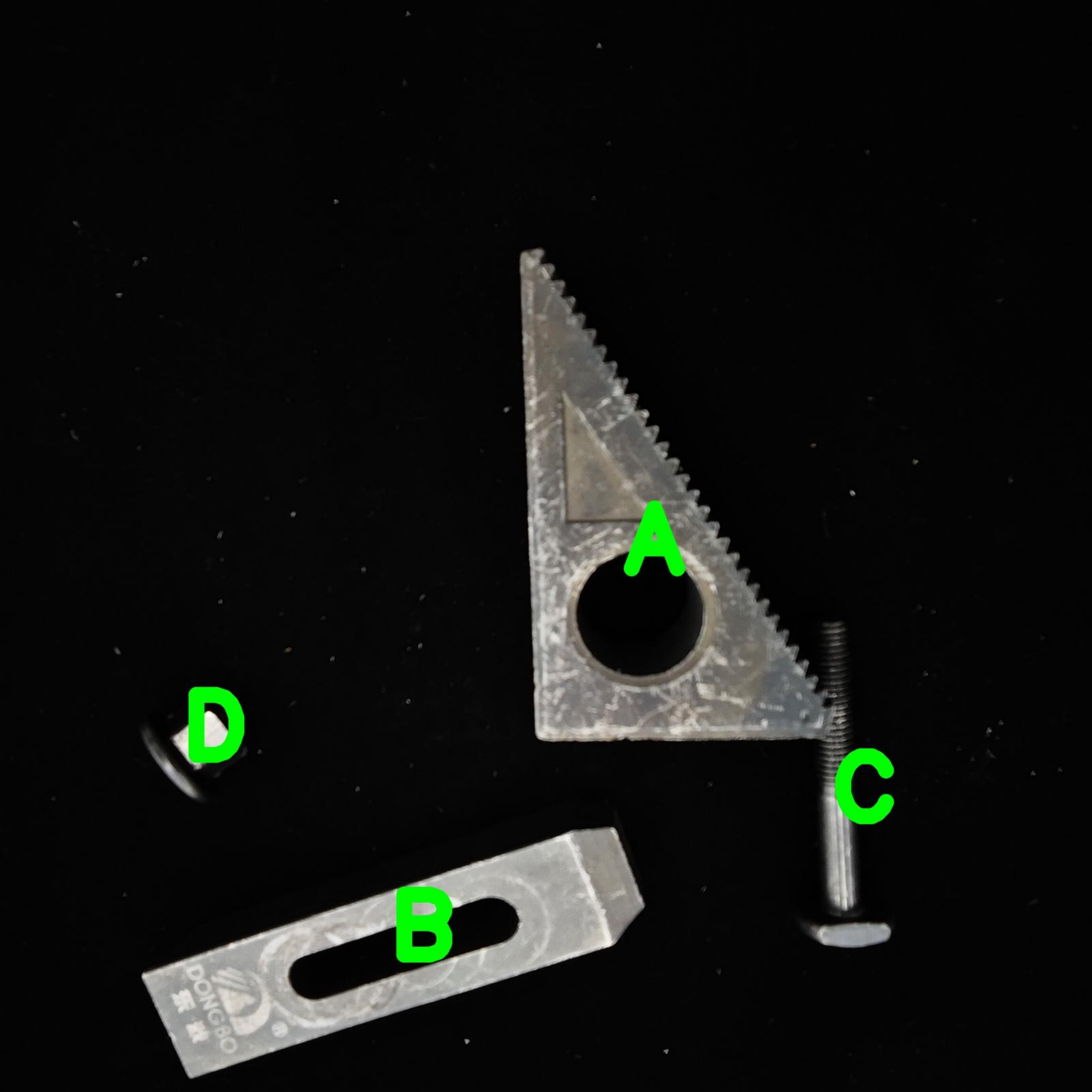}
    }
    \parbox[t][1.5cm][t]{\linewidth}{\small Triangular part}
\end{minipage}

\begin{minipage}{0.23\textwidth}
    \centering
    \parbox[c][3.0cm][c]{\linewidth}{
        \centering\includegraphics[width=\linewidth,height=3.0cm,keepaspectratio]{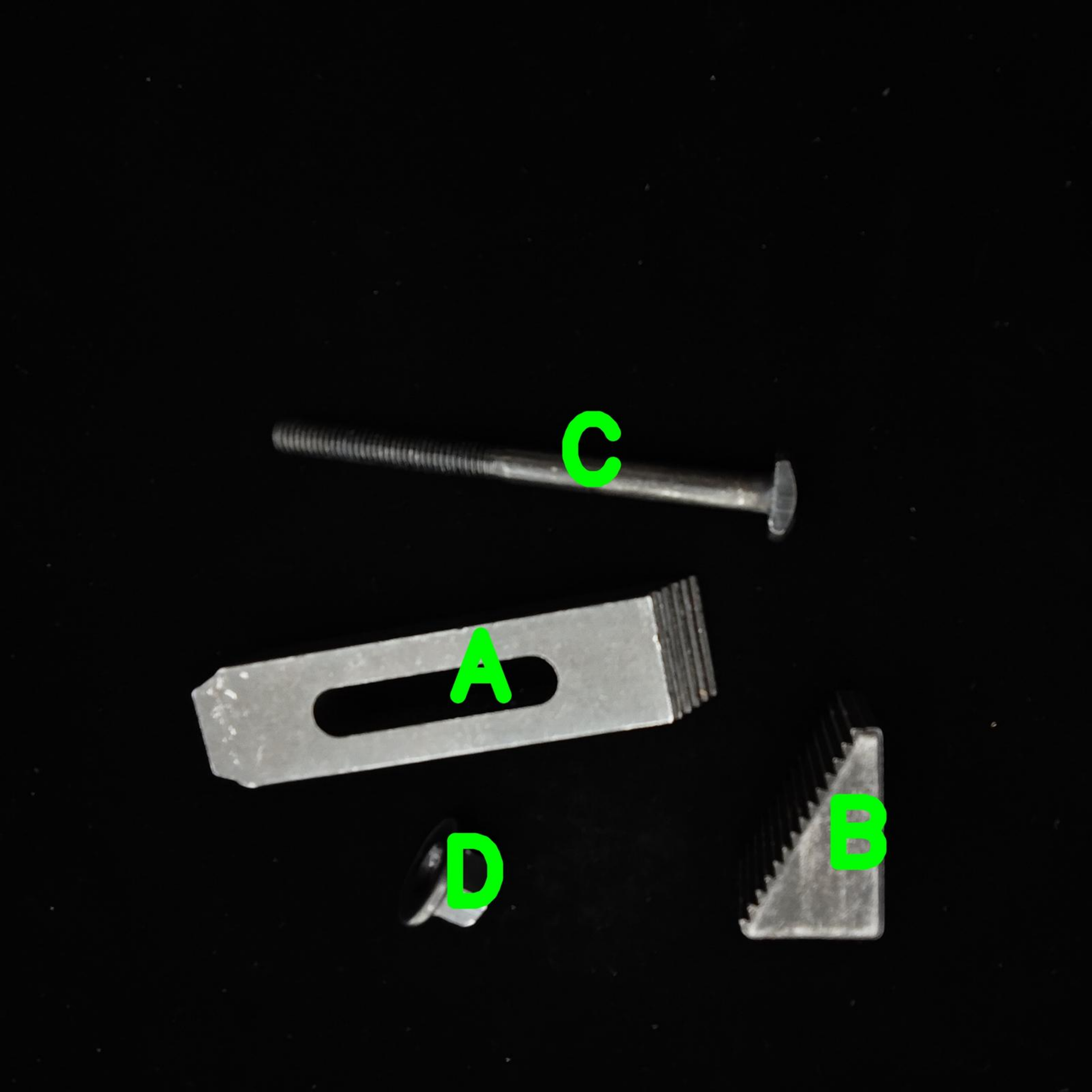}
    }
    \parbox[t][1.5cm][t]{\linewidth}{\small Long screw}
\end{minipage}\hfill
\begin{minipage}{0.23\textwidth}
    \centering
    \parbox[c][3.0cm][c]{\linewidth}{
        \centering\includegraphics[width=\linewidth,height=3.0cm,keepaspectratio]{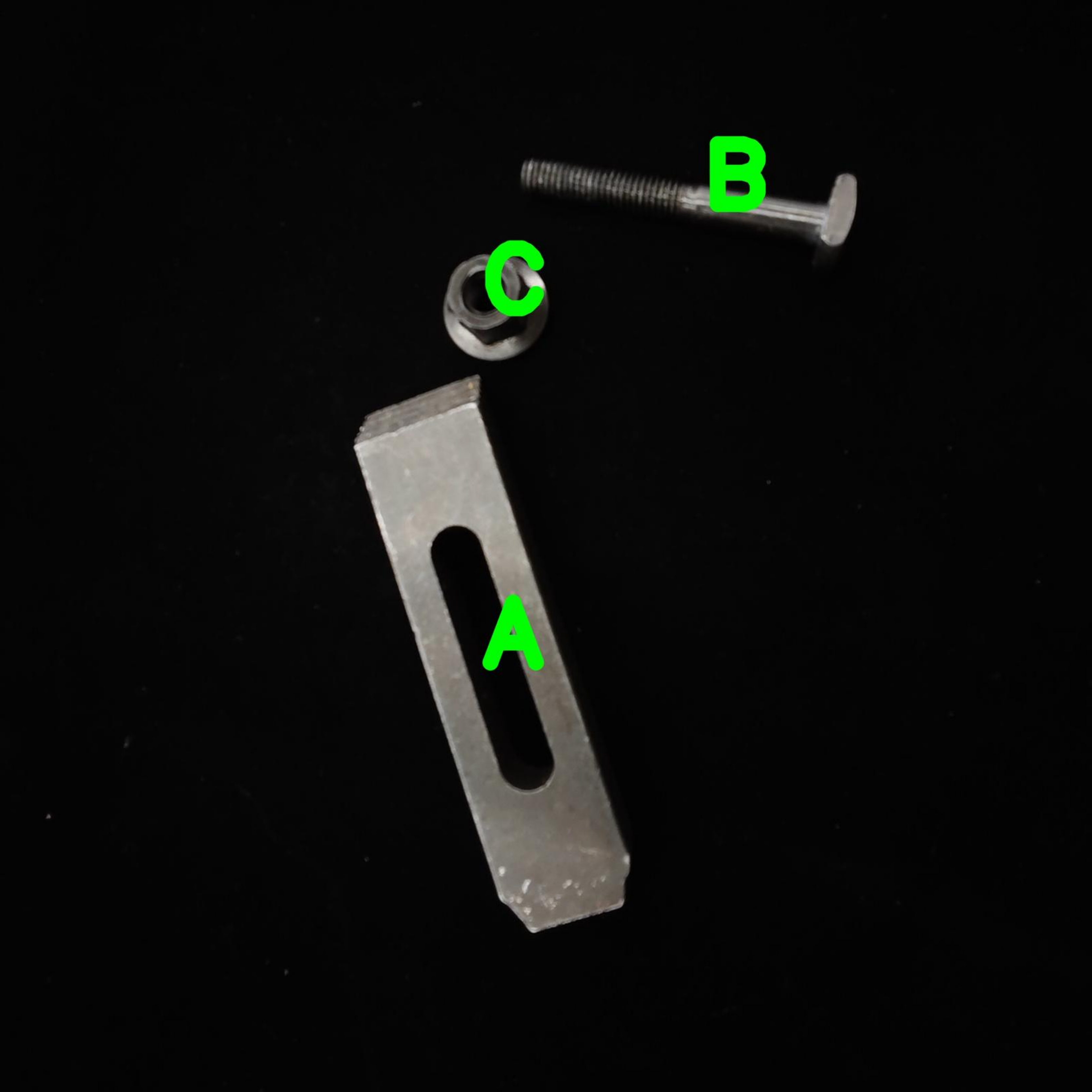}
    }
    \parbox[t][1.5cm][t]{\linewidth}{\small Missing triangular part}
\end{minipage}\hfill
\begin{minipage}{0.23\textwidth}
    \centering
    \parbox[c][3.0cm][c]{\linewidth}{
        \centering\includegraphics[width=\linewidth,height=3.0cm,keepaspectratio]{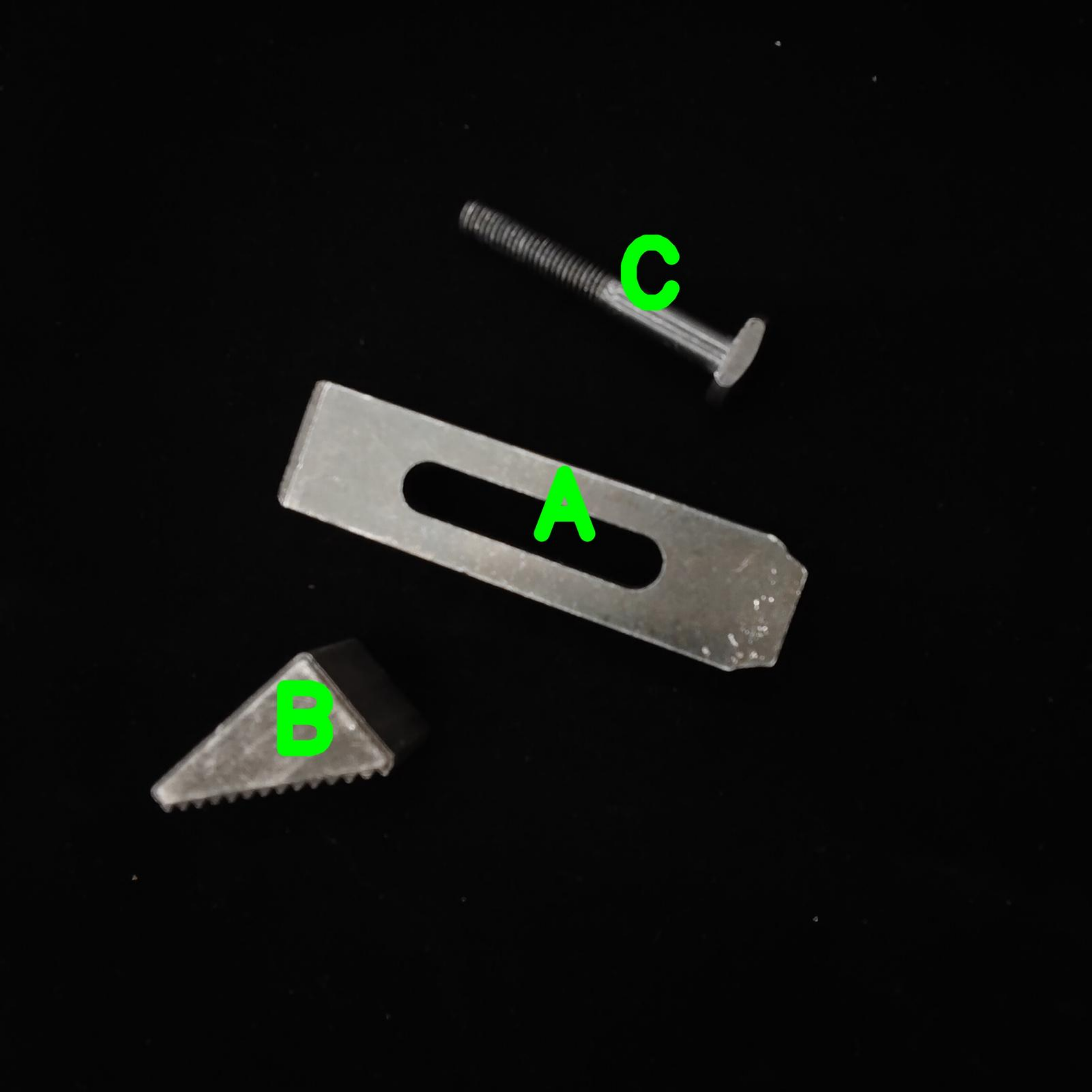}
    }
    \parbox[t][1.5cm][t]{\linewidth}{\small Missing one nut}
\end{minipage}\hfill
\begin{minipage}{0.23\textwidth}
    \centering
    \parbox[c][3.0cm][c]{\linewidth}{
        \centering\includegraphics[width=\linewidth,height=3.0cm,keepaspectratio]{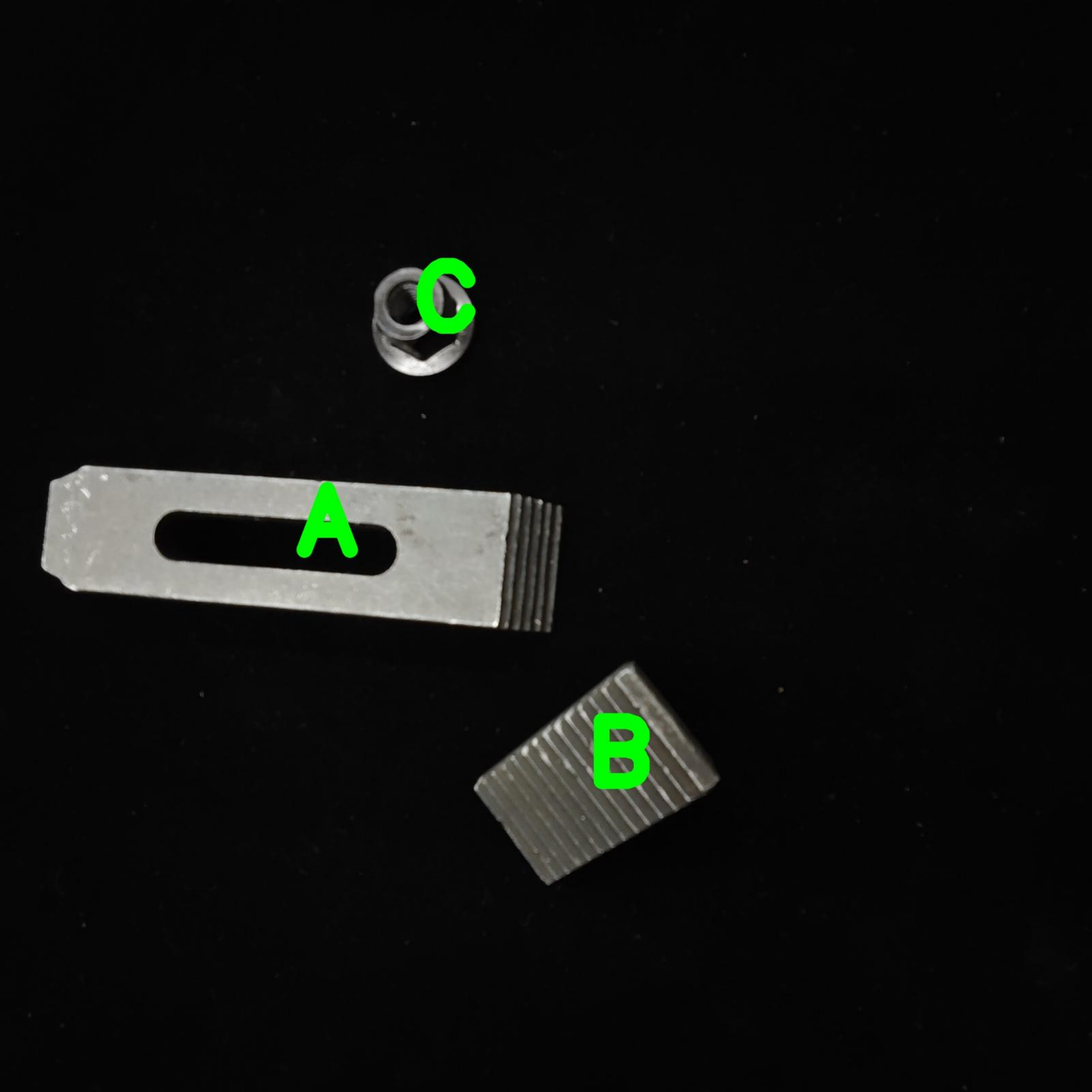}
    }
    \parbox[t][1.5cm][t]{\linewidth}{\small Missing screw}
\end{minipage}

\caption{Random grounding examples from \cncscen}
\label{figab: Manufacturing_scenario_4_saved}
\end{figure}

\begin{figure}[htbp]
\centering

\begin{minipage}{0.23\textwidth}
    \centering
    \parbox[c][3.0cm][c]{\linewidth}{
        \centering\includegraphics[width=\linewidth,height=3.0cm,keepaspectratio]{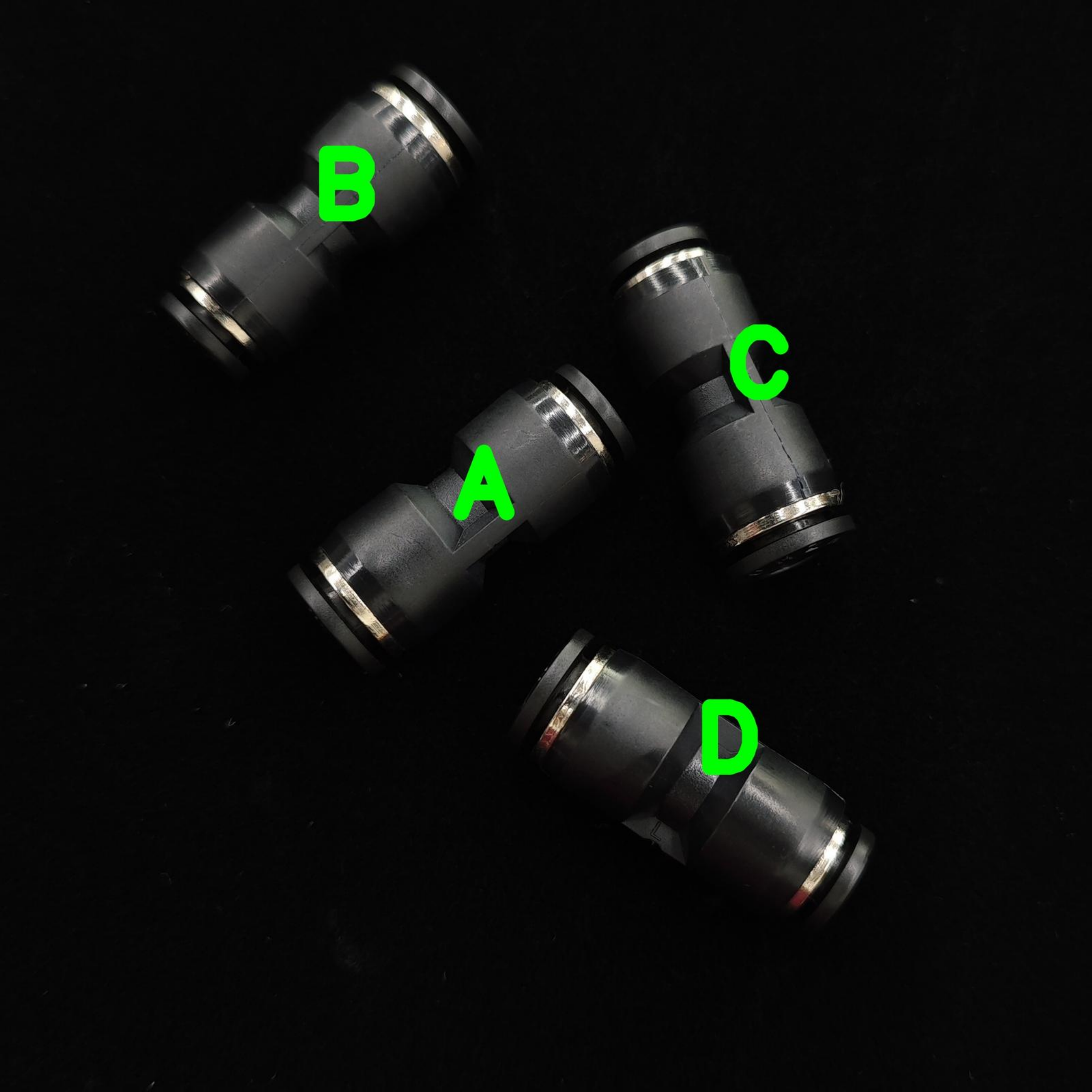}
    }
    \parbox[t][1.5cm][t]{\linewidth}{\small Normal Case}
\end{minipage}\hfill
\begin{minipage}{0.23\textwidth}
    \centering
    \parbox[c][3.0cm][c]{\linewidth}{
        \centering\includegraphics[width=\linewidth,height=3.0cm,keepaspectratio]{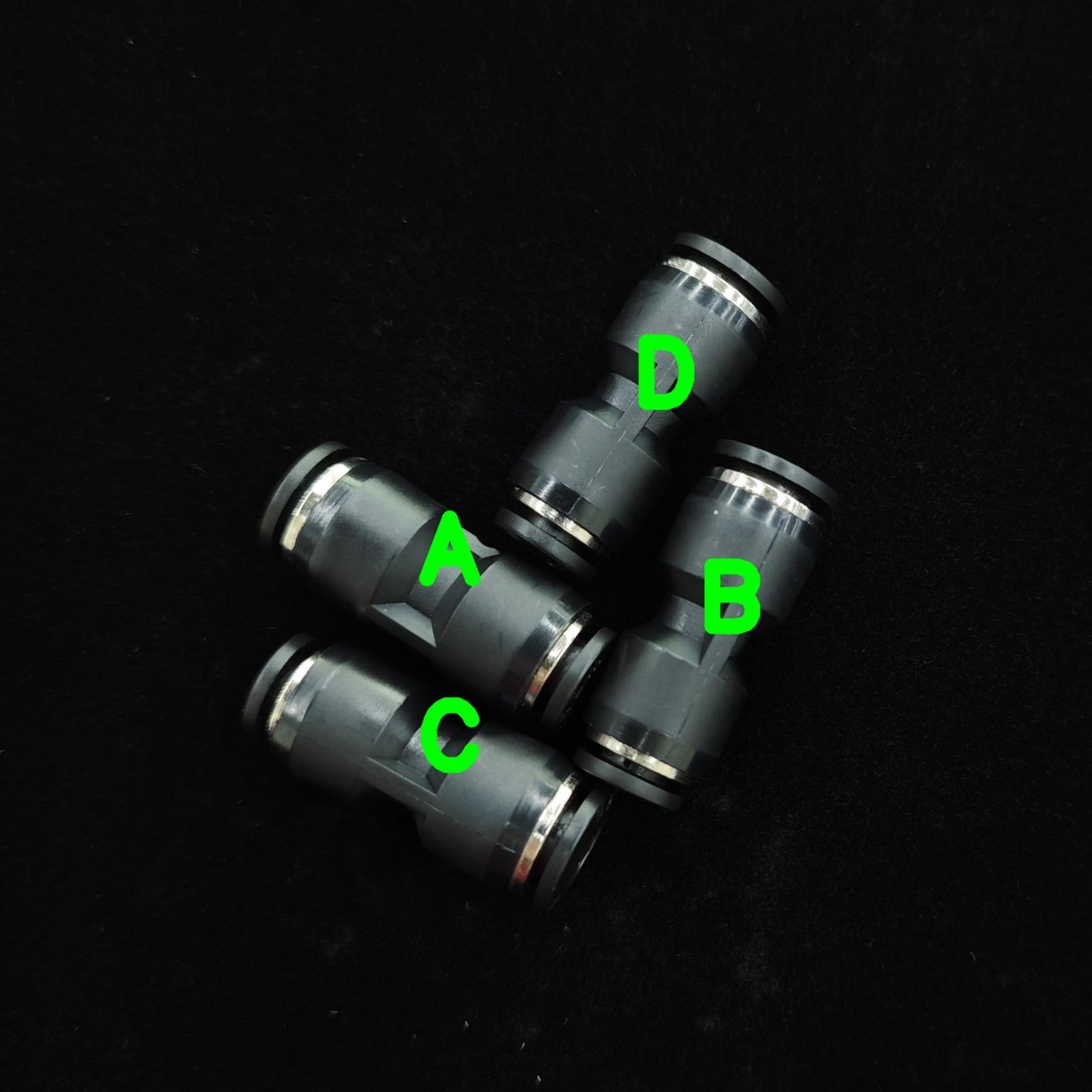}
    }
    \parbox[t][1.5cm][t]{\linewidth}{\small Normal Case}
\end{minipage}\hfill
\begin{minipage}{0.23\textwidth}
    \centering
    \parbox[c][3.0cm][c]{\linewidth}{
        \centering\includegraphics[width=\linewidth,height=3.0cm,keepaspectratio]{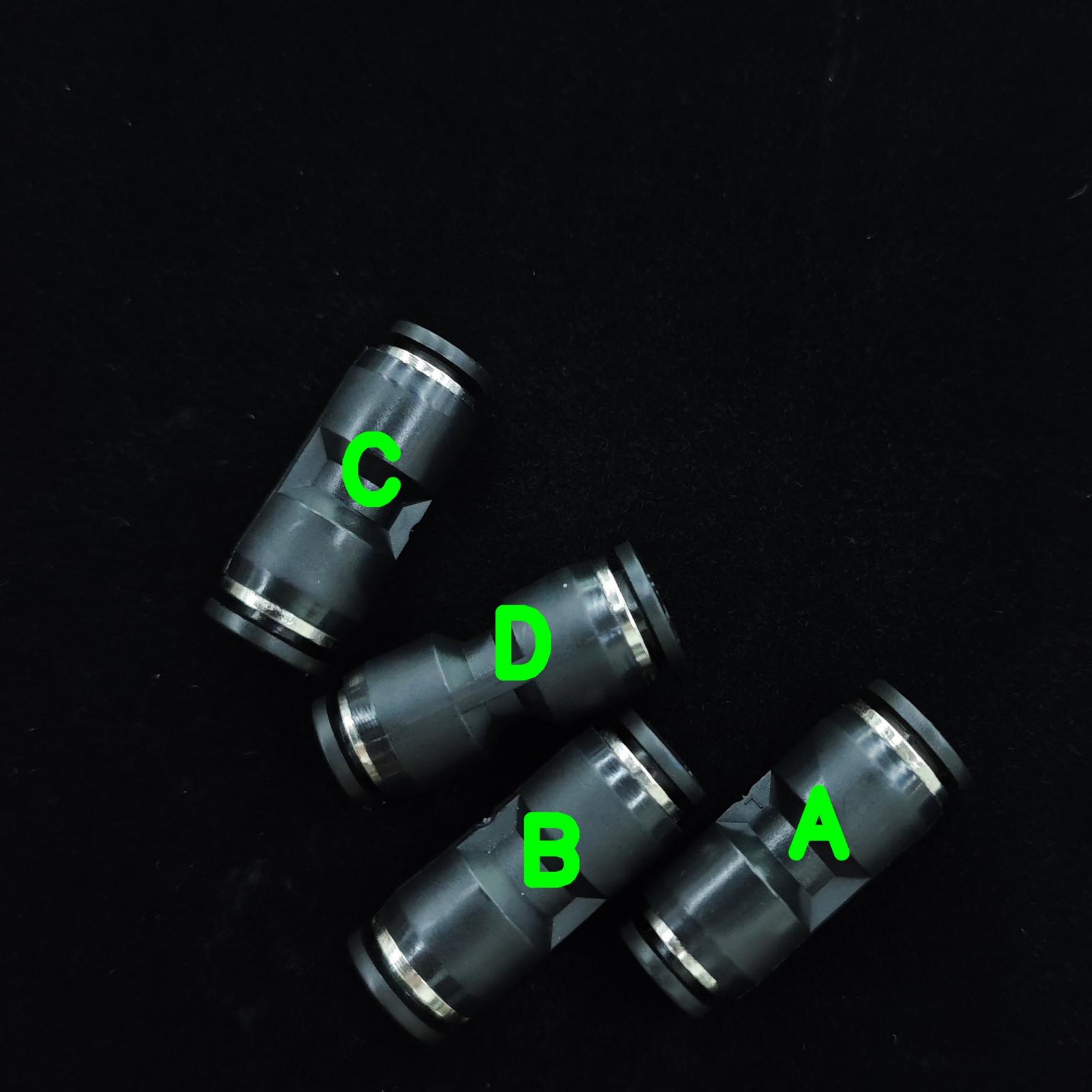}
    }
    \parbox[t][1.5cm][t]{\linewidth}{\small Normal Case}
\end{minipage}\hfill
\begin{minipage}{0.23\textwidth}
    \centering
    \parbox[c][3.0cm][c]{\linewidth}{
        \centering\includegraphics[width=\linewidth,height=3.0cm,keepaspectratio]{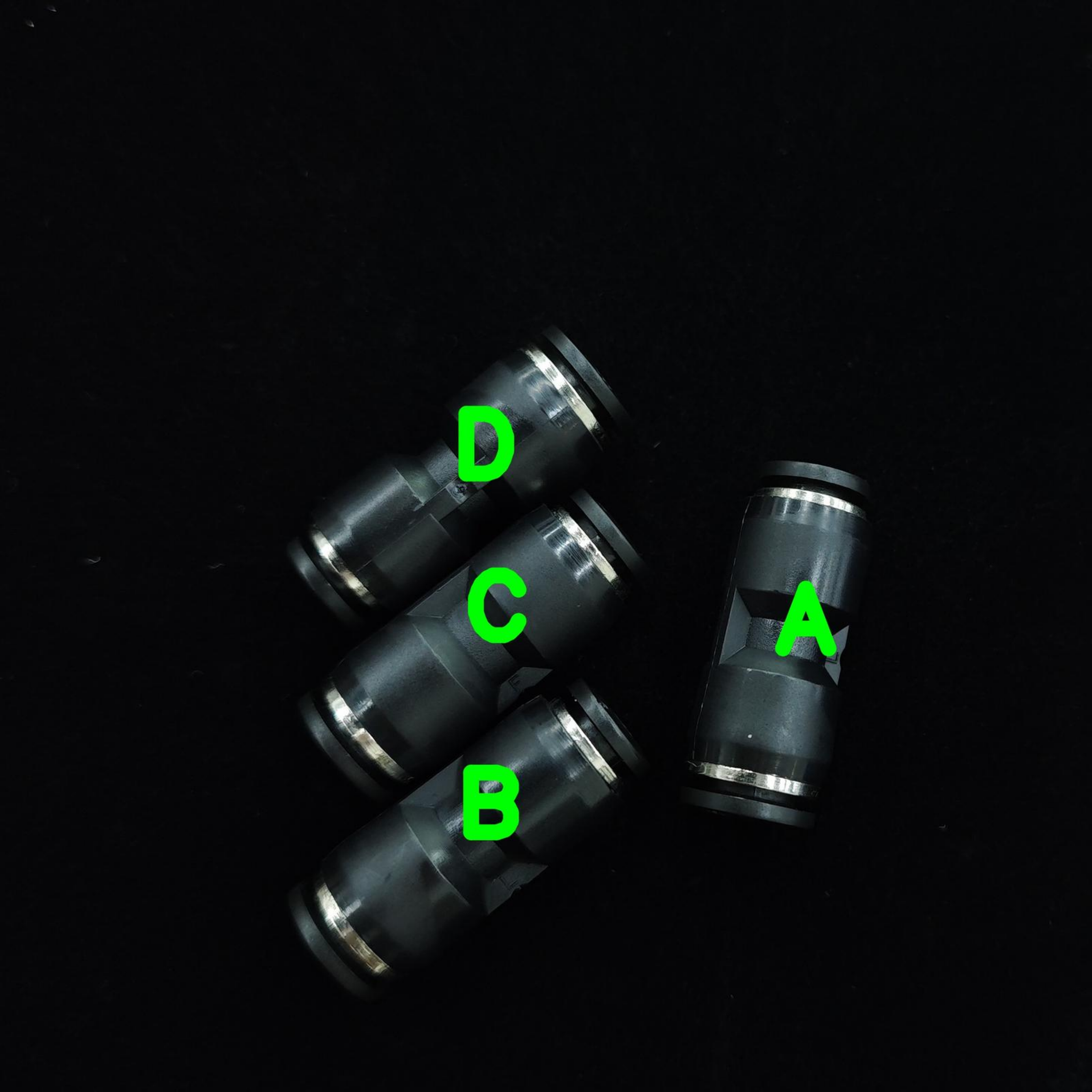}
    }
    \parbox[t][1.5cm][t]{\linewidth}{\small Normal Case}
\end{minipage}

\begin{minipage}{0.23\textwidth}
    \centering
    \parbox[c][3.0cm][c]{\linewidth}{
        \centering\includegraphics[width=\linewidth,height=3.0cm,keepaspectratio]{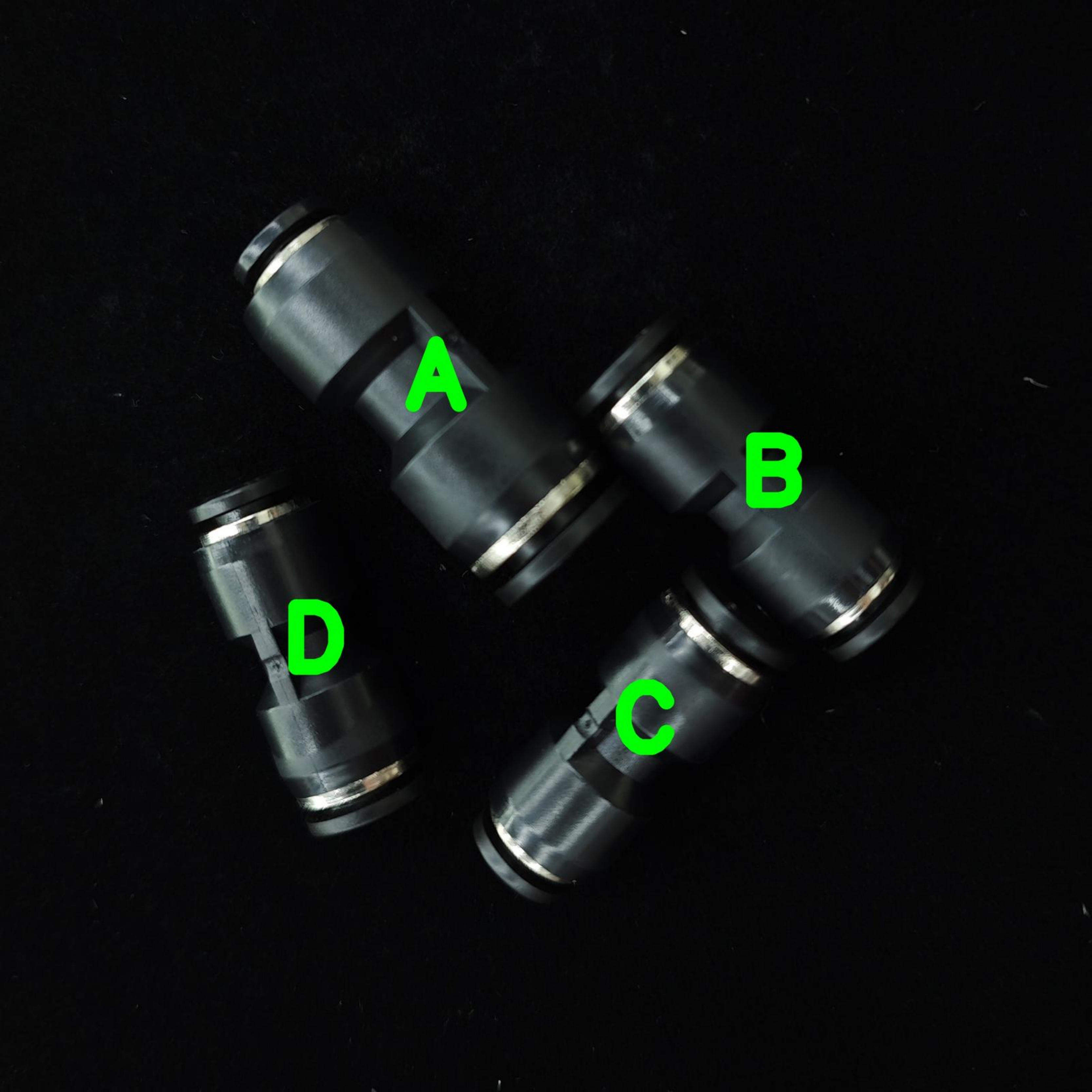}
    }
    \parbox[t][1.5cm][t]{\linewidth}{\small 2-way pneumatic tube connectors (8 to 6), mixed with model (10 to 6)}
\end{minipage}\hfill
\begin{minipage}{0.23\textwidth}
    \centering
    \parbox[c][3.0cm][c]{\linewidth}{
        \centering\includegraphics[width=\linewidth,height=3.0cm,keepaspectratio]{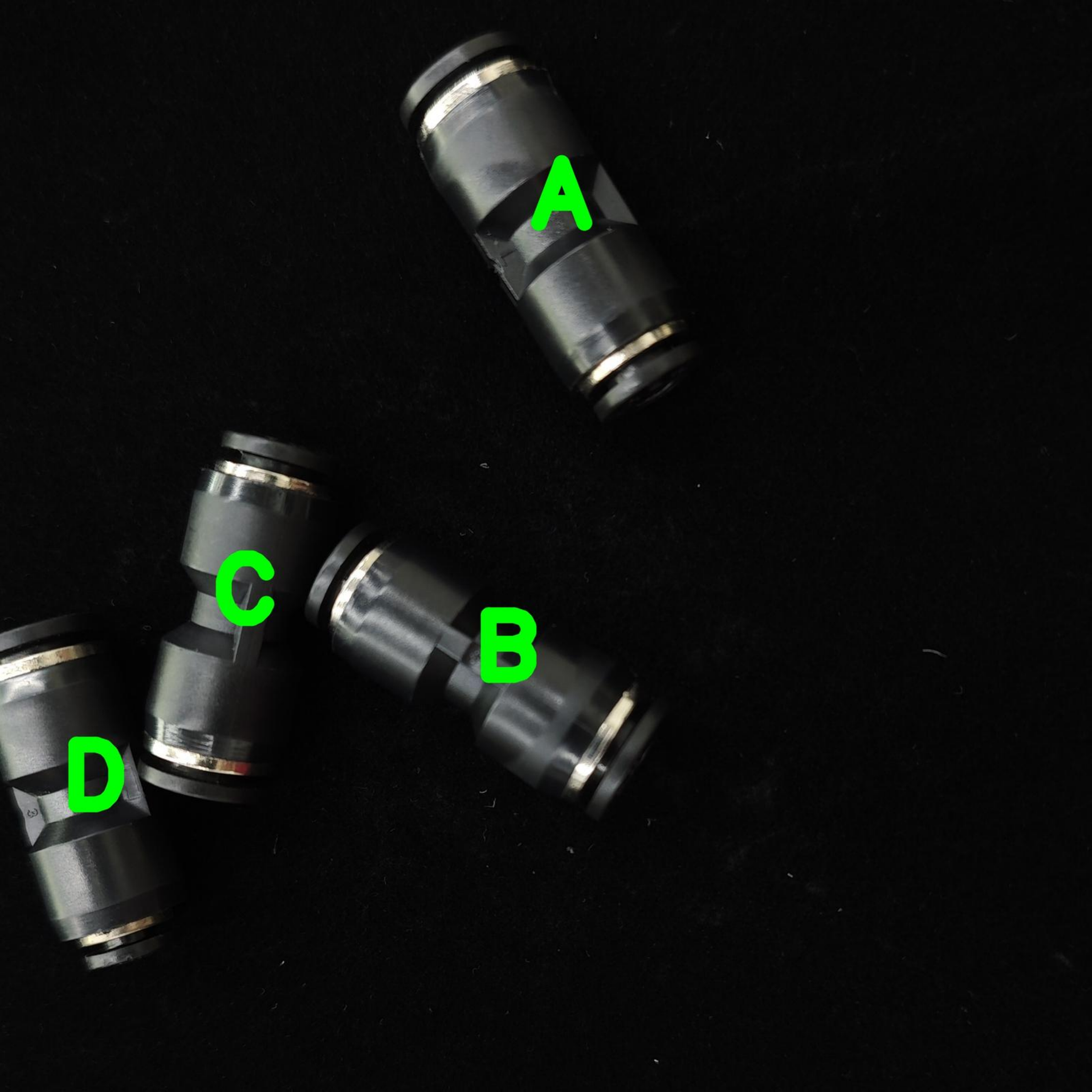}
    }
    \parbox[t][1.5cm][t]{\linewidth}{\small 2-way pneumatic tube connectors (8 to 6), mixed with model (8 to 4)}
\end{minipage}\hfill
\begin{minipage}{0.23\textwidth}
    \centering
    \parbox[c][3.0cm][c]{\linewidth}{
        \centering\includegraphics[width=\linewidth,height=3.0cm,keepaspectratio]{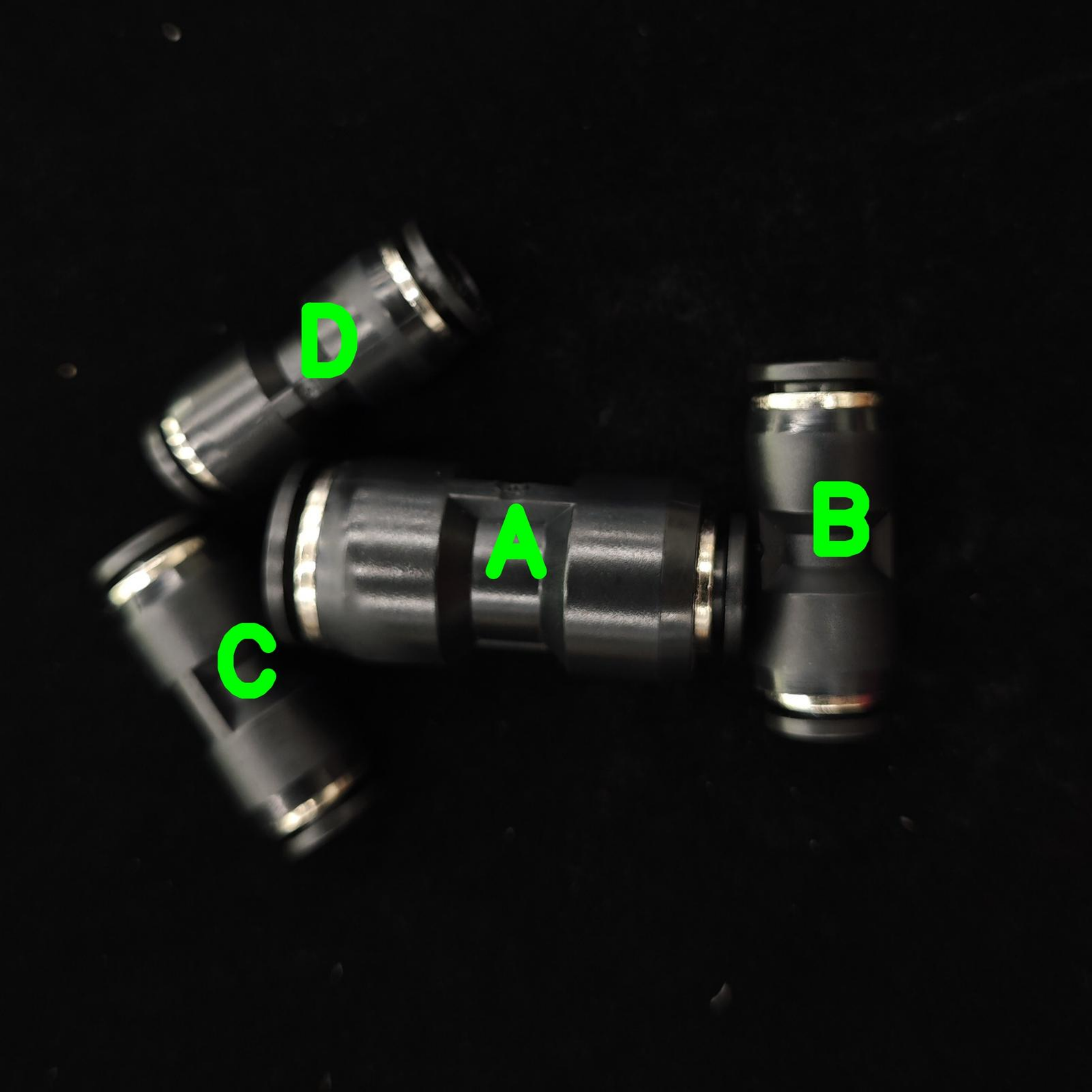}
    }
    \parbox[t][1.5cm][t]{\linewidth}{\small 2-way pneumatic tube connectors (8 to 6), mixed with model (12 to 8)}
\end{minipage}\hfill
\begin{minipage}{0.23\textwidth}
    \centering
    \parbox[c][3.0cm][c]{\linewidth}{
        \centering\includegraphics[width=\linewidth,height=3.0cm,keepaspectratio]{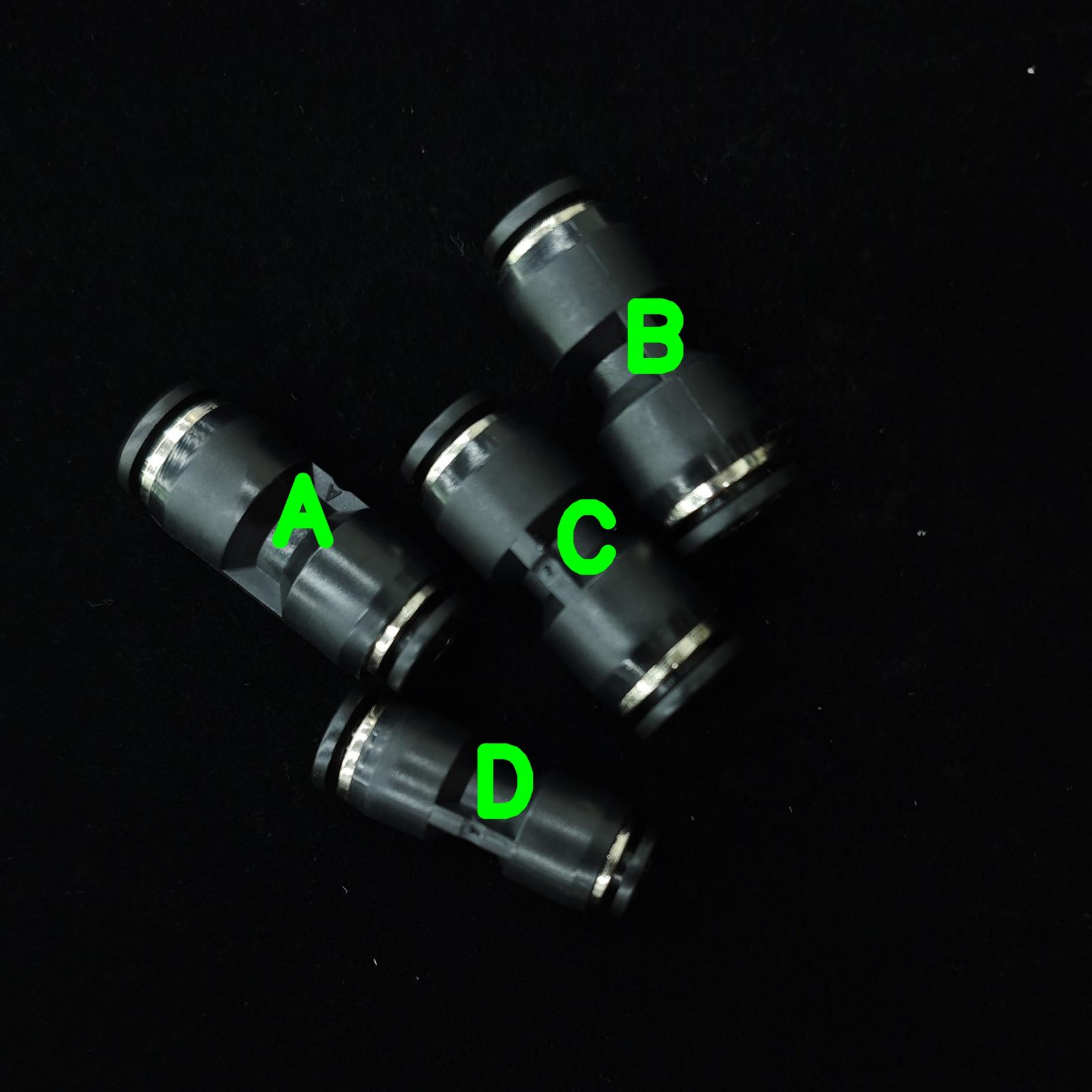}
    }
    \parbox[t][1.5cm][t]{\linewidth}{\small 2-way pneumatic tube connectors (8 to 6), mixed with model (6 to 4)}
\end{minipage}

\begin{minipage}{0.23\textwidth}
    \centering
    \parbox[c][3.0cm][c]{\linewidth}{
        \centering\includegraphics[width=\linewidth,height=3.0cm,keepaspectratio]{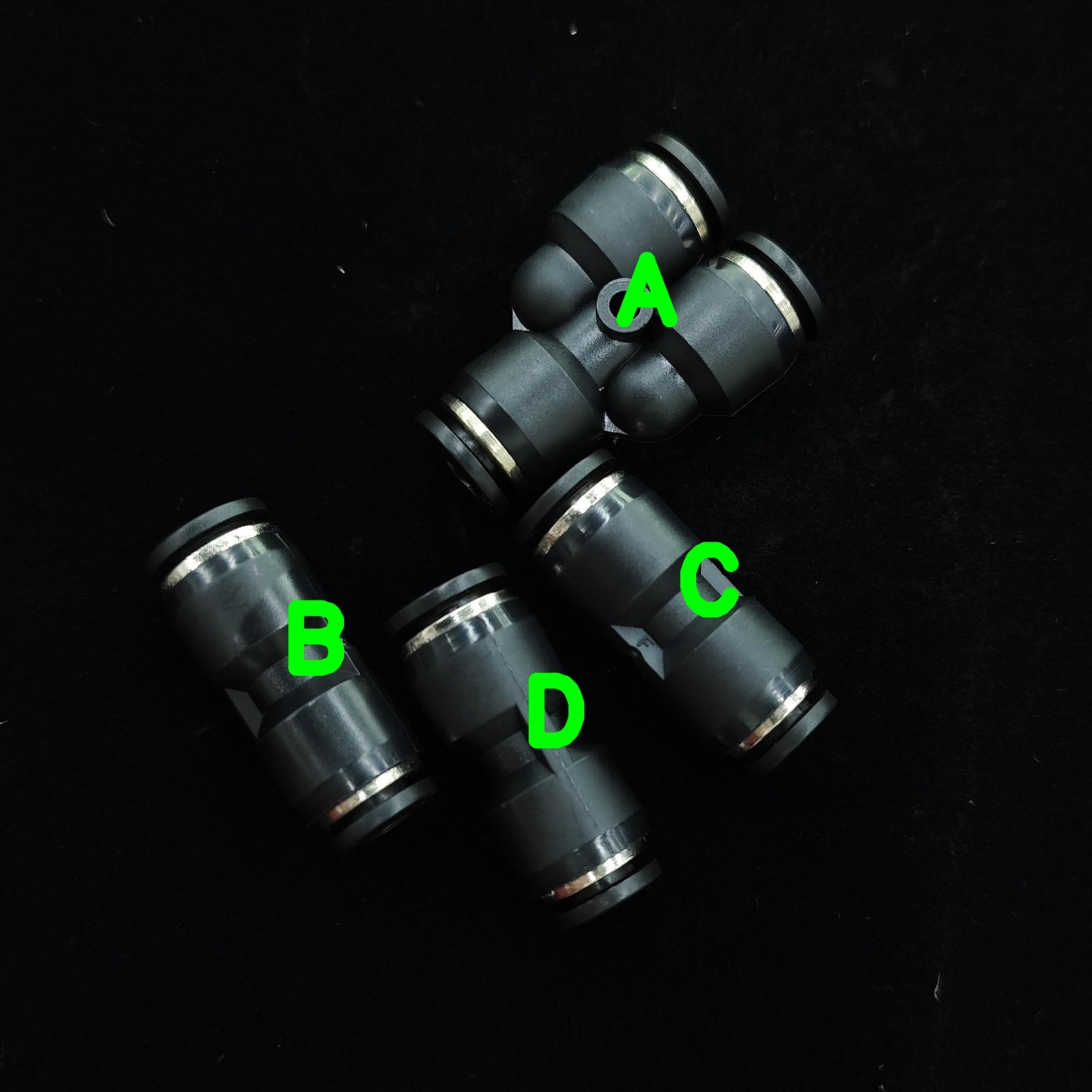}
    }
    \parbox[t][1.5cm][t]{\linewidth}{\small Three 2-way pneumatic tube connectors (8mm to 6mm), mixed with Y-type 3-way}
\end{minipage}\hfill
\begin{minipage}{0.23\textwidth}
    \centering
    \parbox[c][3.0cm][c]{\linewidth}{
        \centering\includegraphics[width=\linewidth,height=3.0cm,keepaspectratio]{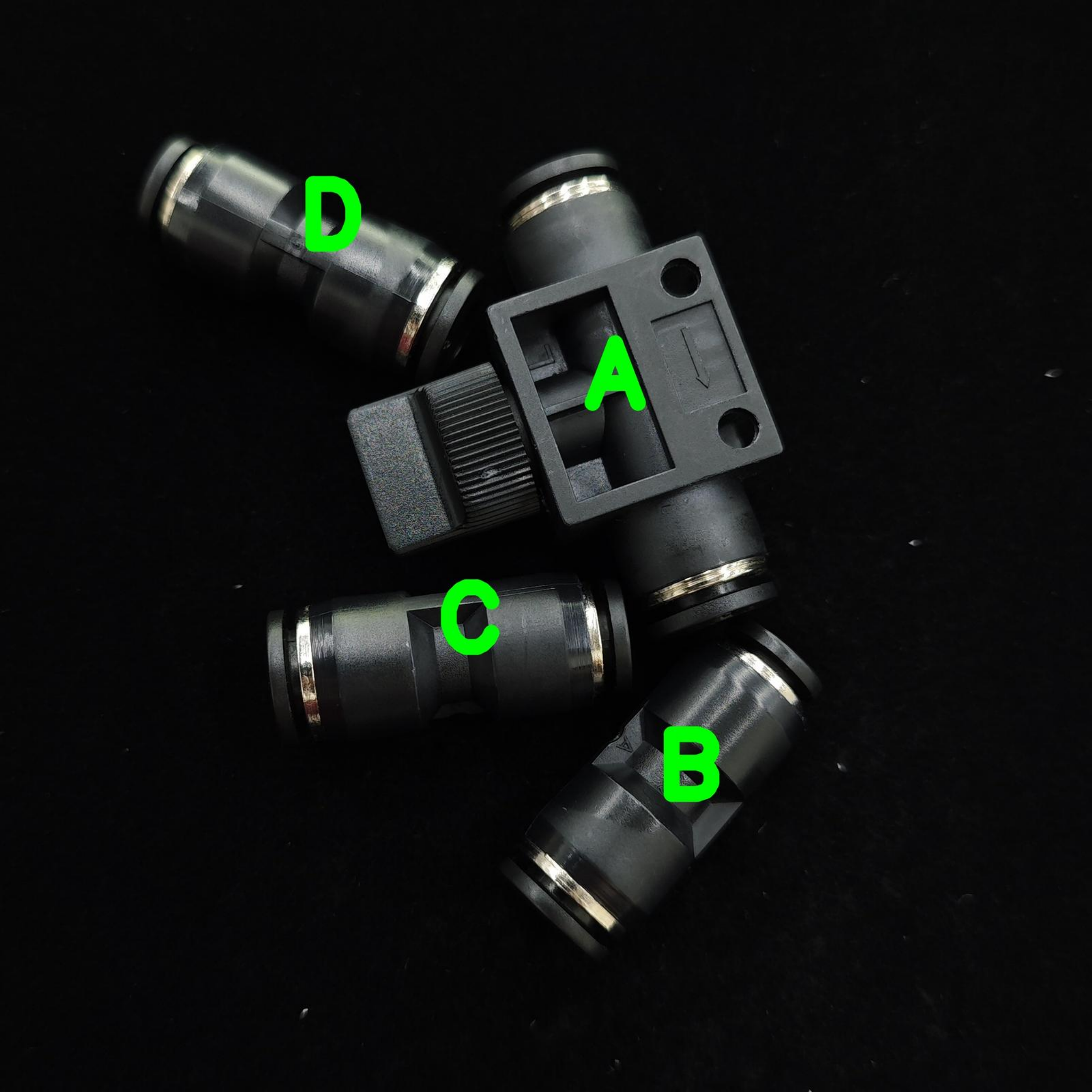}
    }
    \parbox[t][1.5cm][t]{\linewidth}{\small Three 2-way pneumatic tube connectors (8mm to 6mm), mixed with throttle valve}
\end{minipage}\hfill
\begin{minipage}{0.23\textwidth}
    \centering
    \parbox[c][3.0cm][c]{\linewidth}{
        \centering\includegraphics[width=\linewidth,height=3.0cm,keepaspectratio]{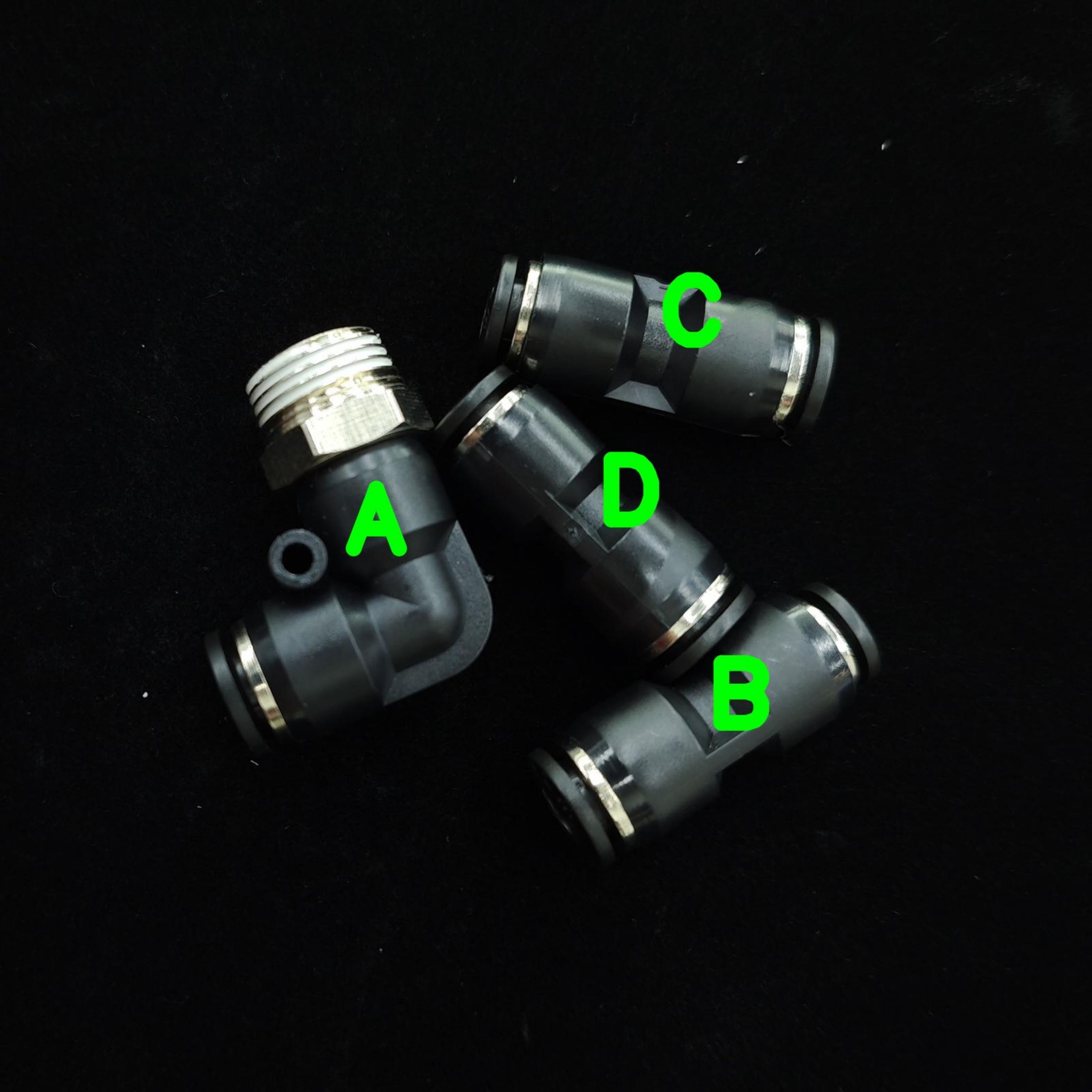}
    }
    \parbox[t][1.5cm][t]{\linewidth}{\small Three 2-way pneumatic tube connectors (8 to 6), mixed with external-thread elbow}
\end{minipage}\hfill
\begin{minipage}{0.23\textwidth}
    \centering
    \parbox[c][3.0cm][c]{\linewidth}{
        \centering\includegraphics[width=\linewidth,height=3.0cm,keepaspectratio]{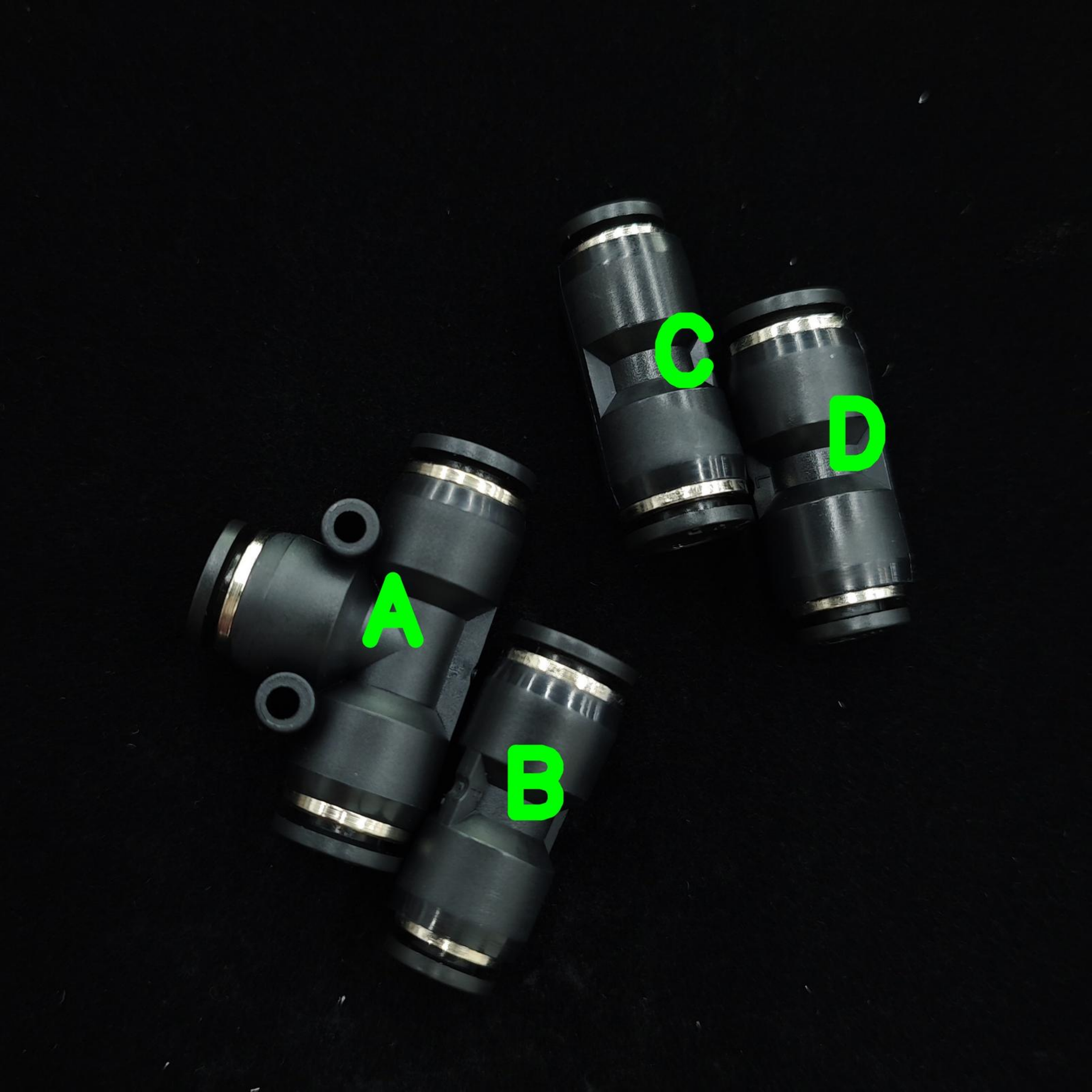}
    }
    \parbox[t][1.5cm][t]{\linewidth}{\small Three 2-way pneumatic tube connectors (8mm to 6mm), mixed with T-type 3-way}
\end{minipage}

\caption{Random grounding examples from \pcsscen}
\label{figab: Manufacturing_task1_scenario_3_saved}
\end{figure}

This subsection shows the Set-of-Mark annotated images used in Bottleneck Analysis~A (Section 4.5 of the main text). Each assembly image is overlaid with distinct letter labels (A, B, C, \etc) on individual components. These annotated images serve as the visual input for both the single-image grounding tasks (C$\to$L and L$\to$C) and the cross-image correspondence tasks (L$\to$L and C$\to$C). We present examples from all four image-based scenarios: \messcen\ (Figure~\ref{figab: Manufacturing_scenario_1_saved}), \pesscen\ (Figure~\ref{figab: Manufacturing_scenario_2_saved}), \cncscen\ (Figure~\ref{figab: Manufacturing_scenario_4_saved}), and \pcsscen\ (Figure~\ref{figab: Manufacturing_task1_scenario_3_saved}).

\newpage

\FloatBarrier

\section{Data examples}\label{app:dataxemplars}

This section provides complete prompt--response examples for all three main benchmark tasks (Sections~\ref{app:exemplars_task1}--\ref{app:exemplars_task3}) and for each of the three bottleneck analyses (Sections~\ref{app:exemplars_BA1}--\ref{app:exemplars_BA3}). Each example includes the system prompt, user prompt, and the model's raw response, showing the full input--output format used in our evaluation pipeline.

\subsection{Examples of \taskone}\label{app:exemplars_task1}

Below are examples of \taskone\ (Workpiece Verification) evaluation data covering both image and three-view modalities. The task requires the model to identify which part in an assembly has the wrong model number or specification.

\textbf{Examples of \pcsscen }

\begin{AIbox}{\taskone \ \pcsscen \ ICD}
  \small
  \begin{itemize}
  \item 
  \textbf{System Prompt:}
  
    General task:
    
    - You will be given one or more reference images showing the correct part type and model/specification,
    followed by a query and a test image that should match that specification.
    
    - Your job is to detect which part in the test image deviates from that specification (or from the majority).
    
    - Instead of returning a coordinate directly, you will be presented with multiple-choice options (A, B, C, ... ), each corresponding to one candidate part location in the test image.
    
    Input format:
    
    - References: one or more images showing correct parts/assemblies that match the stated specification.
    
    - Query: a single line stating the part type and model/specification to check.
    
    - Test Image: the image to evaluate (containing multiple instances of the part).
    
    - Choices: a list of options, each giving a normalized coordinate [x,y] in [0,1] for one part in the test image(top-left is (0,0), bottom-right is (1,1)).
    
    Output format:
    
    - Return exactly one letter in the format "Answer: X" at the end of your response.
    
    - You may include brief reasoning before the final answer.
    
    - Use the reference images only to understand the expected specification; make the final decision based on the Test Image and the provided Choices.

  \end{itemize}
\end{AIbox}

\begin{AIbox}{\taskone \ \pcsscen \ ICD}

  \small
  \begin{itemize}
    
    \item \textbf{User Prompt:}

    Pneumatic Connectors (PC) are specialized components designed for use in pneumatic systems, mainly serving as the interface between air tubes and pneumatic devices. They are widely applied in industrial automation and machinery, including production lines, automated assembly equipment, packaging machines, and various systems powered by compressed air. With the growing demand for efficiency and reliability in modern manufacturing, PCs have become highly popular and indispensable in ensuring stable operations.
    
    The primary function of a PC is to provide secure and convenient connections for pneumatic tubing, ensuring stable and efficient air transmission within the system. For example, the commonly used 2-way pneumatic tube connector creates a reliable passage between two air tubes, allowing smooth airflow and proper system operation. These connectors feature a compact design, easy installation, and quick disassembly, which significantly improves efficiency during maintenance and replacement. As a result, they play a critical role as essential components in any pneumatic system.

    REFERENCE 1 (CORRECT EXAMPLE): These are four correct 2-way pneumatic tube connectors (8mm to 6mm) assembly. All parts meet specifications. Use this as a baseline for comparison.
    
    REFERENCE 2 (CORRECT EXAMPLE): These are four correct 2-way pneumatic tube connectors (8mm to 6mm) assembly. All parts meet specifications. Use this as a baseline for comparison.
    
    REFERENCE 3 (CORRECT EXAMPLE): These are four correct 2-way pneumatic tube connectors (8mm to 6mm) assembly. All parts meet specifications. Use this as a baseline for comparison.

    Based on the provided reference images of correctly assembled 2-way pneumatic tube connectors (8mm to 6mm), check the test image for any mismatches in component specifications. Determine whether the components in the test image are correctly specified and matched, and identify any issues such as incorrect model numbers, wrong sizes, or mismatched specifications. If a problem is found, identify which part has the incorrect specification.
    
    Below are candidate parts with their normalized coordinates in [0,1].
    
    Pick the letter corresponding to the mismatched part.
    
    A. Part at [0.34, 0.43]    
    B. Part at [0.51, 0.77]    
    C. Part at [0.55, 0.54]    
    D. Part at [0.42, 0.24]
    
    Answer: C
    
    Based on the provided reference images of correctly assembled 2-way pneumatic tube connectors (8mm to 6mm), check the test image for any mismatches in component specifications. Determine whether the components in the test image are correctly specified and matched, and identify any issues such as incorrect model numbers, wrong sizes, or mismatched specifications. If a problem is found, identify which part has the incorrect specification.
    
    Below are candidate parts with their normalized coordinates in [0,1].
    
    Pick the letter corresponding to the mismatched part.
    
    A. Part at [0.59, 0.66]
    
    B. Part at [0.41, 0.27]
    
    C. Part at [0.32, 0.71]
    
    D. Part at [0.51, 0.44]

  \end{itemize}
\end{AIbox}

\begin{AIbox}{\taskone \ \pcsscen \ ICD}
  \small
  \begin{itemize}    
      \item \textbf{Model:} Gemini-2.5-Flash
        \item  GT answers:  D
      \item  Pred answer : B
    \item   \textbf{Raw Response:}
    
    The top connector has a noticeably different body profile (more pronounced faceted/stepped center and lacks the same window/slot detailing seen on the other three), indicating it's a different model/spec than the correct 8mm-to-6mm 2-way reducer connectors.

    Answer: B

  \end{itemize}
\end{AIbox}

\iffalse

\begin{figure}[htbp]
\centering

\begin{minipage}{0.23\textwidth}
    \centering
    \parbox[c][3.0cm][c]{\linewidth}{
        \centering\includegraphics[width=\linewidth,height=3.0cm,keepaspectratio]{figs/Manufacturing_task1_scenario_3_saved/figs/Normal_Case_8_to_6__image_167.pdf}
    }
    \parbox[t][1.5cm][t]{\linewidth}{\small Normal Case}
\end{minipage}\hfill
\begin{minipage}{0.23\textwidth}
    \centering
    \parbox[c][3.0cm][c]{\linewidth}{
        \centering\includegraphics[width=\linewidth,height=3.0cm,keepaspectratio]{figs/Manufacturing_task1_scenario_3_saved/figs/Normal_Case_8_to_6__image_209.pdf}
    }
    \parbox[t][1.5cm][t]{\linewidth}{\small Normal Case}
\end{minipage}\hfill
\begin{minipage}{0.23\textwidth}
    \centering
    \parbox[c][3.0cm][c]{\linewidth}{
        \centering\includegraphics[width=\linewidth,height=3.0cm,keepaspectratio]{figs/Manufacturing_task1_scenario_3_saved/figs/Normal_Case_8_to_6__image_227.pdf}
    }
    \parbox[t][1.5cm][t]{\linewidth}{\small Normal Case}
\end{minipage}\hfill
\begin{minipage}{0.23\textwidth}
    \centering
    \parbox[c][3.0cm][c]{\linewidth}{
        \centering\includegraphics[width=\linewidth,height=3.0cm,keepaspectratio]{exemplars/2-way_pneumatic_tube_connectors_(8_to_6),_mixed_with_model_(10_to_6)image_60_gt.jpg}
    }
    \parbox[t][1.5cm][t]{\linewidth}{\small 2-way pneumatic tube connectors  8 to 6 mixed with external-thread elbow}
\end{minipage}

\caption{A test example from Manufacturing image scenario 3}
\label{fig: Manufacturing_task1_scenario_3_test}
\end{figure}
\fi

\begin{AIbox}{\taskone \ \pcsscen \ Zero-Shot}
  \small
  \begin{itemize}
    \item \textbf{User Prompt:}
    
    Pneumatic Connectors (PC) are specialized components designed for use in pneumatic systems, mainly serving as the interface between air tubes and pneumatic devices. They are widely applied in industrial automation and machinery, including production lines, automated assembly equipment, packaging machines, and various systems powered by compressed air. With the growing demand for efficiency and reliability in modern manufacturing, PCs have become highly popular and indispensable in ensuring stable operations.
    
    The primary function of a PC is to provide secure and convenient connections for pneumatic tubing, ensuring stable and efficient air transmission within the system. For example, the commonly used 2-way pneumatic tube connector creates a reliable passage between two air tubes, allowing smooth airflow and proper system operation. These connectors feature a compact design, easy installation, and quick disassembly, which significantly improves efficiency during maintenance and replacement. As a result, they play a critical role as essential components in any pneumatic system.
    
    Based on the provided reference images of correctly assembled 2-way pneumatic tube connectors (8mm to 6mm), check the test image for any mismatches in component specifications. Determine whether the components in the test image are correctly specified and matched, and identify any issues such as incorrect model numbers, wrong sizes, or mismatched specifications. If a problem is found, identify which part has the incorrect specification.
    
    Below are candidate parts with their normalized coordinates in [0,1].
    
    Pick the letter corresponding to the mismatched part.
    
    A. Part at [0.59, 0.66]
    
    B. Part at [0.41, 0.27]
    
    C. Part at [0.32, 0.71]
    
    D. Part at [0.51, 0.44]
    
    Answer:
    
    \end{itemize}

\end{AIbox}

\begin{AIbox}{\taskone \ \pcsscen \ Zero-Shot}
  \small
  \begin{itemize}
    \item \textbf{System Prompt:}
    
    General task:
    
    - You will be given a query naming an industrial part type and its expected model/specification.
    
    - Your job is to detect which part in the provided image deviates from that specification (or from the majority).
    
    - Instead of returning a coordinate directly, you will be presented with a list of multiple-choice options (A, B, C, ...), each describing one candidate part coordinate.
    
    Input format:
    
    - A Query: line stating the part type and model/specification.
    
    - An Image: field containing a single image showing multiple instances of that part.
    
    - A Choices: Section listing multiple options, each with a normalized coordinate for one part in the image (normalized to [0,1]).
    
    Output format:
    
    - Return exactly one letter in the format "Answer: X" at the end of your response.
    
    - You could include your reasoning before the answer.
     
  \end{itemize}
\end{AIbox}

\begin{AIbox}{\taskone \ \pcsscen \ Zero-Shot}
  \small
  \begin{itemize}
        \item \textbf{Model:} Gemini-2.5-Flash
        \item  GT answers:  D
      \item  Pred answer : A
    \item  \textbf{Raw Response 1}: 
    
   Query: 2-way pneumatic tube connectors (8mm to 6mm)

   Image: (Image showing four pneumatic tube connectors)
   
   Choices:
   
   A. Part at [0.59, 0.66]
   
   B. Part at [0.41, 0.27]
   
   C. Part at [0.32, 0.71]
   
   D. Part at [0.51, 0.44]
   
   Analysis:
   
   The query specifies 2-way pneumatic tube connectors with a size specification of 8mm to 6mm. This means the connector should be a reducing union, connecting an 8mm tube on one side and a 6mm tube on the other.

   I need to examine the four connectors in the image to see if they match this specification. Pneumatic push-to-connect fittings typically have different sizes indicated by the diameter of the tube insertion points. Visually, a reducing union (8mm to 6mm) would have one end visibly larger than the other. A straight union (e.g., 8mm to 8mm or 6mm to 6mm) would have both ends the same size.

  \end{itemize}
\end{AIbox}

\begin{AIbox}{\taskone \ \pcsscen \ Zero-Shot}
  \small
  \begin{itemize}
    \item  \textbf{Raw Response 2}:

   1.  **Part B [0.41, 0.27] (Top-left connector):** This connector appears to have two ends of the same size (straight union).
   
   2.  **Part D [0.51, 0.44] (Top-right connector):** This connector also appears to have two ends of the same size (straight union).
   
   3.  **Part A [0.59, 0.66] (Bottom-right connector):** This connector appears to have two ends of the same size (straight union).
   
   4.  **Part C [0.32, 0.71] (Bottom-left connector):** This connector appears to have two ends of the same size (straight union).
    
    All four parts in the image appear to be straight unions (connecting two tubes of the same diameter, e.g., 8mm to 8mm or 6mm to 6mm).
   
   The query specifies a **reducing union (8mm to 6mm)**. Since none of the parts show a clear difference in size between the two ends, it is highly likely that all parts are straight unions (e.g., 8mm-8mm or 6mm-6mm) and thus deviate from the required 8mm to 6mm specification.
    
    However, in a multiple-choice question where the goal is to identify *the* mismatched part, and all parts appear to be the same type (straight union) and thus all mismatch the required specification (reducing union), I must look for any subtle difference or assume the question implies finding *any* part that doesn't meet the specification.
    
    Let's re-examine the parts for any visual cues that might distinguish one from the others, assuming the majority might represent one size (e.g., 8mm-8mm) and one might represent a different size (e.g., 6mm-6mm), or one might actually be the reducing union.
    
    Upon close inspection, all four connectors look identical in form factor (straight union) and overall size relative to each other. They all appear to be the same model, which is a straight union (equal diameter on both sides).
    
    Since the query requires an 8mm to 6mm reducing union, and all parts shown are straight unions (equal diameter on both sides), *all* parts are mismatched specifications. In such a scenario, selecting any one of the options is technically correct as they all fail to meet the required specification.
    
    Let's choose the first option listed, as there is no visual evidence to single out one part as being *more* mismatched than the others.
    
    If the intent of the question is to find a part that deviates from the *majority* shown in the image, and all parts are visually identical straight unions, then there is no deviation within the image itself.
    
    If the intent is to find a part that deviates from the *Query specification* (8mm to 6mm reducing union), then all parts deviate. I will select option A as a representative example of the mismatch against the query specification.
    
    Answer: A
  \end{itemize}
\end{AIbox}

\begin{AIbox}{\taskone \ \pcsscen \ Ref-Cond }
  \small
  \begin{itemize}
    \item \textbf{System Prompt:}
    
    General task:
    
    - You will be given one or more reference images showing the correct part type and model/specification, followed by a query and a test image that should match that specification.
    
    - Your job is to detect which part in the test image deviates from that specification (or from the majority).
    
    - Instead of returning a coordinate directly, you will be presented with multiple-choice options (A, B, C, ...),
    
    each corresponding to one candidate part location in the test image.
    
    Input format:
    
    - References: one or more images showing correct parts/assemblies that match the stated specification.
    
    - Query: a single line stating the part type and model/specification to check.
    
    - Test Image: the image to evaluate (containing multiple instances of the part).
    
    - Choices: a list of options, each giving a normalized coordinate [x,y] in [0,1] for one part in the test image (top-left is (0,0), bottom-right is (1,1)).
    
    Output format:
    
    - Return exactly one letter in the format "Answer: X at the end of your response.
    
    - You may include brief reasoning before the final answer.
    
    - Use the reference images only to understand the expected specification; make the final decision based on the Test Image and the provided Choices.
    
  \end{itemize}
\end{AIbox}

\begin{AIbox}{\taskone \ \pcsscen \ Ref-Cond }
  \small
  \begin{itemize}
    \item \textbf{User Prompt:}
    
    Pneumatic Connectors (PC) are specialized components designed for use in pneumatic systems, mainly serving as the interface between air tubes and pneumatic devices. They are widely applied in industrial automation and machinery, including production lines, automated assembly equipment, packaging machines, and various systems powered by compressed air. With the growing demand for efficiency and reliability in modern manufacturing, PCs have become highly popular and indispensable in ensuring stable operations.
    
    The primary function of a PC is to provide secure and convenient connections for pneumatic tubing, ensuring stable and efficient air transmission within the system. For example, the commonly used 2-way pneumatic tube connector creates a reliable passage between two air tubes, allowing smooth airflow and proper system operation. These connectors feature a compact design, easy installation, and quick disassembly, which significantly improves efficiency during maintenance and replacement. As a result, they play a critical role as essential components in any pneumatic system.
    
    REFERENCE 1 (CORRECT EXAMPLE): There are four correct 2-way pneumatic tube connectors (8mm to 6mm)s. All parts meet specifications. Use this as baseline for comparison.
    
    REFERENCE 2 (CORRECT EXAMPLE): There are four correct 2-way pneumatic tube connectors (8mm to 6mm)s. All parts meet specifications. Use this as baseline for comparison.
    
    REFERENCE 3 (CORRECT EXAMPLE): There are four correct 2-way pneumatic tube connectors (8mm to 6mm)s. All parts meet specifications. Use this as baseline for comparison.
    
    Based on the provided reference images of correctly assembled 2-way pneumatic tube connectors (8mm to 6mm), check the test image for any mismatches in component specifications. Determine whether the components in the test image are correctly specified and matched, and identify any issues such as incorrect model numbers, wrong sizes, or mismatched specifications. If a problem is found, identify which part has the incorrect specification.
    
    Below are candidate parts with their normalized coordinates in [0,1].
    
    Pick the letter corresponding to the mismatched part.
    
    A. Part at [0.59, 0.66]
    
    B. Part at [0.41, 0.27]
    
    C. Part at [0.32, 0.71]
    
    D. Part at [0.51, 0.44]
    
  \end{itemize}
\end{AIbox}

\begin{AIbox}{\taskone \ \pcsscen \ Ref-Cond }
  \small
  \begin{itemize}
       \item  GT answers:  D
      \item  Pred answer : C
    \item \textbf{Raw Response:}
    
    Three connectors share the same overall body profile and end/collet sizing consistent with the reference 8mm-to-6mm reducing unions. The lower-left connector has a noticeably different body geometry (more uniform cylindrical Section and different mid-body detailing), indicating a different connector model/spec (likely not the same 8mm to 6mm reducer).
    
    Answer: C
    
  \end{itemize}
\end{AIbox}

\textbf{Examples of \chsscen}

\begin{AIbox}{\taskone \ \chsscen \ Zero-Shot}
  \small
  \begin{itemize}
    \item  \textbf{System Prompt:}
       
       You are an expert in industrial component quality inspection.
       
       Metal bolts are fundamental fastening components widely used in mechanical structures, construction projects, transportation equipment, and a variety of industrial manufacturing applications. As essential connecting elements, bolts work together with nuts and washers to provide stability and security between different parts. In modern production, bolts are valued for their strength, durability, and reusability, making them indispensable in ensuring the integrity of complex systems.
       
       The primary function of a bolt is to create a strong and reliable connection that maintains the strength and stability of the overall structure. For instance, in steel assembly, bolts firmly secure separate plates, preventing loosening or displacement and ensuring long-term load-bearing performance. Their standardized design allows easy installation, quick removal, and efficient maintenance, which greatly improves productivity and reduces downtime. As a result, bolts play a critical role as core components in any application that demands reliable fastening.
       
       Based on the provided 3V rendered images (showing front, side, and top perspectives) of metal bolts, the task is to compare them with the standard reference images to verify completeness and consistency. The reference images show the expected number of bolts of the same type, all meeting specifications. In the test image, each part is labeled with a letter (A, B, C, D, etc.) and rendered in a distinct color for identification. Your task is to examine the test image and identify any issues such as mismatched components (wrong model number, incorrect dimensions, or different workpiece type). If a problem is identified, return only the letter corresponding to the faulty part (e.g., "B" or "D")."

  \end{itemize}
\end{AIbox}

\begin{AIbox}{\taskone \ \chsscen \  Zero-Shot}
  \small
  \begin{itemize}
    \item  \textbf{User Prompt:}
    Examine the provided three-view rendered image (front, side, and top views merged). Each part is labeled with a letter (A, B, C, D, etc.) and rendered in a distinct color. Identify which part, if any, has the wrong model number or specification compared to the others. Return only the letter corresponding to the mismatched part.
    
    Answer:

  \end{itemize}
\end{AIbox}

\begin{AIbox}{\taskone \ \chsscen \  Zero-Shot}
  \small
  \begin{itemize}
  
  \item \textbf{Model:} Qwen3-VL-235B
  \item \textbf{Gt answers}: D
  \item \textbf{Pred answer}: D
  \item  \textbf{Raw Response:}
  
  So, let's analyze the three views. First, check each part's consistency across views.

Looking at part D: In the front view, D is a light blue bolt overlapping A. In the side view, D is a light blue bolt but the head shape seems different. Wait, in the top view, D is a light blue shape that looks like a different type—maybe the head is not matching. Let's check the other parts. A, B, C have consistent head shapes (like hexagonal or standard bolt heads), but D in the top view has a more rounded or different head. Wait, in the front view, D is a bolt, but in the top view, D's shape is different. Let's check the side view: D is a light blue bolt, but maybe the length or head type is wrong. Wait, the key is that all bolts should be the same type. Let's see: A, B, C have the same head style (the top view for A, B, C shows the head as a certain shape). D in the top view is a different shape—maybe it's a different model. Let's confirm:

- A: In top view, it's a blue shape with a hole (hex head), side view shows the bolt, front view same.

- B: Red, top view is a red shape with a rounded head (maybe hex), side view, front view.

- C: Pink, top view is a pink shape, side and front.

- D: Light blue. In top view, D's shape is different—it looks like a different type of fastener, maybe a screw with a different head. Wait, the other bolts (A, B, C) have heads that in top view are more like hexagonal (A's top view is a hex with a hole), B's top view is a hex, C's top view is a hex. D's top view is a more oval or different shape, which doesn't match. So D is the mismatched part.

  \end{itemize}
\end{AIbox}

\subsection{Examples of \tasktwo}\label{app:exemplars_task2}

Below are examples of \tasktwo\ (Structural Surface Inspection) evaluation data. The task requires the model to classify a workpiece's surface condition into one of five categories: \textit{crack}, \textit{cut}, \textit{deformation}, \textit{dent}, or \textit{good}. We first list the workpiece type descriptions provided to the model, followed by a complete prompt--response example.

\noindent
\begin{AIbox}{Task 2 The Introduction of workpieces.}
  \small
  \begin{itemize}
    \item  \textbf{Corner Bracket:} A corner bracket is an L-shaped metal fitting used to reinforce right-angle joints. Its function is to securely connect two perpendicular components, ensuring structural stability. It is widely used in furniture assembly, building frameworks, shelving, and machinery support.
     \item  \textbf{Countersunk Screw:} A countersunk screw has a conical head that sits flush with the material surface. Its function is to provide a smooth, flat appearance while maintaining strong fastening. It is commonly applied in mechanical assemblies, electronics, and furniture manufacturing.
     \item  \textbf{Cup Head Screw:} A cup head screw features a rounded, smooth head and often includes a square neck to prevent rotation. Its function is to hold wooden or metal parts together securely without turning in the joint. It is widely used in woodworking, construction, and heavy machinery connections.
     \item  \textbf{Eye Bolt:} An eye bolt has a looped head designed for attaching ropes, cables, or hooks. Its function is to serve as a lifting or anchoring point in load-bearing operations. It is commonly found in lifting equipment, construction, marine, and transportation industries.
     \item  \textbf{Flat Washer:} A flat washer is a thin, flat circular disk with a central hole. Its function is to distribute the load of a bolt or nut and protect the connected surface from damage. It is used in mechanical assemblies, construction, and electronic devices.
     \item  \textbf{Hex Nut:} A hex nut is a six-sided fastener commonly paired with bolts. Its function is to secure and clamp parts together through threaded engagement. It is widely used in machinery, automotive manufacturing, construction, and furniture.
     \item  \textbf{Rivet Nut:} A rivet nut is a threaded insert designed for installation in thin materials. Its function is to provide a strong internal thread where tapping is impossible. It is often used in automotive, aerospace, sheet metal, and electronic enclosures.
  \end{itemize}
\end{AIbox}

\begin{AIbox}{Task 2 The Introduction of workpieces.}
  \small
  \begin{itemize}
     
     \item  \textbf{Self-tapping Screw:} A self-tapping screw can cut its own threads into materials without pre-drilling. Its function is to enable fast and easy assembly without requiring separate threading. It is commonly used in metal sheets, plastics, and wood structures.
     \item  \textbf{Spring Washer:} A spring washer is a split ring that provides elastic tension under compression. Its function is to prevent loosening of nuts and maintain constant preload. It is widely applied in machinery, automotive, and electrical equipment.
     \item  \textbf{T Bolt Half-thread Screw:} A T-bolt half-thread screw has a T-shaped head that fits into profile slots for positioning. Its function is to provide adjustable and secure fastening in modular assemblies. It is typically used in machine frames, automation systems, and aluminum profile structures.
     \item  \textbf{T Nut:} A T-nut is designed to fit into T-slots, providing a threaded anchor point. Its function is to enable adjustable fastening within T-slot profiles. It is commonly used in machine tables, aluminum extrusion frames, and assembly lines.
     \item  \textbf{T Screw:} A T-screw has a T-shaped head allowing it to slide into slots for secure fixation. Its function is to work with T-nuts for quick assembly and position adjustment. It is often used in industrial equipment, fixtures, and aluminum frame systems.
     \item  \textbf{Wing Nut:} A wing nut has two projecting “wings” that allow manual tightening without tools. Its function is to enable quick installation and removal by hand. It is frequently used in devices or setups that require frequent adjustments, such as lighting and lab equipment.
     \item  \textbf{Wing Screw:} A wing screw has a head shaped like wings for easy hand operation. Its function is to allow fast manual fastening or adjustment without tools. It is commonly applied in instruments, supports, and lightweight mechanical assemblies.
  \end{itemize}
\end{AIbox}

\textbf{Examples of \tasktwo \ Cup Head Screw }

\begin{AIbox}{\tasktwo \ Cup Head Screw}
  \small
  \begin{itemize}
  \item 
  \textbf{System Prompt:}
  
You are an expert in manufacturing defect analysis based on 3V rendered images of parts.

Cup Head Screw: A cup head screw features a rounded, smooth head and often includes a square neck to prevent rotation. Its function is to hold wooden or metal parts together securely without turning in the joint. It is widely used in woodworking, construction, and heavy machinery connections.

Given a 3V image of a manufactured part, your task is to answer two questions:

**Part 1:** Is the part normal?

Answer: Yes or No

**Part 2:** If the answer to Part 1 is No, identify the defect type by selecting one letter:

A) Crack - a material fracture or micro-crack caused by stress concentration or brittleness

B) Cut - surface scratch or gouge caused by improper machining or external tool impact

C) Deformation - geometric distortion or warping due to excessive load, heat treatment, or clamping stress

D) Dent - local surface depression caused by impact or overpressure during handling or clamping

**Important:** There is at most one type of abnormality present in each part.

Answer format:

Part 1: [Yes/No]

Part 2: [A/B/C/D] (only if Part 1 is No)

If uncertain, choose the option that best matches the dominant feature.

  \end{itemize}
\end{AIbox}

\begin{AIbox}{\tasktwo \ Cup Head Screw}
  \small
  \begin{itemize}
  \item \textbf{User Prompt:}
Here are 3 normal reference images with no defects. Use these to understand what normal (good) parts look like when identifying abnormalities:

NORMAL REFERENCE 1 (No Defect)

NORMAL REFERENCE 2 (No Defect)

NORMAL REFERENCE 3 (No Defect)

Analyze the given part:

Part 1: Is it normal? (Yes/No)

Part 2: If No, select the defect type:

A) Crack

B) Cut

C) Deformation

D) Dent

\item \textbf{Setting:} Ref-Cond 
\item  
  \textbf{Raw response:}
  
Part 1: No

Part 2: B

  \end{itemize}
\end{AIbox}

\subsection{Examples of \taskthree}\label{app:exemplars_task3}

Below are examples of \taskthree\ evaluation data covering both image and point cloud modalities. The task requires the model to detect extra, unnecessary, or incorrectly specified parts in an assembly. We show examples from the \messcen\ and \cncscen\ image scenarios as well as 3D point cloud input.

\textbf{Examples of \messcen }

\begin{AIbox}{\taskthree \ \messcen \ ICD}
  \small
  \begin{itemize}
    \item \textbf{System Prompt:}
    
    General task:
    
    - You will be given one or several reference images of correct assemblies, followed by a query and a test image expected to be the same assembly.
    
    - Your job is to identify which part in the provided test image uses the wrong model number (incorrect size or shape) and thus deviates from the rest.
    
    - Instead of returning a coordinate directly, you will be presented with a list of multiple-choice options (like A, B, C), each describing one candidate part location.
    
    - Choose the letter(s) corresponding to the mismatched part(s).
    
    Input format:
    
    - References: one or more images showing correct assemblies.
    
    - Query: a single line stating the task.
    
    - Test Image: the image to evaluate.
    
    - Choices: a list of options, each with a normalized coordinate for one part in the test image (normalized to [0,1]).
    
    Output format:
    
    - Return exactly the letter of the mismatched part at the end of the answer in a format start with "Answer:" and then letter choice like  Answer: C.
    
    - Also you can return your reasoning based on the provided reference images but be sure to give the final answer only based on the test image.
  \end{itemize}
\end{AIbox}

\begin{AIbox}{\taskthree \ \messcen \ ICD}

  \small
  \begin{itemize}
    \item \textbf{User Prompt 1:}
      
      Based on the provided reference images of correctly assembled Metal expansion bolts, check the test image for any extra, unnecessary, or incorrectly specified parts.

      A standard assembly consists of five key components, each contributing to its performance and reliability: (1)Bolt - the main fastening element that provides tensile strength; (2)Nut - used to tighten and secure the bolt in place; (3)Flat Washer - distributes the load evenly across the surface, reducing the risk of damage to the base material; (4)Spring Washer - prevents loosening caused by vibrations or dynamic loads; (5)Sleeve - the expansion element that deforms outward under tightening, ensuring strong anchorage in the drilled hole. Determine whether the test image contains any components that should not be present or have incorrect specifications. Identify any such problematic parts.

      Below are candidate parts with their normalized coordinates in [0,1].
      
      Pick the letter corresponding to the mismatched part.
      
      A. Part at [0.35, 0.53]
      
      B. Part at [0.52, 0.40]
      
      C. Part at [0.48, 0.24]
      
      D. Part at [0.65, 0.18]
      
      E. Part at [0.68, 0.40]
      
      Answer: A
      
      Based on the provided reference images of correctly assembled Metal expansion bolt, check the test image for any extra, unnecessary, or incorrectly specified parts. A standard assembly consists of five key components, each contributing to its performance and reliability: (1)Bolt - the main fastening element that provides tensile strength; (2)Nut - used to tighten and secure the bolt in place; (3)Flat Washer - distributes the load evenly across the surface, reducing the risk of damage to the base material; (4)Spring Washer - prevents loosening caused by vibrations or dynamic loads; (5)Sleeve - the expansion element that deforms outward under tightening, ensuring strong anchorage in the drilled hole. Determine whether the test image contains any components that should not be present or have incorrect specifications. Identify any such problematic parts.
      
      Below are candidate parts with their normalized coordinates in [0,1].
      
      Pick the letter corresponding to the mismatched part.
      
      A. Part at [0.48, 0.53]
      
      B. Part at [0.53, 0.64]
      
      C. Part at [0.50, 0.42]
      
      D. Part at [0.43, 0.32]
      
      E. Part at [0.66, 0.56]

    \end{itemize}
\end{AIbox}

\begin{AIbox}{\taskthree \ \messcen \ ICD}

  \small
  \begin{itemize}
    \item \textbf{User Prompt 2:}

      Based on the provided reference images of correctly assembled Metal expansion bolts, check the test image for any extra, unnecessary, or incorrectly specified parts.

      A standard assembly consists of five key components, each contributing to its performance and reliability: (1)Bolt - the main fastening element that provides tensile strength; (2)Nut - used to tighten and secure the bolt in place; (3)Flat Washer - distributes the load evenly across the surface, reducing the risk of damage to the base material; (4)Spring Washer - prevents loosening caused by vibrations or dynamic loads; (5)Sleeve - the expansion element that deforms outward under tightening, ensuring strong anchorage in the drilled hole. Determine whether the test image contains any components that should not be present or have incorrect specifications. Identify any such problematic parts.

      Below are candidate parts with their normalized coordinates in [0,1].
      
      Pick the letter corresponding to the mismatched part.
      
      A. Part at [0.35, 0.53]
      
      B. Part at [0.52, 0.40]
      
      C. Part at [0.48, 0.24]
      
      D. Part at [0.65, 0.18]
      
      E. Part at [0.68, 0.40]
      
      Answer: A
      
      Based on the provided reference images of correctly assembled Metal expansion bolt, check the test image for any extra, unnecessary, or incorrectly specified parts. A standard assembly consists of five key components, each contributing to its performance and reliability: (1)Bolt - the main fastening element that provides tensile strength; (2)Nut - used to tighten and secure the bolt in place; (3)Flat Washer - distributes the load evenly across the surface, reducing the risk of damage to the base material; (4)Spring Washer - prevents loosening caused by vibrations or dynamic loads; (5)Sleeve - the expansion element that deforms outward under tightening, ensuring strong anchorage in the drilled hole. Determine whether the test image contains any components that should not be present or have incorrect specifications. Identify any such problematic parts.
      
      Below are candidate parts with their normalized coordinates in [0,1].
      
      Pick the letter corresponding to the mismatched part.
      
      A. Part at [0.48, 0.53]
      
      B. Part at [0.53, 0.64]
      
      C. Part at [0.50, 0.42]
      
      D. Part at [0.43, 0.32]
      
      E. Part at [0.66, 0.56]

    \end{itemize}
\end{AIbox}

\begin{AIbox}{\taskthree \ \messcen \ ICD}
  \small
  \begin{itemize}
  \item \textbf{Model:} Gemini-2.5-Flash
      \item \textbf{Gt answers}: C
      \item \textbf{Pred answer}: D
    \item \textbf{Raw Response:} 
    The standard Metal expansion bolt assembly, as shown in the reference images, consists of five components: Bolt, Nut, Flat Washer, Spring Washer, and Sleeve.
    
    The test image contains the following parts:
    
    1.  **Bolt** (Hex socket cap screw type, different from the references but still a bolt) at approximately [0.43, 0.32] (D).
    
    2.  **Nut** at approximately [0.43, 0.32] (D) - Wait, D is the bolt. The nut is at [0.48, 0.24] in the previous example, or here at [0.48, 0.53] (A).
    
    3.  **Sleeve** at approximately [0.50, 0.42] (C).
    
    4.  **Flat Washer** (large) at approximately [0.53, 0.64] (B).
    
    5.  **Small Washer** (likely a Spring Washer or a second Flat Washer, but much smaller than the others) at approximately [0.66, 0.56] (E).
    
    Let's re-examine the components based on the standard assembly and the reference images:
    
    *   **Reference Images Components:** Bolt (with a conical end), Nut, Sleeve, Flat Washer (large), and Spring Washer (small, split ring).
    
    *   **Test Image Components:**
    
    *   A. Nut (Correct size relative to the bolt threads).
    *   B. Flat Washer (Large, correct size).
    
    *   C. Sleeve (Correct size).
    
    *   D. Bolt (Incorrect type/head shape Hex socket cap screw head instead of the conical head shown in the references, which is crucial for the expansion mechanism).
    
    *   E. Small Washer (This appears to be a small flat washer, not a spring washer, and its inner diameter seems too small for the bolt/sleeve, or it might be a small flat washer intended to be the spring washer replacement, but it is visually different from the spring washers in the references).
    
    Comparing the Bolt (D) in the test image to the Bolts in the reference images:
    
    The reference bolts have a conical or rounded end that engages with the sleeve to cause expansion. The bolt in the test image (D) is a standard hex socket cap screw, which has a flat end and a cylindrical head. This type of bolt head and end shape is incorrect for a standard expansion bolt mechanism where the bolt head/end is designed to push into the sleeve to expand it.
    
    Therefore, the Bolt (D) is the mismatched part due to its incorrect model/shape (Hex socket cap screw instead of the required expansion bolt type).
    
    Answer: D
  \end{itemize}
\end{AIbox}

\begin{AIbox}{\taskthree \ \messcen \ Zero-Shot}
  \small
  \begin{itemize}
    \item \textbf{User Prompt:}
    
    Metal Expansion Bolt (MEB) is one of the most frequently applied workpieces. Its primary function is to ensure strong anchoring by expanding within the drilled hole of the base material, thereby generating high frictional resistance and mechanical interlock.
    
    A typical Metal Expansion Bolt consists of five key components, each contributing to its performance and reliability: (1)Bolt - the main fastening element that provides tensile strength; (2)Nut - used to tighten and secure the bolt in place; (3)Flat Washer - distributes the load evenly across the surface, reducing the risk of damage to the base material; (4)Spring Washer - prevents loosening caused by vibrations or dynamic loads; (5)Sleeve - the expansion element that deforms outward under tightening, ensuring strong anchorage in the drilled hole. Together, these components form a complete fastening system that offers durability, reliability, and ease of installation in demanding construction and industrial environments.
    
    Based on the provided reference images of correctly assembled Metal expansion bolt, check the test image for any extra, unnecessary, or incorrectly specified parts. A standard assembly consists of five key components, each contributing to its performance and reliability: (1)Bolt - the main fastening element that provides tensile strength; (2)Nut - used to tighten and secure the bolt in place; (3)Flat Washer - distributes the load evenly across the surface, reducing the risk of damage to the base material; (4)Spring Washer - prevents loosening caused by vibrations or dynamic loads; (5)Sleeve - the expansion element that deforms outward under tightening, ensuring strong anchorage in the drilled hole. Determine whether the test image contains any components that should not be present or have incorrect specifications. Identify any such problematic parts.
    
    Below are candidate parts with their normalized coordinates in [0,1].
    
    Pick the letter corresponding to the mismatched part.
    
    A. Part at [0.48, 0.53]
    
    B. Part at [0.53, 0.64]
    
    C. Part at [0.50, 0.42]
    
    D. Part at [0.43, 0.32]
    
    E. Part at [0.66, 0.56]
    
    Answer:
    
  \end{itemize}
\end{AIbox}

\begin{AIbox}{\taskthree \ \messcen \ Zero-Shot}
  \small
  \begin{itemize}
    \item \textbf{System Prompt:}
          
          General task:
          
          - You will be given a single image of an assembly containing multiple industrial parts that match each other well.
          
          - Your job is to identify which part in the provided test image uses the wrong model number (incorrect size or shape) and thus deviates from the rest.
          
          - Instead of returning a coordinate directly, you will be presented with a list of multiple-choice options (A, B, C, ...), each describing one candidate part location.
          
          - Choose the letter corresponding to the mismatched part.
          
          Input format:
          
          - Query: a single line stating the task.
          
          - Image: the image to evaluate.
          
          - Choices: a list of options, each with a normalized coordinate for one part in the image (normalized to [0,1]).

          Output format:
          
          - Return exactly one letter in the format "Answer: X" at the end of your response.
          
          - You could include your reasoning before the answer.
  \end{itemize}
\end{AIbox}

\begin{AIbox}{\taskthree \ \messcen \ Zero-Shot}
  \small
  \begin{itemize}
    \item \textbf{Model:} Gemini-2.5-Flash
      \item \textbf{Gt answers}: C
      \item \textbf{Pred answer}: E
    \item  \textbf{Raw Response:} 
    
    The standard assembly for a Metal expansion bolt requires five components: Bolt, Nut, Flat Washer, Spring Washer, and Sleeve.
    
    Let's identify the parts present in the image:
    
    1.  **Bolt**: The threaded component with a cylindrical head (Socket Cap Screw style). (Approx. [0.50, 0.42]) - Matches C.
    
    2.  **Nut**: The hexagonal threaded fastener. (Approx. [0.48, 0.53]) - Matches A.
    
    3.  **Sleeve**: The long cylindrical expansion tube. (Approx. [0.53, 0.64]) - Matches B.
    
    4.  **Flat Washer**: The large, solid ring. (Approx. [0.43, 0.32]) - Matches D.
    
    5.  **Small Washer**: A smaller, solid ring. (Approx. [0.66, 0.56]) - Matches E.
    
    The required components are: Bolt, Nut, Flat Washer, Spring Washer, and Sleeve.
    
    The components present are:
    
    *   Bolt (C)
    
    *   Nut (A)
    
    *   Sleeve (B)
    
    *   Flat Washer (D)
    
    *   Small Washer (E)

    Comparing the required list to the present list:
    
    *   The assembly is missing a **Spring Washer**.
    
    *   The assembly contains two flat washers (D and E), but one of them (E) is significantly smaller than the other (D) and appears to be the wrong size or an extra, unnecessary component, as the standard assembly only requires one Flat Washer and one Spring Washer.
    
    However, the task asks to identify the part that uses the wrong model number (incorrect size or shape) or is extra/unnecessary.
    
    If we assume the larger washer (D) is the correct Flat Washer, then the smaller washer (E) is an extra component and is also likely the wrong size for the bolt/sleeve combination. Furthermore, the required Spring Washer is missing.
    
    In the context of identifying a mismatched part based on size/specification or being extra:
    
    *   Part E ([0.66, 0.56]) is a small flat washer. Given that a standard assembly requires one flat washer (D) and one spring washer (missing), this second, smaller flat washer (E) is an extra, unnecessary component and is likely the wrong size for the bolt/nut combination if it were intended to be the primary washer.
    
    Therefore, the small washer at [0.66, 0.56] is the problematic part.
    
    Answer: E
  \end{itemize}
\end{AIbox}

\begin{AIbox}{\taskthree \ \messcen \  Ref-Cond }
  \small
  \begin{itemize}
    \item  \textbf{System prompt:}
    The same with \taskthree \ Metal Expansion Bolt (MEB) Image In-Context Demonstration.
    \item  \textbf{User prompt:} 
    
     Metal Expansion Bolt (MEB) is one of the most frequently applied products. Its primary function is to ensure strong anchoring by expanding within the drilled hole of the base material, thereby generating high frictional resistance and mechanical interlock.

     A typical Metal Expansion Bolt consists of five key components, each contributing to its performance and reliability: (1)Bolt - the main fastening element that provides tensile strength;   (2)Nut - used to tighten and secure the bolt in place;  (3)Flat Washer - distributes the load evenly across the surface, reducing the risk of damage to the base material;  (4)Spring Washer - prevents loosening caused by vibrations or dynamic loads;  (5)Sleeve - the expansion element that deforms outward under tightening, ensuring strong anchorage in the drilled hole. Together, these components form a complete fastening system that offers durability, reliability, and ease of installation in demanding construction and industrial environments
     
     Reference 1: correct Metal expansion bolt assembly.
     
     Reference 2: correct Metal expansion bolt assembly.
     
     Reference 3: correct Metal expansion bolt assembly.
     
     Based on the provided reference images of correctly assembled Metal expansion bolt, check the test image for any extra, unnecessary, or incorrectly specified parts. A standard assembly consists of five key components, each contributing to its performance and reliability: (1)Bolt - the main fastening element that provides tensile strength; (2)Nut - used to tighten and secure the bolt in place; (3)Flat Washer - distributes the load evenly across the surface, reducing the risk of damage to the base material; (4)Spring Washer - prevents loosening caused by vibrations or dynamic loads; (5)Sleeve - the expansion element that deforms outward under tightening, ensuring strong anchorage in the drilled hole. Determine whether the test image contains any components that should not be present or have incorrect specifications. Identify any such problematic parts.
     
     Below are candidate parts with their normalized coordinates in [0,1].
     
     Pick one choice corresponding to the mismatched part.
     
     A. Part at [0.48, 0.53]
     
     B. Part at [0.53, 0.64]
     
     C. Part at [0.50, 0.42]
     
     D. Part at [0.43, 0.32]
     
     E. Part at [0.66, 0.56]

  \end{itemize}
\end{AIbox}

\begin{AIbox}{\taskthree \ \messcen \  Ref-Cond}
  \small
  \begin{itemize}
    \item \textbf{Model:} Gemini-2.5-Flash
      \item \textbf{Gt answers}: C
      \item \textbf{Pred answer}: C
    \item  \textbf{Raw Response:}
    
    In the reference assemblies, the **bolt** is a countersunk-style expansion bolt with a tapered head and the characteristic shank shape for driving the sleeve. In the test image, the bolt at **[0.50, 0.42]** is a different type (looks like a standard socket-head/cap screw), so it's the wrong model/shape for this expansion bolt set.    
    
    Answer: C

  \end{itemize}
\end{AIbox}
%301": { "case_info": { "normal_case": "raw_full_data/Manufacturing_scenario_2
%603": { "case_info": {  "normal_case": "raw_full_data/Manufacturing_scenario_4

\textbf{Examples of \pesscen}

\begin{AIbox}{\taskthree \ \pesscen \ ICD}
  \small
  \begin{itemize}
    \item \textbf{System Prompt:}

    General task:
    
    - You will be given one or several reference images of correct assemblies, followed by a query and a test image expected to be the same assembly.
    
    - Your job is to identify which part in the provided test image uses the wrong model number (incorrect size or shape) and thus deviates from the rest.
    
    - Instead of returning a coordinate directly, you will be presented with a list of multiple-choice options (like A, B, C), each describing one candidate part location.
    
    - Choose the letter(s) corresponding to the mismatched part(s).

    Input format:
    
    - References: one or more images showing correct assemblies.
    
    - Query: a single line stating the task.
    
    - Test Image: the image to evaluate.
    
    - Choices: a list of options, each with a normalized coordinate for one part in the test image (normalized to [0,1]).

    Output format:
    
    - Return exactly the letter of the mismatched part at the end of the answer in a format start with "Answer:" and then letter choice like Answer: C.
    
    - Also you can return your reasoning based on the provided reference images but be sure to give the final answer only based on the test image.
    
  \end{itemize}
\end{AIbox}

\begin{AIbox}{\taskthree \ \pesscen \ ICD}
  \small
  \begin{itemize}
    \item \textbf{User Prompt:}
  Plastic Expansion Bolts (PEB) stand out as one of the most commonly used variants. Their primary function is to provide a simple yet effective way to anchor screws into walls, panels, or other base materials where direct fastening is not feasible. By expanding inside the drilled hole as the screw is driven in, PEBs create a tight grip that prevents loosening, enhances load-bearing capacity, and ensures durability of the connection. 
  
  The standard assembly must contain three plastic expansion screws, each consisting of two parts: (1) plastic anchors, (2) metal screw.

  Reference 1: Correct Plastic expansion bolt assembly.
  
  Reference 2: Correct Plastic expansion bolt assembly.
  
  Reference 3: Correct Plastic expansion bolt assembly.

  Based on the provided reference images of correctly assembled Plastic expansion bolt, check the test image for any extra, unnecessary, or incorrectly specified parts. A standard assembly must contain three plastic expansion screws, each consisting of two parts: (1) plastic anchors, (2) metal screw. Determine whether the test image contains any components that should not be present or have incorrect specifications. Identify any such problematic parts.

  Below are candidate parts with their normalized coordinates in [0,1].
  
  Pick the letter corresponding to the mismatched part.
  
  A. Part at [0.50, 0.53]
  
  B. Part at [0.40, 0.40]
  
  C. Part at [0.60, 0.52]
  
  D. Part at [0.60, 0.43]
  
  E. Part at [0.35, 0.63]
  
  F. Part at [0.50, 0.64]
  
  Answer: F

    Based on the provided reference images of correctly assembled Plastic expansion bolt, check the test image for any extra, unnecessary, or incorrectly specified parts. A standard assembly must contain three plastic expansion screws, each consisting of two parts: (1) plastic anchors, (2) metal screw. Determine whether the test image contains any components that should not be present or have incorrect specifications. Identify any such problematic parts.
    
    Below are candidate parts with their normalized coordinates in [0,1].

    Pick the letter corresponding to the mismatched part.
    
    A. Part at [0.40, 0.52]
    
    B. Part at [0.43, 0.70]
    
    C. Part at [0.50, 0.44]
    
    D. Part at [0.47, 0.39]
    
    E. Part at [0.65, 0.56]
    
    F. Part at [0.42, 0.62]
    
    Answer:
  \end{itemize}
\end{AIbox}

\begin{AIbox}{\taskthree \ \pesscen \ ICD}
  \small
  \begin{itemize}
   \item \textbf{Model:} Gemini-2.5-Flash
      \item \textbf{Gt answers}: F
      \item \textbf{Pred answer}: F
    \item \textbf{Raw Response1:} 
    
    The test image shows three plastic anchors and four metal screws. A standard assembly, as indicated by the references, should consist of three complete sets, meaning three plastic anchors and three metal screws.
    
    The parts in the test image are:  
    1.  Plastic anchor (Type 1, ribbed) at [0.35, 0.63] (E in the previous set, B in the current set)    
    2.  Plastic anchor (Type 1, ribbed) at [0.50, 0.53] (A in the previous set, C in the current set)    
    3.  Plastic anchor (Type 2, stepped/split) at [0.60, 0.52] (C in the previous set, E in the current set)    
    4.  Metal screw at [0.40, 0.40] (B in the previous set, D in the current set)    
    5.  Metal screw at [0.60, 0.43] (D in the previous set, A in the current set)    
    6.  Metal screw at [0.50, 0.64] (F in the previous set, F in the current set)
    7.  Metal screw at [0.42, 0.62] (F in the current set, this is the same screw as F in the previous set, but the coordinates are slightly different; let's re-evaluate the coordinates for the screws)

    Let's use the coordinates provided in the current prompt:

    A. Part at [0.40, 0.52] (Metal screw)    
    B. Part at [0.43, 0.70] (Plastic anchor, ribbed)     
    C. Part at [0.50, 0.44] (Plastic anchor, ribbed)   
    D. Part at [0.47, 0.39] (Metal screw)   
    E. Part at [0.65, 0.56] (Plastic anchor, stepped/split)    
    F. Part at [0.42, 0.62] (Metal screw)
    
    Wait, there are four screws visible in the image. Let's identify all parts and their types:
    
    1. Plastic Anchor (ribbed, bottom left): [0.43, 0.70] (B)    
    2. Plastic Anchor (ribbed, top center): [0.50, 0.44] (C)    
    3. Plastic Anchor (stepped/split, right): [0.65, 0.56] (E)    
    4. Metal Screw (top right): [0.60, 0.43] (Not listed, but close to A)   
    5. Metal Screw (top center): [0.47, 0.39] (D)    
    6. Metal Screw (bottom center): [0.40, 0.52] (A)    
    7. Metal Screw (bottom left): [0.42, 0.62] (F)
    
    There are 3 plastic anchors (B, C, E) and 4 metal screws (A, D, F, and one unlisted). Since the standard assembly requires 3 plastic anchors and 3 metal screws, one of the metal screws is extra/unnecessary.
  
  \end{itemize}
\end{AIbox}

\begin{AIbox}{\taskthree \ \pesscen \ ICD}
  \small
  \begin{itemize}
   \item \textbf{Model:} Gemini-2.5-Flash
      \item \textbf{Gt answers}: F
      \item \textbf{Pred answer}: F
    \item \textbf{Raw Response2:}

    Let's re-examine the coordinates and the image to confirm the parts listed in the choices:
    
    A. Part at [0.40, 0.52]: Metal screw (bottom center)    
    B. Part at [0.43, 0.70]: Plastic anchor (bottom left)    
    C. Part at [0.50, 0.44]: Plastic anchor (top center)
    D. Part at [0.47, 0.39]: Metal screw (top center)    
    E. Part at [0.65, 0.56]: Plastic anchor (right)    
    F. Part at [0.42, 0.62]: Metal screw (bottom left, partially overlapping with B)

    The image contains 3 plastic anchors and 4 metal screws. Since the expected assembly is 3 sets (3 anchors + 3 screws), one of the metal screws is the mismatched part (extra/unnecessary). All options A, D, and F are metal screws. The fourth screw is located around [0.60, 0.43] (top right).
    
    Since the task is to identify *which part* uses the wrong model number or is extra/unnecessary, and there are 4 screws when only 3 are needed, any of the screws could be considered the extra part if they are all identical. Assuming all screws are the same model, we must choose one of the screw options provided in the choices.
    
    In the previous attempt, the answer was F, which corresponds to the metal screw at [0.42, 0.62]. This screw is indeed one of the four screws present, making it an extra component if the assembly is meant to be 3 sets.
    
    Answer: F
    
  \end{itemize}
\end{AIbox}

\begin{AIbox}{\taskthree \ \pesscen \ Zero-Shot}
  \small
  \begin{itemize}
    \item \textbf{System Prompt:}

    General task:
    
    - You will be given a single image of an assembly containing multiple industrial parts that match each other well.
    
    - Your job is to identify which part in the provided test image uses the wrong model number (incorrect size or shape) and thus deviates from the rest.
    
    - Instead of returning a coordinate directly, you will be presented with a list of multiple-choice options (A, B, C, ...), each describing one candidate part location.
    
    - Choose the letter corresponding to the mismatched part.
    
    Input format:
    
    - Query: a single line stating the task.
    
    - Image: the image to evaluate.
    
    - Choices: a list of options, each with a normalized coordinate for one part in the image (normalized to [0,1]).
    
    Output format:
    
    - Return exactly one letter in the format "Answer: X" at the end of your response.
    
    - You could include your reasoning before the answer.
  \end{itemize}
\end{AIbox}

\begin{AIbox}{\taskthree \ \pesscen \ Zero-Shot}
  \small
  \begin{itemize}
    \item \textbf{User Prompt:}
    
    Plastic Expansion Bolts (PEB) stand out as one of the most commonly used variants. Their primary function is to provide a simple yet effective way to anchor screws into walls, panels, or other base materials where direct fastening is not feasible. By expanding inside the drilled hole as the screw is driven in, PEBs create a tight grip that prevents loosening, enhances load-bearing capacity, and ensures durability of the connection. 
    
    The standard assembly must contain three plastic expansion screws, each consisting of two parts: (1) plastic anchors, (2) metal screw.
    
    Based on the provided reference images of correctly assembled Plastic expansion bolt, check the test image for any extra, unnecessary, or incorrectly specified parts. A standard assembly must contain three plastic expansion screws, each consisting of two parts: (1) plastic anchors, (2) metal screw. Determine whether the test image contains any components that should not be present or have incorrect specifications. Identify any such problematic parts.
    
    Below are candidate parts with their normalized coordinates in [0,1].

    Pick the letter corresponding to the mismatched part.
    
    A. Part at [0.40, 0.52]
    
    B. Part at [0.43, 0.70]
    
    C. Part at [0.50, 0.44]
    
    D. Part at [0.47, 0.39]
    
    E. Part at [0.65, 0.56]
    
    F. Part at [0.42, 0.62]
    
    Answer:
  \end{itemize}
\end{AIbox}

\begin{AIbox}{\taskthree \ \pesscen \ Zero-Shot}
  \small
  \begin{itemize}
     \item \textbf{Model:} Gemini-2.5-Flash
      \item \textbf{Gt answers}: F
      \item \textbf{Pred answer}: E
    \item \textbf{Raw Response:}

    The image shows an assembly of plastic expansion bolts, which should consist of three plastic anchors and three metal screws.
    
    **Analysis of Components:**
    
    1.  **Plastic Anchors (Yellow):** There are three plastic anchors present.
    
    *   Anchor 1 (Top-center): Appears to be a standard ribbed anchor.
    
    *   Anchor 2 (Bottom-left): Appears to be a standard ribbed anchor.
    
    *   Anchor 3 (Right-center, [0.65, 0.56]): This anchor has a distinct, square-wave or zig-zag pattern along its body, which is different in shape/design from the other two anchors.
    
    2.  **Metal Screws (Silver):** There are three metal screws present. All three screws appear to be standard wood/self-tapping screws of similar size and shape.
    
    **Identifying the Mismatched Part:**
    
    The task is to identify the part that deviates from the rest or is incorrectly specified. The plastic anchor at the right-center position has a different design (square-wave pattern) compared to the other two plastic anchors (ribbed pattern). This difference in model/shape makes it the deviating component.
    
    The candidate options for the plastic anchors are:
    
    A. Part at [0.40, 0.52] (Metal screw) - Incorrect type
    
    B. Part at [0.43, 0.70] (Metal screw) - Incorrect type
    
    C. Part at [0.50, 0.44] (Plastic anchor, top-center) - Matches Anchor 2
    
    D. Part at [0.47, 0.39] (Metal screw) - Incorrect type
    
    E. Part at [0.65, 0.56] (Plastic anchor, right-center) - **Mismatched shape/model**
    
    F. Part at [0.42, 0.62] (Plastic anchor, bottom-left) - Matches Anchor 1
    
    The plastic anchor at [0.65, 0.56] (Option E) is the one with the different design compared to the other two plastic anchors.
    
    Answer: E
  \end{itemize}
\end{AIbox}

\begin{AIbox}{\taskthree \ \pesscen \ Ref-Cond }
  \small
  \begin{itemize}
    \item \textbf{System prompt:} 
        The same with \taskthree \  \ \pcsscen \ ICD.

        \item \textbf{User prompt:} 
        
         Plastic Expansion Bolts (PEB) stand out as one of the most commonly used variants. Their primary function is to provide a simple yet effective way to anchor screws into walls, panels, or other base materials where direct fastening is not feasible. By expanding inside the drilled hole as the screw is driven in, PEBs create a tight grip that prevents loosening, enhances load-bearing capacity, and ensures durability of the connection.
         
         The standard assembly must contain three plastic expansion screws, each consisting of two parts: 
         
         (1) plastic anchors,
         
         (2) metal screw
         
         Reference 1: Correct plastic expansion bolt assembly.
         
         Reference 2: Correct plastic expansion bolt assembly.

         Reference 3: Correct plastic expansion bolt assembly.

         Based on the provided reference images of correctly assembled Plastic expansion bolt, check the test image for any extra, unnecessary, or incorrectly specified parts. A standard assembly must contain three plastic expansion screws, each consisting of two parts:
         
         (1) plastic anchors,
         
         (2) metal screw.

         Determine whether the test image contains any components that should not be present or have incorrect specifications. Identify any such problematic parts.
         
         Below are candidate parts with their normalized coordinates in [0,1].
         
         Pick one choice corresponding to the mismatched part.
         
         A. Part at [0.40, 0.52]
         
         B. Part at [0.43, 0.70]
         
         C. Part at [0.50, 0.44]
         
         D. Part at [0.47, 0.39]
         
         E. Part at [0.65, 0.56]
         
         F. Part at [0.42, 0.62]

  \end{itemize}
\end{AIbox}

\begin{AIbox}{\taskthree \ \pesscen \ Ref-Cond }
  \small
  \begin{itemize}
       \item \textbf{Model:} Gemini-2.5-Flash
      \item \textbf{Gt answers}: F
      \item \textbf{Pred answer}: F
    \item \textbf{Raw Response:} 
    
    The test image contains three metal screws that look consistent with the reference (same general length and countersunk head style). 
    
    However, among the three plastic anchors, the one at **[0.42, 0.62]** differs in shape/size (it has a noticeably different body profile and proportions compared to the other two anchors that match the reference style), indicating it's the wrong model.
    
    Answer: F

  \end{itemize}
\end{AIbox}

\textbf{Examples of \cncscen }

\begin{AIbox}{\taskthree \ \cncscen \ ICD}
  \small
  \begin{itemize}
    \item \textbf{System Prompt:}
    
    General task:
    
    - You will be given one or several reference images of correct assemblies, followed by a query and a test image expected to be the same assembly.
    
    - Your job is to identify which part in the provided test image uses the wrong model number (incorrect size or shape) and thus deviates from the rest.
    
    - Instead of returning a coordinate directly, you will be presented with a list of multiple-choice options (like A, B, C), each describing one candidate part location.
    
    - Choose the letter(s) corresponding to the mismatched part(s).
    
    Input format:- References: one or more images showing correct assemblies.
    
    - Query: a single line stating the task.
    
    - Test Image: the image to evaluate.
    
    - Choices: a list of options, each with a normalized coordinate for one part in the test image (normalized to [0,1]).
    
    Output format:
    
    - Return exactly the letter of the mismatched part at the end of the answer in a format start with "Answer:" and then letter choice like Answer: C.
    
    - Also you can return your reasoning based on the provided reference images but be sure to give the final answer only based on the test image.
  \end{itemize}
\end{AIbox}

\begin{AIbox}{\taskthree \ \cncscen \ ICD}
  \small
  \begin{itemize}
    \item \textbf{User Prompt:}
    
    In CNC machining, the primary role of fixtures is to ensure that the workpiece is accurately positioned and firmly secured during various cutting, milling, or drilling operations.
    
    A typical CNC fixture is composed of several fundamental elements that work together to deliver reliable clamping performance. These include: (1) the flange nut, (2) the T-bolt half-thread screw, (3) the pressure plate, and (4) the triangular wedge.
    
    Reference 1: Correct Computer Numerical Control (CNC) fixture assembly.
    
    Reference 2: Correct Computer Numerical Control (CNC) fixture assembly.
    
    Reference 3: Correct Computer Numerical Control (CNC) fixture assembly.
    
    Based on the provided reference images of correctly assembled Computer Numerical Control (CNC) fixture, check the test image for any extra, unnecessary, or incorrectly specified parts. Determine whether the test image contains any components that should not be present or have incorrect specifications. Identify any such problematic parts.
    
    Below are candidate parts with their normalized coordinates in [0,1].
    
    Pick the letter corresponding to the mismatched part.
    
    A. Part at [0.24, 0.50]
    
    B. Part at [0.59, 0.60]
    
    C. Part at [0.47, 0.83]
    
    D. Part at [0.33, 0.36]

    Answer: C

     Based on the provided reference images of correctly assembled Computer Numerical Control (CNC) fixture, check the test image for any extra, unnecessary, or incorrectly specified parts. Determine whether the test image contains any components that should not be present or have incorrect specifications. Identify any such problematic parts.
     
     Below are candidate parts with their normalized coordinates in [0,1].
     
     Pick one choice corresponding to the mismatched part.

     A. Part at [0.70, 0.44]
     
     B. Part at [0.69, 0.81]
     
     C. Part at [0.45, 0.40]
     
     D. Part at [0.29, 0.58] 
            
  \end{itemize}
\end{AIbox}

\begin{AIbox}{\taskthree \ \cncscen \ ICD}
  \small
  \begin{itemize}
         \item \textbf{Model:} Gemini-2.5-Flash
      \item \textbf{Gt answers}: D
      \item \textbf{Pred answer}: B
    \item \textbf{Raw Response:} 
    
    The reference images show a CNC fixture assembly typically consisting of a T-slot bolt, a clamping strap (with a slot), a serrated block (or similar work support/grip), and a nut.

    Let's examine the parts in the test image and compare them to the references:
    
    1.  **Part A (Clamping Strap):** Located at [0.70, 0.44]. This is a metal strap with a central slot. This part appears consistent in shape and function with the clamping strap seen in all reference images.
    
    2.  **Part C (T-slot Bolt):** Located at [0.45, 0.40]. This is a bolt with a T-shaped head and threaded shaft. This part appears consistent with the bolt seen in all reference images.
    
    3.  **Part D (Serrated Block/Work Support):** Located at [0.29, 0.58]. This is a rectangular block with serrations on one face and a hole in the center. This part is consistent with the serrated block/work support seen in Reference 1 and Reference 3 (though Reference 2 and 4 show a different, triangular serrated part, the rectangular serrated block is a valid component in this type of kit, as shown in References 1 and 3).
    
    4.  **Part B (Nut):** Located at [0.69, 0.81]. This is a small hexagonal nut.
    
    Now, let's look closely at the nut (Part B) in the test image and compare it to the nuts in the reference images.
    
    *   In Reference 1, 2, 3, and 4, the nut is a standard hexagonal nut, often a flange nut or a standard hex nut, designed to thread onto the bolt.
    
    *   In the Test Image, the part at [0.69, 0.81] (Part B) is a small, low-profile hexagonal piece. It appears to be a **T-nut** or a similar specialized nut, but its size and shape relative to the bolt (Part C) suggest it might be too small or the wrong type of nut for this specific assembly, especially when compared to the larger, more robust nuts shown in the references (e.g., Reference 2 and 4 show a clearly larger nut).
    
    However, upon closer inspection of the test image, the part at [0.69, 0.81] (Part B) is a **T-nut** (or a similar specialized nut) which is typically used to slide into a T-slot on a machine table. This type of nut is not typically used as the clamping nut *on top* of the clamping strap in this assembly; a standard hex nut or flange nut is used there (as seen in the references).
    
    Let's re-examine the parts in the first set of images provided in the prompt (which are the same as the second set, just different crops/views).
    
    In the test image:
    
    A. Part at [0.70, 0.44] - Clamping Strap (Correct)
    
    B. Part at [0.69, 0.81] - Nut (Mismatched type/size compared to references)
    
    C. Part at [0.45, 0.40] - T-slot Bolt (Correct)
    
    D. Part at [0.29, 0.58] - Serrated Block (Correct type, seen in References 1 and 3)
    
    The nut (Part B) is significantly smaller and appears to be a different type (possibly a T-nut or a very small hex nut) than the nuts shown in the reference images, which are larger and more standard hex nuts for clamping. Therefore, Part B is the mismatched component.
    
    Answer: B
  \end{itemize}
\end{AIbox}

\begin{AIbox}{\taskthree \ \cncscen \  Ref-Cond }
  \small
  \begin{itemize}
    \item \textbf{System prompt:} 
    The same with \taskthree \ Computer Numerical Control (CNC) fixture Image  In-Context Demonstration.
    
    \item \textbf{User prompt:}

     In CNC machining, the primary role of fixtures is to ensure that the workpiece is accurately positioned and firmly secured during various cutting, milling, or drilling operations.
     
     A typical CNC fixture is composed of several fundamental elements that work together to deliver reliable clamping performance. These include: 
     
     (1) the flange nut, 
     
     (2) the T-bolt half-thread screw,
     
     (3) the pressure plate, and 
     
     (4) the triangular wedge.

     Reference 1: correct Computer Numerical Control (CNC) fixture assembly.

     Reference 2: correct Computer Numerical Control (CNC) fixture assembly.
     
     Reference 3: correct Computer Numerical Control (CNC) fixture assembly.

     Based on the provided reference images of correctly assembled Computer Numerical Control (CNC) fixture, check the test image for any extra, unnecessary, or incorrectly specified parts. Determine whether the test image contains any components that should not be present or have incorrect specifications. Identify any such problematic parts.
     
     Below are candidate parts with their normalized coordinates in [0,1].
     
     Pick one choice corresponding to the mismatched part.

     A. Part at [0.70, 0.44]
     
     B. Part at [0.69, 0.81]
     
     C. Part at [0.45, 0.40]
     
     D. Part at [0.29, 0.58]

  \end{itemize}
\end{AIbox}

\begin{AIbox}{\taskthree \ \cncscen \  Ref-Cond }
  \small
  \begin{itemize}
    \item \textbf{Raw Response:}
    
     To determine the mismatched part, we compare the test image with the reference images of correct CNC fixture assemblies. The references consistently show four key components: a rectangular slotted piece, a bolt, a circular nut with a hexagonal interior, and a rectangular serrated block. 
     
     - **Option A** ([0.70, 0.44]): This corresponds to the rectangular slotted piece, which matches the shape and size seen in all references.
     
     - **Option C** ([0.45, 0.40]): This is the bolt, which aligns with the standard bolt shape and size in the references.
     
     - **Option D** ([0.29, 0.58]): This is the rectangular serrated block, which matches the serrated edge and overall form in references 1.
     
     - **Option B** ([0.69, 0.81]): This is the small hex nut. In the references, the nut is consistently larger and has a circular shape with a hexagonal interior (not a small hex nut). The size and shape of this part deviate from the correct specification.
     
     Thus, the mismatched part is **B**, as it does not match the expected size or shape of the nut in the reference assemblies.
     
     Answer: B

  \end{itemize}
\end{AIbox}

\textbf{Examples of  \swnscen }

\begin{AIbox}{\taskthree \ \swnscen \ ICD}
  \small
  \begin{itemize}
      
    \item \textbf{System Prompt:}
   
   You are an expert in industrial assembly quality inspection.
   
   Manufacturing plays a crucial role in modern society, supporting everything from everyday consumer products to advanced industrial machinery. None of these achievements would be possible without precise manufacturing processes, where a wide range of workpieces form the backbone of production. These workpieces are fundamental not only to the structural integrity of equipment but also to ensuring safe, stable, and long-term functionality. From household appliances to large-scale mechanical systems, such components are essential in holding structures together and enabling smooth operation across diverse applications.
   
   Among the most widely used workpieces are bolts and nuts. A bolt engages with a nut to create fastening, which provides strong connections and stability between components. However, bolts and nuts alone are often insufficient for reliable assembly. Complementary parts such as washers are also required. Flat washers help distribute pressure evenly and protect contact surfaces, while spring washers are designed to prevent loosening during operation, thereby improving durability and safety. By combining these elements, mechanical assemblies achieve higher precision and resilience.
   
   In practical manufacturing and assembly, bolts, nuts, and washers are commonly paired to form standardized fastening sets. The most typical combination includes: (1) bolt, (2) nut, (3) flat washer, (4) spring washer. This combination ensures that fastening systems remain secure and consistent across a wide range of industrial and everyday applications.
   
   Based on the provided 3V rendered reference images (showing front, side, and top perspectives) of a correctly assembled common manufacturing workpiece combination, check the test image for any mismatches in component configuration. A standard assembly must contain four M18 components: (1) M18 bolt, (2) M18 nut, (3) M18 flat washer, (4) M18 spring washer. In the test image, each part is labeled with a letter (A, B, C, ..., E, etc.) and rendered in a distinct color for identification. Your task is to determine whether the components are complete and correctly matched, and identify any issues such as missing specifications, incorrect components, or extra parts. If a problem is found, return only the letter corresponding to the faulty part (e.g., "B" or "E").
  \end{itemize}
\end{AIbox}

\begin{AIbox}{\taskthree \ \swnscen \ ICD}
  \small
  \begin{itemize}
    \item \textbf{User Prompt:}

    REFERENCE 1 (CORRECT EXAMPLE): This is a correctly assembled product. All parts meet specifications and no extra parts are present. Use this as baseline for comparison.
    
    REFERENCE 2 (CORRECT EXAMPLE): This is a correctly assembled product. All parts meet specifications and no extra parts are present. Use this as baseline for comparison.
    
    REFERENCE 3 (CORRECT EXAMPLE): This is a correctly assembled product. All parts meet specifications and no extra parts are present. Use this as baseline for comparison.

    Examine the provided 3V rendered image. Each part is labeled with a letter (A, B, C, ..., E, etc.) and rendered in a distinct color. Identify which part is extra (unnecessary) or has wrong specifications. Return only the letter.
    
    Answer: C

    Examine the provided 3V rendered image. Each part is labeled with a letter (A, B, C, ..., E, etc.) and rendered in a distinct color. Identify which part is extra (unnecessary) or has wrong specifications. Return only the letter.

    Answer:

    \end{itemize}
\end{AIbox}

\begin{AIbox}{\taskthree \ \swnscen \ ICD}
  \small
  \begin{itemize}
        \item \textbf{Model:} Qwen3-VL-235B
      \item \textbf{Gt answers}: C
      \item \textbf{Pred answer}: C
    \item \textbf{Raw Response:}  So, let's analyze this. The standard assembly needs four M18 components: bolt, nut, flat washer, spring washer. Let's check each part.
    
    First, identify each part by color and shape. In the reference examples, the bolt is the long threaded part (blue in some references), nut is the hexagonal part (red), flat washer is a simple ring (light blue), spring washer is a split ring (pink).
    
    Looking at the current test image:
    
    - A: In Front View, A is on the blue bolt? Wait, no. Wait, in Front View, the blue long part is the bolt (A? Wait, no. Wait, the bolt should be the main threaded part. Let's check the references. In Reference 1, the bolt is blue (A in some views?), but maybe better to check the counts.
    
    Standard has 4 parts: bolt, nut, flat washer, spring washer. So 4 components. Let's count the labeled parts here: A, B, C, D. Wait, but maybe one is extra or wrong.
    
    Wait, in the Front View, there's a blue bolt (A?), red nut (B), light blue flat washer (D), pink spring washer (C). But wait, standard is bolt, nut, flat washer, spring washer. So 4 parts. Let's check if all are present.
    
    Wait, the problem is if there's an extra part or wrong spec. Wait, in the current image, are there more than 4 parts? No, A, B, C, D are four. Wait, but maybe one is the wrong type.
    
    Wait, in the references, the spring washer is pink (C in references), flat washer is light blue (D), nut is red (B), bolt is blue (A). So standard: A=bolt, B=nut, C=spring washer, D=flat washer.
    
    Now check the current image. Front View: blue bolt (A) is the long part, red nut (B) is hexagonal, light blue D is flat washer, pink C is spring washer. Wait, but in the Front View, the pink C is a flat ring, not a spring washer. Spring washers are split (like a C-shape), but here C is a full ring (flat washer), and D is also a ring. Wait, no—wait, spring washers are typically split, so they look like a "C" shape. Flat washers are solid rings.
    
    Looking at the current image:
    
    - C: pink, in Front View is a full ring (flat washer shape), but spring washer should be split. So maybe C is supposed to be spring washer but is flat, or D is wrong.

    Wait, in the reference examples, let's check. For example, Reference 1: C is pink, and in Side View, it's a split washer (spring washer). So C should be spring washer (split), D is flat washer (solid ring).
   
  \end{itemize}
\end{AIbox}

\begin{AIbox}{\taskthree \ \swnscen \ ICD}
  \small
  \begin{itemize}
  
    \item

    In the current test image, let's check the shapes:
    
    - C: pink. In Front View, it's a full ring (solid), which would be a flat washer. But spring washer should be split. So C is supposed to be spring washer but is flat, so it's the wrong type. Alternatively, maybe C is an extra part, but there are four parts. Wait, no—standard is four parts. Wait, maybe the issue is that C is a flat washer but should be spring, and D is... Wait, D is light blue. In Front View, D is a solid ring (flat washer), but maybe there are two flat washers, which is wrong.
    
    Wait, standard assembly is bolt, nut, flat washer, spring washer. So two washers: flat (D) and spring (C). So C should be spring (split), D flat (solid).

    In the current image, let's check the shapes:
    
    - C (pink): In Front View, it's a solid ring (so flat washer), Side View shows it as a split? Wait, Side View: C is pink, looks like a split (C-shape), so maybe it is a spring washer. Wait, Front View: C is pink, solid ring? No, maybe the rendering. Wait, in the Front View, C is a pink ring, but if it's a spring washer, it should have a split. Wait, maybe in the image, C is a solid ring (flat washer), and D is also a solid ring (flat washer), so there are two flat washers, which is wrong. The standard requires one flat and one spring.
    
    So if C and D are both flat washers (solid rings), then one of them is wrong. Let's check the references. In Reference 1, D is light blue (flat washer, solid), C is pink (spring washer, split). So in the test image, if C is a solid ring (flat), then C is the wrong type (should be spring, but is flat), so C is the faulty part.
    
    Alternatively, maybe the bolt is labeled wrong. Wait, A is blue, which in references is the bolt. B is red, which is the nut. So A=bolt, B=nut, D=flat washer, C should be spring washer. If C is not a spring washer (i.e., it's a flat washer), then C is the wrong component.
    
    Yes, so the issue is that C is supposed to be a spring washer but is a flat washer (or vice versa), but in the standard, we need one of each. If both C and D are flat washers, then C is the extra or wrong one. Wait, no—there should be two washers: flat and spring. So if C is a flat washer (same as D), then there's an extra flat washer, so C is the faulty part (should be spring, but is flat).
    
    Therefore, the faulty part is C. 
  \end{itemize}
\end{AIbox}

\begin{AIbox}{\taskthree \ \swnscen \ Zero-Shot}
  \small
  \begin{itemize}
    \item  \textbf{System Prompt:} Remains the same.
    \item \textbf{User Prompt:}

    Examine the provided 3V rendered image (front, side, and top views merged). Each part is labeled with a letter (A, B, C, ..., E, etc.) and rendered in a distinct color. Identify which part, if any, is extra (unnecessary) or has wrong specifications. Return only the letter corresponding to the problematic part.
    
    Answer:
    
  \end{itemize}
\end{AIbox}

\begin{AIbox}{\taskthree \ \swnscen \ Zero-Shot}
  \small
  \begin{itemize}
    
    \item  \textbf{Raw Response 1:}

    The standard assembly requires four M18 components: (1) M18 bolt, (2) M18 nut, (3) M18 flat washer, and (4) M18 spring washer.

    Let's identify the components in the image:  
    
    *   **A (Dark Blue):** This is the M18 bolt (visible in all three views).  
    
    *   **B (Red):** This is the M18 nut (hexagonal shape, visible in all three views).    
    
    *   **C (Pink):** This is a flat, ring-shaped component, which is the M18 flat washer.   
    
    *   **D (Light Blue):** This is a split, ring-shaped component, which is the M18 spring washer.

    The image shows four components: a bolt (A), a nut (B), a flat washer (C), and a spring washer (D). This matches the required set of four components.
    
    However, the assembly configuration is incorrect. In a standard assembly, the components are typically stacked in the order: Bolt Head $\rightarrow$ Workpiece $\rightarrow$ Flat Washer $\rightarrow$ Spring Washer $\rightarrow$ Nut.
    
    In the provided image, the components are not assembled but are shown as separate, floating parts in the three views. The prompt asks to check the test image for any mismatches in component configuration, and to identify any issues such as missing specifications, incorrect components, or **extra parts**.
    
    Upon closer inspection of the labels in the Front View:  
    *   The dark blue component is the bolt (A).    
    *   The red component is the nut (B).    
    *   The pink component is the flat washer (C).    
    *   The light blue component is the spring washer (D).
    
    The labels A and B are placed on the nut (B) in the Front View, suggesting that the nut (B) might be labeled twice or that one of the labels refers to a feature or an internal component. However, looking at the Side View and Top View, A clearly labels the bolt, and B clearly labels the nut.
    
    Re-examining the Front View:  
    *   The large dark blue object is the bolt (A).    
    *   The large red object is the nut (B).    
    *   The pink ring is the flat washer (C).    
    *   The light blue ring is the spring washer (D).
    
    The labels A and B are placed inside the red nut (B) in the Front View. This is highly ambiguous.
    
    Let's assume the labels A, B, C, ... refer to the four distinct physical components:
    
    A = Bolt    
    
    B = Nut    
    
    C = Flat Washer    
    
    D = Spring Washer

  \end{itemize}
\end{AIbox}

\begin{AIbox}{\taskthree \ \swnscen \ Zero-Shot}
  \small
  \begin{itemize}
    
    \item  \textbf{Raw Response 2:}

     The set is complete and contains the correct types of components.
    
    If we interpret the prompt as checking for *extra* parts based on the labels provided in the image, we must look for components that are not part of the standard set. Since the standard set is (Bolt, Nut, Flat Washer, Spring Washer), and the image shows exactly these four distinct components (A, B, C, ...), there are no extra components.
    
    However, if we strictly follow the labels as they appear in the image, especially in the Front View, the label 'A' is placed on the nut (B) in the Front View, while 'A' labels the bolt in the Side and Top Views. This suggests a labeling error or an ambiguity.
    
    Let's re-examine the components based on their visual representation and standard function:
    1.  **Bolt (A):** Present.
    2.  **Nut (B):** Present.
    3.  **Flat Washer (C):** Present.
    4.  **Spring Washer (D):** Present.
    
    The set is complete. There are no extra components.
    
    If the issue is a mismatch in specifications (e.g., size), all components are visually represented as being of the same nominal size (M18, as specified in the prompt).
    
    If the issue is an incorrect component type:
    *   A is a bolt.
    *   B is a nut.
    *   C is a flat washer.
    *   D is a spring washer.
    All types are correct.
    
    The only remaining possibility is that one of the components is redundant or incorrectly labeled/represented, leading to an "extra part" interpretation. Since the standard set requires four parts, and four distinct parts are shown, none are technically "extra."
    
    However, in quality inspection tasks involving visual representations, sometimes a component is included that is not required for the *specific* assembly type, or a component is duplicated.
    
    Let's assume the image is intended to show a standard assembly set, and we must identify a component that is redundant if the assembly only required three parts, or if one of the washers is redundant.

  \end{itemize}
\end{AIbox}

\begin{AIbox}{\taskthree \ \swnscen \ Zero-Shot}
  \small
  \begin{itemize}
    
    \item  \textbf{Raw Response 3:}

    The standard combination is (1) bolt, (2) nut, (3) flat washer, (4) spring washer.
    
    If we look at the components C (Flat Washer) and D (Spring Washer), they are often used together, but sometimes only one is required. If the requirement was only for a flat washer, the spring washer (D) would be extra. If the requirement was only for a spring washer, the flat washer (C) would be extra. Since the prompt explicitly states the standard combination includes *both* flat and spring washers, neither C nor D is extra based on the definition provided.
    
    Given the constraints of the problem (identifying a faulty part, which includes extra parts), and the fact that the set is complete and correct according to the definition, the only potential "fault" is the ambiguity or redundancy in the labeling itself, or a subtle visual error.
    
    Let's assume the visual representation is flawed and one component is redundant. In many industrial applications, the spring washer (D) is used to prevent loosening, and the flat washer (C) is used for load distribution. If the application only required load distribution, the spring washer (D) would be considered extra. If the application only required anti-loosening, the flat washer (C) would be considered extra.
    
    Since the prompt defines the standard set as including all four, we must look for an error in the components themselves.
    
    Component D (Spring Washer) is visually represented as a split ring, which is correct for a spring washer.
    Component C (Flat Washer) is visually represented as a solid ring, which is correct for a flat washer.
    
    If we assume the standard assembly only requires one type of washer, and the flat washer (C) is the primary component for load distribution, then the spring washer (D) is the most likely candidate for being "extra" if the assembly was simplified.
    
    If we strictly adhere to the provided definition: "The most typical combination includes: (1) bolt, (2) nut, (3) flat washer, (4) spring washer," then all four components (A, B, C, ...) are required.
    
    If the image is a test image that must be checked against the standard, and the standard is the four-part set, then the image is correct in terms of component count and type.
    
    However, in typical quality inspection scenarios where a component is flagged as "faulty," it often means it is redundant or incorrect. Given the ambiguity, and the common practice of sometimes omitting one washer type, let's consider the possibility that one washer is redundant.
    
    In many contexts, the spring washer (D) is considered an optional anti-loosening feature, while the flat washer (C) is more fundamental for surface protection and load distribution. If one must be flagged as extra, it is often the spring washer.
    
    Let's assume the intended error is the inclusion of the spring washer (D) as an extra component beyond a basic bolt/nut/flat washer assembly.
    
    D
  \end{itemize}
\end{AIbox}

\begin{AIbox}{\taskthree \ \swnscen \ Ref-Cond}
  \small
  \begin{itemize}
    \item \textbf{System Prompt:}     remains the same 

    \item \textbf{User Prompt:}
    
    REFERENCE 1 (CORRECT EXAMPLE): This is a correctly assembled product. All parts meet specifications and no extra parts are present. Use this as baseline for comparison.
    
    REFERENCE 2 (CORRECT EXAMPLE): This is a correctly assembled product. All parts meet specifications and no extra parts are present. Use this as baseline for comparison.
    
    REFERENCE 3 (CORRECT EXAMPLE): This is a correctly assembled product. All parts meet specifications and no extra parts are present. Use this as baseline for comparison.
    
    Based on the reference images showing correct assemblies, examine the TEST IMAGE above. Each part is labeled with a letter (A, B, C, ..., E, etc.) and rendered in a distinct color. Identify which part, if any, is extra (unnecessary) or has wrong specifications compared to the standard assembly. Return only the letter corresponding to the problematic part.
    
    Answer:
    \end{itemize}
\end{AIbox}

\begin{AIbox}{\taskthree \ \swnscen \ Ref-Cond}
  \small
  \begin{itemize}
    \item \textbf{Model:} Qwen3-VL-235B
    
    \item \textbf{Gt answers}: C
    
    \item \textbf{Pred answer}: A
    
    \item  \textbf{Raw Response 1:}
    
    So, let's try to figure out which part is problematic. First, I need to recall the standard assembly: it should have four M18 components: bolt (B), nut (A), flat washer (D), spring washer (C). Wait, let me check the references.
    
    Looking at the reference images. In each correct example, the parts should be: bolt (the long threaded part), nut (hexagonal), flat washer (circular, flat), spring washer (circular but with a split, like a spring). Let's identify each label.
    
    In the references, let's see:

    - Bolt is the long part with threads. In references, B is often the bolt? Wait, no. Wait, in the first reference image (top left), the red long part is labeled B. So B is the bolt.
    
    - Nut is the hexagonal part. In the first reference, the blue hexagonal part is A. So A is the nut.
    
    - Flat washer (D): in the first reference, D is the light blue circular washer.

    - Spring washer (C): pink circular with a split, so C is spring washer.

    Now, the test image is the fourth set (the last one with front, side, top views). Let's check each part.

    Test image:

    Front View: blue bolt (A?), red nut (B), light blue washer (D), pink spring washer (C). Wait, but let's check the labels.
    
    Wait, in the test image's front view: the blue long threaded part is labeled A. Wait, but in the references, the bolt is usually the red or blue long part. Wait, maybe the color coding is consistent.

    \end{itemize}
\end{AIbox}

\begin{AIbox}{\taskthree \ \swnscen \ Ref-Cond}
  \small
  \begin{itemize}

    \item  \textbf{Raw Response 2:}
   
    Wait, in the correct references, the bolt is the long threaded component. Let's check the references:

    Reference 1 (first correct example):

    - Front view: red bolt (B), blue nut (A), pink spring washer (C), light blue flat washer (D).
    
    Yes, so B is bolt, A is nut, C is spring washer, D is flat washer.
    
    So standard parts: B (bolt), A (nut), C (spring washer), D (flat washer).
    
    Now, test image:

    Check each part. Let's list all parts in test image.
    
    Front View:
    
    - Blue long threaded part: labeled A. But in references, A is the nut (hexagonal), so this A is the bolt? Wait, no. Wait, in references, A is the nut (hexagonal). So if the test image has a blue long threaded part labeled A, that's wrong because A should be the nut (hexagonal), not the bolt.

    Wait, let's check the test image's parts:

    Test image front view:
    
    - A is on the blue long threaded part (bolt). But in references, A is the nut (hexagonal). So A is mislabeled here. Wait, no. Wait, maybe the labels are consistent across all images. The labels A, B, C, D are fixed per part type.

    Wait, the problem says each part is labeled with a letter (A, B, C, D, E, etc.) and rendered in a distinct color. So A is a specific part type, B another, etc.
    
    In the correct references, the four parts are A, B, C, D. Let's confirm:
    
    In reference 1 (first correct example):

    - A is the nut (hexagonal, blue in front view).

    - B is the bolt (long, red).

    - C is spring washer (pink, circular with split).

    - D is flat washer (light blue, circular).
    
    Yes, that's consistent.
    
    Now, test image:
    
    We need to check if all four parts are present (A, B, C, D) and no extras.
    
    Test image:

    Front View:

    - A is on the blue bolt (long threaded part). Wait, in references, A is the nut, so this A is in the bolt's position. That's a problem.
    
    Wait, let's list all parts in test image:

    - A: appears as a blue long threaded part (bolt) in front view, and as a blue hexagonal part in side view? Wait no.

    Wait, test image front view:

    - Leftmost is blue long threaded (A), then red hexagonal (B), then light blue circular (D), then pink circular (C).

    Side view:
    
    - Top: D (light blue, flat washer), then C (pink, spring washer), B (red hexagonal, nut), A (blue, which looks like a bolt head? Wait, A in side view is a blue part that's the bolt head? Wait, no.

    Wait, in the test image's side view:
    
    - The blue part labeled A is the bolt head (since it's the large end of the bolt). So A is the bolt? But in references, A is the nut.
    \end{itemize}
\end{AIbox}

\begin{AIbox}{\taskthree \ \swnscen \ Ref-Cond}
  \small
  \begin{itemize}

    \item  \textbf{Raw Response 3:}

    Wait, this is key. In the correct assembly:
    
    - A should be the nut (hexagonal, fits onto the bolt).
    
    - B should be the bolt (long threaded).
    
    - C: spring washer.
    
    - D: flat washer.
    
    But in the test image, let's check the parts:
    
    In test image, let's see which parts are present.
    
    Test image:
    
    Front View:

    - A: blue, long threaded (so A is the bolt here)
    
    - B: red, hexagonal (so B is the nut)
    
    - C: pink, spring washer
    
    - D: light blue, flat washer
    
    Wait, but in the references, the bolt is B, nut is A. So here, A is the bolt, B is the nut. That's swapped.
    
    But the problem is about wrong specifications or extra parts. Let's check if all four parts are present.
    
    Wait, in test image, do we have all four: A, B, C, D?

    Yes: A, B, C, D are all present. But the issue is that in the correct assembly, the bolt should be a specific type. Wait, no: the problem says the standard assembly must contain four M18 components: (1) M18 bolt, (2) M18 nut, (3) M18 flat washer, (4) M18 spring washer. The labels A, B, C, D are assigned to these parts in the references.
    
    In the references, A is the nut, B is the bolt, C is spring washer, D is flat washer.
    
    In the test image, let's check:

    - Is there a bolt? Yes, but it's labeled A.
    
    - Is there a nut? Yes, labeled B.
    
    - C is spring washer.
    
    - D is flat washer.
    
    But the problem is about "wrong specifications" or "incorrect components". Wait, no—the labels are part of the identification. Each letter corresponds to a part type. So if in the test image, the part labeled A is not the nut (which it should be), then A is wrong.
    
    Wait, but maybe the issue is an extra part or missing part.
    
    Wait, let's count the parts. In correct assembly: 4 parts (A, B, C, D). Test image: are there 4 parts?
    
    Test image front view: A, B, C, D are all there. Side view: A, B, C, D. Top view: A, B, C, D. So no extra parts.

    But wait, in the test image, let's check the configuration. In the correct assembly, the bolt (B) should have the nut (A) at the end, with washers in between. Wait, the standard combination is bolt, then flat washer, spring washer, then nut? Or the order may vary but the parts must be present.
    \end{itemize}
\end{AIbox}

\begin{AIbox}{\taskthree \ \swnscen \ Ref-Cond}
  \small
  \begin{itemize}

    \item  \textbf{Raw Response 4:}
    
    Wait, the key is: the test image may have a part that is not supposed to be there, or a part that's missing, or a wrong type.

    Wait, looking at the test image, let's see the side view. In the side view, there's a part D (flat washer), C (spring washer), B (nut), and A (bolt). But in the correct references, when assembled, the bolt (B) is the main shaft, nut (A) is at the end. In the test image, the bolt is labeled A, which is supposed to be the nut. So A is misused as the bolt.

    Wait, but the problem says "identify any issues such as missing specifications, incorrect components, or extra parts".

    Wait, another angle: in the correct references, each part is one of the four. In the test image, do we have all four? Let's see:
    
    - Bolt: should be present. In test image, A is the bolt (blue long).

    - Nut: B is the nut (red hexagonal).

    - Flat washer: D (light blue).

    - Spring washer: C (pink).
    
    So all four parts are present. But the problem is that the labels are assigned to wrong parts. Wait, no—the labels are fixed. A is always supposed to be the nut. So if in the test image, A is the bolt, then A is the wrong component.

    Wait, let's check the reference images again.

    Reference 1 (first correct example):

    - Front view: A is blue hexagon (nut), B is red bolt, C is pink spring washer, D is light blue flat washer.
    
    Yes, so A = nut, B = bolt, C = spring washer, D = flat washer.

    In test image:
    
    - A is the blue bolt (long threaded), so A is not the nut. Therefore, A is the wrong component—it's a bolt but should be a nut.
    
    Wait, but maybe the bolt is supposed to be B. So B should be the bolt. In test image, B is the red hexagon (nut), so B is the nut. Therefore, B is correct as the nut. A is the bolt, but A should be the nut. So A is a bolt where a nut should be. Therefore, A is the incorrect part.
    
    Alternatively, maybe there's an extra part. Wait, no, all four parts are present.
    
    Wait, let's look at the test image's side view. In the side view, there's a blue part labeled A which is the bolt (since it's the long shaft), and a red part labeled B which is the nut. But in the correct assembly, the bolt should be B, so A should not be the bolt. Therefore, A is the wrong part because it's a bolt but labeled as A (which should be nut).

    Yes, that's the issue. So the problematic part is A.

  \end{itemize}
\end{AIbox}

\subsection{Examples of Bottleneck Analysis A}\label{app:exemplars_BA1}

Below are examples from the cross-image visual grounding analysis (Bottleneck Analysis A in Section 4.5 of the main text). Each example shows a pair of Set-of-Mark annotated images where the model must match a part identified in the first image to its visual counterpart in the second image. We present both coordinate-to-coordinate (C$\to$C) and letter-to-letter (L$\to$L) cases from Mistral-3-8B, illustrating typical failure modes.

\noindent
\begin{AIbox}{Grounding example of  \pcsscen \ 1 }
  \small
  %G: grounding_ablation_cross_mistral_8b\coord_to_coord case_idx: 0 ref_key: Task1/Manufacturing_task1_scenario_3/Not_Normal_Case/Different_Model_number/2-way_pneumatic_tube_connectors_(8_to_6),_mixed_with_model_(10_to_6)/image_21.jpg test_key: Task1/Manufacturing_task1_scenario_3/Not_Normal_Case/Different_Model_number/2-way_pneumatic_tube_connectors_(8_to_6),_mixed_with_model_(10_to_6)/image_25.jpg task: Task1 scenario: Manufacturing_task1_scenario_3 folder: 2-way_pneumatic_tube_connectors_(8_to_6),_mixed_with_model_(10_to_6)
  \begin{itemize}
    \item \textbf{Setting:} Zero-shot; Coordinates to Coordinates (\textbf{C$\to$C})
    \item \textbf{Gt answer}: A 
    \item \textbf{Pred answer:} C
    \item \textbf{Raw response}: In Image 1, the part at coordinates [0.54, 0.59] is labeled **C**. This part is a cylindrical component with a hexagonal section near the top, which is distinct from the other parts.

In Image 2, we need to identify the same type of component (cylindrical with a hexagonal section near the top). The part labeled **C** in Image 2 matches this description.

Thus, the correct answer is:

Answer: C
  \end{itemize}
\end{AIbox}

\noindent
\begin{AIbox}{Grounding example of  \pcsscen \ 2}
  \small
  %\grounding_ablation_cross_mistral_8b\letter_to_letter case_idx: 0 ref_key: Task1/Manufacturing_task1_scenario_3/Not_Normal_Case/Different_Model_number/2-way_pneumatic_tube_connectors_(8_to_6),_mixed_with_model_(10_to_6)/image_21.jpg test_key: Task1/Manufacturing_task1_scenario_3/Not_Normal_Case/Different_Model_number/2-way_pneumatic_tube_connectors_(8_to_6),_mixed_with_model_(10_to_6)/image_25.jpg task: Task1 scenario: Manufacturing_task1_scenario_3 folder: 2-way_pneumatic_tube_connectors_(8_to_6),_mixed_with_model_(10_to_6)
  \begin{itemize}
  \item \textbf{Setting:} Zero-shot; letters to letters (\textbf{L$\to$L})
    \item \textbf{Gt answer}: B 
    \item \textbf{Pred answer:} C
    \item \textbf{Raw response}:  In Image 1, Part D is a cylindrical component with a smooth outer surface and a threaded inner section that appears to be a barrel or sleeve type with a slightly tapered end.

Comparing this to Image 2:
- Part A is cylindrical but has a different texture and shape, resembling a threaded connector with a wider end.
- Part B is a long cylindrical component with a smooth surface and a wider middle section, not matching Part D.
- Part C is a cylindrical component with a threaded outer section and a smooth inner section, similar to Part D in Image 1.
- Part D in Image 2 is a cylindrical component with a smooth surface and a wider middle section, similar to Part B in Image 1.

The correct match for Part D in Image 1 is Part C in Image 2.

Answer: C
  \end{itemize}
\end{AIbox}

\subsection{Examples of Bottleneck Analysis B}\label{app:exemplars_BA2}

Below is an example from the missing-part detection analysis (Bottleneck Analysis B in Section 4.5 of the main text). The model is given an explicit assembly specification listing all required components, and must identify which part is absent from the three-view rendering. This example shows a Qwen3-VL-235B response on a \swnscen\ case.

\noindent
\begin{AIbox}{Missing part of \swnscen}
  \small
  % 0 key: test_image value: /home/x2jian/industryVLM/raw_full_data/three-view/Task3/manufacturing_scenario_1_pointcloud/Normal_Case/point_cloud_56_cluster_merged.png qwen3-vl-235b-a22b-thinking zero_shot
  \begin{itemize} 

    \item \textbf{System Prompt:}   You are an expert manufacturing quality inspector. Your task is to examine three-view rendered images (front, side, top) of an assembly and determine whether all required components are present. Each visible part is labeled with a letter and colored distinctly. You will be given a description of what a complete assembly should contain, and you must identify which part (if any) is missing from the test image. Answer with ONLY the letter of the correct option (e.g., 'Answer: C').
    
    \item \textbf{User Prompt:}

  You are inspecting a three-view rendering of an assembly that should be a complete Metal Expansion Bolt.
  
  Assembly Description:
  A standard fastening assembly consists of four M18 components: (1) M18 Bolt -- the main fastening element; (2) M18 Nut -- engages with the bolt to create a secure connection; (3) M18 Flat Washer -- distributes pressure evenly and protects contact surfaces; (4) M18 Spring Washer -- prevents loosening during operation. A complete assembly must contain all four components.
  
  The test image shows front, side, and top views of the assembly. Each part is labeled with a letter and rendered in a distinct color. Compare the visible parts against the expected components listed above. Count the parts and determine if any required component is missing.
  
  Which of the following parts is missing from this assembly?
  A. No part is missing
  
  B. Nut
  C. Flat Washer
  D. Spring Washer
  E. Bolt
  
  Answer with ONLY the letter of your choice (e.g., 'Answer: C').
     \item \textbf{Setting:} Zero-shot; 
    \item \textbf{Model} Qwen3-VL-235B; 

    \item \textbf{Gt answer}: A
    \item \textbf{Pred answer}: A
  \end{itemize}
\end{AIbox}
\noindent
\begin{AIbox}{Missing part of \swnscen}
  \small
  % 0 key: test_image value: /home/x2jian/industryVLM/raw_full_data/three-view/Task3/manufacturing_scenario_1_pointcloud/Normal_Case/point_cloud_56_cluster_merged.png qwen3-vl-235b-a22b-thinking zero_shot
  \begin{itemize}
  
    \item  \textbf{Raw Response 1:}  Got it, let's analyze this. The assembly should have four components: Bolt, Nut, Flat Washer, Spring Washer. The parts in the image are labeled A, B, C, D. Let's map each label to the components.

    First, let's recall the standard parts. A bolt is the main fastening element (probably the long red one, B). A nut engages with the bolt (maybe the red rounded part in side view? Wait, side view has B overlapping, but maybe B is the bolt. Then washers: flat washer is a flat ring, spring washer is a split ring. Wait, but the problem says four M18 components: Bolt, Nut, Flat Washer, Spring Washer.

    Looking at the labels:

    - A: Blue, in front view is a circle, side view is a ring (so washer), top view is a small blue shape. So A is likely a washer (flat or spring).
    
    - B: Red, long in front and top views, so that's the bolt (main fastening element).

    - C: Pink, in all views, maybe the nut? Wait, nut is a hexagonal part that engages with the bolt. In side view, C is a pink shape next to B.
    
    - D: Light blue, in front, side, top views as a small shape. Maybe the other washer.

    Wait, the assembly needs four parts: Bolt (B?), Nut (maybe C?), Flat Washer (A?), Spring Washer (D?).

    But wait, the problem states four components: Bolt, Nut, Flat Washer, Spring Washer. So all four must be present.

    Let's count the labels: A, B, C, D. Four labels. But wait, are all four parts actually present? Wait, maybe one of the labels is a duplicate or one is missing.

    Wait, no—the labels are A, B, C, D. So four parts. But the question is which is missing. Wait, the options include "No part is missing" (A), but let's check.
    
    Wait, the assembly description says four components: Bolt, Nut, Flat Washer, Spring Washer. So each label should correspond to one of these. Let's map:

    - Bolt: B (red, long, main fastening element)
    
    - Nut: C (pink, maybe the hex nut)
    
    - Flat Washer: A (blue, ring)
    
    - Spring Washer: D (light blue, maybe the spring washer)

    But wait, the problem is to check if all four are present. The labels are A, B, C, D—so four parts. But wait, maybe one of the components is not present. Wait, no—the labels are distinct, so four parts. But the options include "No part is missing" as option A. But wait, the problem says "which of the following parts is missing". Wait, maybe the labels correspond to different parts.

    Wait, let's think again. The four required components are Bolt, Nut, Flat Washer, Spring Washer. Let's see which is missing.

      \end{itemize}
\end{AIbox}

\noindent
\begin{AIbox}{Missing part of \swnscen}
  \small
  % 0 key: test_image value: /home/x2jian/industryVLM/raw_full_data/three-view/Task3/manufacturing_scenario_1_pointcloud/Normal_Case/point_cloud_56_cluster_merged.png qwen3-vl-235b-a22b-thinking zero_shot
  \begin{itemize}
  
    \item  \textbf{Raw Response 2:} 
    If the assembly has all four, then option A. But maybe one of the labels is not a required component. Wait, no—the labels are the parts in the assembly. Wait, the problem says "each visible part is labeled with a letter". So if there are four labels (A,B,C,D), but maybe one of the required components is not present, meaning that one of the labels is not a required part, but no—wait, the required parts are four, so the assembly should have four parts. If there are four parts labeled, but one of the required components is missing, that would mean that one of the labels is a different part, but the problem states that the assembly should consist of those four. Wait, maybe the labels correspond to the four, but one of the required components is not present.

    Wait, let's list the required components:

    1. M18 Bolt – main fastening element (B, since it's the long red part)

    2. M18 Nut – engages with bolt (maybe C, pink)

    3. M18 Flat Washer – distributes pressure (A, blue ring)
    
    4. M18 Spring Washer – prevents loosening (D, light blue)

    So if all four are present, then A (no part missing). But wait, the options are A. No part missing, B. Spring Washer, C. Flat Washer, D. Bolt, E. Nut.

    Wait, but let's check the side view. In side view, there's A (blue ring), B (red, overlapping), C (pink), D (light blue at top). So all four labels are present. So why would a part be missing? Wait, maybe one of the components is not present. Wait, but the labels are A, B, C, D. So four parts. The required is four. So maybe all are present. But wait, the problem says "a complete assembly must contain all four components". So if all four are present, answer A.

    But wait, let's check the assembly description again. The four components are Bolt, Nut, Flat Washer, Spring Washer. Let's see which part is which.

    Bolt is the main fastening element (B), Nut is the one that threads onto the bolt (probably C), Flat Washer (A), Spring Washer (D). So all four are present. But wait, the options include A. No part missing. So why would the answer not be A?
    
    Wait, maybe I made a mistake. Let's count the parts. The labels are A, B, C, D. Four parts. The required is four. So all are present. But maybe one of the components is not present. Wait, perhaps the Spring Washer is missing. Wait, in the side view, D is a light blue shape at the top. Is D the Spring Washer? Let's think: Spring Washers are typically helical or split, but in a rendering, maybe it's a different shape. Flat Washer is a flat ring (A), which matches the blue ring in side view. Bolt is B, Nut is C. Then D would be the Spring Washer. So all four are present. So answer A.
    
    But wait, the problem says "determine whether all required components are present". If all four labels correspond to the four required components, then no part is missing. But let's check the options. Option A is "No part is missing".

      \end{itemize}
\end{AIbox}

\noindent
\begin{AIbox}{Missing part of \swnscen}
  \small
  % 0 key: test_image value: /home/x2jian/industryVLM/raw_full_data/three-view/Task3/manufacturing_scenario_1_pointcloud/Normal_Case/point_cloud_56_cluster_merged.png qwen3-vl-235b-a22b-thinking zero_shot
  \begin{itemize}
  
    \item  \textbf{Raw Response 3:} 
     Wait, but maybe the assembly is supposed to have four parts, but in reality, one of the labels is not a required component. No, the problem states that the assembly should consist of those four, so each required component must be present. Since there are four labeled parts, and four required, they must map one-to-one. Therefore, no part is missing. So answer A.
    
    But wait, let's re-examine. The user might have a trick. Let's see: the assembly is a Metal Expansion Bolt. Wait, the description says "a standard fastening assembly consists of four M18 components: (1) M18 Bolt... (2) M18 Nut... (3) M18 Flat Washer... (4) M18 Spring Washer".

    Wait, but a Metal Expansion Bolt might not require a nut? Wait, no, the problem states that the assembly should have all four. So regardless of real-world, per problem statement, four parts are required.

    Looking at the image: front view has A, B, C, D. Side view has A, B, C, D. Top view has A, B, C, D. So all four parts are present. So answer A.

    But wait, the options list A as "No part is missing". So the answer should be A.

    Wait, but maybe I'm wrong. Let's check again. Let's list each component:

    - Bolt (B): present (red, long)
    
    - Nut (C): present (pink)
    
    - Flat Washer (A): present (blue ring)
    
    - Spring Washer (D): present (light blue)

    All four are present. So no part missing. Answer A
    
  \end{itemize}
\end{AIbox}

\subsection{Examples of Bottleneck Analysis C}\label{app:exemplars_BA3}

Below is an example from the raw point cloud text input analysis (Bottleneck Analysis C in Section 4.5 of the main text). The model receives 3D coordinates serialized as integer-scaled text tables and must perform the standard \taskthree\ task without any visual rendering. This example shows a Qwen3-235B response on a \swnscen\ case.

\begin{AIbox}{Pure point input example of \swnscen}
  \small
  %      "ori_image": "raw_full_data/task3_point/Task3/manufacturing_scenario_1_pointcloud/pointcloudnew_test/Not_Normal_Case/Different_Model_number/Flat_Washer_M14/point_cloud_39.ply",

  \begin{itemize}

    \item \textbf{System Prompt:}
    You are an expert at analyzing 3D point cloud data for industrial quality control.
    
    Your task:
    
    - Identify which part (if any) deviates from the expected specification based on 3D point-cloud data.
    
    - At most one part deviates from the others.
    
    Input format:
    
    - "Description of context:" (optional) - Background information about the parts and their function.
    
    - "Task:" - The specific task instruction.
    
    - "Query:" - Indicates that the 3D point cloud data follows.
    
    - "Data:" - 3D coordinates of each part's point cloud, formatted as:
    
    Part A:
    
    x, y, z
    
    1234, 5678, 9012
    
    1235, 5679, 9011
    
    ...
    
    Part B:
    
    x, y, z
    
    ...
    
    Output format
    
    - Return the Part ID (e.g., "Part A") of the deviating part.
    
    - If all parts match specifications, state that all parts are correct.

    \item \textbf{User Prompt:}
    
    Task: There are 4 parts in total. Identify the only part whose 3D point-cloud data deviates from the specified model's design compared to the others.

           Query: The 3D point cloud data for all parts is provided below.

           Data:

           Part A:
           
           x, y, z
           
           3930, -3952, 42338
           
           4664, -4289, 42390
           
           3381, -3782, 42378

           ...

           Part B:
           
            ...

           5180, -698, 32471
           
           4244, -1049, 32322

           Please answer the query by thinking about the shape, size or other features of the part that represented by the point cloud data.

           You may use any reasoning approach you feel is needed to make an accurate determination.
           
           Output format:
           
           - Return the Part ID of the part in question like: "Part X". Be sure to give the result at the end of your response.",

     \item \textbf{Brief info:} Zero-shot; 
    \item \textbf{Model:}  Qwen3-235B
    \item \textbf{Gt answer}: C 
    \item \textbf{Pred answer:} C
  \end{itemize}
\end{AIbox}

\begin{AIbox}{Pure point input example of \swnscen}
  \small
  %      "ori_image": "raw_full_data/task3_point/Task3/manufacturing_scenario_1_pointcloud/pointcloudnew_test/Not_Normal_Case/Different_Model_number/Flat_Washer_M14/point_cloud_39.ply",

  \begin{itemize}

    \item \textbf{Raw response}: 
    
    To identify the deviating part from the 3D point cloud data, I analyzed the dimensional characteristics, aspect ratios, and spatial distribution of points for each part. 
    
    First, I examined the coordinate ranges for each dimension (x, y, z) across all parts:
    
    Part A:
    
    - x: \textasciitilde2500-5800 (span \textasciitilde3300)
    
    - y: \textasciitilde-4300 to -1400 (span \textasciitilde2900)
    
    - z: \textasciitilde40700-42600 (span \textasciitilde1900)
    
    Part B:
    
    - x: \textasciitilde9400-12100 (span \textasciitilde2700)
    
    - y: \textasciitilde1000-3600 (span \textasciitilde2600)
    
    - z: \textasciitilde37000-38300 (span \textasciitilde1300)
    
    Part C:
    
    - x: \textasciitilde6300-16400 (span \textasciitilde10100)
    
    - y: \textasciitilde-6400-600 (span \textasciitilde7000)
    
    - z: \textasciitilde43700-49900 (span \textasciitilde6200)

    Part D:
    
    - x: \textasciitilde4200-7100 (span \textasciitilde2900)

    - y: \textasciitilde-2900-300 (span \textasciitilde3200)
    
    - z: \textasciitilde31300-33900 (span \textasciitilde2600)

    The most striking difference is in the z dimension span:
    
    - Part C's z span is \textasciitilde6200, which is 3.26 larger than Part A's (1900), 4.77 larger than Part B's (1300), and 2.38 larger than Part D's (2600)
    
    When examining aspect ratios (normalized by the smallest dimension):
    
    - y:z ratio: Part A (1.53:1), Part B (2.00:1), Part C (1.13:1), Part D (1.23:1)
    
    - Part C has a significantly smaller y:z ratio compared to the others, indicating it's much "taller" relative to its width
    
    Part C also shows an unusual pattern in point distribution:

    - Contains numerous points with extreme z values (up to 49900) that are well beyond the z ranges of other parts
    
    - Has points with both highly negative y values (down to -6400) and positive/near-zero values, creating a wider y distribution

    Most critically, Part C's absolute size in all dimensions is substantially larger than the other parts:
    
    - x span is \textasciitilde3.7 larger than the average of other parts
    
    - y span is \textasciitilde2.4 large
    
    - z span is \textasciitilde3.3 larger

    This disproportionate scaling across all dimensions, combined with the abnormal y:z aspect ratio, indicates that Part C has a fundamentally different shape than the other three parts. In industrial quality control, such significant dimensional deviation from the expected model would classify this as a defective part.

    Part C

      \end{itemize}
\end{AIbox}

%\input{Related Works Results.tex}

% \bibliographystyle{plainnat}
% \bibliography{main}
\end{document}